\documentclass[10pt,twocolumn,letterpaper]{article}

\usepackage{iccv}
\usepackage{times}
\usepackage{epsfig}
\usepackage{graphicx}
\usepackage{amsmath}
\usepackage{amssymb}
\usepackage{booktabs}
\usepackage{epigraph}
\usepackage{url}            
\usepackage{amsfonts}       
\usepackage{nicefrac}       
\usepackage{xcolor,colortbl}         
\usepackage{xspace}
\usepackage{multirow}
\usepackage{mathtools}
\usepackage{overpic}
\usepackage{caption}

\usepackage{microtype}

\usepackage[pagebackref=true,breaklinks=true,letterpaper=true,colorlinks,bookmarks=false]{hyperref}

\iccvfinalcopy 

\ificcvfinal\fi

\usepackage[capitalize]{cleveref}
\crefname{section}{Sec.}{Secs.}
\Crefname{section}{Section}{Sections}
\Crefname{table}{Table}{Tables}
\crefname{table}{Tab.}{Tabs.}

\definecolor{red}{rgb}{0.95,0.4,0.4}
\definecolor{purered}{rgb}{1,0,0}
\definecolor{darkblue}{rgb}{0,0,0.8}
\definecolor{darkred}{rgb}{1,0,0}
\definecolor{darkgreen}{rgb}{0,0.5,0}
\definecolor{grey}{rgb}{0.6,0.6,0.6}
\definecolor{col1}{RGB}{232, 161, 148}
\definecolor{col2}{RGB}{148, 187, 232}
\definecolor{lightgrey}{rgb}{0.85,0.85,0.85}
\definecolor{lightlightgrey}{rgb}{0.9,0.9,0.9}
\definecolor{verylightBG}{rgb}{0.9,0.99,0.99}
\definecolor{darkgreen}{rgb}{0.3, 0.75, 0.3}
\definecolor{darkgrey}{rgb}{0.8,0.35,0.35}
\definecolor{darkorange}{rgb}{0.95,0.5,0.2}

\definecolor{maroon}{cmyk}{0,0.87,0.68,0.32}
\definecolor{grey}{rgb}{0.6,0.6,0.6}
\definecolor{lightmaroon}{rgb}{0.98,0.86,0.86}

\newcommand{\PAR}[1]{\vskip4pt \noindent {\bf #1~}}
\newcommand{\PARit}[1]{\vskip4pt \noindent {\it #1~}}

\newcommand*{\lidog}{LiDOG\@\xspace}

\newcommand*{\source}{source\@\xspace}
\newcommand*{\target}{target\@\xspace}

\newcommand*{\marksrc}{\cellcolor{red!20}}
\newcommand*{\marktar}{\cellcolor{blue!10}}

\begin{document}

\title{Walking Your LiDOG: A Journey Through Multiple Domains for LiDAR Semantic Segmentation}



\author{Cristiano Saltori$^1$, Aljoša Ošep$^2$, Elisa Ricci$^{1,3}$, Laura Leal-Taixé$^4$\\
$^1$ University of Trento, $^2$ TU Munich, $^3$ Federazione Bruno Kessler, $^4$ NVIDIA\\
{\tt\small cristiano.saltori@unitn.it}
}


\ificcvfinal\thispagestyle{empty}\fi

\ificcvfinal\thispagestyle{empty}\fi
\newcommand{\ckm}{\checkmark}
\twocolumn[{%
\renewcommand\twocolumn[1][]{#1}%
\maketitle
\begin{center}
\vspace{-0.5cm}
\includegraphics[width=0.99\linewidth]{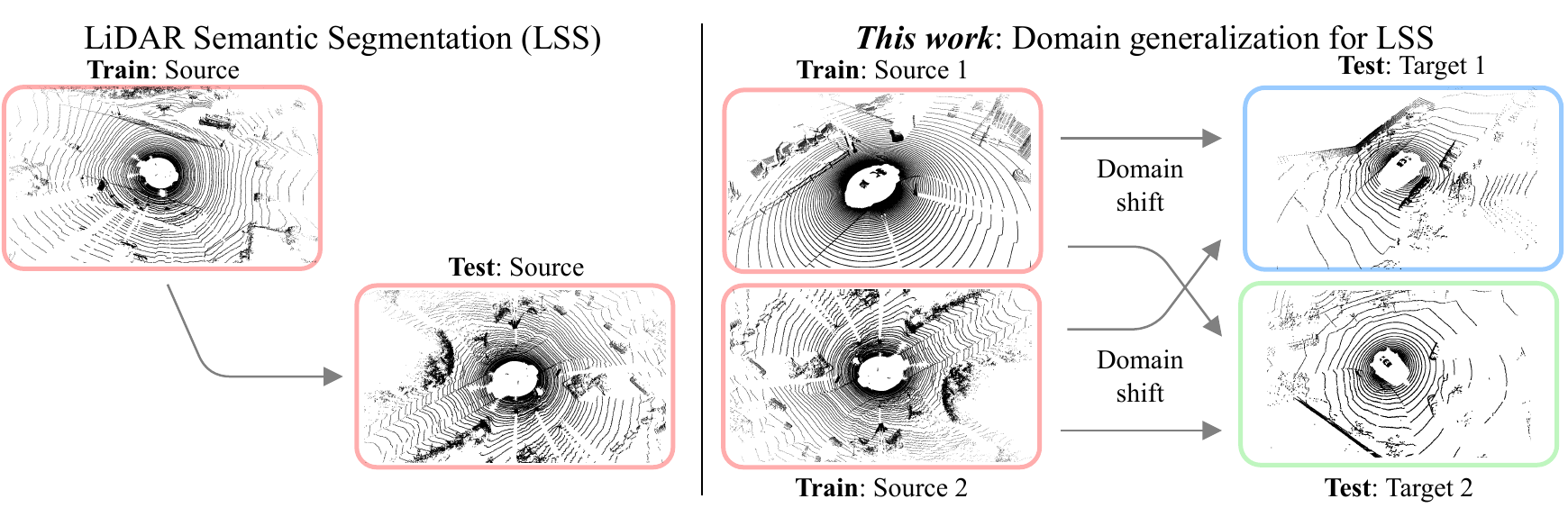}
\vspace{-10pt}
%
%
\captionof{figure}{\textbf{Domain Generalization for LiDAR Semantic Segmentation (DG-LSS)}. \textit{Left}: Existing LSS methods are trained and evaluated on point clouds drawn from the same domain. \textit{Right}: We focus on studying LSS under domain shifts, where the test samples are drawn from a different data distribution. Our paper aims to address the generalization aspect of this task.}

\label{fig:teaser}
\vspace{0.1cm}
\end{center}%
}]

\begin{abstract}
\vspace{-12pt}
The ability to deploy robots that can operate safely in diverse environments is crucial for developing embodied intelligent agents. As a community, we have made tremendous progress in within-domain \textit{LiDAR semantic segmentation}. However, do these methods generalize \textit{across} domains? 
To answer this question, we design the first experimental setup for studying \textit{domain generalization} (DG) for \textit{LiDAR semantic segmentation} (DG-LSS). Our results confirm a significant gap between methods, evaluated in a cross-domain setting: for example, a model trained on the source dataset (SemanticKITTI) obtains $26.53$ mIoU on the target data, compared to $48.49$ mIoU obtained by the model trained on the target domain (nuScenes). 
To tackle this gap, we propose the first method specifically designed for DG-LSS, which obtains $34.88$ mIoU on the target domain, outperforming all baselines. Our method augments a sparse-convolutional encoder-decoder 3D segmentation network with an additional, dense 2D convolutional decoder that learns to classify a birds-eye view of the point cloud. This simple auxiliary task encourages the 3D network to learn features that are robust to sensor placement shifts and resolution, and are transferable across domains. With this work, we aim to inspire the community to develop and evaluate future models in such cross-domain conditions.
\end{abstract}

\section{Introduction}


We address the challenge of achieving accurate \textit{and} robust semantic segmentation of LiDAR point clouds. LiDAR semantic segmentation (LSS) is one of the most fundamental perception problems in mobile robot navigation, with applications ranging from mapping~\cite{li2022hdmapnet}, localization~\cite{ma2019exploiting}, and online dynamic situational awareness \cite{thrun2006stanley}.\footnote{Semantic point classifier was part of the perception stack in early autonomous vehicles, such as Stanley, that won the DARPA challenge in 2005.} 

\textbf{Status Quo.} State-of-the-art LSS methods \cite{choy20194d, hu2020randla, thomas2019kpconv, zhu2021cylindrical} perform well when trained \textit{and} evaluated using the same sensory setup and environment (\ie, \source domain, Fig.~\ref{fig:teaser}, \textit{left}). 
However, their performance degrades significantly in the presence of domain shifts (Fig.~\ref{fig:teaser}, \textit{right}), commonly caused by differences in sensory settings, \eg, a new type of sensor, or recording environments, \eg, geographic regions with different road layouts or types of vehicles. 
One way to mitigate this is to collect multi-domain datasets for pre-training, similar to what has been done in the image domain using inexpensive cameras that are widely available~\cite{Deng09CVPR,Lin14ECCV}. However, building a crowd-sourced collection of multi-domain LiDAR datasets is currently not feasible. 

\textbf{Stirring the pot.} 
As a first step towards LiDAR segmentation models that are robust to domain shifts, we present the first-of-its-kind experimental test-bed for studying \textit{Domain Generalization} (DG) in the context of \textit{LiDAR Semantic Segmentation} (LSS). In our DG-LSS setup we train and evaluate models on different domains, including two synthetic~\cite{saltori2022gipso} and two real-world densely labeled datasets~\cite{Behley19ICCV,fong21arxiv}, recorded in different geographic regions with different sensors. 
This evaluation setup reveals a significant gap in terms of mean intersection-over-union between models trained on the \source and \target domains: for example, a model transferred from SemanticKITTI to nuScenes dataset obtains $26.53$ mIoU compared $48.49$ mIoU obtained by the model fully trained on the target domain. \textit{Could this gap be alleviated with DG techniques?}

%
%

\textbf{Insights.} 
To address this challenge, we propose \textit{LiDOG} (\textbf{Li}DAR \textbf{DO}main \textbf{G}eneralization) as a simple yet effective method specifically designed for DG-LSS. 
%
%
In addition to reasoning about the scene semantics in 3D space, LiDOG projects features from the sparse 3D decoder onto the 2D bird's-eye-view (BEV) plane along the vertical axis and learns to estimate a dense 2D semantic layout of the scene. In this way, LiDOG encourages the 3D network to learn features that are robust to variations in, \eg, type of sensor or geo locations, and thereby can be transferred across different domains. 
%
This directly leads to increased robustness toward domain shifts and yields $+8.35$ mIoU improvement on the \target domain, confirming the efficacy of our approach. Our experimental evaluation confirms this approach is consistently more effective compared to prior efforts in data augmentations~\cite{zhang2022pointcutmix, Nekrasov213DV, saltori2022cosmix}, domain adaptation techniques~\cite{wang2020train, langer2020iros}, and image-based DG techniques~\cite{choi2021robustnet, pan2018two}, applied to the LiDAR semantic segmentation. 

\textbf{Contributions.} We make the following key contributions. We (i) present the \textit{first} study on \textit{domain generalization} in the context of \textit{LiDAR semantic segmentation}. 
To this end, we (ii) carefully construct a test-bed for studying DG-LSS using two synthetic and two densely-labeled real-world datasets recorded in different cities with different sensors. 
This allows us to (iii) rigorously study how prior efforts proposed in related domains can be used to tackle domain shift and (iv) propose \lidog, a simple yet strong baseline that learns robust, generalizable, and domain-invariant features by learning semantic priors in the 2D birds-eye view. 
Despite its simplicity, we achieve state-of-the art performance in all the generalization directions. 
\footnote{Our code is available at \href{https://github.com/saltoricristiano/LiDOG}{https://saltoricristiano.github.io/lidog/}.} 

\section{Related Work}

\begin{figure*}[t]
    \centering
    \includegraphics[scale=0.9]{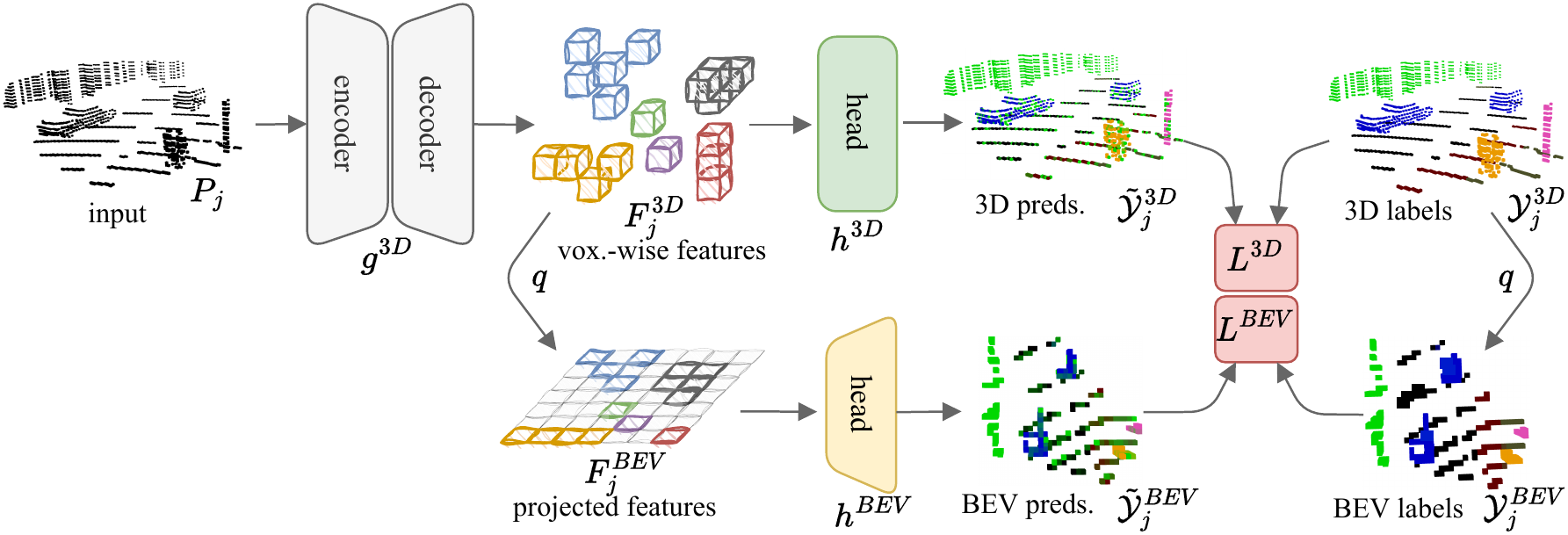}
    \vspace{-7pt}
    \caption{\textbf{\lidog overview}. 
    %
    We encode our input LiDAR scan $P_j$ using the 3D backbone $g^{3D}$ to learn the occupied voxels' feature representations $F^{3D}$.
    (\textit{Upper branch - main task}) We apply a sparse segmentation head on $F^{3D}$ and supervise with 3D semantic labels, $\mathcal{Y}_j^{3D}$.
    (\textit{Lower branch - auxiliary task}) We project those features along the height-axis to obtain a dense 2D bird's-eye (BEV) view features $F^{BEV}$, and apply several 2D convolutional layers to learn the 2D BEV representation. We supervise the BEV auxiliary task by using BEV-view of semantic labels, $\mathcal{Y}_j^{BEV}$.
    We train jointly on both $L^{3D}$ and $L^{BEV}$.}
    \label{fig:lidog}
\end{figure*}

\PAR{LiDAR Semantic Segmentation (LSS).}
LiDAR-based semantic scene understanding has been used in autonomous vehicle perception since the dawn of autonomous driving~\cite{thrun2006stanley}. State-of-the-art methods are data-driven and rely on different LiDAR data representations and network architectures. Point-based methods~\cite{qi2017pointnet++,hu2020randla,thomas2019kpconv,zhao2021point, wu2022point} learn representations directly from unordered point sets, whereas range-view-based methods~\cite{milioto2019rangenet++,wu2018squeezeseg, wu2019squeezesegv2,zhang2020polarnet,alonso2020MiniNet3D} learn a representation from a range pseudo-image that represents the LiDAR point cloud. State-of-the-art methods~\cite{choy20194d,3DSemanticSegmentationWithSubmanifoldSparseConvNet,zhu2021cylindrical,li2022panoptic,tang2020spvnas} map 3D points into a 3D quantized grid and learn representations using sparse 3D convolutions~\cite{3DSemanticSegmentationWithSubmanifoldSparseConvNet, choy20194d}. 
The works mentioned above focus on \textit{within-domain} performance, \ie, they train and test on data from the same dataset and do not evaluate robustness to domain shifts. By contrast, we focus on \textit{cross-domain} robustness, \ie, we train on a source domain and test on a target domain.

\PAR{Domain Adaptation for LSS.}
Domain shift in LiDAR perception~\cite{wang2020train, saltori2020sf, langer2020domain, qin2019pointdan} is a well-documented issue as sensors are expensive, and crowd-sourcing data collection is difficult.
\textit{Unsupervised} domain adaptation (UDA) techniques allow us to reduce the domain shift by looking at data from the \target domain, but without needing any labels in that domain.
Several methods reduce domain shift by mimicking sensory characteristics of the \target domain data, often in a data-driven manner~\cite{langer2020iros, zhao2021epointda, synlidar, xiao2022transfer}. 
Alternatively, domain shift can be reduced by feature alignment or prediction consistency~\cite{wu2019squeezesegv2, jaritz2019xmuda, jaritz2020xmuda}, or by training a point cloud completion network to map domains into a canonical domain~\cite{yi2021complete}. 
More recently, mixup strategies \cite{mixup2018, yun2019cutmix} have been introduced in the context of UDA for LSS~\cite{saltori2022cosmix, xiao2022polarmix}.
In LiDAR-based 3D object detection, scaling~\cite{wang2020train} and self-training~\cite{saltori2020sf, yang2021st3d} are often used strategies.
All aforementioned methods must be exposed to (unlabeled) data from the \target domain to adapt the learned representation.
In this work, we study DG-LSS with the goal of learning as-general-and-robust-as-possible representations in the first place, before exposure to \textit{any data from a particular target domain} during training.

\PAR{Domain Generalization.} Alternatively, robustness to domain shifts can be tackled via \textit{Domain Generalization} (DG). In DG, we aim to learn domain-invariant features by training solely on the \source domain data~\cite{zhou2021domain}. Due to the importance of learning robust representations, DG has been widely studied in the image domain~\cite{choi2021robustnet, zhou2021domain, zhao2022shade, pan2018two} in both, multi-~\cite{ghifary2016scatter, li2018learning, zhou2020deep} and single-source~\cite{carlucci2019domain, bucci2021self, zhao2022shade, wang2019learning_iclr, laskin2020curl} settings. Image-based DG approaches reduce domain shift via domain alignment~\cite{li2018deep, ghifary2016scatter}, meta-learning~\cite{li2018learning, balaji2018metareg}, data and style augmentations~\cite{shankar2018generalizing, zhou2020deep, zhao2022shade}, ensembling techniques~\cite{d2019domain, liu2020ms}, disentangled learning~\cite{khosla2012undoing, wang2020cross}, self-supervised learning~\cite{carlucci2019domain, bucci2021self}, regularization strategies~\cite{wang2019learning_iclr, huang2020self} and, reinforcement learning~\cite{huang2020self, laskin2020curl}. Data augmentations have also been employed in the context of DG in LiDAR-based 3D object detection in~\cite{lehner20223d} to learn to detect vehicles that have been damaged in car accidents.
In our work, we propose the first method specifically designed for DG-LSS and introduce a dense 2D convolutional decoder that learns features robust to domain shift.
In our experimental evaluation (Sec.~\ref{sec:experimental}), we adapt and compare to several state-of-the-art data augmentation techniques, such as Mix3D~\cite{Nekrasov213DV}, PointCutMix~\cite{zhang2022pointcutmix}, and PolarMix~\cite{xiao2022polarmix} that apply mixup strategies~\cite{mixup2018, yun2019cutmix} to 3D point clouds. Moreover, we adapt state-of-the-art methods from the image domain RobustNet~\cite{choi2021robustnet} and IBN~\cite{pan2018two}. We show that our method has substantially better generalization across all experiments. Finally, \textit{parallel work} by \cite{sanchez2022domain} accumulates consecutive point clouds over time to introduce robustness towards point clouds of different domains and, consequentially, improve generalization capabilities.

\section{A LiDAR Journey Across Domains}

In this section, we first recap the standard LiDAR Semantic Segmentation (LSS) setting for completeness. Next, we formally introduce the problem of Domain Generalization for LiDAR Semantic Segmentation (DG-LSS), which we study in this paper.

\PAR{LiDAR Semantic Segmentation.} 
In LSS, we are given a set of labeled training instances in the form of point clouds  $\mathcal{T}_{train} = \{ P_j = \{ \mathbf{p}_i \in \mathbb{R}^3\}_{i=1}^{N_j} \}$. Each point cloud $P_j \sim \mathcal{D}$ consists of $N_j$ points, and is drawn from a certain point cloud distribution $\mathcal{D}$. 
Moreover, point clouds in the training set are labeled according to a predefined semantic class vocabulary $\mathcal{K} = \{1, \ldots, K\}$ of $K$ categorical labels. 
Given such a training set $\mathcal{T}_{train}$ along with class vocabulary $\mathcal{K}$, the goal of LSS is to learn a function $f(\mathbf{p}_i) \to \mathcal{K}$, that maps each point $\mathbf{p}_i \in P_j,  \forall j,$ to one of $K$ semantic classes, such that the prediction error on the \textit{same domain} $\mathcal{D}$ is minimized. 
In other words, LSS models are trained \textit{and} evaluated on (disjunct) sets of point clouds that are drawn from the same data distribution $\mathcal{D}$.
We note that it is standard practice for LSS methods to be evaluated on multiple datasets\footnote{At the moment, SemanticKITTI and nuScenes-lidarseg are commonly used for the evaluation.}. However, standard multi-dataset evaluation follows the setting described above: models are trained \textit{and} evaluated for \textit{each} domain \textit{separately}.

\PAR{Domain Generalization for LiDAR Semantic Segmentation.}
In this work, we release the assumption that train and test splits were sampled from the same domain and explicitly study cross-\textit{domain generalization} in LSS. 
We are given $M$ labeled datasets $\mathcal{T}_m \sim \mathcal{D}_m$, $m \in \{ 1, \ldots, M \}$ sampled from different data distributions $\mathcal{D}_m$. 
Importantly, all datasets must be labeled according to a common semantic vocabulary $\mathcal{K}$, \ie, we are studying DG in closed-set conditions~\cite{zhou2021domain}. 
Then, the task can be defined as follows: we use $\mathcal{T}_{\{m\}}$ datasets for training, and the held-out datasets $\mathcal{T}_{\{m'\}}, \{m'\} \cap \{m\} = \emptyset$, solely for testing. 

The goal of DG-LSS is, therefore, to train a mapping $f(\mathbf{p}_i) \to \mathcal{K}$, such that the prediction performance on the target datasets $\mathcal{T}_{\{m'\}} \sim \mathcal{D}_{\{m'\}}$ is maximized. 
We only have access to data and labels from the \target datasets during evaluation. The model has never \textit{seen any target data} before evaluation, neither labeled nor unlabeled.


\section{Walking Your LiDOG}

\PAR{Overview.} We provide a schematic overview of \lidog: \textit{Li}dar \textit{DO}main \textit{G}eneralization network in Fig.~\ref{fig:lidog}. LiDOG leverages a 3D sparse-convolutional encoder-decoder neural network (Fig.~\ref{fig:lidog}, $g^{3D}$) that encodes the input point cloud $P_j$ into sparse 3D features $F^{3D}$ (Fig.~\ref{fig:lidog}). 
%
In order to tackle the domain shift problem, we propose to augment this network with a dense top-down prediction auxiliary task during model training. In particular, we project sparse 3D features $F^{3D}$ from the decoder network along the height axis to obtain a 2D bird's-eye view (BEV) representation $F^{BEV}$ of the features learned by the 3D network. Then, we jointly optimize the losses for the 3D and BEV network heads. 
Importantly, the auxiliary BEV head is not required during model inference. Its sole purpose is to effectively learn domain-agnostic features during training. 



\subsection{3D Segmentation Branch}
\label{sec:3d_segmentation}
The upper branch of \lidog (Fig.~\ref{fig:lidog}) aims to learn the main segmentation task, \ie, it learns the mapping function $f(\mathbf{p}_i) \to \mathcal{K}$. 
We first process the input LiDAR point cloud $P_j$ into a 3D occupancy (voxel) grid $V_j$. During this phase, the continuous 3D input space is evenly sampled into discrete 3D cells. Points falling into the same cell are merged.
We then learn a feature representation of the occupancy grid $V_j$ via a 3D encoder-decoder network $g^{3D}$ based on the well-established sparse-convolutional backbone~\cite{choy20194d}. 
The encoder consists of a series of sparse 3D convolutional downsampling layers that learn a condensed 3D representation, while the decoder upsamples the features back to the original resolution with sparse convolutional upsampling layers. We use sparse batch-normalization layers. 
The output of $g^{3D}$ is a set of voxel-wise features $F^{3D}_j = g^{3D}(V_j)$, where each feature vector is associated with a voxel cell in $V_j$.
We employ a sparse segmentation head to obtain semantic posteriors that predict the voxel-wise categorical distribution from the voxel features $F^{3D}_j$: $\tilde{\mathcal{Y}^{3D}_j} = p(\mathcal{K}|V_j)= \sigma(h^{3D}(F^{3D}_j))$, where $\sigma$ denotes the softmax activation function.

\subsection{Dense Auxiliary Task}
\label{sec:time_bev}

In the lower branch of \lidog, we add the auxiliary task that encourages the network to learn a representation that is robust to domain shifts. We transform 3D features to bird's-eye view (BEV) space, and train a network to estimate BEV semantic scene layout.


\PAR{Sparse-to-dense feature projection.} We start our dense-to-sparse segmentation task by projecting voxel-wise features $F_j^{3D}$ into dense BEV-features feature space $F_j^{BEV}$.
%
%
Given a voxel $v_i \in V_j$ center coordinates $(x_i, y_i, z_i)$, we compute its corresponding BEV coordinates $(x^{BEV}_i, z^{BEV}_i)$ as:
\begin{equation}
    q(x_i, z_i) = (x^{BEV}_i, z^{BEV}_i) =  \Bigl(\Bigl\lfloor\frac{x_i}{x_q} \Bigl\rfloor, \Bigl\lfloor\frac{z_i}{z_q} \Bigl\rfloor\Bigr),
    \label{eq:feature_projection}
\end{equation}
%
where $q$ is the projection function, and $x_q$ and $z_q$ are the quantization parameters. We project only voxels within projection bounds $B^{3D} = [(-b_x, b_x), (-b_z, b_z)]$, where $b_x$ and $b_z$ are the bound values along the $x$ and $z$ axis, and shift projected coordinates to fit in the BEV-features height and width.
Consequently, we define BEV features $F_j^{BEV}$ by initializing them to zero and by filling non-empty cells as $F_j^{BEV}(x^{BEV}_i, z^{BEV}_i) = F^{3D}_i$.
When multiple voxels project to the same cell, we randomly keep one.

\PAR{Sparse-to-dense label projection.}
We obtain source labels by following the same procedure used for $F_j^{BEV}$. After voxelization, each voxel $v_i$ is associated to a label category $y^{3D}_i \in \mathcal{Y}^{3D}_j$. We use Eq.~\ref{eq:feature_projection} and the same bounds $B^{3D}$ to compute the label coordinates $(x^{BEV}_i, z^{BEV}_i)$ of BEV image pixels. Then, we define BEV labels as: $\mathcal{Y}_j^{BEV}(x^{BEV}_i, z^{BEV}_i) = y^{3D}_i$.
%

\PAR{BEV predictions.}
We predict BEV semantic posteriors by employing a standard 2D segmentation head $h^{BEV}$ which predicts semantic classes in the dense 2D BEV plane and takes as input BEV features $F_j^{BEV}$.
First, we reduce the dimensionality of $F_j^{BEV}$ through a pooling operation. Then, we feed $h^{BEV}$ with $F_j^{BEV}$ and obtain BEV semantic predictions: $\tilde{\mathcal{Y}^{BEV}_j} = \sigma(h^{BEV}(F_j^{BEV}))$, where $\sigma$ denotes the softmax activation function.

\subsection{Losses}
\label{sec:joint_objective}

Given 3D semantic predictions $\tilde{\mathcal{Y}_j^{3D}}$ and labels $\mathcal{Y}_j^{3D}$, as well as BEV semantic predictions $\tilde{\mathcal{Y}_j^{BEV}}$ and labels $\mathcal{Y}_j^{BEV}$, we train both \lidog network heads jointly using soft DICE loss~\cite{Jadon2020}, \ie, $L^{3D} = dice(\tilde{\mathcal{Y}_j^{3D}}, \mathcal{Y}_j^{3D})$, and $L^{BEV} = dice(\tilde{\mathcal{Y}_j^{BEV}}, \mathcal{Y}_j^{BEV})$. 
Finally, we average contributions of both losses: $L_{tot} = \frac{1}{2}(L^{BEV} + L^{3D})$. 
DICE loss has been shown in literature~\cite{Jadon2020, saltori2022cosmix} to perform favorably for rarer classes compared to the cross-entropy loss. 


\begin{figure}[t]
    \centering
    \setlength\tabcolsep{1.pt}
    \begin{tabular}{cccc}
    \raggedright
        \begin{overpic}[width=0.28\columnwidth]{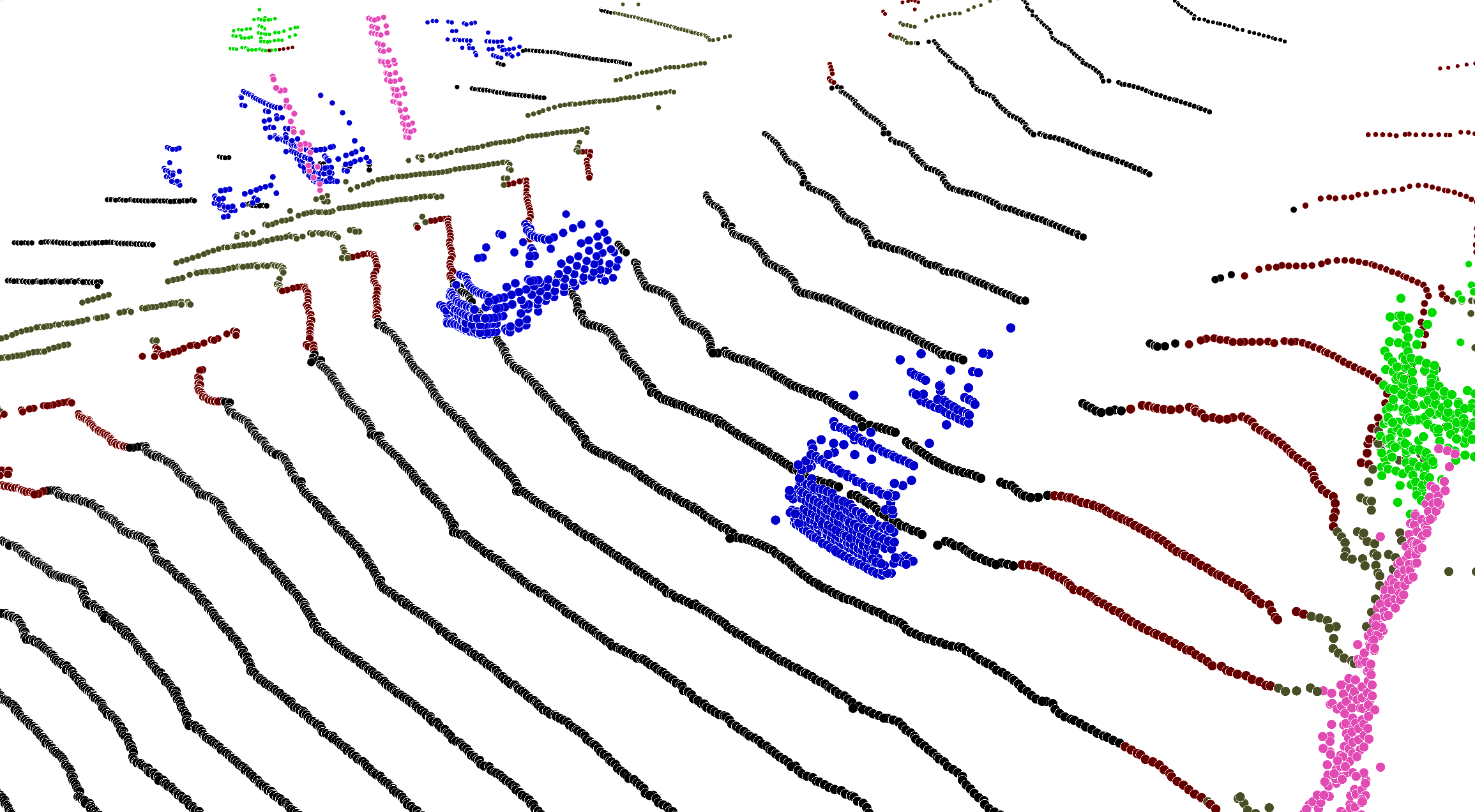}
        \end{overpic} &
        \hspace{1mm}
        \begin{overpic}[width=0.17\columnwidth]{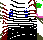}
        \end{overpic}&
        \hspace{1mm}
        \begin{overpic}[width=0.28\columnwidth]{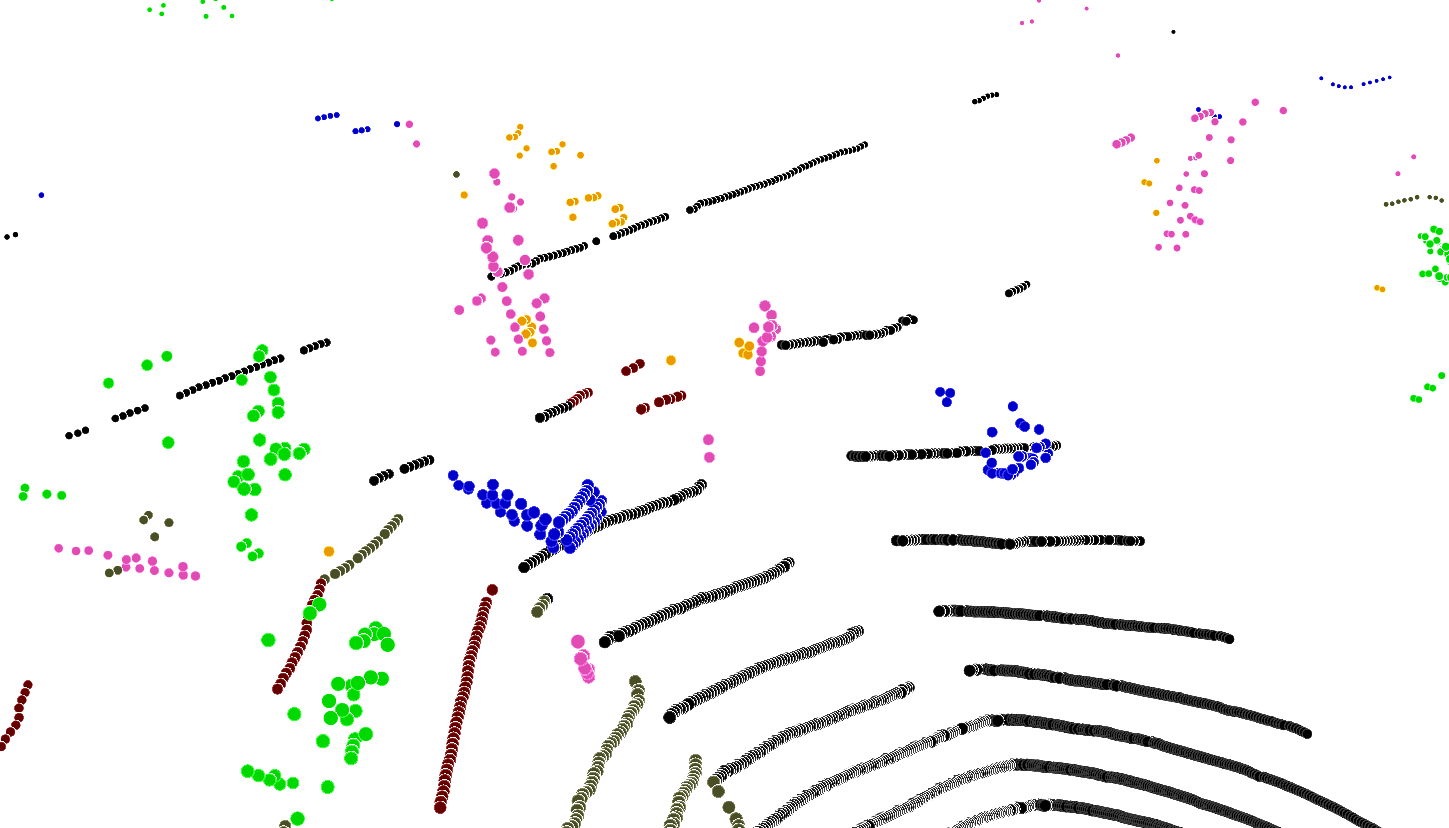}
        \end{overpic} &
        \hspace{1mm}
        \begin{overpic}[width=0.16\columnwidth]{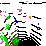}
        \end{overpic}
    \end{tabular}
    \vspace{-10pt}
    \caption{LiDAR point clouds and their corresponding BEV views: SemanticKITTI (\textit{left}) and nuScenes (\textit{right}). After projection, BEV images are geometrically more similar.}
    \label{fig:bev_motivations}
\end{figure}

\subsection{Why Does This Work?}
\begin{figure}[t]
\centering
    \setlength\tabcolsep{1pt}
    \begin{tabular}{cc}
    \raggedright
        \begin{overpic}[width=0.46\columnwidth]{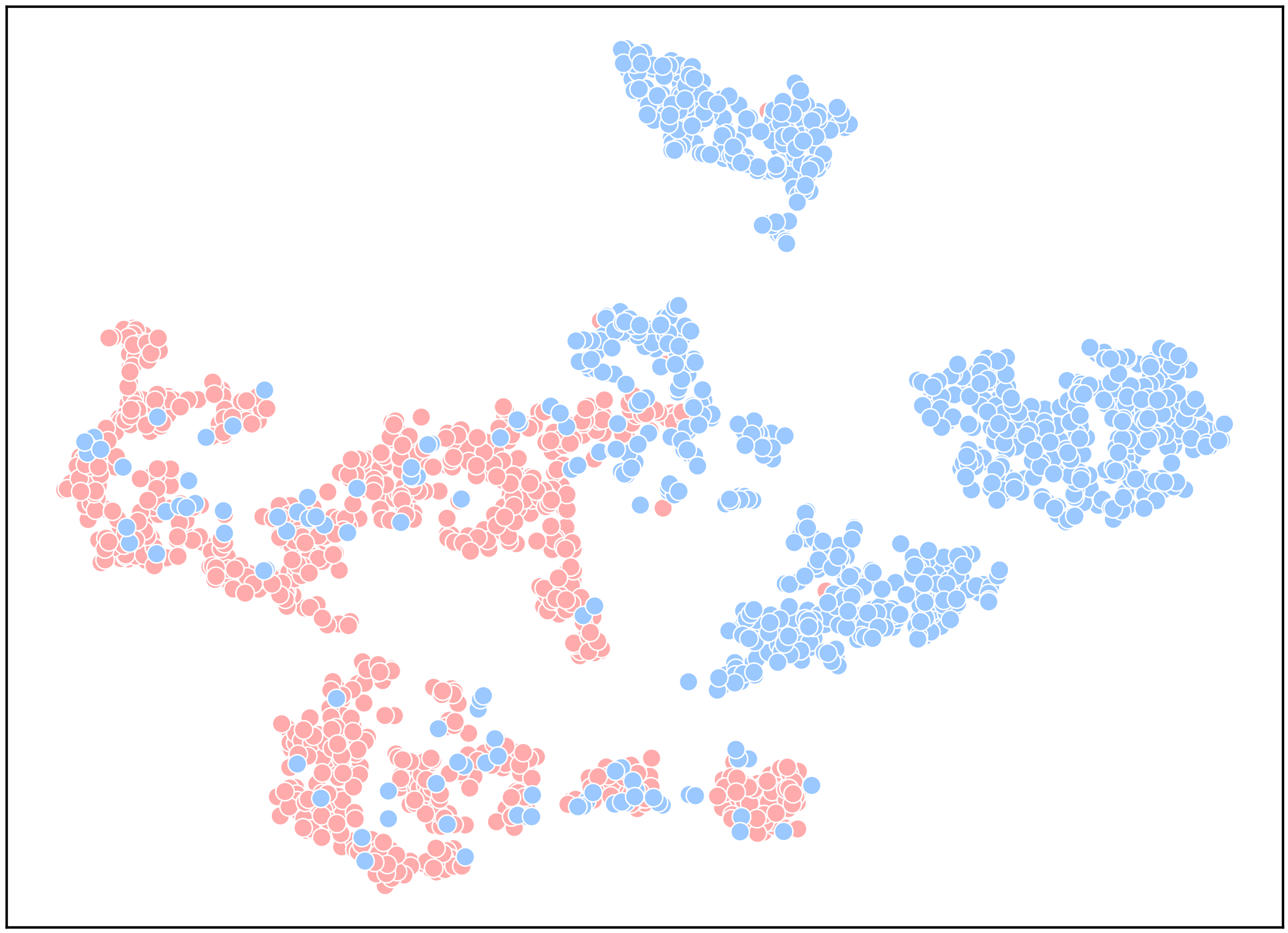}
        \put(2,65){\color{black}\footnotesize w/o BEV}
        \put(-7,2){\rotatebox{90}{\color{black}\footnotesize Synth4D-k.$\to$S.KITTI}}
        \end{overpic} &  
        \begin{overpic}[width=0.46\columnwidth]{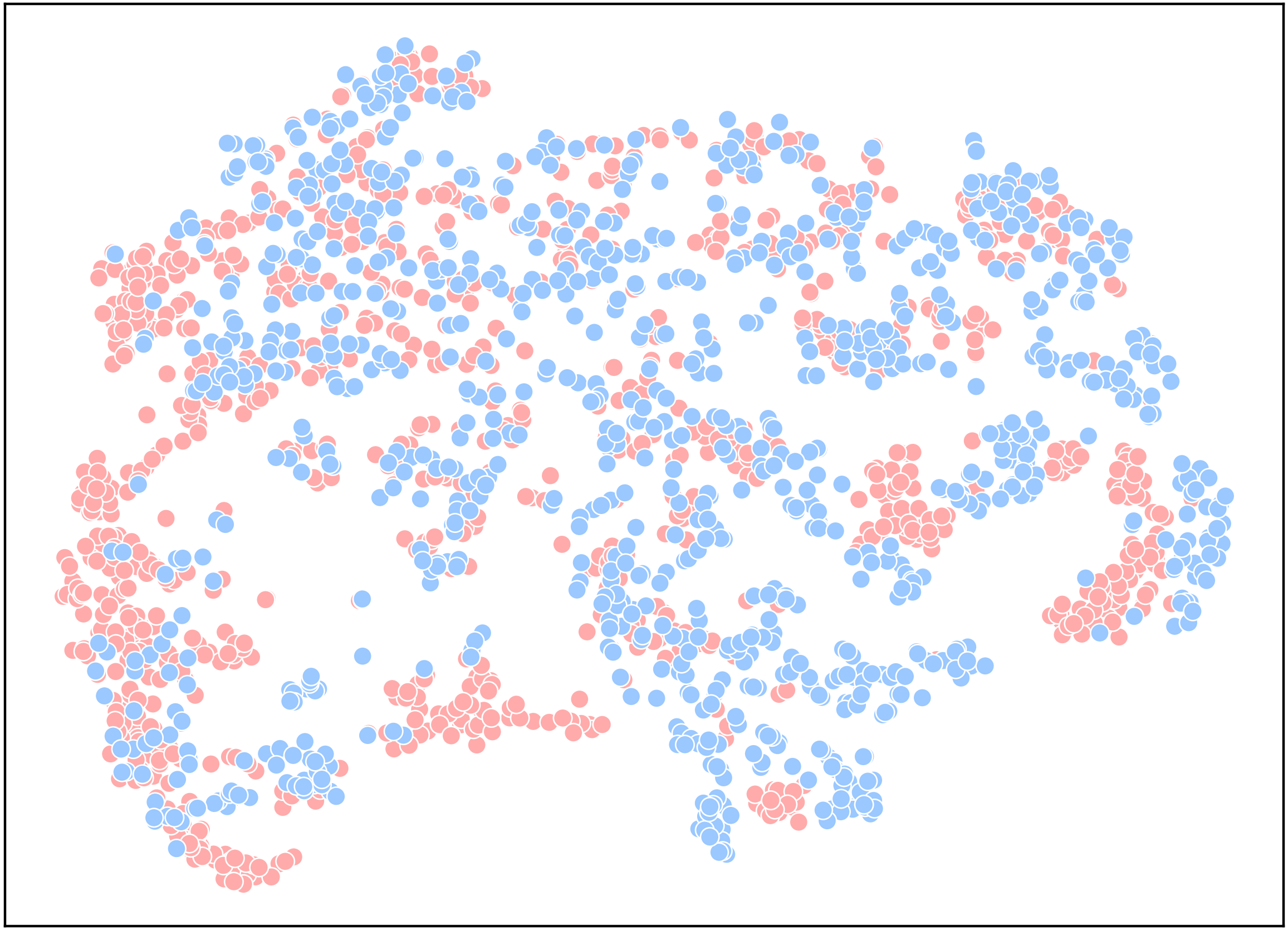}
        \put(2,65){\color{black}\footnotesize BEV}
        \end{overpic}\\
        \begin{overpic}[width=0.46\columnwidth]{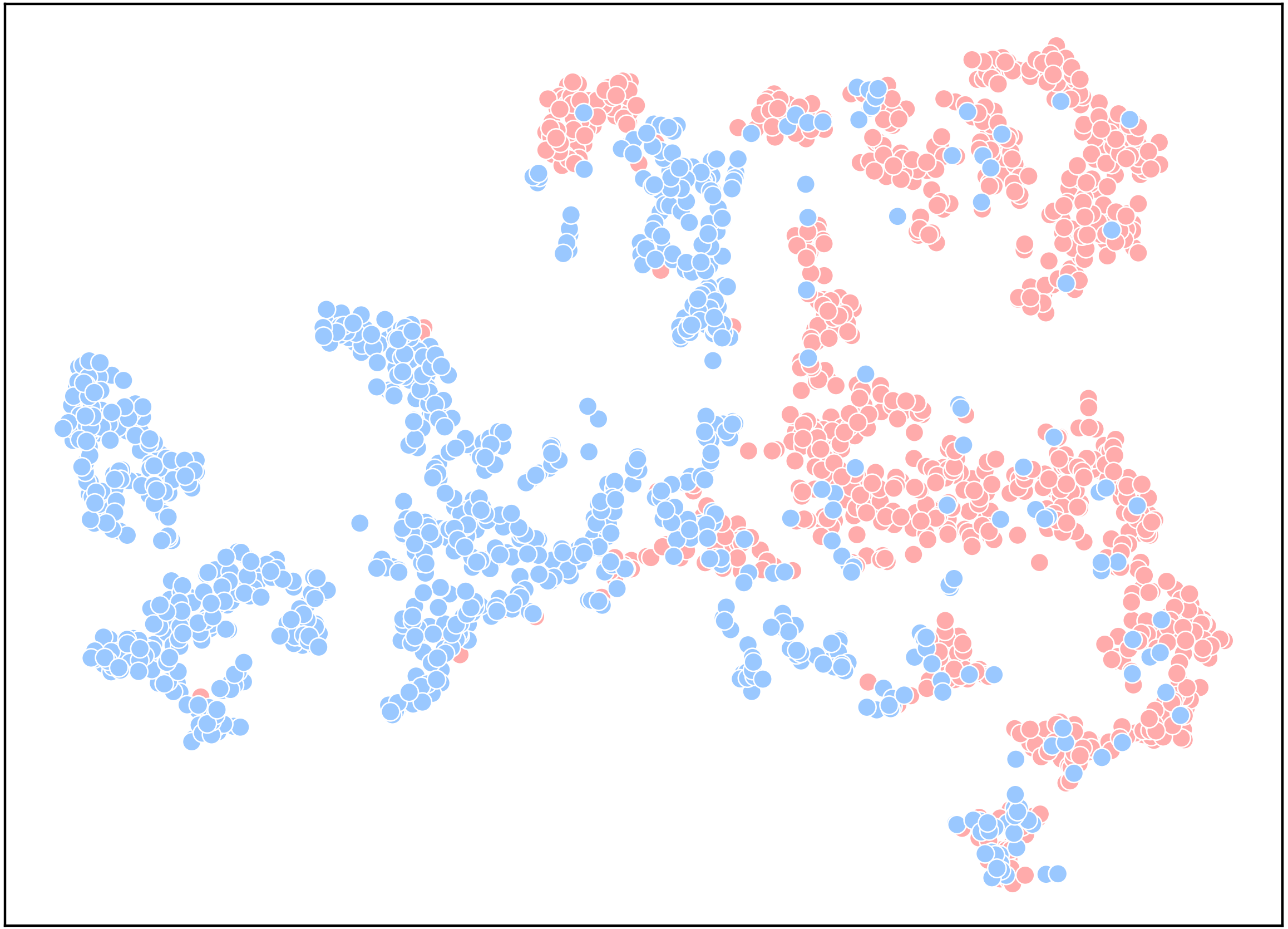}
        \put(2,65){\color{black}\footnotesize w/o BEV}
        \put(-7,5){\rotatebox{90}{\color{black}\footnotesize Synth4D-k.$\to$nuSc.}}
        \end{overpic}& 
        \begin{overpic}[width=0.46\columnwidth]{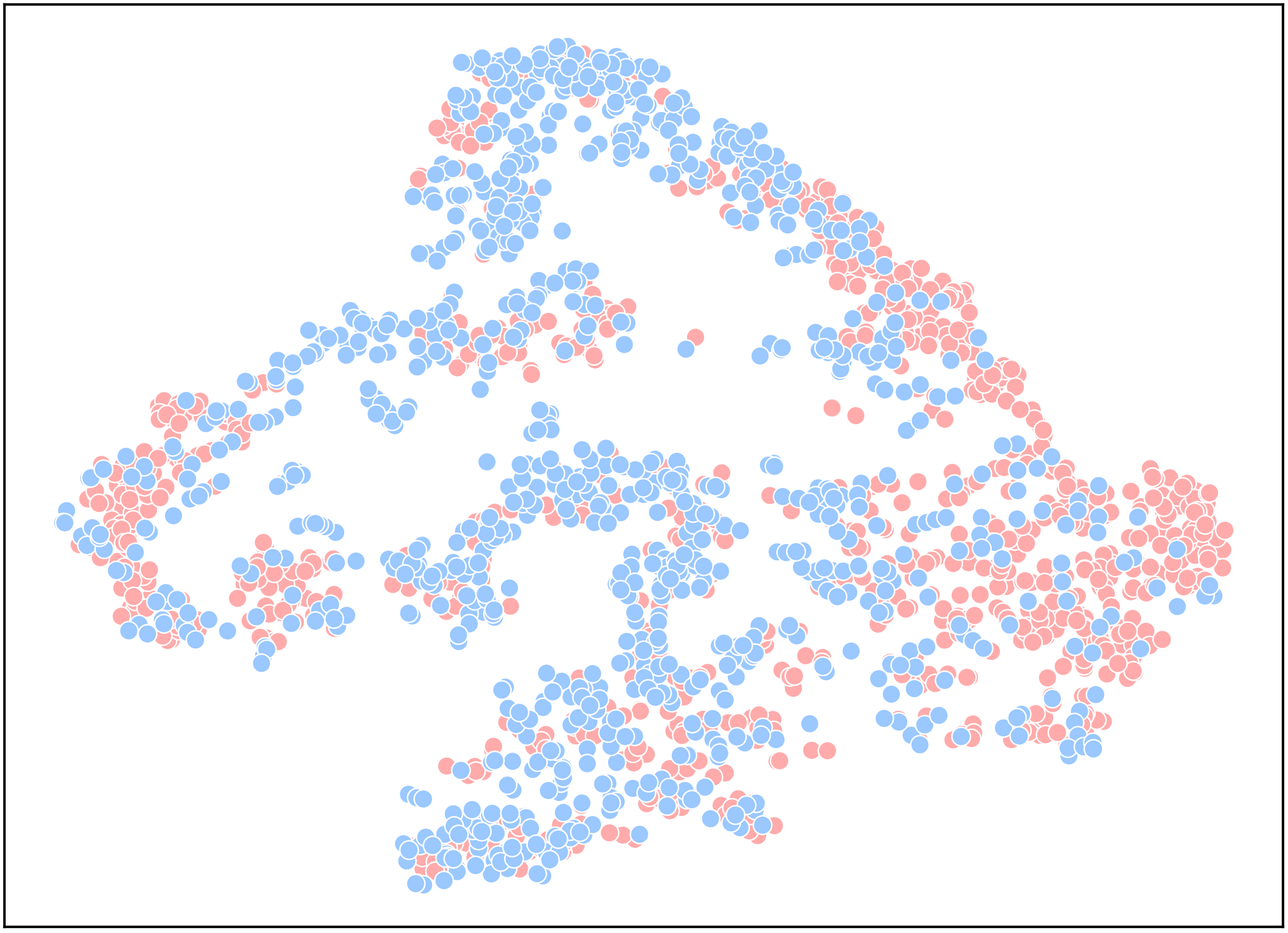}
        \put(2,65){\color{black}\footnotesize BEV}
        \end{overpic}
    \end{tabular}
    \vspace{-10pt}
    \caption{\textbf{Feature visualization}: t-SNE visualization~\cite{Maaten08JMLR} of the point embeddings 
    for the \textit{road} class, obtained by training our network without (w/o BEV, \textit{left}) and with BEV task (BEV, \textit{right}). \textit{Top:} Synth4D-KITTI$\to$SemanticKITTI. \textit{Bottom:} Synth4D-KITTI$\to$nuScenes. As shown, projected features are well-aligned when training the network with the auxiliary BEV task.}
    \label{fig:tsne}
\end{figure}

We demonstrate empirically in Sec.~\ref{sec:experimental} that the dense BEV prediction proxy task significantly improves model performance in terms of cross-domain generalization. Why does this simple objective yield, in practice, more robust feature representations? There are several possible sources of domain shift, \eg, different sensors, different environments, and scans are visibly different (Fig.~\ref{fig:bev_motivations}).
After the projection, we obtain a denser label space that is less sensitive to individual sensor characteristics: in the BEV space scans from different domains look \textit{more alike}, Fig.~\ref{fig:bev_motivations}. We effectively regularize the model training by promoting the network robustness to changes in the sensor acquisition patterns. 
%
%
We validate this intuition in Fig.~\ref{fig:tsne} by visualizing learned point embeddings~\cite{Maaten08JMLR} for the \textit{road} class with and without BEV auxiliary task, across multiple domains. As can be seen, projected embeddings are consistently closer when employing BEV auxiliary task.

\section{Experimental Evaluation}
\label{sec:experimental}

In this section, we report our experimental evaluation. In Sec.~\ref{sec:test-bed}, we describe our evaluation protocol, datasets and baselines. 
In Sec.~\ref{sec:synth2real}-\ref{sec:real2real}, we report our \textit{synth-to-real} and \textit{real-to-real} results, and discuss qualitative results.
In Sec.~\ref{sec:ablations}, we ablate our proposed approach.
We provide implementation details in the appendix.

\subsection{Experimental Protocol}
\label{sec:test-bed}

We base our evaluation protocol on two synthetic (Synth4D~\cite{saltori2022gipso}) and two real-world datasets (SemanticKITTI~\cite{Behley19ICCV} and nuScenes~\cite{fong21arxiv}). We study (i) generalization from \textit{synthetic$\rightarrow$real data} in \textit{single-} and \textit{multi-source} settings, \ie, trained on one or multiple source datasets, as well as (ii) \textit{real$\rightarrow$real} generalization. 

\PAR{Synthetic datasets.} Synth4D~\cite{saltori2022gipso} is a synthetic dataset generated using CARLA~\cite{Dosovitskiy17} simulator. It provides two synthetic source domains, each one with $20k$ labeled scans, emulating Velodyne64HDLE and Velodyne32HDLE sensors, similar to those used to record SemanticKITTI (\textit{Synth4D-KITTI}) and nuScenes-lidarseg datasets (\textit{Synth4D-nuScenes}). These datasets were generated in the same synthetic environment, mimicking different sensor types. We use the official training splits in all experiments. 

\PAR{Real datasets.} SemanticKITTI~\cite{Behley19ICCV} was recorded in various regions of Karlsruhe, Germany, with Velodyne64HDLE sensor, and provides $20k$ labeled scans. nuScenes~\cite{fong21arxiv} was recorded in Boston, USA, and Singapore, with a sparser Velodyne32HDLE sensor and provides $40k$ labeled LiDAR scans. We use the official training and validation splits for both datasets in all experiments.

\PAR{Class vocabulary.} To ensure this task is well-defined, we formalize cross-dataset consistent and compatible semantic class vocabulary, which ensures there is a one-to-one mapping between all semantic classes. We follow~\cite{saltori2022gipso} which provides a common label space among Synth4D, SemanticKITTI and nuScenes-lidarseg consisting of seven semantic classes: \texttt{vehicle}, \texttt{person}, \texttt{road}, \texttt{sidewalk}, \texttt{terrain}, \texttt{manmade} and \texttt{vegetation}. We perform all evaluation with respect to these classes. 

\PAR{Baselines.} We are the first in studying DG-LSS, therefore, we construct our set of baselines by considering the previous efforts in bridging the domain shift in other LiDAR tasks as well DG methods for images.
%
%

\PARit{Data augmentation baselines (Aug).} Data augmentations are used in supervised learning to improve model generalization and, in DA, to reduce domain shift. We re-implement and evaluate in the DG context three data augmentation approaches: Mix3D~\cite{Nekrasov213DV}, PointCutMix~\cite{zhang2022pointcutmix} and CoSMix~\cite{saltori2022cosmix}. Mix3D concatenates both points and labels of different scenes, obtaining a single mixed-up scene. PointCutMix and CoSMix follow the same strategy but mix either patches or semantic regions, respectively. We follow the official implementations and apply these data augmentation strategies during source training. For single-source training, we mix between samples of the same domain, while for multi-source training, we mix between samples of the two source domains. 
\PARit{Image-based baselines (2D DG).}
Image-based DG is a widely explored field. However, not all the methods can be extended to DG-LSS due to their image-related assumptions, \eg, style consistency.
We identify and re-implement two image-based approaches that can be extended to DG-LSS: IBN~\cite{pan2018two} and RobustNet~\cite{choi2021robustnet}.
IBN makes use of both batch and instance normalization layers in the network blocks to boost generalization performance. We follow the IBN-C block scheme and use it to build the sparse model. 
RobustNet~\cite{choi2021robustnet} makes use of IBN blocks and introduces an instance whitening loss. Similarly, we start from our IBN implementation and use their instance-relaxed whitening loss~\cite{choi2021robustnet} during training.

%
\PARit{UDA for LiDAR baselines (3D UDA).} Most of the UDA for LSS approaches assume (unlabeled) target data available for training/adaptation~\cite{yang2021st3d}, hence, we cannot adapt and evaluate them in the DG setting fairly. We identify two UDA baselines that assume only weak supervision from the target domain, \eg, vehicle dimensions or sensor specs. We re-implement and extend SN~\cite{wang2020train} and RayCast~\cite{langer2020domain}.
SN~\cite{wang2020train} uses the knowledge of average vehicle dimensions from source and target domains to re-scale source instances. We employ DBSCAN to isolate vehicle instances in both source and target domains and estimate the ratio between their dimensions. At training time, we follow their pipeline and train our segmentation model on the re-scaled point clouds.
RayCast~\cite{langer2020iros} employs ray casting to re-sample source data by mimicking target sensor sampling. We use the official code to re-sample source data and use mapped data during training.
Notice that both these baselines use a-priori knowledge from the target domain, \ie, vehicle dimensions or target sensor specifications, and can thus be considered as weakly-supervised baselines. 

\subsection{Synth $\to$ Real Evaluation}
\label{sec:synth2real}

\begin{table*}[t]
    \centering
    \caption{\textbf{Synth4D-KITTI$\rightarrow$Real, single-source.} Our approach (\lidog) improves upon \textit{Source} model on both real datasets: $+19.49$ mIoU for SemanticKITTI and $+16.52$ mIoU for nuScenes, outperforming all baselines. 
    Lower bound~(\textit{red}): a model trained on the source domain without the help of DG techniques. Upper bound~(\textit{blue}): model directly trained on the target data.}
    \vspace{-10pt}
    \small 
    \setlength{\tabcolsep}{3pt}
\resizebox{0.99\linewidth}{!}{
\begin{tabular}{cc|ccccccccc|c}
\toprule 
S & T & Info & Method & Vehicle & Person & Road & Sidewalk & Terrain & Manmade & Vegetation & mIoU\tabularnewline
\midrule 

& & \marksrc Lower bound & \marksrc Source & \marksrc 34.33 & \marksrc 4.47 & \marksrc 36.38 & \marksrc 13.27 & \marksrc 19.48 & \marksrc 23.1 & \marksrc 41.81 & \marksrc 24.69\tabularnewline
\cmidrule{3-12}
\parbox[t]{2mm}{\multirow{9}{2mm}{\rotatebox{90}{Synth4D-KITTI}}} & \parbox[t]{2mm}{\multirow{9}{2mm}{\rotatebox{90}{SemanticKITTI}}}& \multirow{3}{*}{Aug} & Mix3D~\cite{Nekrasov213DV} & 59.99 & 14.02 & 55.75 & 17.71 & 25.67 & 39.26 & 53.00 & 37.92 \tabularnewline
 & &  & PointCutMix~\cite{zhang2022pointcutmix} & 58.26 & 14.35 & 67.74 & 13.91 & \textbf{27.26} & \textbf{49.21} & \textbf{64.22} & 42.14\tabularnewline
 & &  & CoSMix~\cite{saltori2022cosmix} & 43,01 & 14.16 & 61.84 & 12.38 & 18.84 & 18.00 & 57.56 & 32.26\tabularnewline
\cmidrule{3-12} 
&  & \multirow{2}{*}{2D DG} & IBN~\cite{pan2018two} & 24.95 & 9.18 & 56.82 & 17.85 & 7.21 & 21.59 & 44.65 & 26.04\tabularnewline
 & &  & RobustNet~\cite{choi2021robustnet} & 50.85 & 14.97 & 58.71 & 7.83 & 19.96 & 42.58 & 44.52 & 34.20\tabularnewline
\cmidrule{3-12} 
 & & \multirow{2}{*}{3D UDA} & SN~\cite{wang2020train} & 49,38 & 14.83 & 68.53 & 18.45 & 25.62 & 37.49 & 59.48 & 39.11\tabularnewline
 & &  & RayCast~\cite{langer2020domain} & 51.73 & 4.33 & 56.43 & 18.07 & 23.91 & 36.23 & 40.52 & 33.03\tabularnewline
 \cmidrule{3-12}
 & & 3D DG & Ours (\lidog) & \textbf{72.86} & \textbf{17.1} & \textbf{71.85} & \textbf{28.48} & 11.53 & 46.02 & 61.42 & \textbf{44.18}\tabularnewline
\cmidrule{3-12}
 & & \marktar Upper bound & \marktar Target & \marktar 81.57 & \marktar 23.23 & \marktar 82.09 & \marktar 57.64 & \marktar 46.84 & \marktar 64.14 & \marktar 75.21 & \marktar 61.53\tabularnewline
\midrule
\midrule
& & \marksrc Lower bound & \marksrc Source & \marksrc 19.99 & \marksrc 6.80 & \marksrc 32.85 & \marksrc 6.81 & \marksrc 11.95 & \marksrc 45.07 & \marksrc 20.86 & \marksrc 20.62\tabularnewline
\cmidrule{3-12}
\parbox[t]{2mm}{\multirow{9}{*}{\rotatebox{90}{Synth4D-KITTI}}} &\parbox[t]{2mm}{\multirow{9}{*}{\rotatebox{90}{nuScenes}}} & \multirow{3}{*}{Aug} & Mix3D~\cite{Nekrasov213DV} & 26,8 & \textbf{22.68} & 41.90 & 7.31 & 13.54 & 48.17 & 52.09 & 30.36\tabularnewline
 & &  & PointCutMix~\cite{zhang2022pointcutmix} & 23,97 & 19.49 & 44.27 & 6.29 & 12.14 & 48.85 & 55.44 & 30.06\tabularnewline
 & &  & CoSMix~\cite{saltori2022cosmix} & 16.94 & 15.78 & 52.97 & 2.09 & 5.55 & 15.6 & 43.85 & 21.83\tabularnewline
\cmidrule{3-12}
 & & \multirow{2}{*}{2D DG} & IBN~\cite{pan2018two} & 21,32 & 11.96 & 36.83 & 5.39 & 11.25 & 35.91 & 37.46 & 22.87\tabularnewline
 & &  & RobustNet~\cite{choi2021robustnet} & 26.19 & 7.47 & 43.85 & 2.29 & 13.93 & 43.63 & 46.32 & 26.24\tabularnewline
\cmidrule{3-12}
 & & \multirow{2}{*}{3D UDA} & SN~\cite{wang2020train} & 23.14 & 14.08 & 51.60 & 11.38 & 14.02 & 46.91 & 50.83 & 30.28\tabularnewline
 & &  & RayCast~\cite{langer2020domain} & 19.54 & 11.78 & 56.53 & 6.66 & 8.45 & 45.66 & 36.99 & 26.52\tabularnewline
 \cmidrule{3-12}
 & & 3D DG & Ours (\lidog)  & \textbf{31.29} & 19.62 & \textbf{64.66} & \textbf{14.21} & \textbf{15.63} & \textbf{57.3} & \textbf{57.27} & \textbf{37.14}\tabularnewline
\cmidrule{3-12}
 & & \marktar Upper bound & \marktar Target & \marktar 47.42 & \marktar 20.18 & \marktar 80.89 & \marktar 37.02 & \marktar 34.99 & \marktar 64.27 & \marktar 54.67 & \marktar 48.49\tabularnewline
\bottomrule 
\end{tabular}
}
\label{tab:synth4d-semkitti}
\end{table*}

\label{sec:qualitative}
\begin{figure*}[t]
\centering
    \setlength\tabcolsep{1.pt}
    \begin{tabular}{cccc}
    \raggedright
        \begin{overpic}[width=0.24\textwidth]{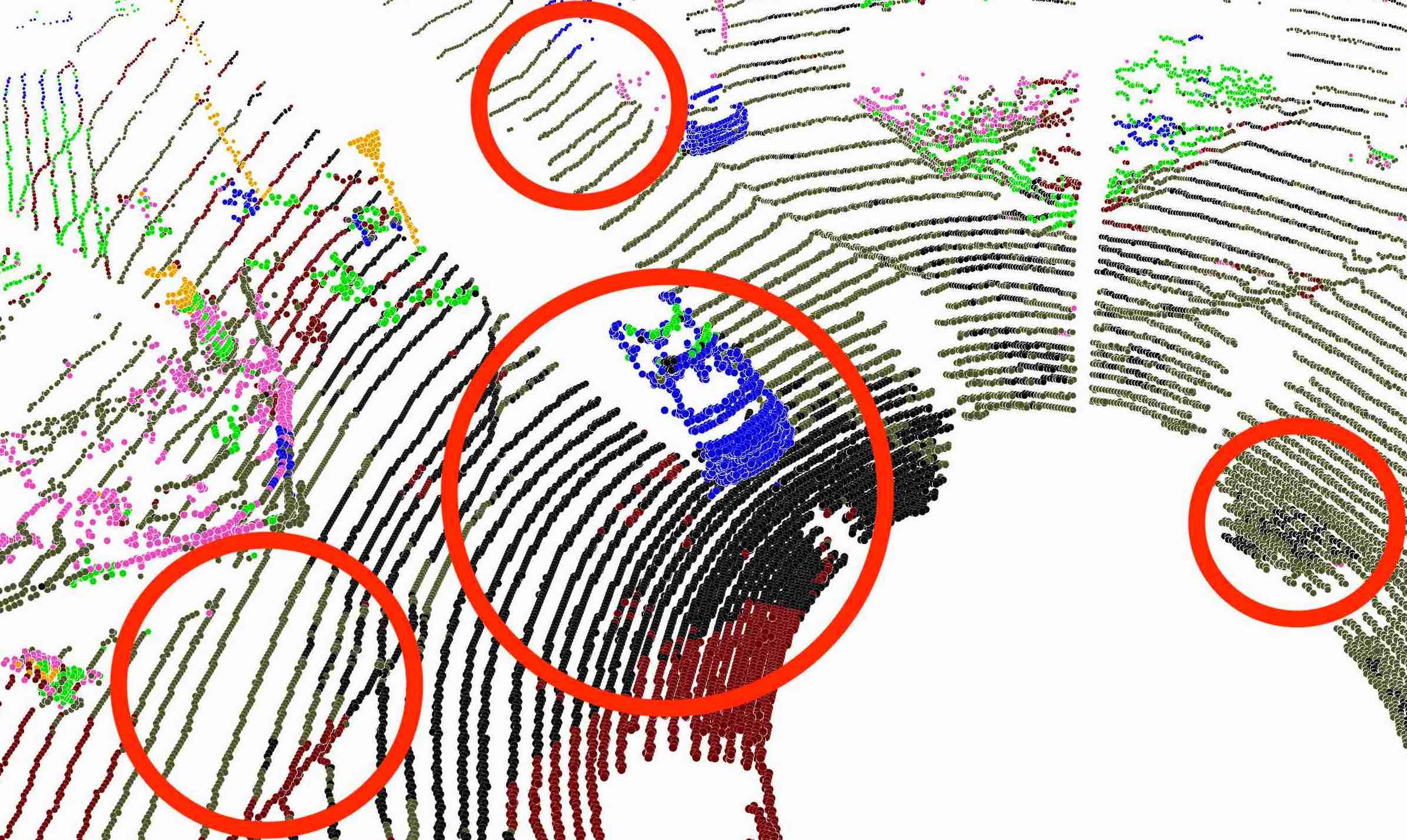}
        \end{overpic} &  
        \begin{overpic}[width=0.24\textwidth]{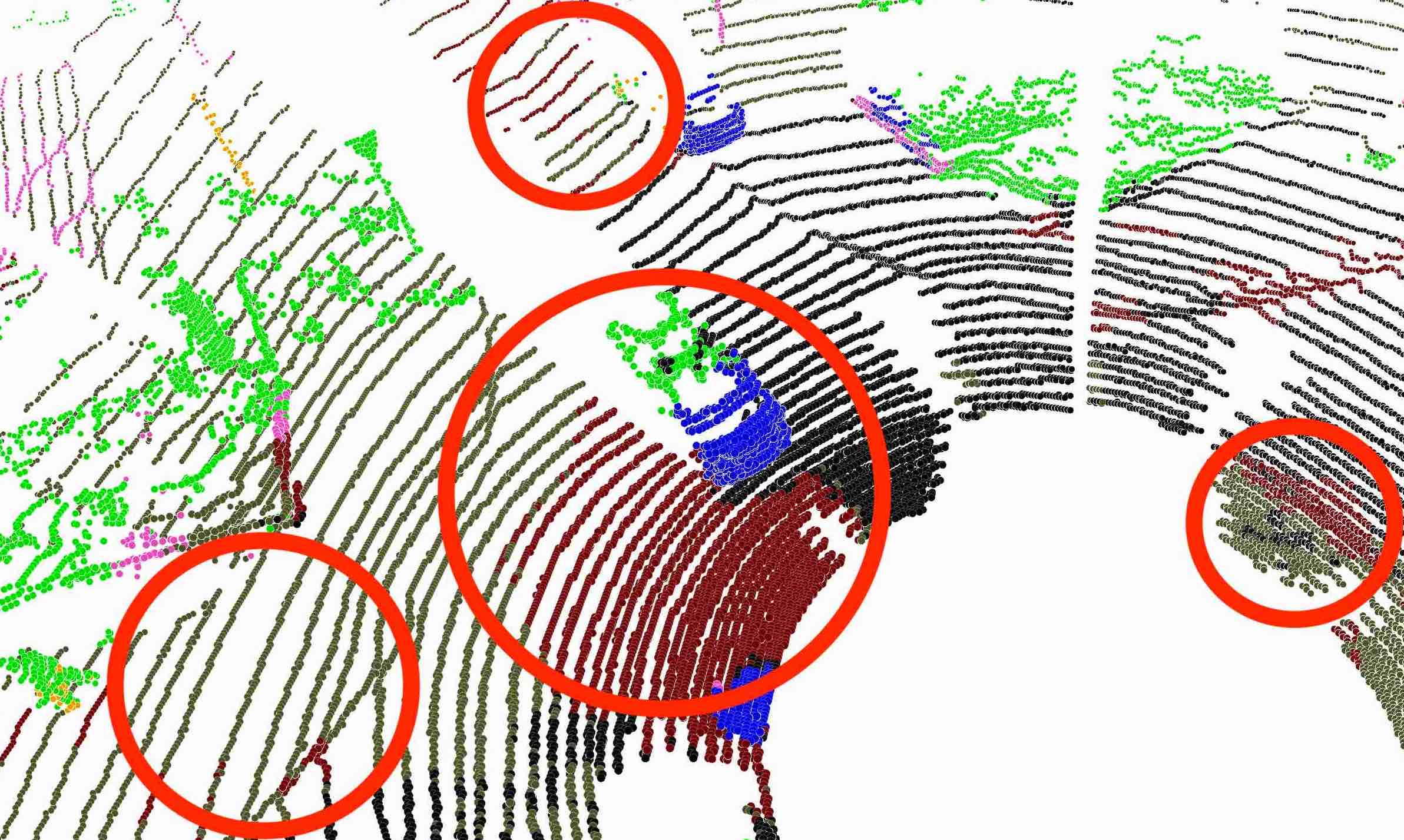}
        \end{overpic} &
        \begin{overpic}[width=0.24\textwidth]{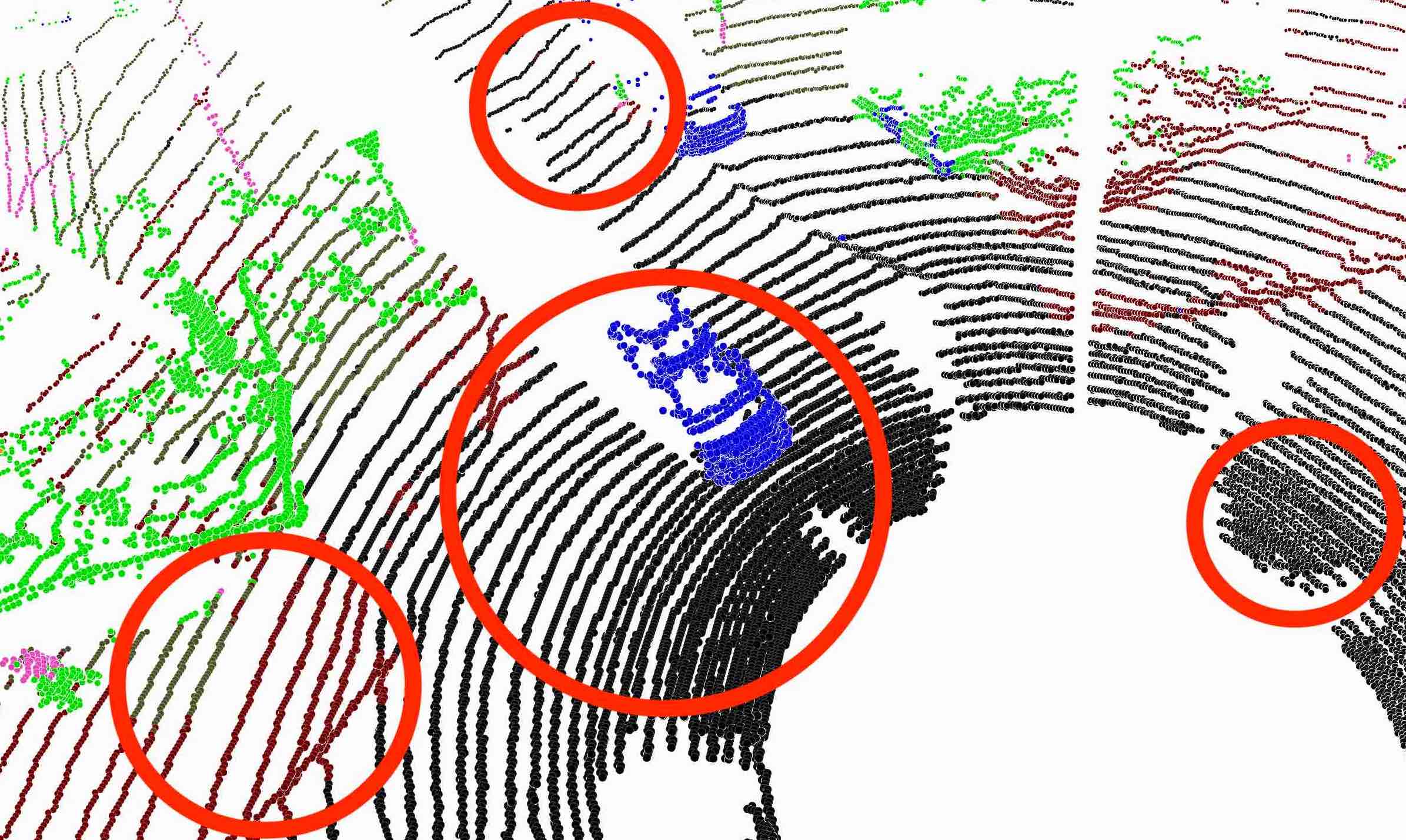}
        \end{overpic}& 
        \begin{overpic}[width=0.24\textwidth]{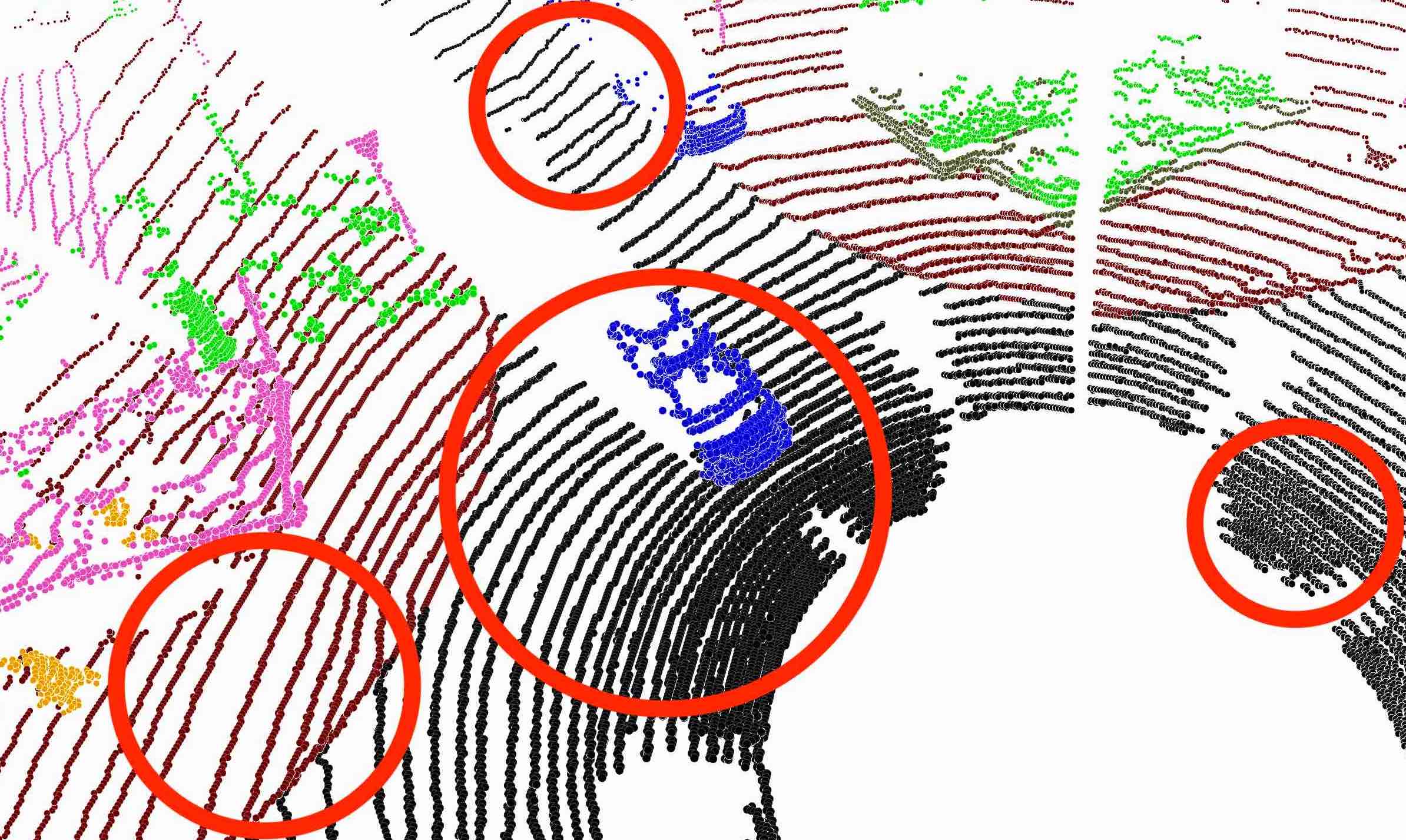}
        \end{overpic}\\
        \begin{overpic}[width=0.24\textwidth]{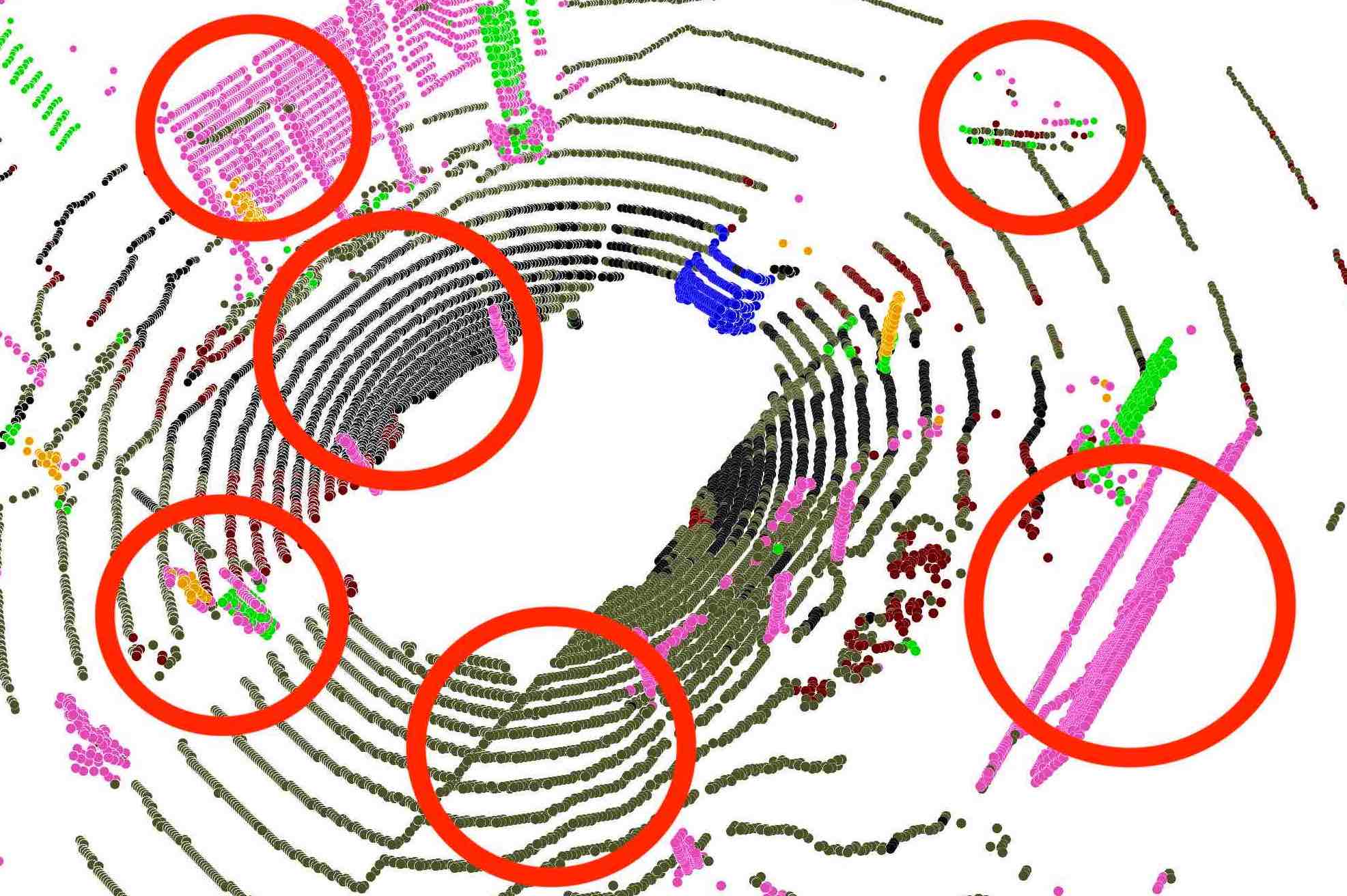}
        \put(40,-5){\color{black}\footnotesize \textbf{source}}
        \put(145,-5){\color{black}\footnotesize \textbf{mix3D}}
        \put(250,-5){\color{black}\footnotesize \textbf{ours}}
        \put(350,-5){\color{black}\footnotesize \textbf{gt}}
        \end{overpic} &  
        \begin{overpic}[width=0.24\textwidth]{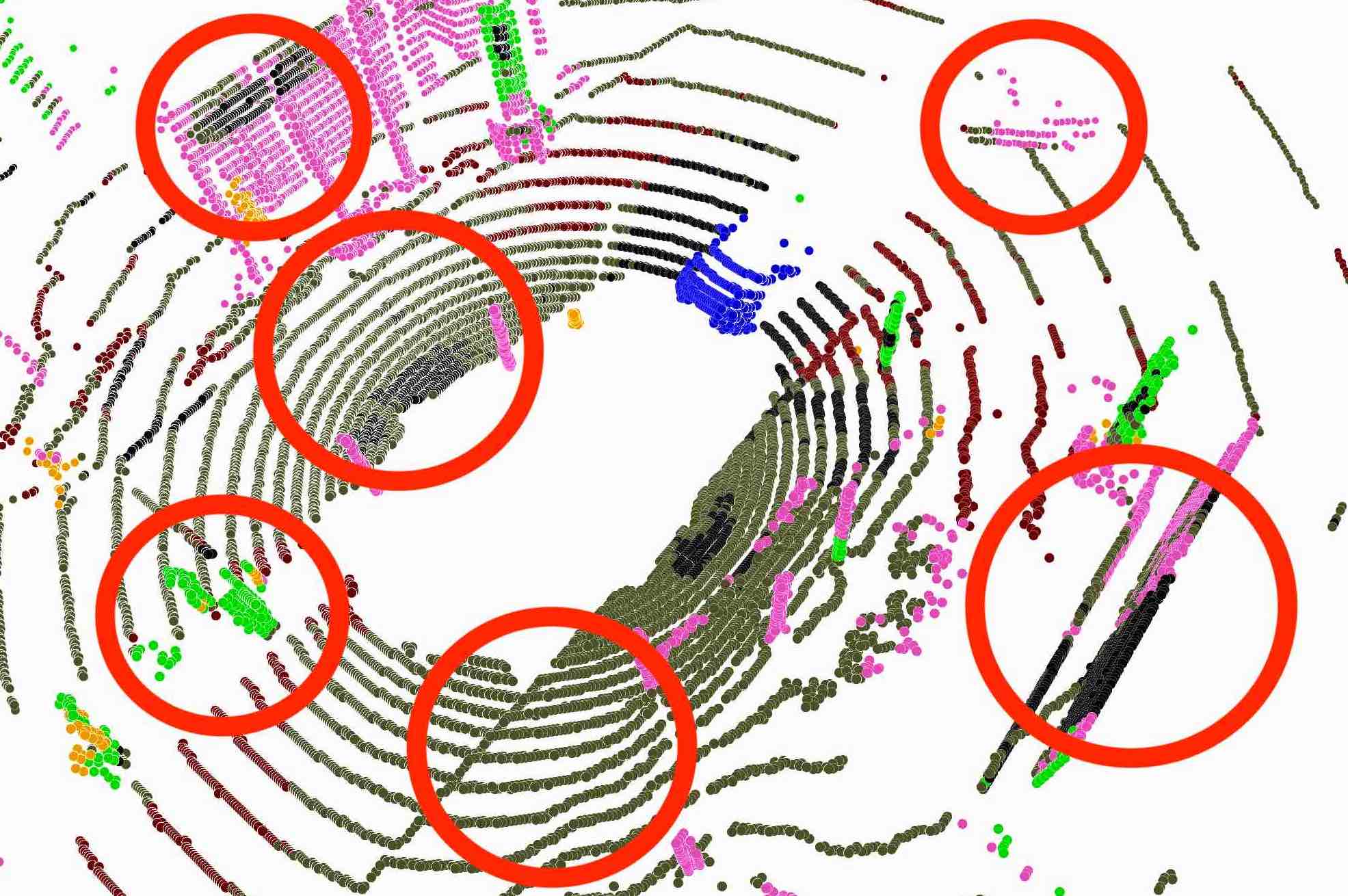}
        \end{overpic} &
        \begin{overpic}[width=0.24\textwidth]{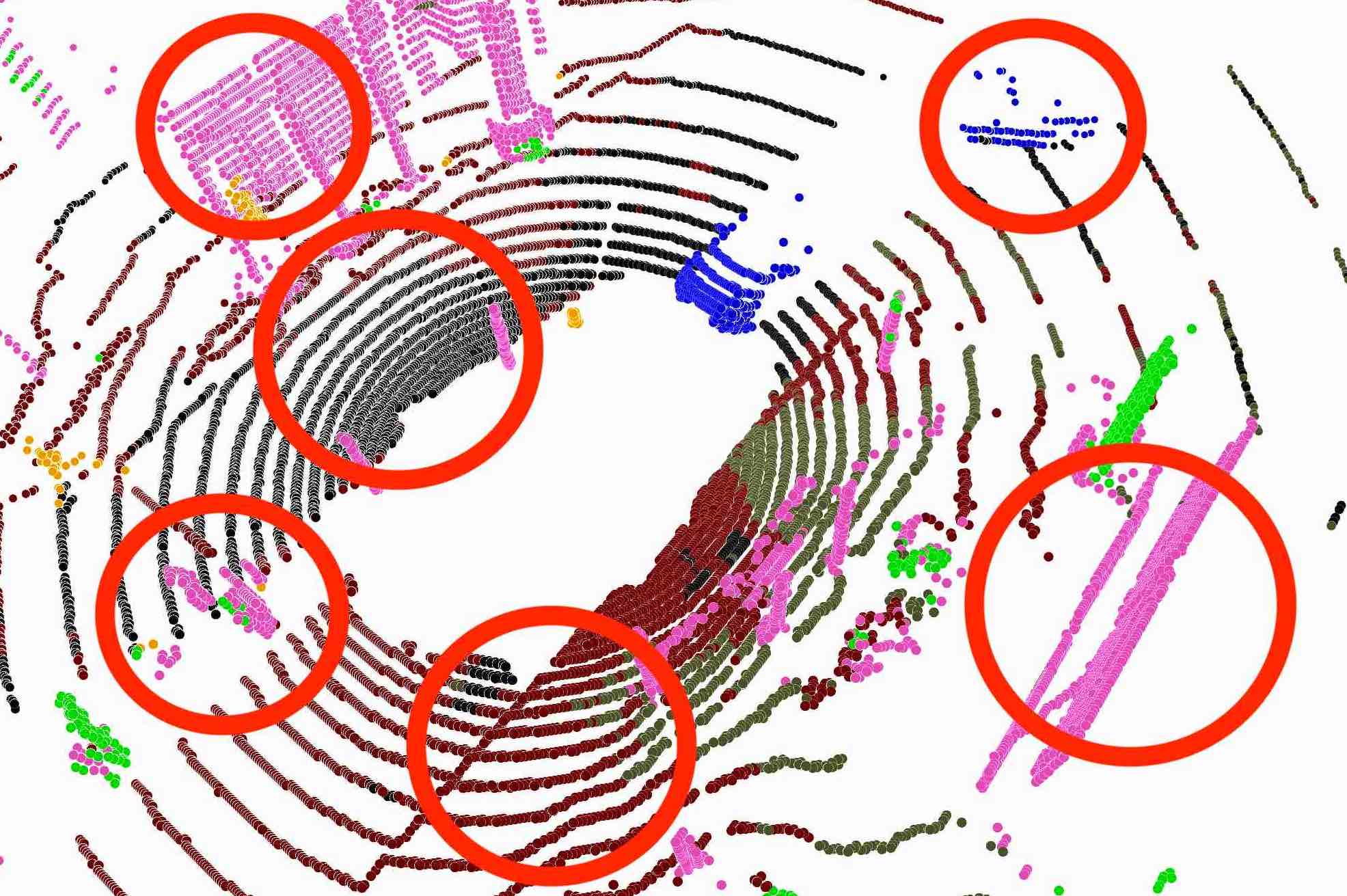}
        \end{overpic}& 
        \begin{overpic}[width=0.24\textwidth]{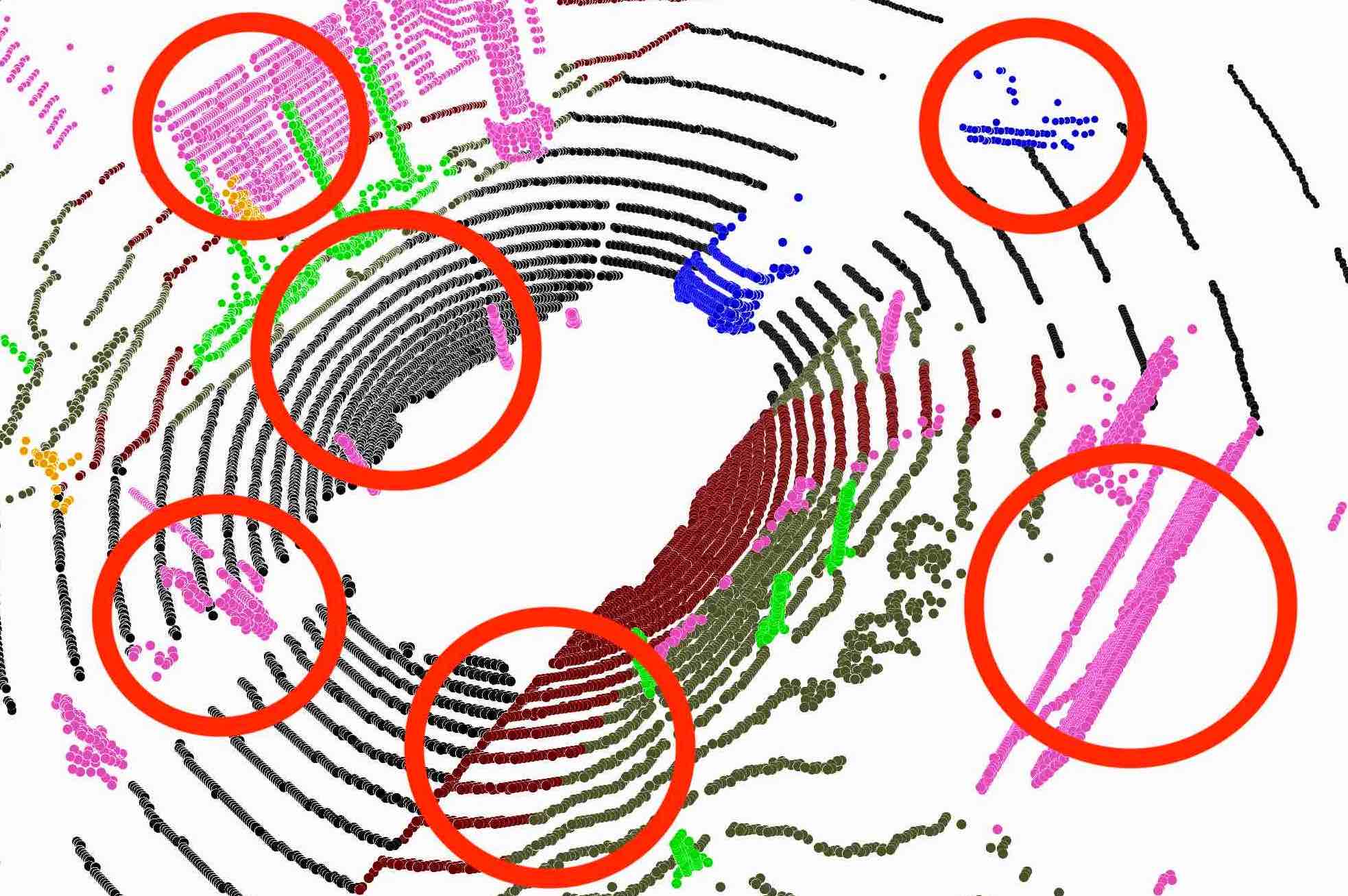}
        \end{overpic}\\
    \end{tabular}
    
    \caption{\textbf{Qualitative results.} \textit{Top:} Synth4D-KITTI$\rightarrow$SemanticKITTI, \textit{bottom:} Synth4D-KITTI$\rightarrow$nuScenes. \lidog improves consistently improves results over the \textit{source} model and outperforms Mix3D~\cite{Nekrasov213DV} with more homogeneous predictions.}
    \label{fig:qualitative}
\end{figure*}

In this section, we train our model on one or more synthetic \source domains and evaluate on both real \target domains, \ie, SemanticKITTI and nuScenes-lidarseg. We report two models as lower and upper bounds in all the tables for reference: \textit{source} (\ie, a model trained on the source domain without the help of DG techniques) and \textit{target} (\ie, model directly trained on the target data).

\begin{table*}[ht]
    \centering
    \caption{\textbf{(Synth4D-nuScenes $+$ Synth4D-KITTI)$\rightarrow$Real, multi-source.} Baselines significantly improve performance relative to the \textit{source} model. Specifically, with \lidog we observe $+10.62$ mIoU improvement on SemanticKITTI and $+14.63$ mIoU on nuScenes. Our approach (\lidog) outperforms all the compared approaches. 
    Lower bound~(\textit{red}): a model trained on the source domain without the help of DG techniques. Upper bound~(\textit{blue}): model directly trained on the target data.}
    \vspace{-10pt}
    \small 
    \setlength{\tabcolsep}{3pt}
\resizebox{1.0\linewidth}{!}{
\begin{tabular}{cc|ccccccccc|c}
\toprule 
S & T & Info & Method & Vehicle & Person & Road & Sidewalk & Terrain & Manmade & Vegetation & mIoU\tabularnewline
\midrule 

\parbox[t]{2mm}{\multirow{10}{*}{\rotatebox{90}{Synth4D-nuScenes+Synth4D-KITTI}}} & & \marksrc Lower bound & \marksrc Source & \marksrc 45.14 & \marksrc 8.39 & \marksrc 41.74 & \marksrc 17.28 & \marksrc 18.44 & \marksrc 33.81 & \marksrc 57.93 & \marksrc 31.82\tabularnewline
\cmidrule{3-12}
 & \parbox[t]{2mm}{\multirow{9}{*}{\rotatebox{90}{SemanticKITTI}}} & \multirow{3}{*}{Aug} & Mix3D~\cite{Nekrasov213DV} & 45.64 & 7.72 & 59.11 & 22.12 & 16.86 & \textbf{50.03} & 51.59 & 36.15 \tabularnewline
 & &  & PointCutMix~\cite{zhang2022pointcutmix} & 48.38 & 6.28 & 60.8 & 20.53 & 7.56 & 33.62 & 54.32 & 33.07 \tabularnewline
 & &  & CoSMix~\cite{saltori2022cosmix} & 45.81 & 0.93 & 59.28 & 22.81 & 18.13 & 43.14 & 40.91 & 33.00 \tabularnewline
\cmidrule{3-12}
 & & \multirow{2}{*}{2D DG} & IBN~\cite{pan2018two} & 58.9 & 12.02 & 67.09 & 27.89 & 7.53 & 33.91 & 57.62 & 37.85 \tabularnewline
 & &  & RobustNet~\cite{choi2021robustnet} & 59.28 & 13.11 & 66.55 & 7.75 & \textbf{30.87} & 35.46 & \textbf{62.75} & 39.40 \tabularnewline
\cmidrule{3-12}
 & & \multirow{2}{*}{3D UDA} & SN~\cite{wang2020train} & 35.07 & 5.95 & 60.36 & \textbf{27.96} & 13.3 & 26.88 & 54.88 & 32.06 \tabularnewline
 & &  & RayCast~\cite{langer2020domain} & 58.06 & 3.3 & 63.41 & 25.33 & 22.03 & 35.39 & 41.64 & 35.59 \tabularnewline
 \cmidrule{3-12}
%

 & & 3D DG & Ours (\lidog) & \textbf{66.23} & \textbf{18.87} & \textbf{67.39} & 24.13 & 15.22 & 46.21 & 59.03 & \textbf{42.44} \tabularnewline
\cmidrule{3-12} 
 & & \marktar Upper bound & \marktar Target & \marktar 81.57 & \marktar 23.23 & \marktar 82.09 & \marktar 57.64 & \marktar 46.84 & \marktar 64.14 & \marktar 75.21 & \marktar 61.53 \tabularnewline
\midrule
\midrule
 \parbox[t]{2mm}{\multirow{10}{*}{\rotatebox{90}{Synth4D-nuScenes+Synth4D-KITTI}}} &  & \marksrc Lower bound & \marksrc Source & \marksrc 20.64 & \marksrc 13.4 & \marksrc 28.44 & \marksrc 7.61 & \marksrc 10.23 & \marksrc 52.88 & \marksrc 46.01 & \marksrc 25.60 \tabularnewline
\cmidrule{3-12}
&\parbox[t]{2mm}{\multirow{10}{*}{\rotatebox{90}{nuScenes}}} & \multirow{3}{*}{Aug} & Mix3D~\cite{Nekrasov213DV} & 28.35 & 24.6 & 57.16 & 13.96 & 9.48 & 60.19 & 55.98 & 35.67 \tabularnewline
 & &  & PointCutMix~\cite{zhang2022pointcutmix} & 25.6 & 14.8 & 54.29 & 11.58 & 6.66 & 58.09 & 51.53 & 31.79 \tabularnewline
 & &  & CoSMix~\cite{saltori2022cosmix} & 23.13 & 8.26 & 52.24 & 11.04 & 10.88 & 58.29 & 53.43 & 31.04 \tabularnewline
\cmidrule{3-12}
 & & \multirow{2}{*}{2D DG} & IBN~\cite{pan2018two} & 26.11 & 16.95 & 55.16 & 13.59 & 12.67 & 56.64 & 52.92 & 33.43 \tabularnewline
 & &  & RobustNet~\cite{choi2021robustnet} & 27.03 & 16.48 & 50.61 & 6.66 & 15.73 & 57.68 & \textbf{56.47} & 32.95 \tabularnewline
\cmidrule{3-12}
 & & \multirow{2}{*}{3D UDA} & SN~\cite{wang2020train} & 19.08 & 13.89 & 53.18 & 14.73 & 9.69 & 53.54 & 45.39 & 29.93 \tabularnewline
 & &  & RayCast~\cite{langer2020domain} & 25.37 & 9.03 & 60.27 & 10.03 & 11.15 & 54.72 & 45.56 & 30.87 \tabularnewline
 \cmidrule{3-12}
%
 & & 3D DG & Ours (\lidog) & \textbf{30.78} & \textbf{25.3} & \textbf{73.2} & \textbf{19.83} & \textbf{16.03} & \textbf{62.65} & 53.80 & \textbf{40.23} \tabularnewline
\cmidrule{3-12}
 & & \marktar Upper bound & \marktar Target & \marktar 47.42 & \marktar 20.18 & \marktar 80.89 & \marktar 37.02 & \marktar 34.99 & \marktar 64.27 & \marktar 54.67 & \marktar 48.49 \tabularnewline
\bottomrule 
\end{tabular}
}
\label{tab:multi-src}
\end{table*}

\begin{table*}[ht]
    \centering    \caption{\textbf{SemanticKITTI$\rightarrow$nuScenes, single-source.} We train our model on SemanticKITTI and evaluate it on the nuScenes dataset. \lidog improves over the \textit{source} model by $+8.35$ mIoU.
    Lower bound~(\textit{red}): a model trained on the source domain with-out the help of DG techniques. Upper bound~(\textit{blue}): model directly trained on the target data.}
    \vspace{-10pt}
    \small 
    \setlength{\tabcolsep}{3pt}
\resizebox{1.0\linewidth}{!}{
\begin{tabular}{cc|ccccccccc|c}
\toprule 
S & T & Info & Method & Vehicle & Person & Road & Sidewalk & Terrain & Manmade & Vegetation & mIoU\tabularnewline
\midrule

&  & \marksrc Lower bound & \marksrc Source & \marksrc 22.92 & \marksrc 0.03 & \marksrc 63.33 & \marksrc 16.09 & \marksrc 7.42 & \marksrc 35.40 & \marksrc 40.53 & \marksrc 26.53 \tabularnewline
\cmidrule{3-12}
\parbox[t]{2mm}{\multirow{9}{*}{\rotatebox{90}{SemanticKITTI}}} & \parbox[t]{2mm}{\multirow{9}{*}{\rotatebox{90}{nuScenes}}}  & \multirow{3}{*}{Aug} & Mix3D~\cite{Nekrasov213DV} & 33.74 & 11.22 & 58.54 & 12.91 & 5.28 & \textbf{50.36} & 48.59 & 31.52 \tabularnewline
&  &  & PointCutMix~\cite{zhang2022pointcutmix} & 22.75 & 2.68 & 59.37 & 10.47 & 7.04 & 27.9 & 42.74 & 24.71\tabularnewline
&  &  & CoSMix~\cite{saltori2022cosmix} & \textbf{35.91} & 0.00 & 58.13 & 11.57 & 8.95 & 45.17 & 49.11 & 29.83 \tabularnewline
\cmidrule{3-12}
&  & \multirow{2}{*}{2D DG} & IBN~\cite{pan2018two} & 29.93 & 0.01 & 56.77 & 18.70 & 12.09 & 37.67 & 33.83 & 27.00 \tabularnewline
&  &  & RobustNet~\cite{choi2021robustnet} & 25.50 & 0.03 & 62.40 & 15.12 & 9.40 & 29.58 & 43.62 & 26.52 \tabularnewline
\cmidrule{3-12}
&  & \multirow{2}{*}{3D UDA} & SN~\cite{wang2020train} & 21.35 & 0.01 & 60.48 & 15.06 & 6.16 & 31.85 & 45.69 & 25.80 \tabularnewline
&  &  & RayCast~\cite{langer2020domain} & 28.82 & 0.00 & 59.25 & 16.08 & 12.51 & 49.72 & 49.82 & 30.89 \tabularnewline
\cmidrule{3-12}
%

&  & 3D DG & Ours (\lidog) & 23.97 & \textbf{14.86} & \textbf{70.63} & \textbf{24.59} & \textbf{13.97} & 45.27 & \textbf{50.85} & \textbf{34.88} \tabularnewline
\cmidrule{3-12}
&  & \marktar Upper bound & \marktar Target & \marktar 47.42 & \marktar 20.18 & \marktar 80.89 & \marktar 37.02 & \marktar 34.99 & \marktar 64.27 & \marktar 54.67 & \marktar 48.49 \tabularnewline
\bottomrule 
\end{tabular}
}
\label{tab:kitti2nusc}
\end{table*}

\begin{table*}[ht]
    \centering
    \caption{\textbf{nuScenes$\rightarrow$SemanticKITTI, single-source.} We train our model on nuScenes and evaluate it on the SemanticKITTI dataset. \lidog improves over the \textit{source} model by $+11.67$ mIoU. 
    Lower bound~(\textit{red}): a model trained on the source domain with-out the help of DG techniques. Upper bound~(\textit{blue}): model directly trained on the target data.}
    \vspace{-10pt}
    \small 
    \setlength{\tabcolsep}{3pt}
\resizebox{1.0\linewidth}{!}{
\begin{tabular}{cc|ccccccccc|c}
\toprule 
S & T & Info & Method & Vehicle & Person & Road & Sidewalk & Terrain & Manmade & Vegetation & mIoU\tabularnewline
\midrule

&  & \marksrc Lower bound & \marksrc Source & \marksrc 27.16 & \marksrc 6.12 & \marksrc 29.19 & \marksrc 8.75 & \marksrc 22.87 & \marksrc 51.27 & \marksrc 61.47 & \marksrc 29.55  \tabularnewline
\cmidrule{3-12}
\parbox[t]{2mm}{\multirow{9}{*}{\rotatebox{90}{nuScenes}}} & \parbox[t]{2mm}{\multirow{9}{*}{\rotatebox{90}{SemanticKITTI}}} & \multirow{3}{*}{Aug} & Mix3D~\cite{Nekrasov213DV} & 37.86 & 6.74 & 41.95 & 5.73 & 27.59 & 41.21 & 65.41 & 32.36  \tabularnewline
& &  & PointCutMix~\cite{zhang2022pointcutmix} & 55.22 & \textbf{13.92} & 45.96 & 5.43 & 30.47 & \textbf{56.50} & \textbf{70.98} & 39.78 \tabularnewline
& &  & CoSMix~\cite{saltori2022cosmix} & 44.58 &	13.88 &	36.10 &	10.19 &	29.32 &	54.43 &	69.08 &	36.80 \tabularnewline
\cmidrule{3-12}
& & \multirow{2}{*}{2D DG} & IBN~\cite{pan2018two} & 22.00 & 11.32 & 37.24 & 0.21 & 13.11 & 21.76 & 50.33 & 22.28 \tabularnewline
& &  & RobustNet~\cite{choi2021robustnet} &32.94 & 10.98 & 39.85 & 14.70 & 28.27 & 50.42 & 58.47 & 33.66\tabularnewline
\cmidrule{3-12}
& & \multirow{2}{*}{3D UDA} & SN~\cite{wang2020train} & 25.69 & 5.46 & 19.59 & 2.17 & 23.47 & 27.65 & 61.07 & 23.58\tabularnewline
& &  & RayCast~\cite{langer2020domain} & 28.30 & 16.09 & 45.80 & 9.44 & 20.56 & 38.56 & 61.83 & 31.51\tabularnewline
\cmidrule{3-12}
%

& & 3D DG & Ours (\lidog) & \textbf{60.07} & 9.03 & \textbf{47.44} & \textbf{16.40} & \textbf{32.58} & 54.21 & 68.82 & \textbf{41.22} \tabularnewline
\cmidrule{3-12}
& & \marktar Upper bound & \marktar Target & \marktar 81.57 & \marktar 23.23 & \marktar 82.09 & \marktar 57.64 & \marktar 46.84 & \marktar 64.14 & \marktar 75.21 & \marktar 61.53  \tabularnewline

\bottomrule 
\end{tabular}
}
%
%
\label{tab:nusc2kitti}
\end{table*}

\PAR{Single-source.} In Tab.~\ref{tab:synth4d-semkitti}, we report the results for single-source \textit{Synth4D-KITTI$\rightarrow$Real}. We report analogous \textit{Synth4D-nuScenes$\rightarrow$Real} results with similar findings in the appendix. 
We observe a $36.84$ mIoU gap (SemanticKITTI) between the \textit{source} ($24.69$ mIoU) and \textit{target} ($61.53$ mIoU) models. Among the baselines, augmentation-based methods are the most effective: PointCutMix and Mix3D achieve $42.14$ mIoU and $30.36$ mIoU on \textit{Synth4D-KITTI}$\rightarrow$\textit{Real}, respectively.
Surprisingly, these approaches outperform 3D DA baselines, that leverage prior information about the target domain, \eg, vehicle dimensions (SN) or target sensor specs (RayCast), confirming the efficacy of data augmentations for DG.  
\lidog significantly reduces the domain gap between \textit{source} and \textit{target} models and outperforms all the compared baselines in all the scenarios. For example, on \textit{Synth4D-KITTI$\rightarrow$Real} (Tab.~\ref{tab:synth4d-semkitti}), we obtain $44.18$ mIoU, a $+19.49$ improvement over the \textit{source} model. 

In Fig.~\ref{fig:qualitative}, we discuss qualitative results, comparing the \textit{source} model, the top-performing baseline, Mix3D, our proposed \lidog, and the ground-truth labels. As we can see, the \textit{source} model predictions are often incorrect and mingled in many small and large regions. Mix3D brings (limited) improvements. For example, notice how \textit{road} (Fig.~\ref{fig:qualitative}~\textit{top}) is incorrectly segmented. 
\lidog achieves overall top performance, improving on both small (\textit{vehicle}) and large (\textit{building} and \textit{road}) areas. We report additional qualitative results in the appendix. 

\PAR{Multi-source.} In Tab.~\ref{tab:multi-src}, we report the results for the multi-source training, where we train our model on both synthetic datasets, and evaluate the model on both real datasets. Compared to the single source (Tab.~\ref{tab:synth4d-semkitti}), in this setting we train models on synthetic data that mimics two different sensor types. The \textit{source} model improves from $24.69\to 31.82$ mIoU on SemanticKITTI and from $20.62\to25.60$ mIoU on nuScenes as a result of training on multiple synthetic datasets. 
As can be seen, \lidog consistently improves over top-scoring baselines, RobustNet ($+3.04$ mIoU) and Mix3D ($+4.56$ mIoU). 
On average, \lidog significantly improves over the \textit{source} model with $+10.62$ mIoU and $+14.63$ mIoU on SemanticKITTI and nuScenes, respectively.
We conclude that \lidog is consistently a top-performer for classes and scenarios over the \textit{source} model, except for the \textit{terrain} class. This may be a limitation of our BEV projection. We assume it occurs when multiple classes are spatially overlapping due to the top-down projection, \eg, mixing \textit{terrain} with \textit{vegetation}.

\subsection{Real $\to$ Real Evaluation} 
\label{sec:real2real}


In this setting, we only train and evaluate our models on real-world recordings. We report \textit{SemanticKITTI$\rightarrow$nuScenes} results in Tab.~\ref{tab:kitti2nusc} and \textit{nuScenes$\rightarrow$SemanticKITTI} results in Tab.~\ref{tab:nusc2kitti}. Overall, we observe that the performance gap of \textit{Real$\rightarrow$Real} (Tab.~\ref{tab:kitti2nusc}-\ref{tab:nusc2kitti}) is lower compared to the gap in \textit{Synth$\rightarrow$Real} setting (Tab.~\ref{tab:synth4d-semkitti}-\ref{tab:multi-src} and Tab. \textcolor{red}{1} in the appendix).
%
\PAR{SemanticKITTI$\rightarrow$nuScenes.} We observe a performance gap of $21.96$ mIoU between \textit{source} ($26.53$ mIoU) and \textit{target} ($48.49$ mIoU) models. All the baselines consistently improve over the \textit{source} model, however, by a smaller margin as compared to \textit{Synth$\rightarrow$Real}. 
Again, \lidog improves on all the classes compared to the \textit{source} performance and achieves top generalization performance on all the classes except for \textit{vehicle} and \textit{manmade}. 
On average, \lidog obtains $34.88$ mIoU and obtains $+8.35$ mIoU improvement over the \textit{source} model. 
\PAR{nuScenes$\rightarrow$SemanticKITTI.} We observe a performance gap of $31.98$ mIoU with \textit{source} and \textit{target} models achieving $29.55$ mIoU and $61.53$ mIoU, respectively.
Augmentation-based methods are once again the most effective baselines, with PointCutMix achieving $39.78$ mIoU.
\lidog outperforms all the compared baselines, achieves $41.22$ mIoU and obtains $+11.67$ over the \textit{source} model.

 \begin{figure}[t]
\centering
    \setlength\tabcolsep{1.pt}
    \begin{tabular}{cccc}
    \raggedright
        \begin{overpic}[width=0.5\columnwidth]{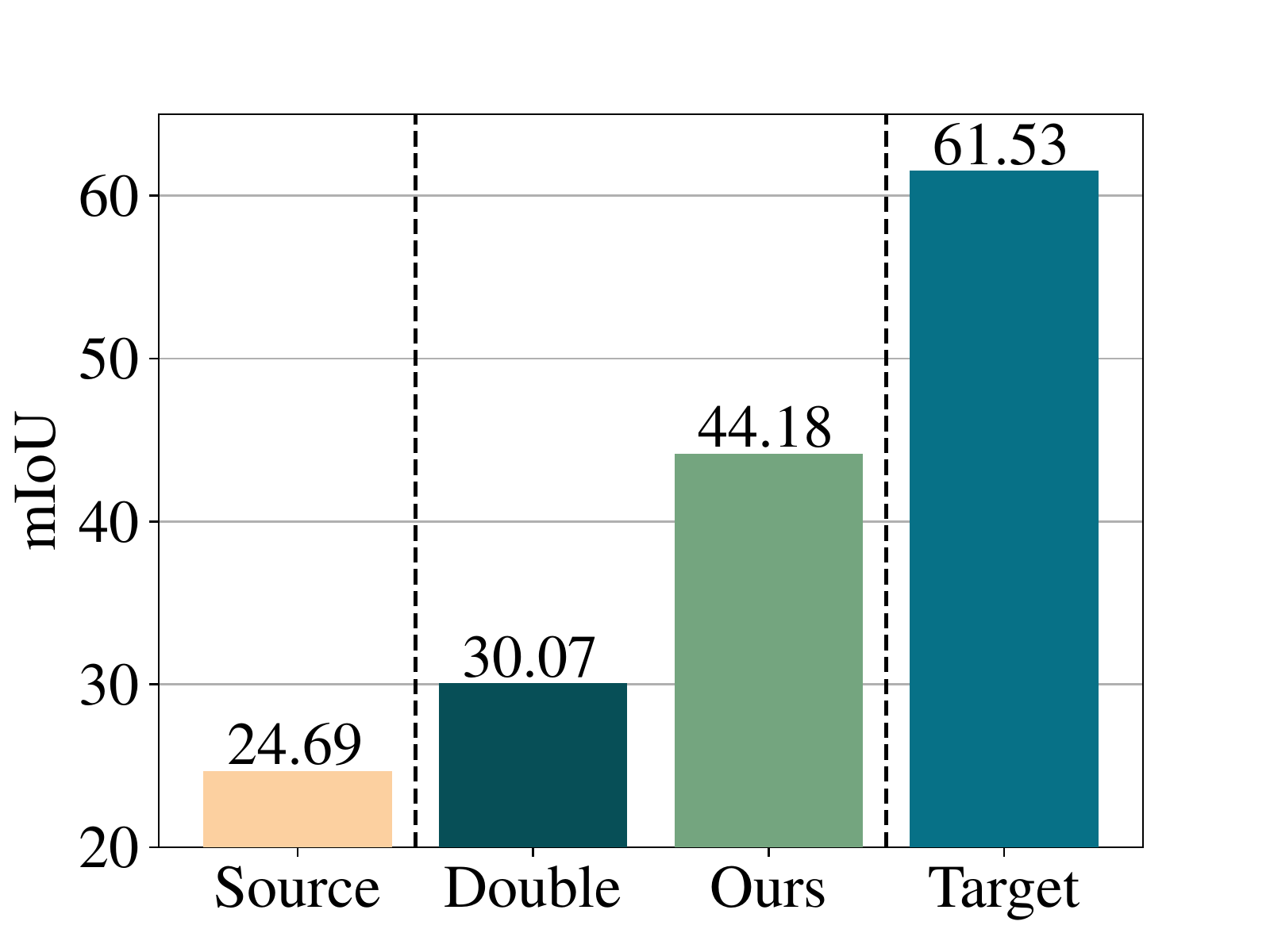}
        \end{overpic} &  
        \begin{overpic}[width=0.5\columnwidth]{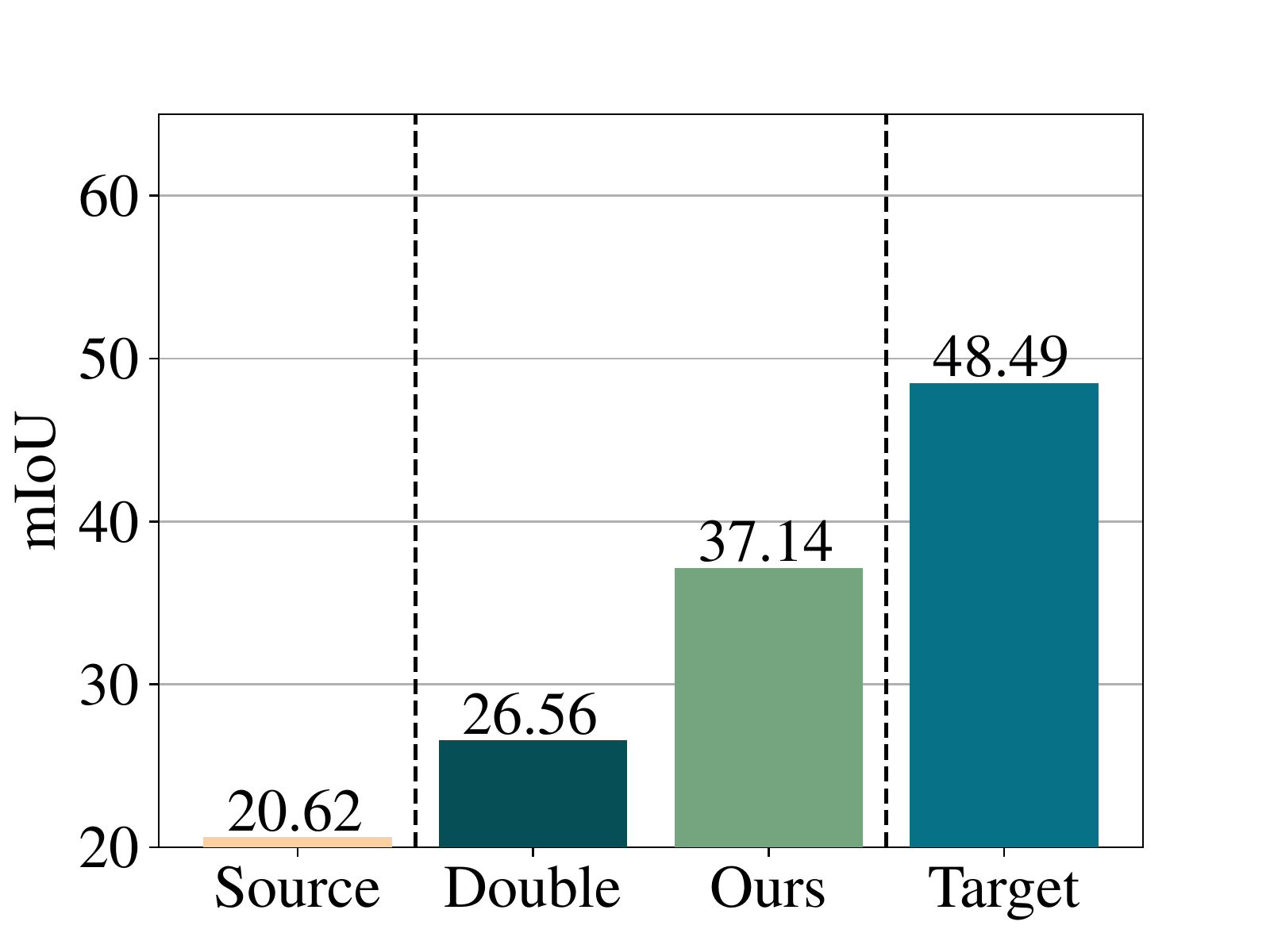}
        \end{overpic}
    \end{tabular}
    \vspace{-10pt}
    \caption{\textbf{Effectivenes of the BEV head:} We compare 2D BEV decoder (\textit{Ours}) to simply adding an additional 3D segmentation head (\textit{Double}) on SemanticKITTI (\textit{left}) and nuScenes (\textit{right}). Source: $Synth4D-KITTI$.}
    \label{fig:ablation_design}

\end{figure}

\subsection{Ablations}
\label{sec:ablations}
%
\PAR{BEV auxiliary task.} 
We study whether \lidog generalization capabilities are due to the additional segmentation head, or specifically due to the dense BEV segmentation head. The use of multiple prediction heads for ensembling is a well-known practice for improving prediction robustness~\cite{caron2020unsupervised, fini2021unified}. To this purpose, we remove the lower branch of \lidog and attach an additional 3D branch. The resulting architecture has two sparse 3D heads $h^{3D}$ taking as input $F^{3D}$.
During inference, predictions are obtained by averaging the predictions from both heads. Fig.~\ref{fig:ablation_design} reports the generalization performance of this alternate version (\textit{Double}) compared to \lidog (\textit{Ours}). As can be seen, by simply learning two decoders and ensembling predictions we obtain lower performance as compared to utilizing the proposed BEV network heads to improve the generalization. 

\begin{figure}[t]
    \centering
 \includegraphics[width=0.99\columnwidth]{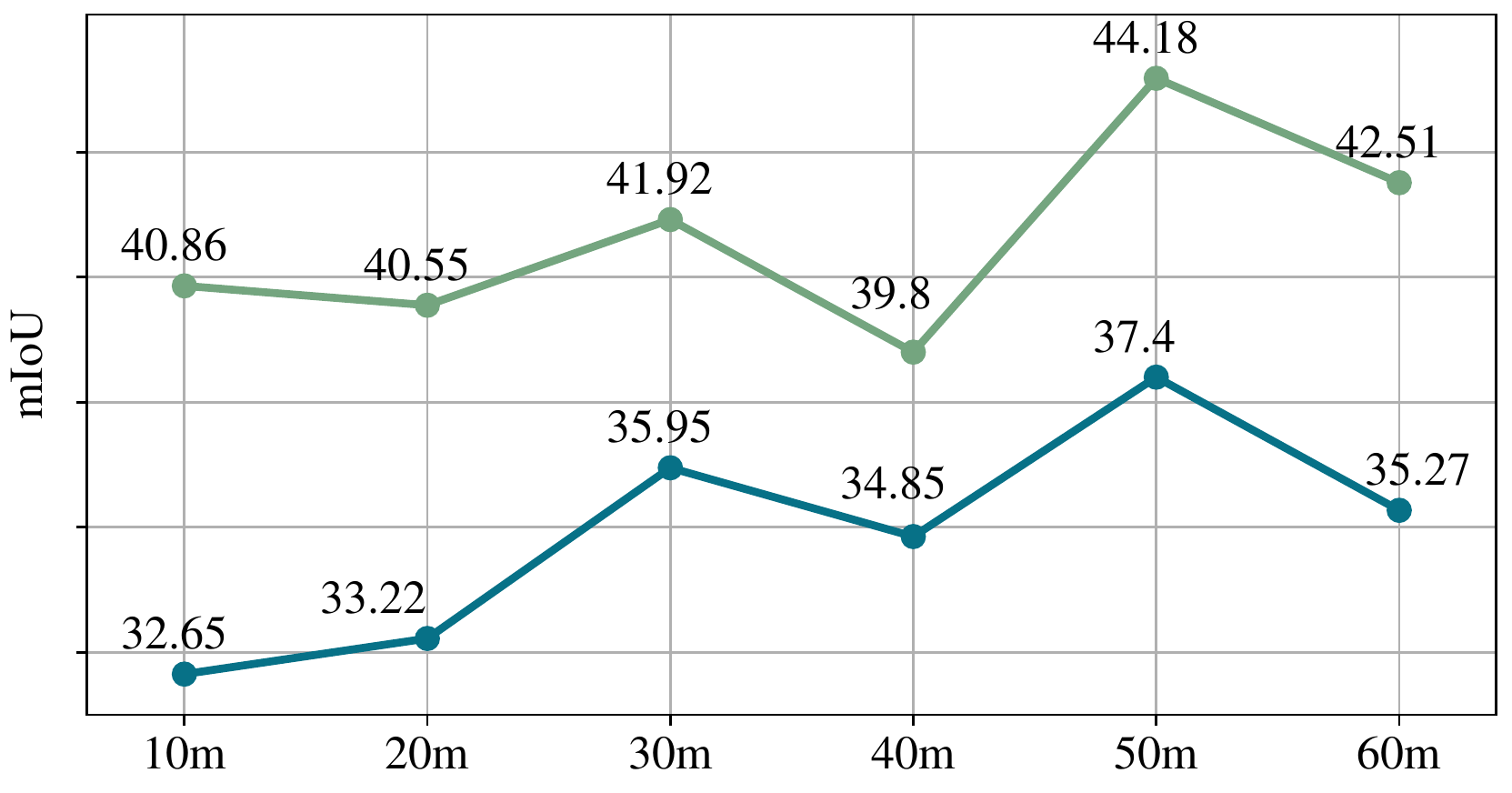}
    \vspace{-10pt}
    \caption{\textbf{BEV prediction area:} We study the impact of BEV area size on SemanticKITTI (\textit{top}) and nuScenes (\textit{bottom}). $50x50 m$ area is consistently the best-performing option on both datasets. Source: $Synth4D-KITTI$.}
    \label{fig:ablation_area}
    \vspace{-1mm}
\end{figure}

\PAR{BEV prediction area.}
First, we study the impact on the generalization ability of \lidog \wrt BEV area size in Synth4D-KITTI$\rightarrow$Real setting. We experiment with projection bounds $B^{3D}$ ranging from $10x10$ to $60x60 m$. As can be seen in Fig.~\ref{fig:ablation_area}, there is a near-linear relation between the BEV area and generalization performance. 

\section{Conclusions}
\vspace{-7pt}
\PAR{Impact.} With the advent of autonomous agents, there will be an increasing need for robust perception models that can generalize across different environments and sensory configurations.
Robustness to domain shift in semantic LiDAR scene understanding is therefore an important challenge that needs to be addressed. 
Fully supervised fine-tuning can be used to bridge domain gaps, however, it comes at the high cost of collecting and annotating data for specific domains and is therefore not a scalable solution.
\PAR{Domain generalization.} To the best of our knowledge, this is the first work that investigates domain generalization techniques to learn robust and generalizable representations for LiDAR semantic segmentation. We evaluate and discuss a comprehensive set of baselines from related fields and study their generalization across synthetic and real datasets, proving the need for more effective solutions. 
To this end, we propose LiDOG that encourages a 3D semantic backbone to learn features robust to domain shifts and is consistently more accurate than the compared baselines. 
\PAR{Limitations.} The main limitation of LiDOG is that its core contribution, the auxiliary bird's-eye prediction task, may lead to overlap after the projection to the BEV plane when classes overlap vertically, \eg, the terrain below vegetation. This results in sub-optimal performance for these classes. This issue could be mitigated with the introduction of soft labels for overlapping semantic classes, or introducing additional losses on multiple BEV height slices (\cf, ~\cite{yang18cvprPixor}).

\footnotesize{\PAR{Acknowledgments.} This project was partially funded by the Sofja Kovalevskaja Award of the Humboldt Foundation, the EU ISFP project PRECRISIS (ISFP-2022-TFI-AG-PROTECT-02-101100539), the PRIN project LEGO-AI (Prot. 2020TA3K9N) and the MUR PNRR project FAIR - Future AI Research (PE00000013) funded by the NextGenerationEU. It was carried out in the Vision and Learning joint laboratory of FBK-UNITN and used the CINECA, NVIDIA-AI TC clusters to run part of the experiments. We thank T. Meinhardt and I. Elezi for their feedback on this manuscript. The authors of this work take full responsibility for its content.}

{\small
\bibliographystyle{ieee_fullname}
\bibliography{refs}
}

\clearpage

\setcounter{page}{1}

\twocolumn[
\centering
\Large
\textbf{Walking Your LiDOG: A Journey Through Multiple Domains for LiDAR Semantic Segmentation} \\
\vspace{0.5em}Supplementary Material\\
\vspace{1.0em}
] 
\appendix

\begin{abstract}
We provide supplementary material in support of the main paper.
We organize the content as follows:
\begin{itemize}
    \item In Sec.~\ref{sec:supp_experiments}, we report the implementation details of our experiments, the additional experimental evaluations on Synth4D-nuScenes$\to$Real, and an additional ablation of \lidog.
    \item In Sec.~\ref{sec:supp_qualitative}, we report and discuss qualitative results comparing the performance of \lidog with all the compared baselines and ground truth annotations.
\end{itemize}

\end{abstract}

\section{Experimental Evaluation}\label{sec:supp_experiments}
We provide the implementation details of our experiments in Sec.~\ref{sec:implementation}.
In Sec.~\ref{sec:supp_synthnusc}, we complete the set of experiments of the main paper and report the experimental results obtained on Synth4D-nuScenes$\to$Real.
In Sec.~\ref{sec:supp_resolution}, we further ablate how the input resolution affects \lidog performance.

\subsection{Implementation details}
\label{sec:implementation}
We implement our method and all the baselines by using the PyTorch framework. We use MinkowskiNet~\cite{choy20194d} as a sparse convolutional backbone in all our experiments and train until convergence with voxel size $0.05$m, total batch size of $16$, learning rate $0.01$ and ADAM optimizer~\cite{Kingma15ICLR}. In the experiments we use random rotation, scaling, and downsampling for better convergence of baselines and \lidog. We set rotation bounds between $[-\pi/2, \pi/2]$, scaling between $[0.95, 1.05]$, and perform random downsampling for 80\% of the patch points.
We set projection bounds $B^{3D}$ based on the input resolution. We set $b_x$ and $b_z$ to $50m$ in the denser source domains of Synth4D-KITTI and SemanticKITTI and $b_x$ and $b_z$ to $30m$ in the sparser domains of Synth4D-nuScenes and nuScenes. Quantization parameters $x_q$ and $z_q$ are computed based on the spatial bounds in order to obtain BEV labels of $168x168$ pixels. 
For downsampling dense features, we use a max pooling layer with window size $5$, stride $3$, and padding $1$. 
The \lidog 2D decoder is implemented with a series of three 2D convolutional layers interleaved by batch normalization layers. ReLU activation function is used on all the layers while softmax activation is applied on the last layer.

\subsection{Synth4D-nuScenes$\to$Real}
\label{sec:supp_synthnusc}

In Tab.~\ref{tab:synth4d-nuscenes}, we report the results for single-source \textit{Synth4D-nuScenes$\rightarrow$Real}.
We observe a $41.82$ mIoU gap (SemanticKITTI) between the \textit{source} ($19.71$ mIoU) and \textit{target} ($61.53$ mIoU) models on Synth4D-nuScenes$\to$SemanticKITTI and a $23.92$ mIoU gap (nuScenes), between \textit{source} ($24.57$ mIoU) and \textit{target} ($48.49$ mIoU) on Synth4D-nuScenes$\to$nuScenes. Among the baselines, augmentation-based methods are the most effective with Mix3D achieving $31.64$ mIoU (SemanticKITTI) and $31.23$ mIoU (nuScenes). 
\lidog reduces the domain gap between \textit{source} and \textit{target} models and outperforms all the compared baselines in all the scenarios. For example, on \textit{Synth4D-nuScenes$\rightarrow$Real} (Tab.~\ref{tab:synth4d-nuscenes}), we obtain $34.79$ mIoU, a $+15.08$ improvement over the \textit{source} model. 
\begin{table*}[t]
    \centering
    \caption{\textbf{Synth4D-nuScenes$\rightarrow$Real, single-source.} We train our model on Synth4D-nuScenes and test on SemanticKITTI and nuScenes. \lidog improves over the \textit{source} models by $+15.08$ mIoU on SemanticKITTI and by $+9.21$ mIoU on nuScenes. \lidog outperforms all the compared baselines. Lower bound (\textit{red}): a model trained n the source domain without the help of DG techniques. Upper bound (\textit{blue}): a model directly trained on target data.}
    \small 
    \setlength{\tabcolsep}{3pt}
    
\resizebox{1.0\linewidth}{!}{
\begin{tabular}{cc|ccccccccc|c}
\toprule 
S & T & Info & Method & Vehicle & Person & Road & Sidewalk & Terrain & Manmade & Vegetation & mIoU\tabularnewline
\hline 
\hline 

\parbox[t]{2mm}{\multirow{10}{*}{\rotatebox{90}{Synth4D-nuScenes}}} & \parbox[t]{2mm}{\multirow{10}{*}{\rotatebox{90}{SemanticKITTI}}}  & \marksrc Lower bound & \marksrc Source & \marksrc 14.54 & \marksrc 2.41 & \marksrc 32.78 & \marksrc 14.86 & \marksrc 6.4 & \marksrc 30.89 & \marksrc 36.07 & \marksrc 19.71\tabularnewline
\cline{3-12}
& & \multirow{3}{*}{Aug} & Mix3D~\cite{Nekrasov213DV} & 37.38 & 7.26 & 56.90 & 21.06 & 10.60 & 34.46 & 53.79 & 31.64\tabularnewline
& &  & PointCutMix~\cite{zhang2022pointcutmix} & 27.51 & 4.32 & 56.04 & 21.37 & 7.02 & 24.38 & 45.35 & 26.57\tabularnewline
& &  & CoSMix~\cite{saltori2022cosmix} & 16.13 & 6.42 & 39.71 & 14.63 & 13.04 & 23.54 & 30.57 & 20.58\tabularnewline
\cline{3-12}
& & \multirow{2}{*}{2D DG} & IBN~\cite{pan2018two} & 47.15 & 7.81 & 50.74 & 4.51 & 15.20 & 29.53 & 51.35 & 29.47\tabularnewline
& &  & RobustNet~\cite{choi2021robustnet} & 21.19 & 8.56 & 44.46 & 10.80 & 15.06 & 11.95 & 30.59 & 20.37\tabularnewline
\cline{3-12}
& & \multirow{2}{*}{3D UDA} & SN~\cite{wang2020train} & 11.56 & 2.38 & 37.50 & 10.86 & 5.19 & 20.70 & 39.23 & 18.20\tabularnewline
& &  & RayCast~\cite{langer2020domain} & 28,89 & 6.34 & 53.59 & 12.94 & \textbf{15.86} & 21.74 & 41.85 & 25.89\tabularnewline
\cline{3-12}
%

& & 3D DG & Ours (\lidog) & \textbf{55.08} & \textbf{11.42} & \textbf{59.48} & \textbf{26.10} & 2.78 & \textbf{34.83} & \textbf{53.82} & \textbf{34.79}\tabularnewline
\cline{3-12}
& & \marktar Upper bound & \marktar Target & \marktar 81.57 & \marktar 23.23 & \marktar 82.09 & \marktar 57.64 & \marktar 46.84 & \marktar 64.14 & \marktar 75.21 & \marktar 61.53\tabularnewline
 \midrule
\midrule
\parbox[t]{2mm}{\multirow{10}{*}{\rotatebox{90}{Synth4D-nuScenes}}} & \parbox[t]{2mm}{\multirow{10}{*}{\rotatebox{90}{nuScenes}}} & \marksrc Lower bound & \marksrc Source & \marksrc 17.73 & \marksrc 8.94 & \marksrc 36.53 & \marksrc 7.28 & \marksrc 5.78 & \marksrc 52.13 & \marksrc 43.62 & \marksrc 24.57\tabularnewline
\cline{3-12}
& & \multirow{3}{*}{Aug} & Mix3D~\cite{Nekrasov213DV} & \textbf{33.83} & 17.85 & 47.74 & 8.93 & 9.71 & 56.35 & 44.18 & 31.23\tabularnewline
& &  & PointCutMix~\cite{zhang2022pointcutmix} & 21.51 & 13.12 & 53.72 & 8.86 & 8.45 & 54.83 & 48.81 & 29.90\tabularnewline
& &  & CoSMix~\cite{saltori2022cosmix} & 22.00 & 15.43 & 57.40 & 8.86 & 9.08 & 56.24 & 47.16 & 30.88\tabularnewline
\cline{3-12}
& & \multirow{2}{*}{2D DG} & IBN~\cite{pan2018two} & 23.07 & 14.13 & 44.96 & 7.10 & 9.83 & 53.70 & \textbf{49.49} & 28.90\tabularnewline
& &  & RobustNet~\cite{choi2021robustnet} & 21.59 & 12.29 & 48.52 & 8.14 & 6.22 & 51.33 & 47.68 & 27.97\tabularnewline
\cline{3-12}
& & \multirow{2}{*}{3D UDA} & SN~\cite{wang2020train} & 24.71 & 8.45 & 5.,08 & 5.66 & \textbf{11.04} & 47.00 & 39.05 & 26.57\tabularnewline
& &  & RayCast~\cite{langer2020domain} & 19.65 & 12.24 & 58.08 & 7.58 & 9.71 & 46.43 & 41.37 & 27.86\tabularnewline
\cline{3-12}
%
& & 3D DG & Ours (\lidog) & 26.79 & \textbf{18.68} & \textbf{63.28} & \textbf{15.81} & 6.57 & \textbf{58.8} & 46.5 & \textbf{33.78}\tabularnewline
\cline{3-12}
& & \marktar Upper bound & \marktar Target & \marktar 47.42 & \marktar 20.18 & \marktar 80.89 & \marktar 37.02 & \marktar 34.99 & \marktar 64.27 & \marktar 54.67 & \marktar 48.49\tabularnewline
\bottomrule 
\end{tabular}
}
\label{tab:synth4d-nuscenes}
\end{table*}



\subsection{BEV resolution}
\label{sec:supp_resolution}
 \begin{figure}[t]
\centering
    \setlength\tabcolsep{1.pt}
    \begin{tabular}{cccc}
    \raggedright
        \begin{overpic}[width=0.5\columnwidth]{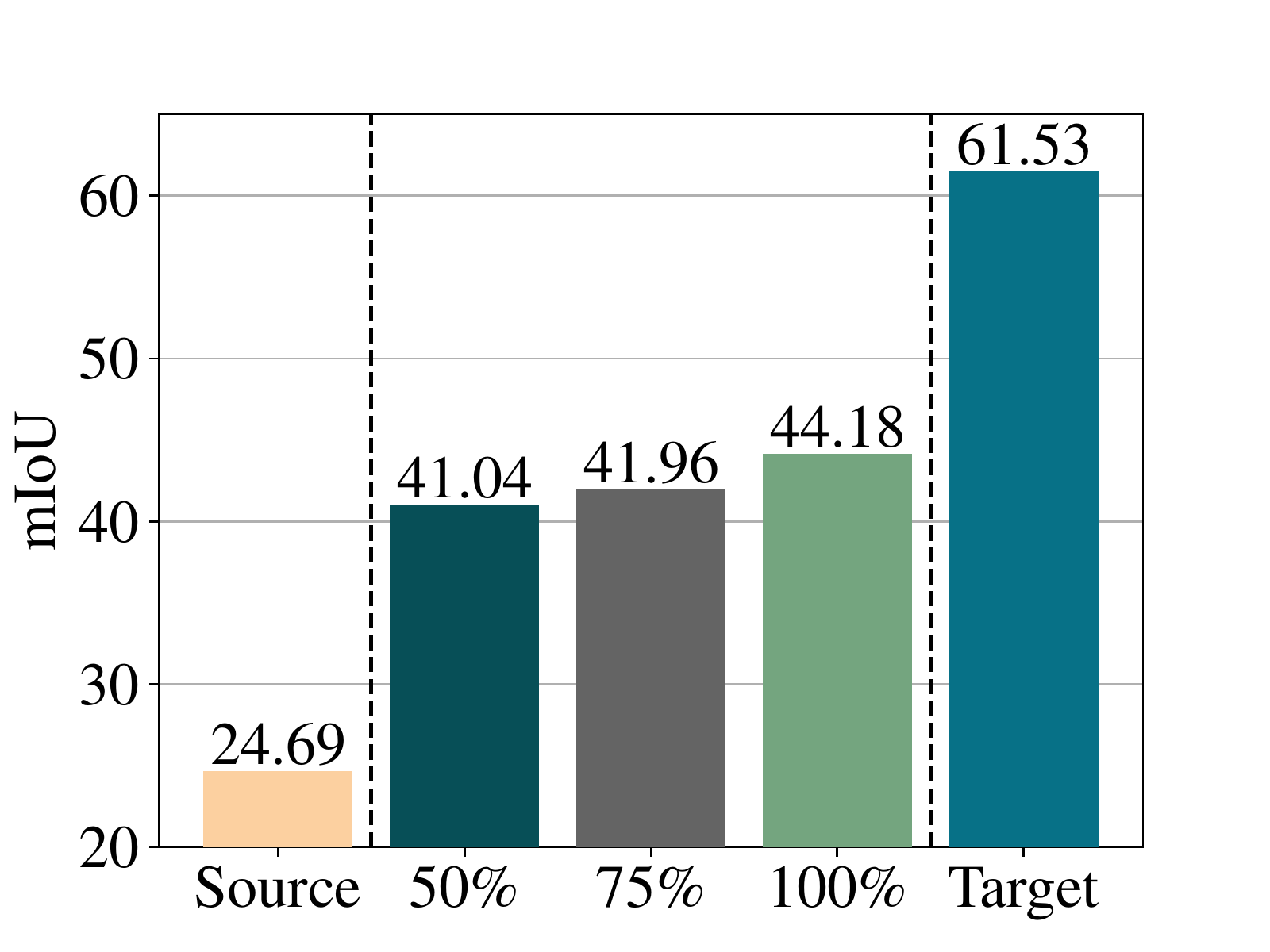}
        \end{overpic} &  
        \begin{overpic}[width=0.5\columnwidth]{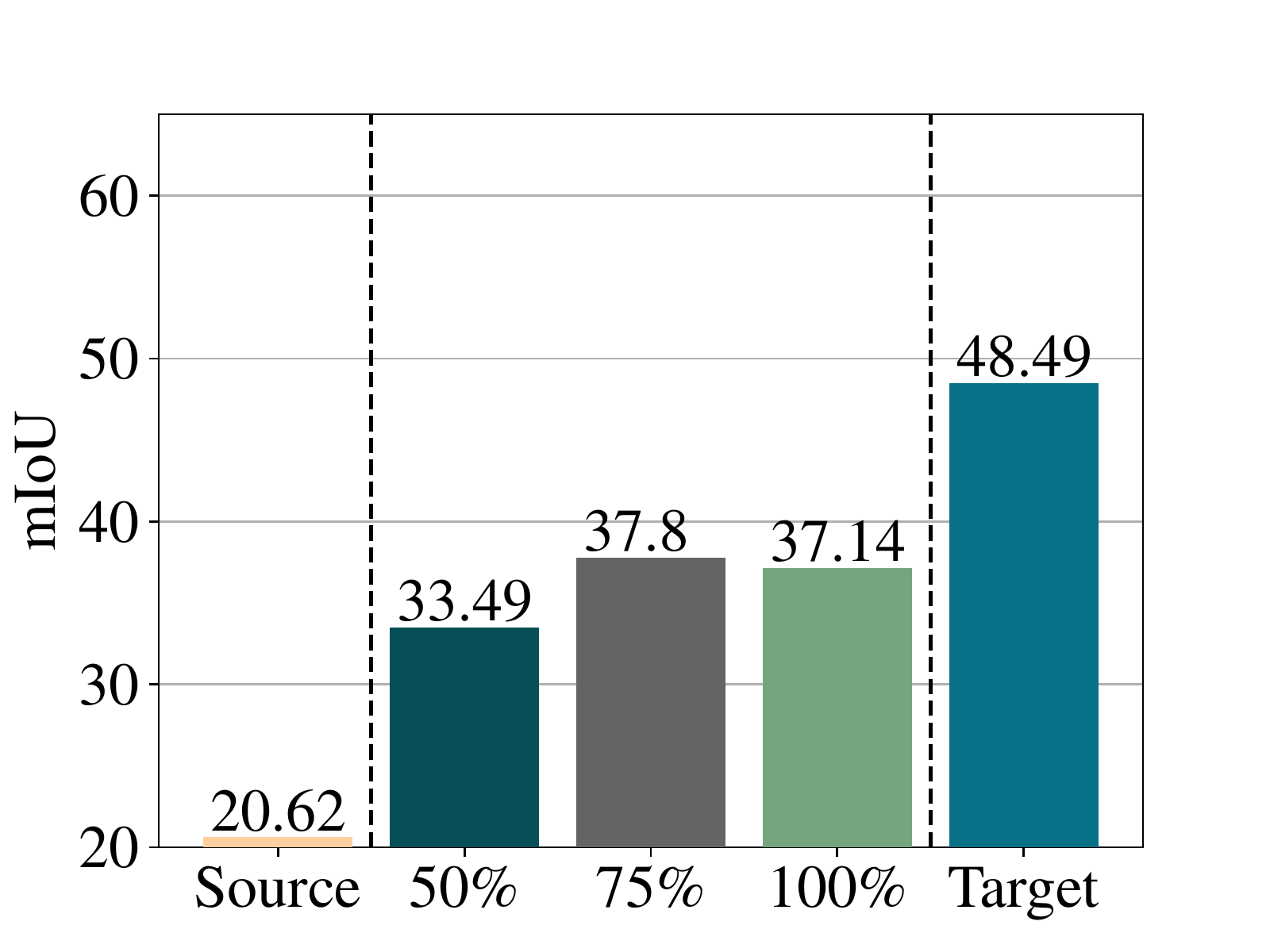}
        \end{overpic}
    \end{tabular}
    \vspace{-10pt}
    \caption{\textbf{BEV image resolution:} We compare the performance while changing the BEV image resolution on SemanticKITTI (\textit{left}) and nuScenes (\textit{right}), Source: \textit{Synth4D-KITTI}}
    \label{fig:ablation_resolution}
\end{figure}

We study the impact of BEV image resolution on the \lidog performance.
We adapt the feature resolution by applying several pooling steps and the label resolution via the quantization step size. 
In Fig.~\ref{fig:ablation_resolution}, we report the results obtained on Synth4D-KITTI$\rightarrow$Real, when using BEV features with $50\%$ and $75\%$ of the initial resolution ($100\%$).  Interestingly, on Synth4D-KITTI$\rightarrow$nuScenes we observe a slight improvement with $75\%$ resolution. However, we obtain consistently top performance on both domains with the full resolution ($100\%$).

\section{Qualitative evaluation}
\label{sec:supp_qualitative}

\begin{figure*}[t]
\centering
    \setlength\tabcolsep{1.pt}
    \begin{tabular}{cccc}
    \raggedright
        \begin{overpic}[width=0.21\textwidth]{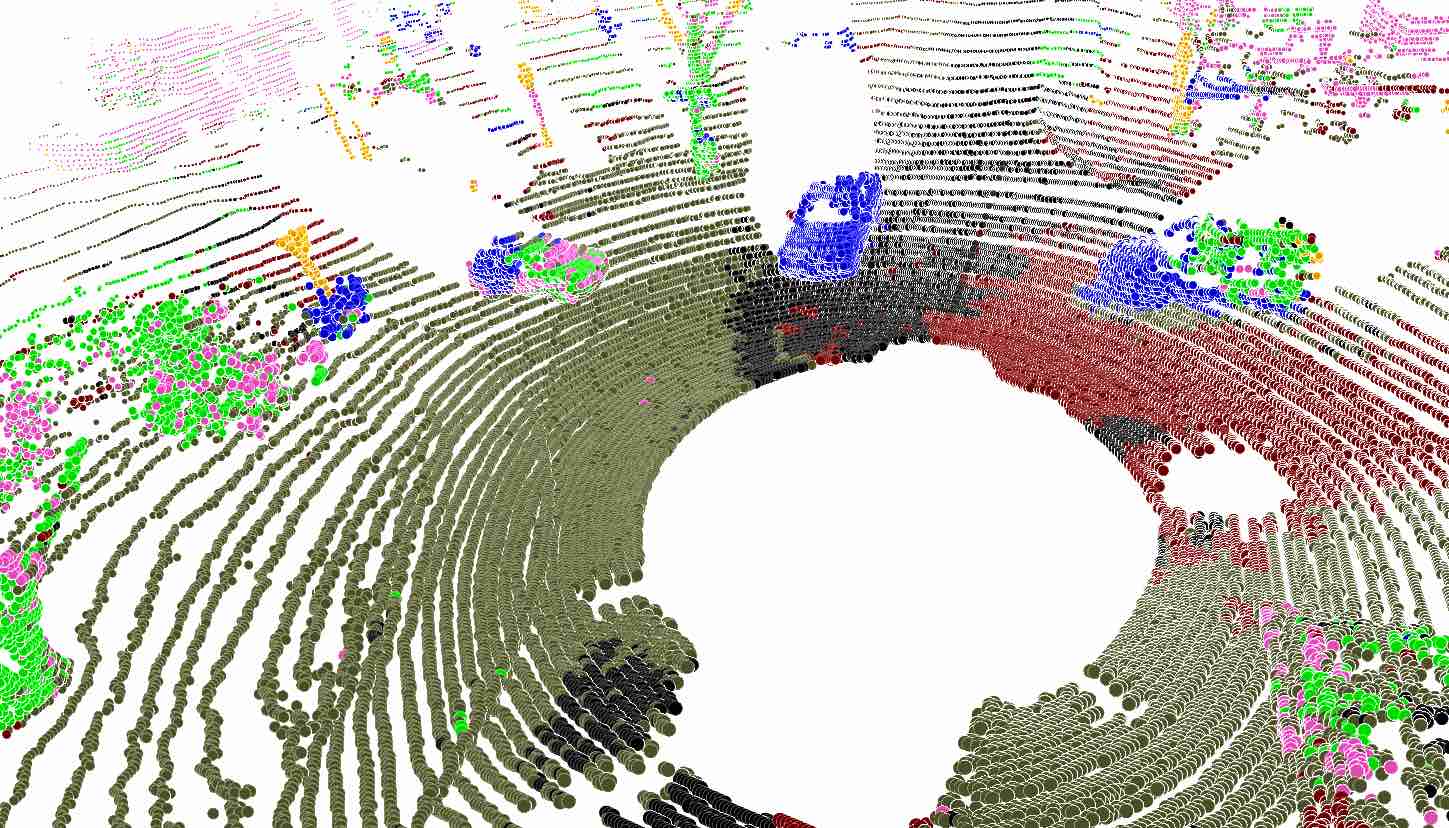}
        \end{overpic} &  
        \begin{overpic}[width=0.21\textwidth]{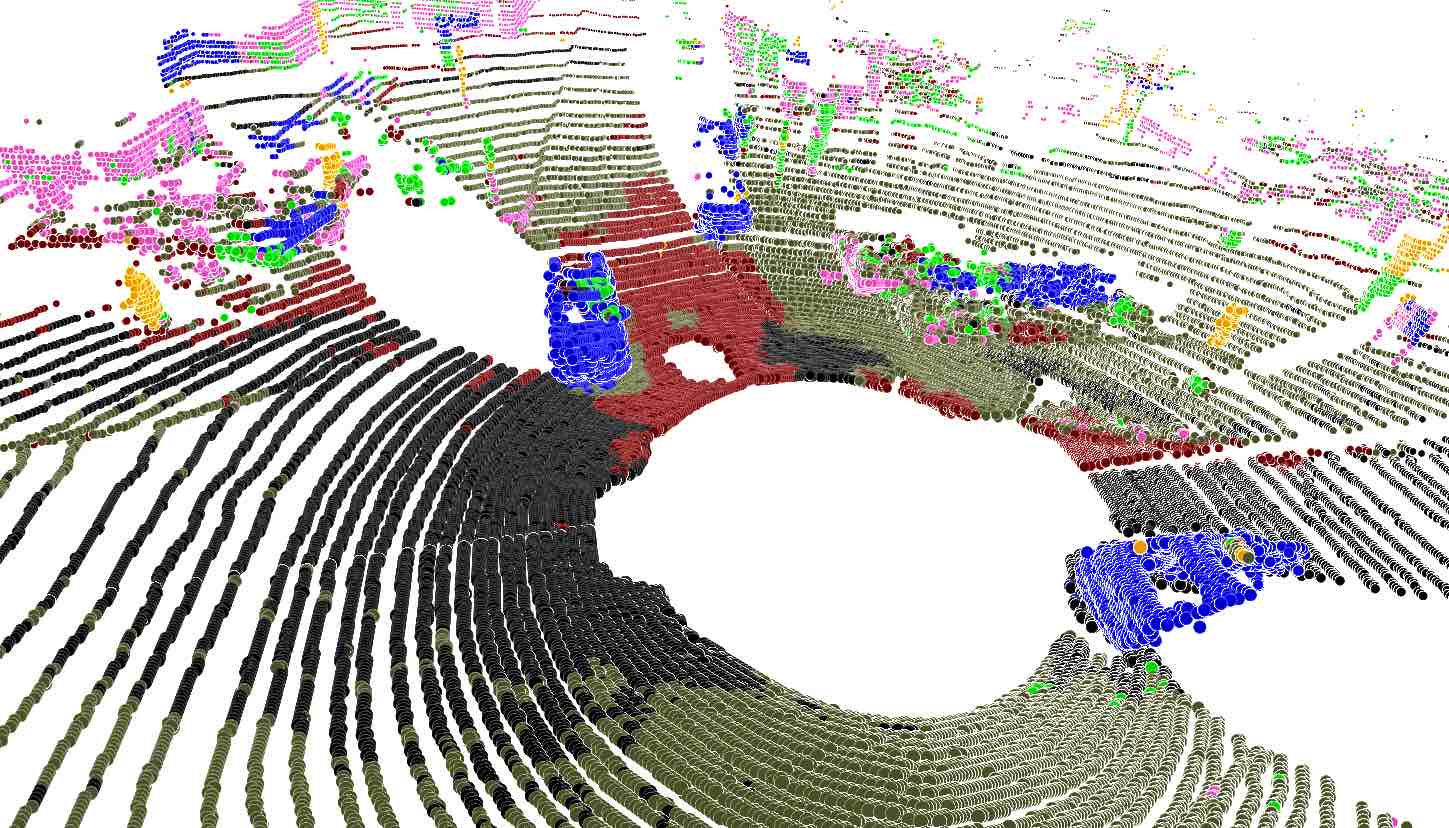}
        \end{overpic} &
        \begin{overpic}[width=0.21\textwidth]{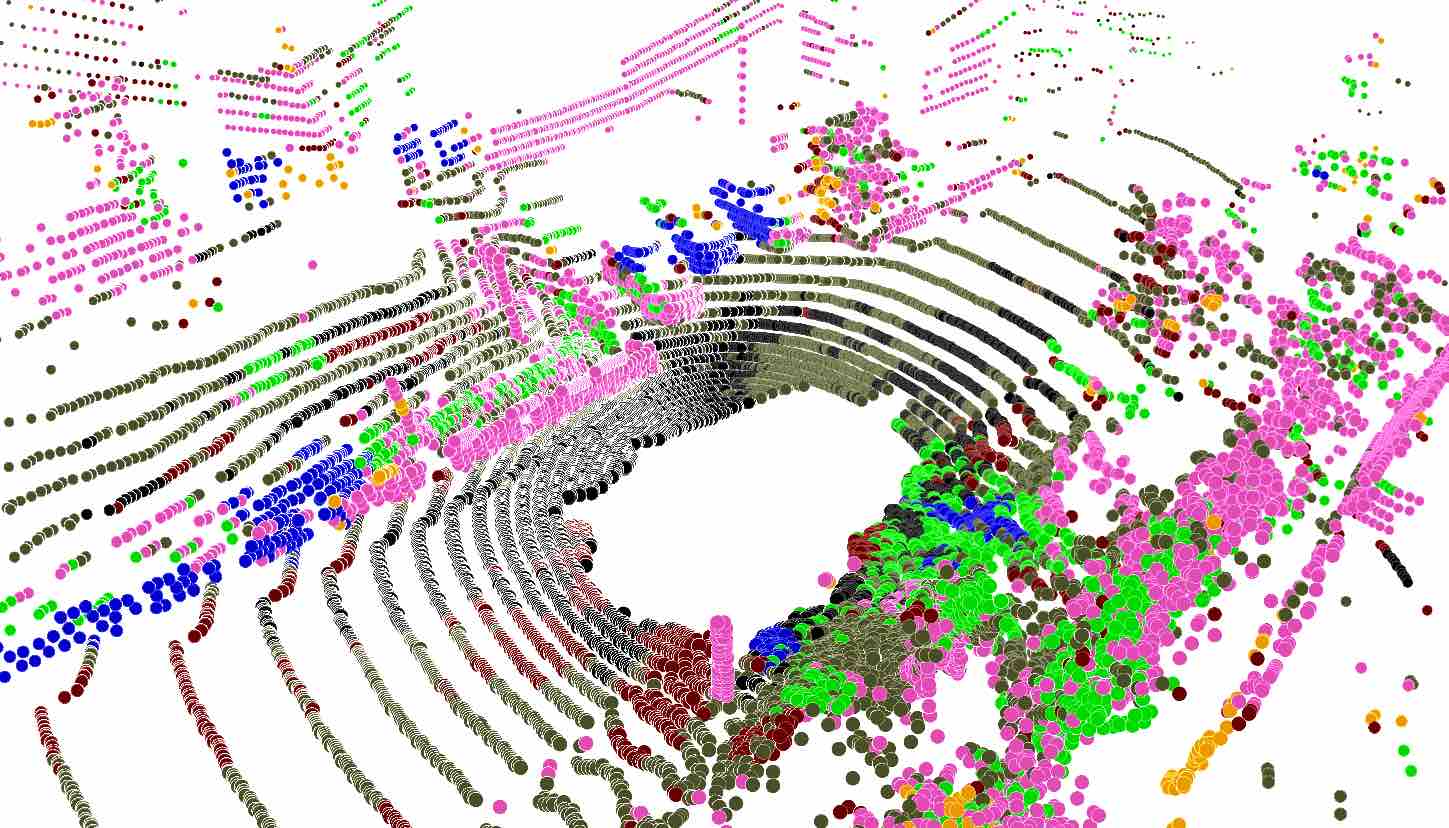}
        \end{overpic}& 
        \begin{overpic}[width=0.21\textwidth]{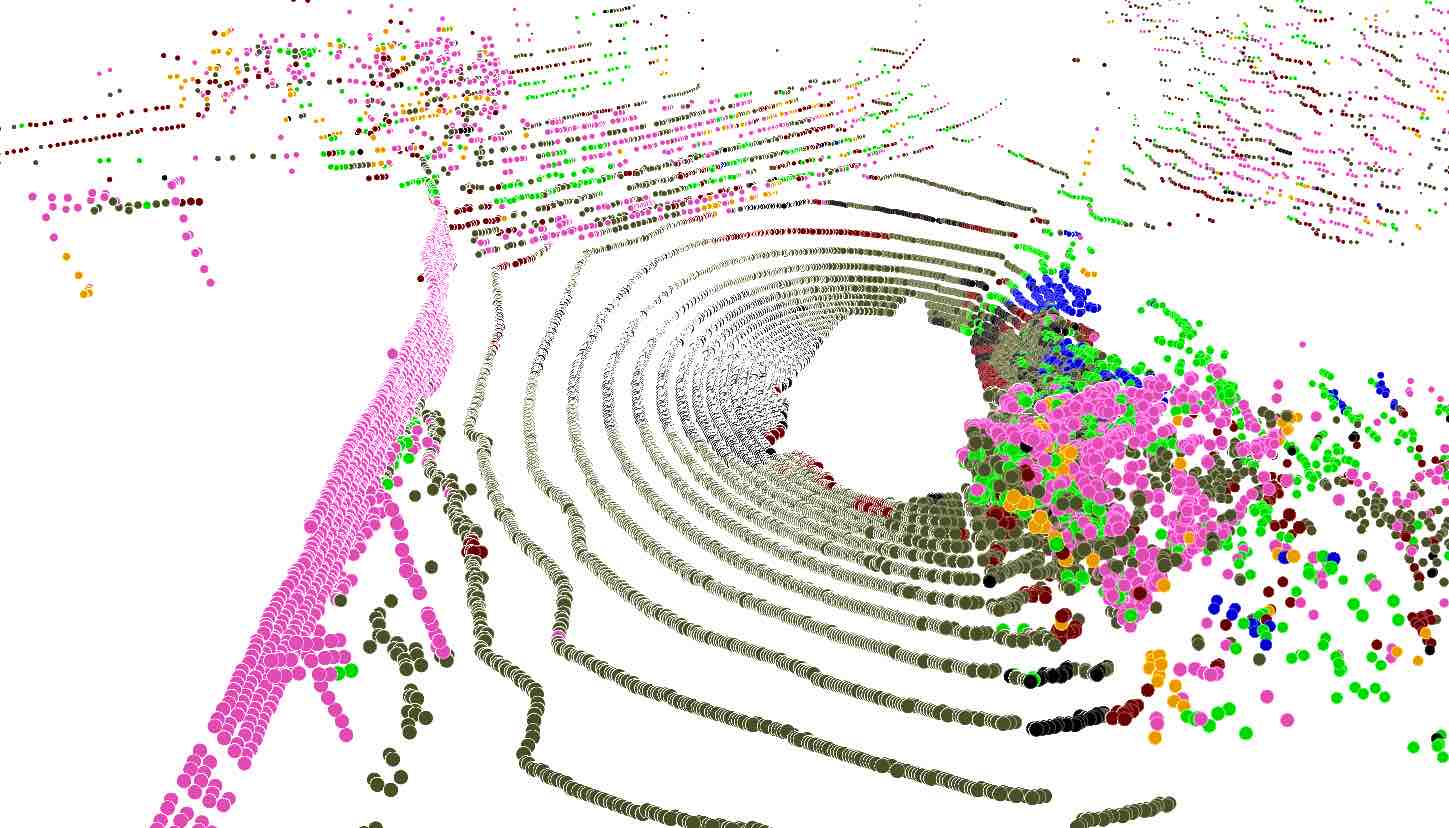}
        \end{overpic}\\
        \begin{overpic}[width=0.21\textwidth]{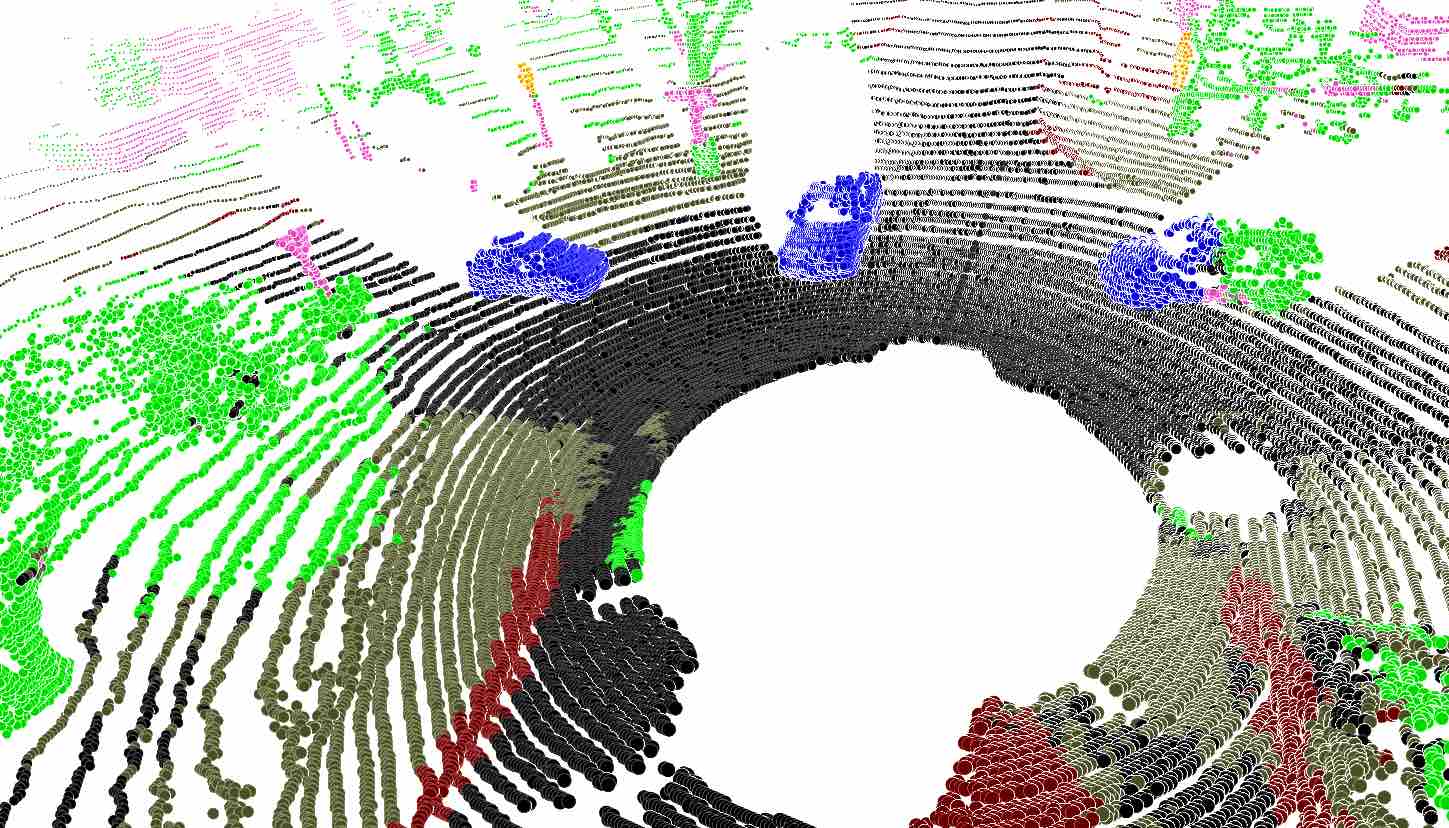}
        \end{overpic} &  
        \begin{overpic}[width=0.21\textwidth]{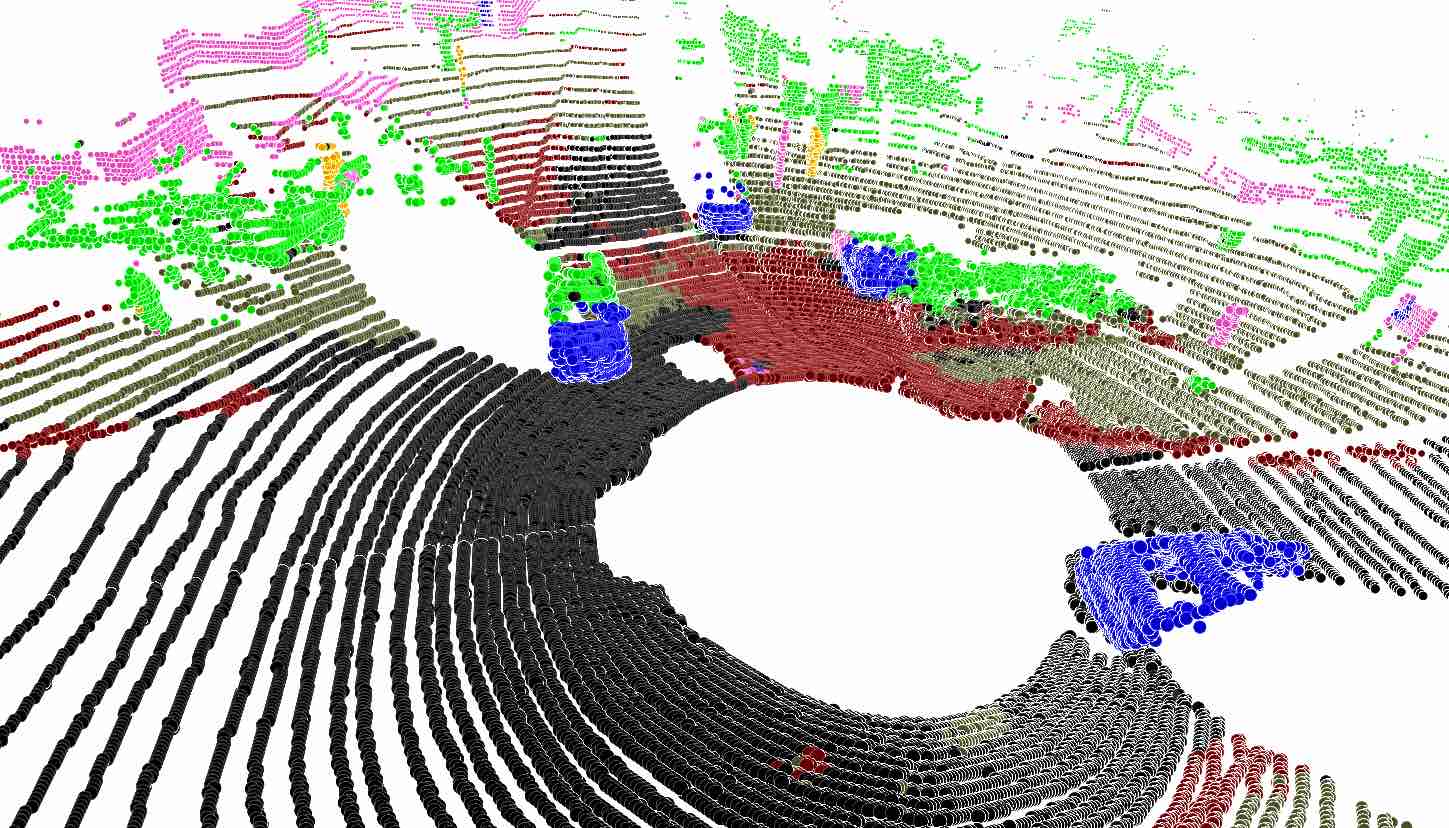}
        \end{overpic} &
        \begin{overpic}[width=0.21\textwidth]{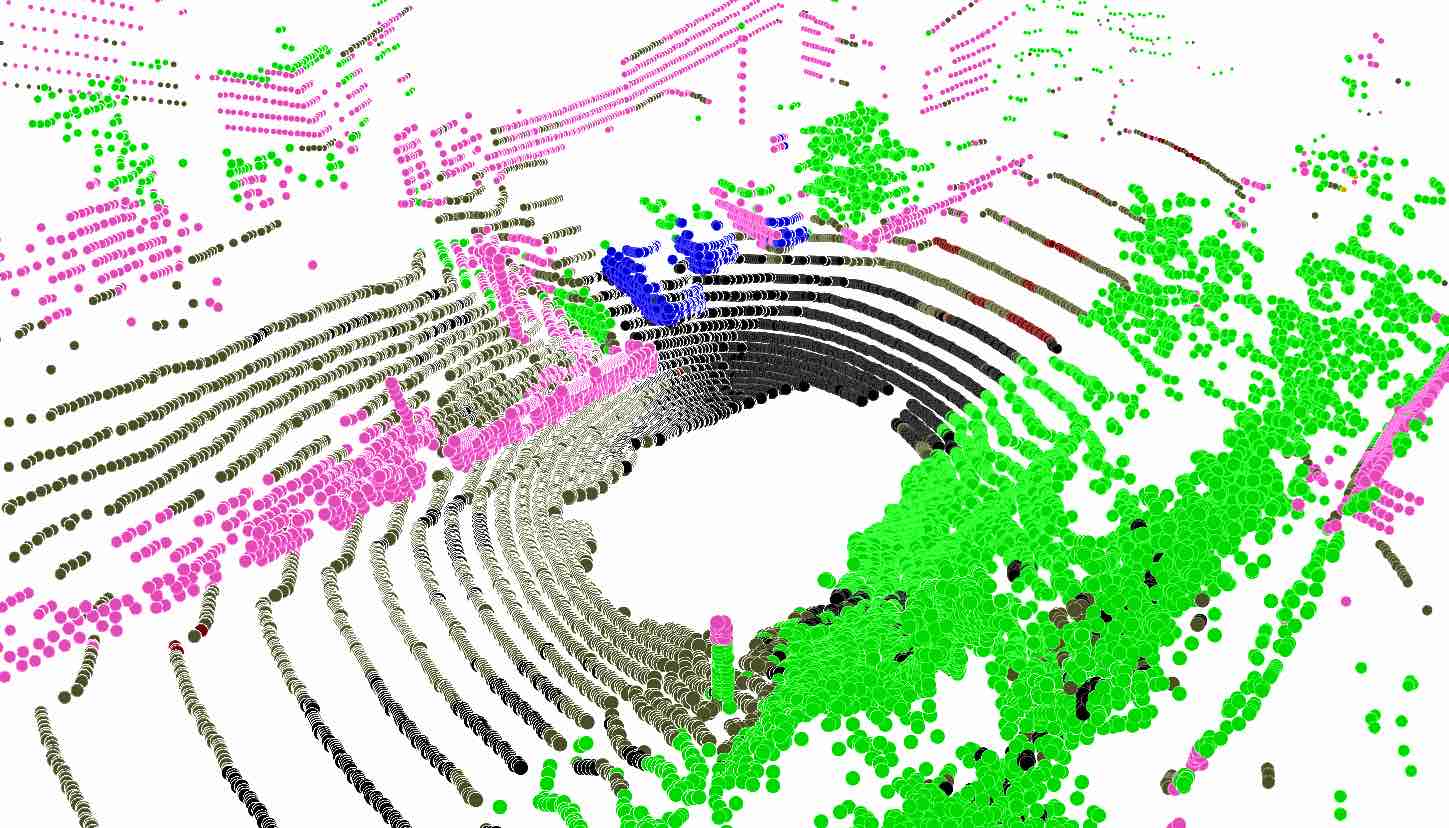}
        \end{overpic}& 
        \begin{overpic}[width=0.21\textwidth]{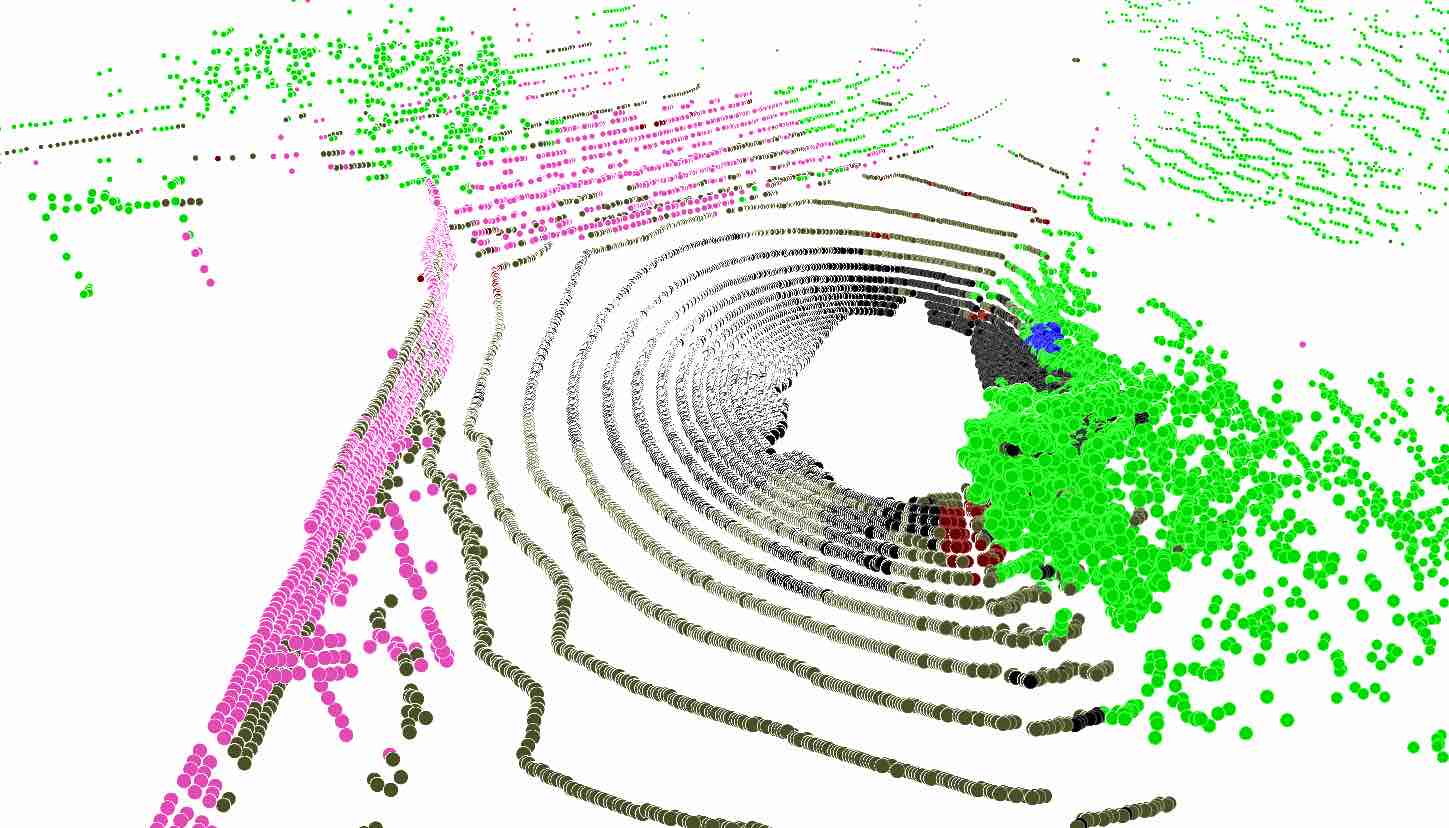}
        \end{overpic}\\
        \begin{overpic}[width=0.21\textwidth]{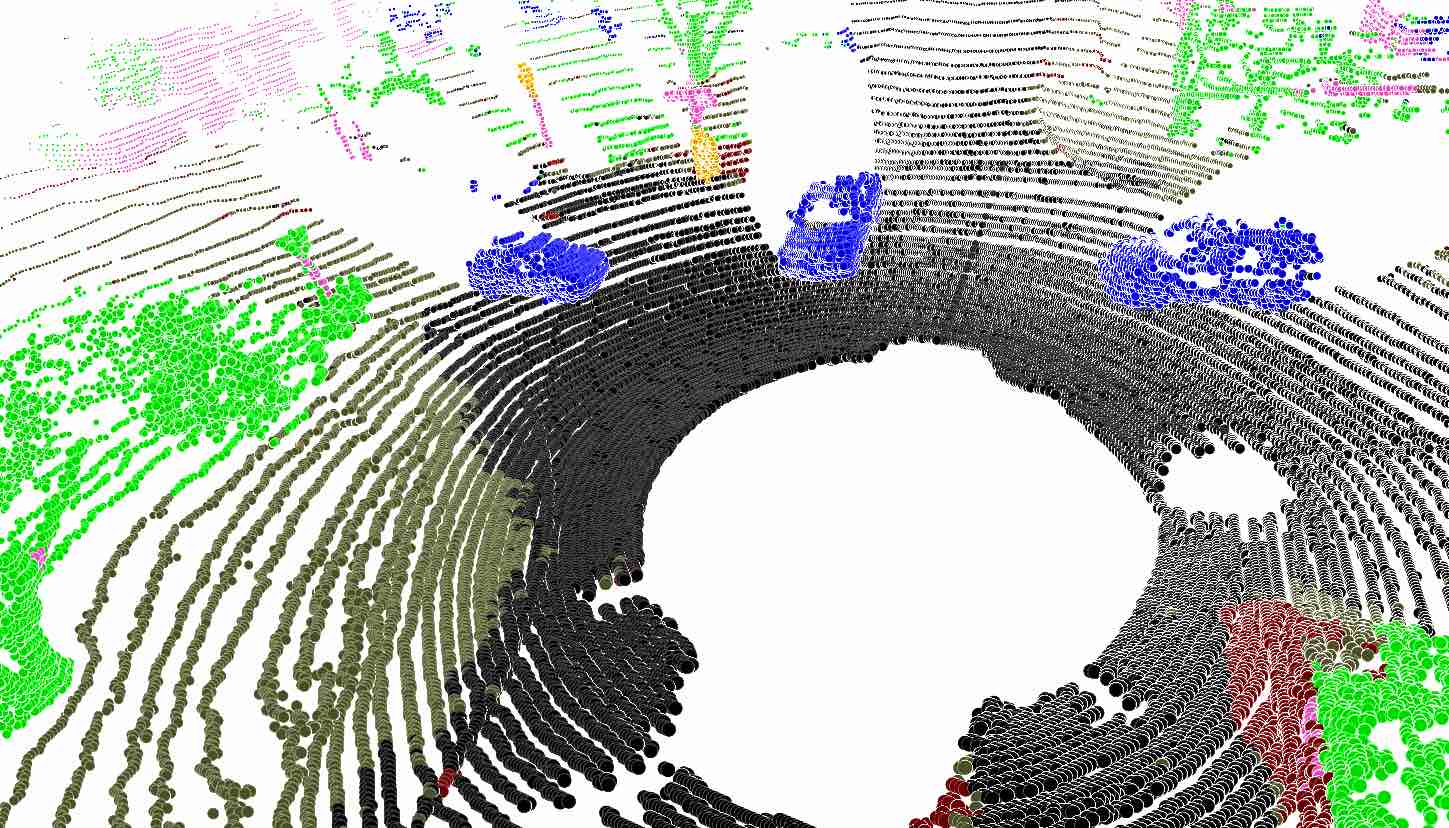}
        \end{overpic} &  
        \begin{overpic}[width=0.21\textwidth]{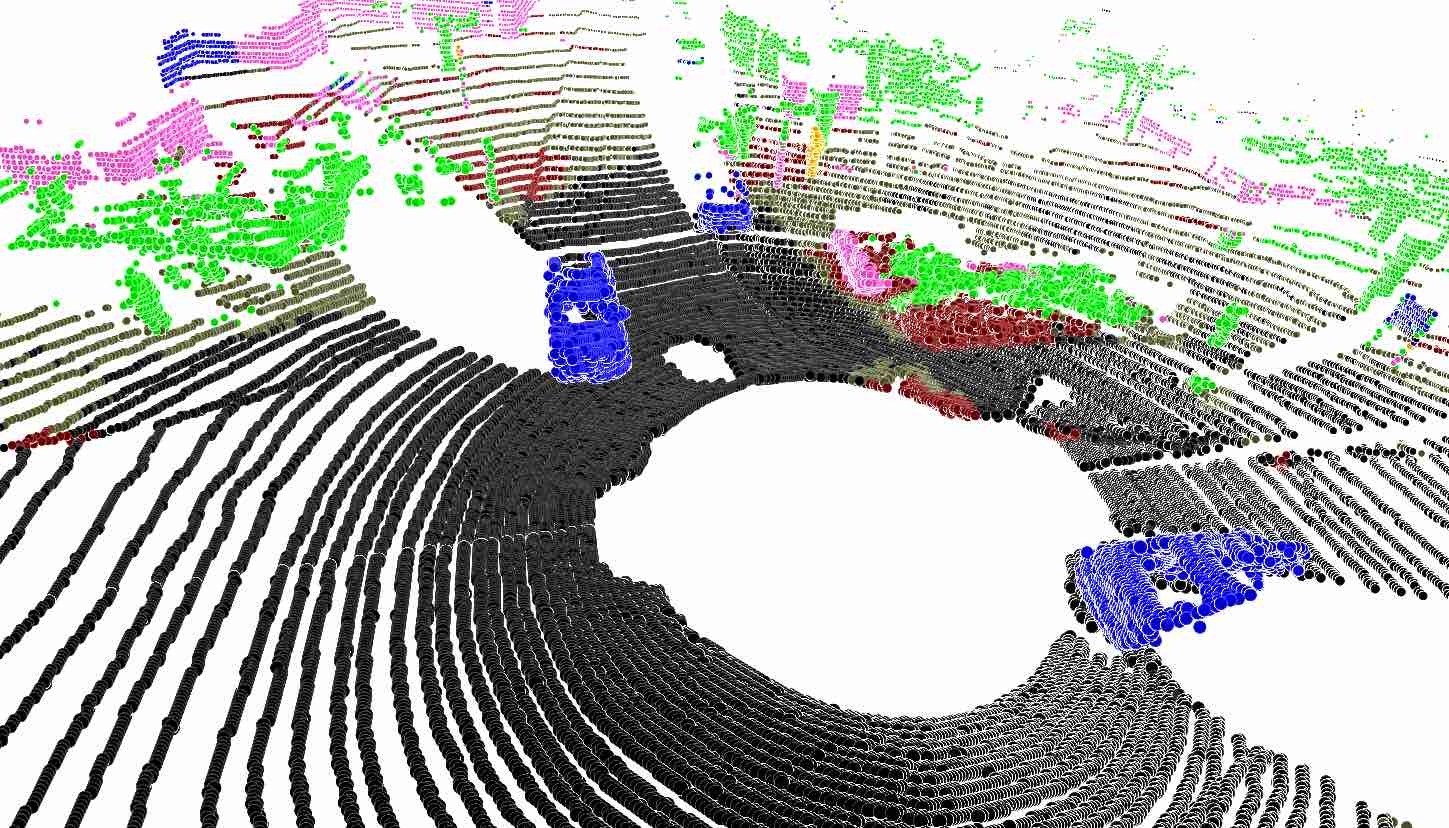}
        \end{overpic} &
        \begin{overpic}[width=0.21\textwidth]{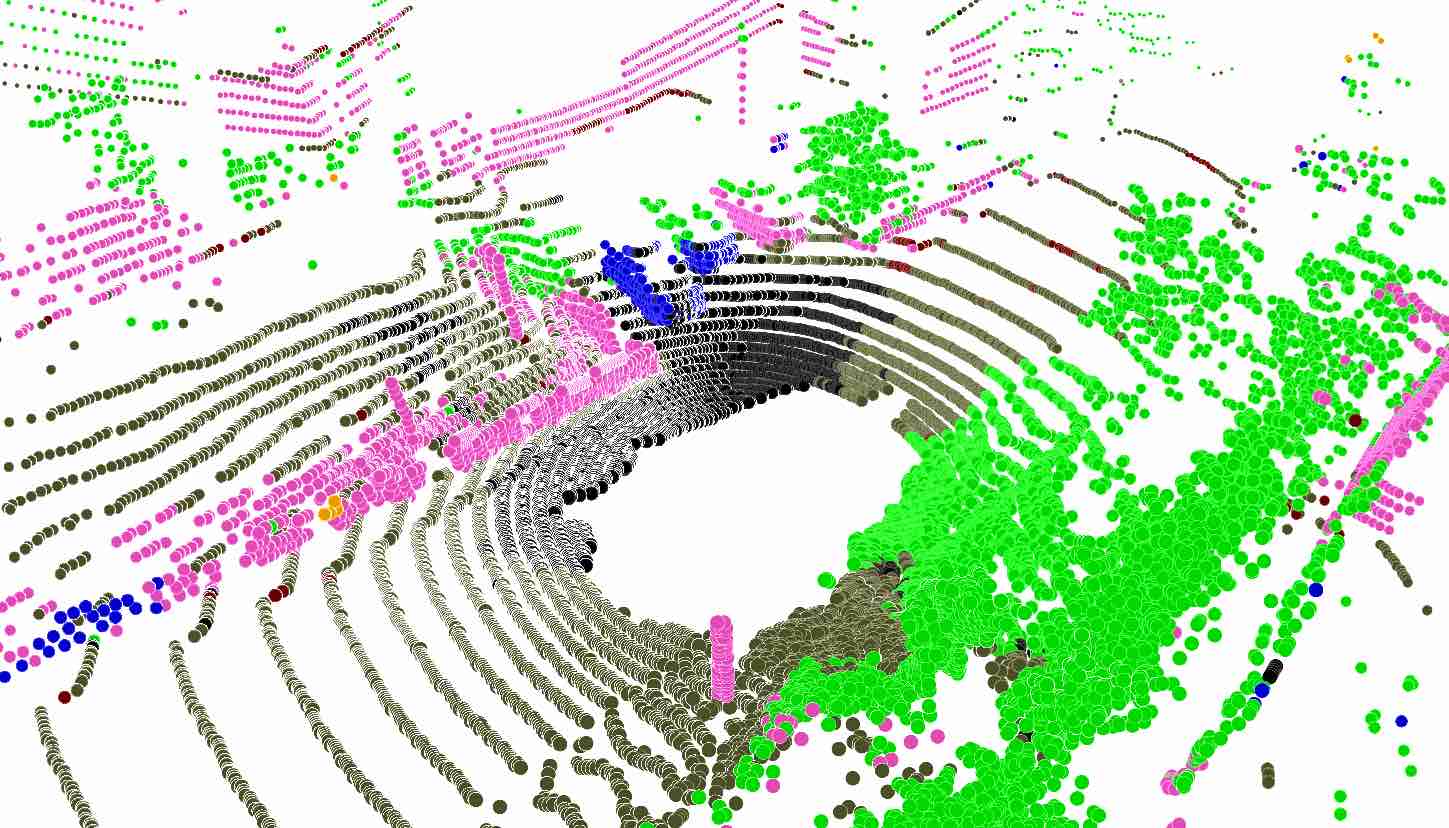}
        \end{overpic}& 
        \begin{overpic}[width=0.21\textwidth]{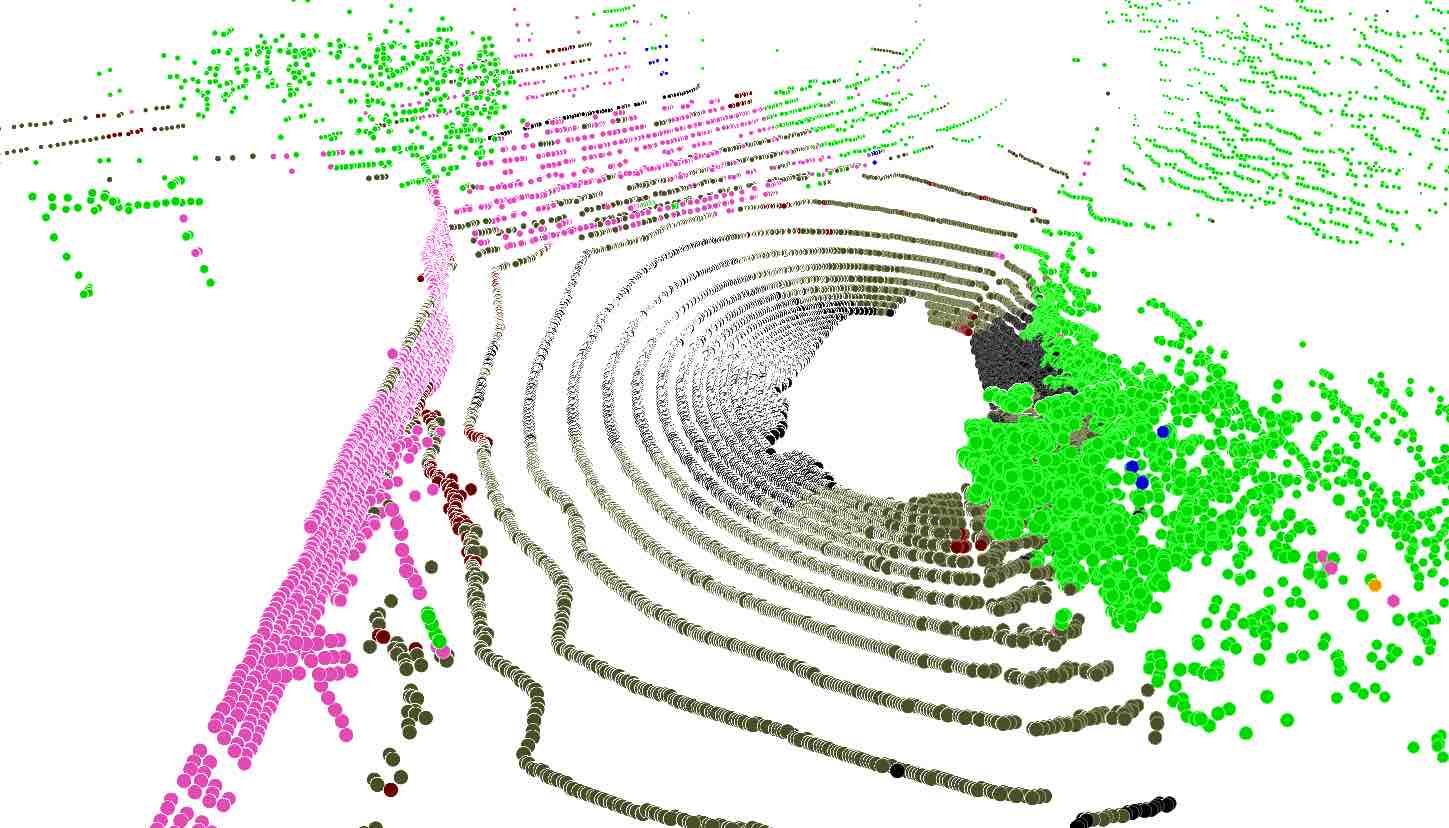}
        \end{overpic}\\
        \begin{overpic}[width=0.21\textwidth]{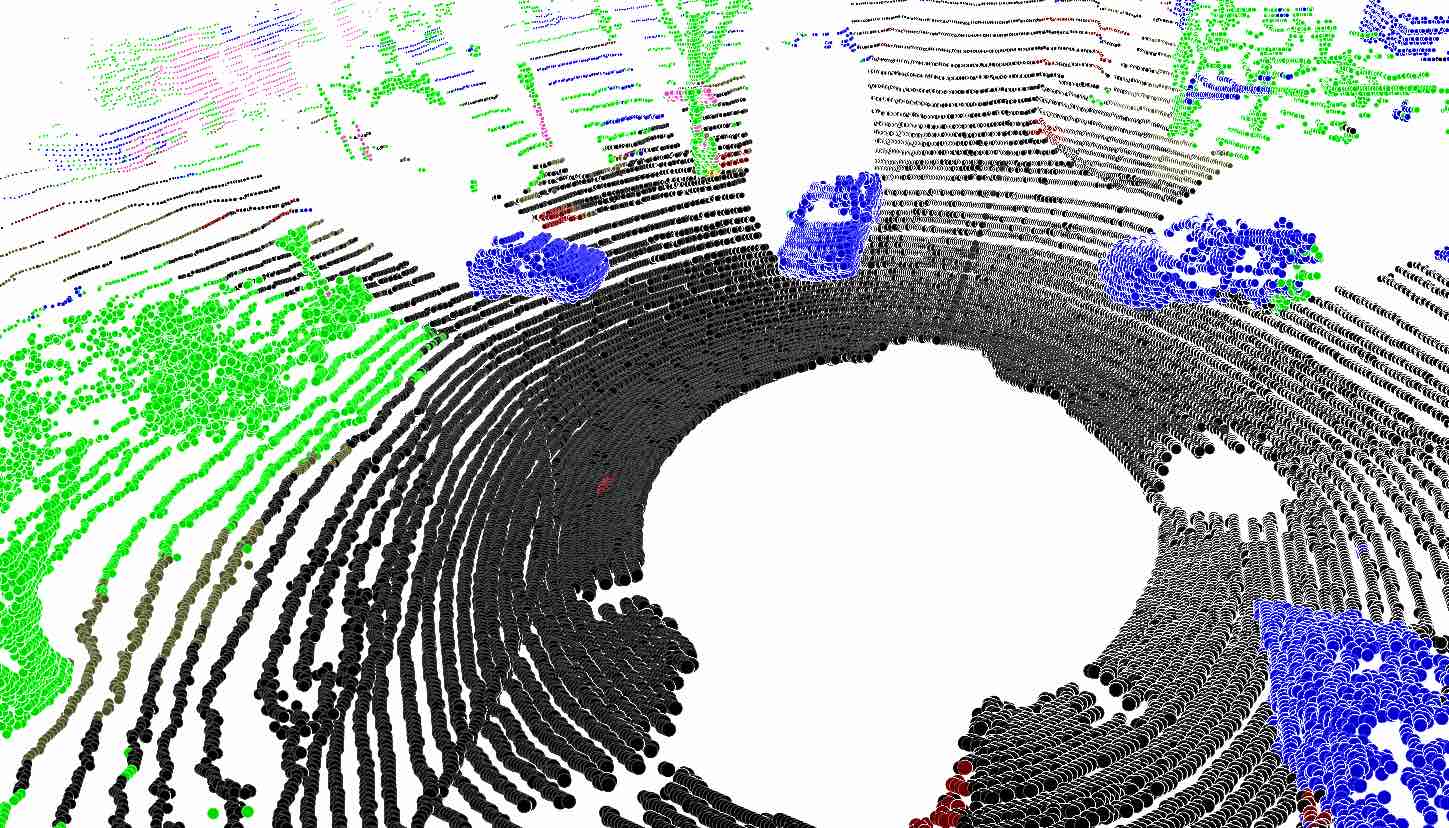}
        \end{overpic} &  
        \begin{overpic}[width=0.21\textwidth]{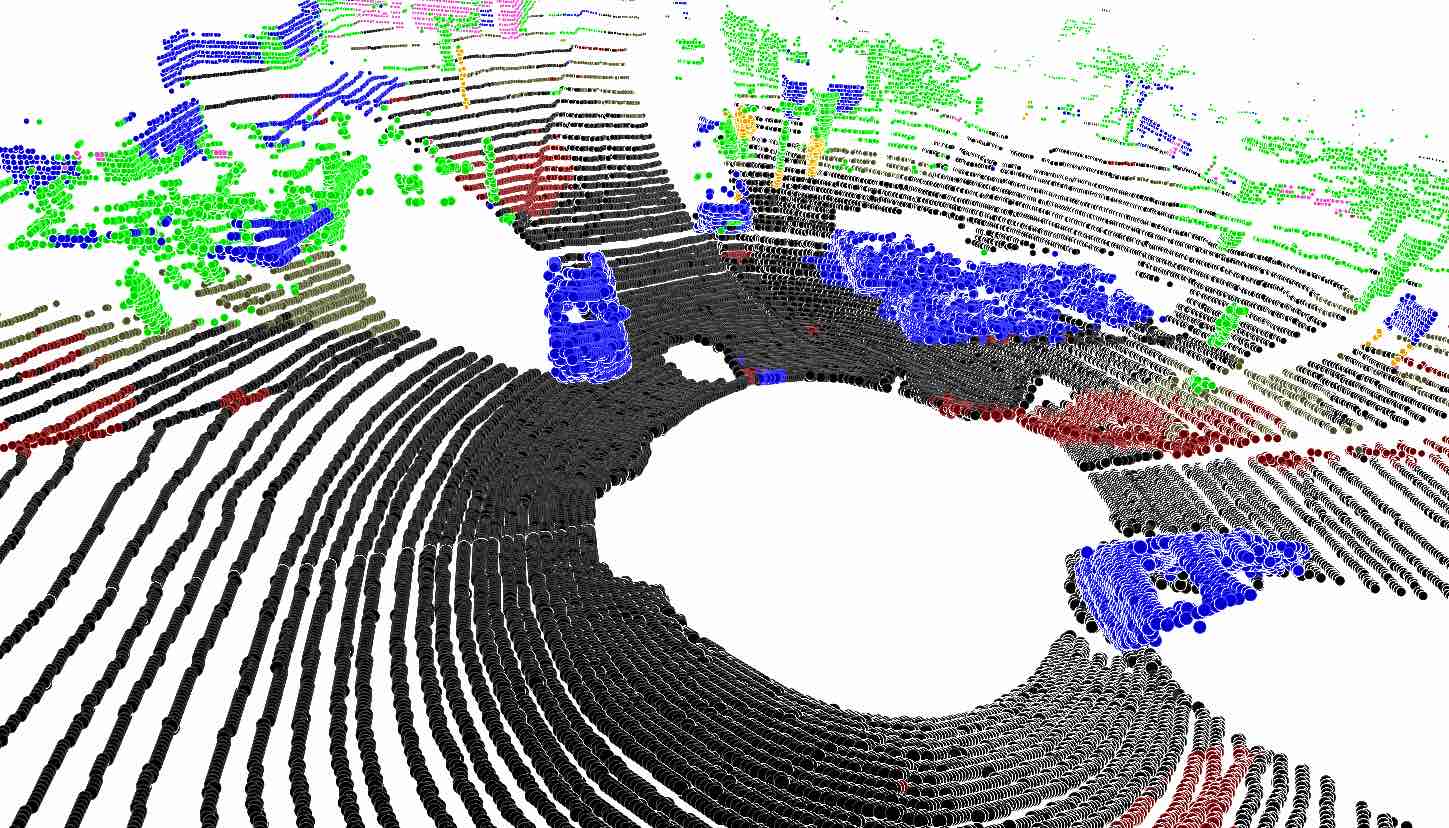}
        \end{overpic} &
        \begin{overpic}[width=0.21\textwidth]{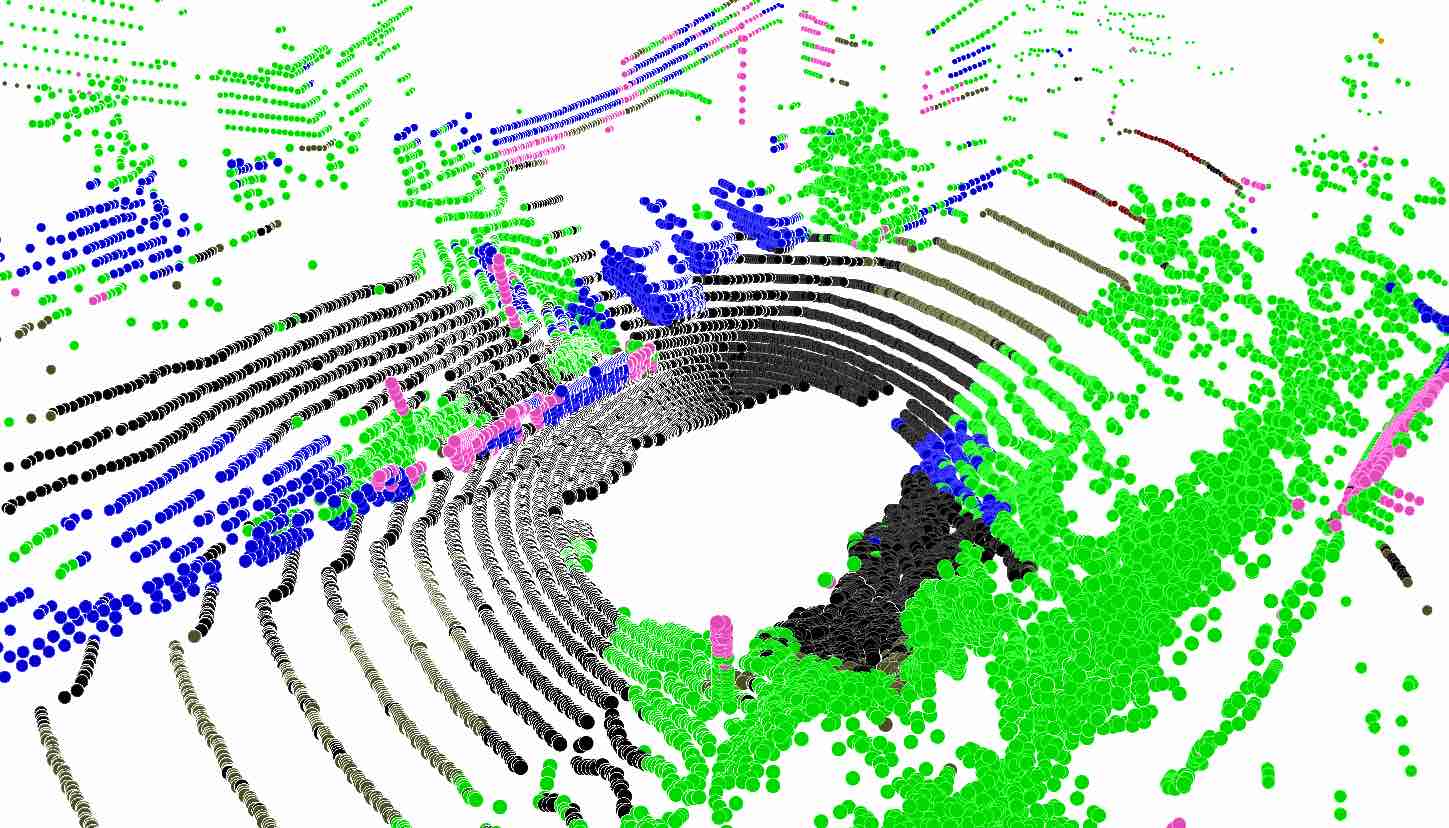}
        \end{overpic}& 
        \begin{overpic}[width=0.21\textwidth]{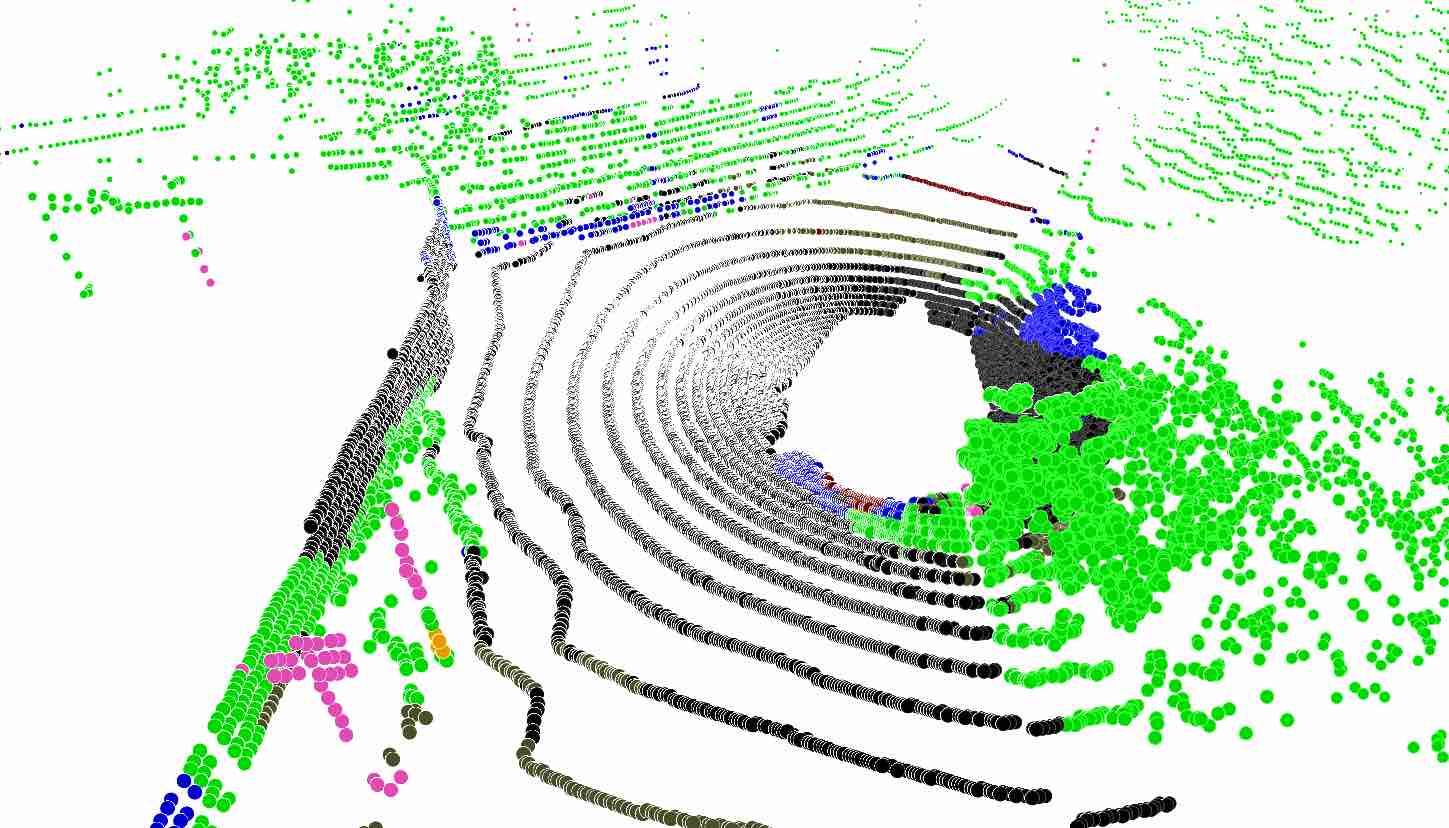}
        \end{overpic}\\
        \begin{overpic}[width=0.21\textwidth]{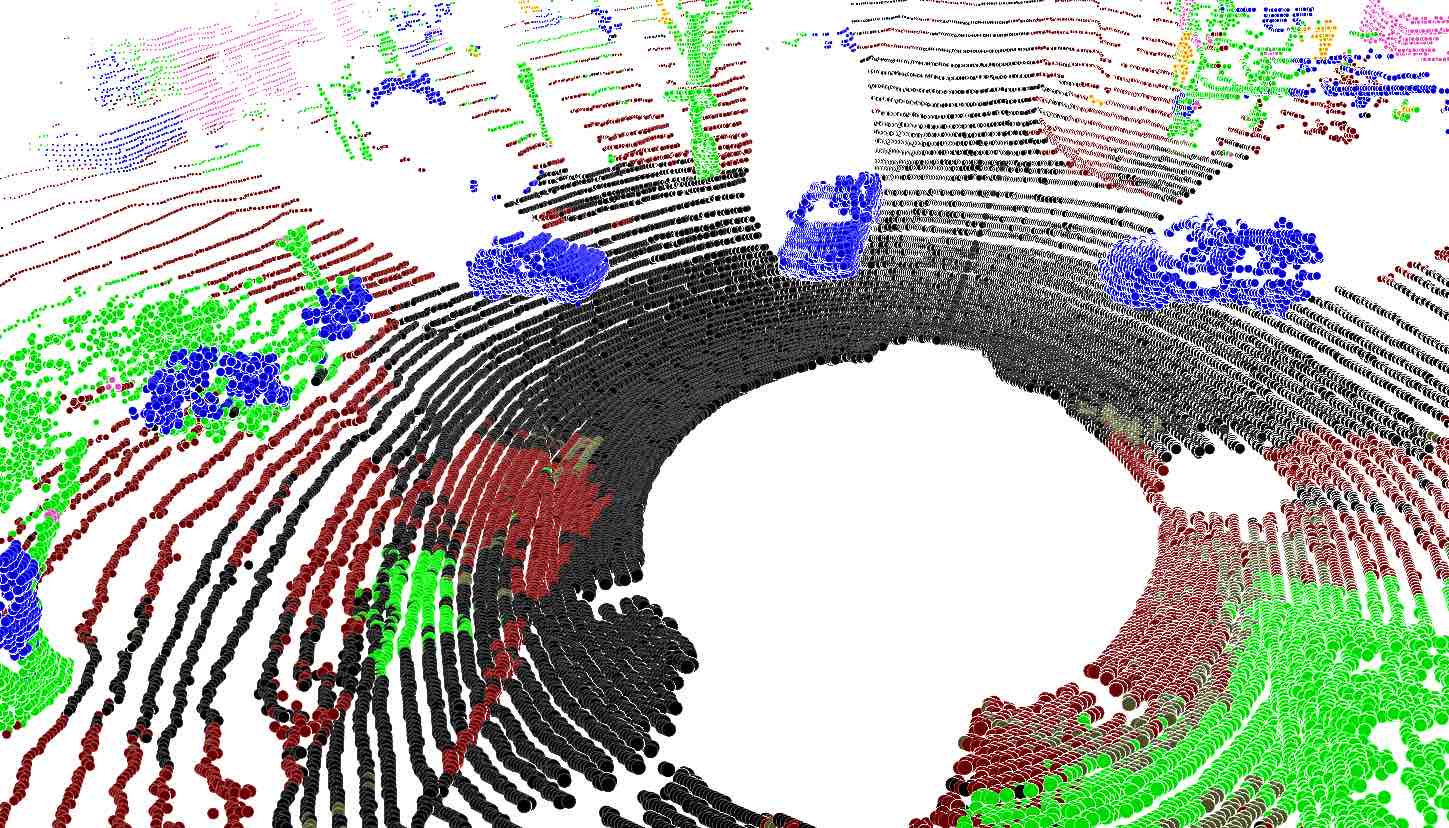}
        \end{overpic} &  
        \begin{overpic}[width=0.21\textwidth]{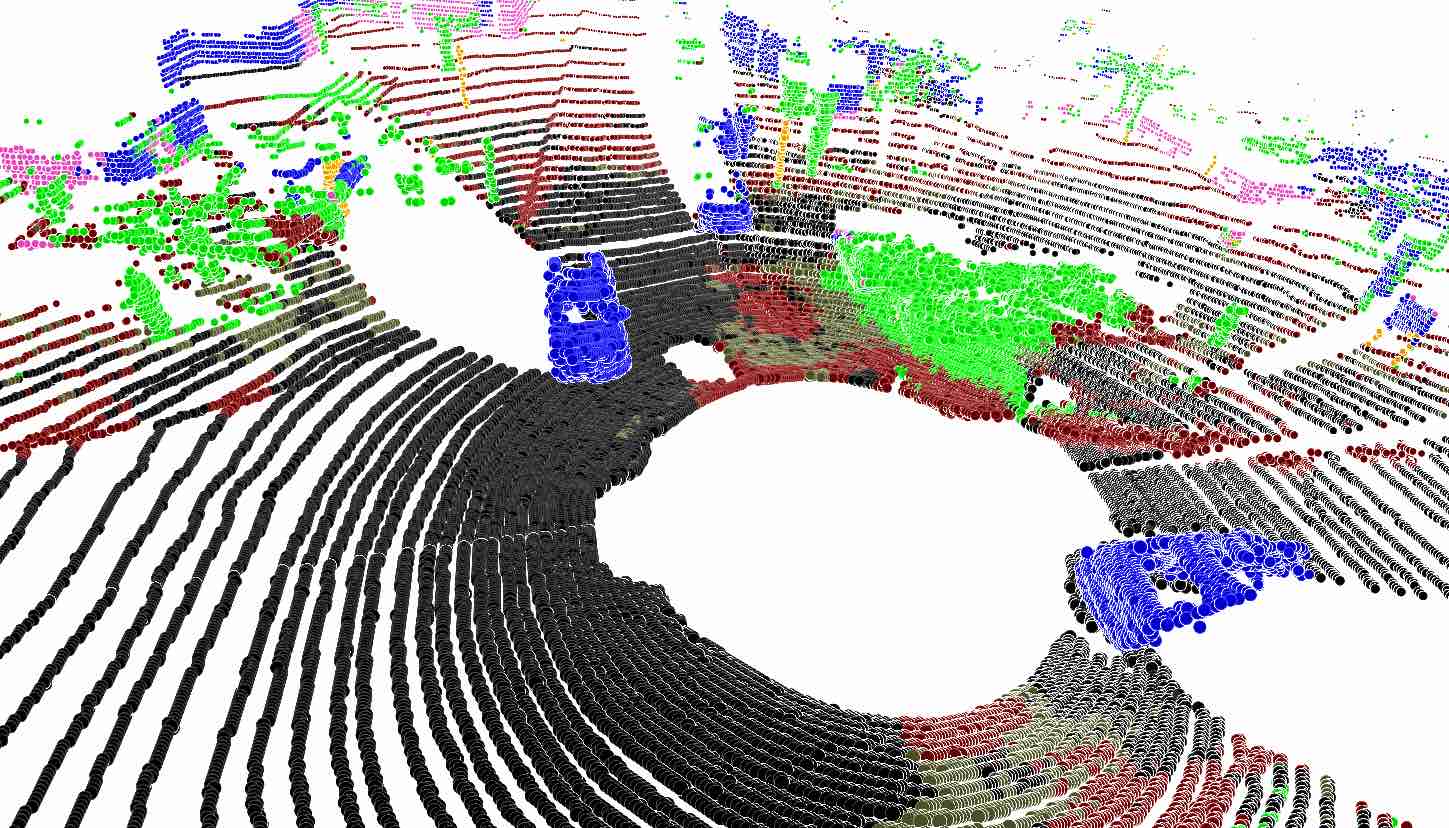}
        \end{overpic} &
        \begin{overpic}[width=0.21\textwidth]{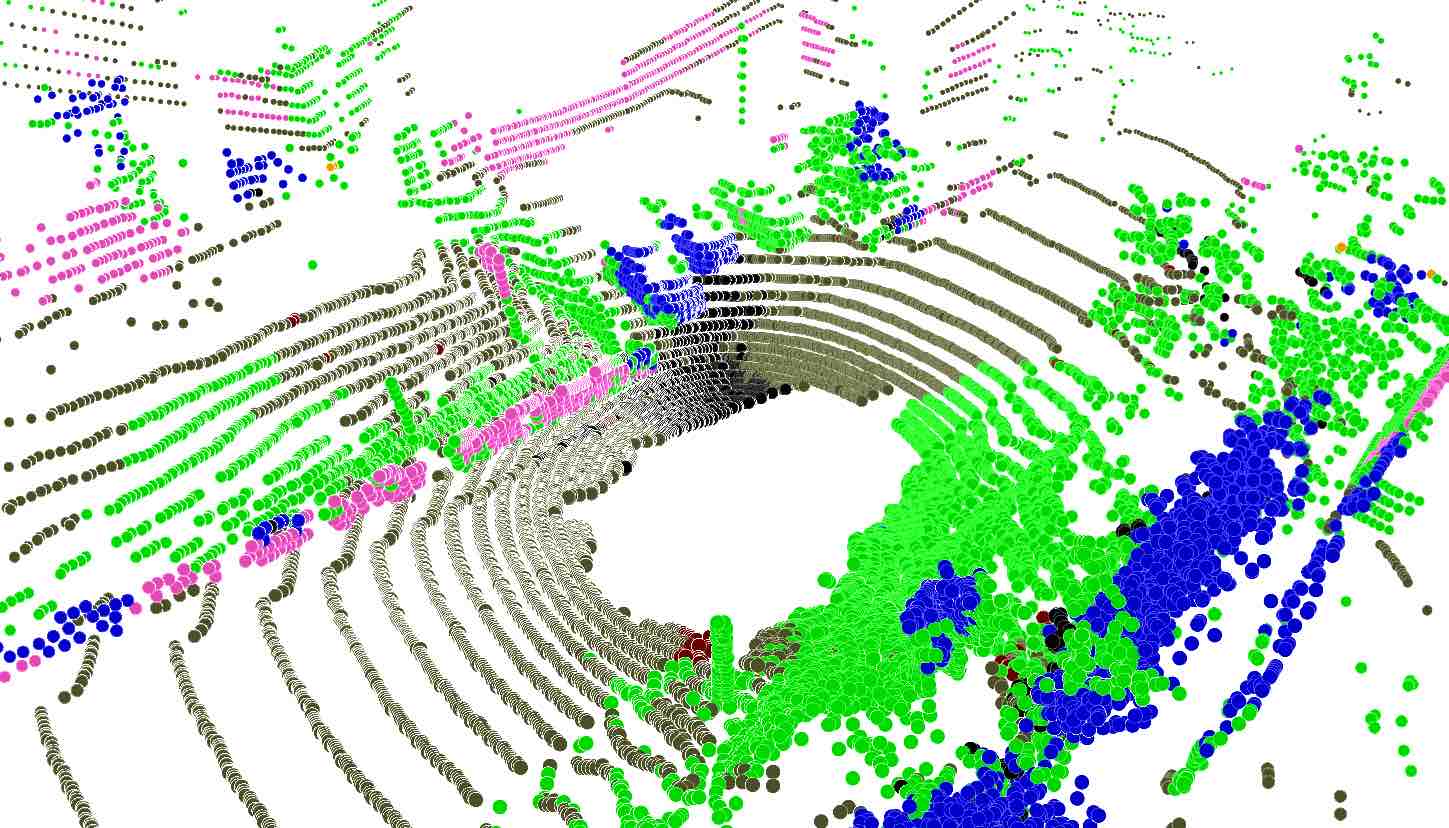}
        \end{overpic}& 
        \begin{overpic}[width=0.21\textwidth]{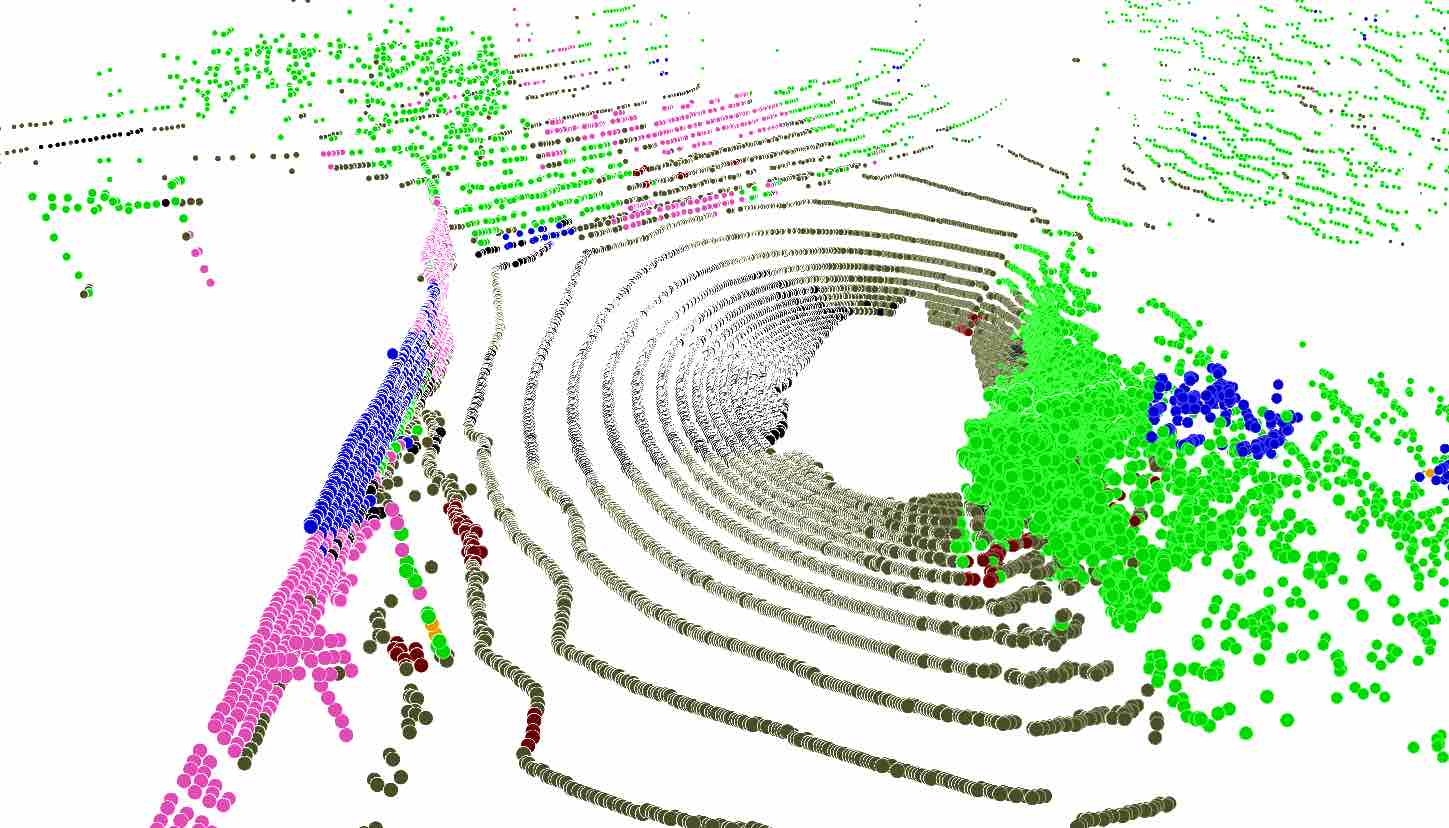}
        \end{overpic}\\
        \begin{overpic}[width=0.21\textwidth]{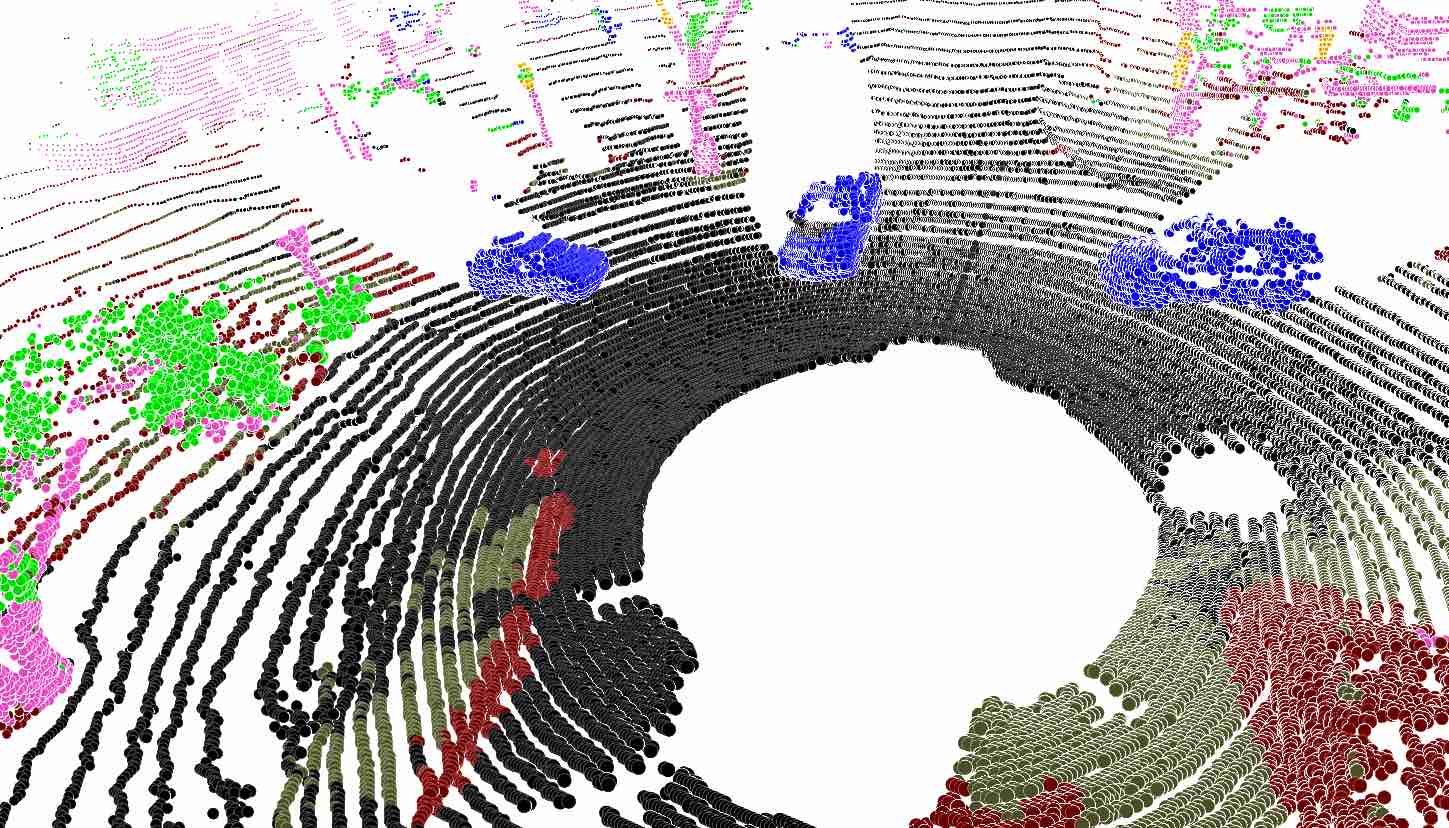}
        \end{overpic} &  
        \begin{overpic}[width=0.21\textwidth]{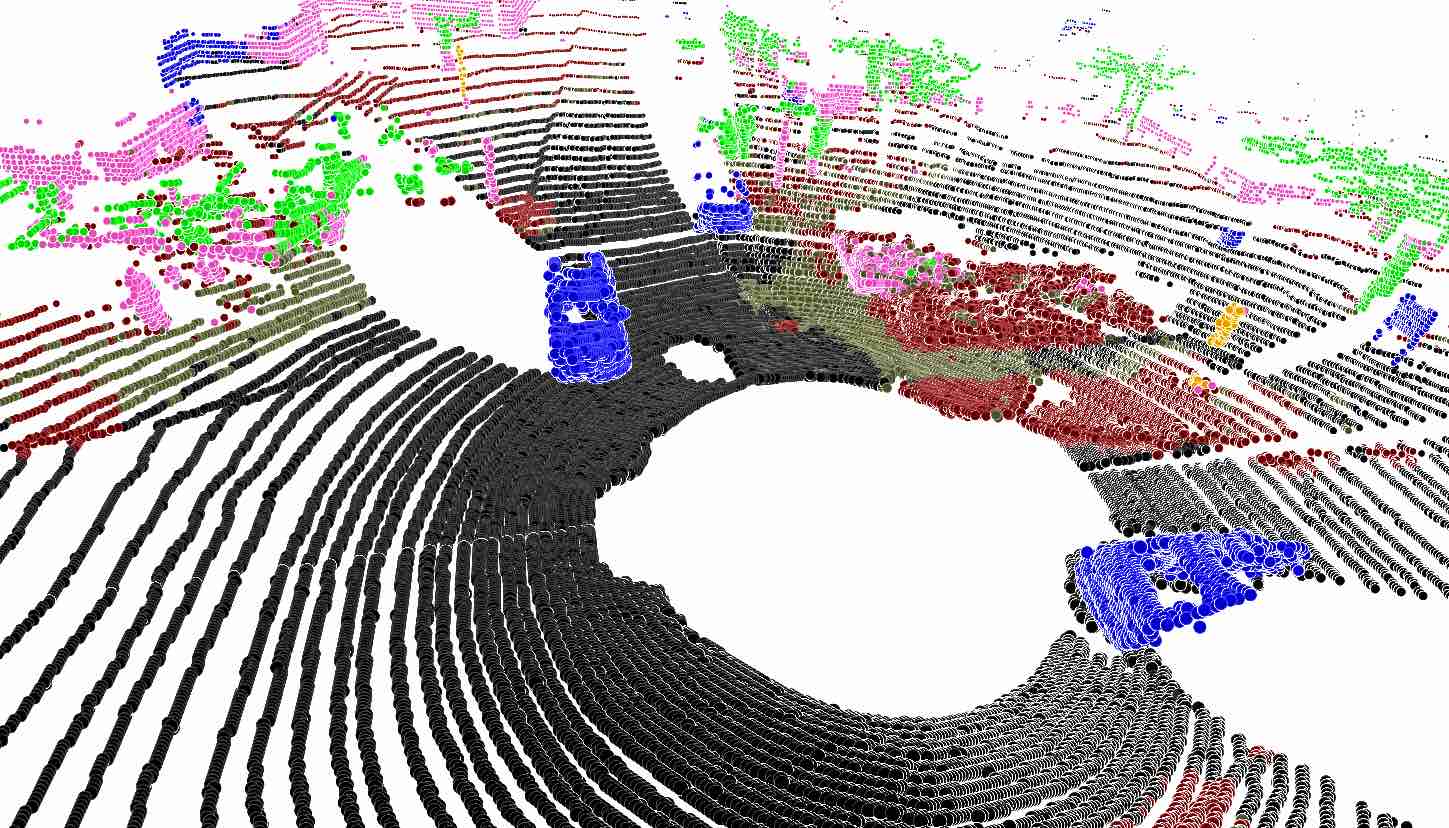}
        \end{overpic} &
        \begin{overpic}[width=0.21\textwidth]{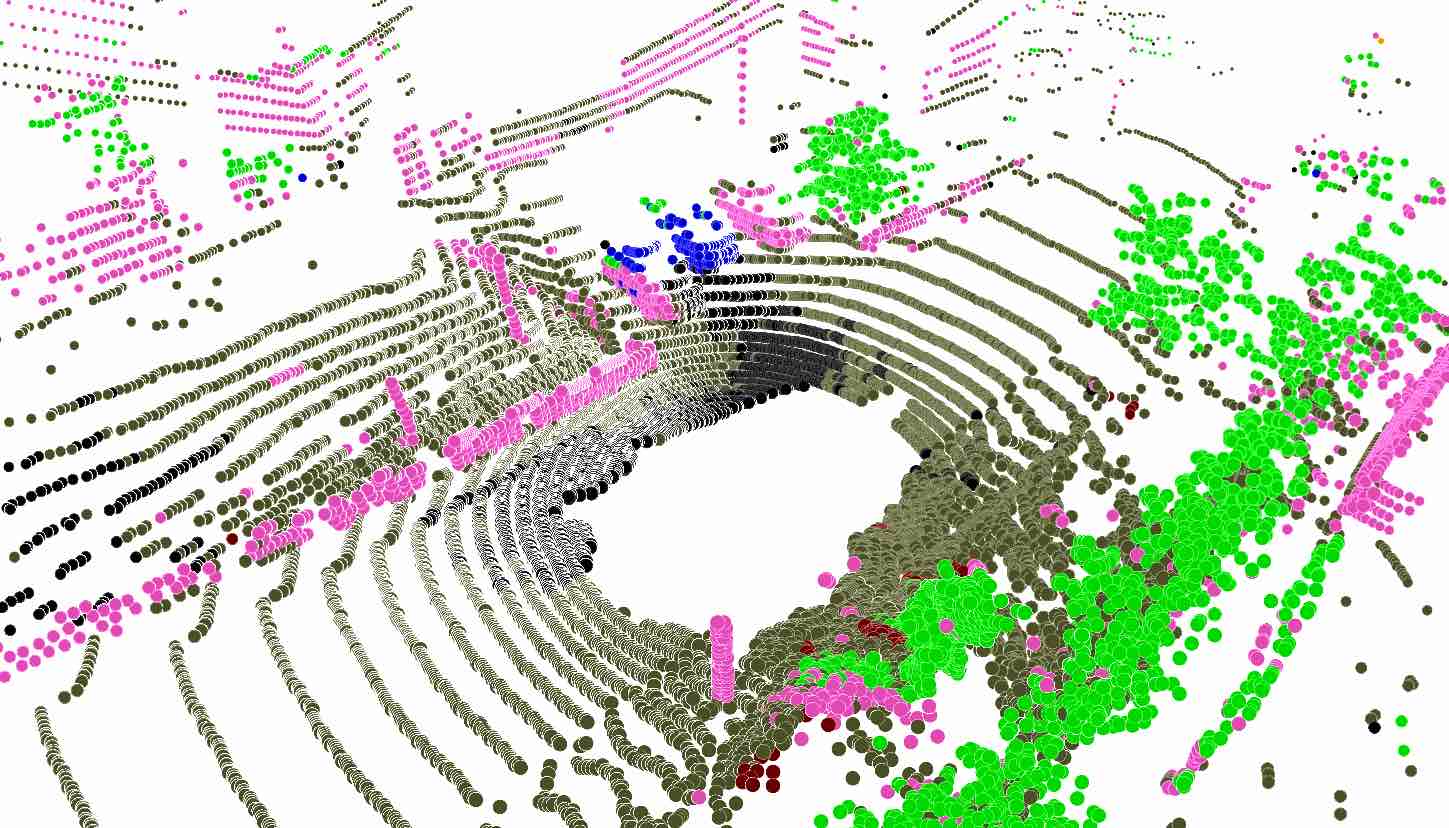}
        \end{overpic}& 
        \begin{overpic}[width=0.21\textwidth]{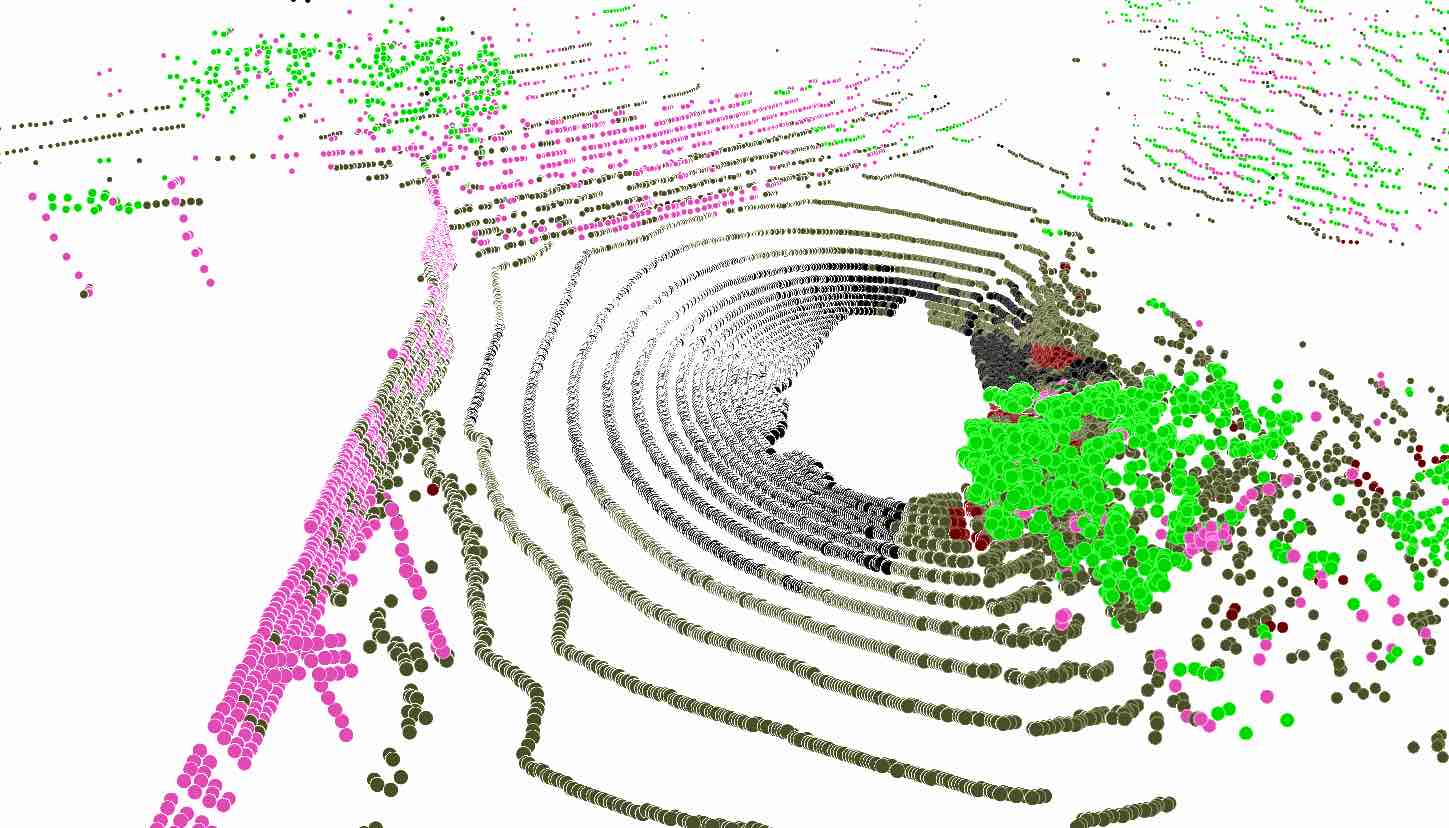}
        \end{overpic}\\
        \begin{overpic}[width=0.21\textwidth]{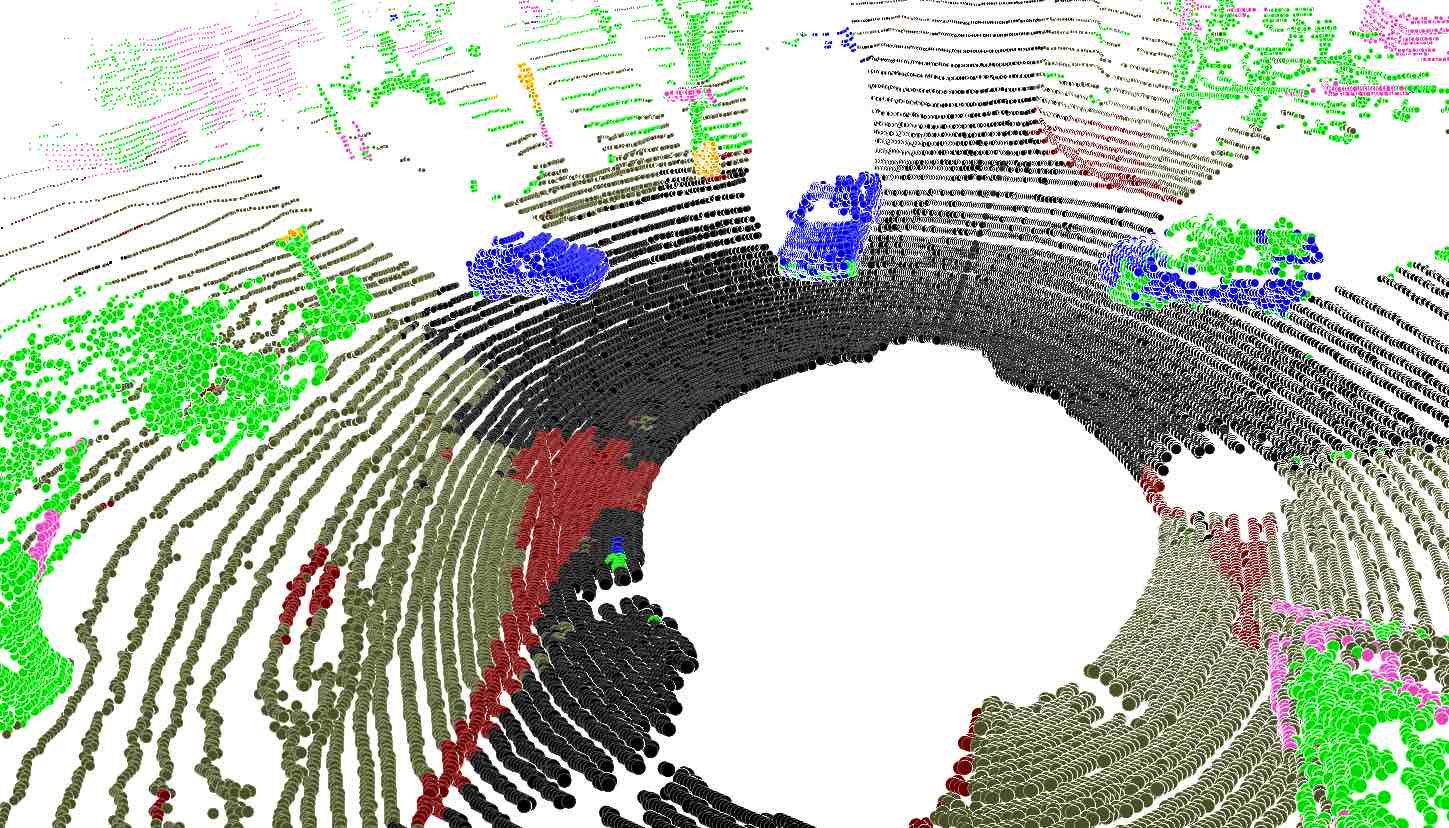}
        \end{overpic} &  
        \begin{overpic}[width=0.21\textwidth]{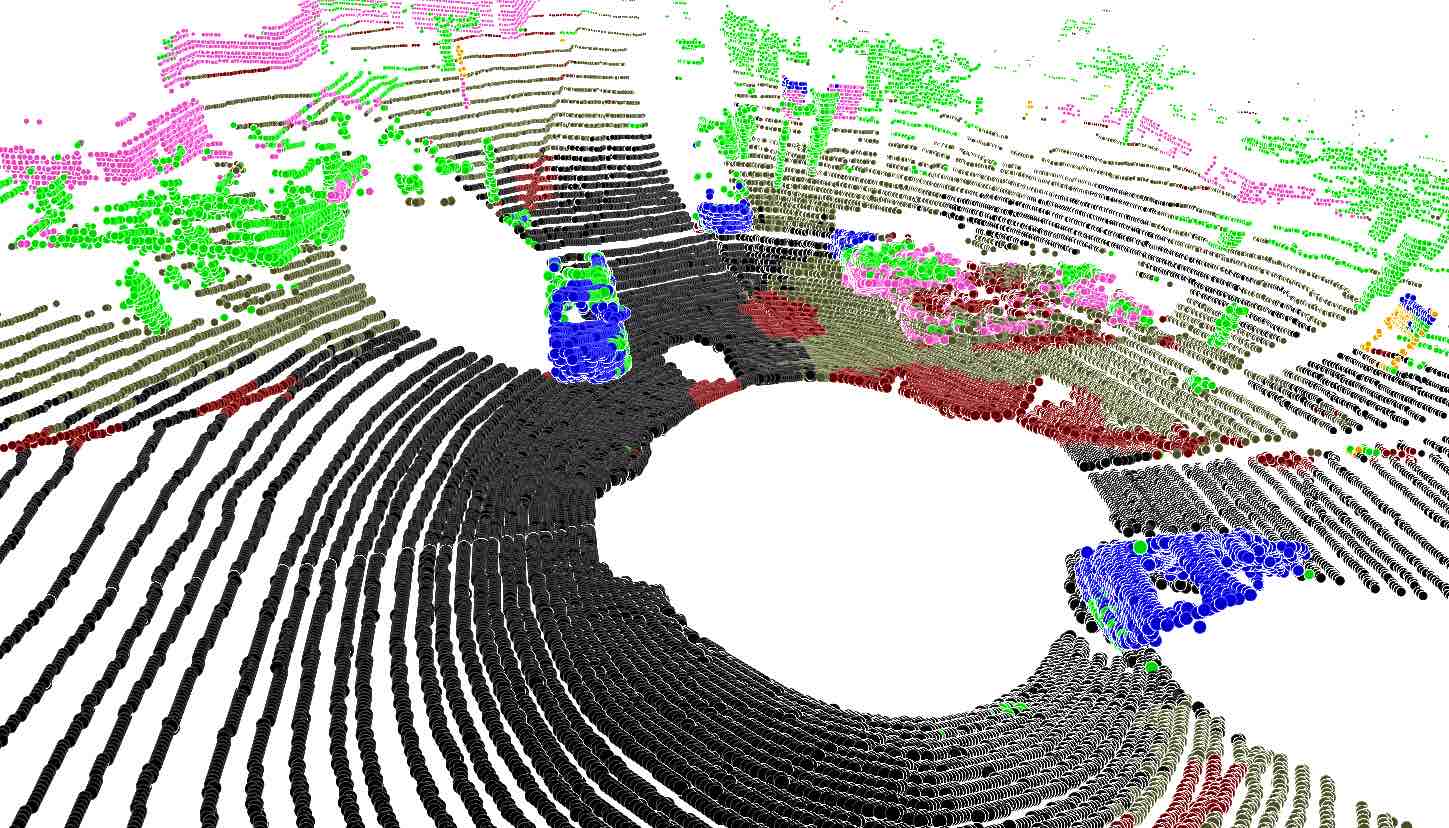}
        \end{overpic} &
        \begin{overpic}[width=0.21\textwidth]{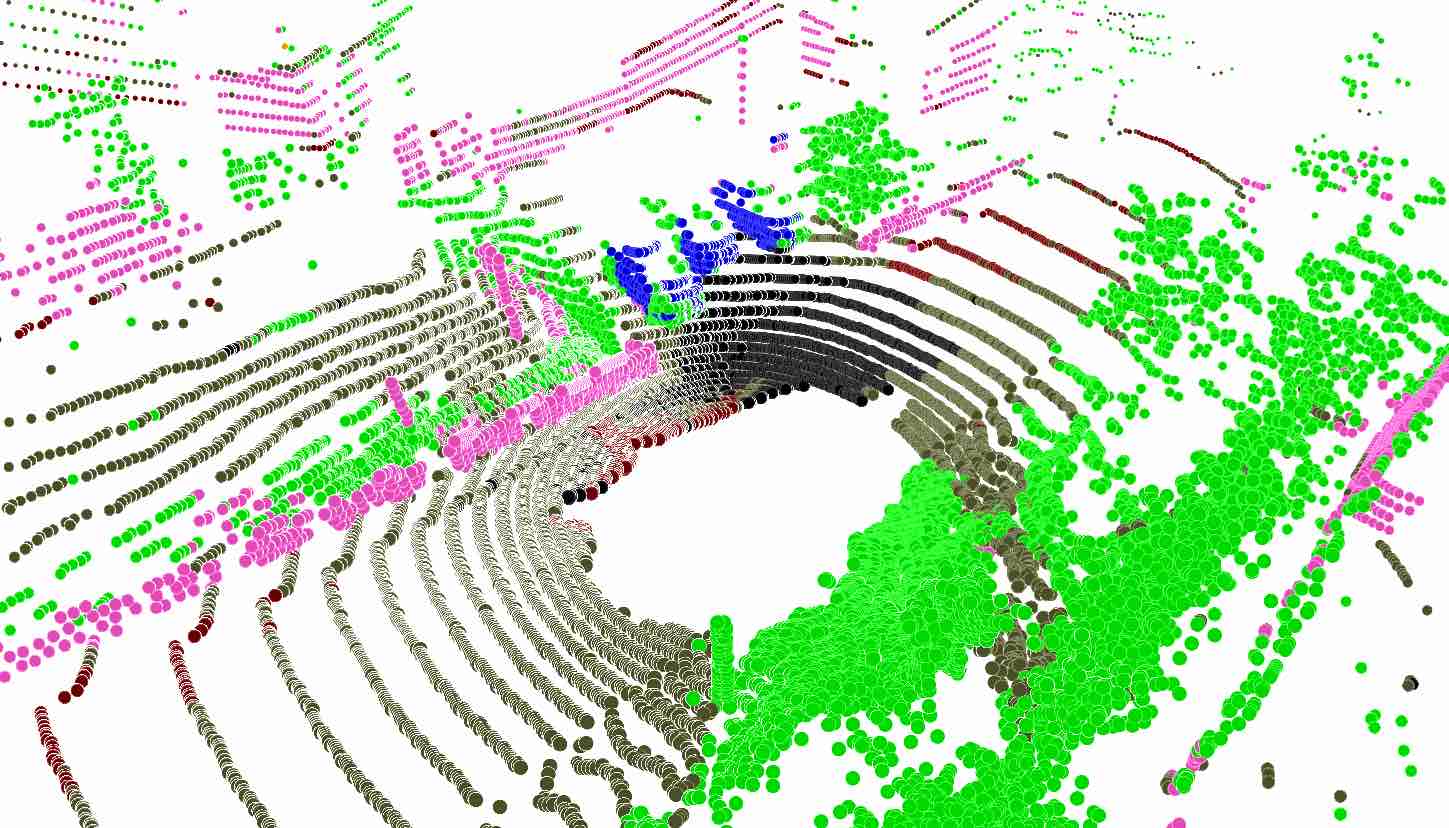}
        \end{overpic}& 
        \begin{overpic}[width=0.21\textwidth]{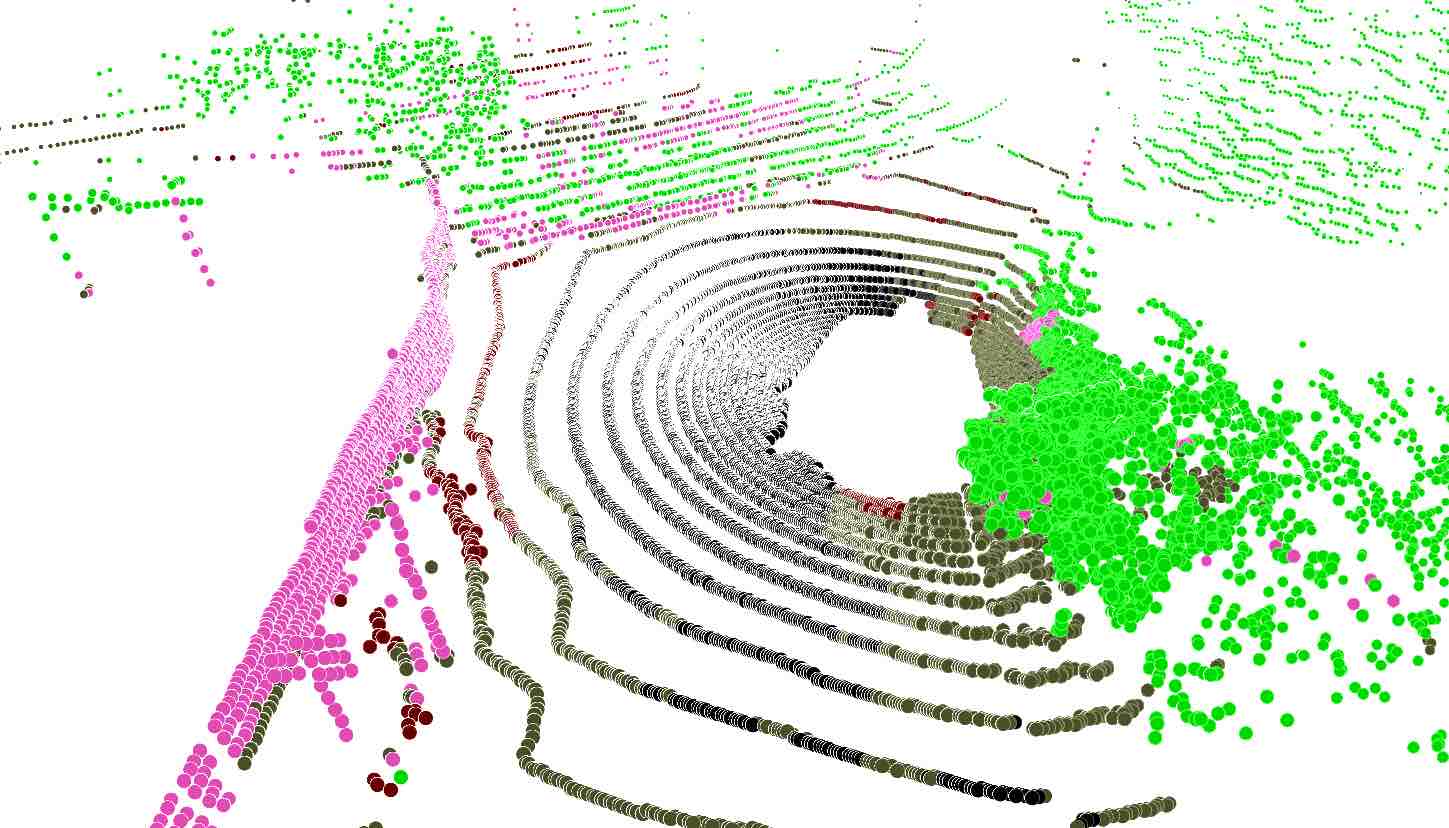}
        \end{overpic}\\
        \begin{overpic}[width=0.21\textwidth]{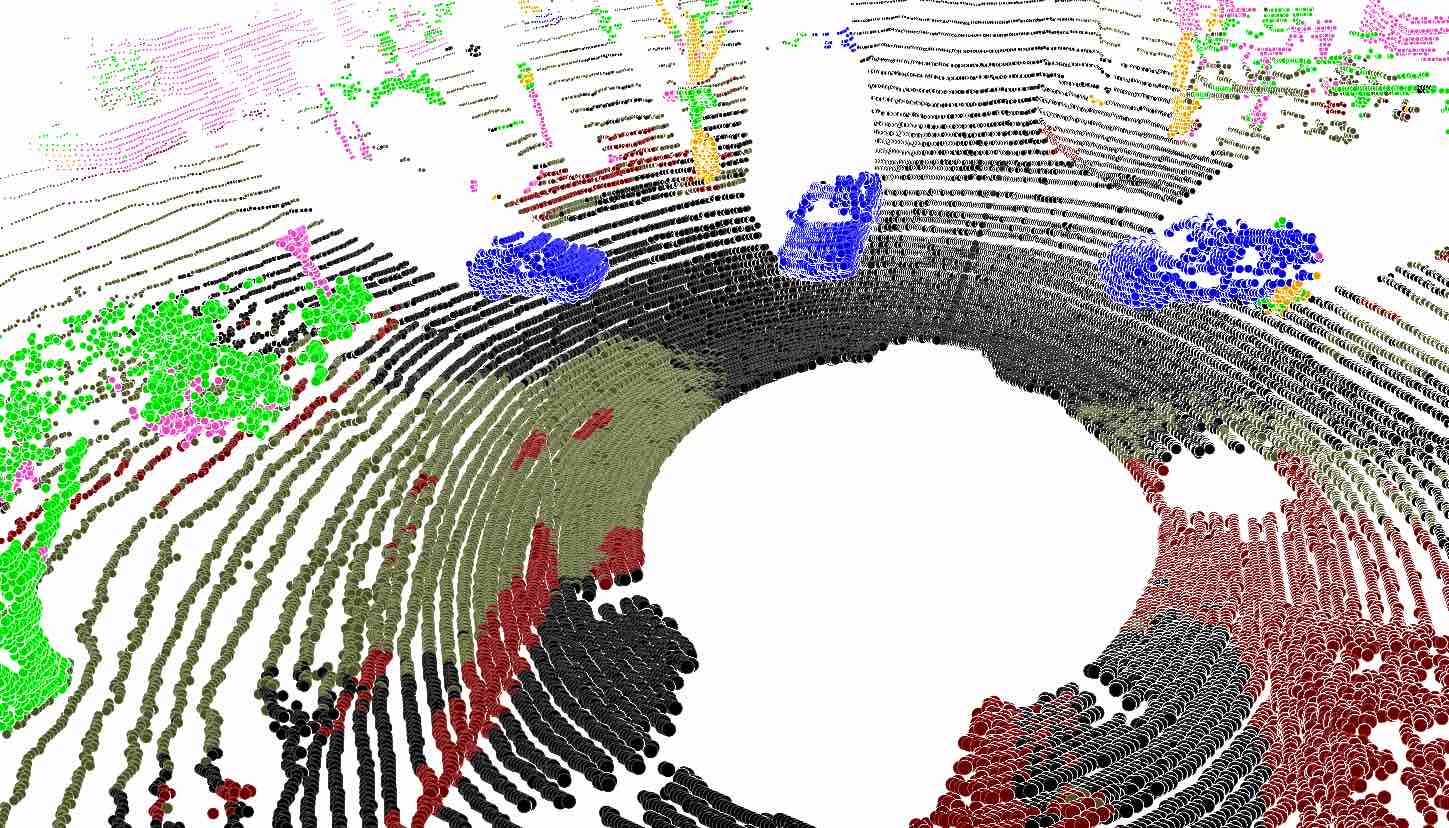}
        \end{overpic} &  
        \begin{overpic}[width=0.21\textwidth]{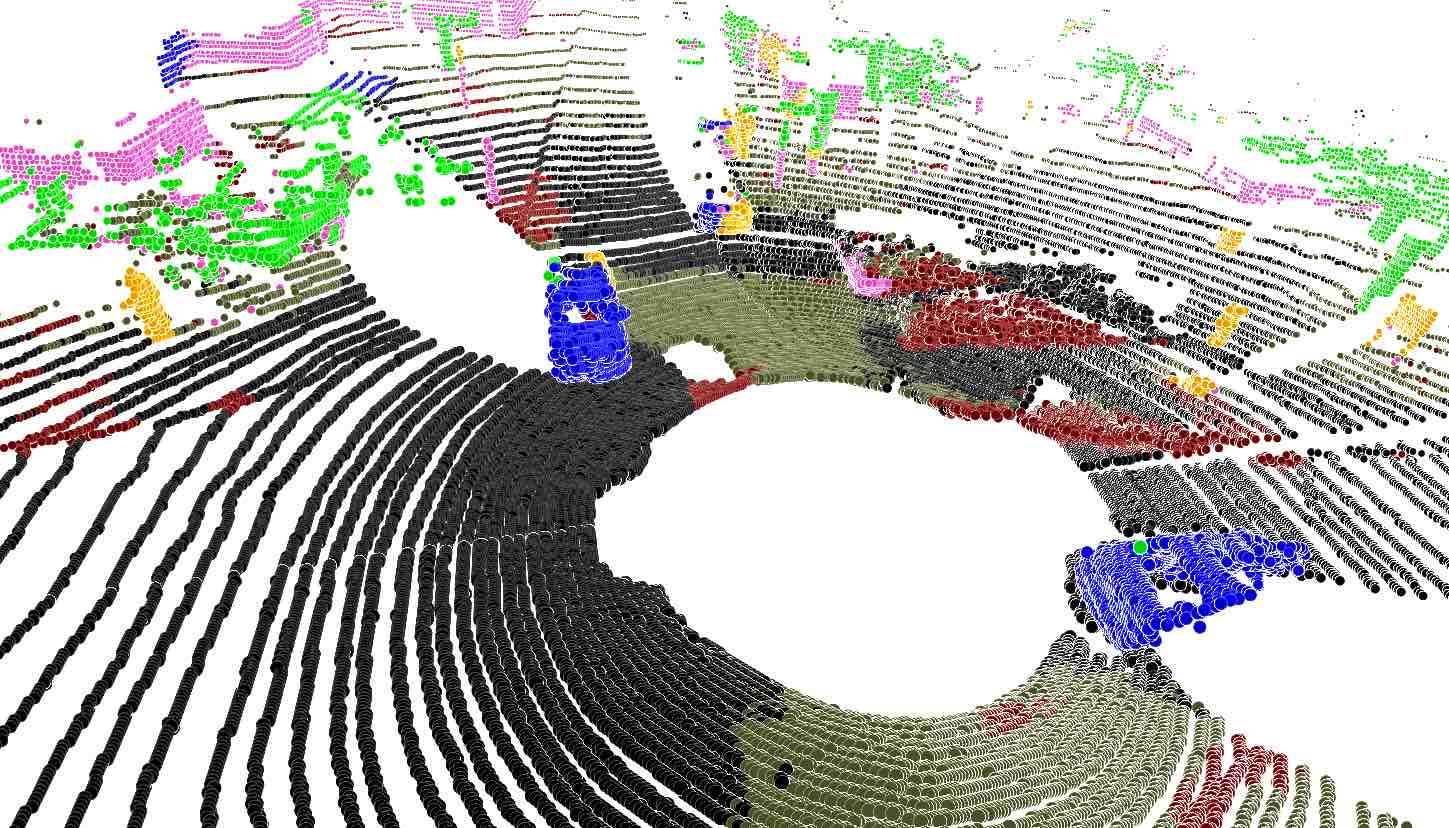}
        \end{overpic} &
        \begin{overpic}[width=0.21\textwidth]{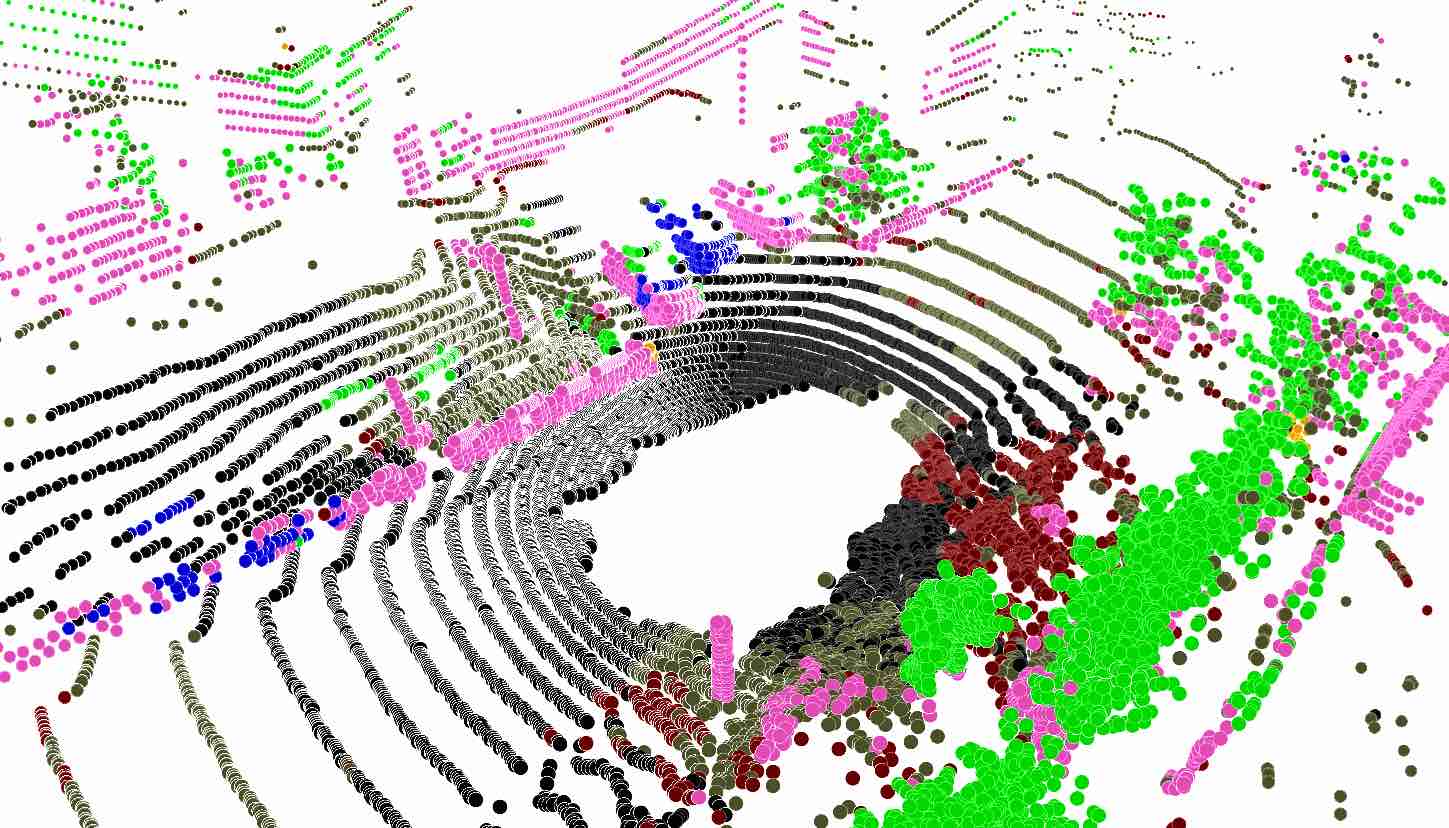}
        \end{overpic}& 
        \begin{overpic}[width=0.21\textwidth]{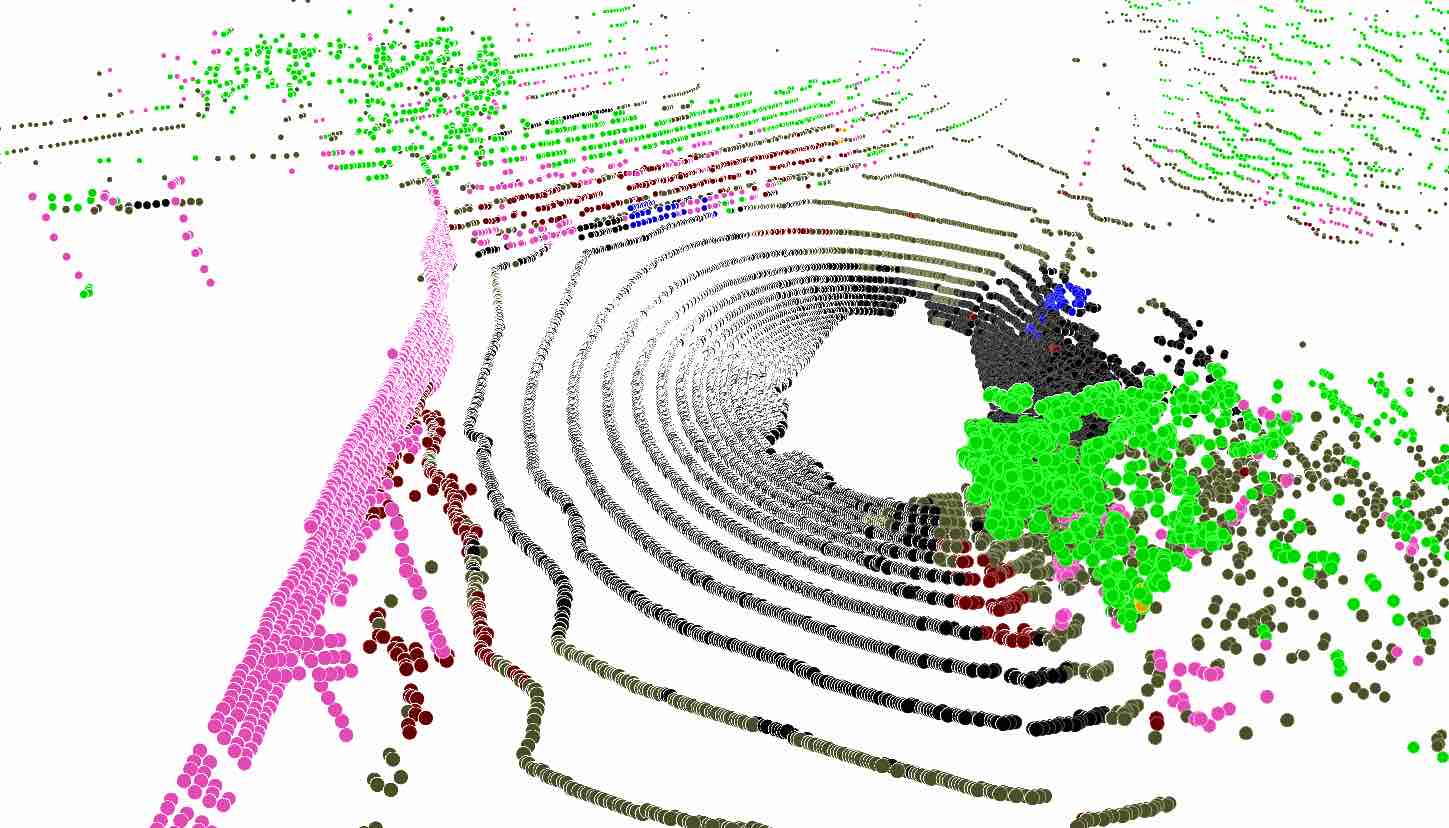}
        \end{overpic}\\
        \begin{overpic}[width=0.21\textwidth]{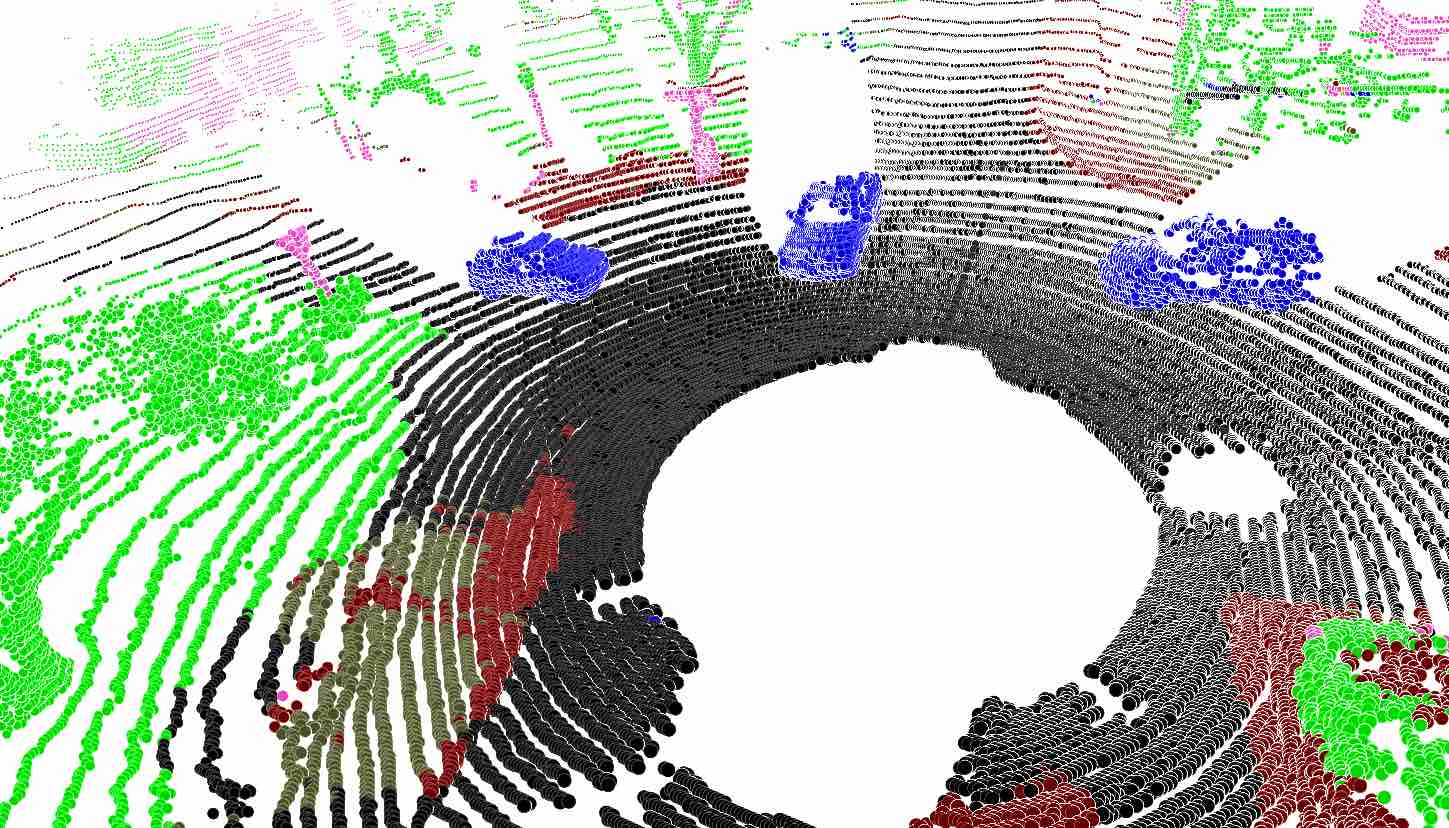}
        \put(-4.5,500){\rotatebox{90}{\color{black}\footnotesize \textbf{source}}}
        \put(-6,440){\rotatebox{90}{\color{black}\footnotesize \textbf{mix3D}}}
        \put(-6,375){\rotatebox{90}{\color{black}\footnotesize \textbf{p.cutmix}}}
        \put(-6,317){\rotatebox{90}{\color{black}\footnotesize \textbf{cosmix}}}
        \put(-6,260){\rotatebox{90}{\color{black}\footnotesize \textbf{ibn}}}
        \put(-6,200){\rotatebox{90}{\color{black}\footnotesize \textbf{robust.}}}
        \put(-5,140){\rotatebox{90}{\color{black}\footnotesize \textbf{sn}}}
        \put(-6,75){\rotatebox{90}{\color{black}\footnotesize \textbf{raycast}}}
        \put(-5,17){\rotatebox{90}{\color{black}\footnotesize \textbf{ours}}}
        \put(-6,-35){\rotatebox{90}{\color{black}\footnotesize \textbf{gt}}}
        \put(80,541){\color{black}\footnotesize \textbf{SemanticKITTI}}
        \put(290,541){\color{black}\footnotesize \textbf{nuScenes}}
        \end{overpic} &  
        \begin{overpic}[width=0.21\textwidth]{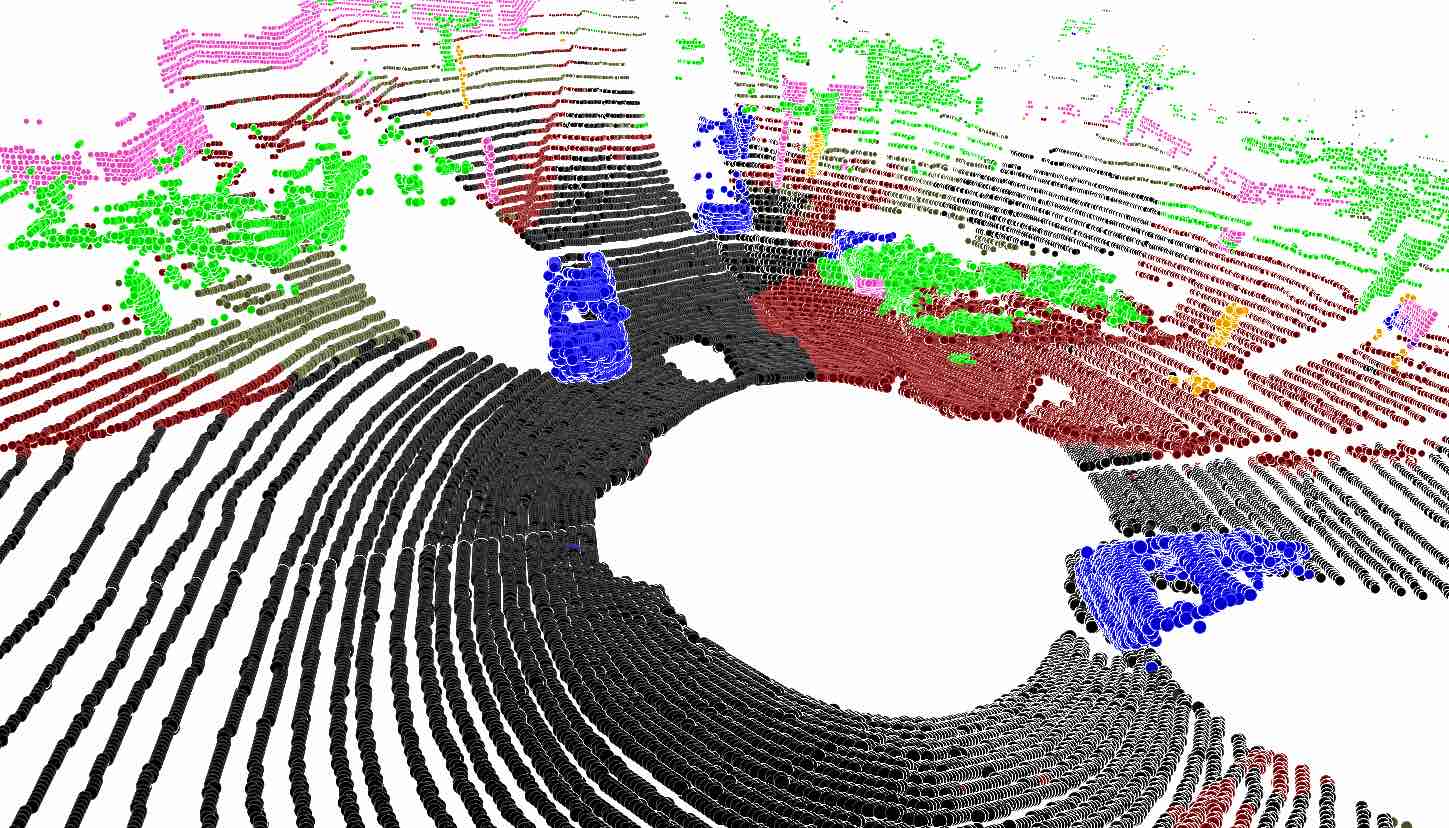}
        \end{overpic} &
        \begin{overpic}[width=0.21\textwidth]{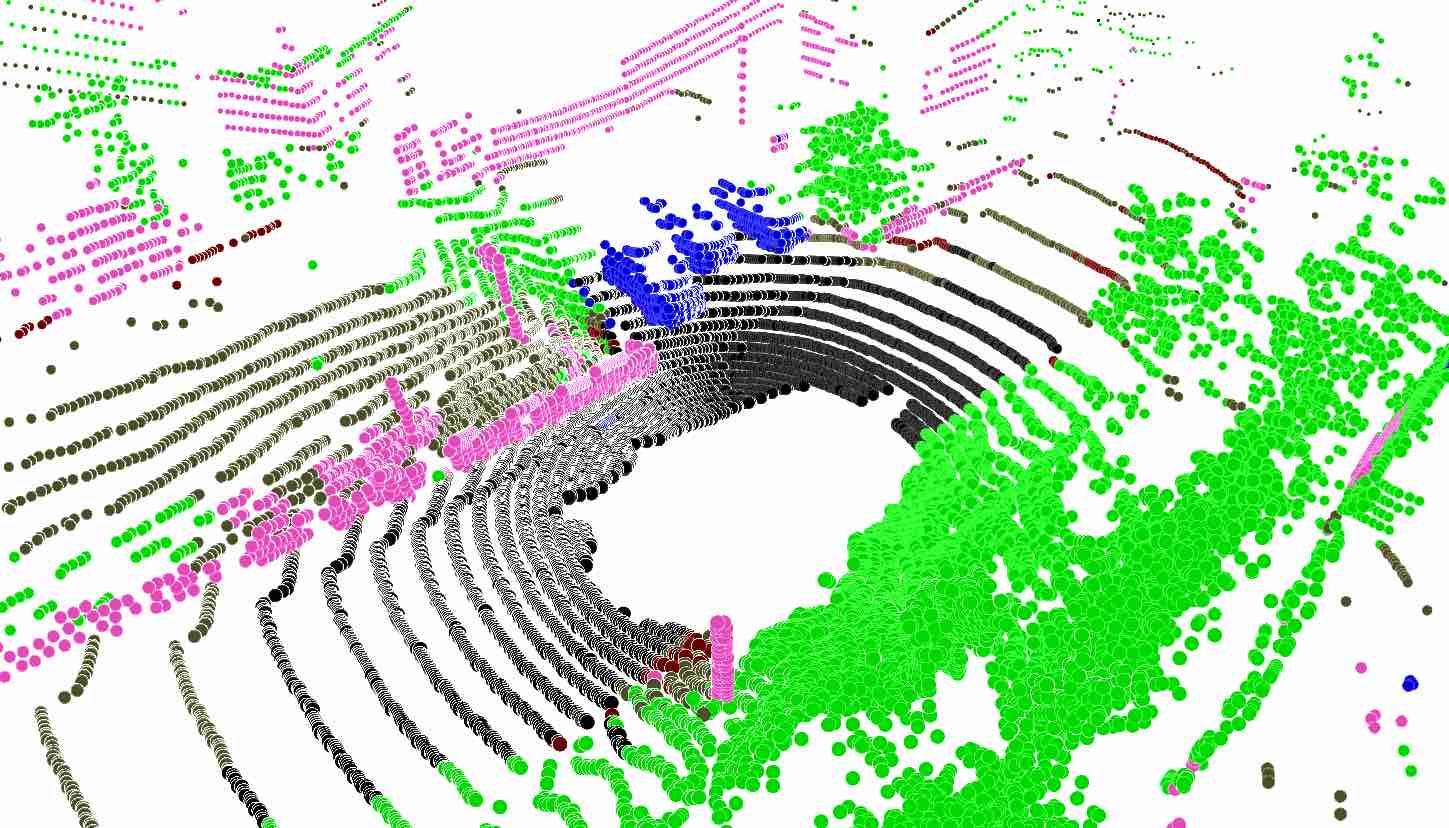}
        \end{overpic}& 
        \begin{overpic}[width=0.21\textwidth]{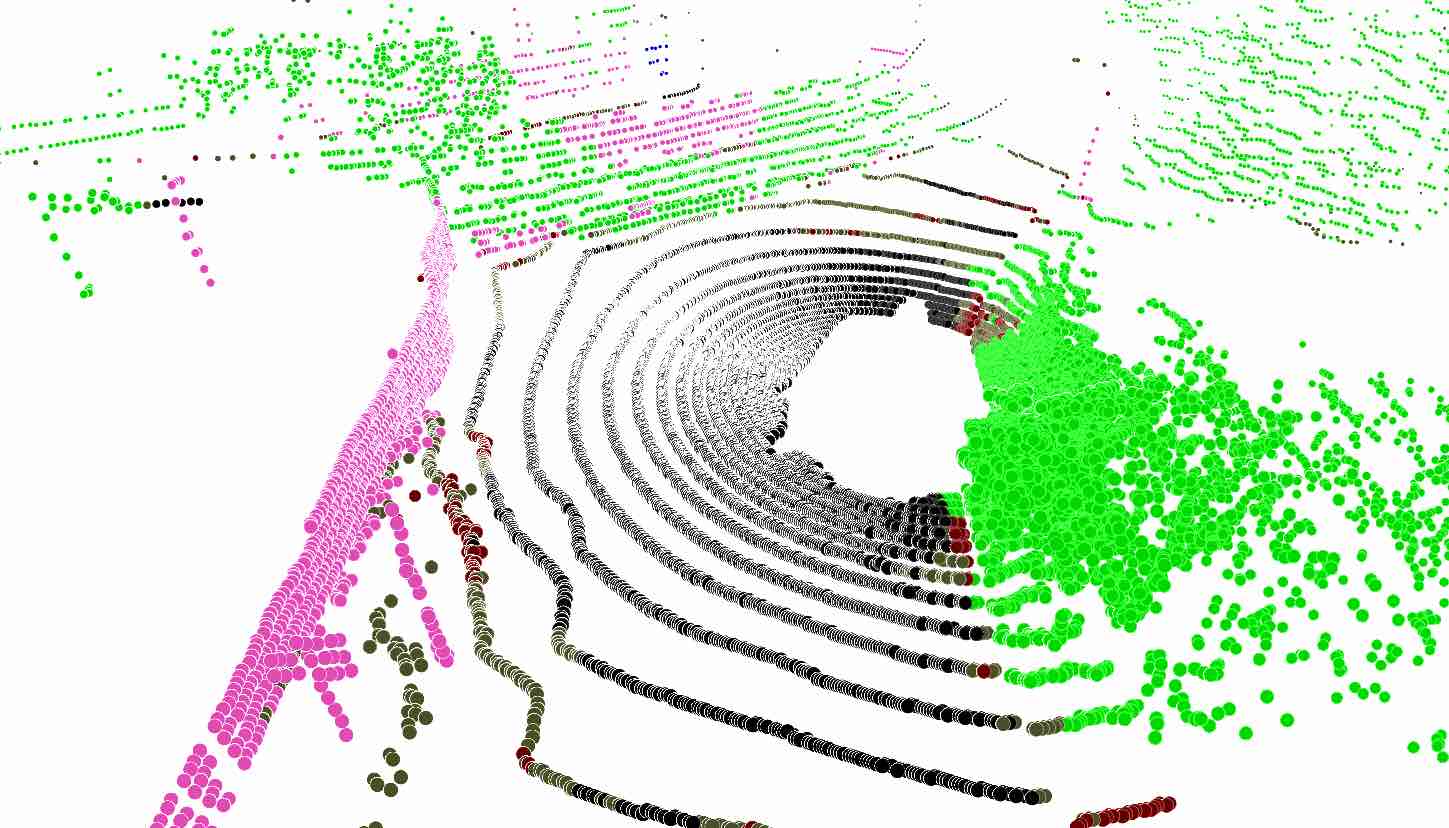}
        \end{overpic}\\
        \begin{overpic}[width=0.21\textwidth]{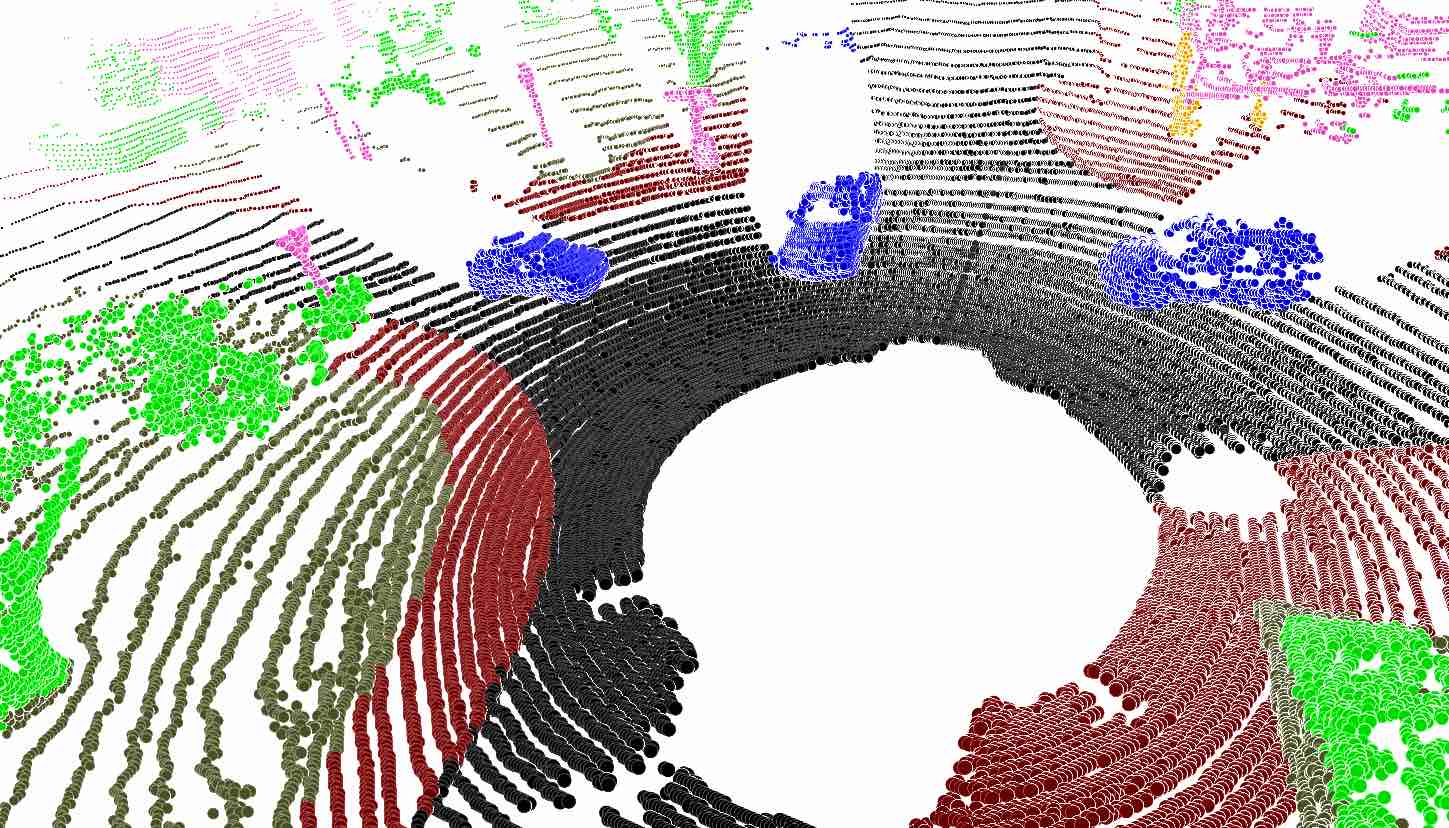}
        \end{overpic} &  
        \begin{overpic}[width=0.21\textwidth]{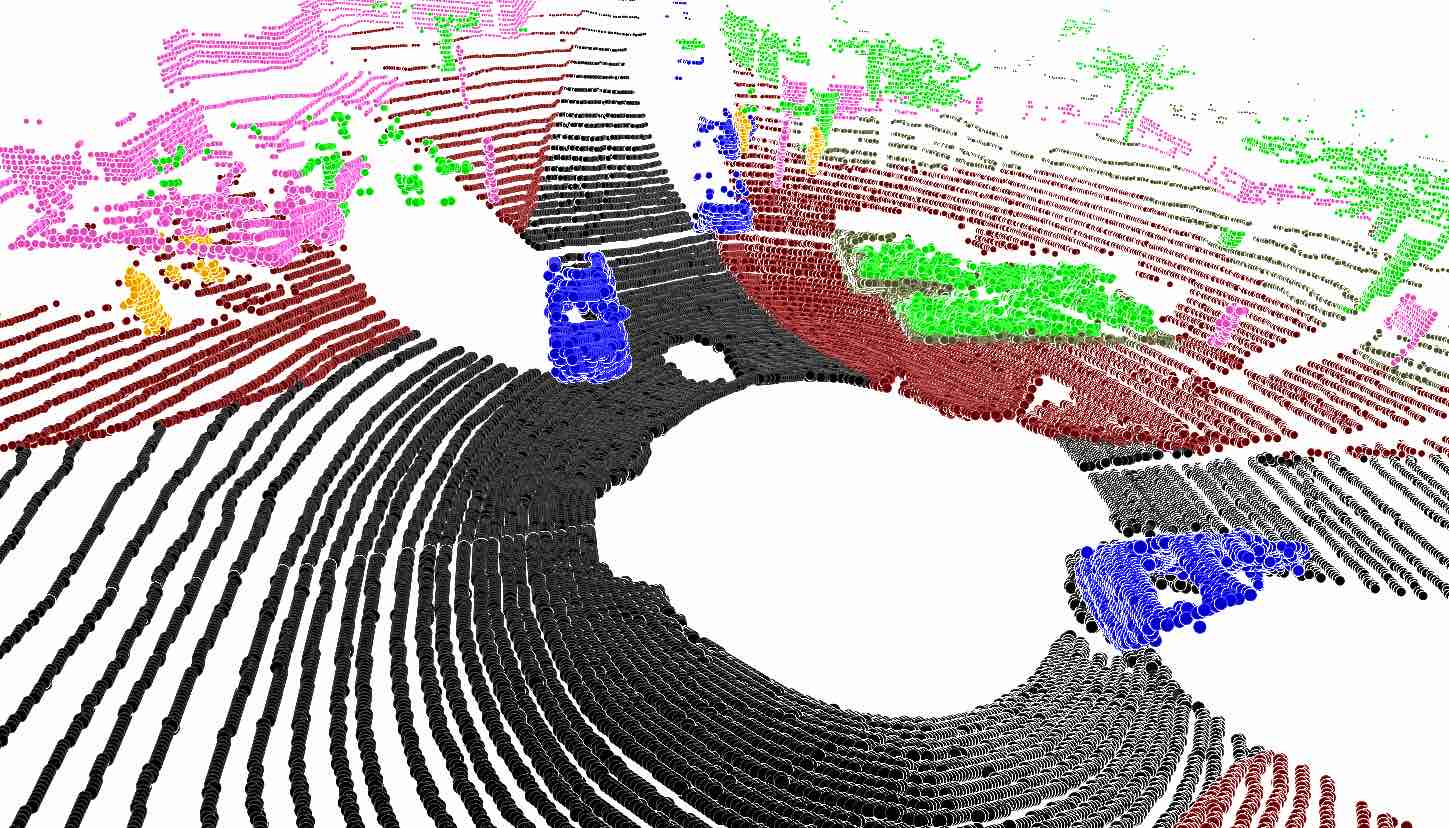}
        \end{overpic} &
        \begin{overpic}[width=0.21\textwidth]{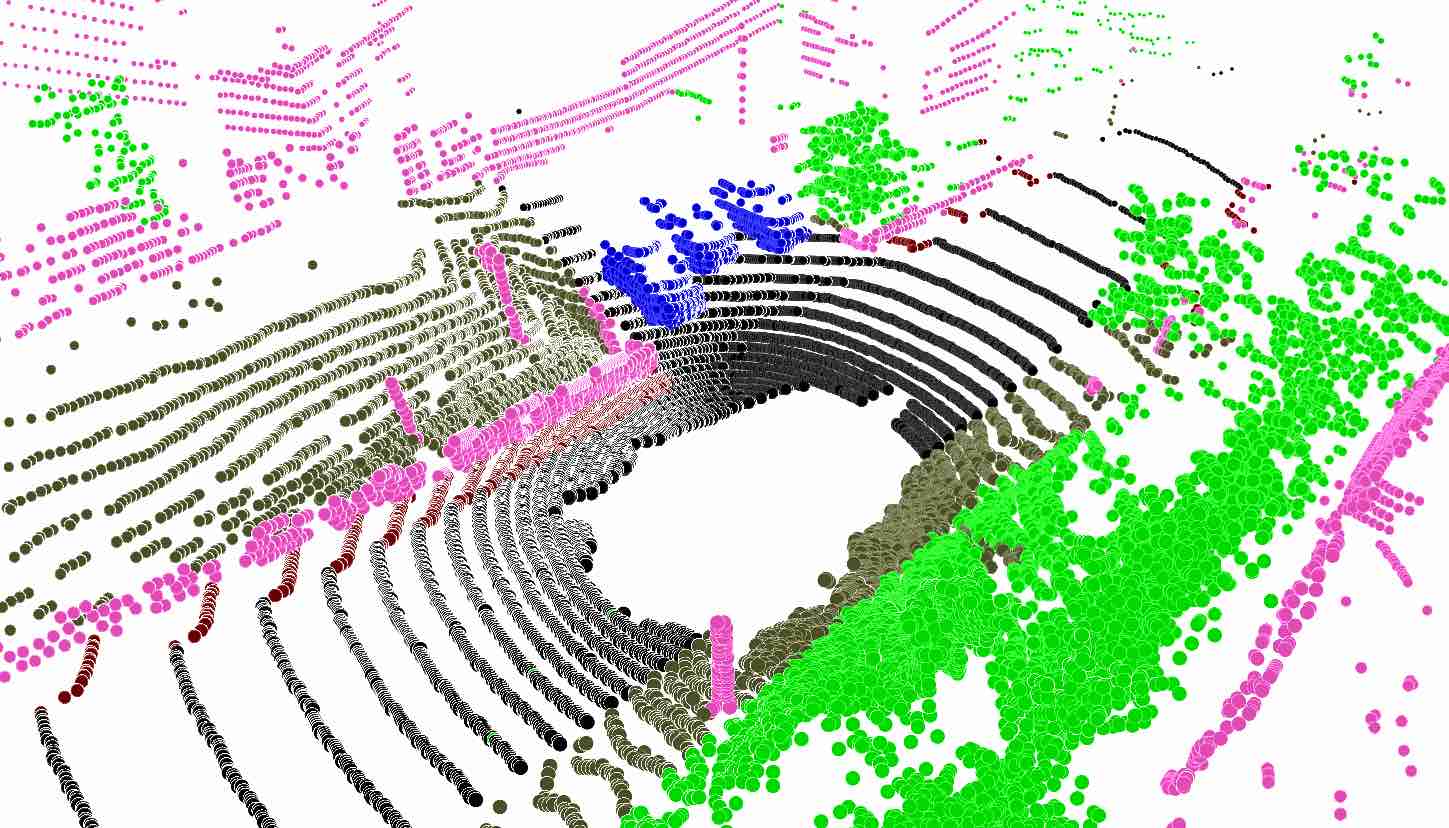}
        \end{overpic}& 
        \begin{overpic}[width=0.21\textwidth]{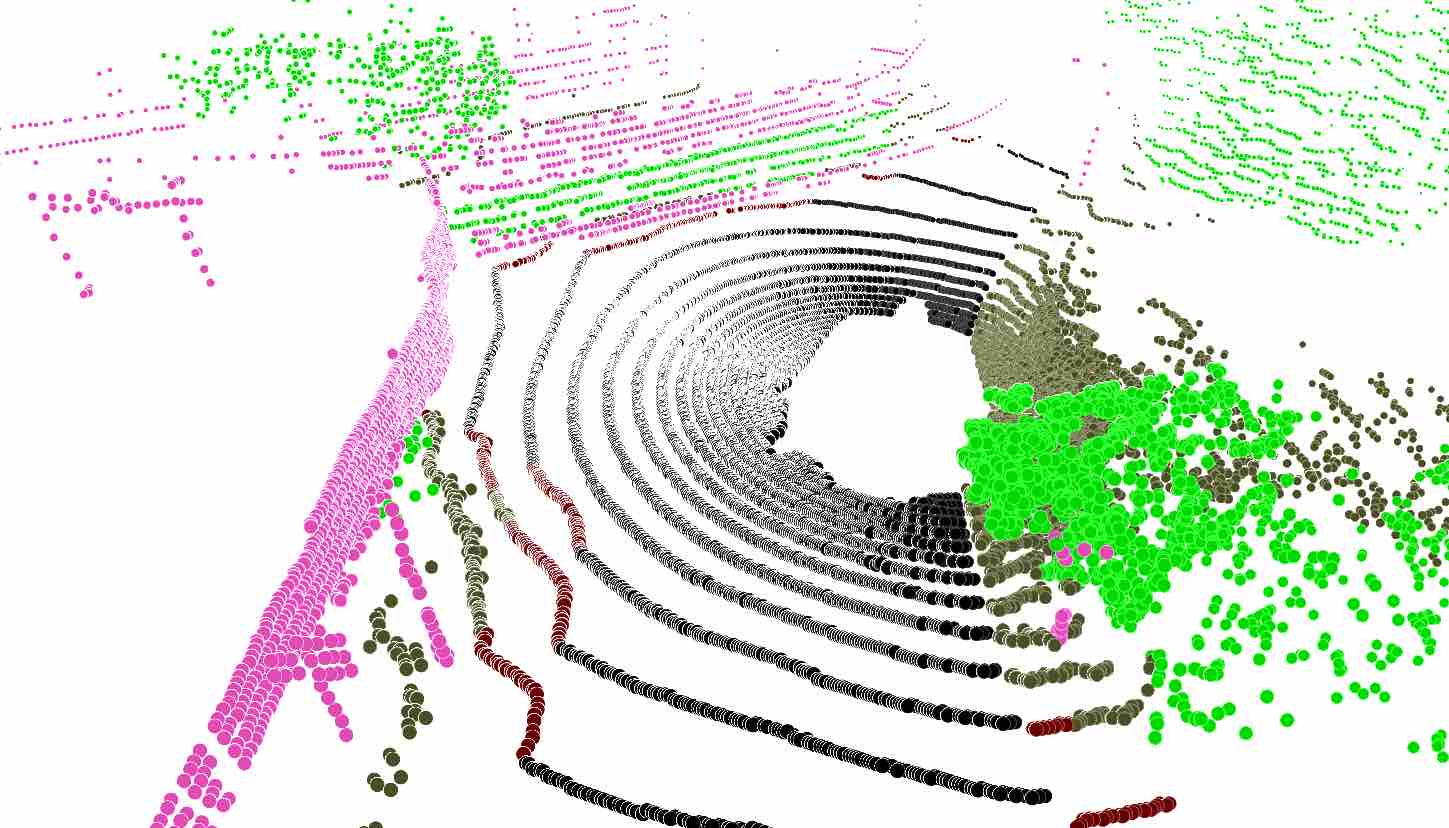}
        \end{overpic}\\
    \end{tabular}
    \vspace{-4mm}
    \caption{\textbf{Qualitative results.} \textit{Top:} Synth4D-kitti$\rightarrow$SemanticKITTI, \textit{bottom:} Synth4D-kitti$\rightarrow$nuScenes. \lidog improves over source and baselines, \eg, we observe the improvements of \textit{road} in SemanticKITTI and \textit{vehicle} in nuScenes.}
    \label{fig:supp_qualitative_synth4dkitti}
\end{figure*}

\begin{figure*}[t]
\centering
    \setlength\tabcolsep{1.pt}
    \begin{tabular}{cccc}
    \raggedright
        \begin{overpic}[width=0.21\textwidth]{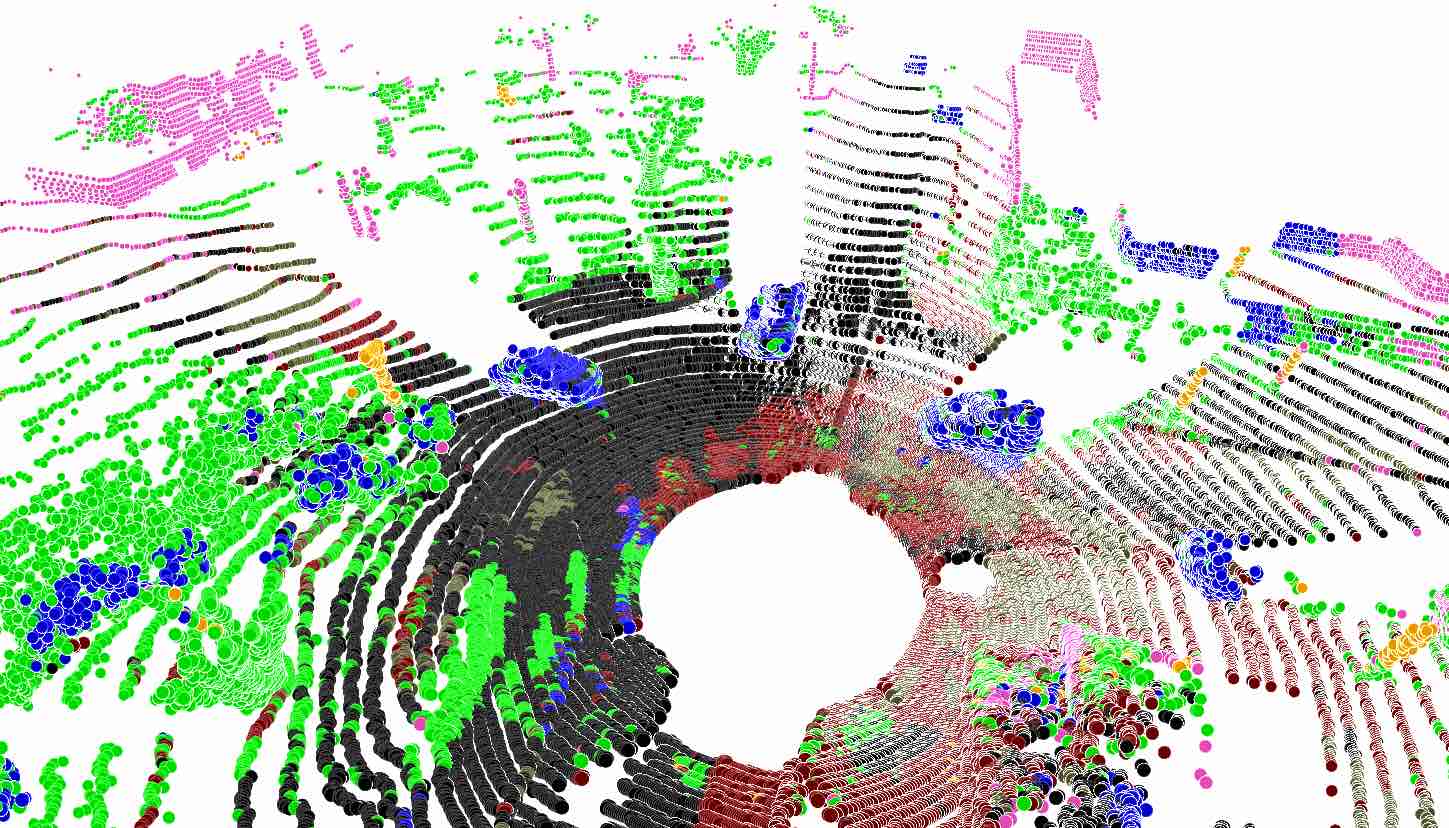}
        \end{overpic} &  
        \begin{overpic}[width=0.21\textwidth]{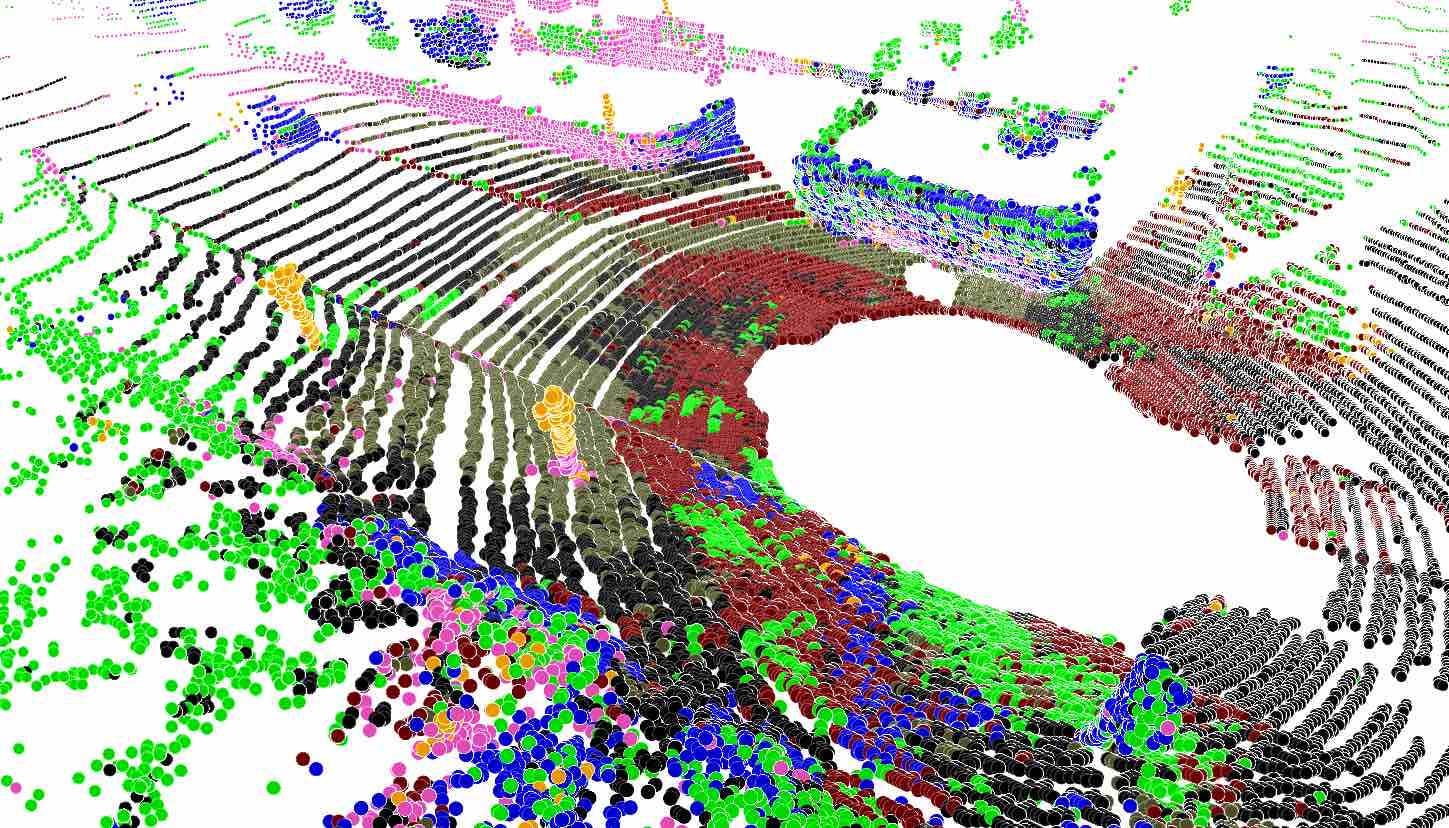}
        \end{overpic} &
        \begin{overpic}[width=0.21\textwidth]{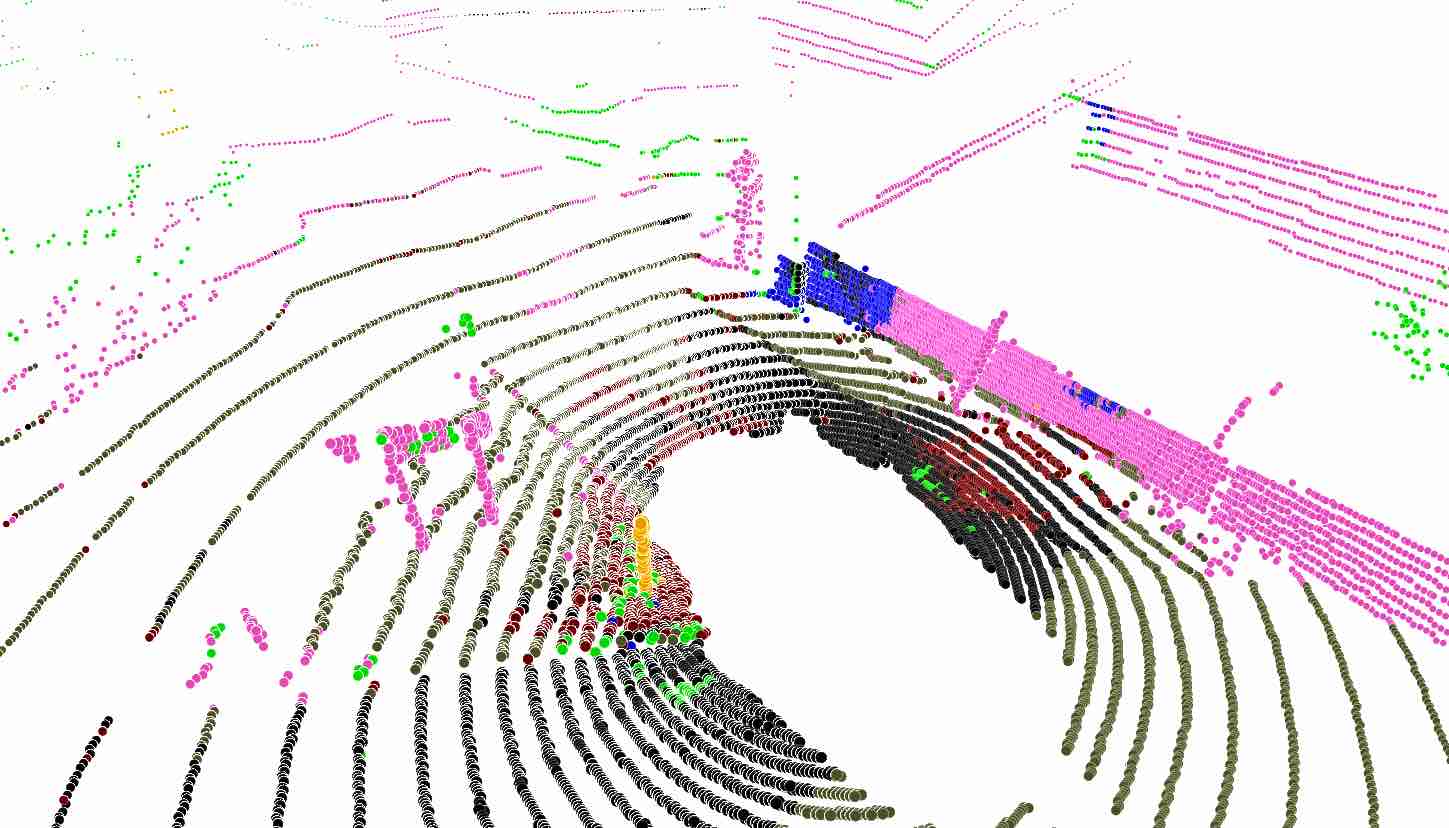}
        \end{overpic}& 
        \begin{overpic}[width=0.21\textwidth]{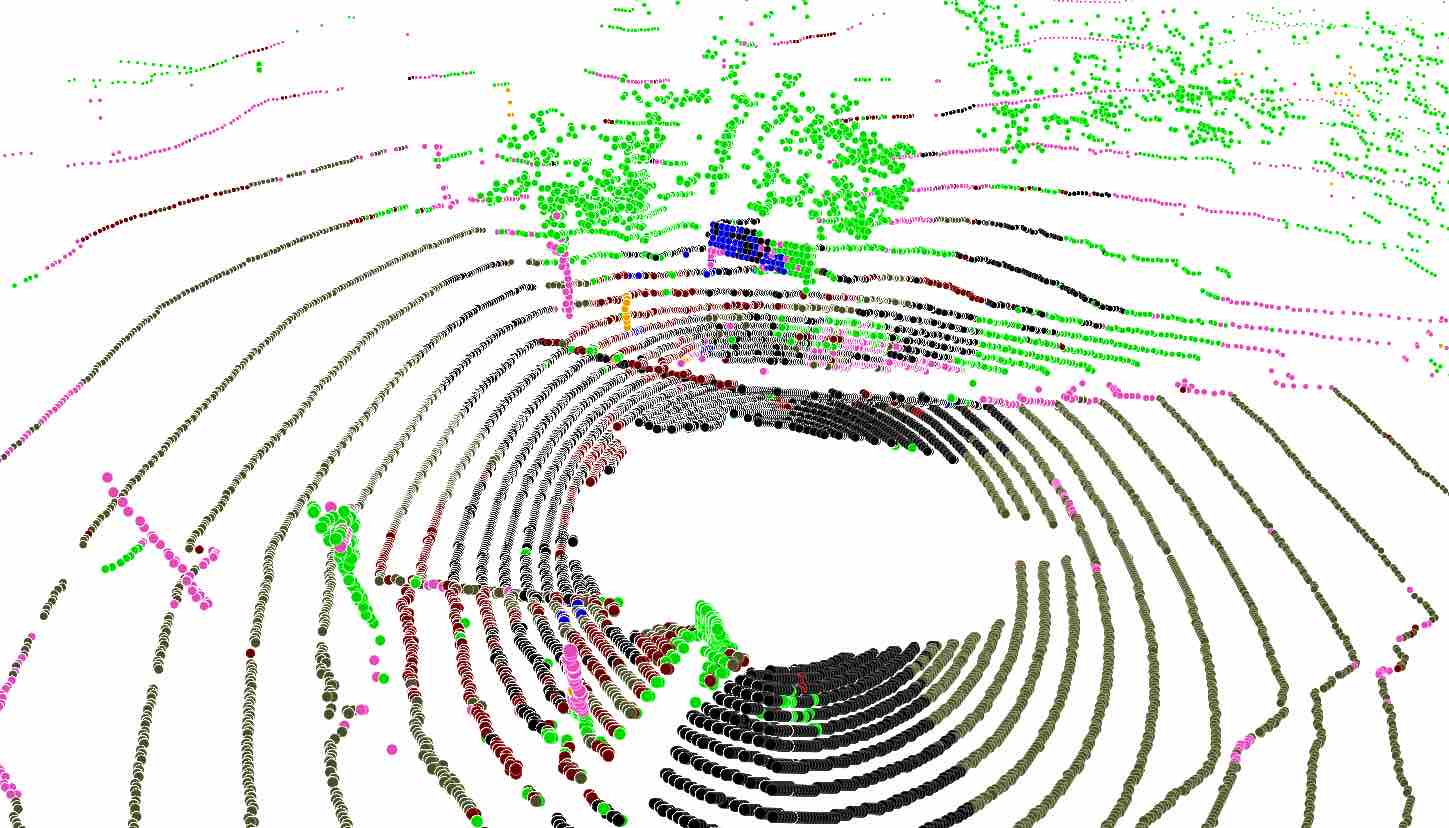}
        \end{overpic}\\
        \begin{overpic}[width=0.21\textwidth]{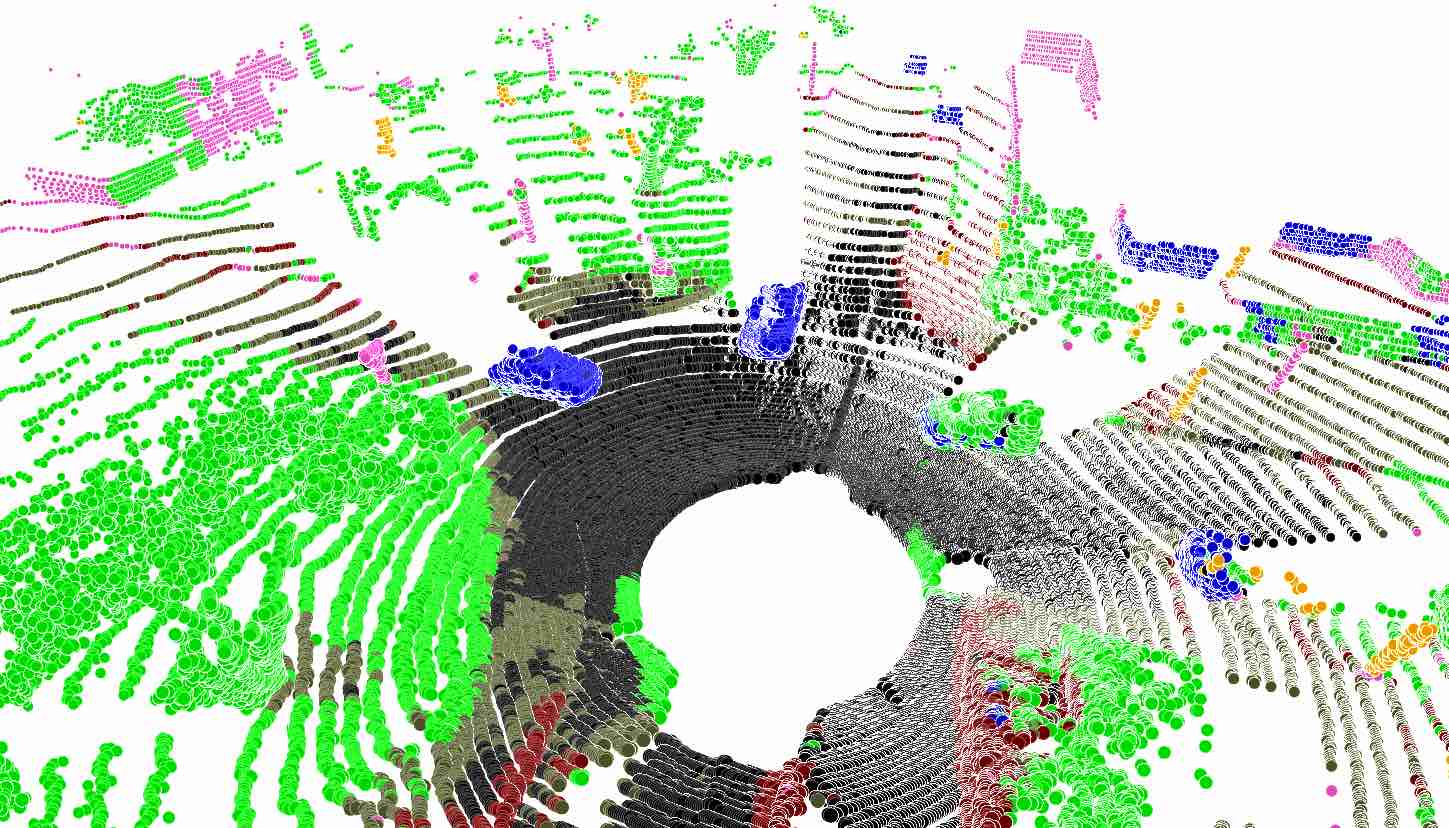}
        \end{overpic} &  
        \begin{overpic}[width=0.21\textwidth]{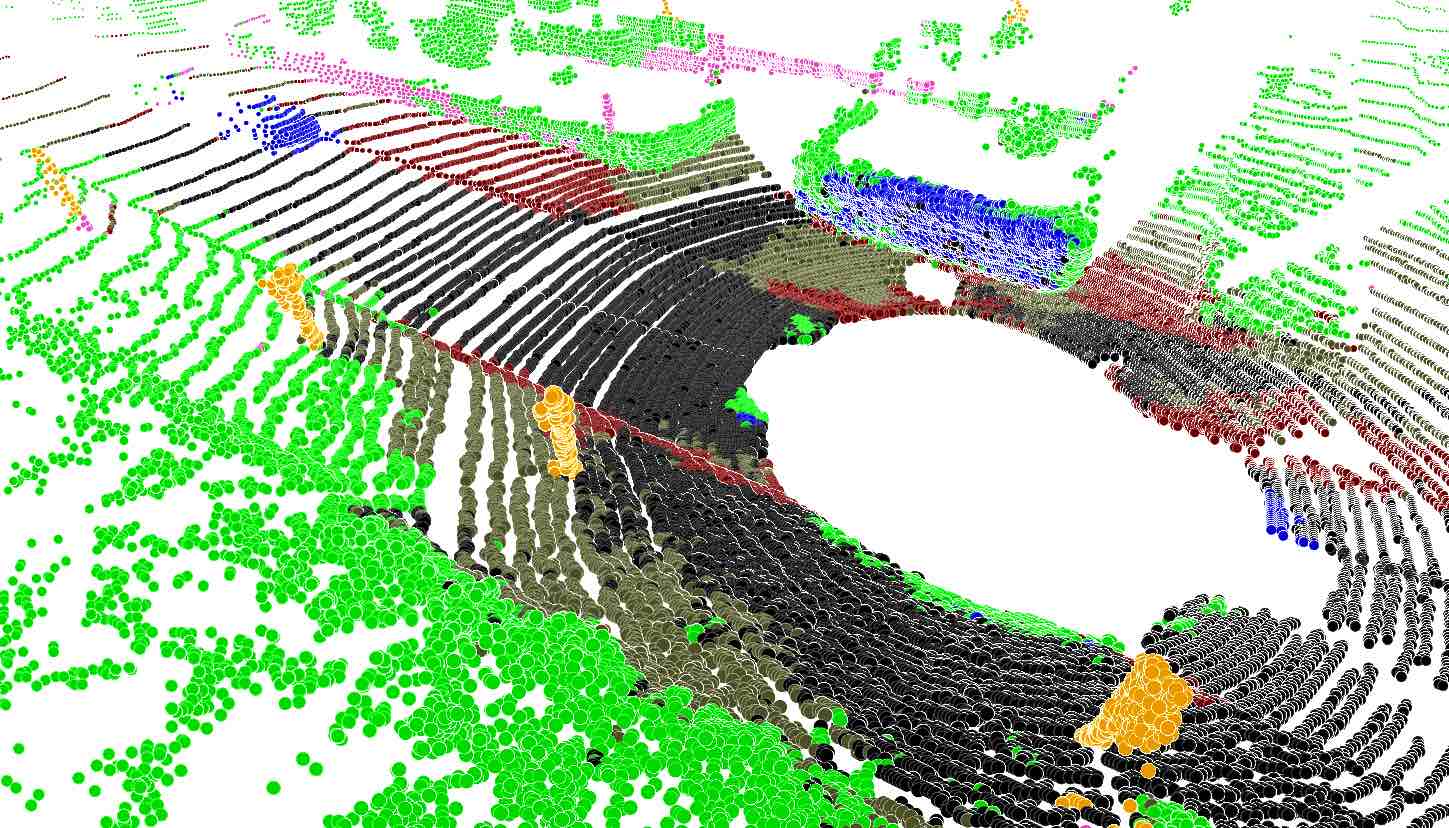}
        \end{overpic} &
        \begin{overpic}[width=0.21\textwidth]{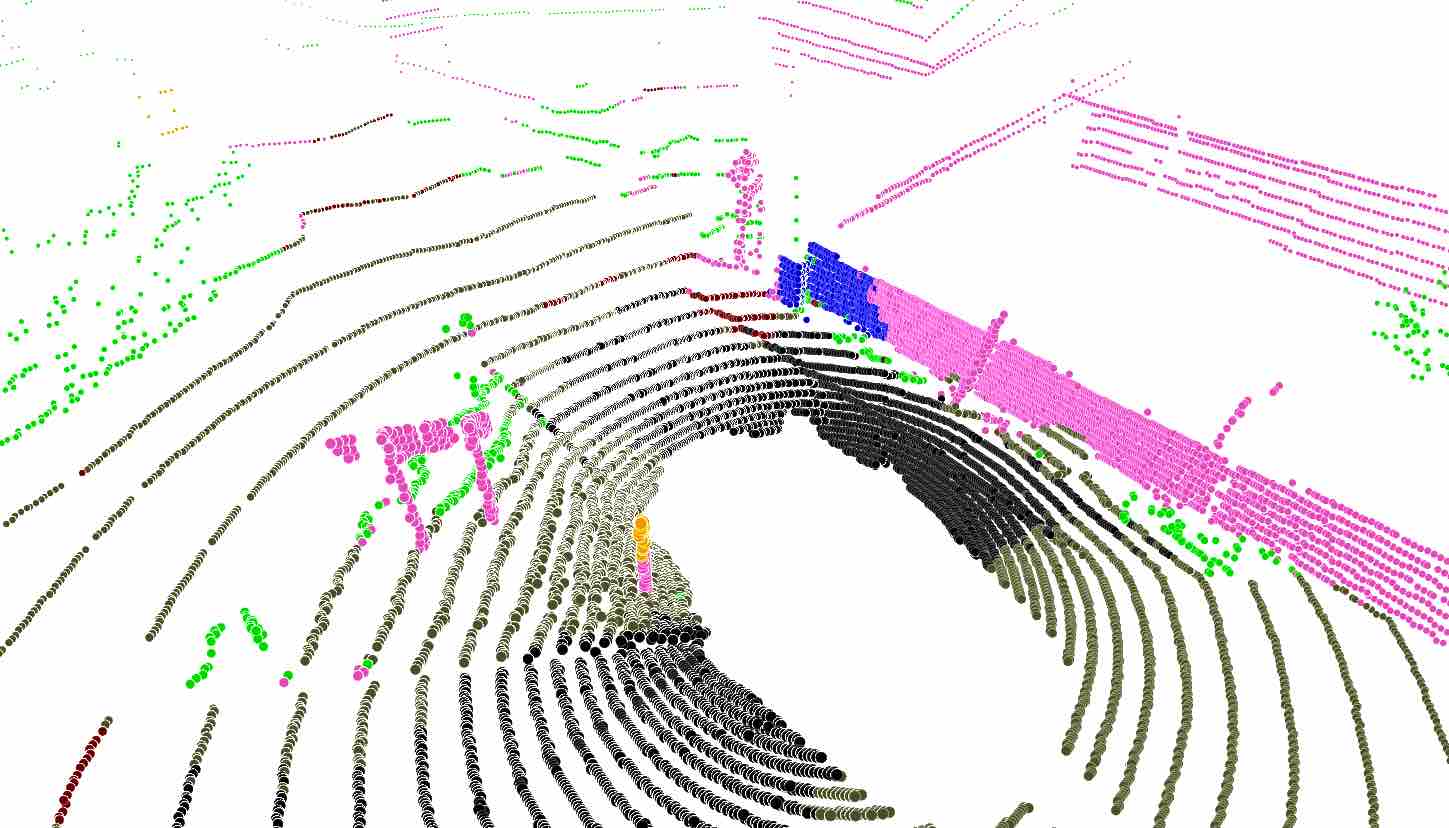}
        \end{overpic}& 
        \begin{overpic}[width=0.21\textwidth]{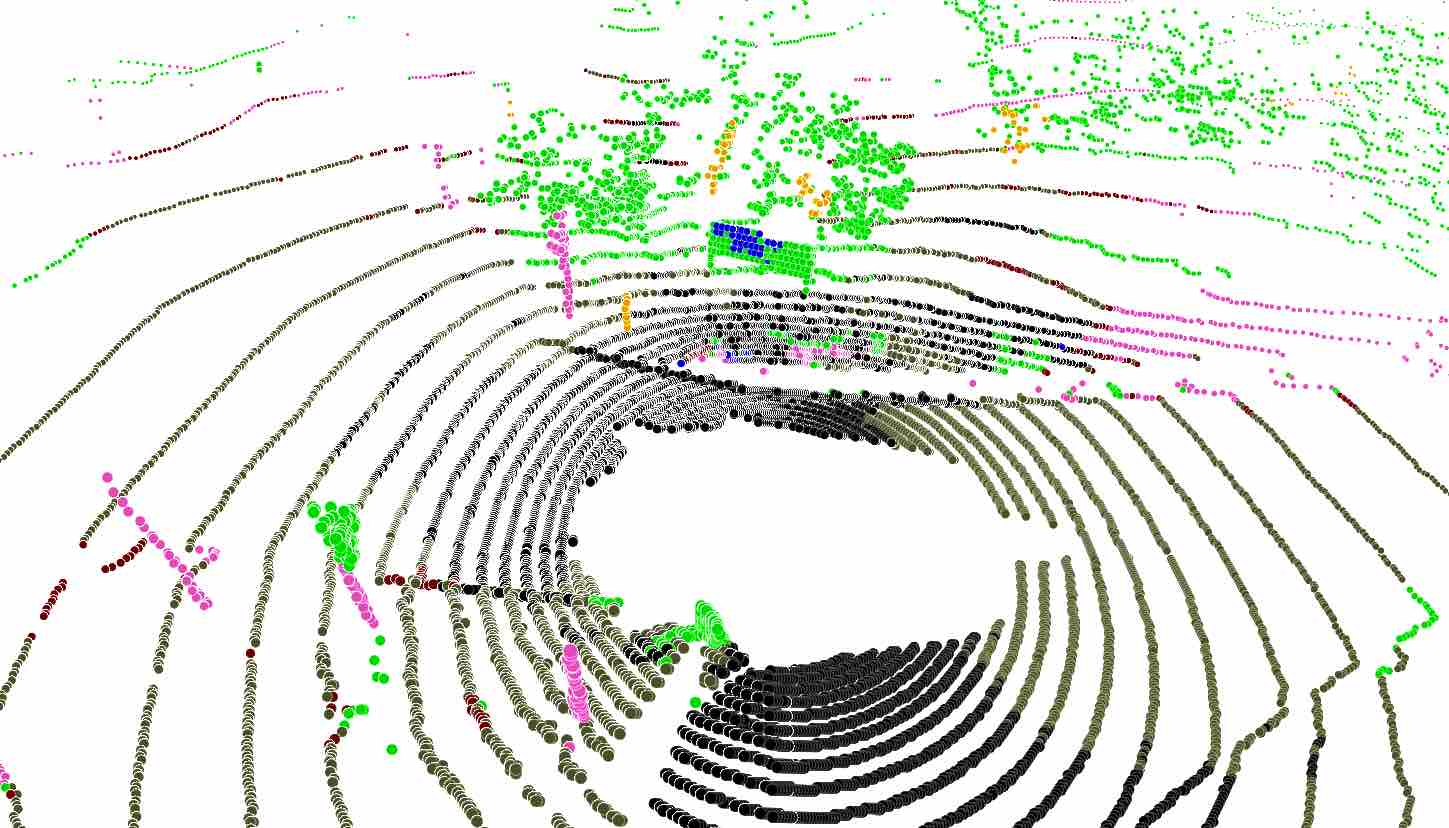}
        \end{overpic}\\
        \begin{overpic}[width=0.21\textwidth]{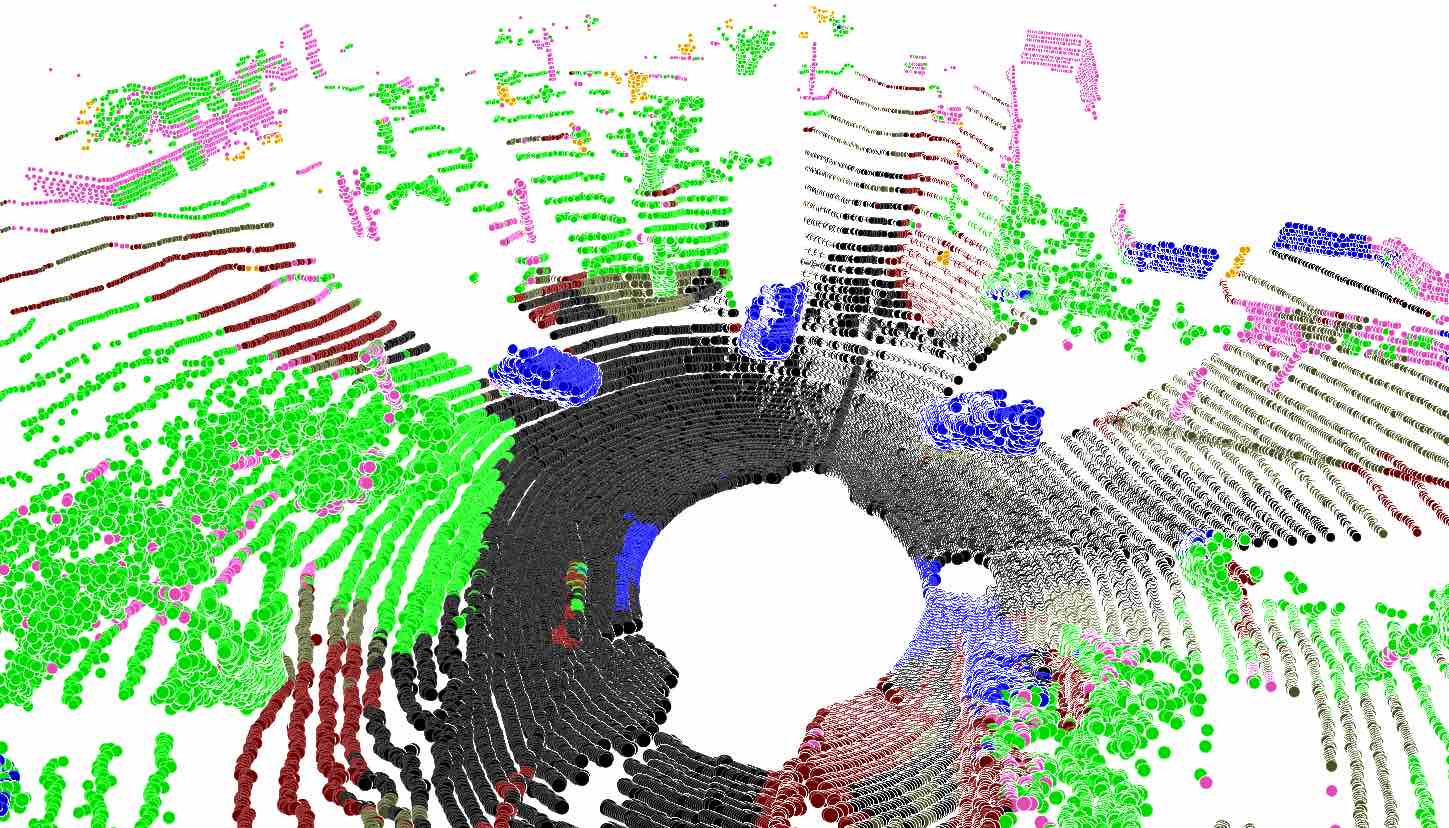}
        \end{overpic} &  
        \begin{overpic}[width=0.21\textwidth]{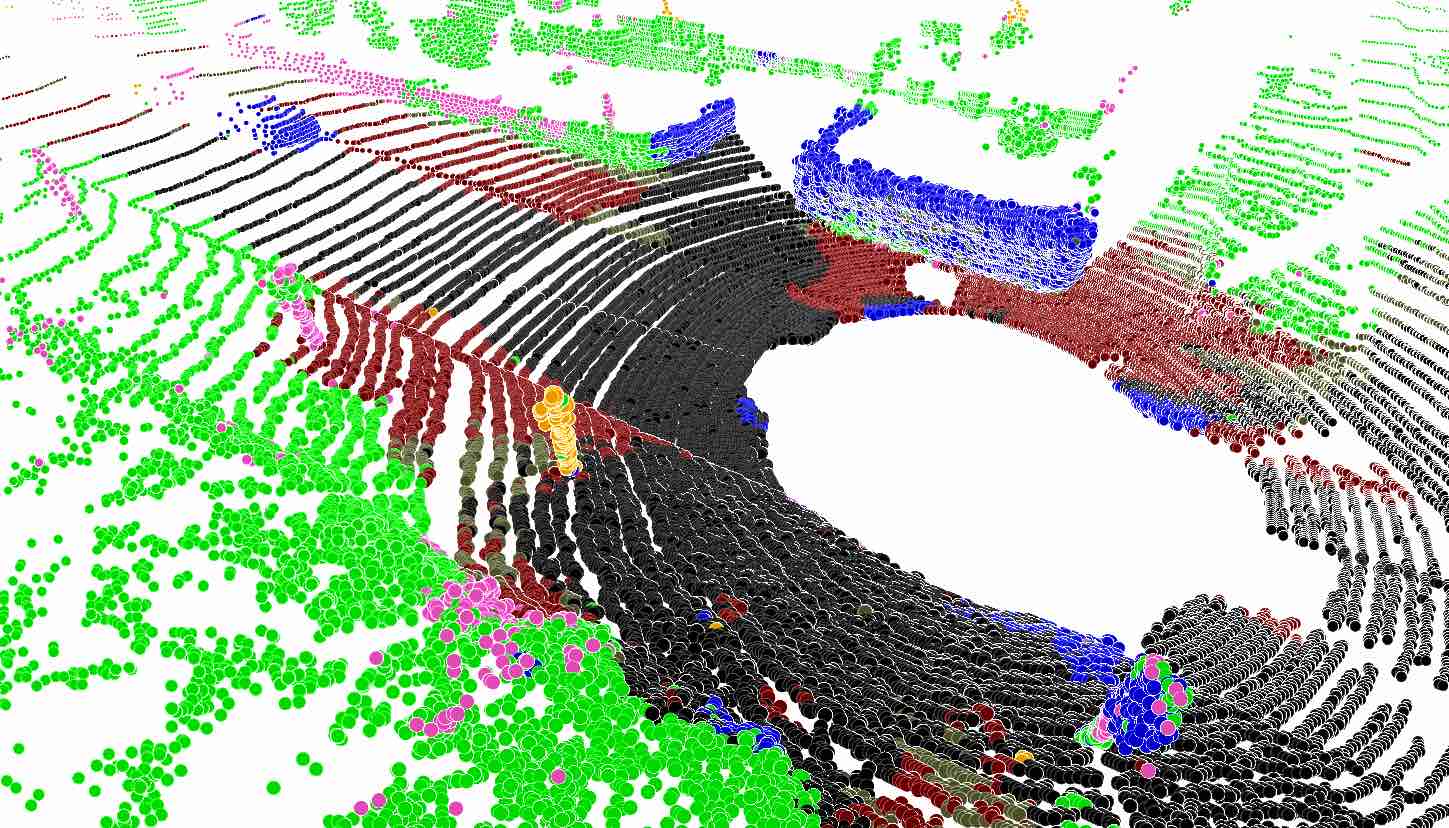}
        \end{overpic} &
        \begin{overpic}[width=0.21\textwidth]{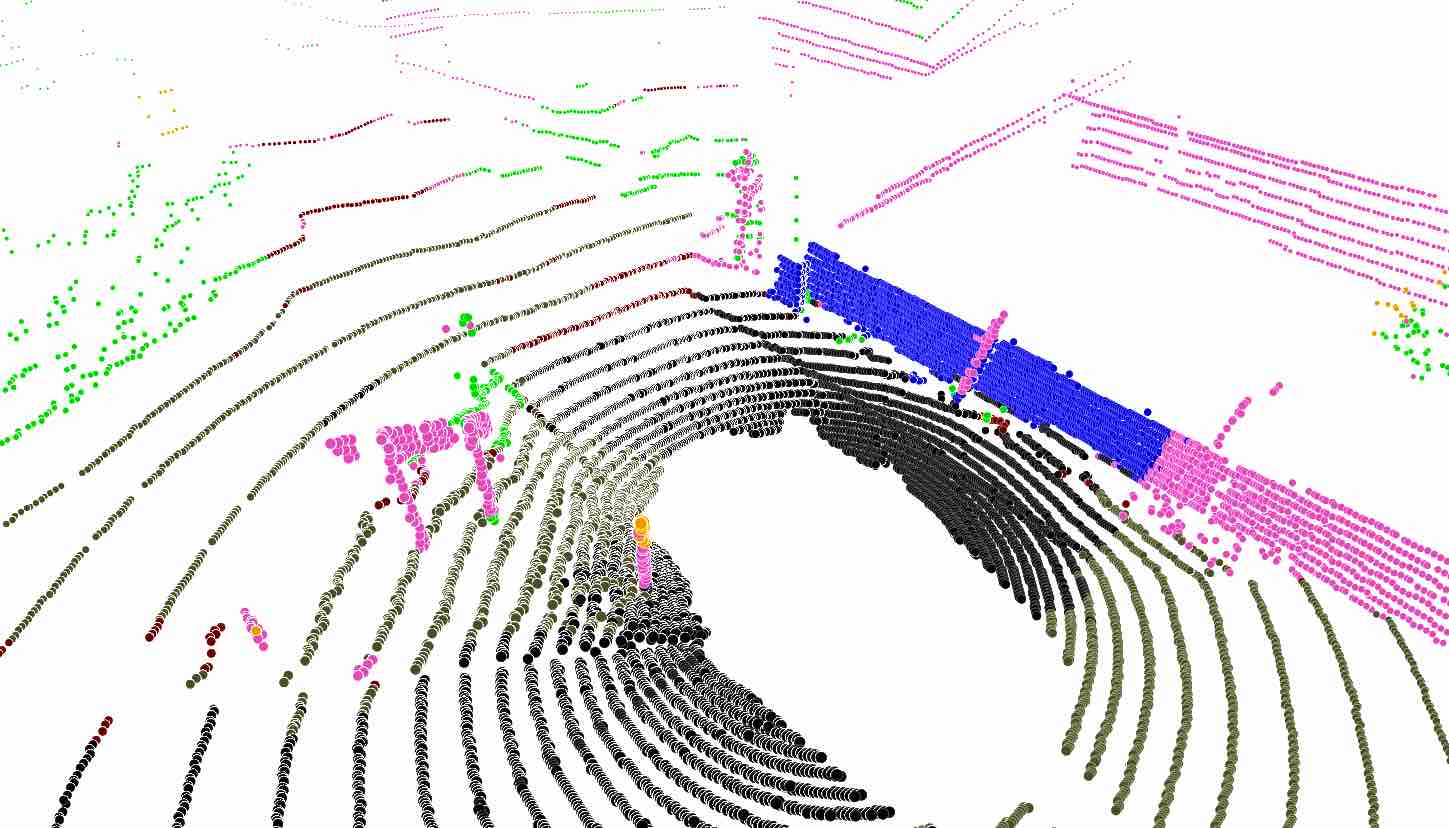}
        \end{overpic}& 
        \begin{overpic}[width=0.21\textwidth]{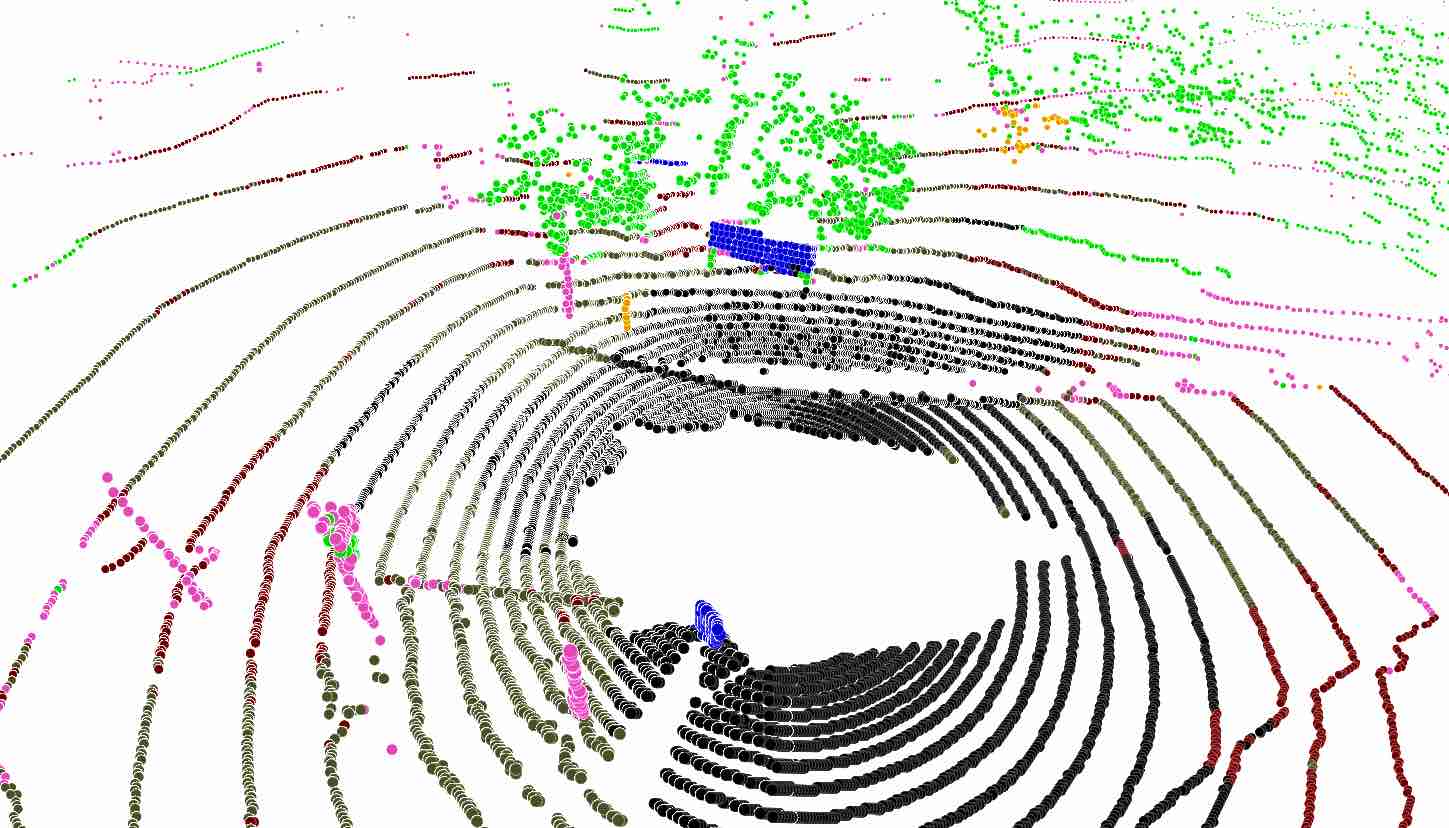}
        \end{overpic}\\
        \begin{overpic}[width=0.21\textwidth]{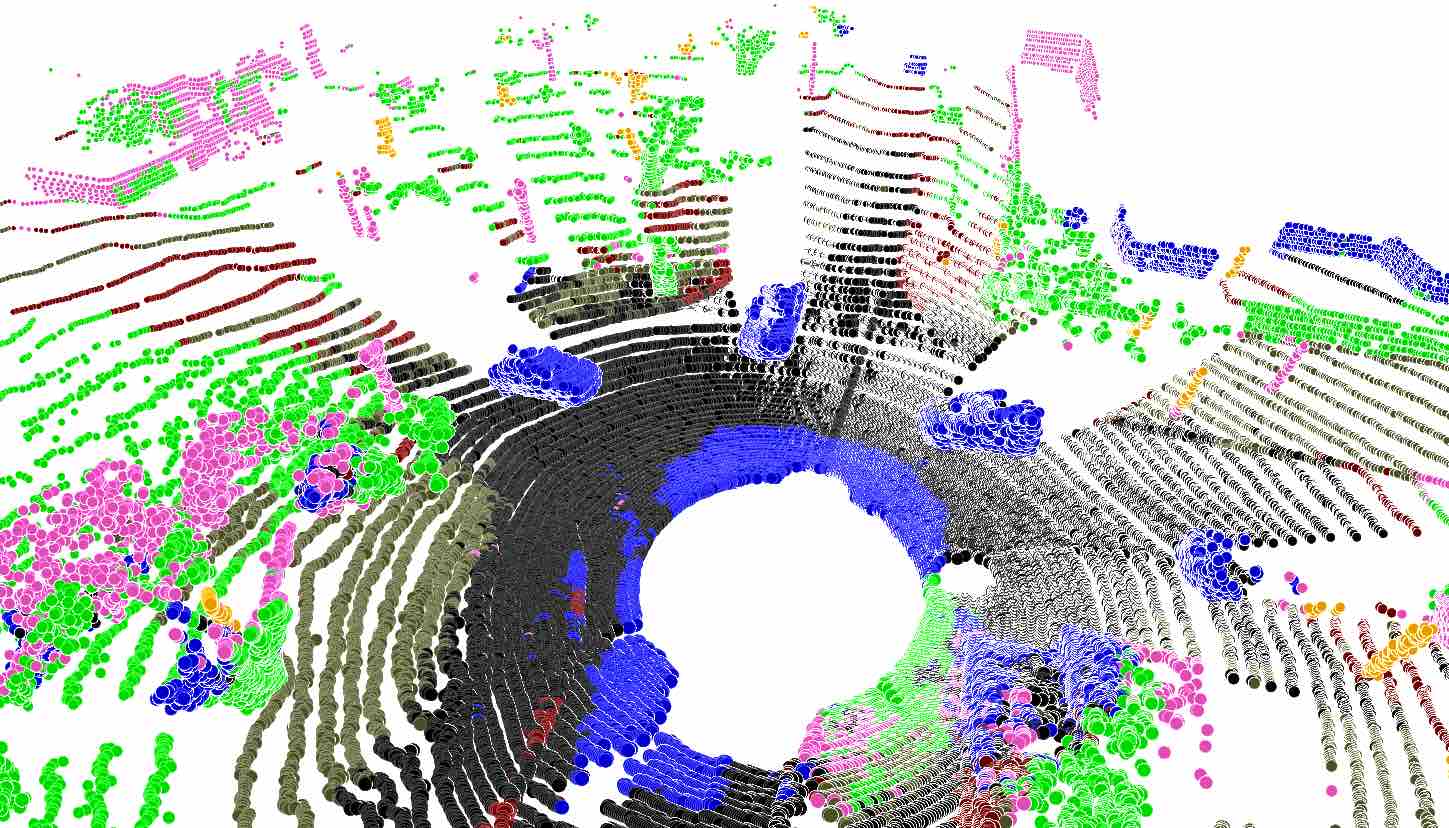}
        \end{overpic} &  
        \begin{overpic}[width=0.21\textwidth]{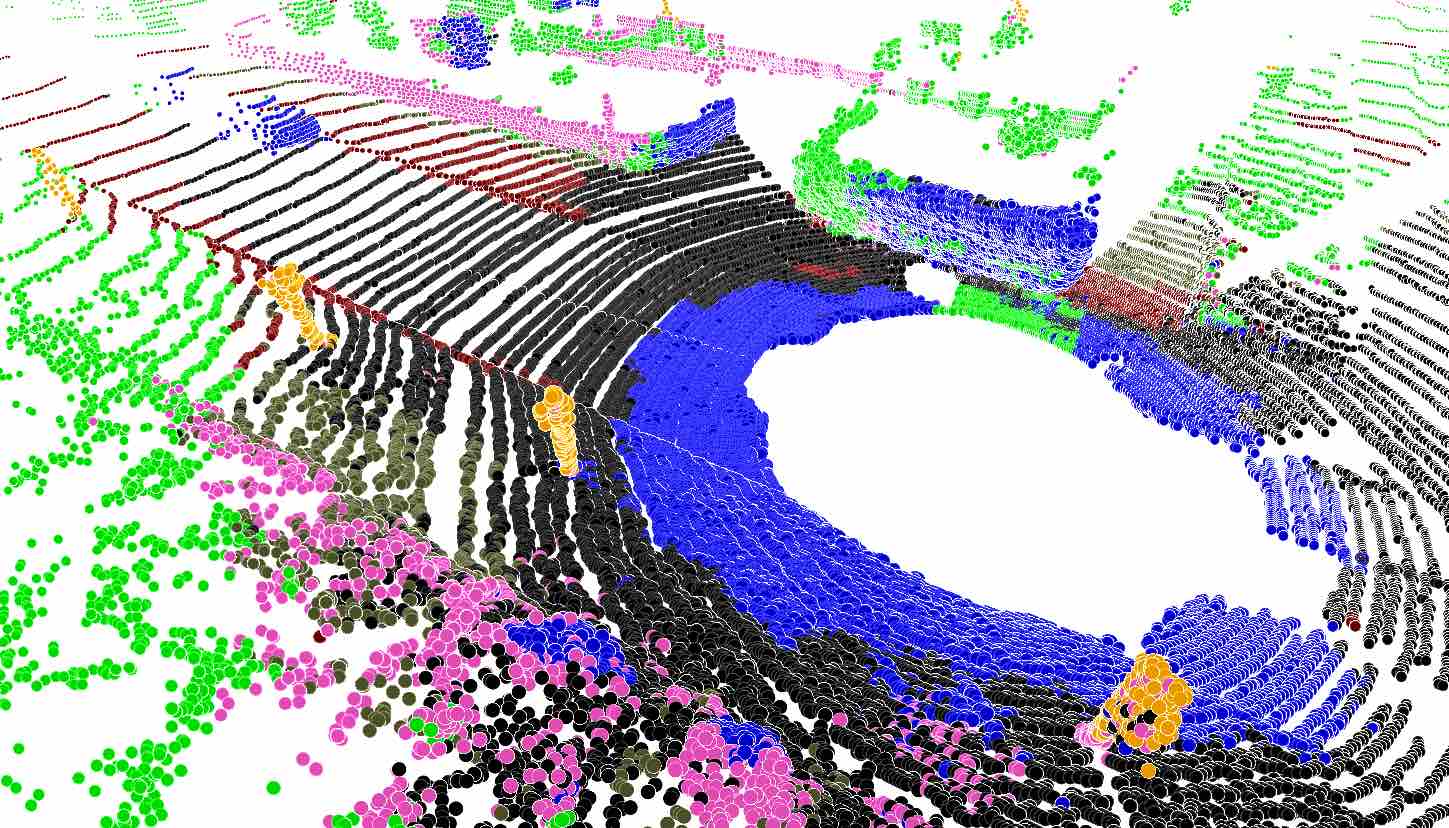}
        \end{overpic} &
        \begin{overpic}[width=0.21\textwidth]{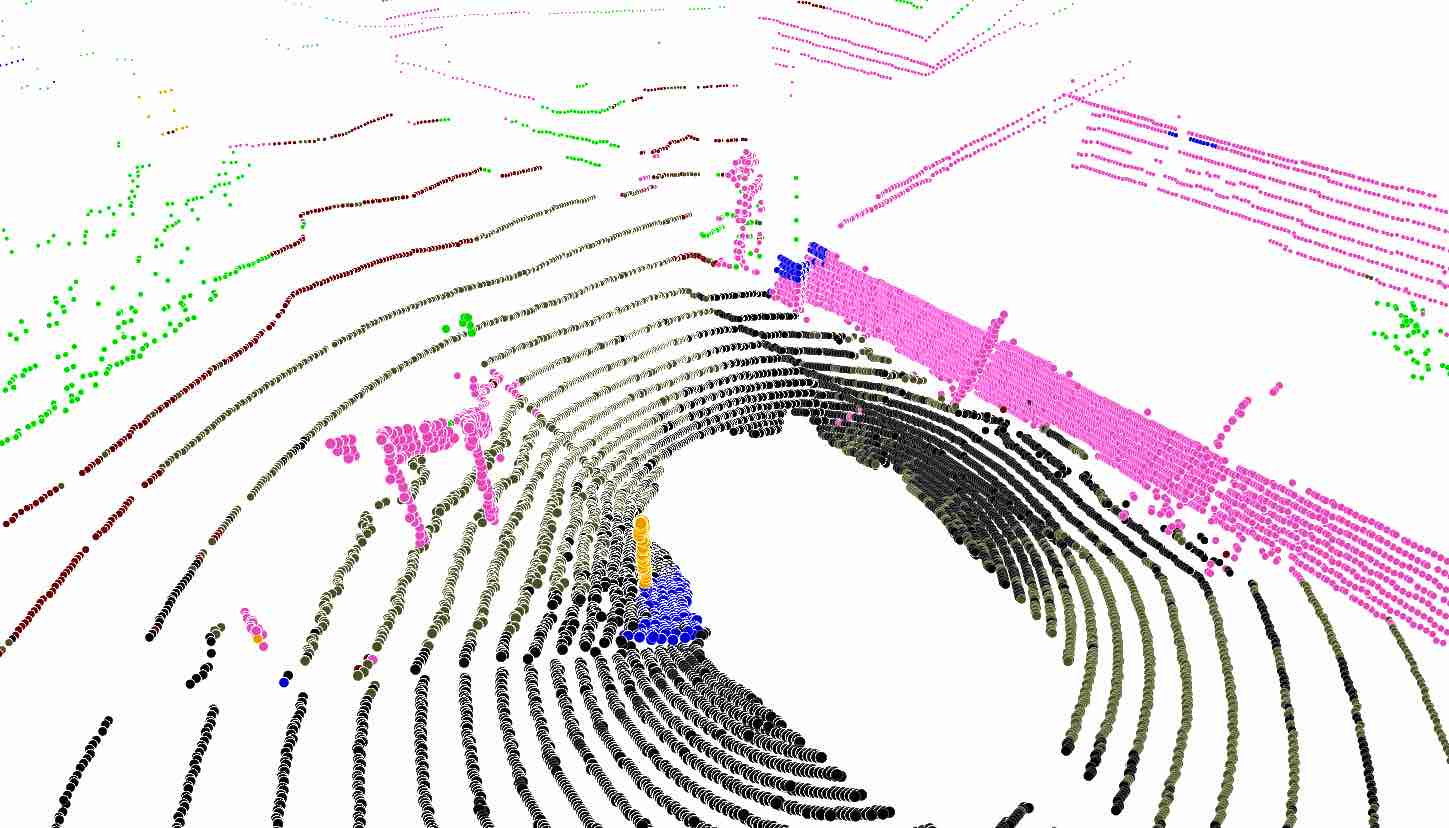}
        \end{overpic}& 
        \begin{overpic}[width=0.21\textwidth]{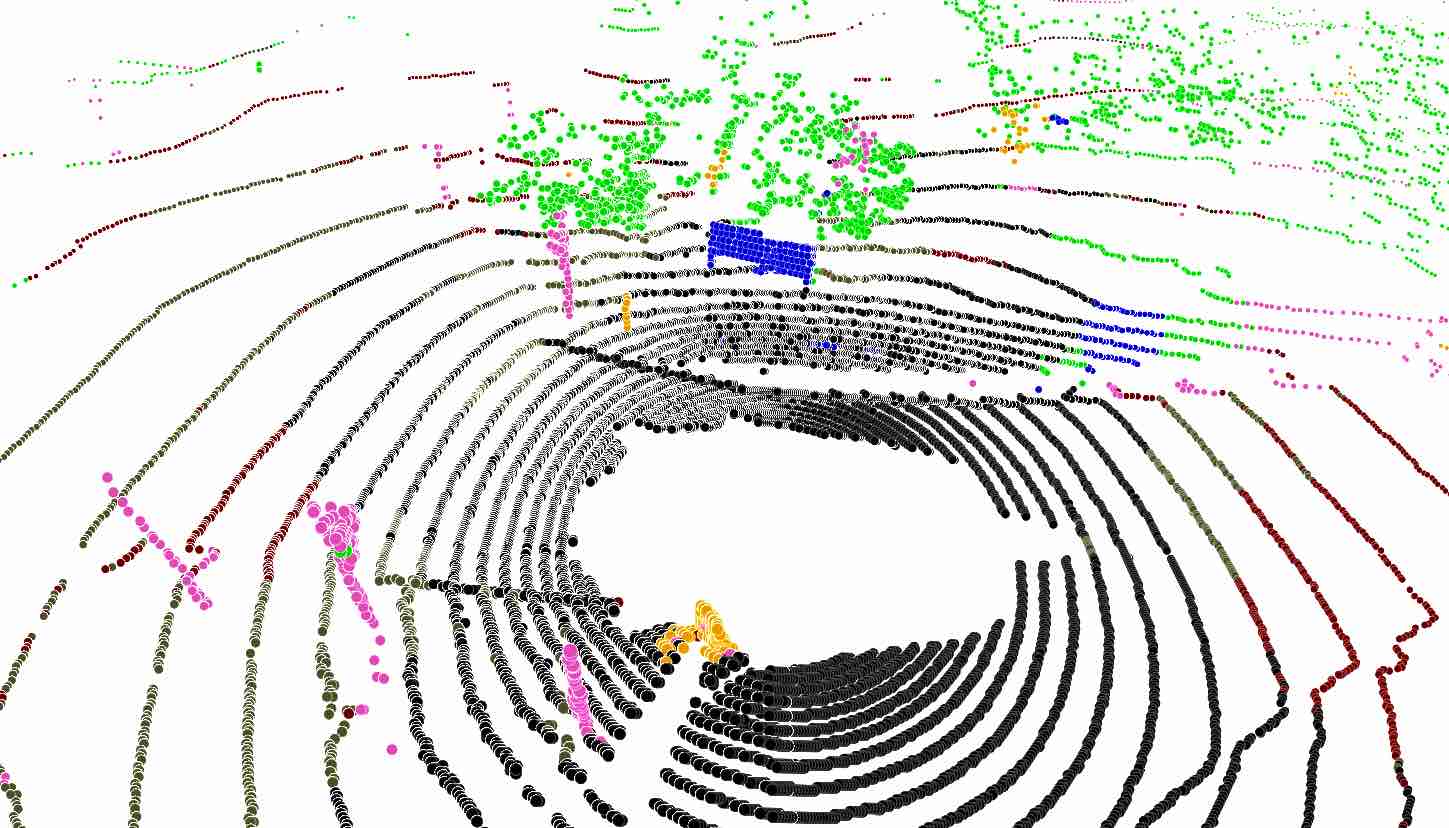}
        \end{overpic}\\
        \begin{overpic}[width=0.21\textwidth]{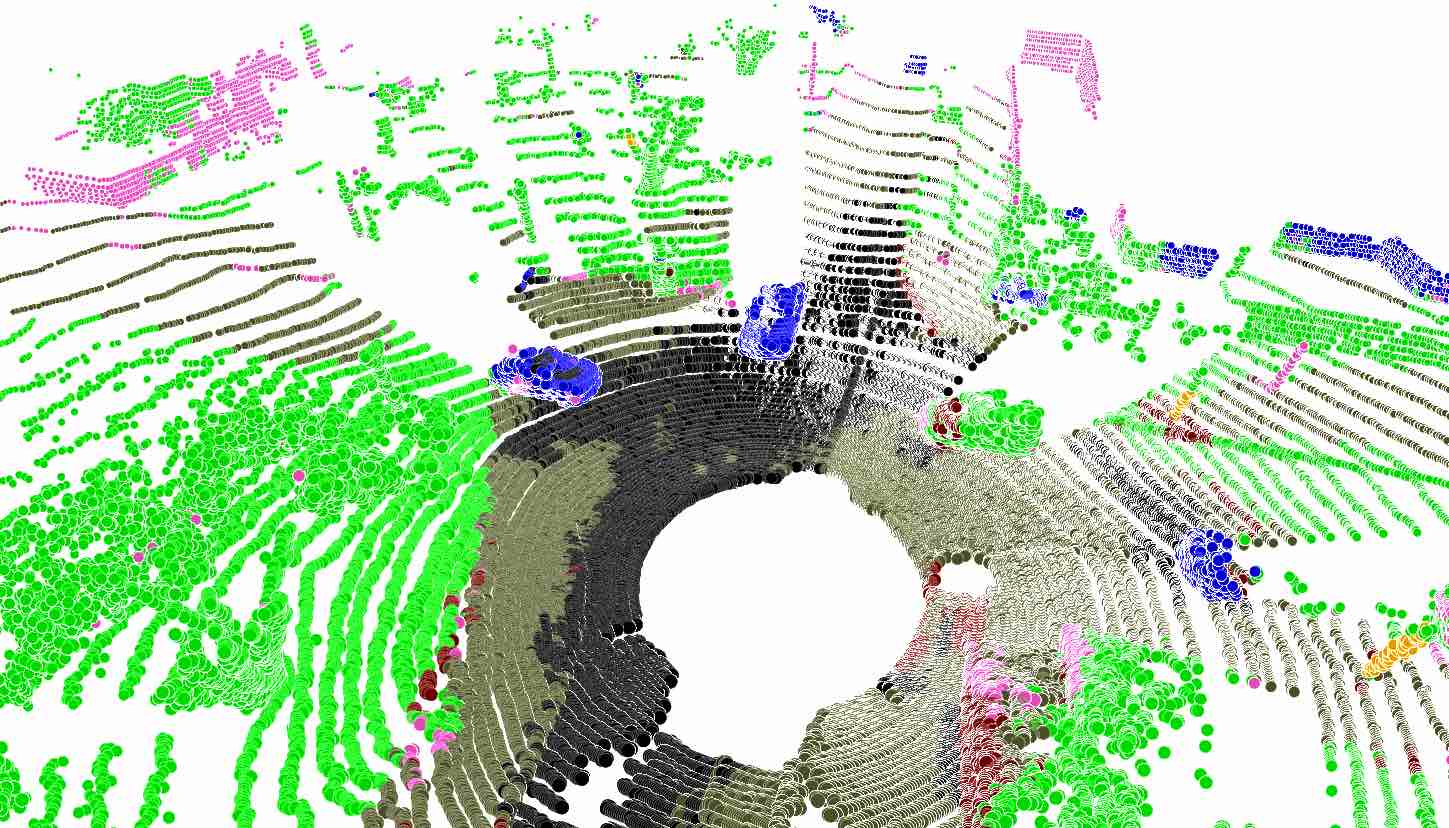}
        \end{overpic} &  
        \begin{overpic}[width=0.21\textwidth]{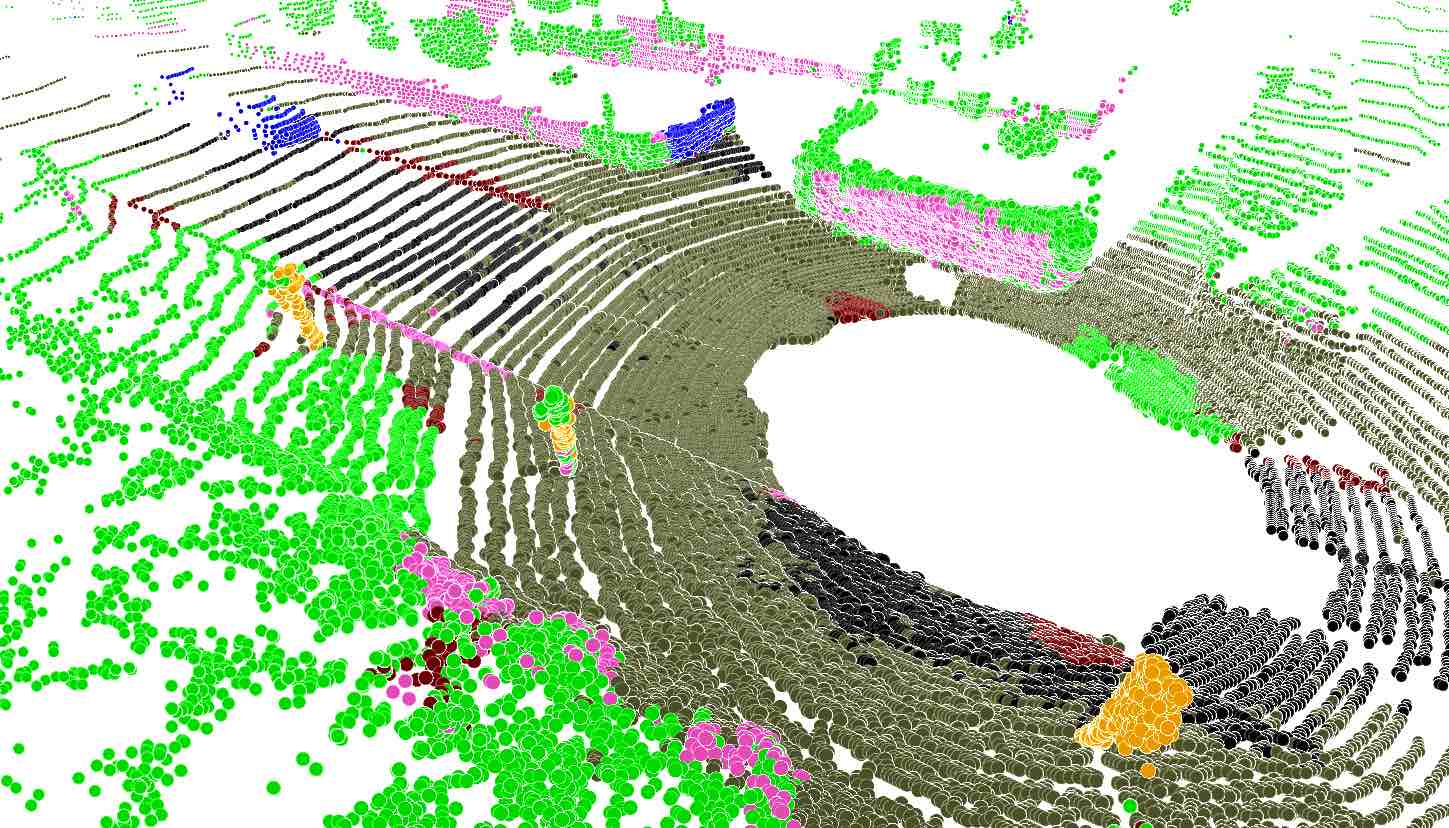}
        \end{overpic} &
        \begin{overpic}[width=0.21\textwidth]{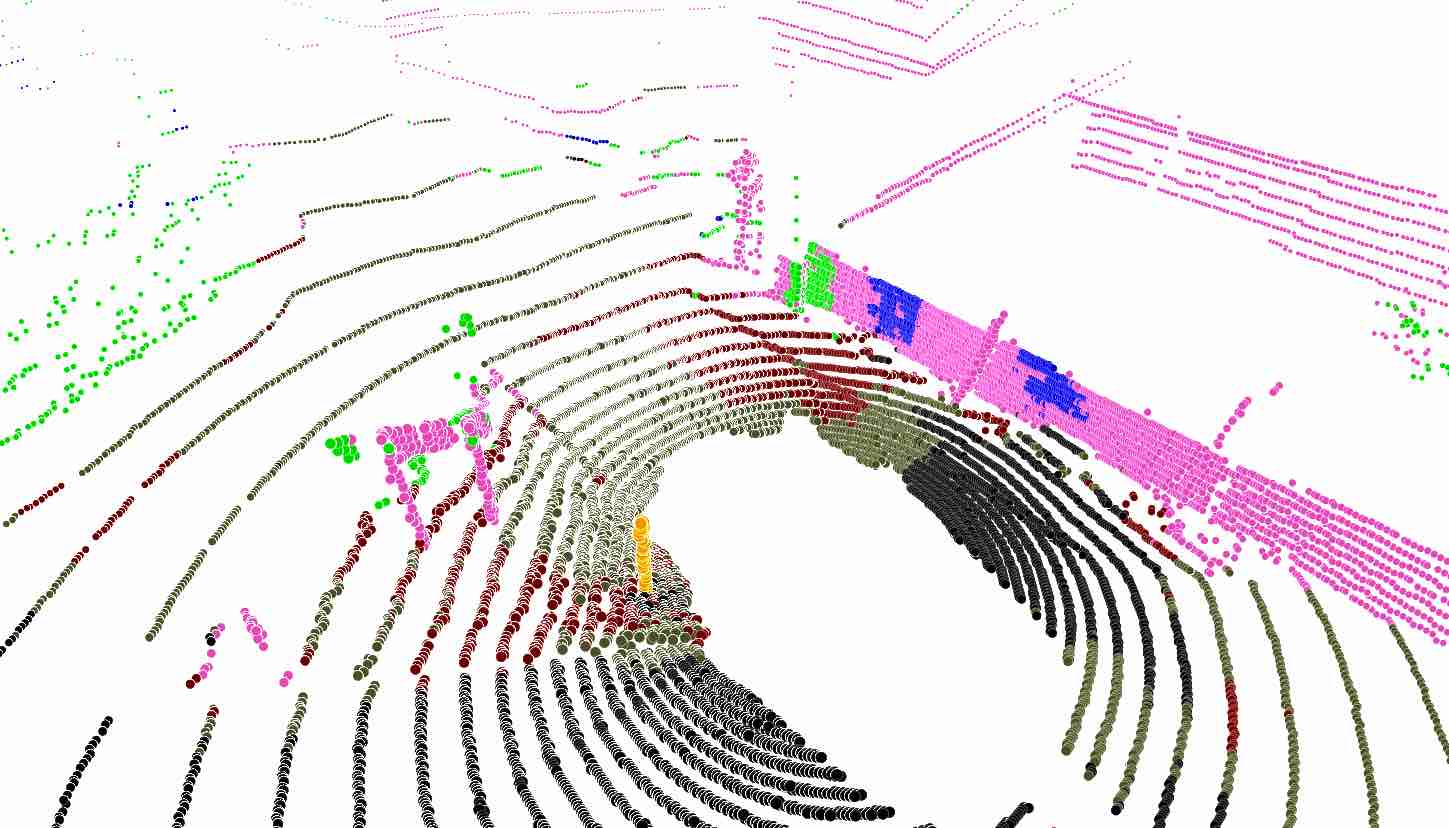}
        \end{overpic}& 
        \begin{overpic}[width=0.21\textwidth]{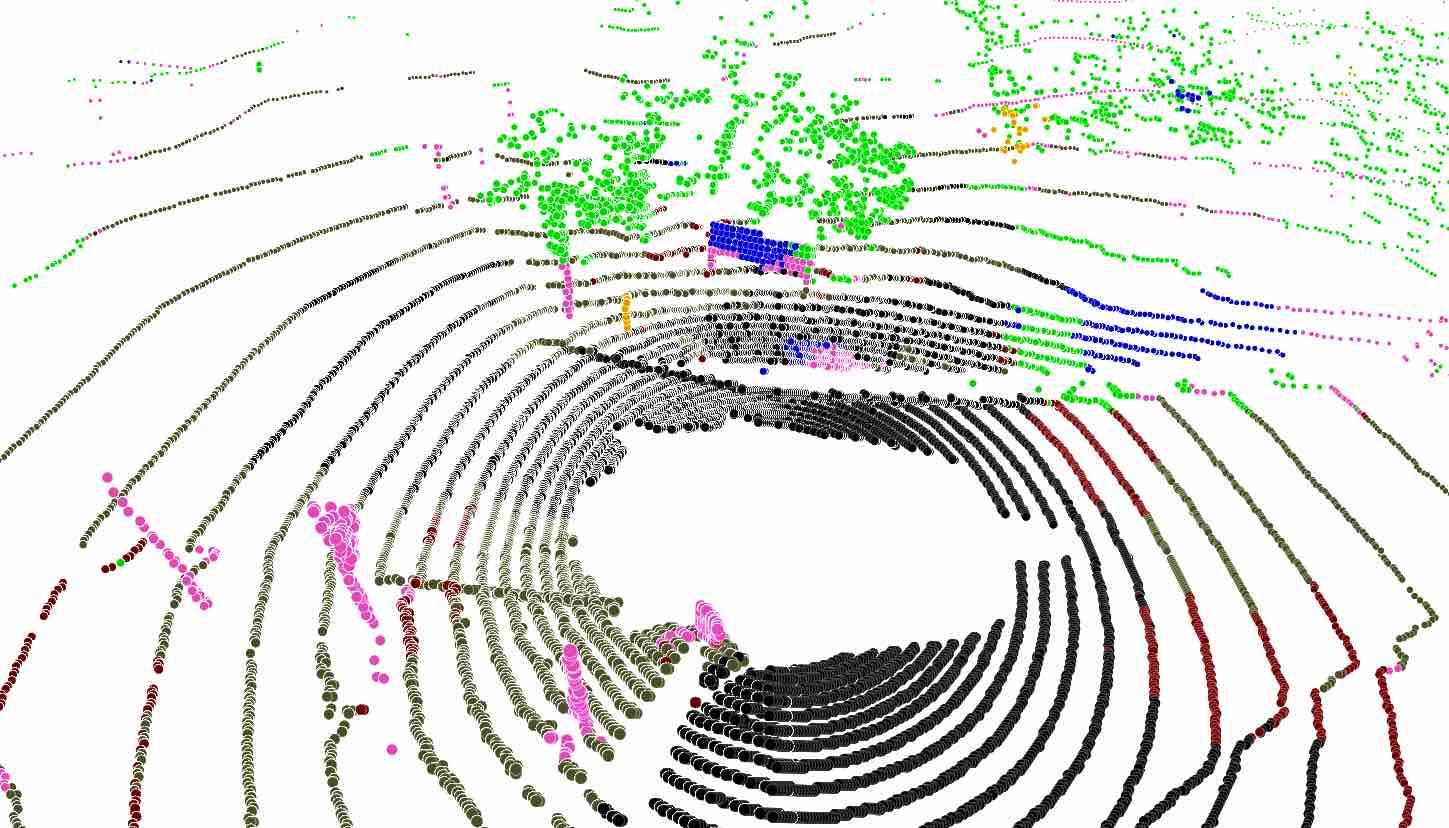}
        \end{overpic}\\
        \begin{overpic}[width=0.21\textwidth]{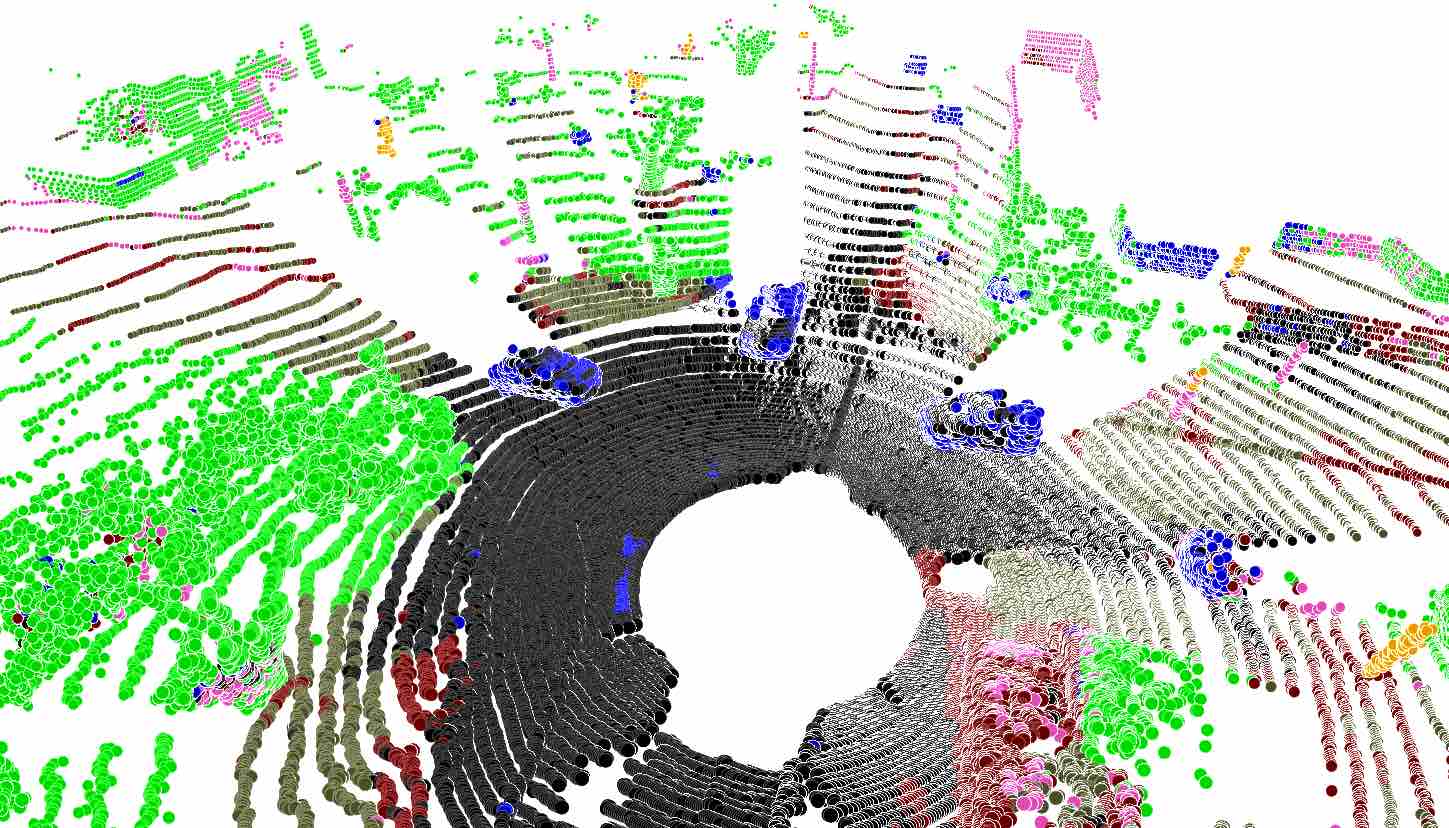}
        \end{overpic} &  
        \begin{overpic}[width=0.21\textwidth]{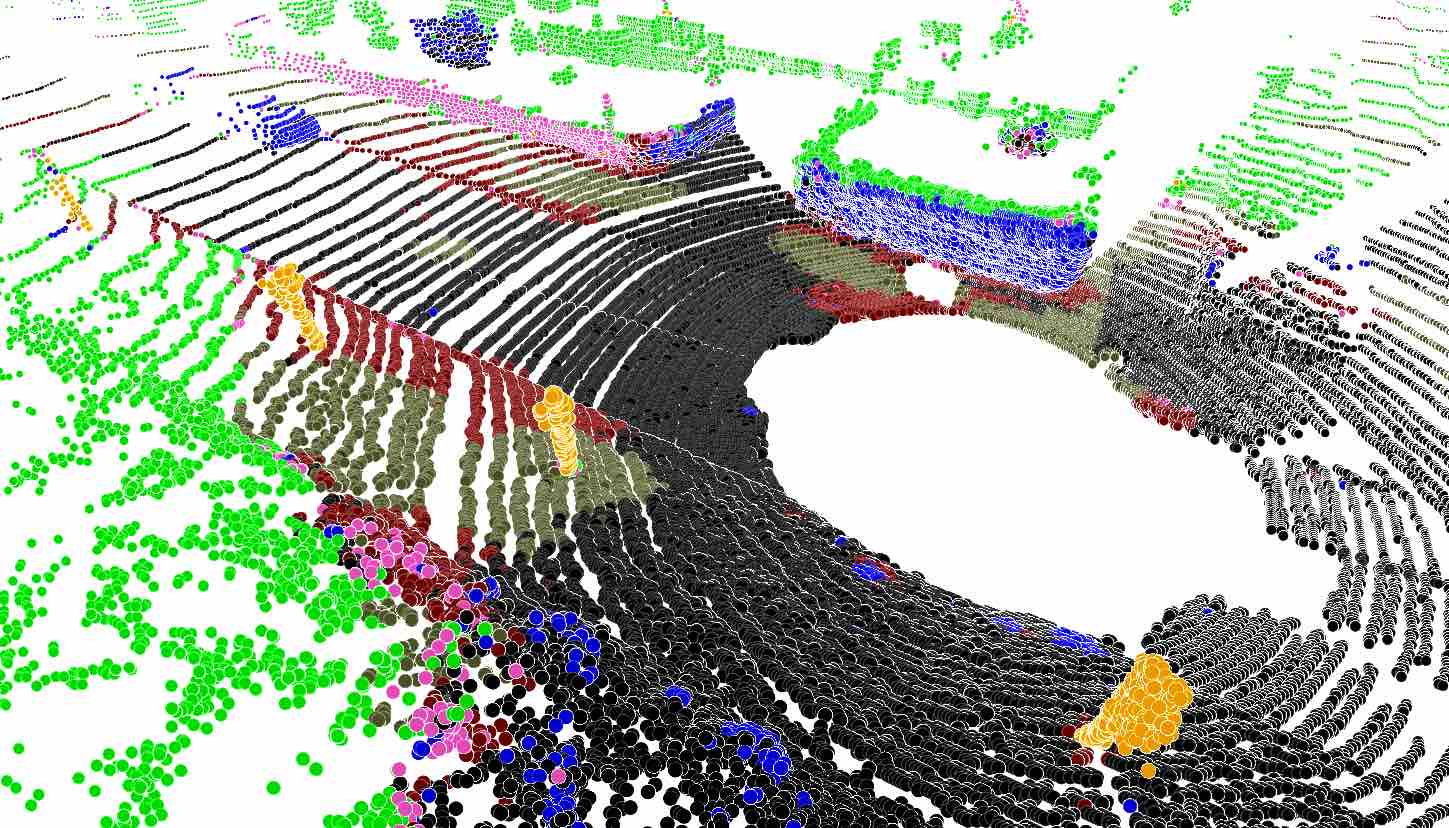}
        \end{overpic} &
        \begin{overpic}[width=0.21\textwidth]{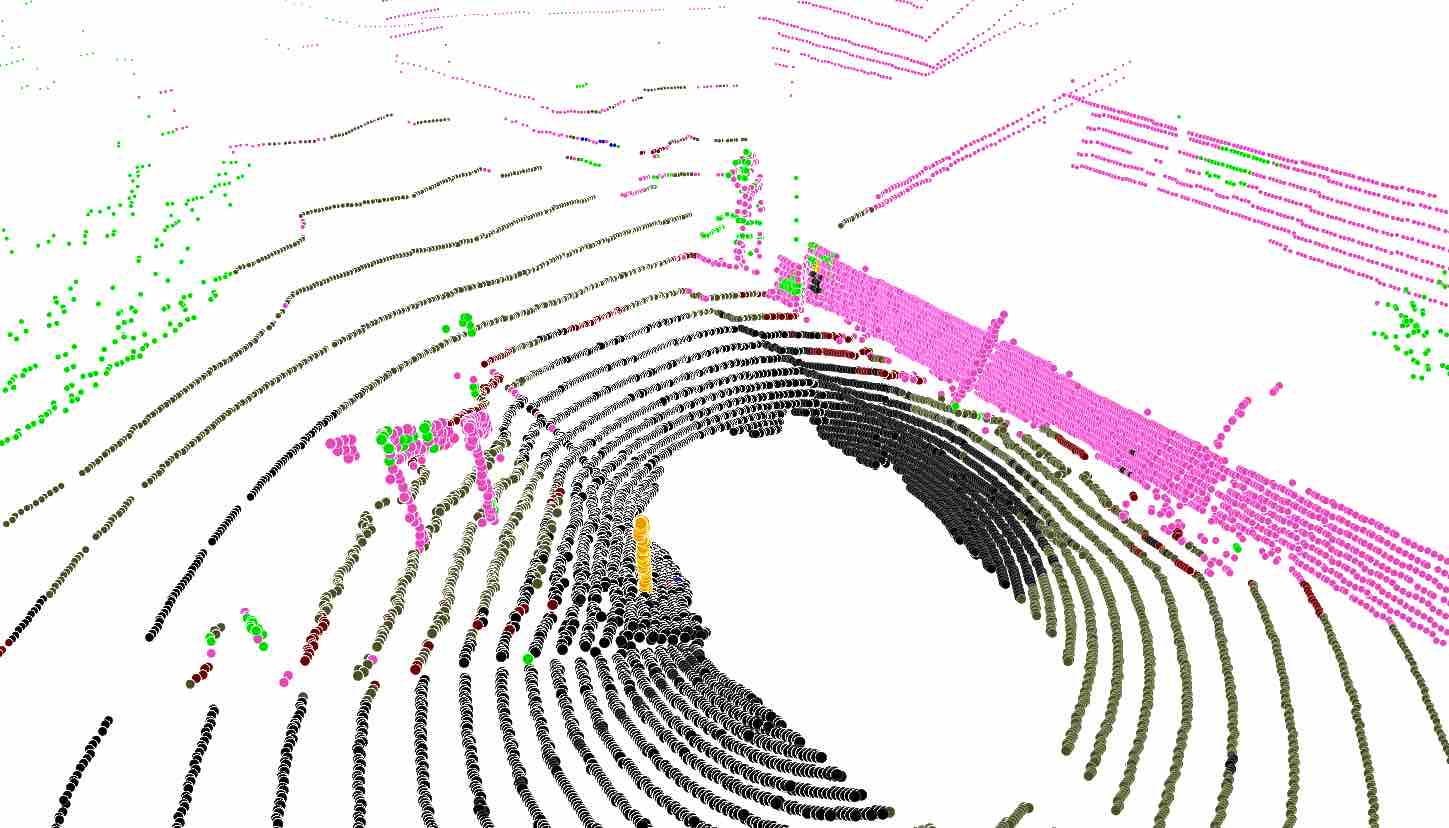}
        \end{overpic}& 
        \begin{overpic}[width=0.21\textwidth]{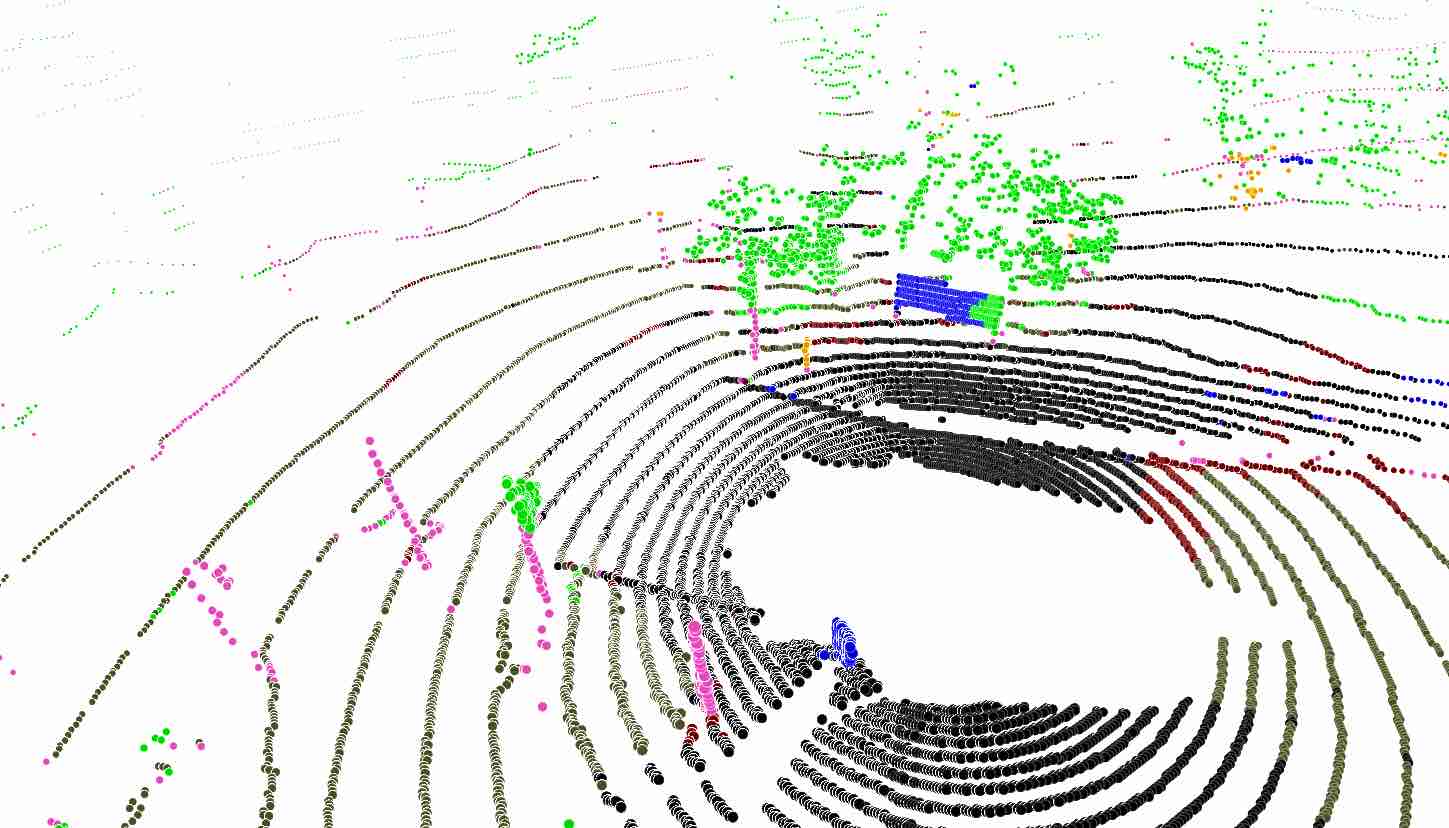}
        \end{overpic}\\
        \begin{overpic}[width=0.21\textwidth]{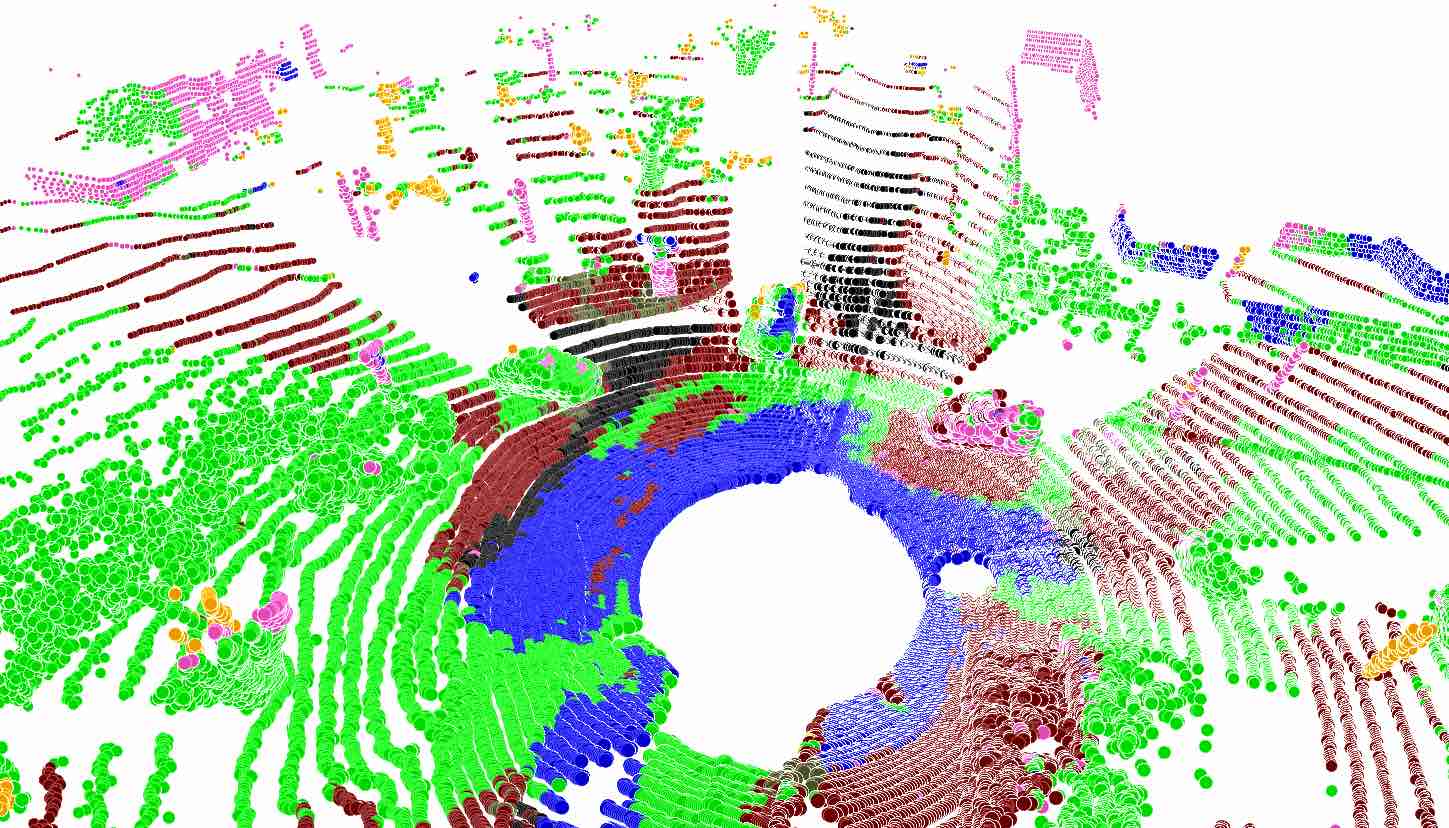}
        \end{overpic} &  
        \begin{overpic}[width=0.21\textwidth]{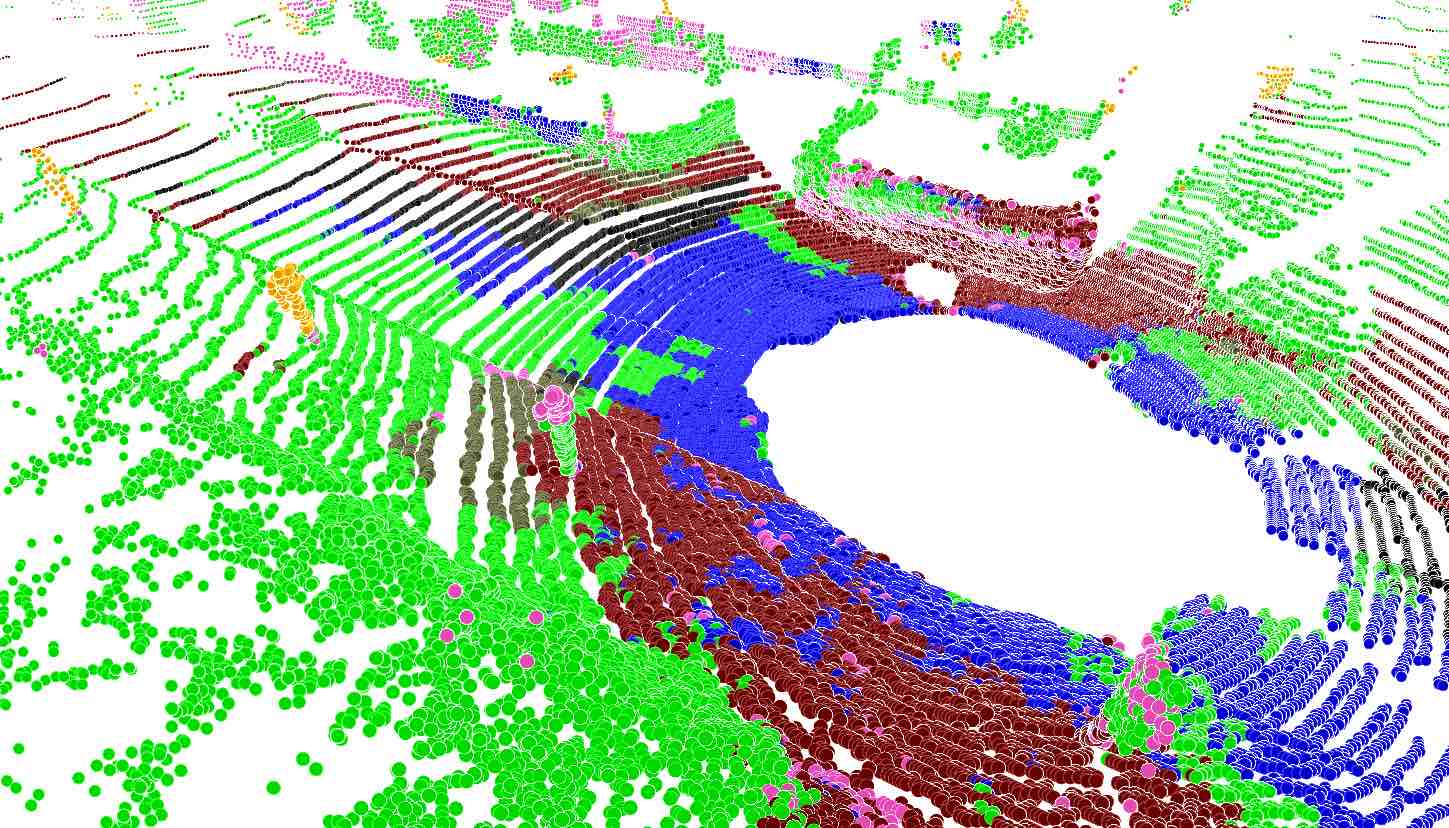}
        \end{overpic} &
        \begin{overpic}[width=0.21\textwidth]{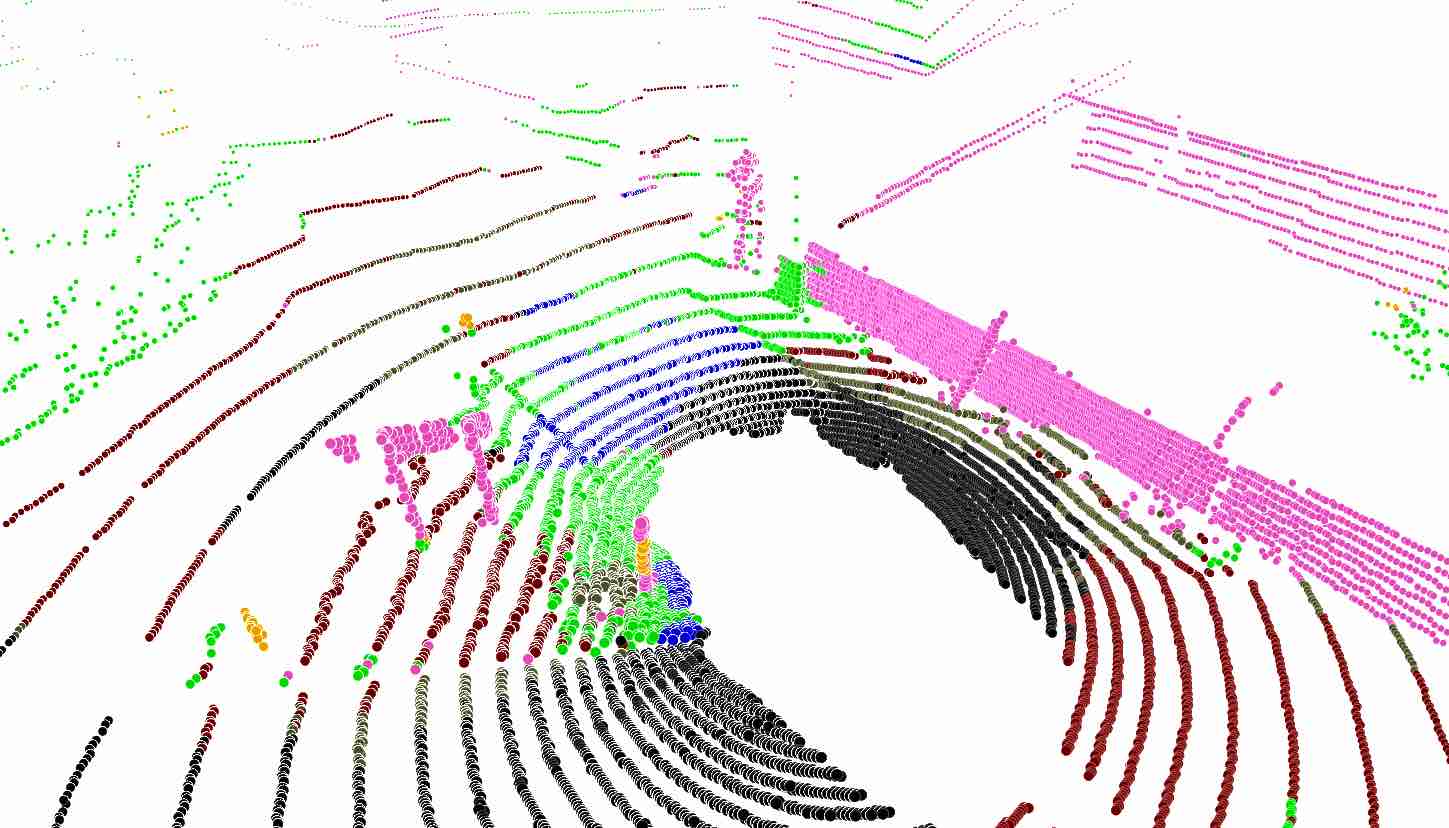}
        \end{overpic}& 
        \begin{overpic}[width=0.21\textwidth]{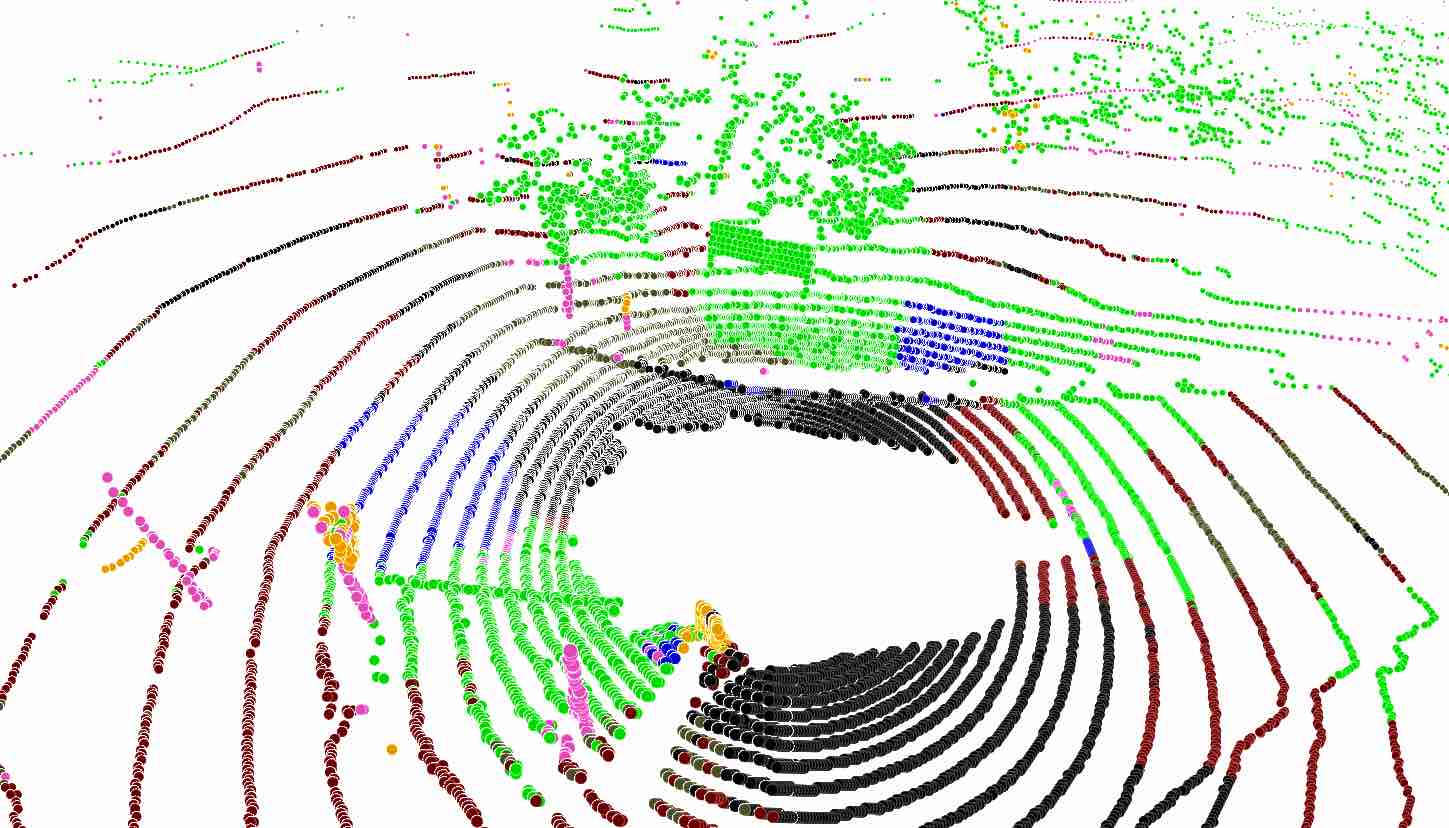}
        \end{overpic}\\
        \begin{overpic}[width=0.21\textwidth]{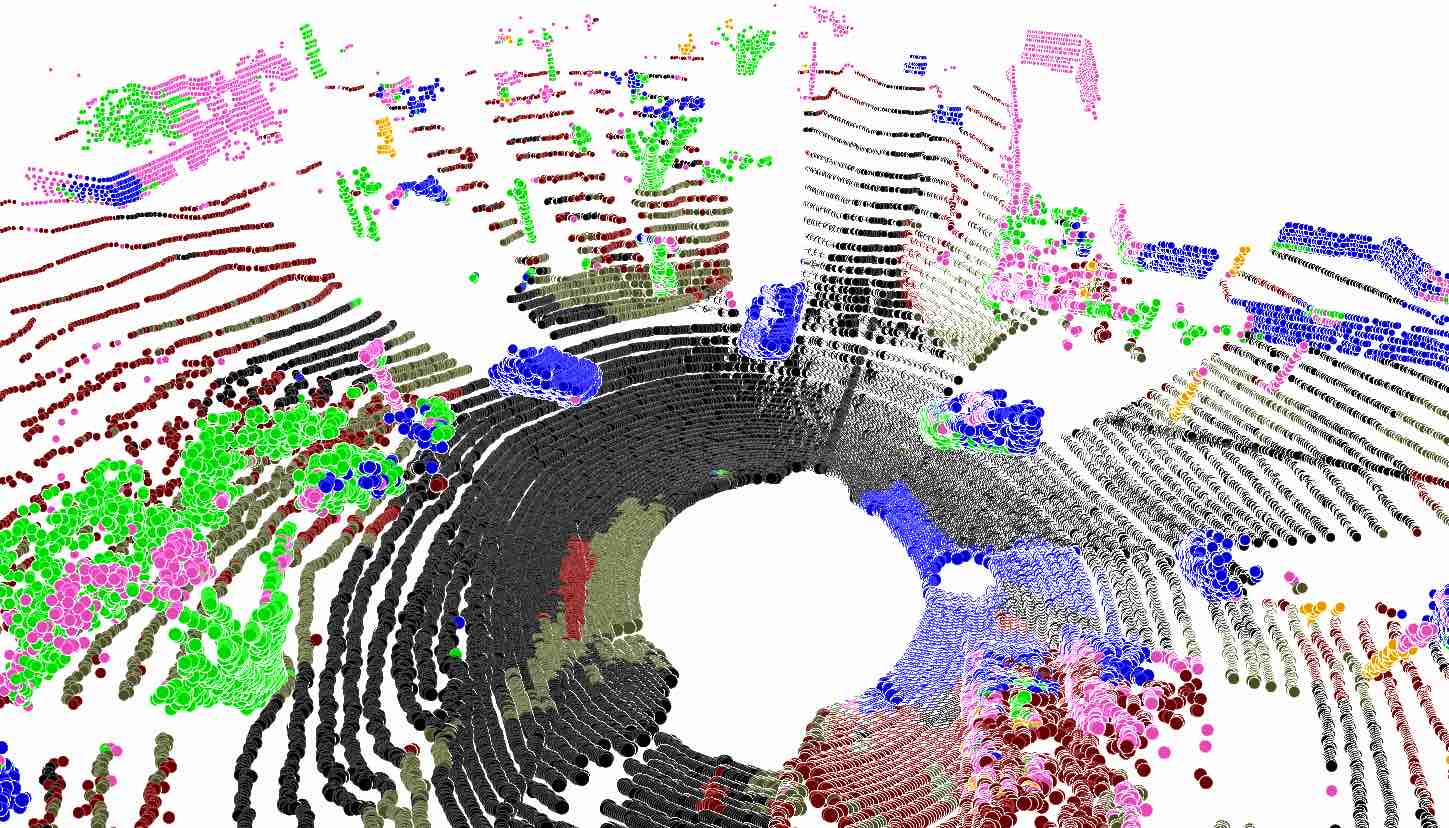}
        \end{overpic} &  
        \begin{overpic}[width=0.21\textwidth]{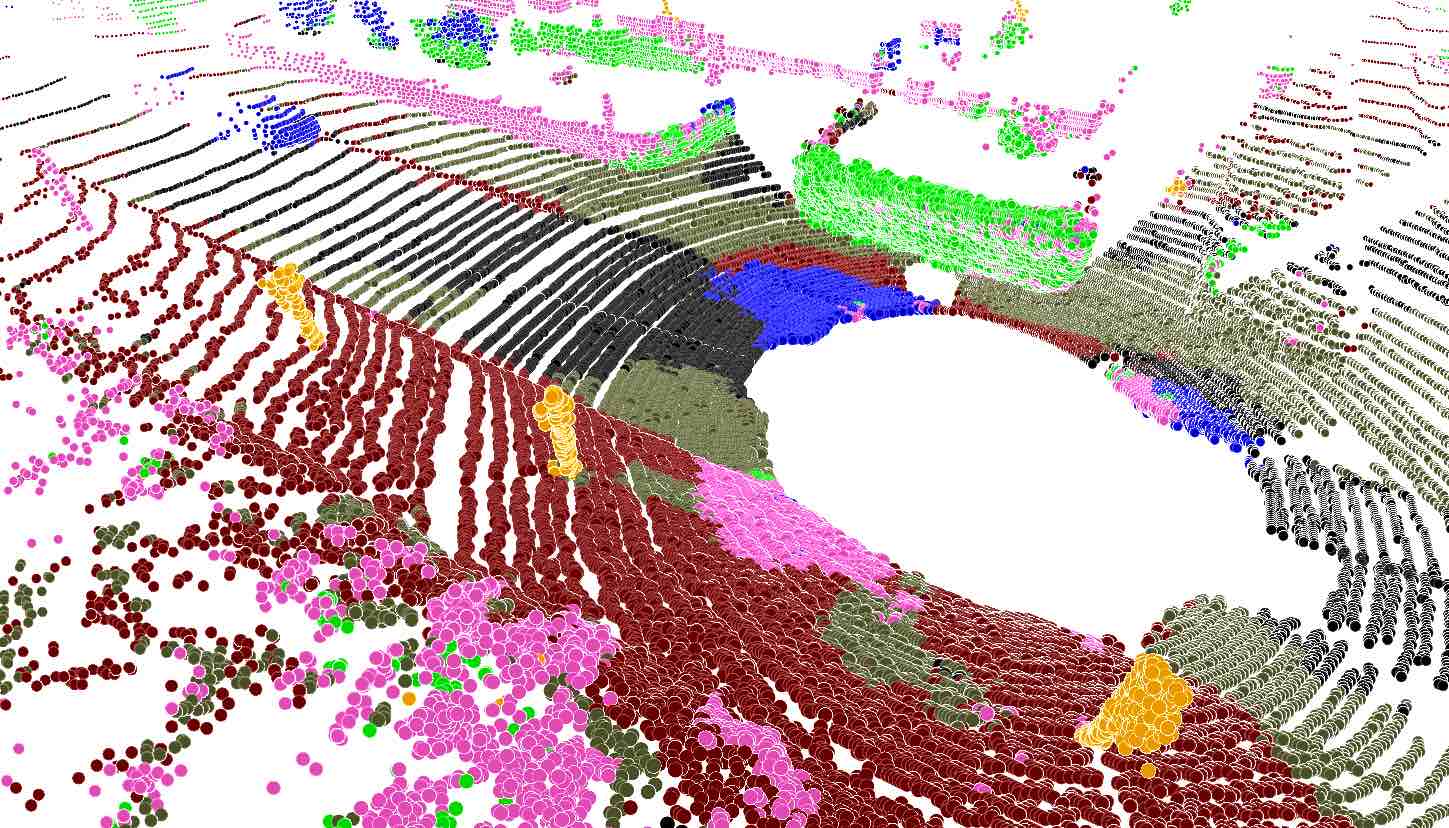}
        \end{overpic} &
        \begin{overpic}[width=0.21\textwidth]{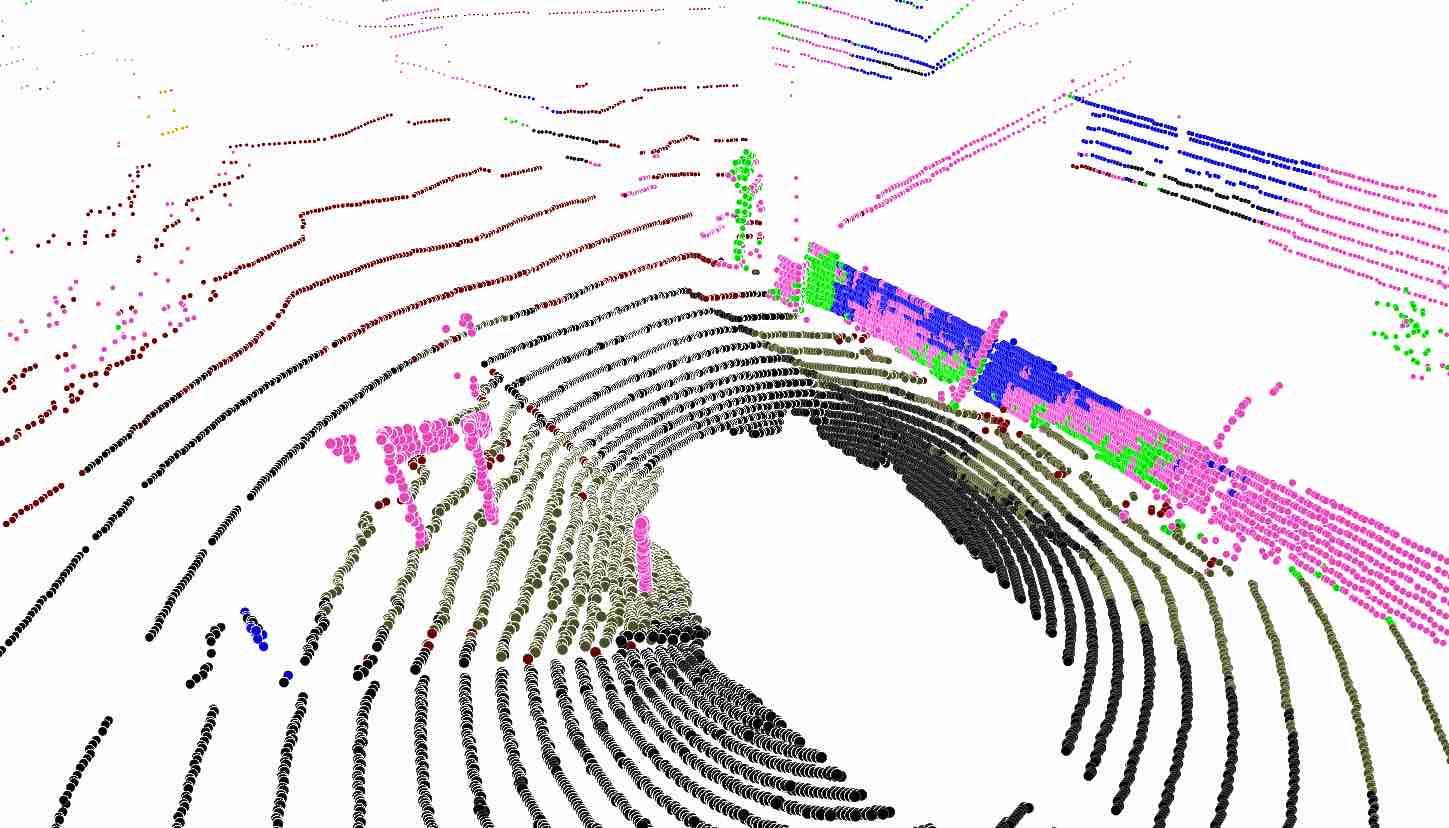}
        \end{overpic}& 
        \begin{overpic}[width=0.21\textwidth]{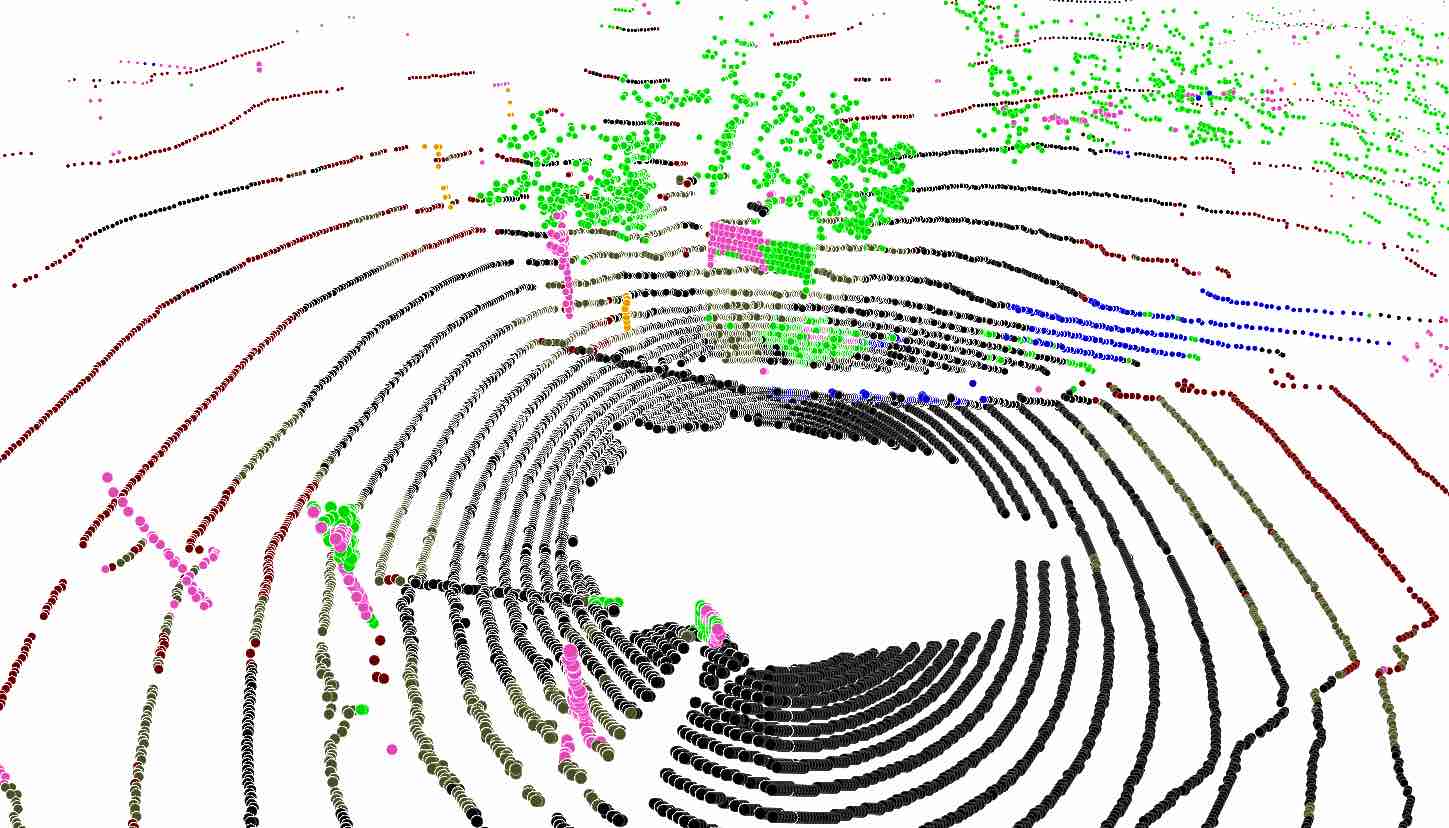}
        \end{overpic}\\
        \begin{overpic}[width=0.21\textwidth]{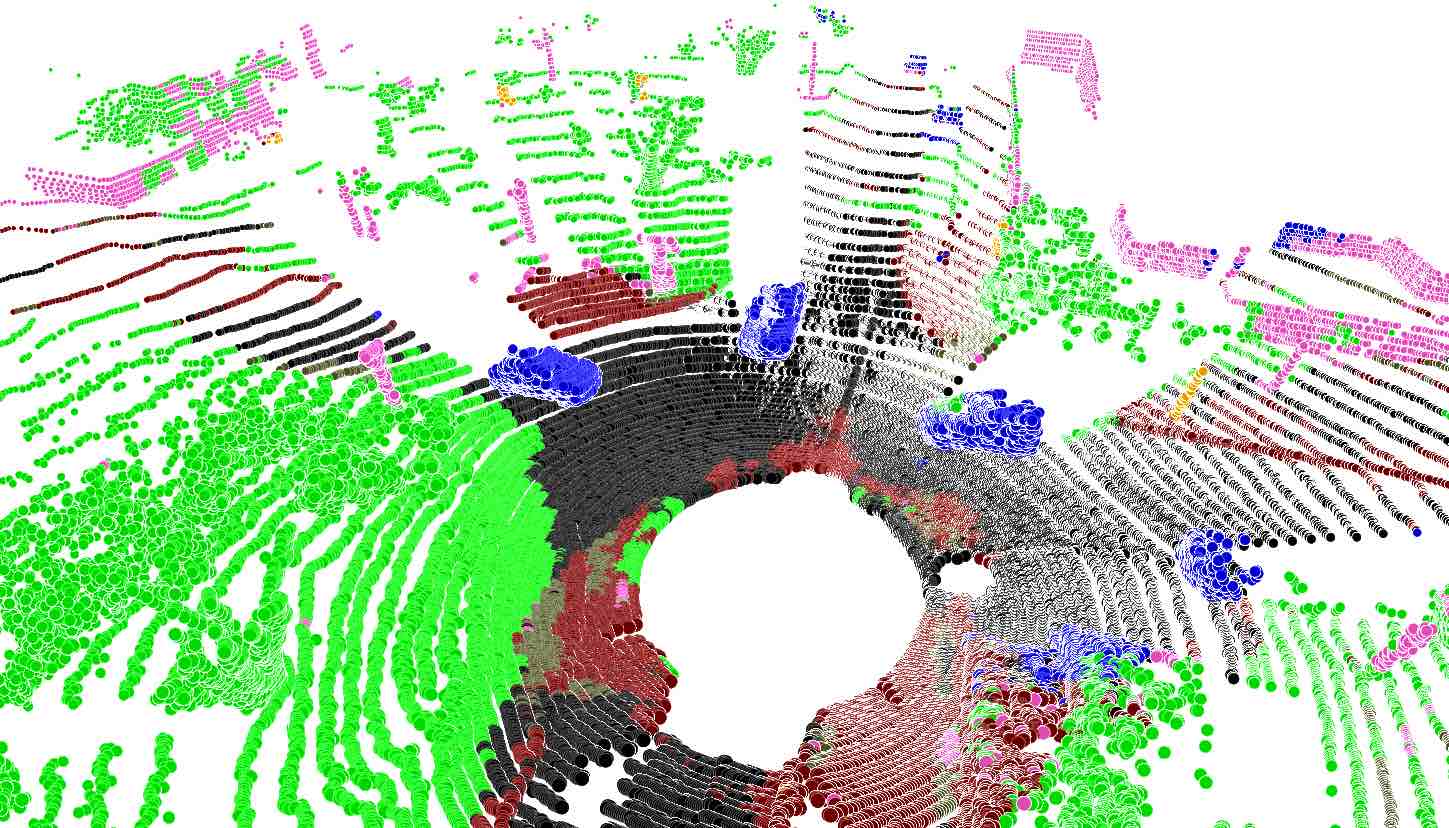}
        \put(-4.5,500){\rotatebox{90}{\color{black}\footnotesize \textbf{source}}}
        \put(-6,440){\rotatebox{90}{\color{black}\footnotesize \textbf{mix3D}}}
        \put(-6,375){\rotatebox{90}{\color{black}\footnotesize \textbf{p.cutmix}}}
        \put(-6,317){\rotatebox{90}{\color{black}\footnotesize \textbf{cosmix}}}
        \put(-6,260){\rotatebox{90}{\color{black}\footnotesize \textbf{ibn}}}
        \put(-6,200){\rotatebox{90}{\color{black}\footnotesize \textbf{robust.}}}
        \put(-5,140){\rotatebox{90}{\color{black}\footnotesize \textbf{sn}}}
        \put(-6,75){\rotatebox{90}{\color{black}\footnotesize \textbf{raycast}}}
        \put(-5,17){\rotatebox{90}{\color{black}\footnotesize \textbf{ours}}}
        \put(-6,-35){\rotatebox{90}{\color{black}\footnotesize \textbf{gt}}}
        \put(80,541){\color{black}\footnotesize \textbf{SemanticKITTI}}
        \put(290,541){\color{black}\footnotesize \textbf{nuScenes}}
        \end{overpic} &  
        \begin{overpic}[width=0.21\textwidth]{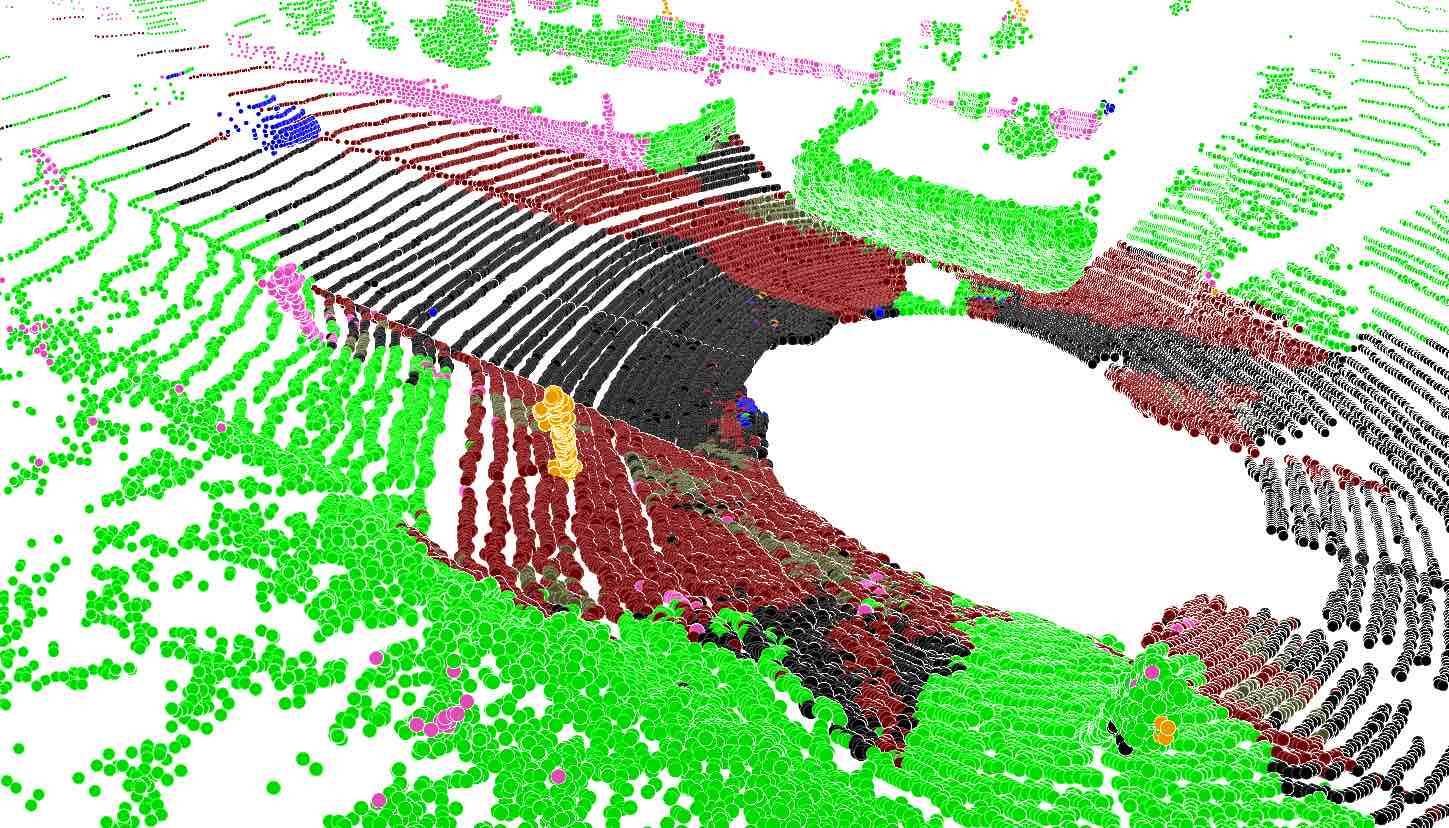}
        \end{overpic} &
        \begin{overpic}[width=0.21\textwidth]{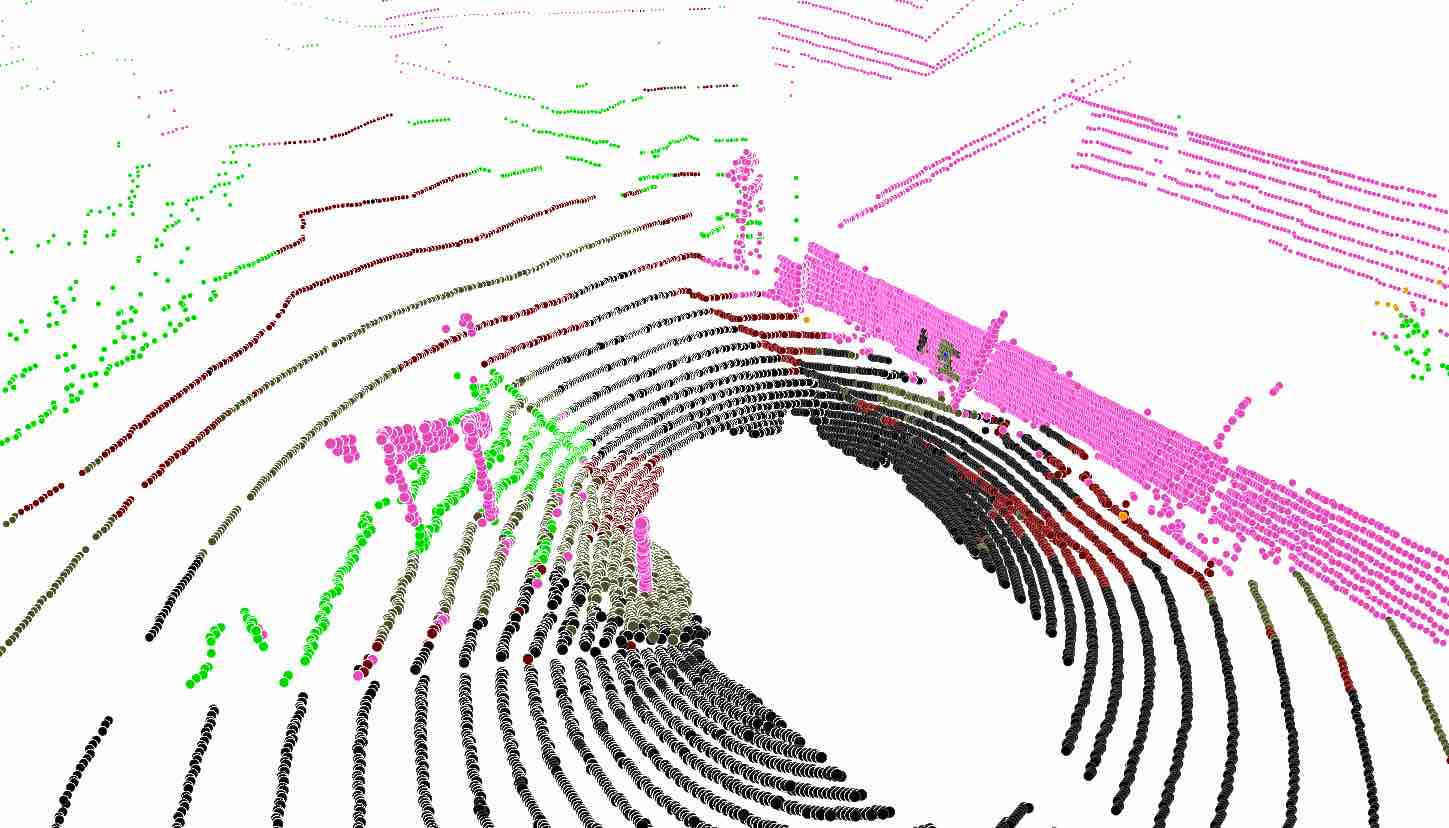}
        \end{overpic}& 
        \begin{overpic}[width=0.21\textwidth]{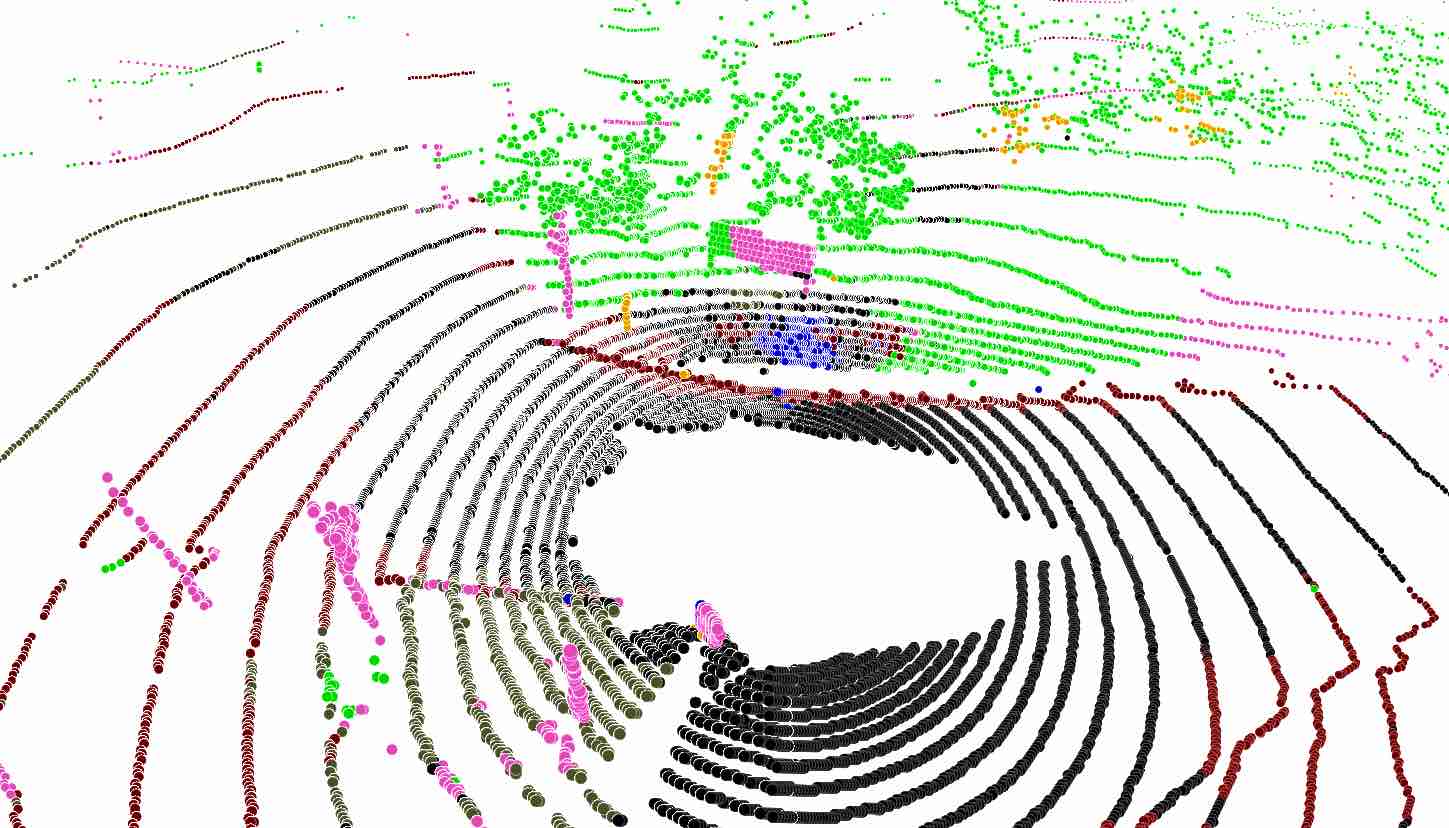}
        \end{overpic}\\
        \begin{overpic}[width=0.21\textwidth]{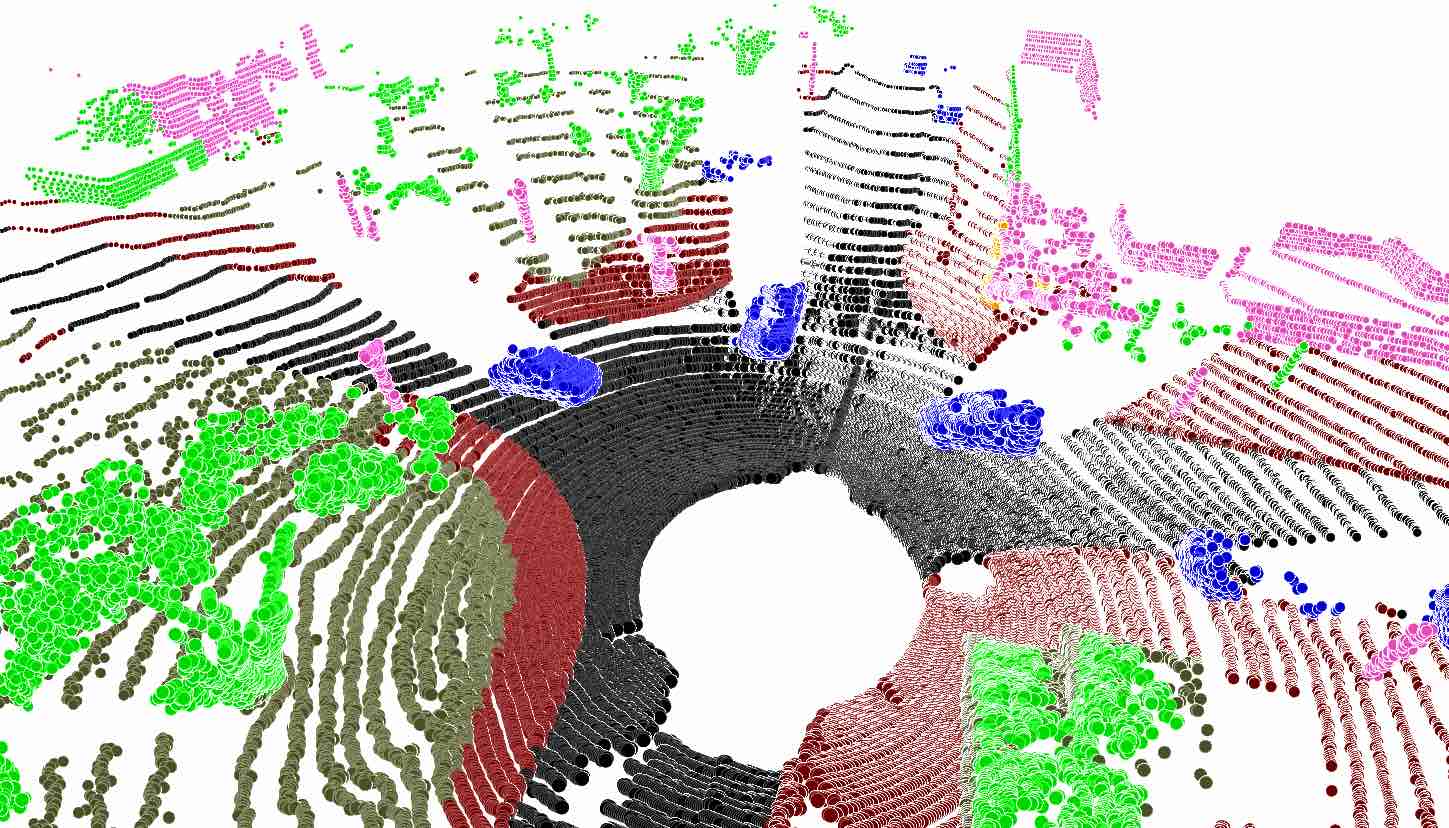}
        \end{overpic} &  
        \begin{overpic}[width=0.21\textwidth]{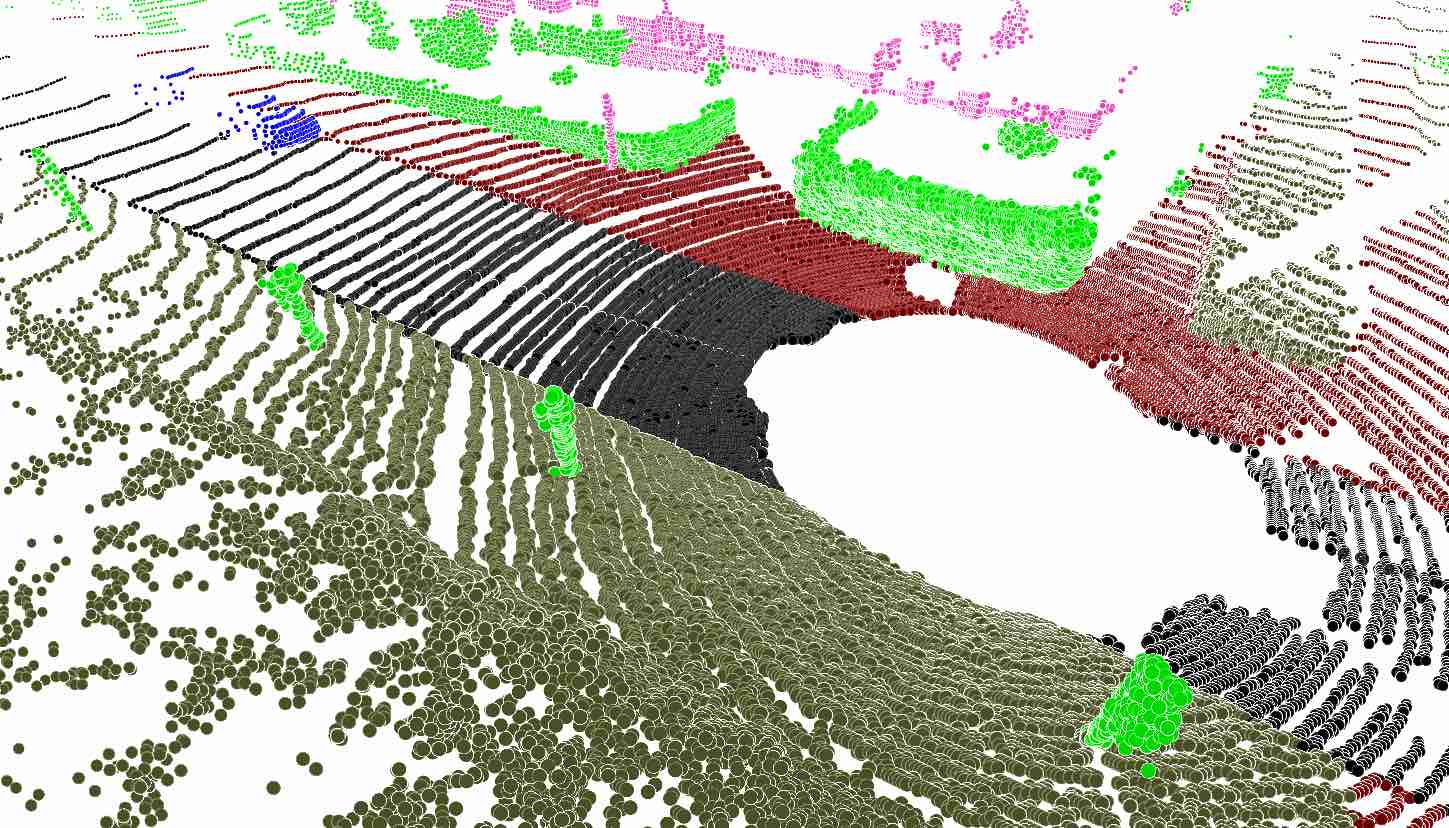}
        \end{overpic} &
        \begin{overpic}[width=0.21\textwidth]{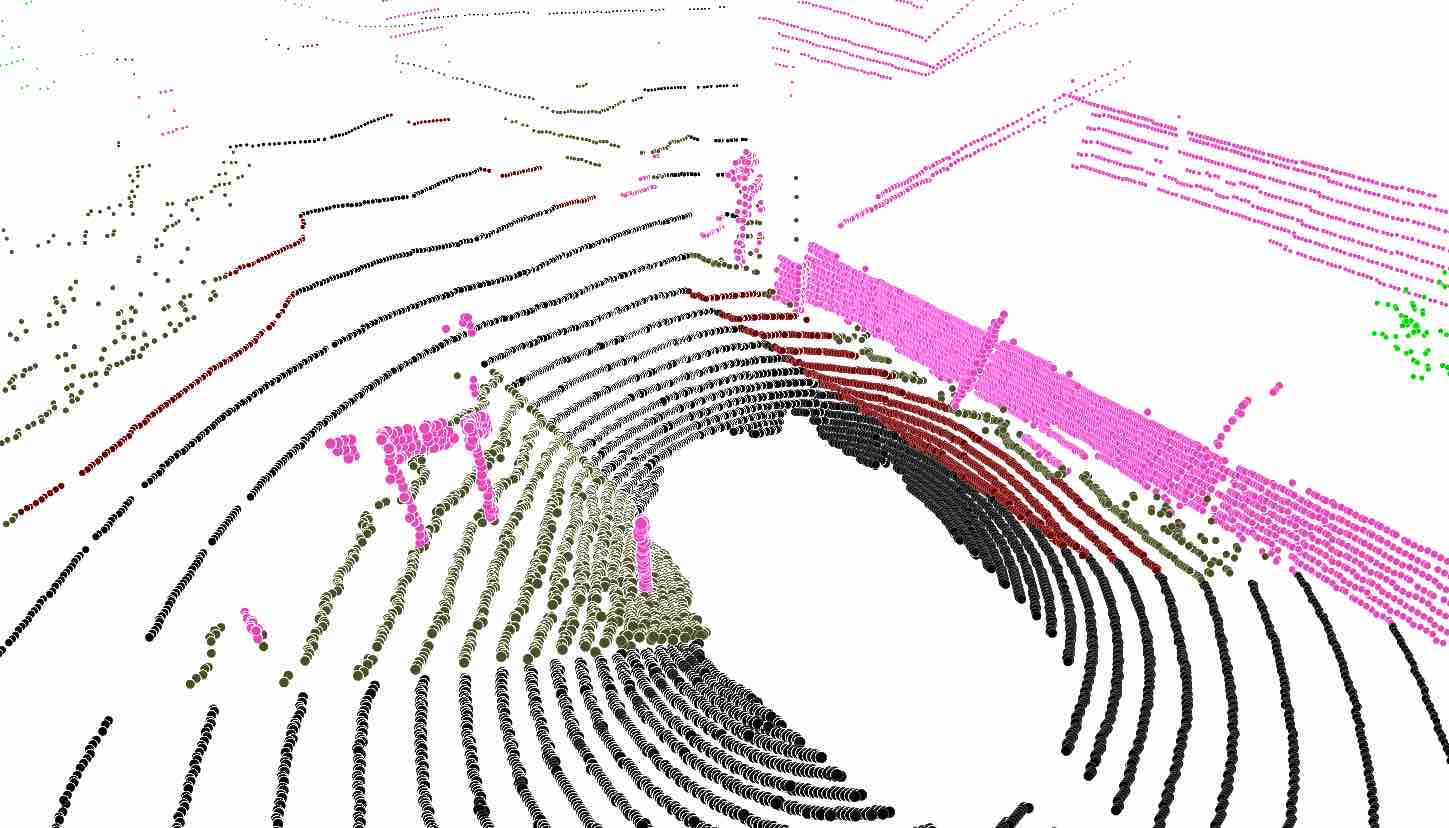}
        \end{overpic}& 
        \begin{overpic}[width=0.21\textwidth]{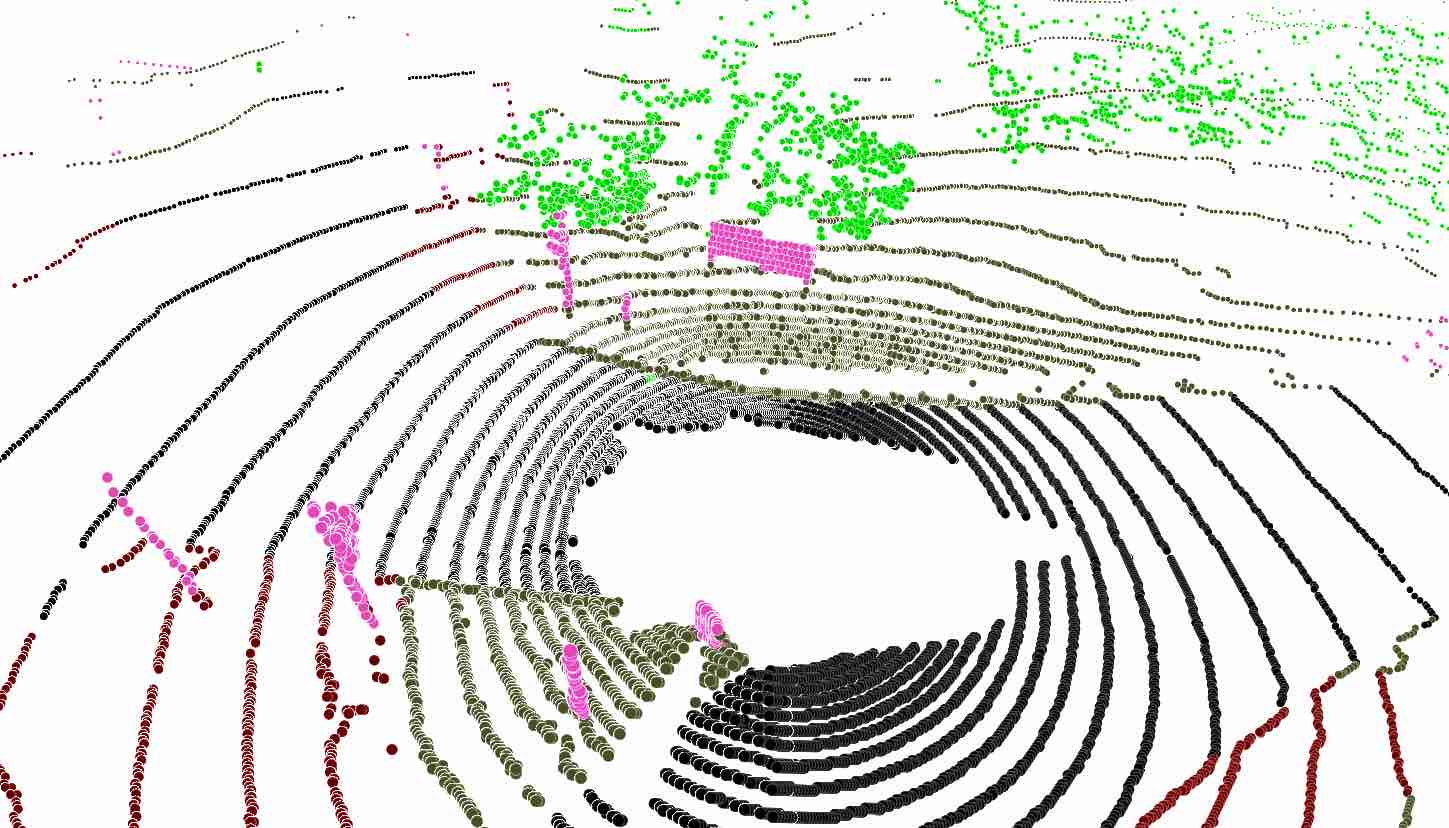}
        \end{overpic}\\
    \end{tabular}
    \vspace{-4mm}
    \caption{\textbf{Qualitative results.} \textit{Left:} Synth4D-nuScenes$\rightarrow$SemanticKITTI, \textit{right:} Synth4D-nuScenes$\rightarrow$nuScenes. \lidog improves over source and baselines, \eg, we observe the improvements of \textit{road} in SemanticKITTI and \textit{manmade} in nuScenes.}
    \label{fig:supp_qualitative_synth4dnusc}
\end{figure*}

\begin{figure*}[t]
\centering
    \setlength\tabcolsep{1.pt}
    \begin{tabular}{cccc}
    \raggedright
        \begin{overpic}[width=0.21\textwidth]{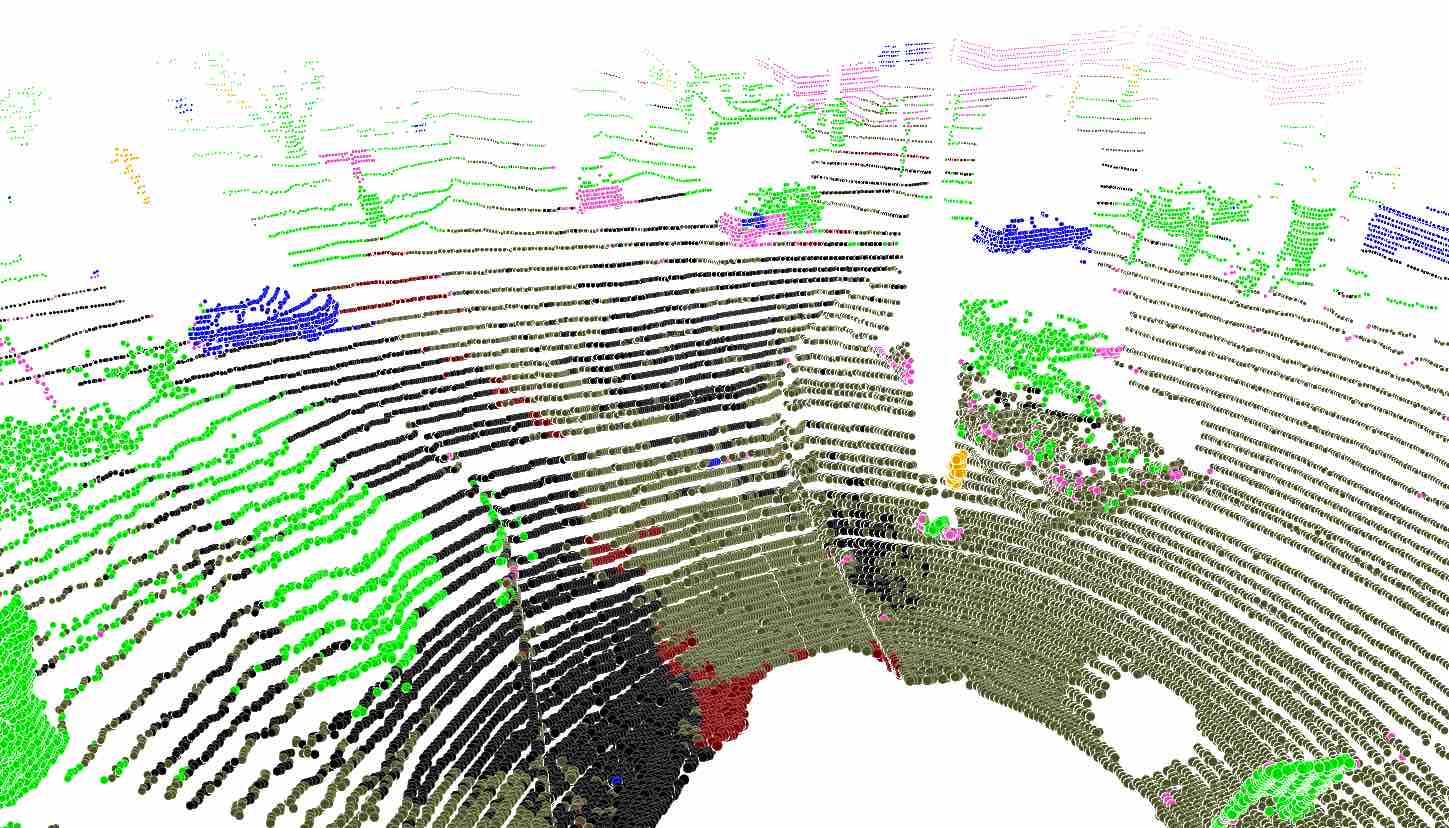}
        \end{overpic} &  
        \begin{overpic}[width=0.21\textwidth]{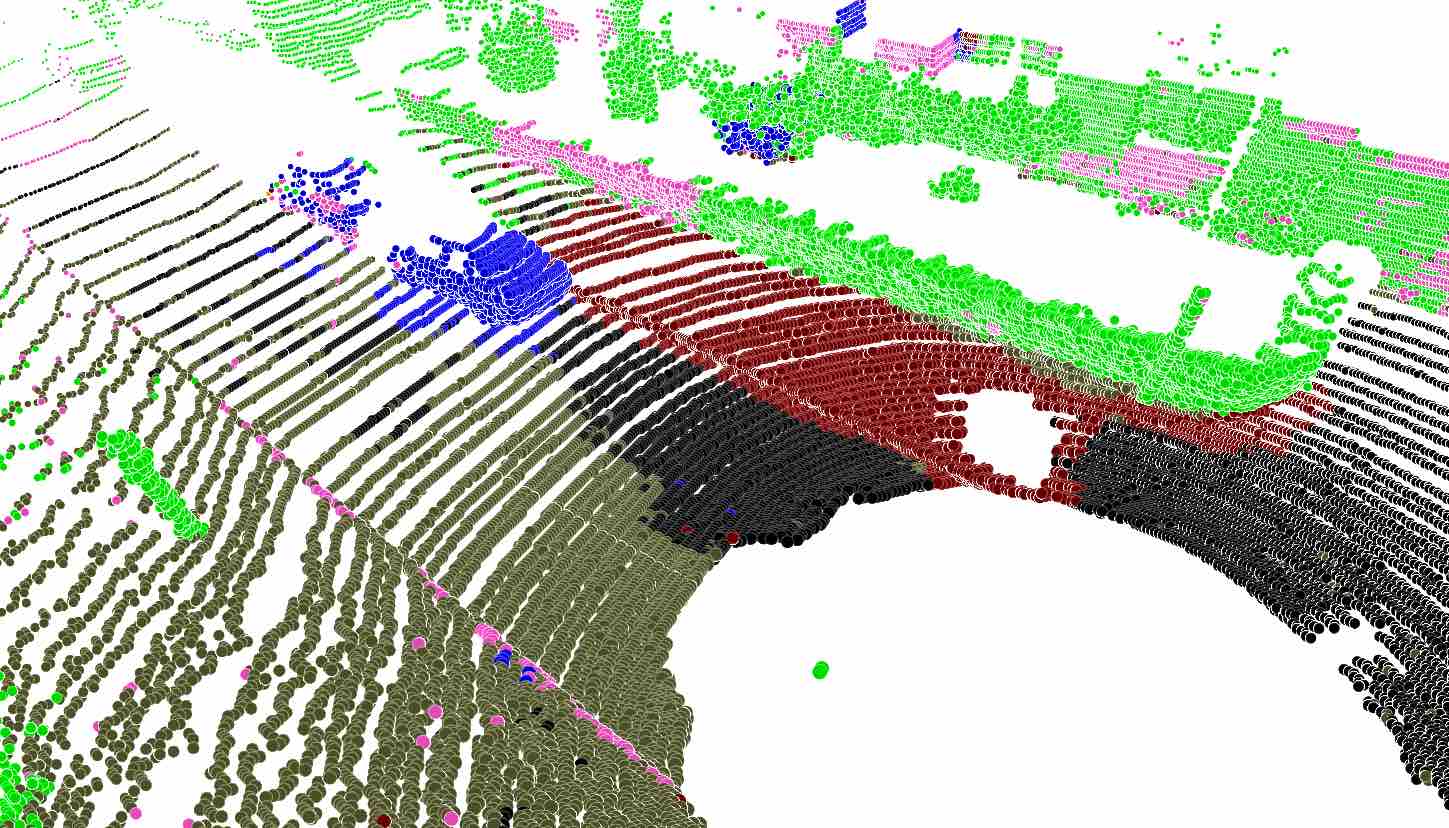}
        \end{overpic} &
        \begin{overpic}[width=0.21\textwidth]{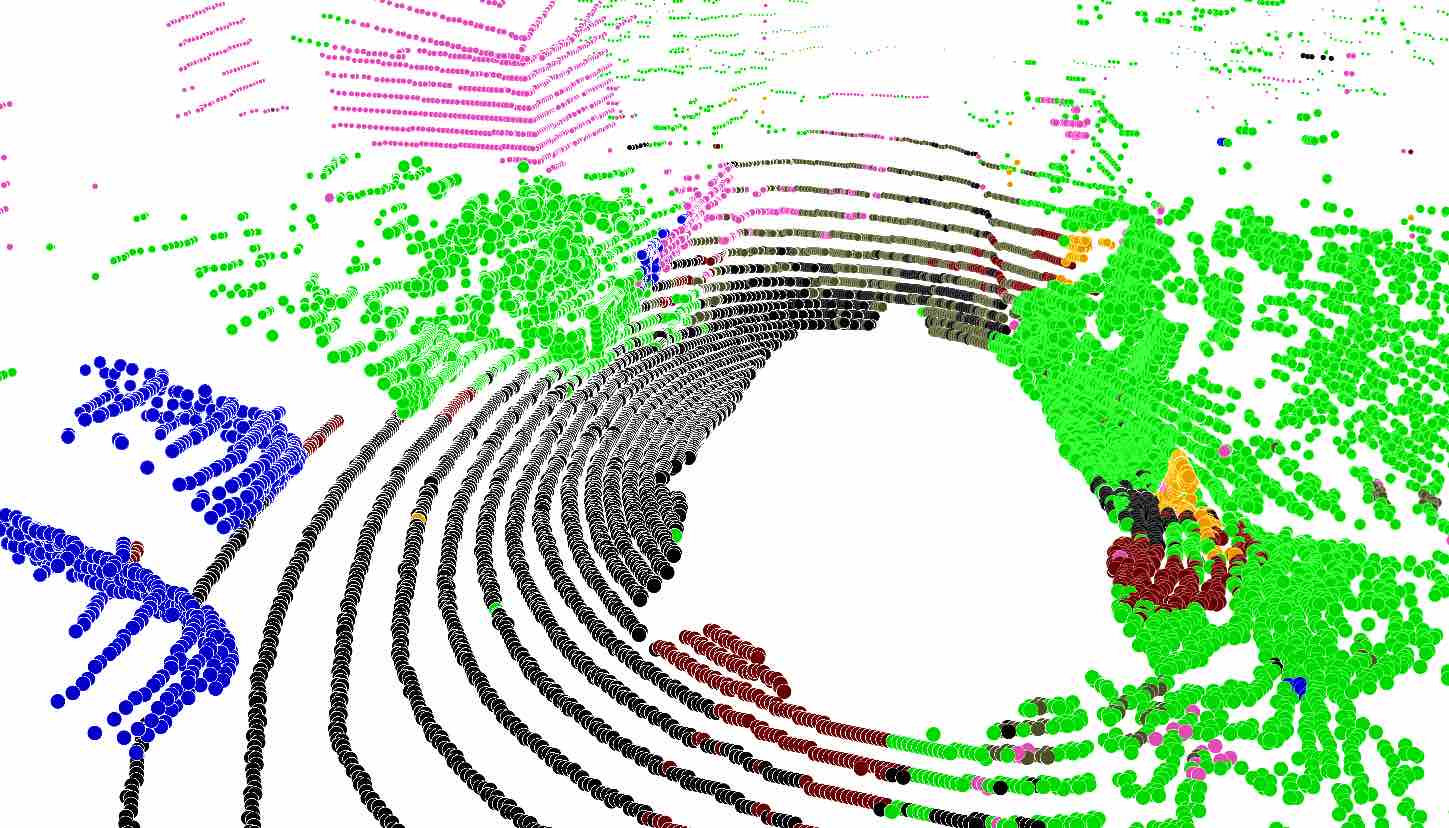}
        \end{overpic}& 
        \begin{overpic}[width=0.21\textwidth]{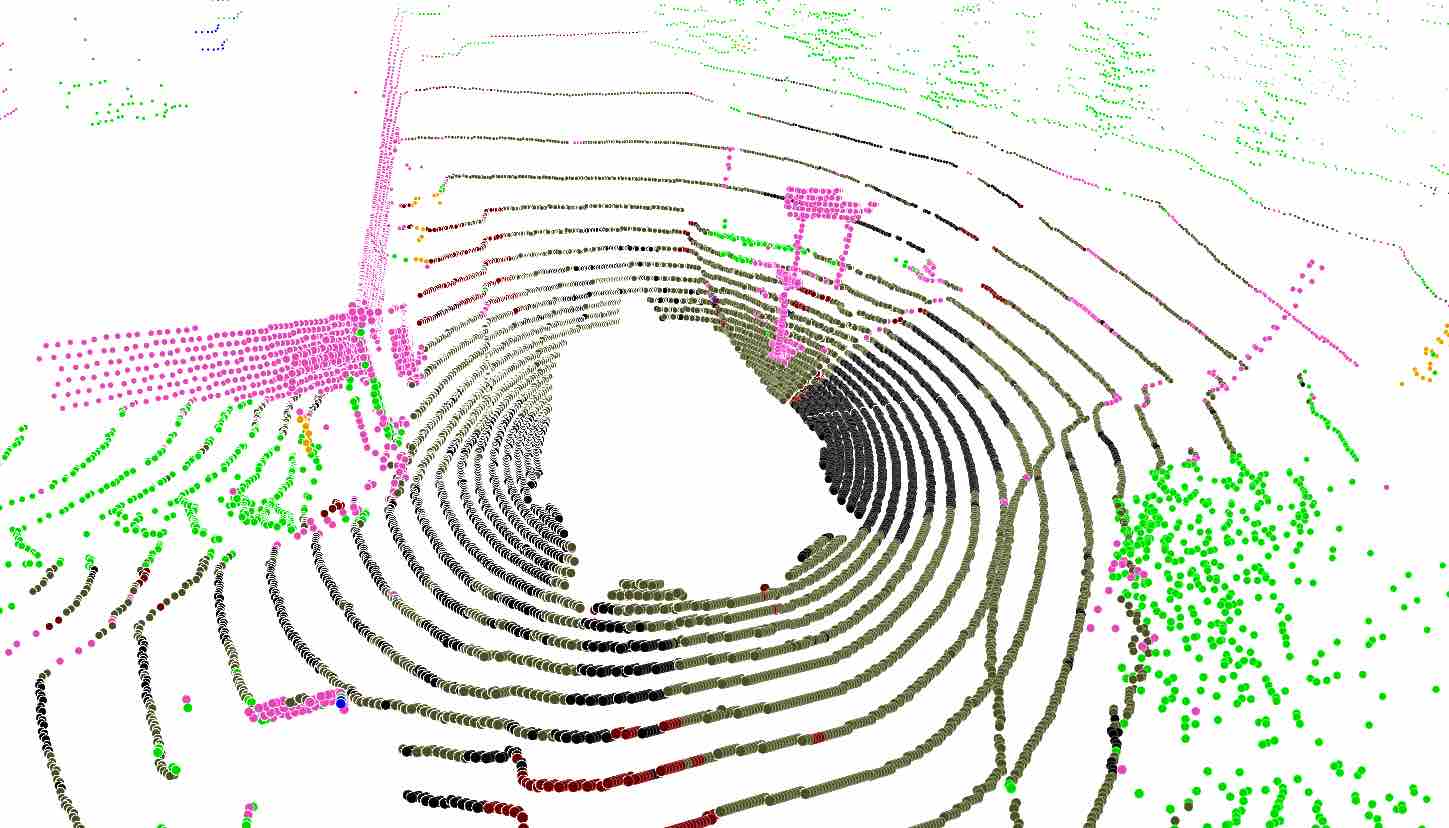}
        \end{overpic}\\
        \begin{overpic}[width=0.21\textwidth]{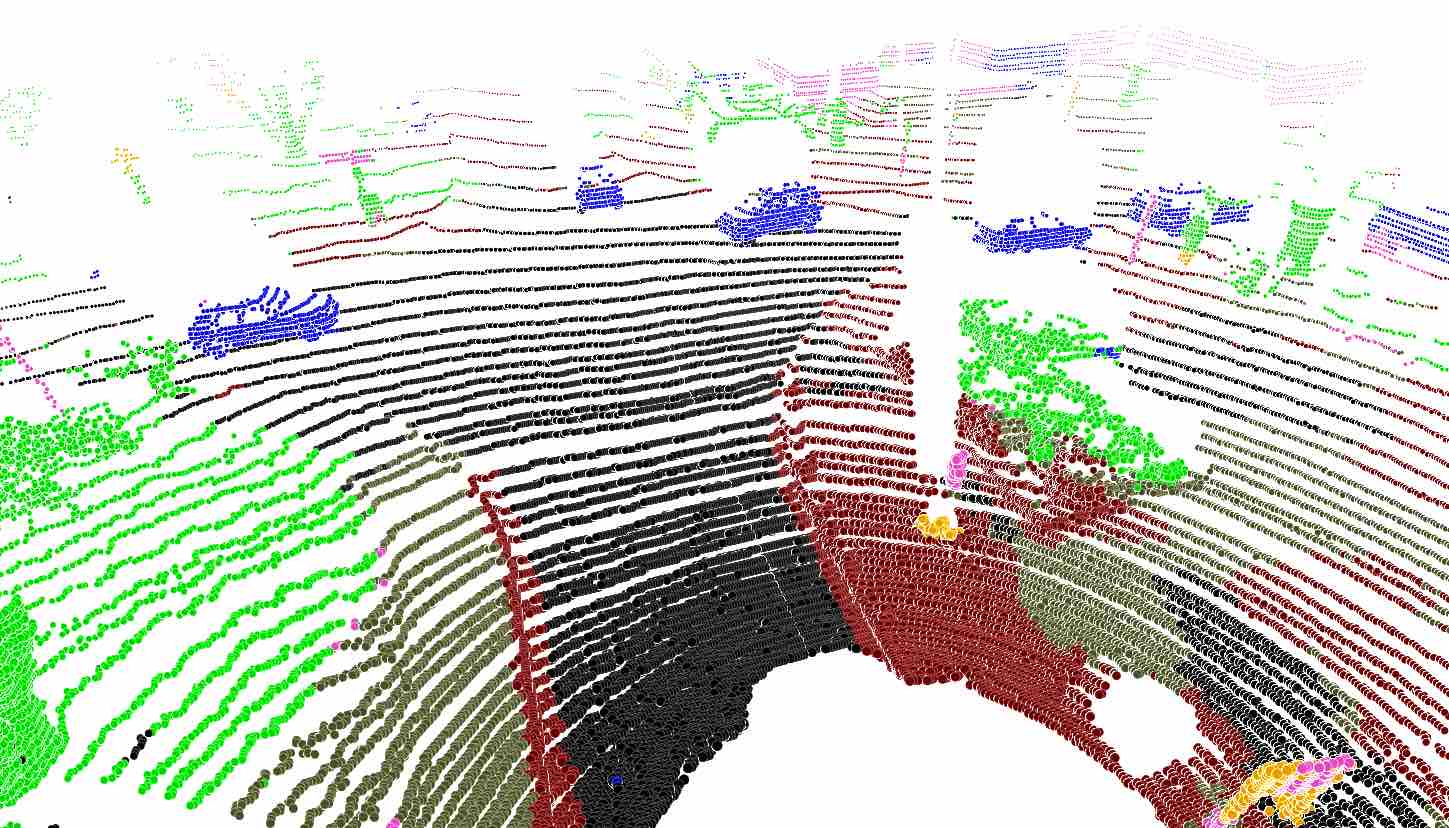}
        \end{overpic} &  
        \begin{overpic}[width=0.21\textwidth]{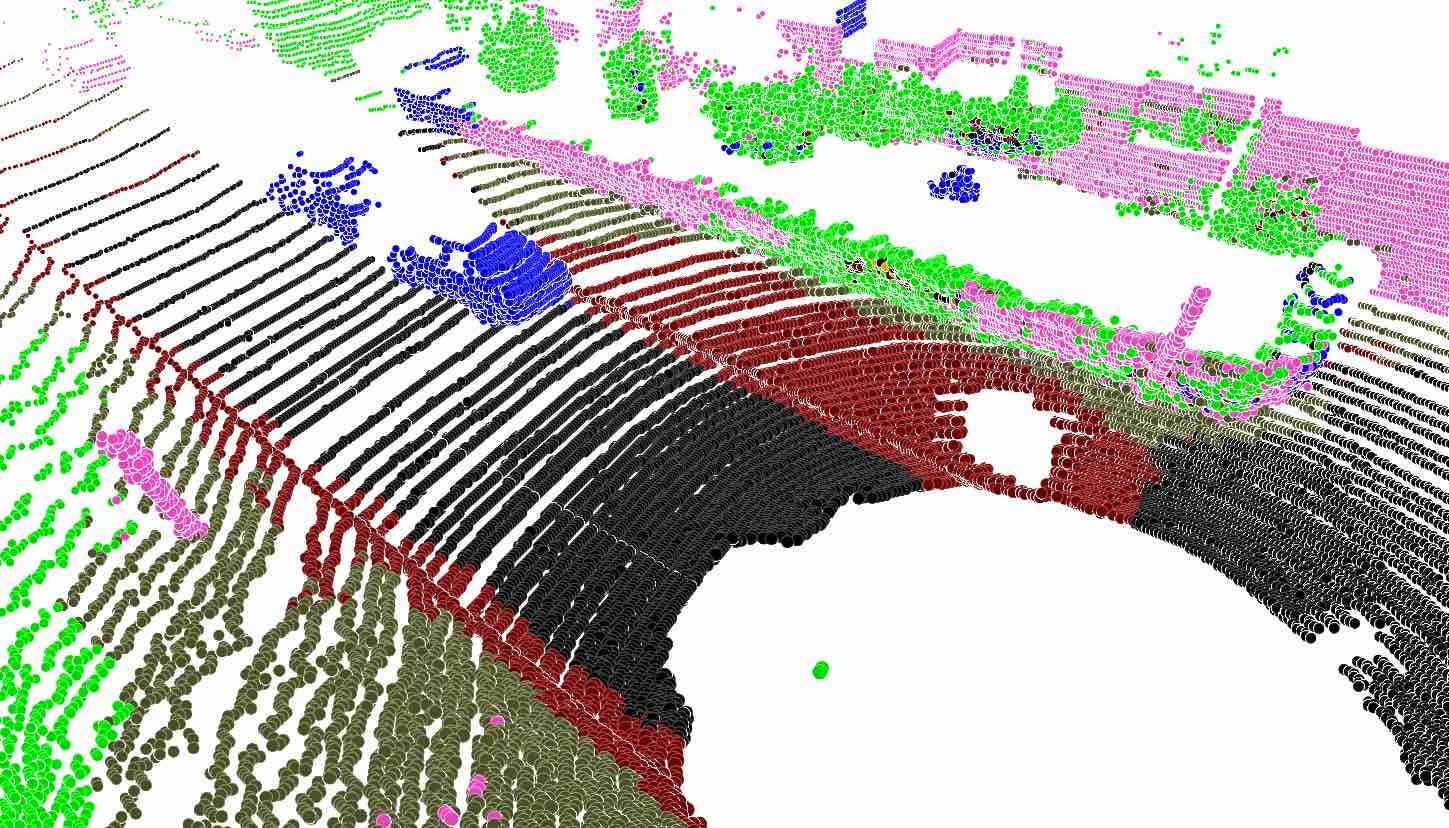}
        \end{overpic} &
        \begin{overpic}[width=0.21\textwidth]{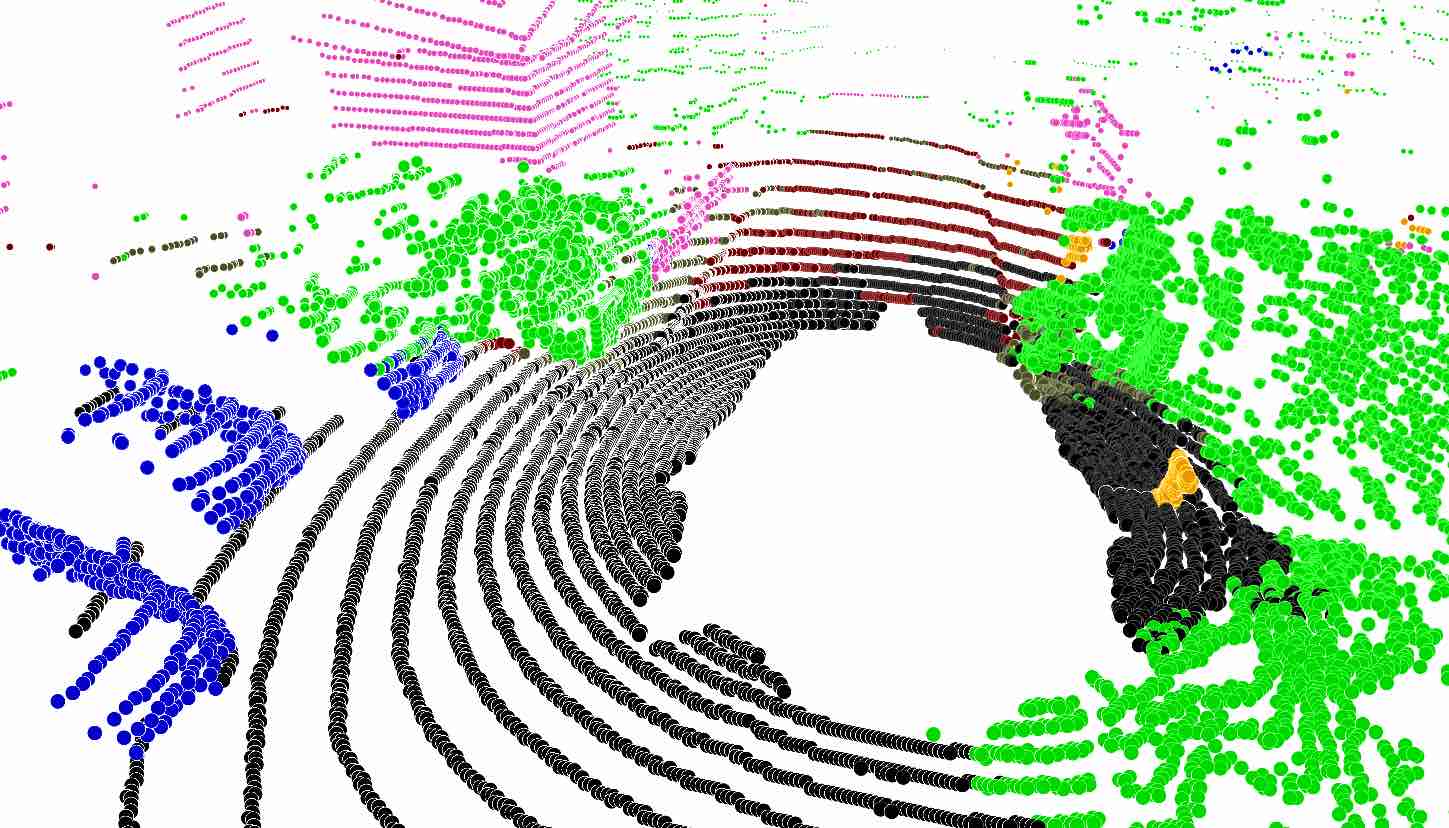}
        \end{overpic}& 
        \begin{overpic}[width=0.21\textwidth]{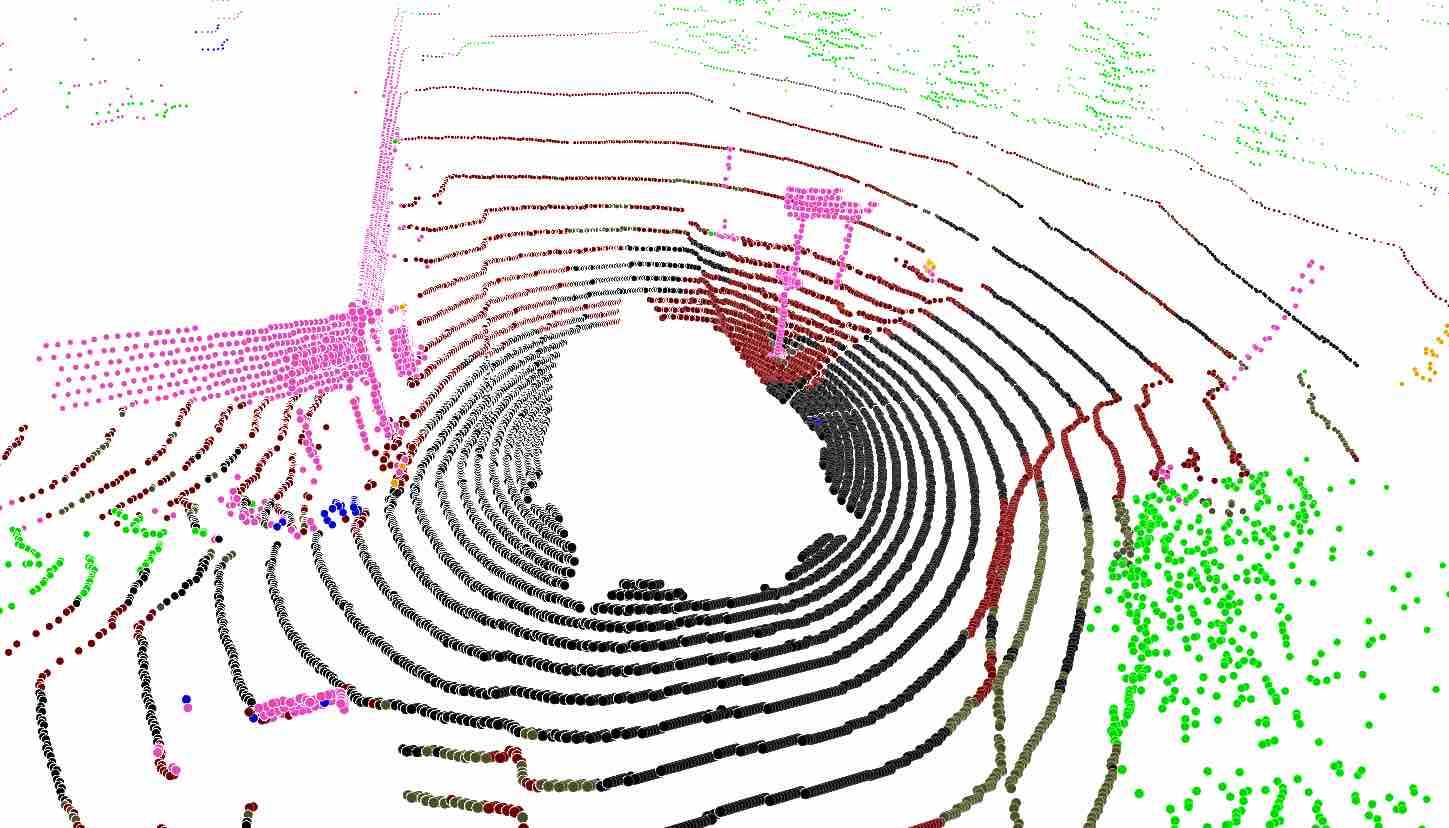}
        \end{overpic}\\
        \begin{overpic}[width=0.21\textwidth]{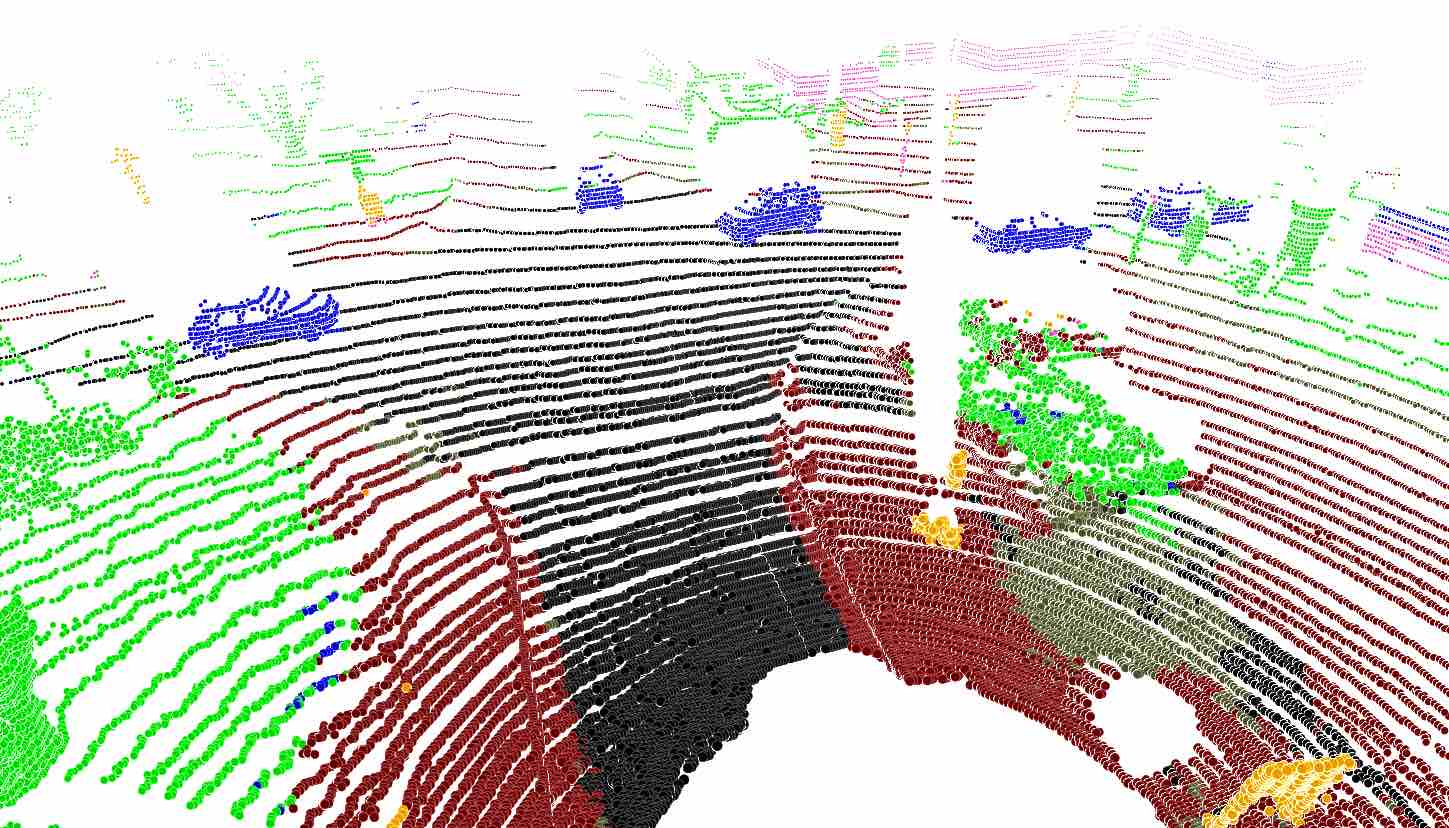}
        \end{overpic} &  
        \begin{overpic}[width=0.21\textwidth]{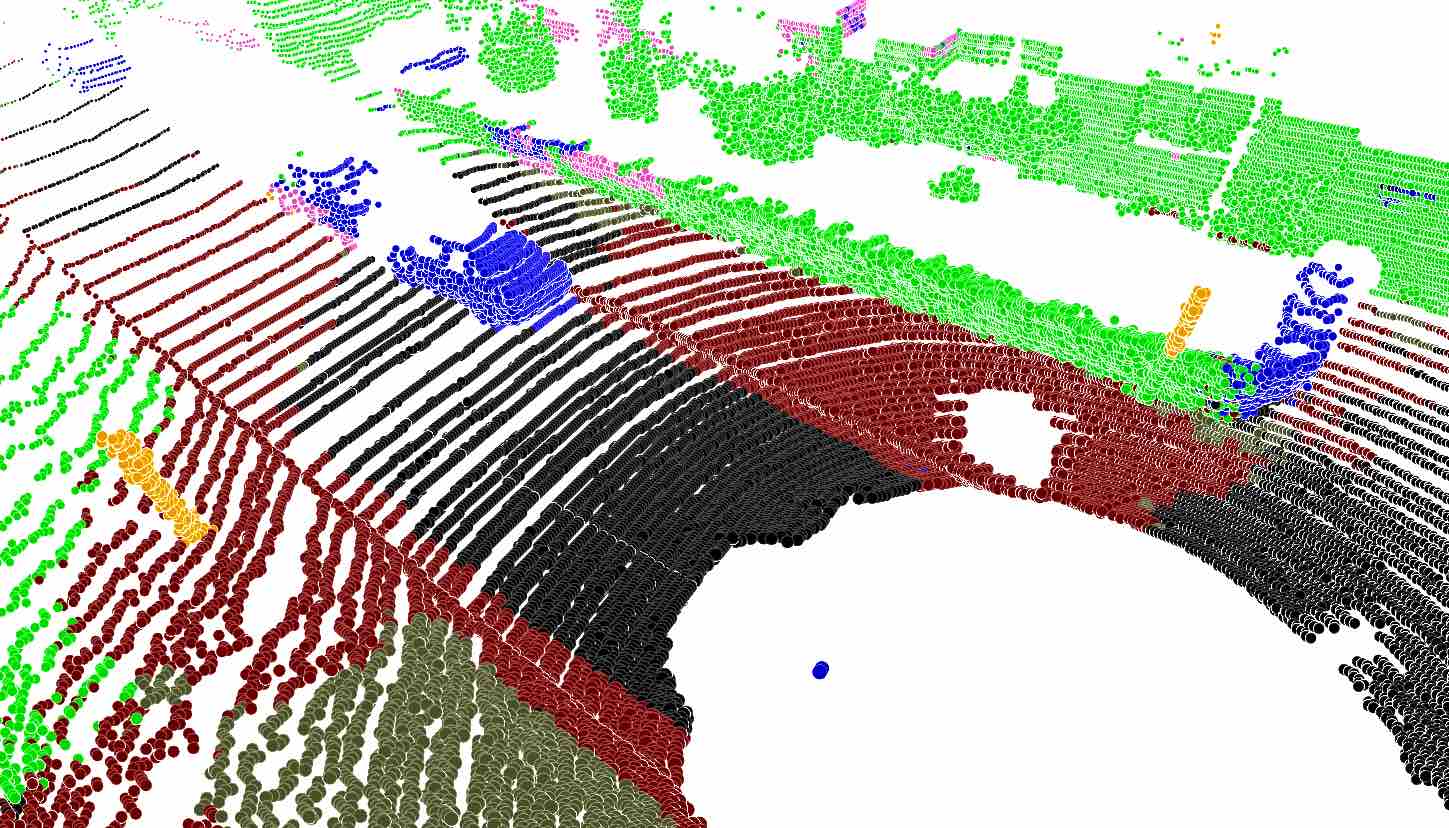}
        \end{overpic} &
        \begin{overpic}[width=0.21\textwidth]{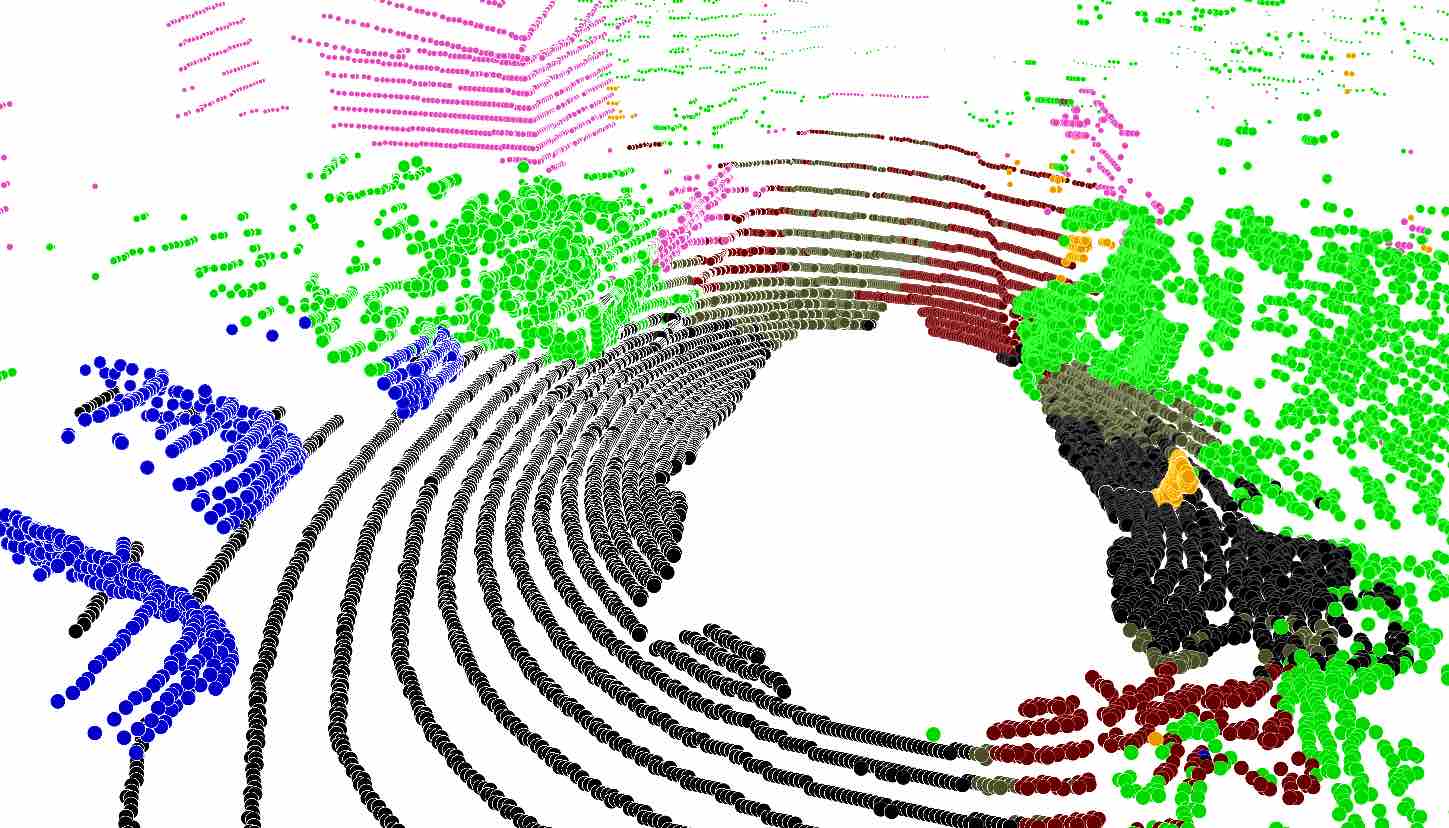}
        \end{overpic}& 
        \begin{overpic}[width=0.21\textwidth]{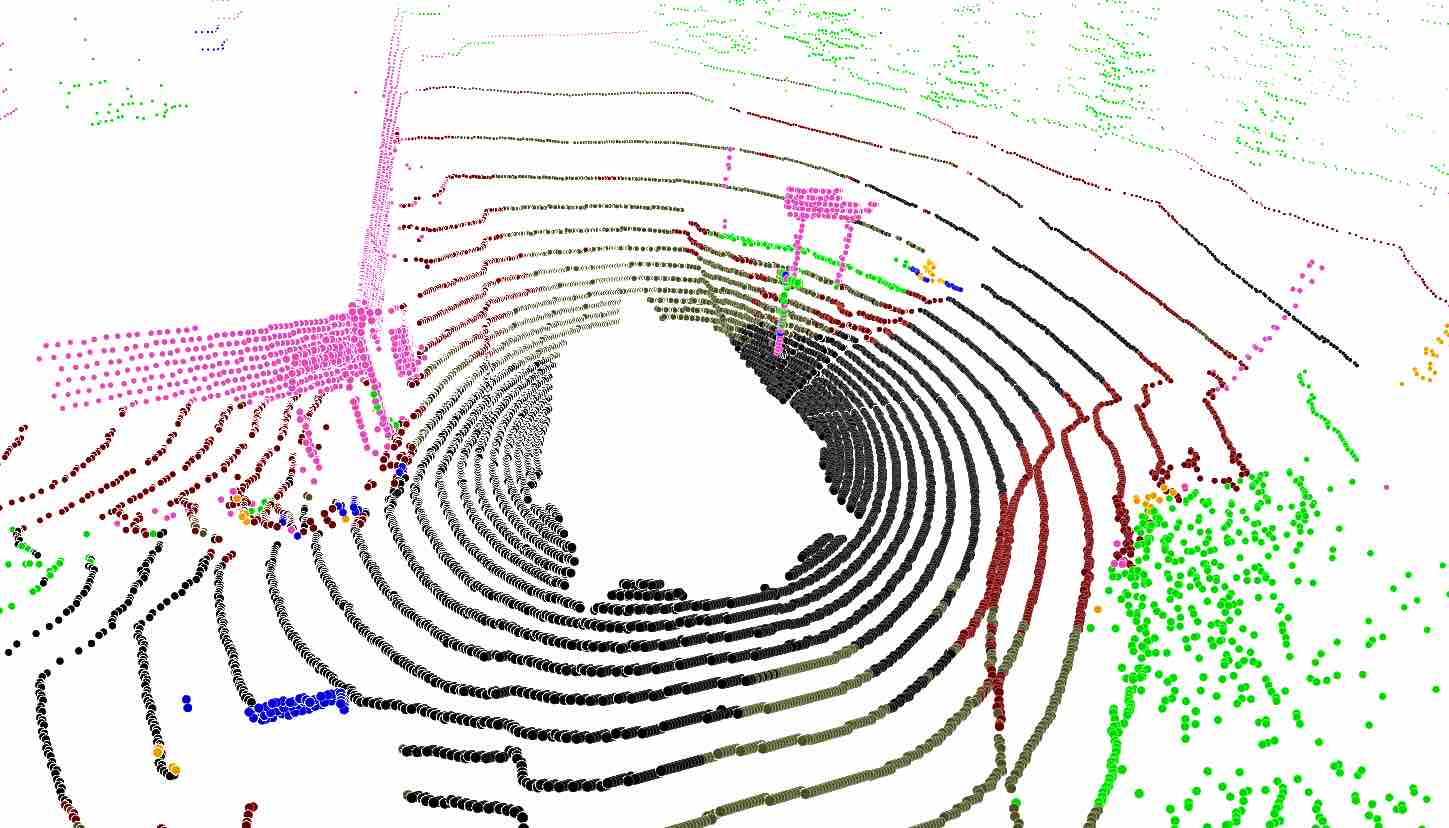}
        \end{overpic}\\
        \begin{overpic}[width=0.21\textwidth]{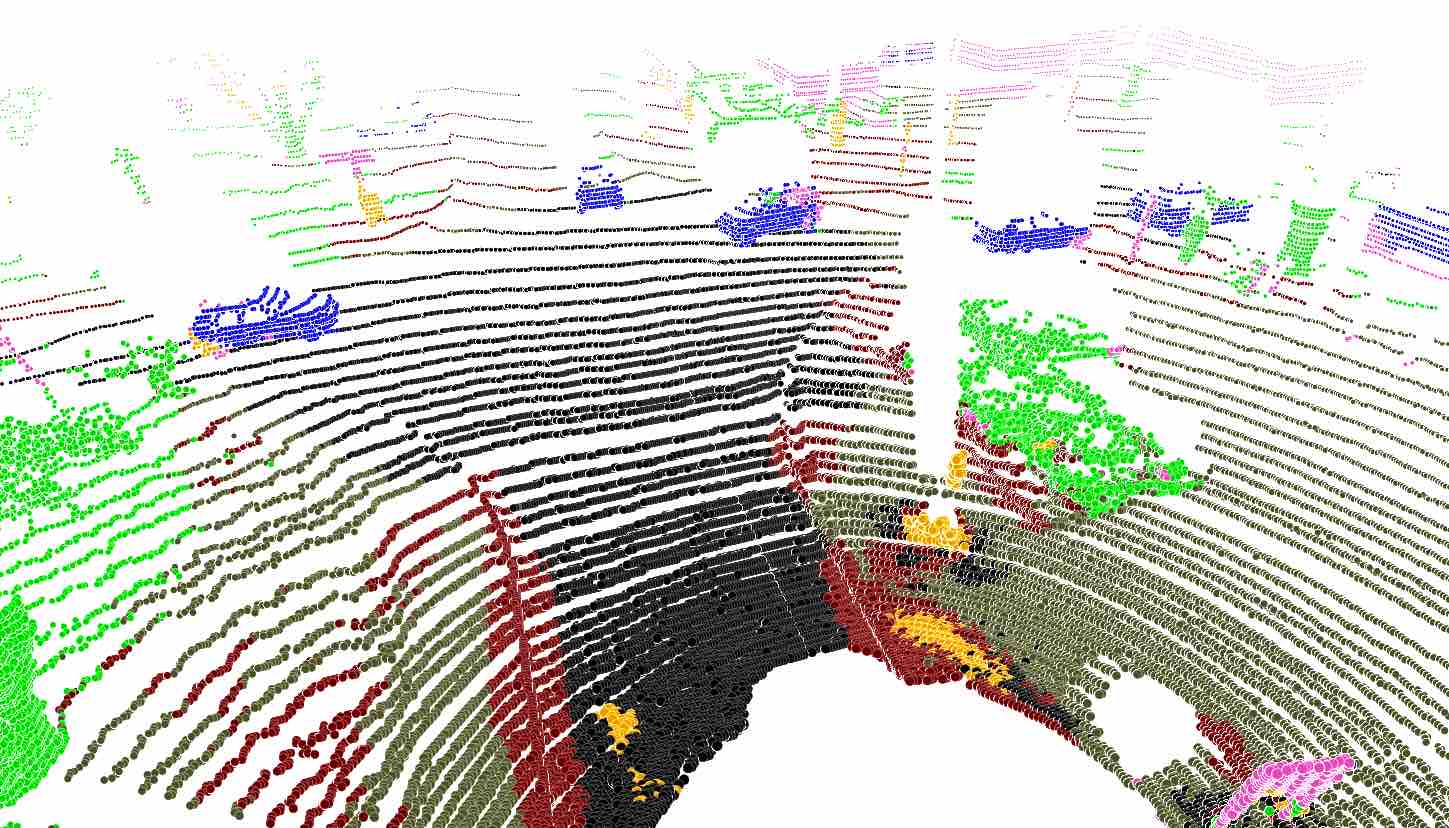}
        \end{overpic} &  
        \begin{overpic}[width=0.21\textwidth]{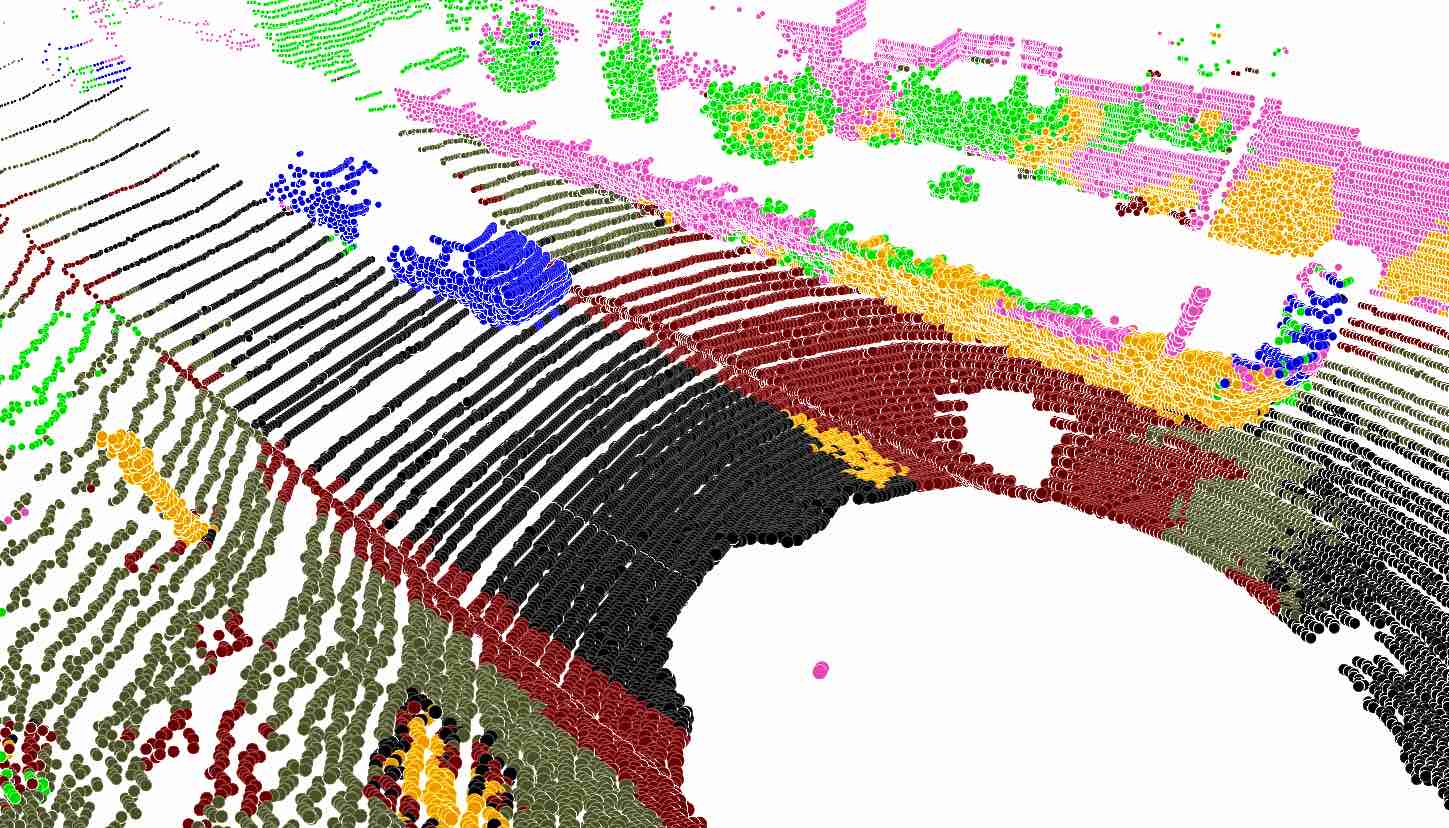}
        \end{overpic} &
        \begin{overpic}[width=0.21\textwidth]{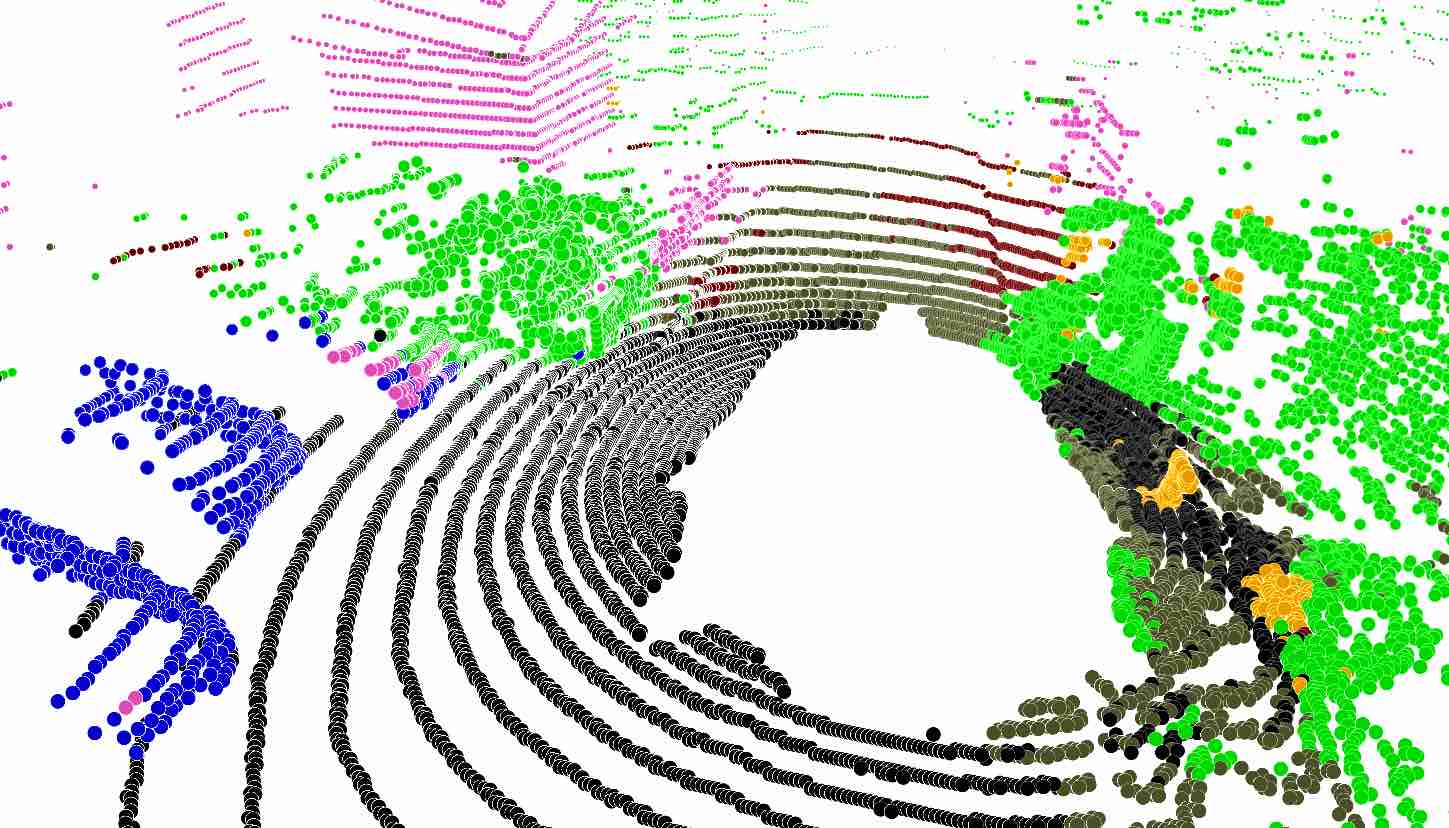}
        \end{overpic}& 
        \begin{overpic}[width=0.21\textwidth]{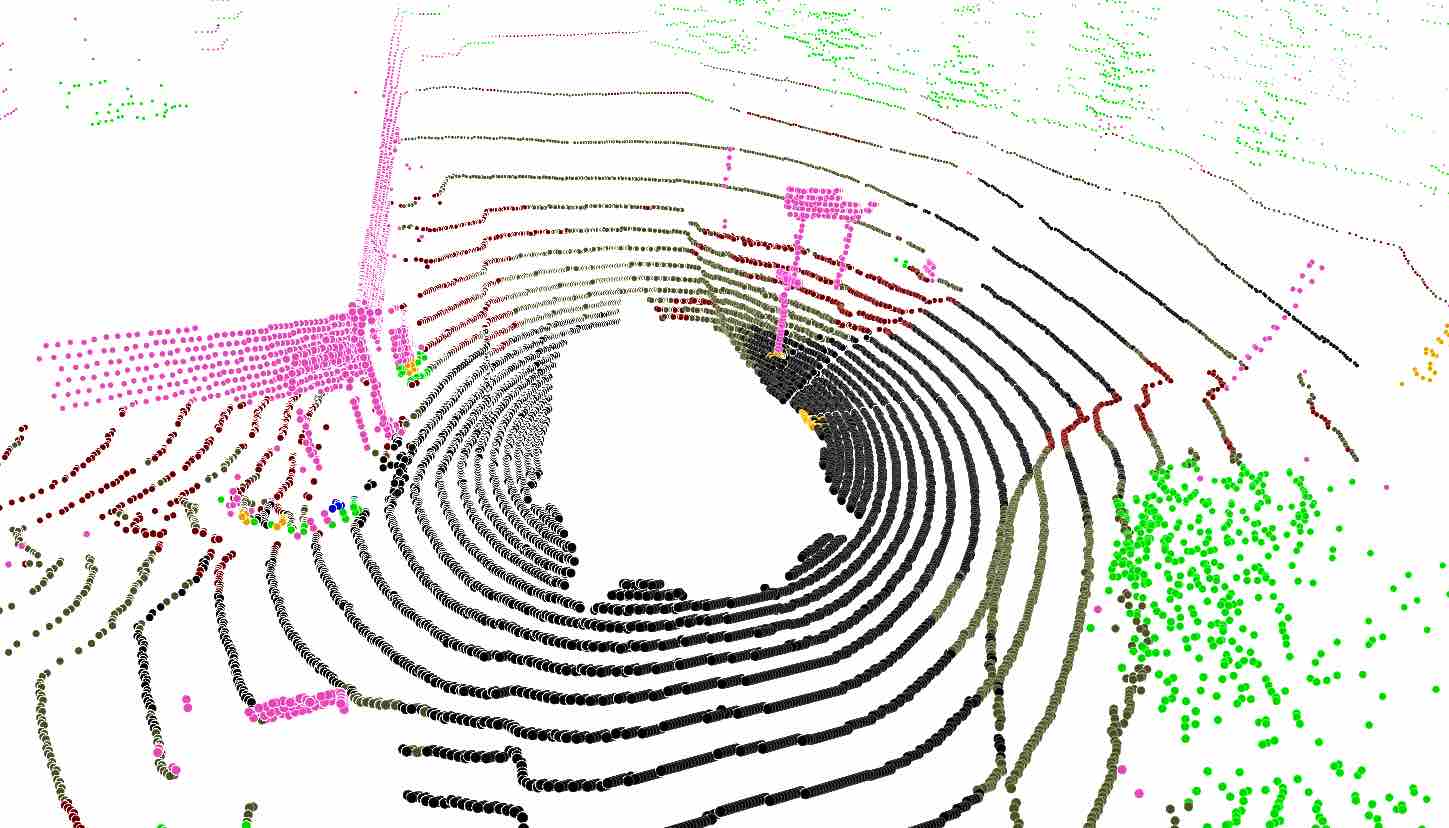}
        \end{overpic}\\
        \begin{overpic}[width=0.21\textwidth]{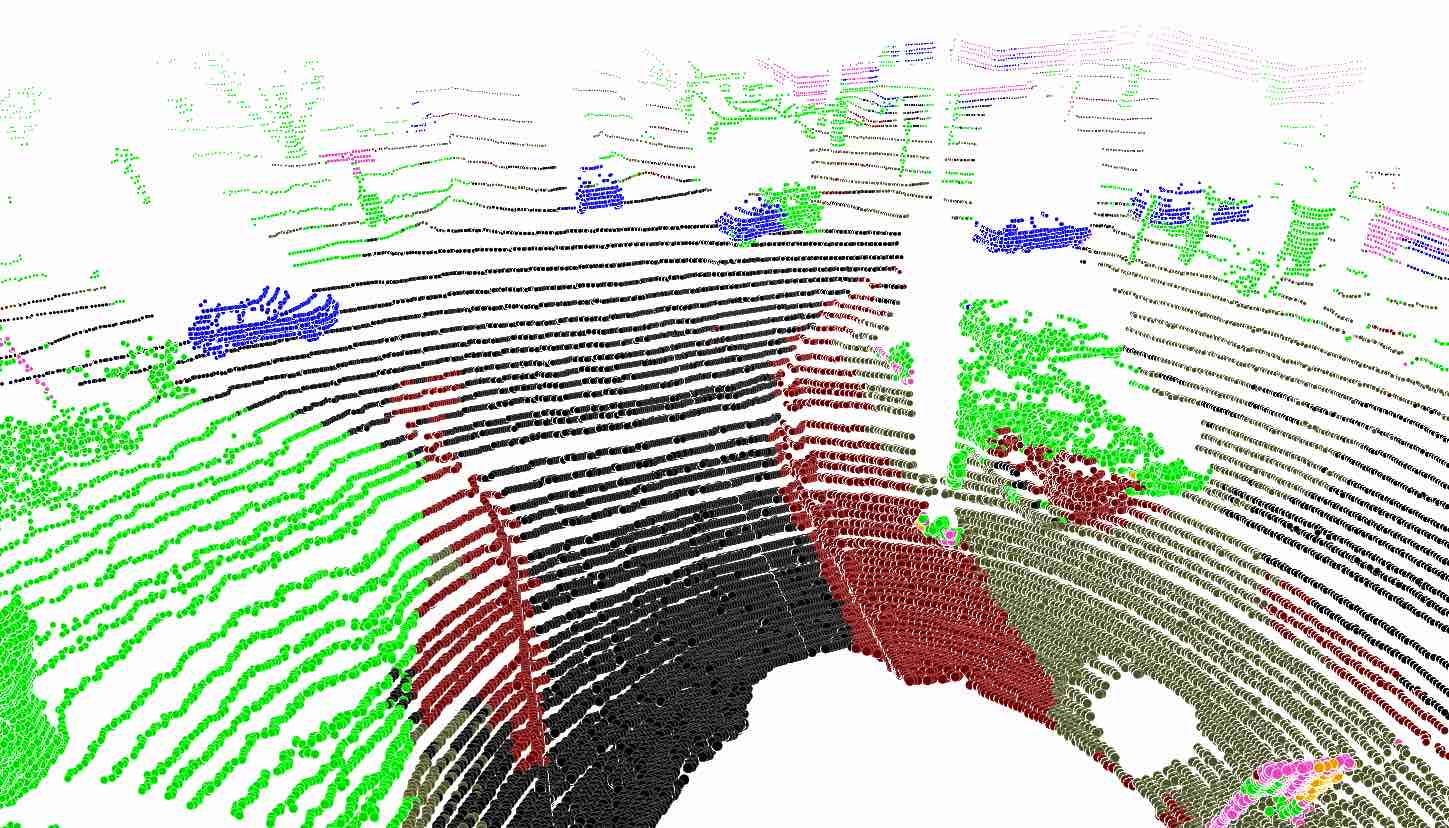}
        \end{overpic} &  
        \begin{overpic}[width=0.21\textwidth]{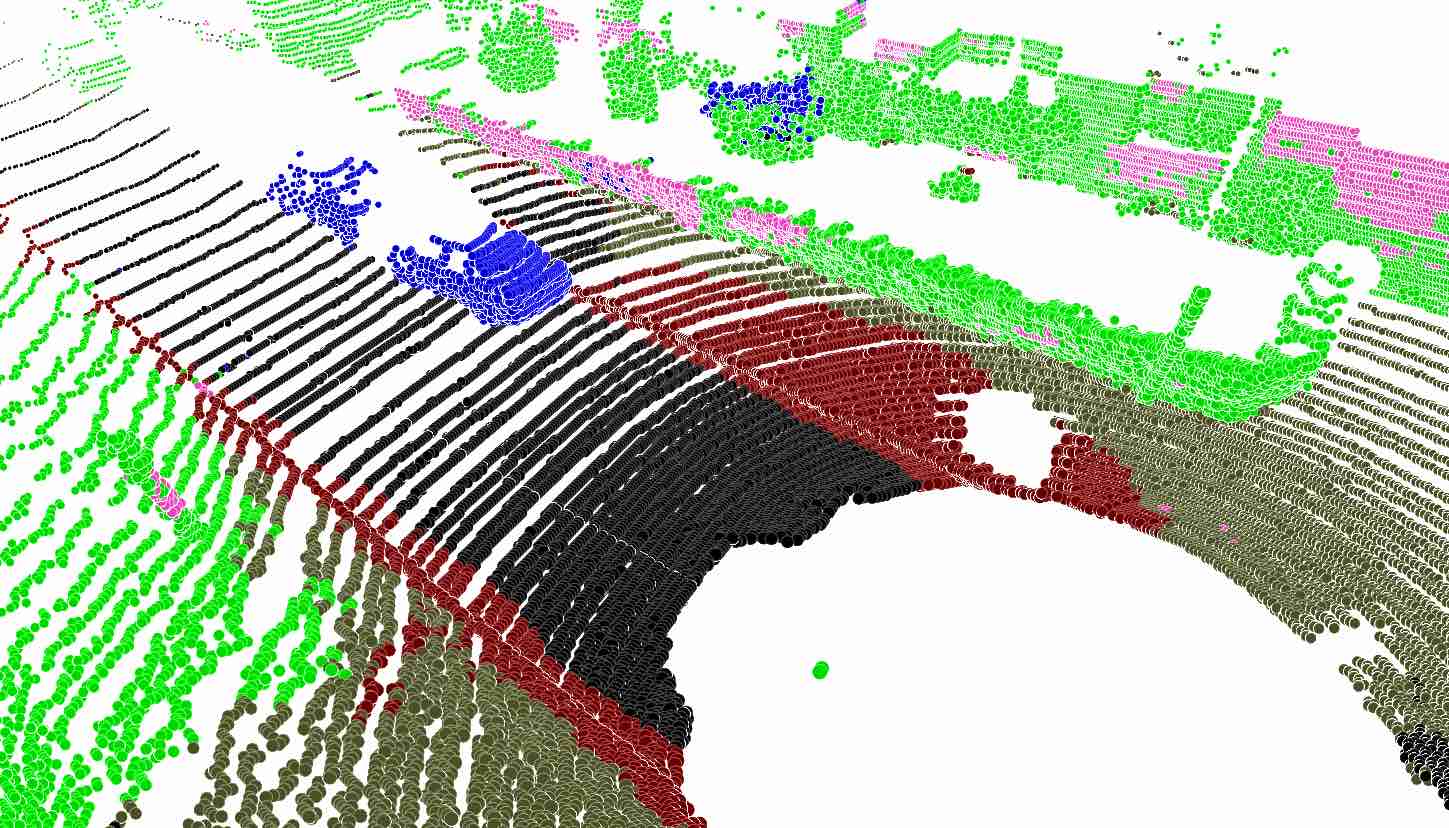}
        \end{overpic} &
        \begin{overpic}[width=0.21\textwidth]{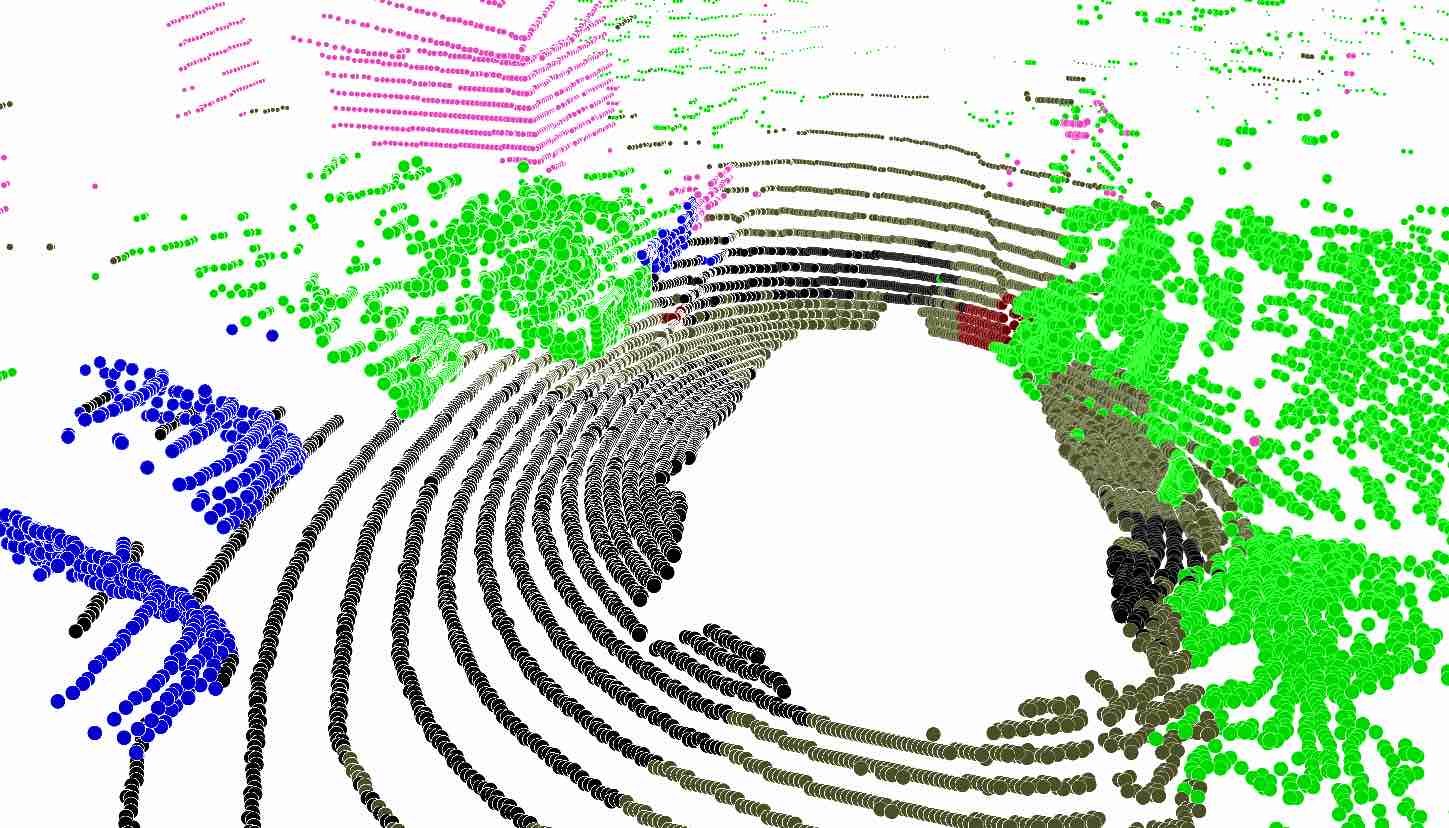}
        \end{overpic}& 
        \begin{overpic}[width=0.21\textwidth]{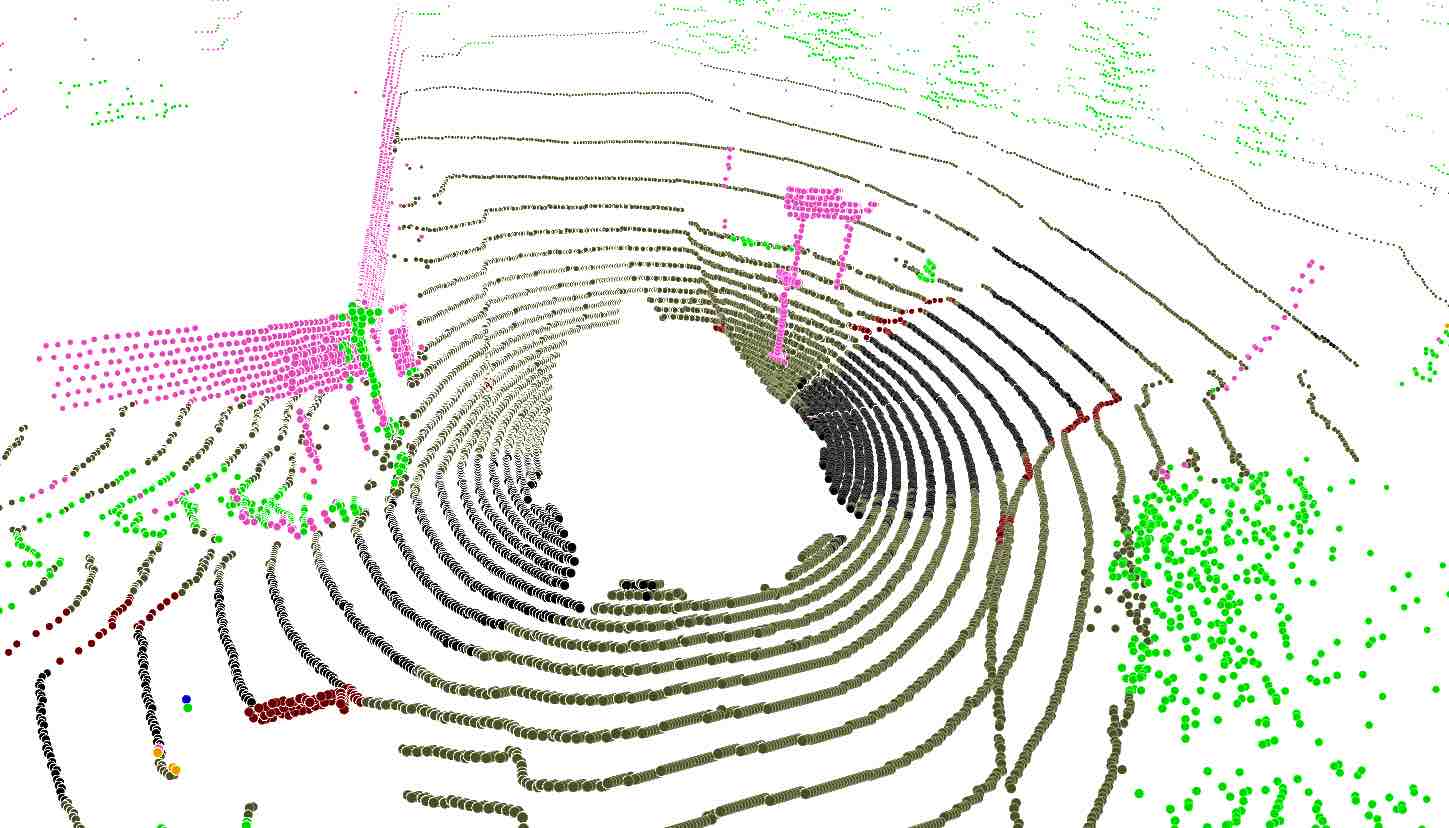}
        \end{overpic}\\
        \begin{overpic}[width=0.21\textwidth]{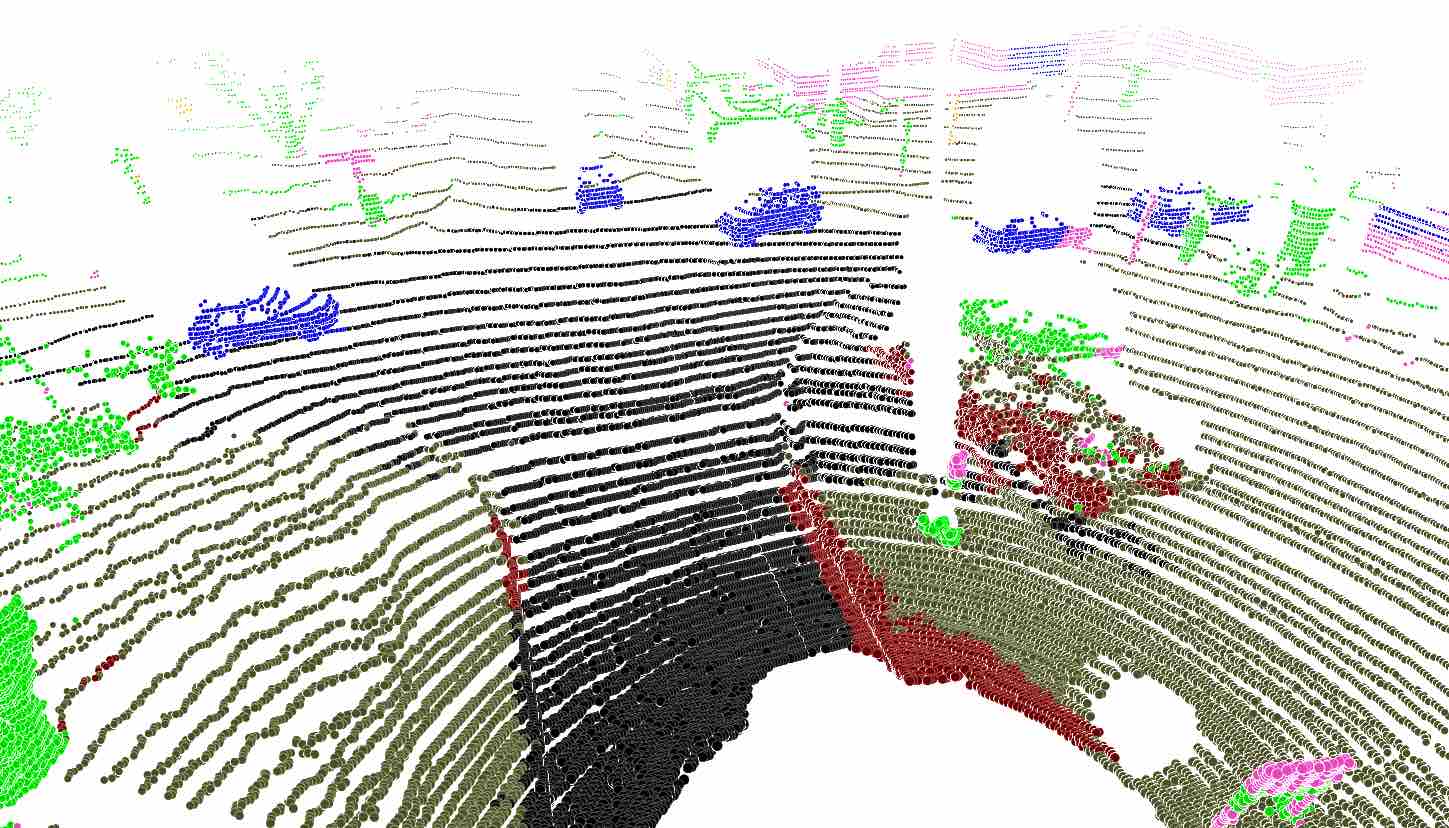}
        \end{overpic} &  
        \begin{overpic}[width=0.21\textwidth]{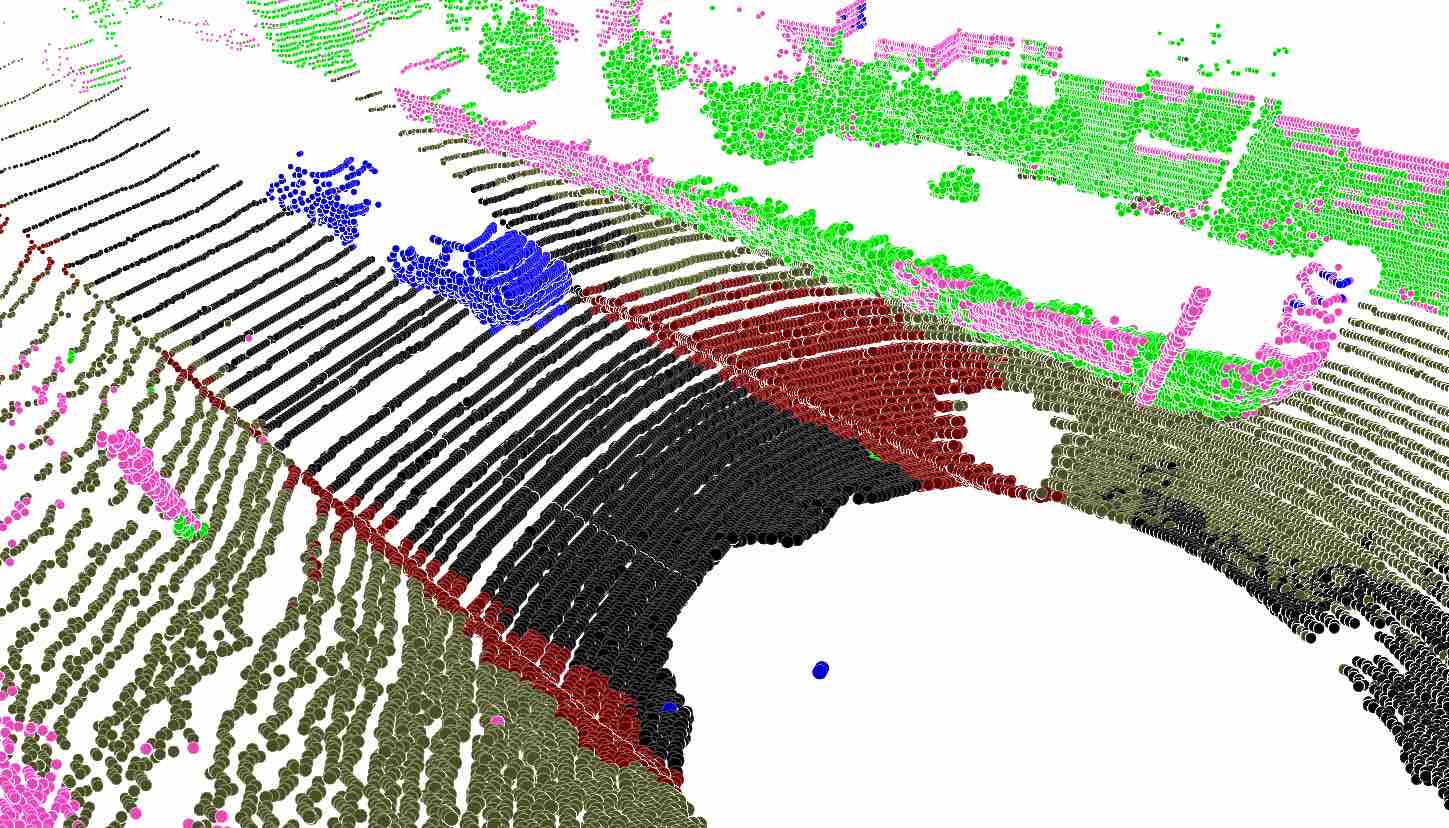}
        \end{overpic} &
        \begin{overpic}[width=0.21\textwidth]{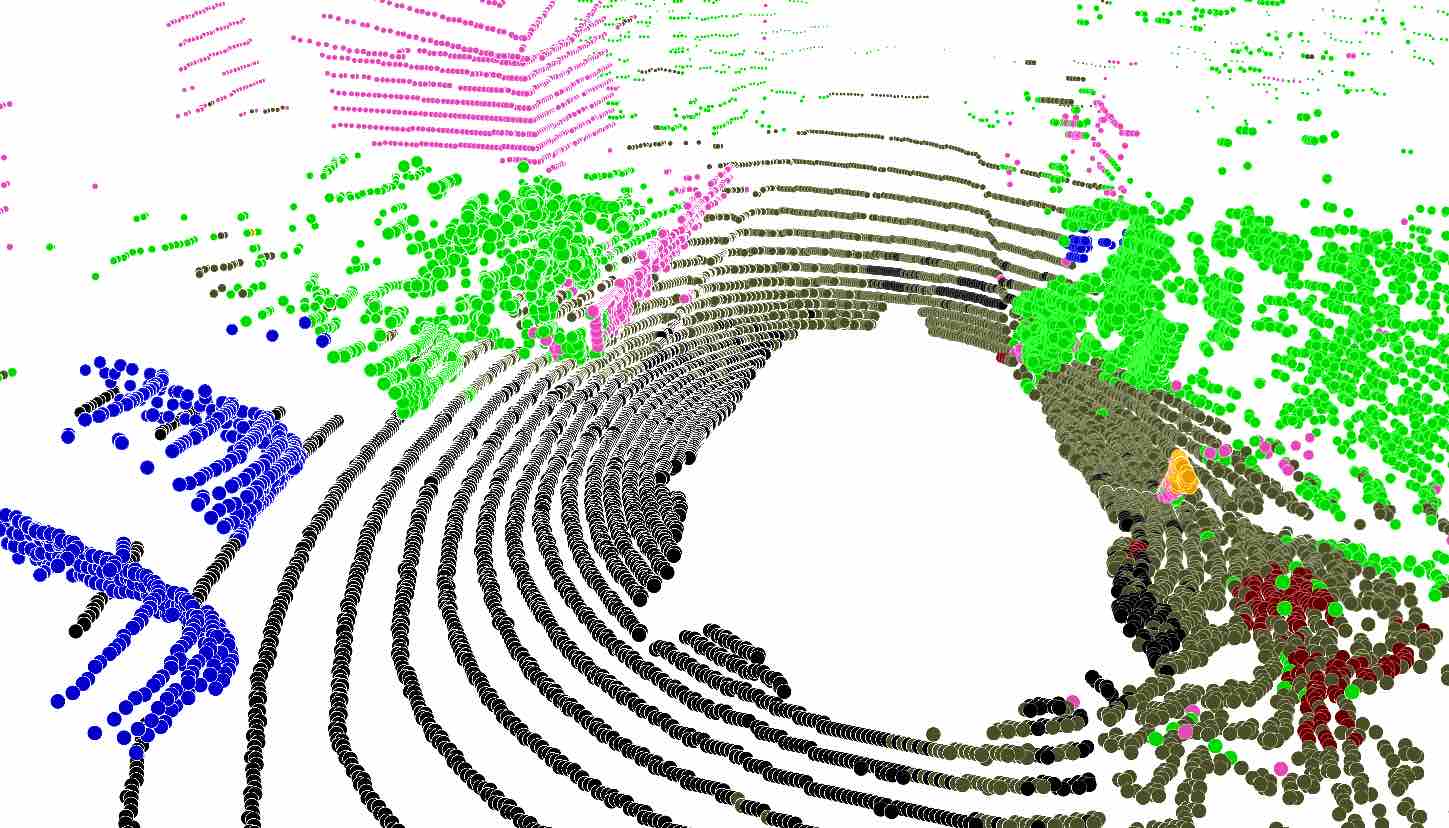}
        \end{overpic}& 
        \begin{overpic}[width=0.21\textwidth]{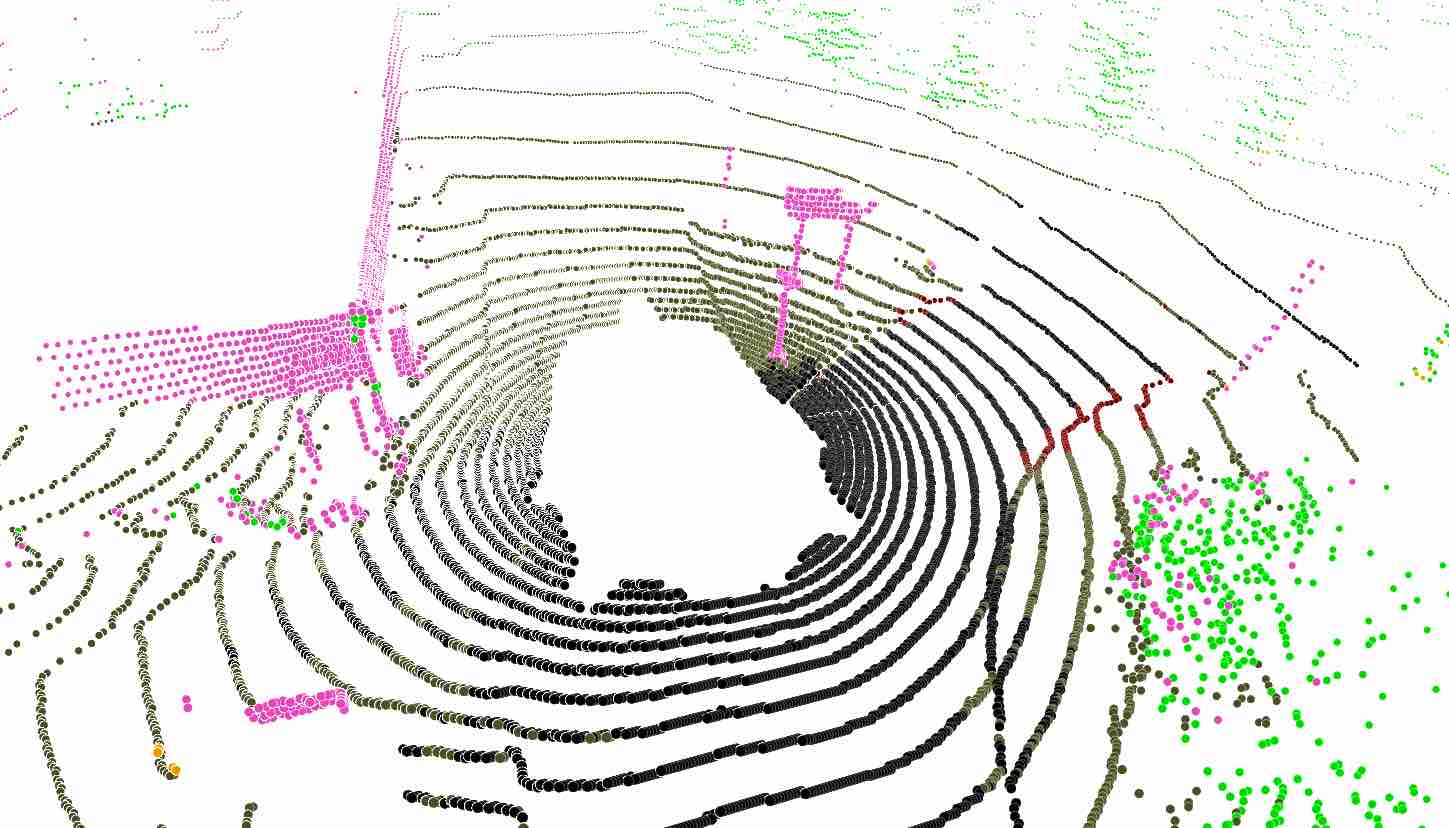}
        \end{overpic}\\
        \begin{overpic}[width=0.21\textwidth]{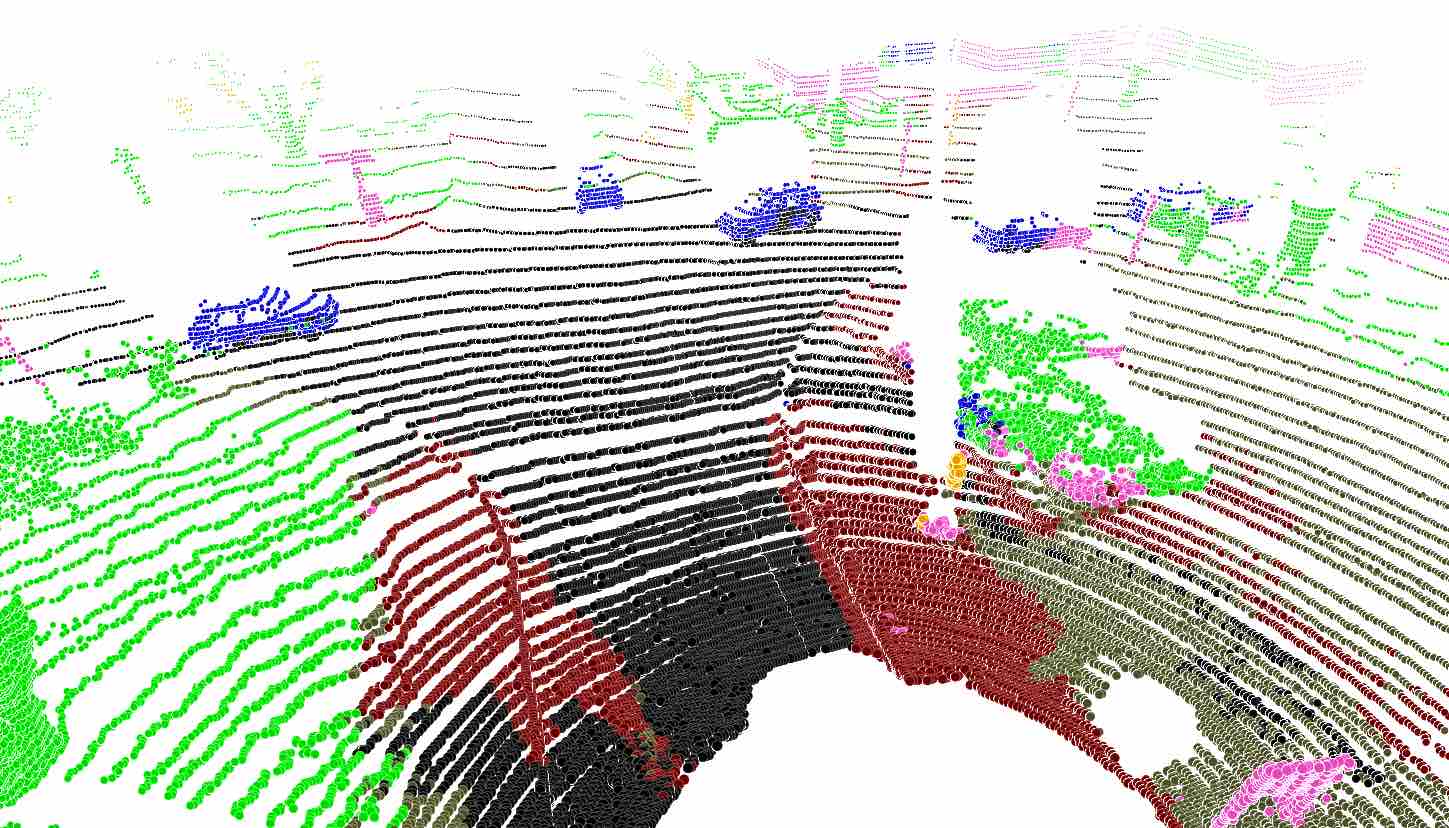}
        \end{overpic} &  
        \begin{overpic}[width=0.21\textwidth]{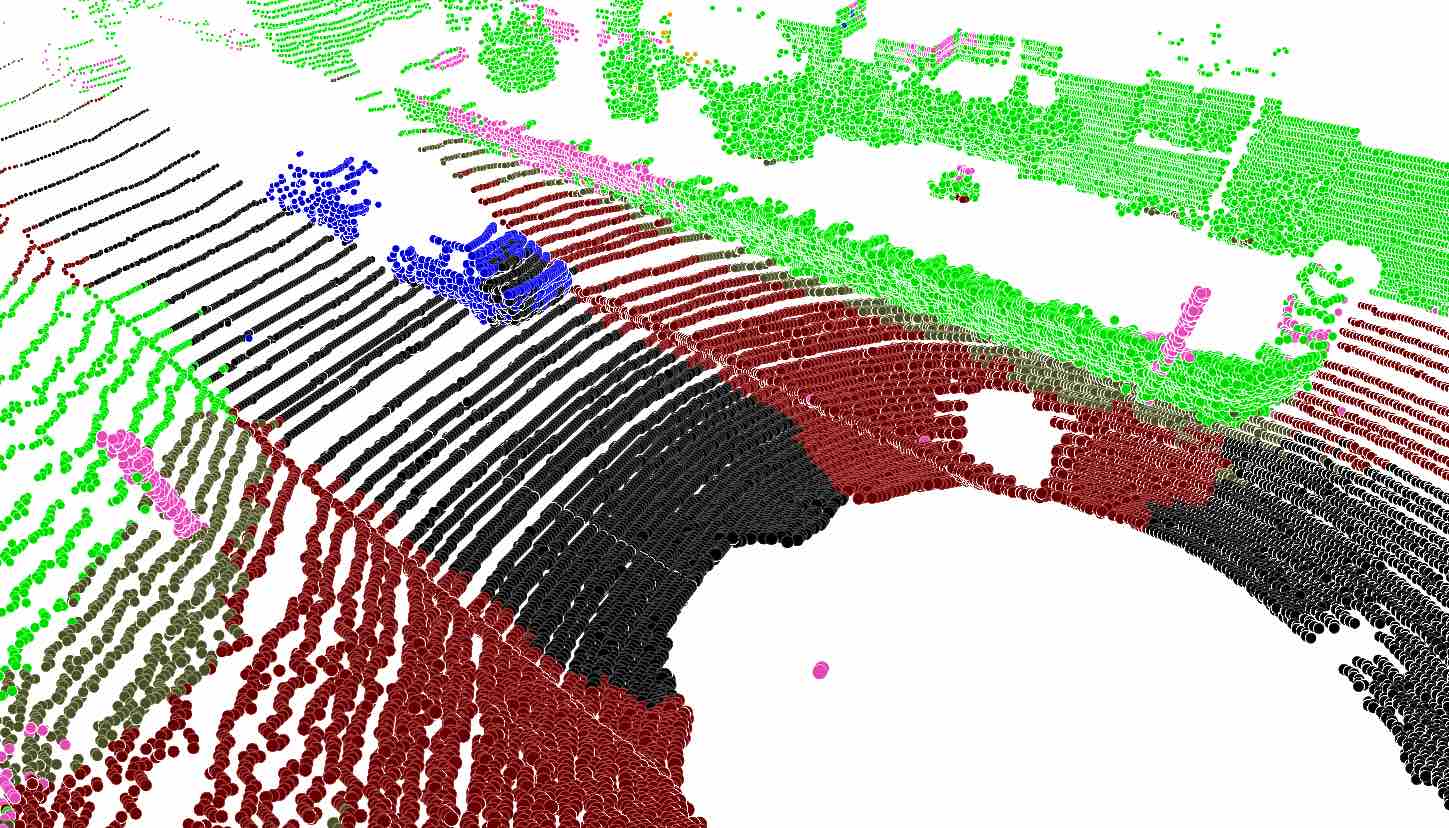}
        \end{overpic} &
        \begin{overpic}[width=0.21\textwidth]{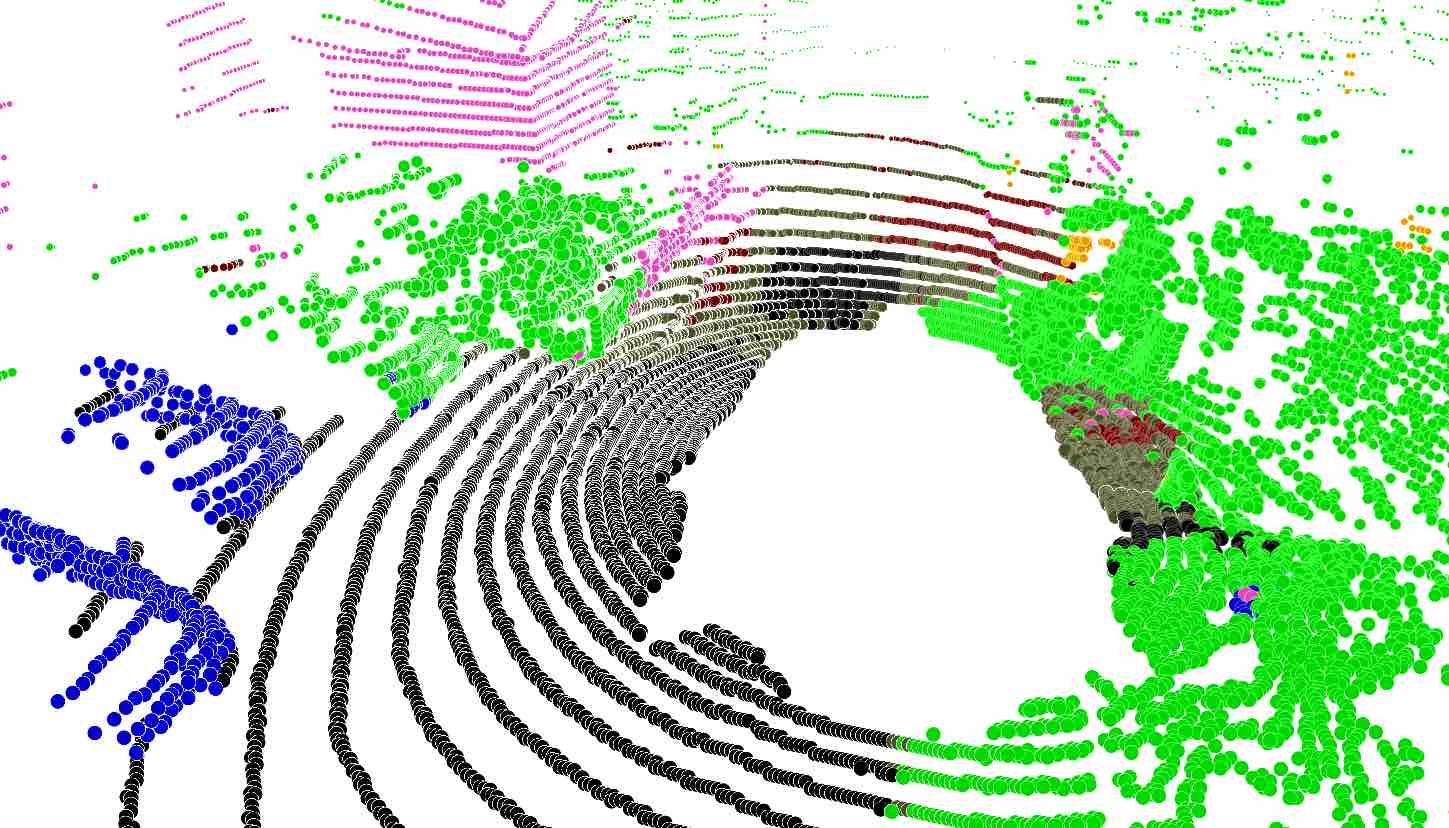}
        \end{overpic}& 
        \begin{overpic}[width=0.21\textwidth]{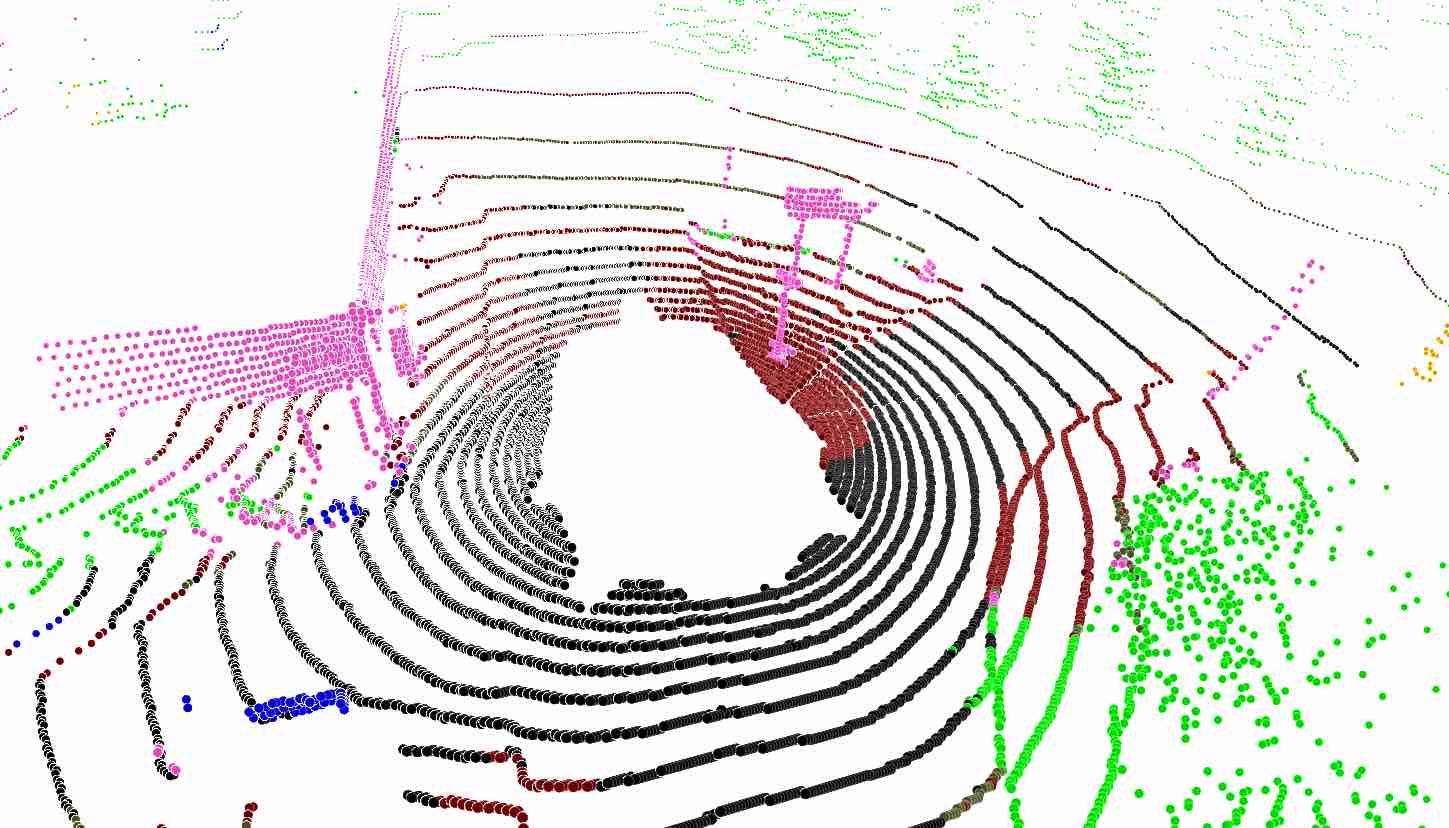}
        \end{overpic}\\
        \begin{overpic}[width=0.21\textwidth]{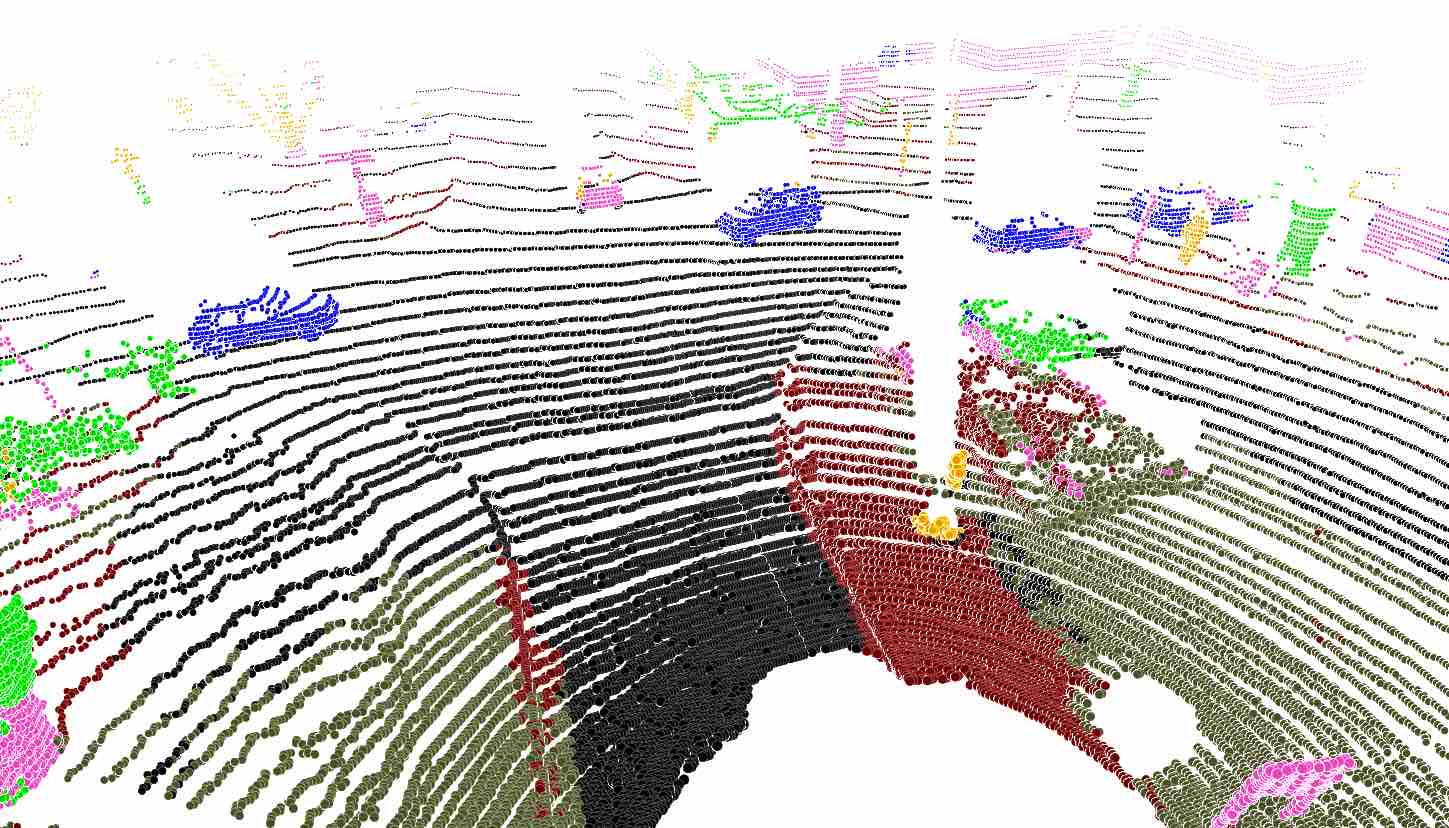}
        \end{overpic} &  
        \begin{overpic}[width=0.21\textwidth]{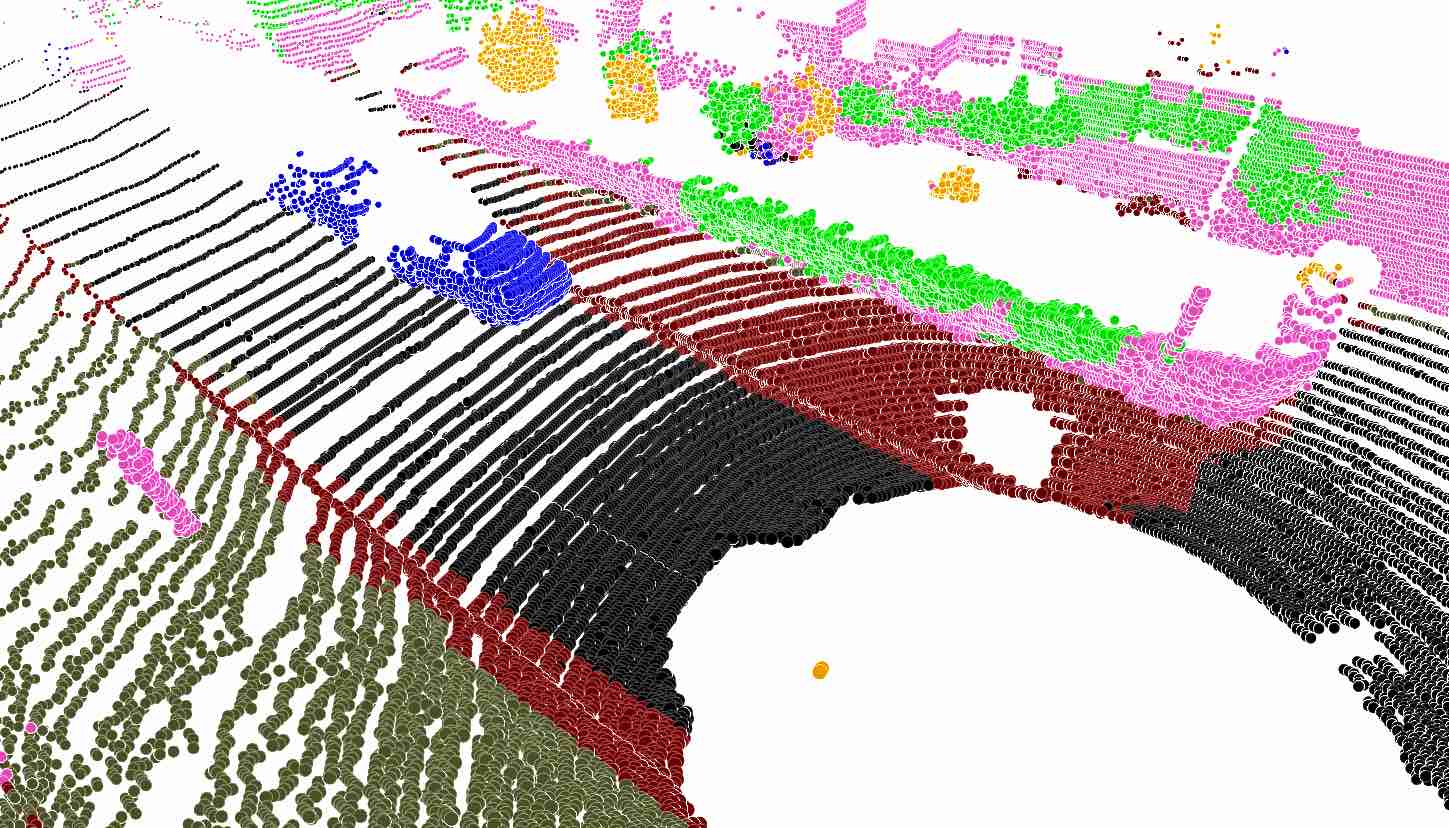}
        \end{overpic} &
        \begin{overpic}[width=0.21\textwidth]{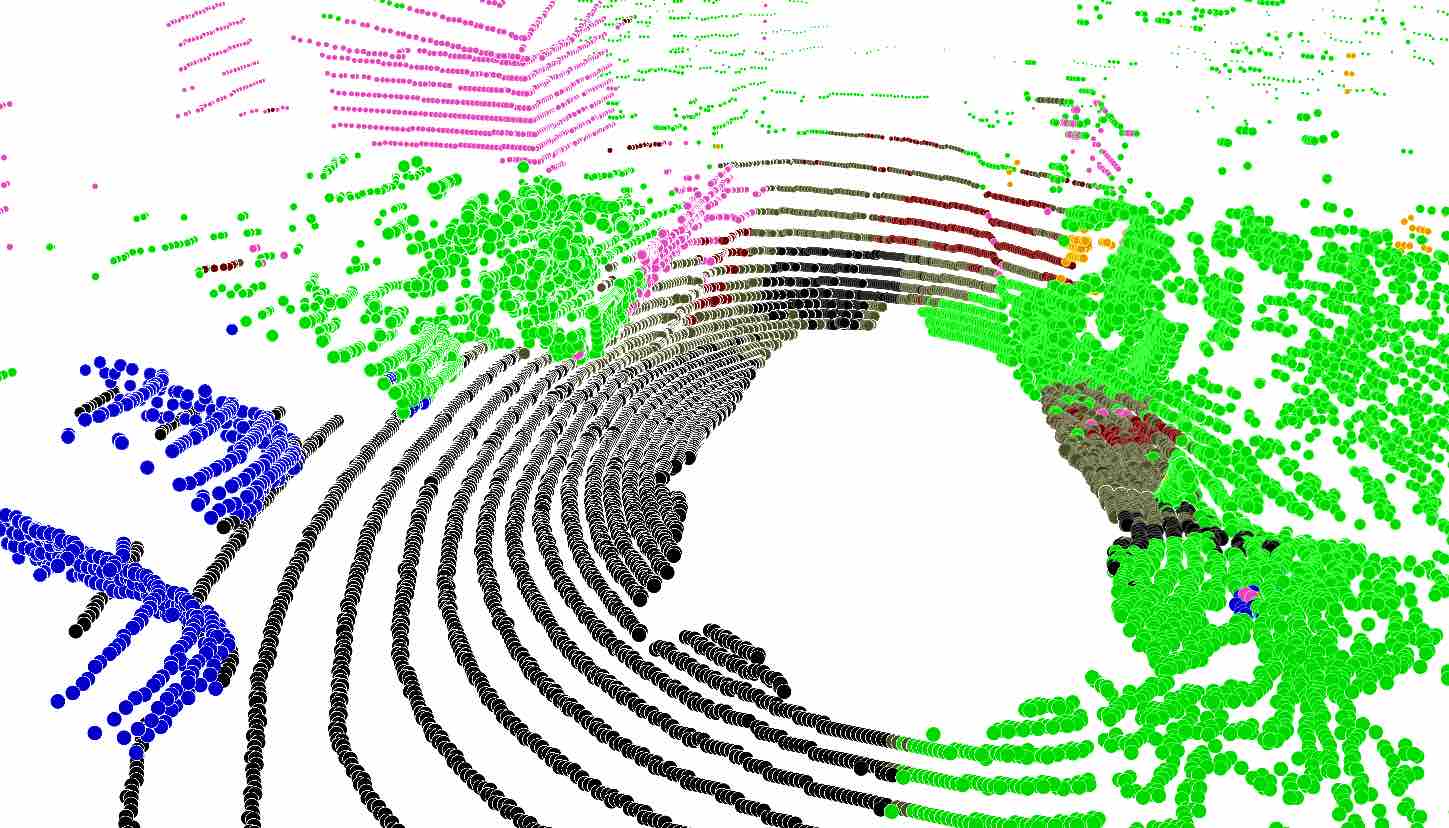}
        \end{overpic}& 
        \begin{overpic}[width=0.21\textwidth]{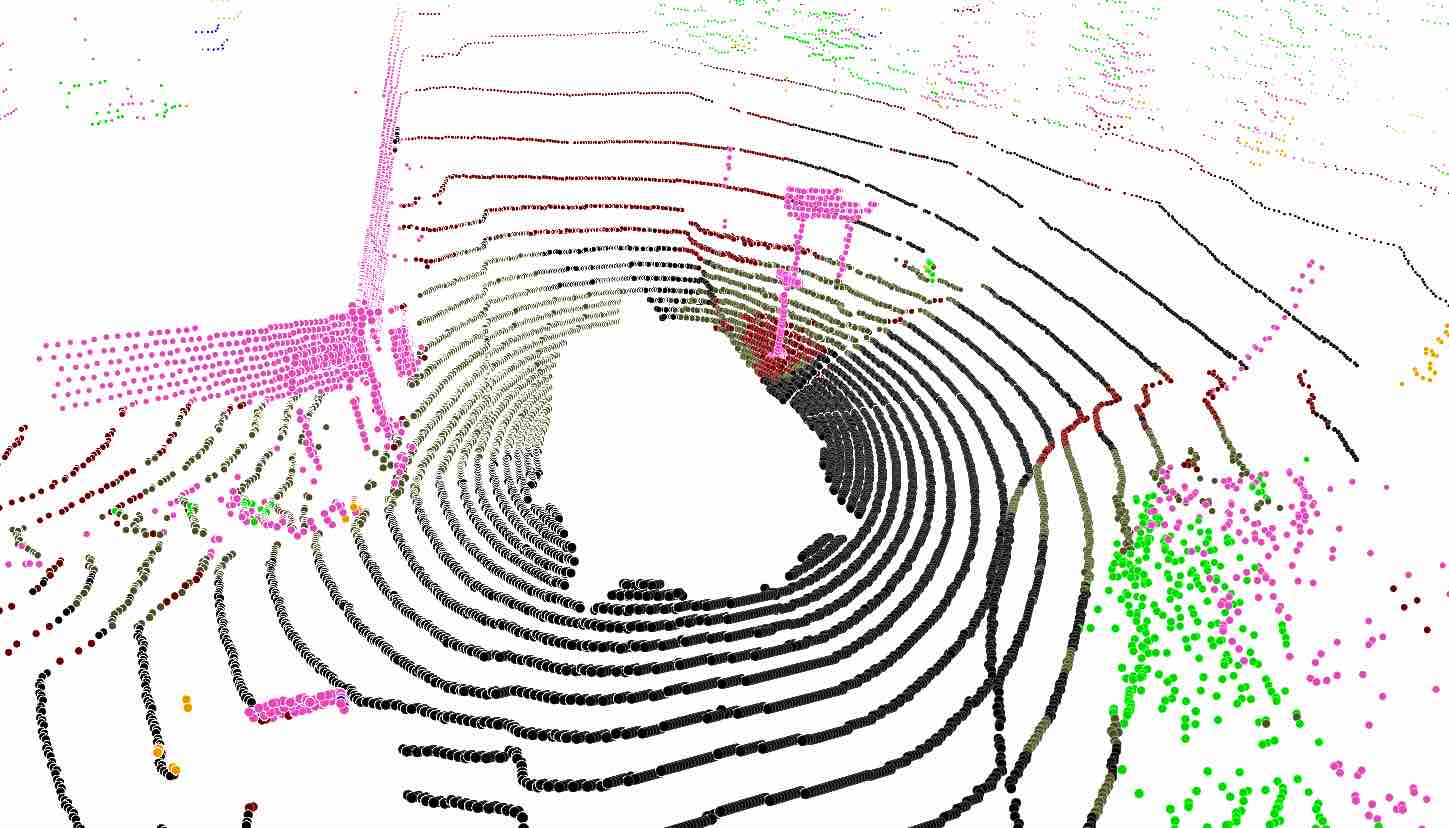}
        \end{overpic}\\
        \begin{overpic}[width=0.21\textwidth]{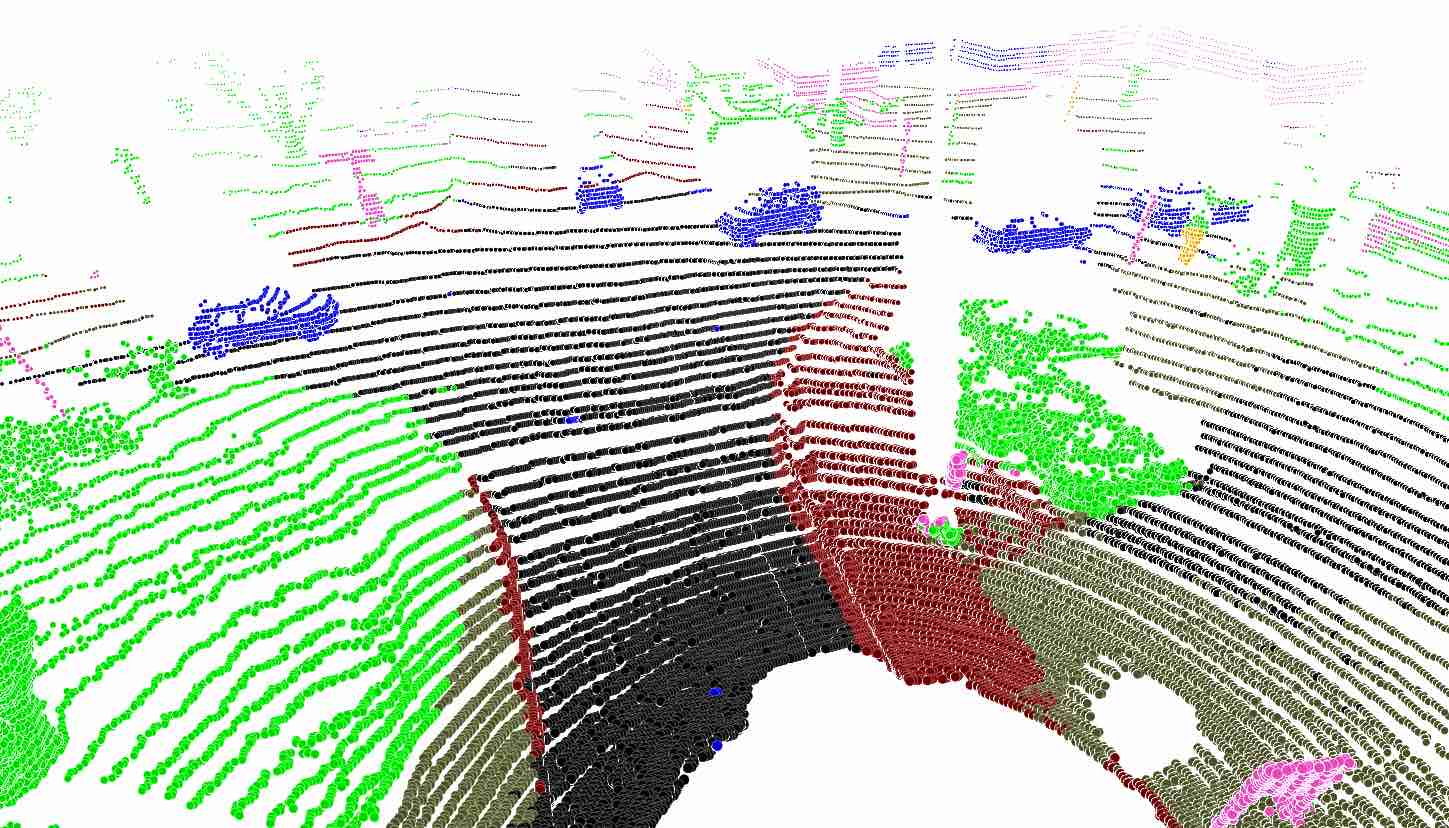}
        \put(-4.5,500){\rotatebox{90}{\color{black}\footnotesize \textbf{source}}}
        \put(-6,440){\rotatebox{90}{\color{black}\footnotesize \textbf{mix3D}}}
        \put(-6,375){\rotatebox{90}{\color{black}\footnotesize \textbf{p.cutmix}}}
        \put(-6,317){\rotatebox{90}{\color{black}\footnotesize \textbf{cosmix}}}
        \put(-6,260){\rotatebox{90}{\color{black}\footnotesize \textbf{ibn}}}
        \put(-6,200){\rotatebox{90}{\color{black}\footnotesize \textbf{robust.}}}
        \put(-5,140){\rotatebox{90}{\color{black}\footnotesize \textbf{sn}}}
        \put(-6,75){\rotatebox{90}{\color{black}\footnotesize \textbf{raycast}}}
        \put(-5,17){\rotatebox{90}{\color{black}\footnotesize \textbf{ours}}}
        \put(-6,-35){\rotatebox{90}{\color{black}\footnotesize \textbf{gt}}}
        \put(80,541){\color{black}\footnotesize \textbf{SemanticKITTI}}
        \put(290,541){\color{black}\footnotesize \textbf{nuScenes}}
        \end{overpic} &  
        \begin{overpic}[width=0.21\textwidth]{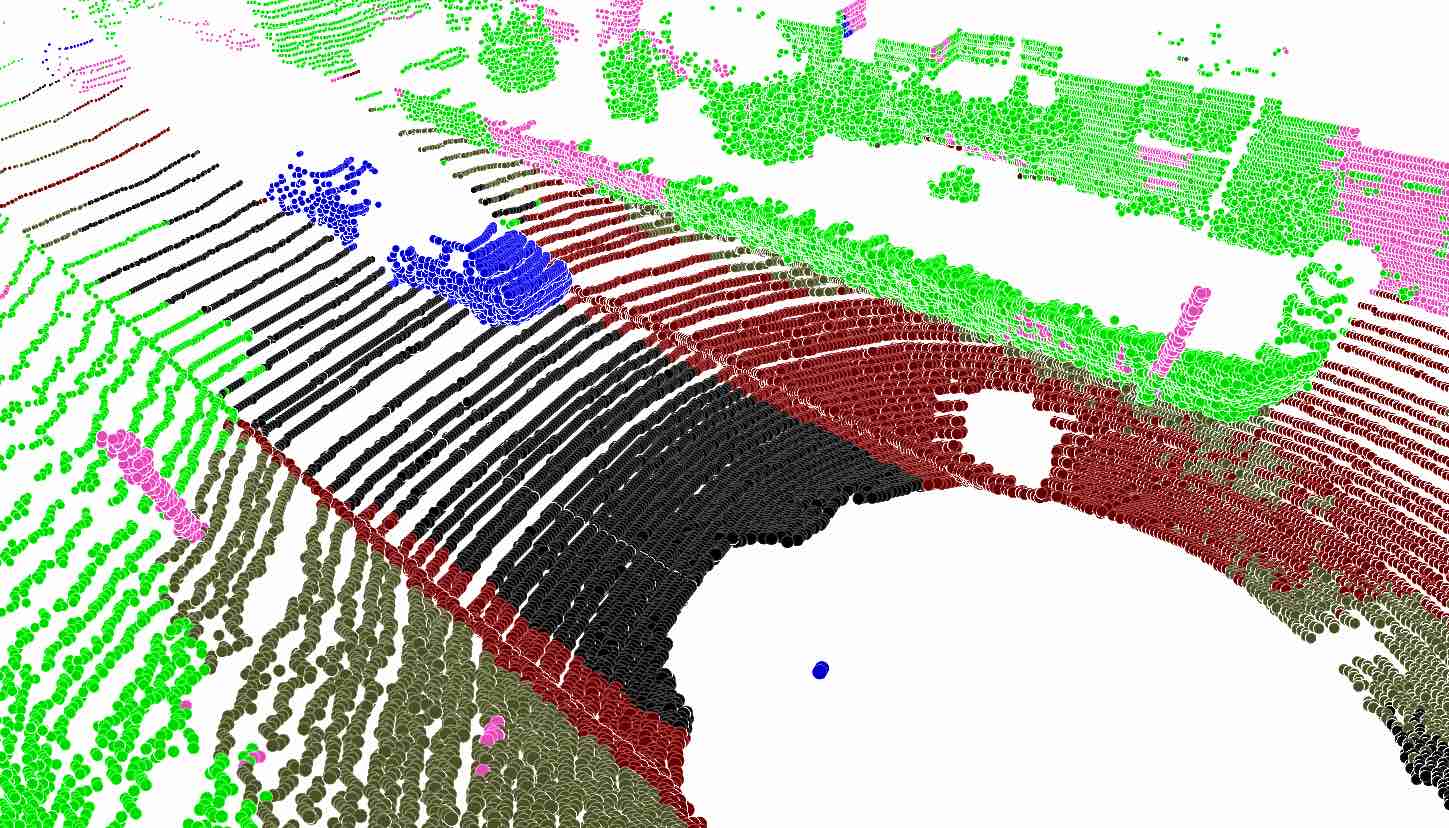}
        \end{overpic} &
        \begin{overpic}[width=0.21\textwidth]{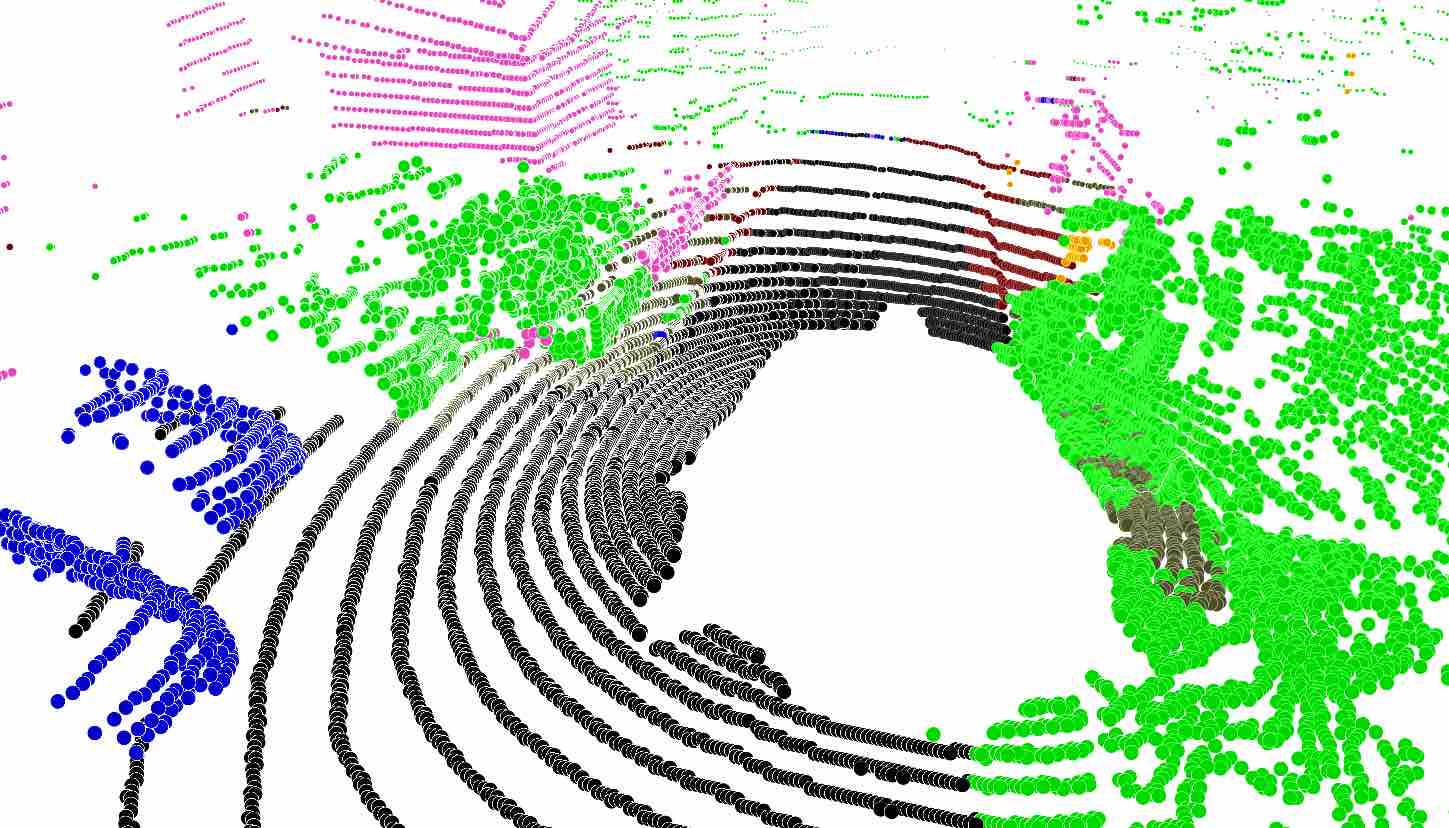}
        \end{overpic}& 
        \begin{overpic}[width=0.21\textwidth]{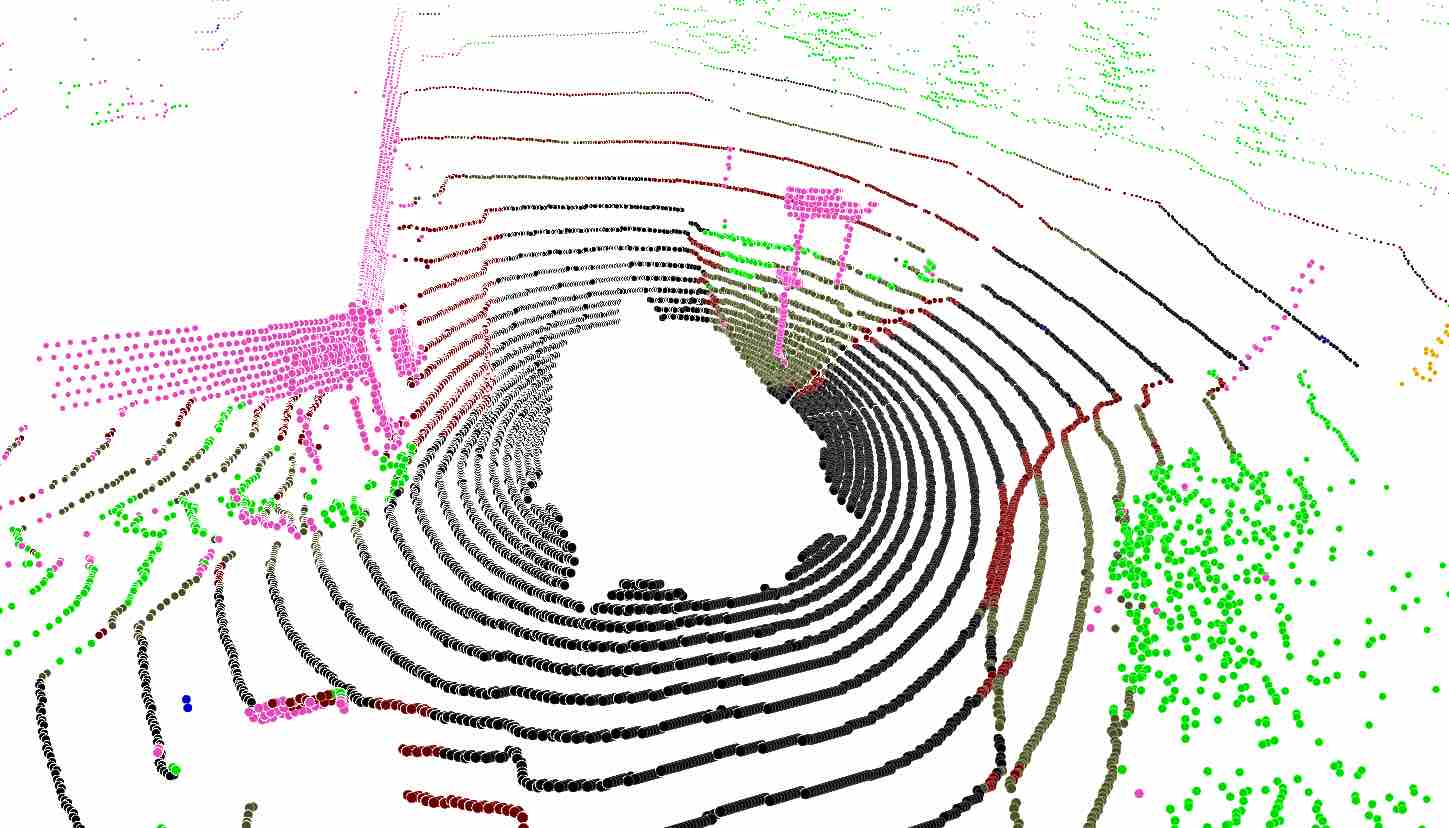}
        \end{overpic}\\
        \begin{overpic}[width=0.21\textwidth]{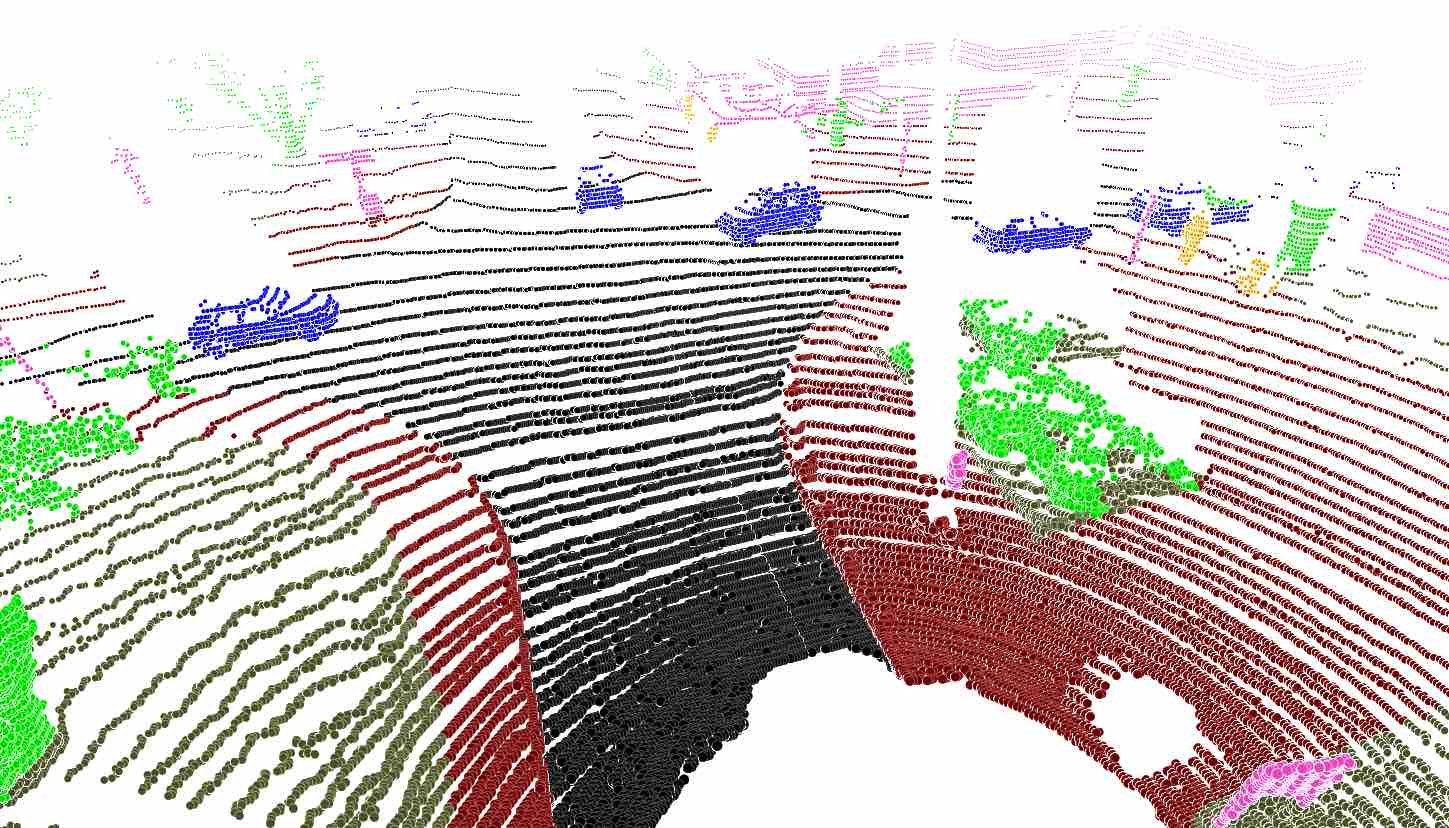}
        \end{overpic} &  
        \begin{overpic}[width=0.21\textwidth]{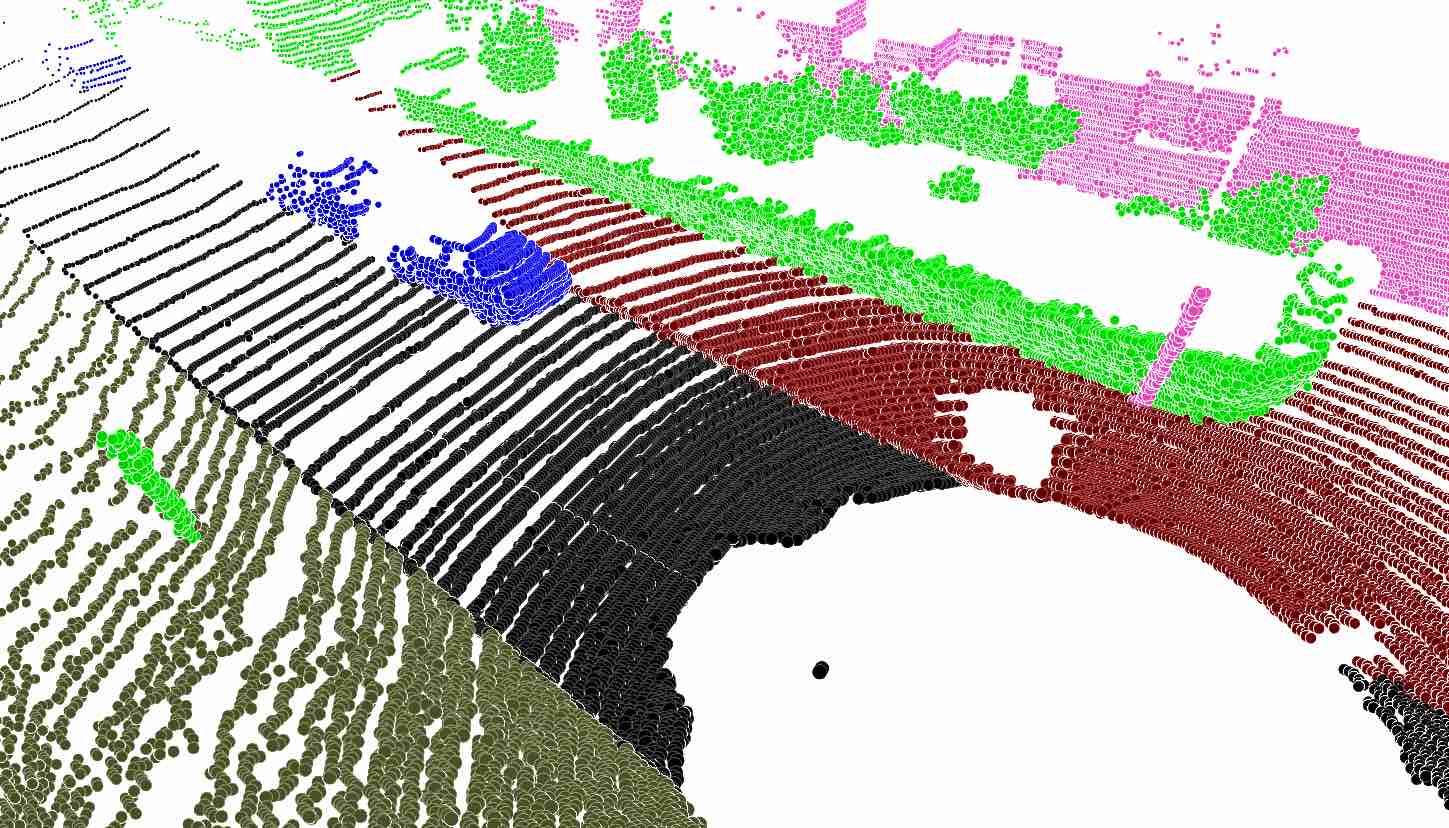}
        \end{overpic} &
        \begin{overpic}[width=0.21\textwidth]{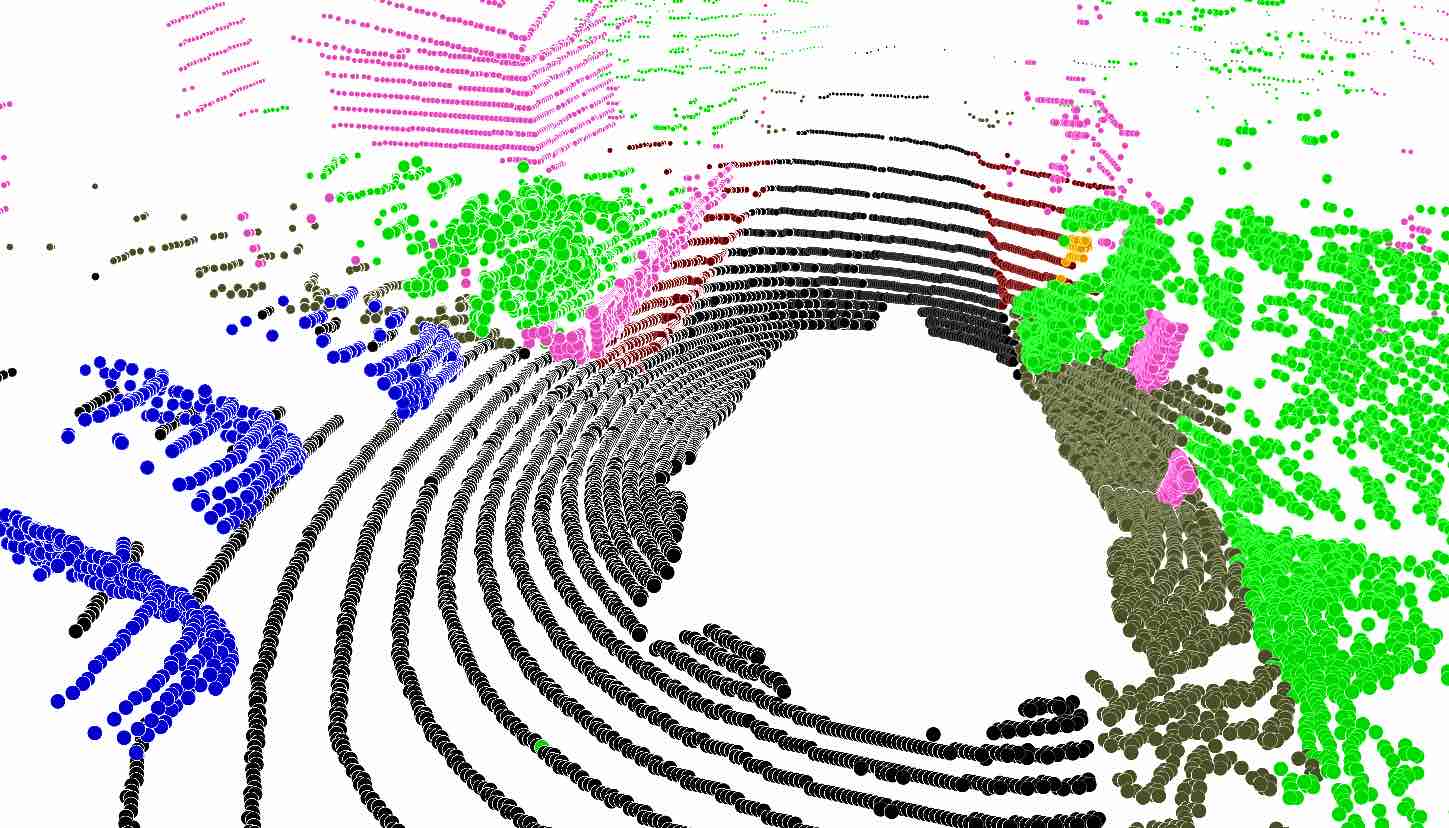}
        \end{overpic}& 
        \begin{overpic}[width=0.21\textwidth]{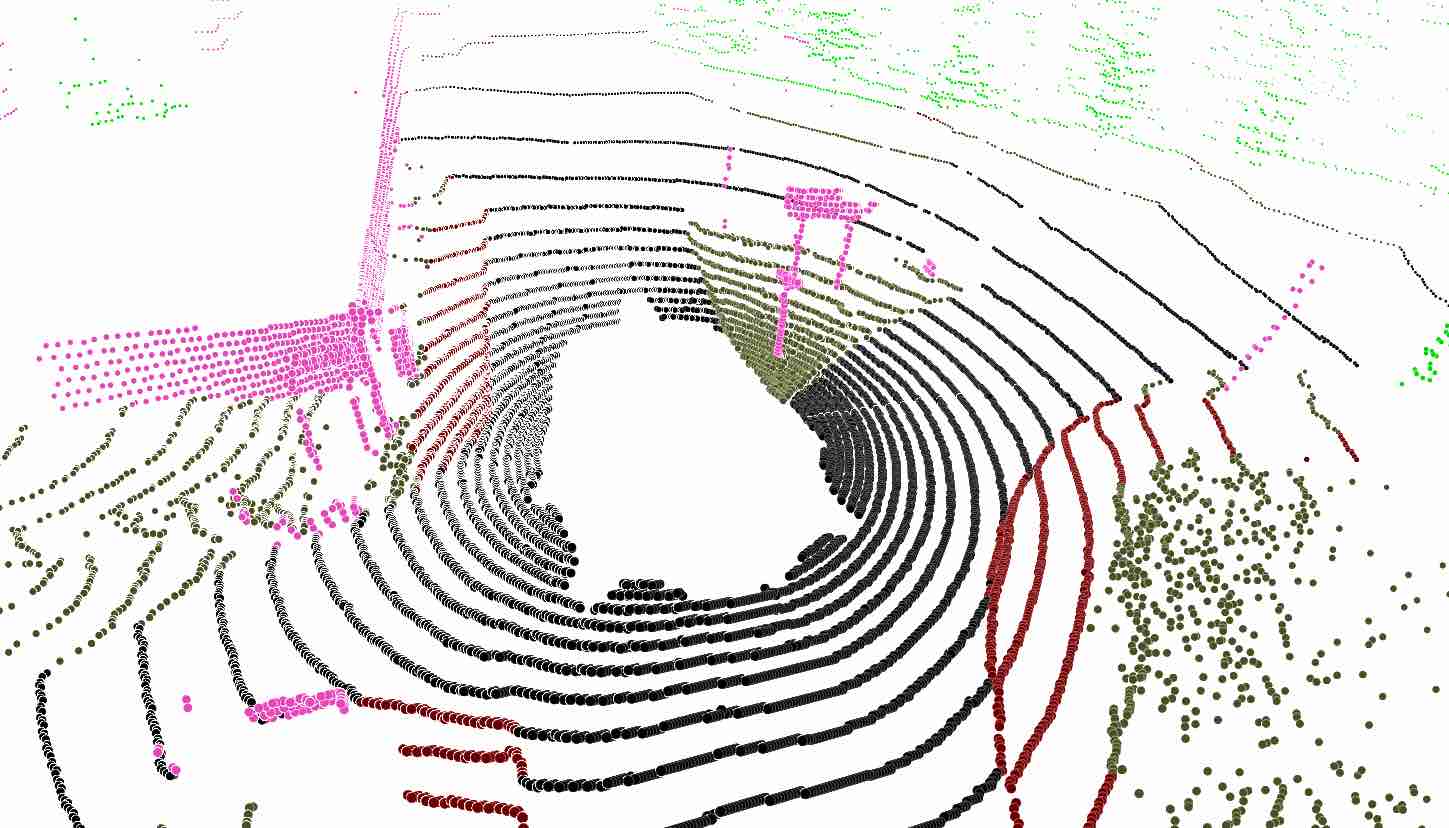}
        \end{overpic}\\
    \end{tabular}
    \vspace{-4mm}
    \caption{\textbf{Qualitative results.} \textit{Top:} Synth4D-kitti$+$Synth4D-nuScenes$\to$SemanticKITTI, \textit{bottom:} Synth4D-kitti$+$Synth4D-nuScenes$\to$nuScenes. \lidog improves over source and baselines, \eg, we observe the improvements of \textit{vegetation} in SemanticKITTI and \textit{road} in nuScenes.}
    \label{fig:supp_qualitative_multi}
\end{figure*}

\begin{figure*}[t]
\centering
    \setlength\tabcolsep{1.pt}
    \begin{tabular}{cccc}
    \raggedright
        \begin{overpic}[width=0.21\textwidth]{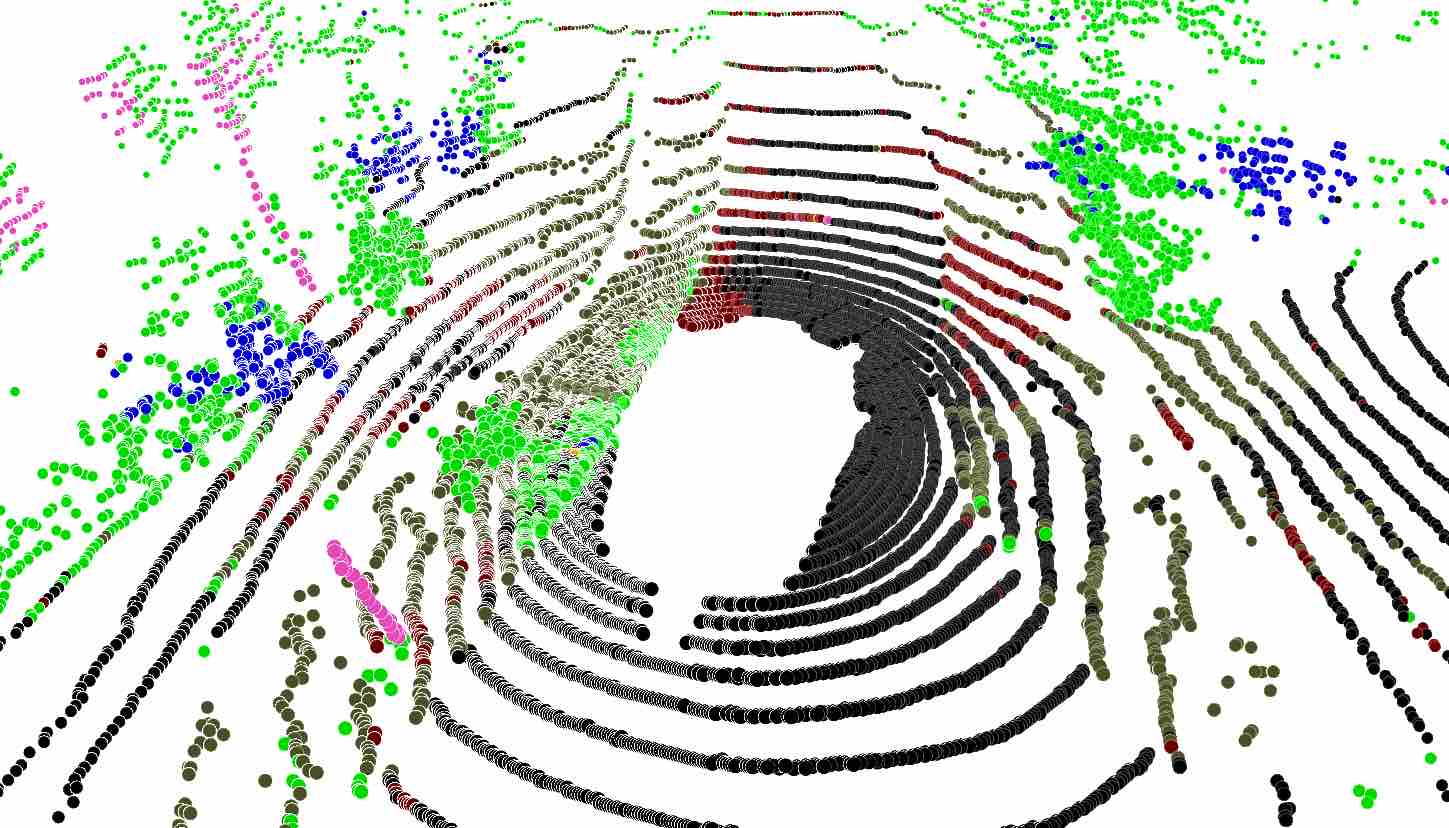}
        \end{overpic} &  
        \begin{overpic}[width=0.21\textwidth]{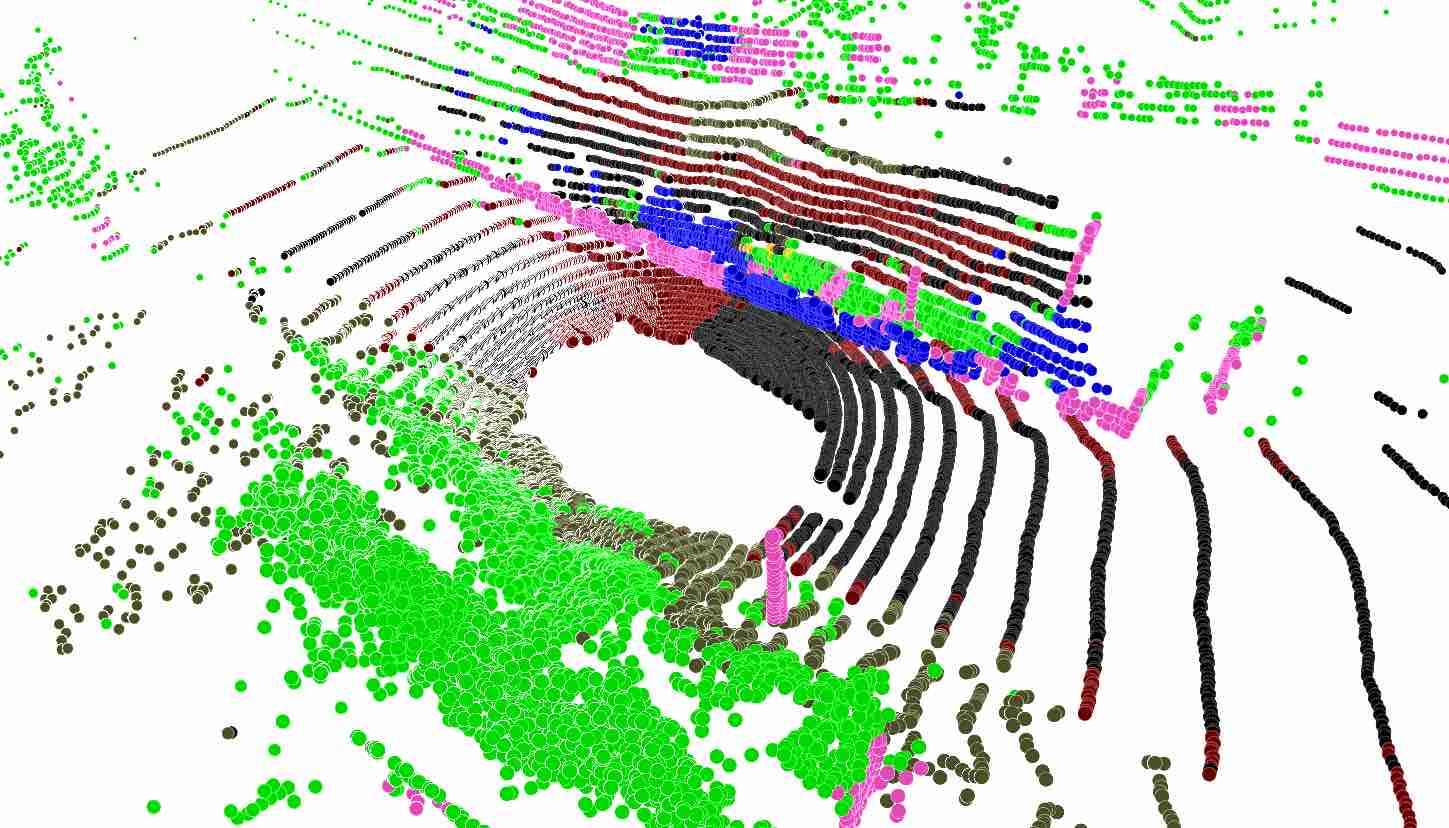}
        \end{overpic} &
        \begin{overpic}[width=0.21\textwidth]{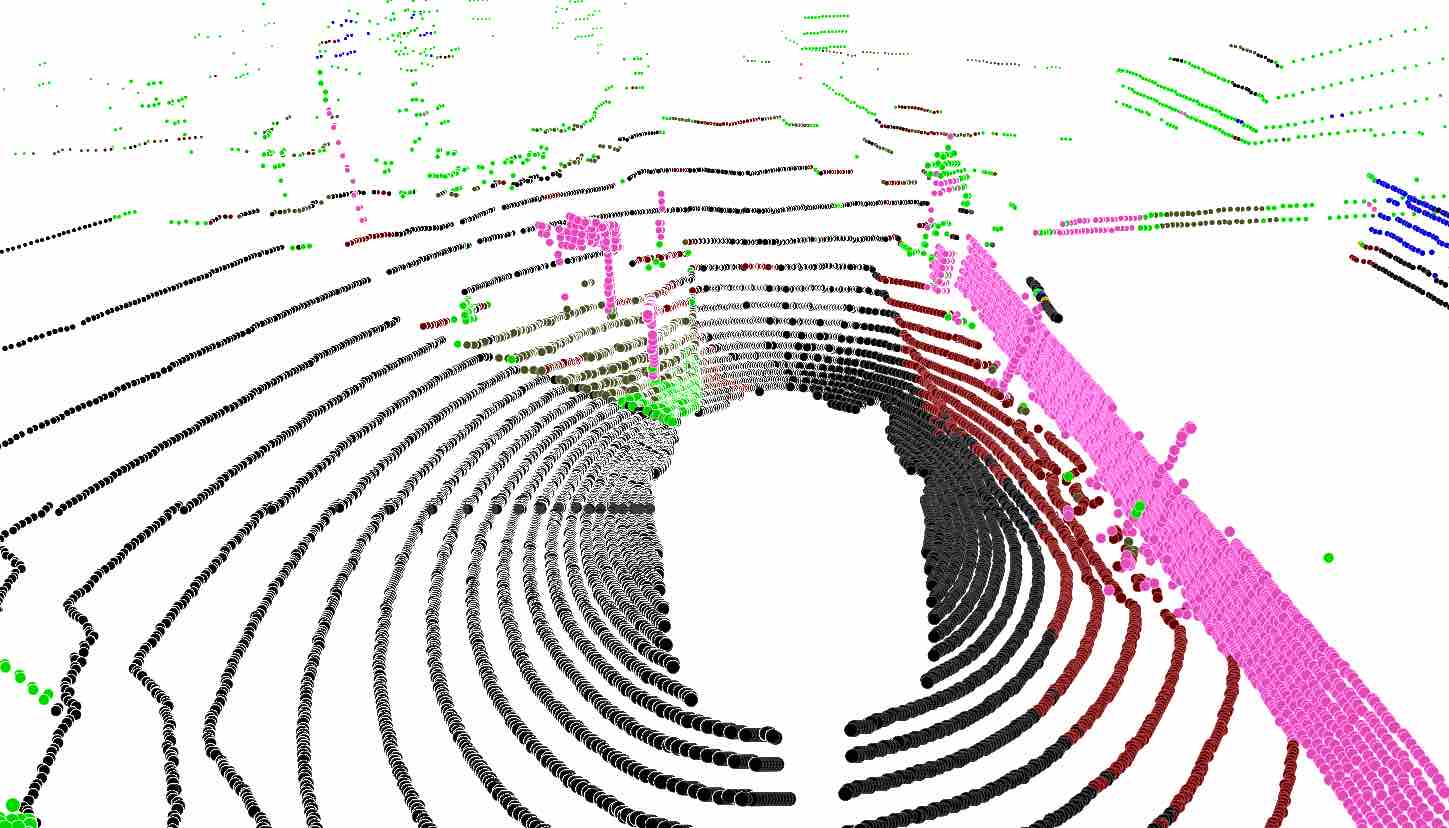}
        \end{overpic}& 
        \begin{overpic}[width=0.21\textwidth]{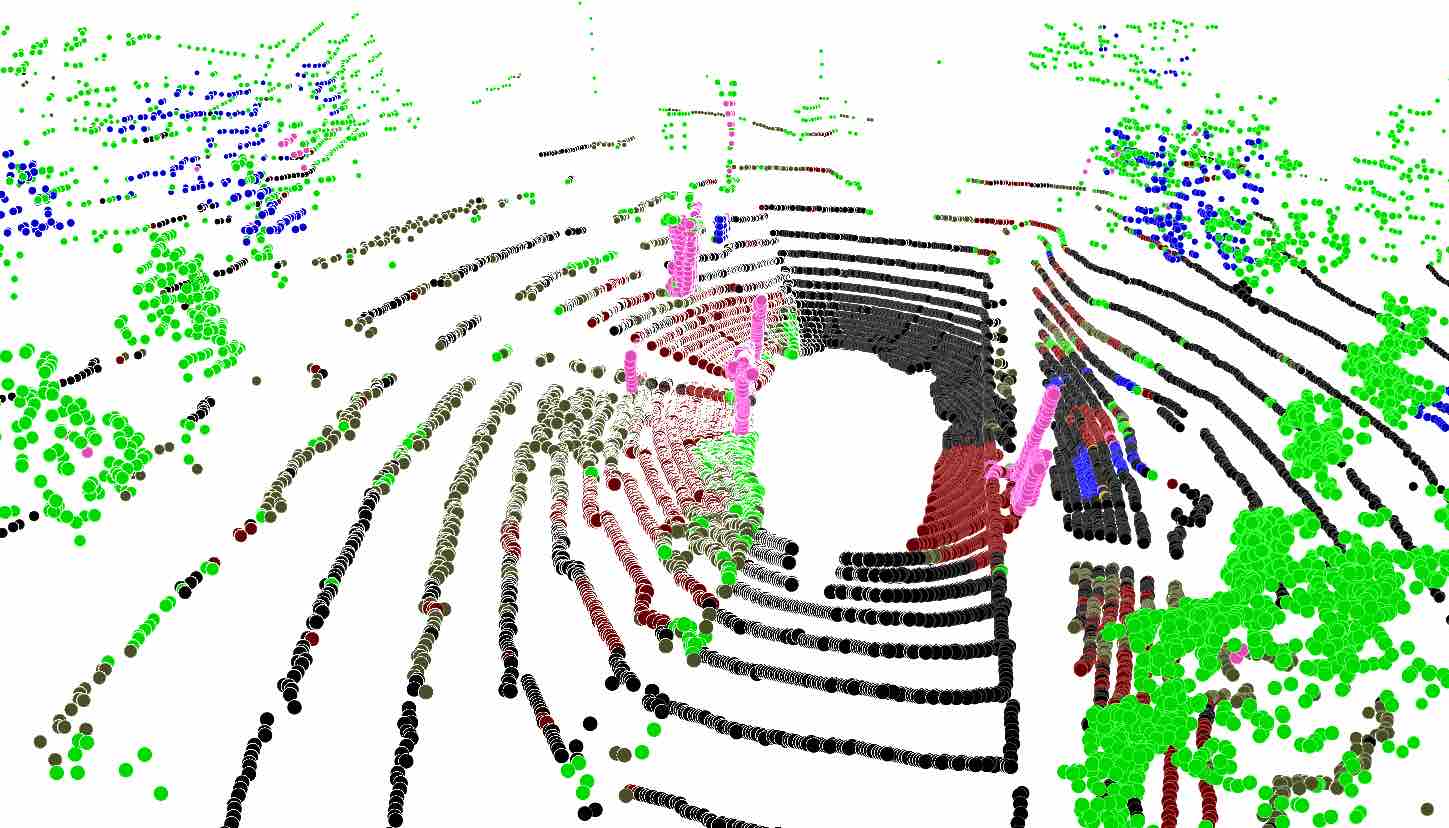}
        \end{overpic}\\
        \begin{overpic}[width=0.21\textwidth]{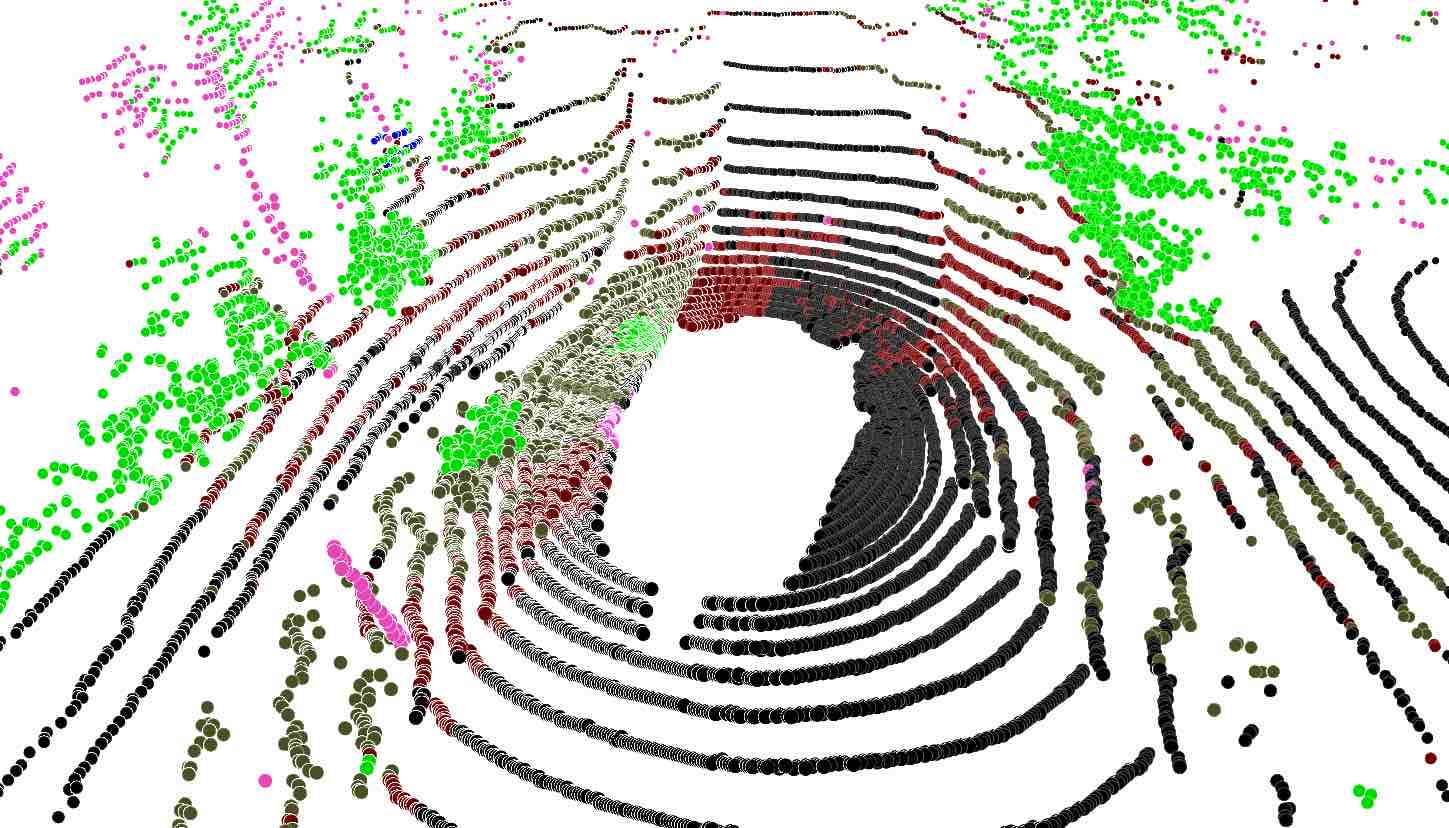}
        \end{overpic} &  
        \begin{overpic}[width=0.21\textwidth]{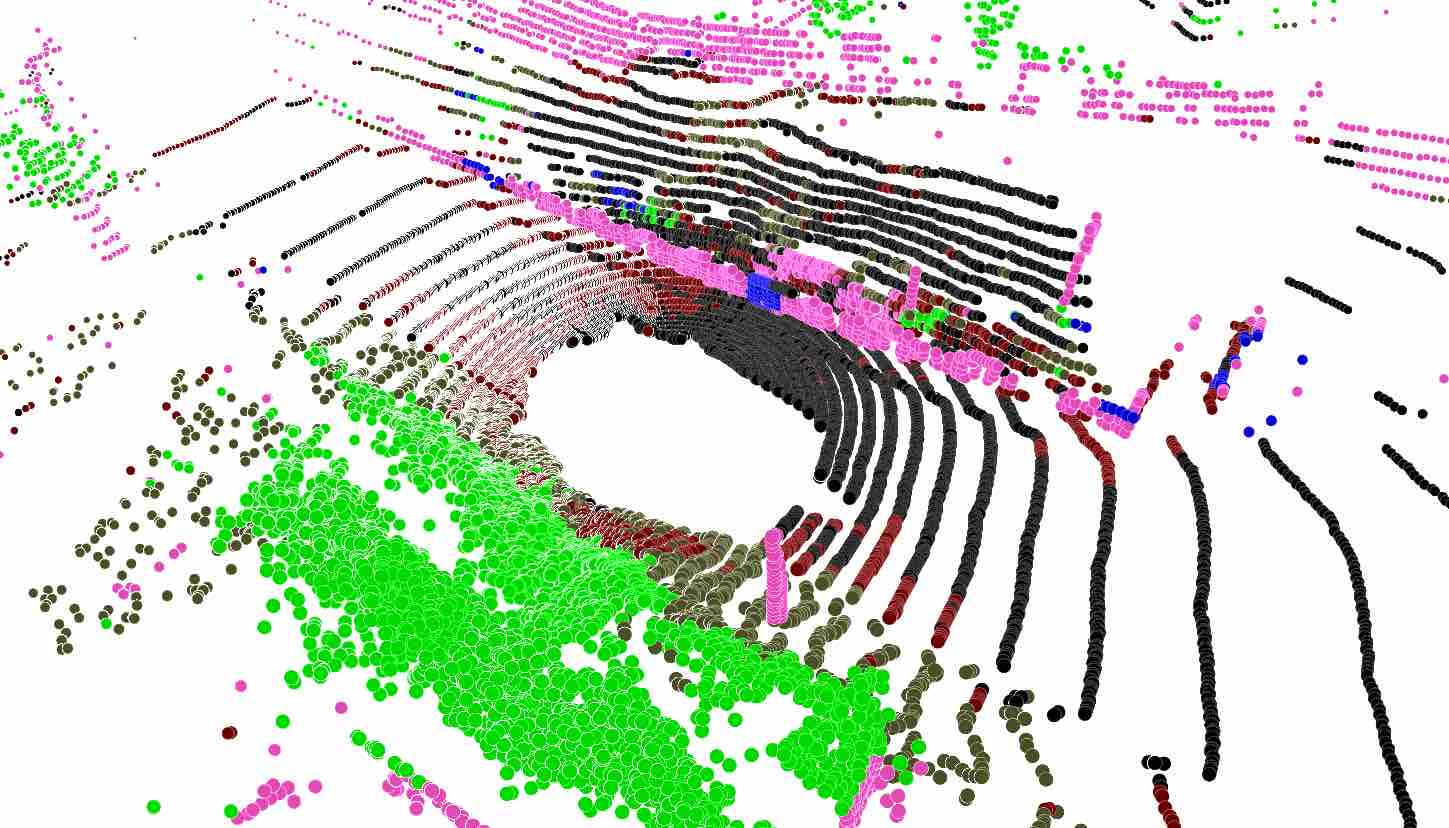}
        \end{overpic} &
        \begin{overpic}[width=0.21\textwidth]{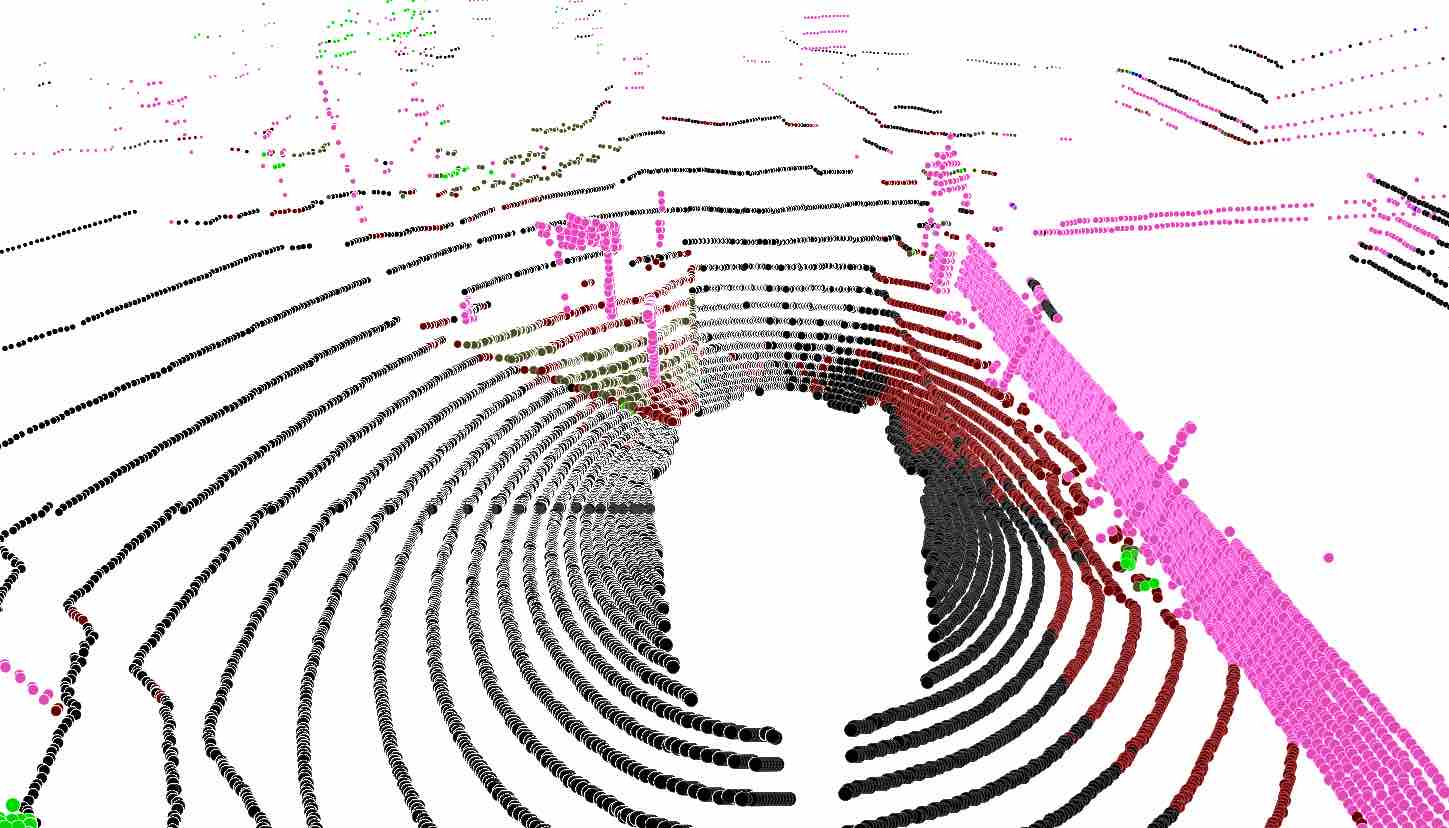}
        \end{overpic}& 
        \begin{overpic}[width=0.21\textwidth]{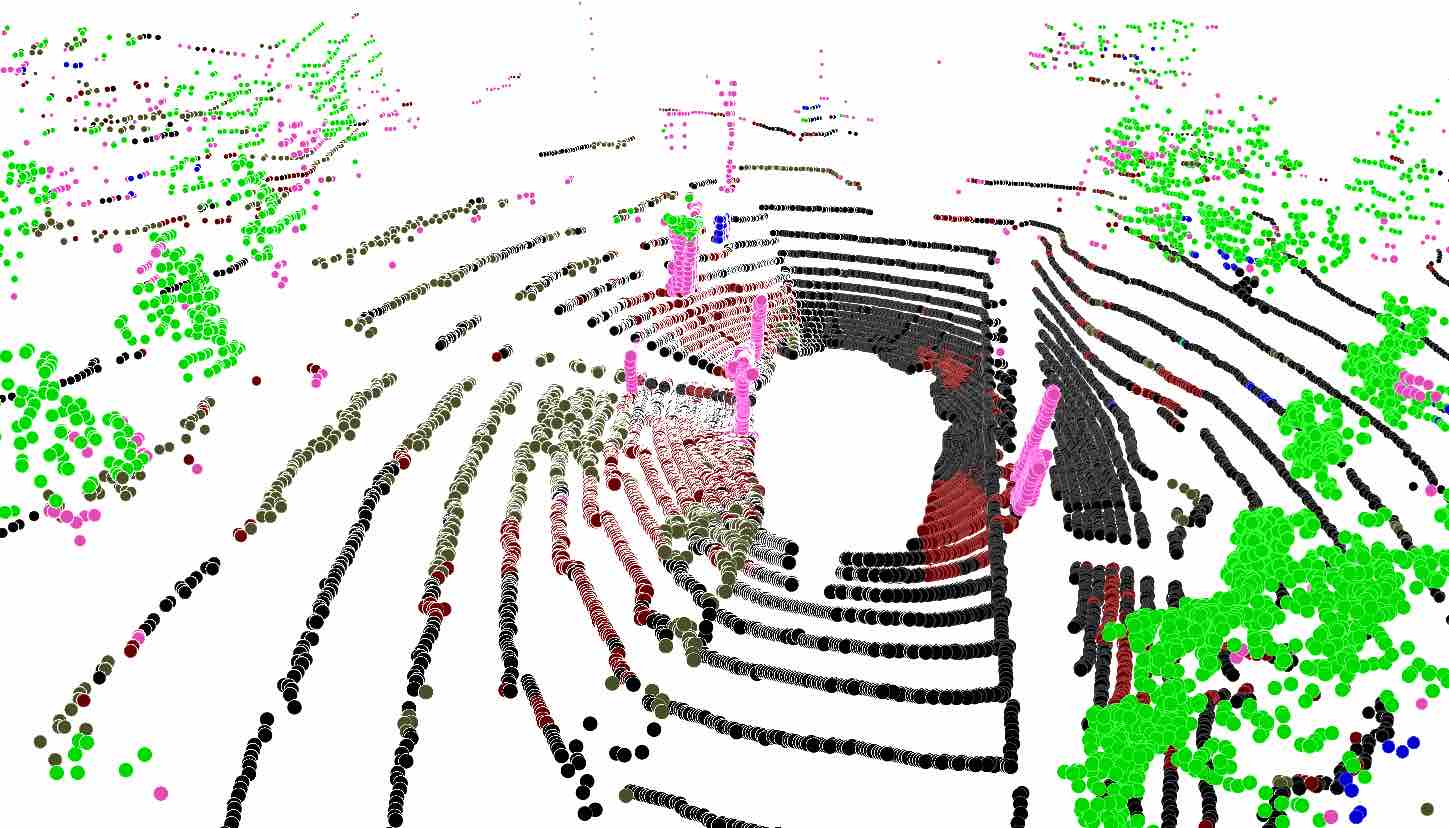}
        \end{overpic}\\
        \begin{overpic}[width=0.21\textwidth]{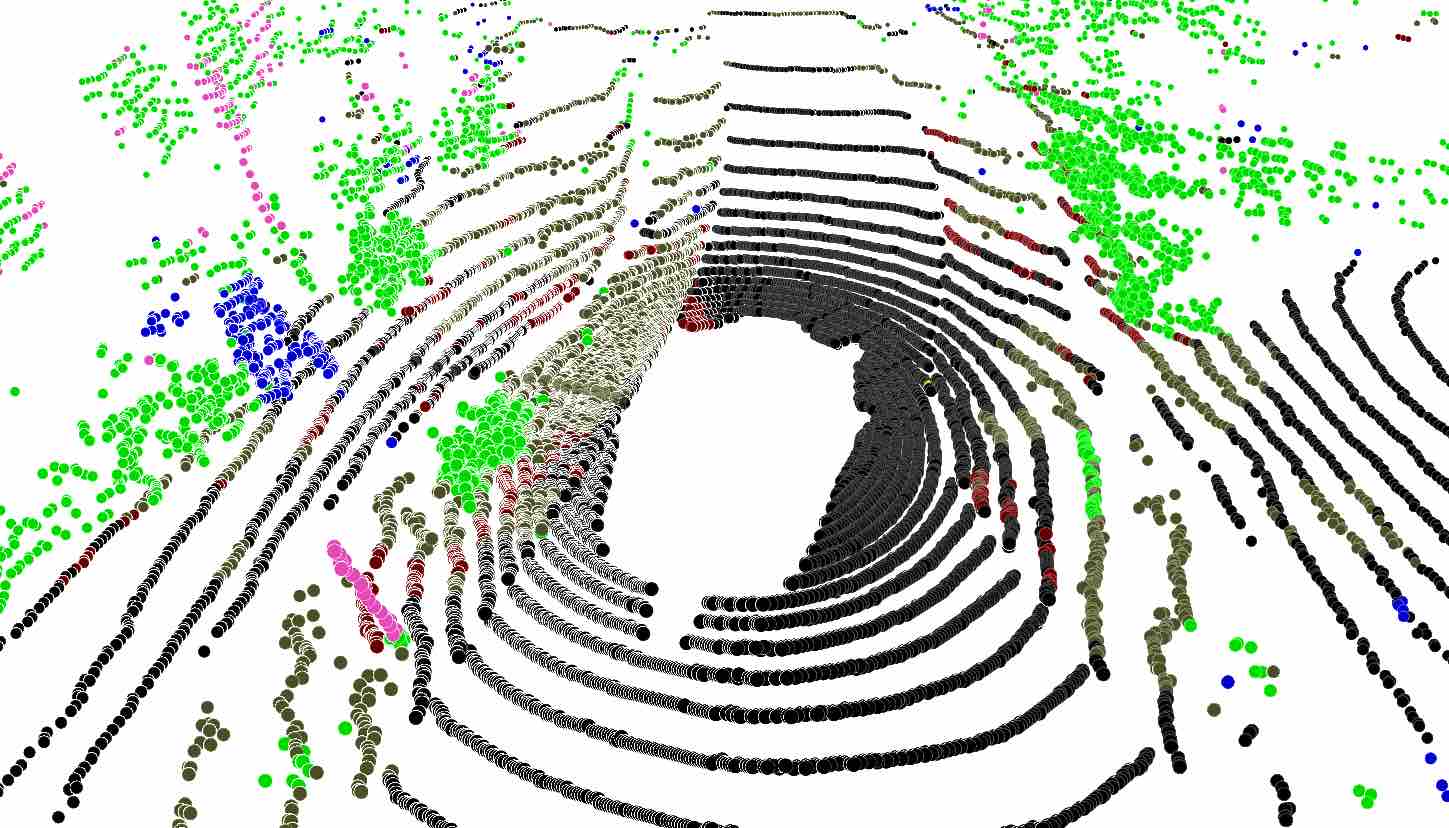}
        \end{overpic} &  
        \begin{overpic}[width=0.21\textwidth]{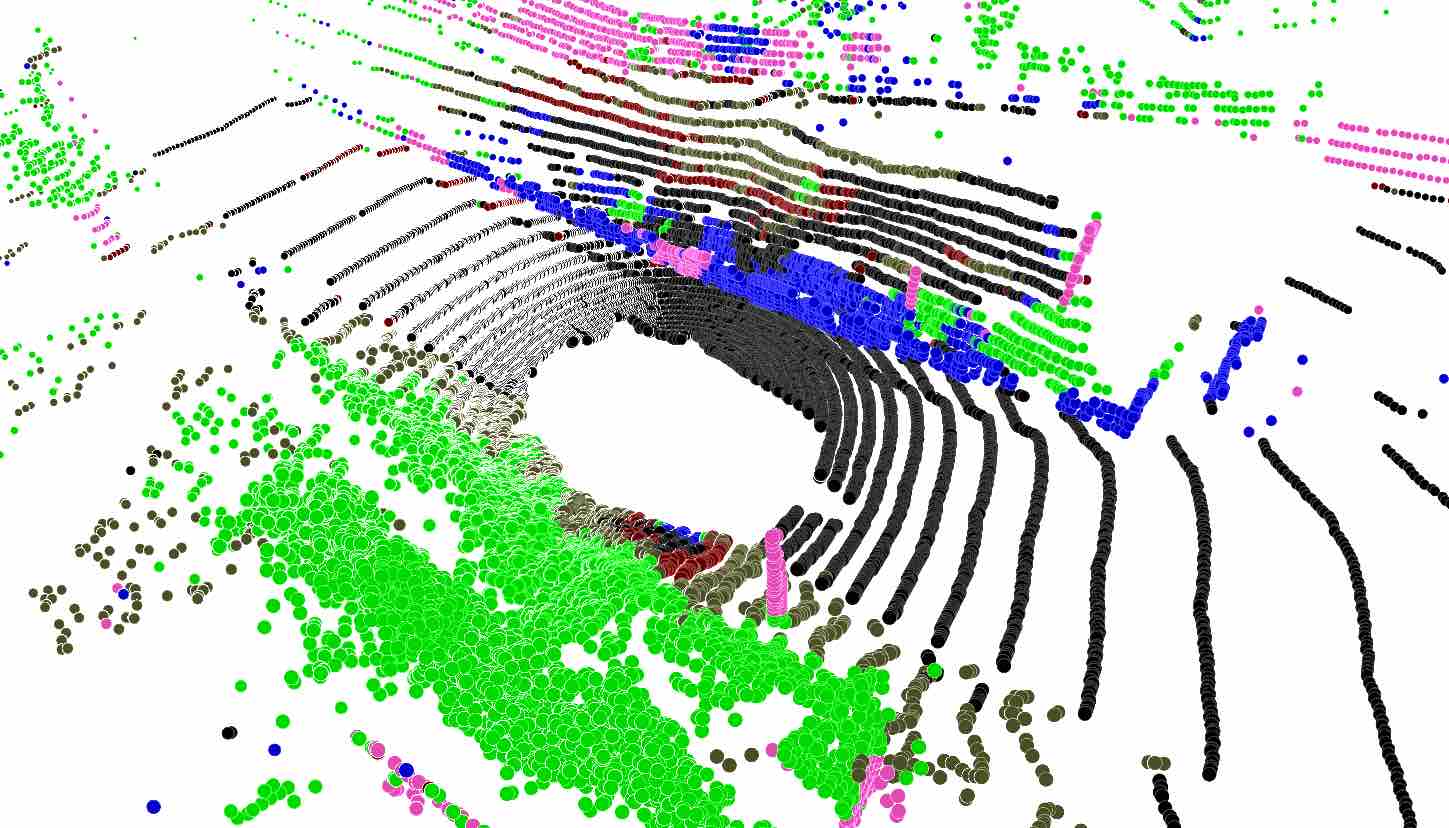}
        \end{overpic} &
        \begin{overpic}[width=0.21\textwidth]{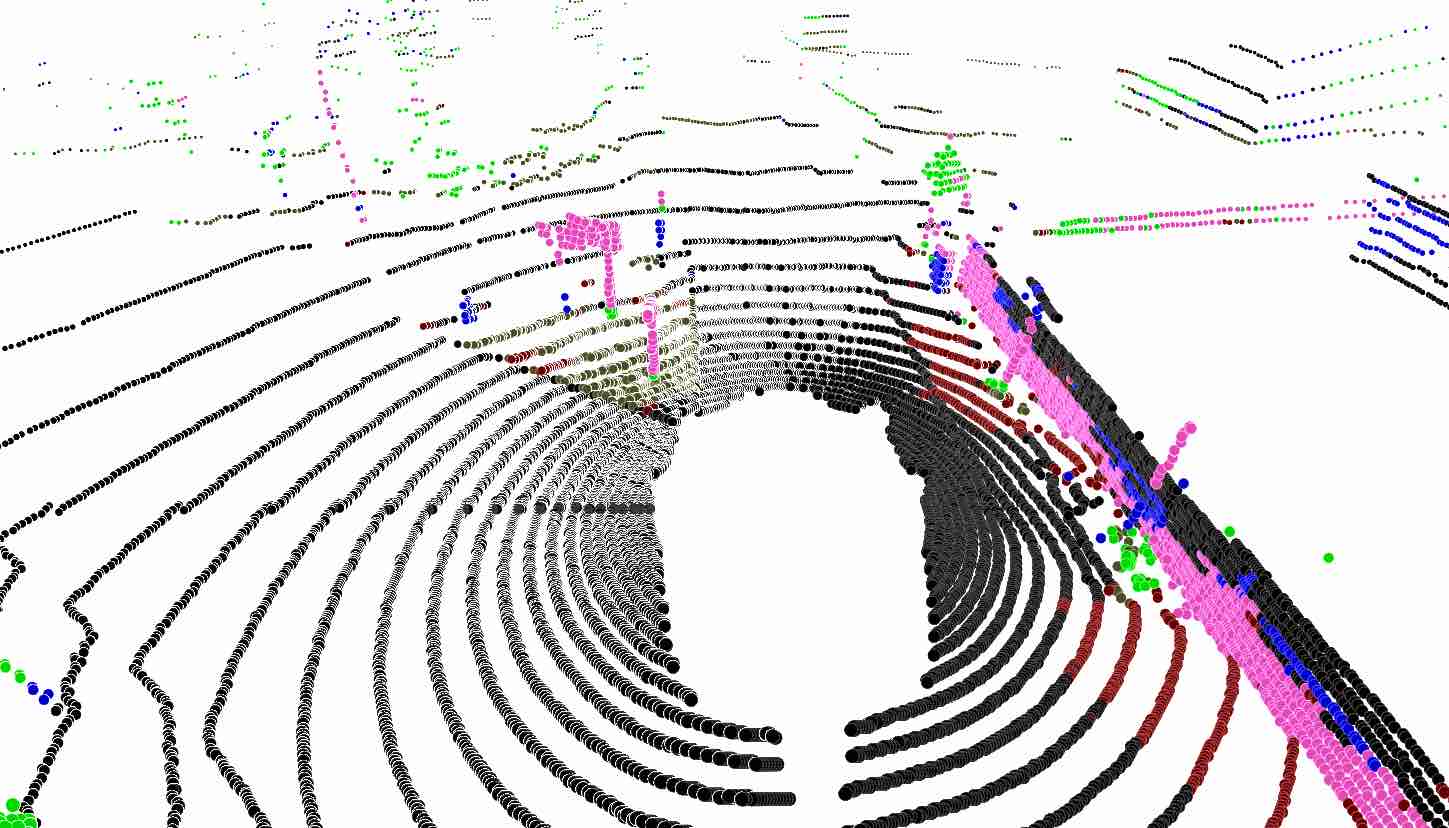}
        \end{overpic}& 
        \begin{overpic}[width=0.21\textwidth]{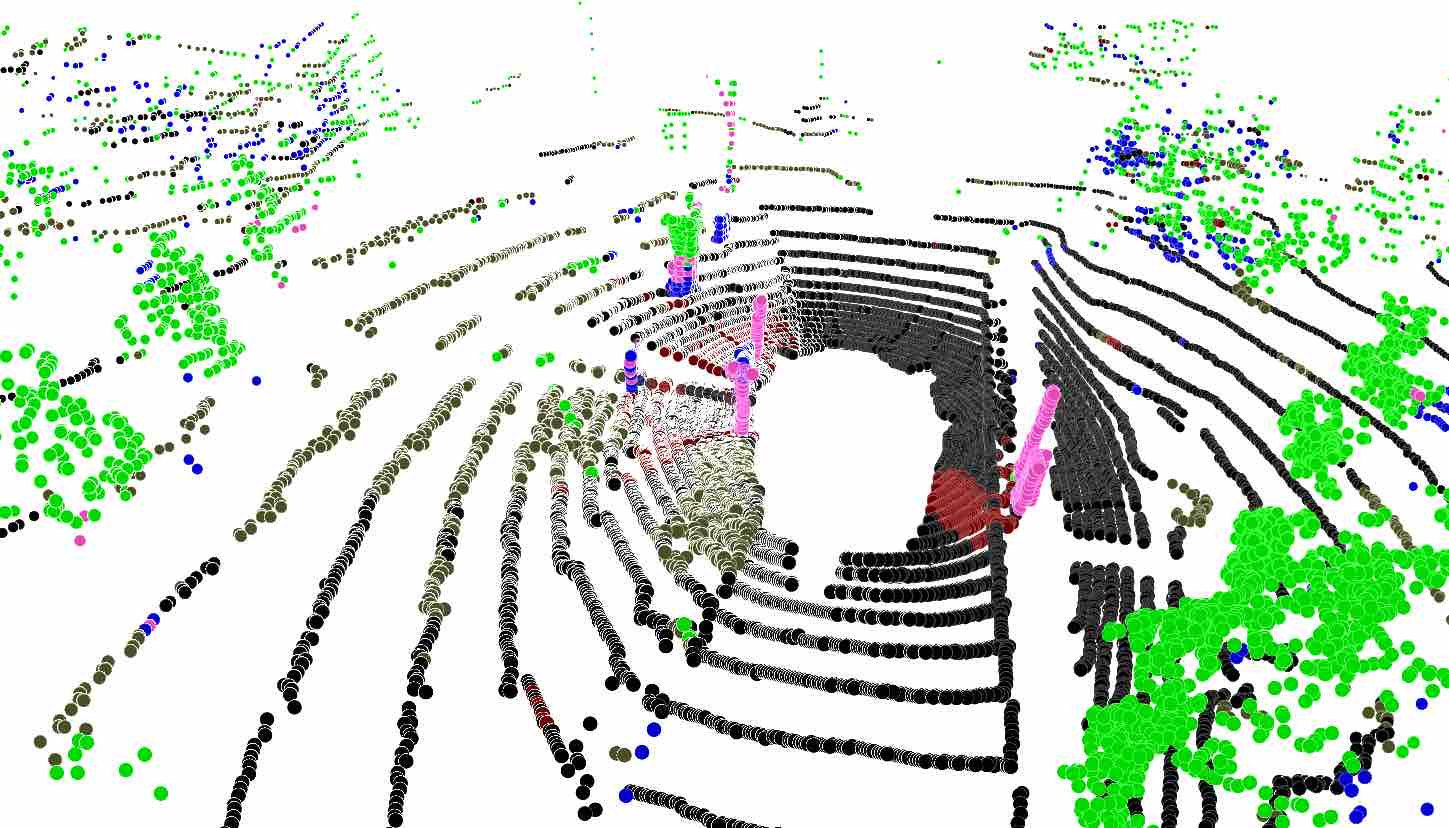}
        \end{overpic}\\
        \begin{overpic}[width=0.21\textwidth]{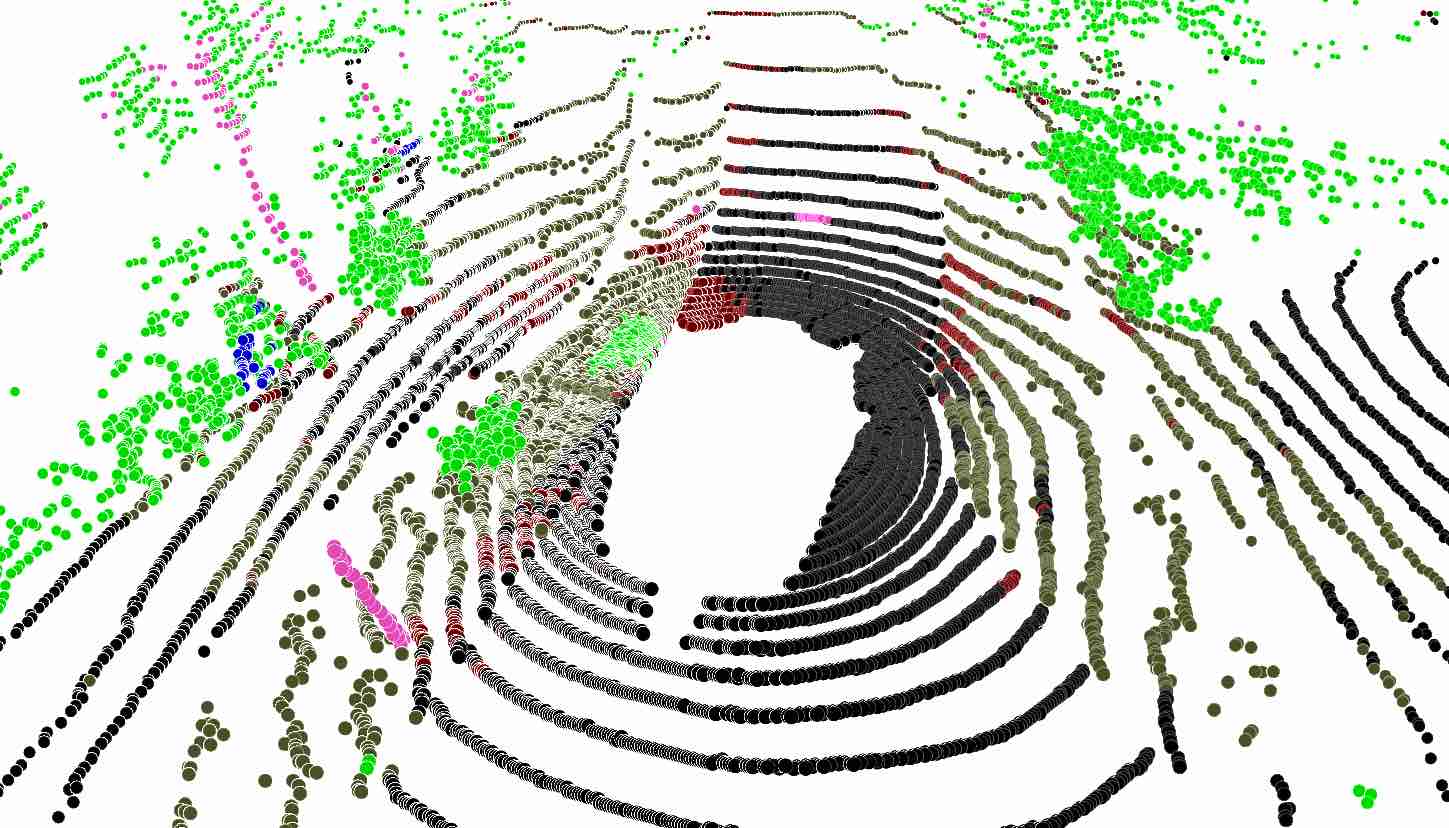}
        \end{overpic} &  
        \begin{overpic}[width=0.21\textwidth]{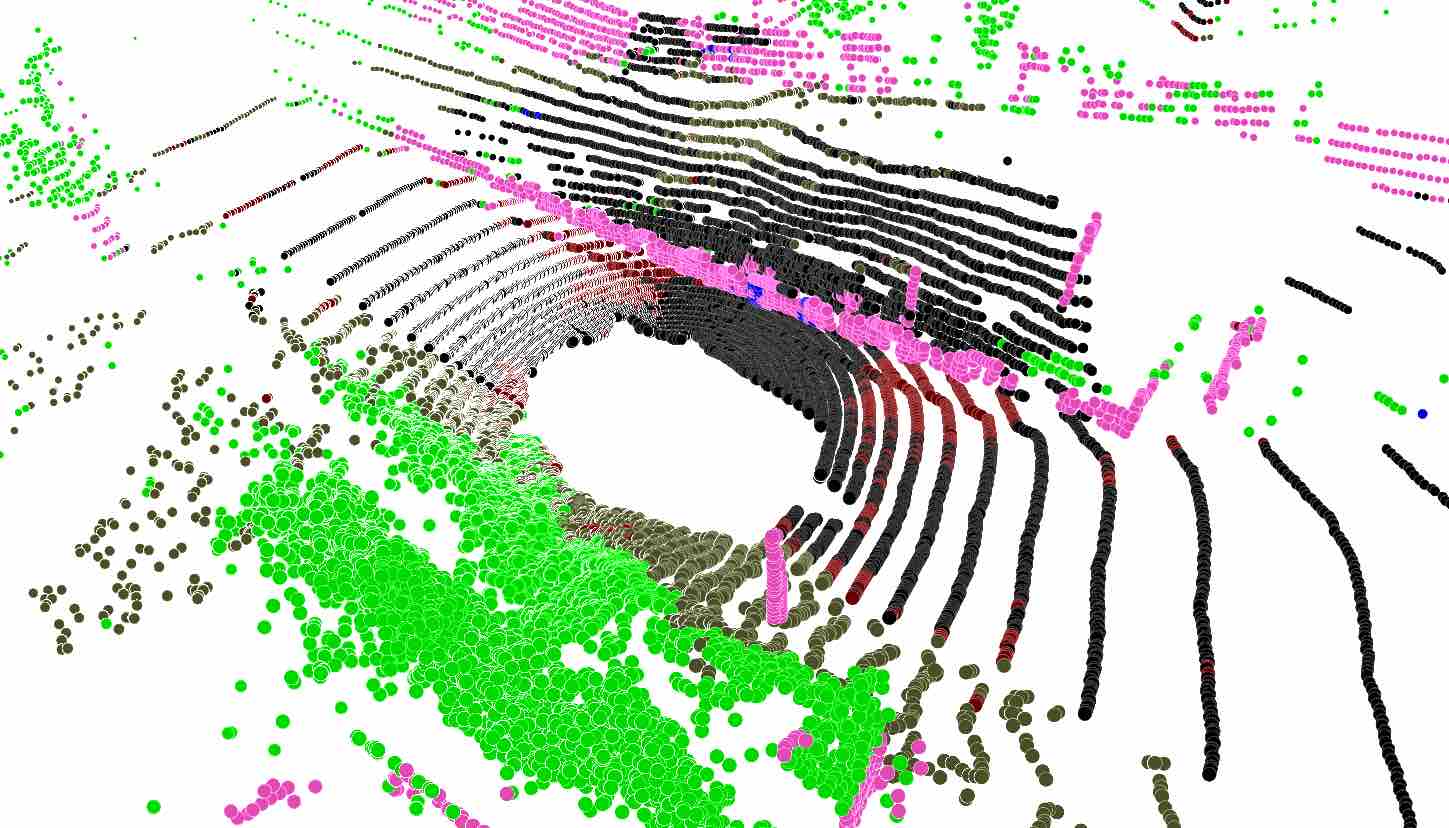}
        \end{overpic} &
        \begin{overpic}[width=0.21\textwidth]{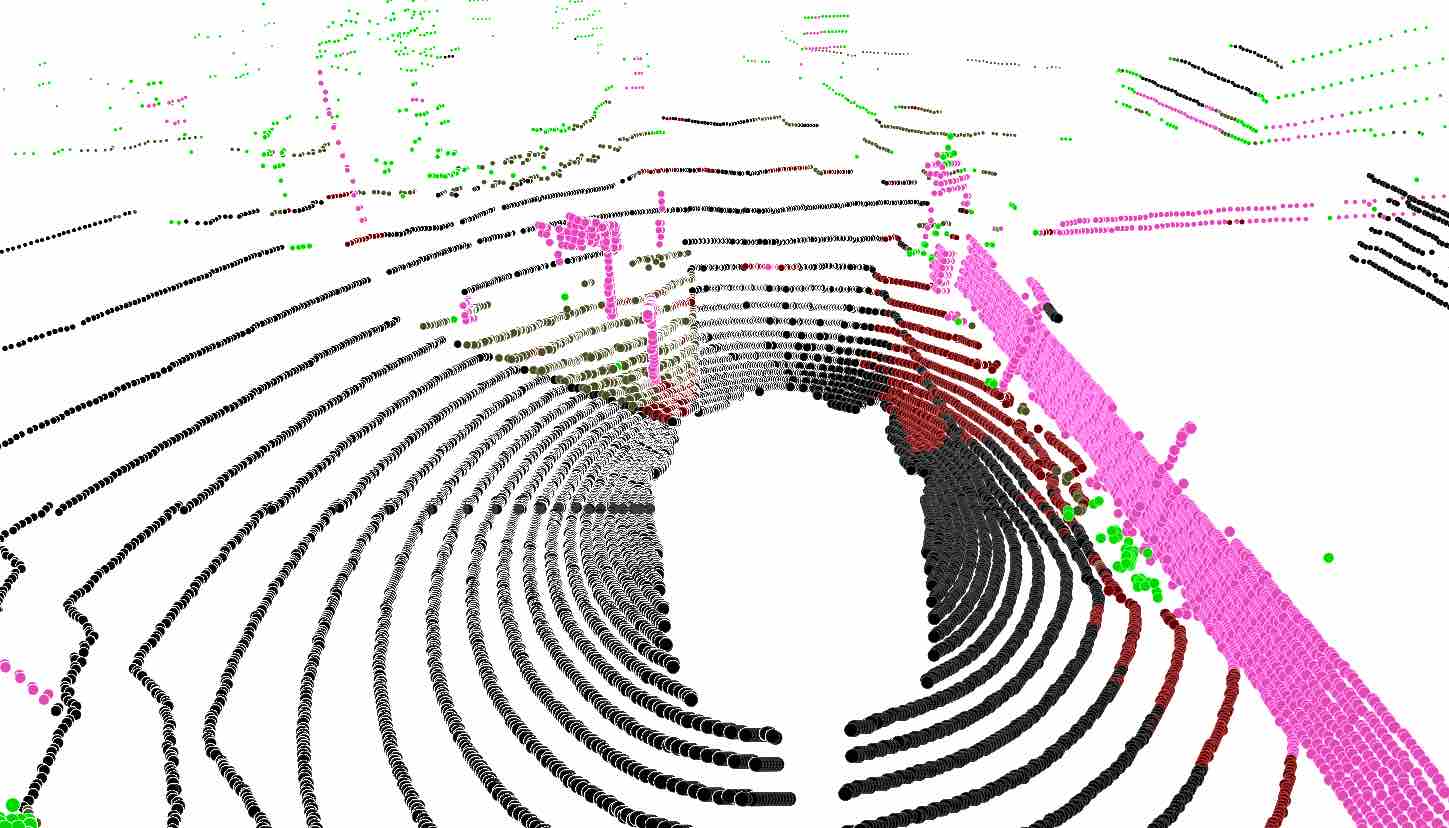}
        \end{overpic}& 
        \begin{overpic}[width=0.21\textwidth]{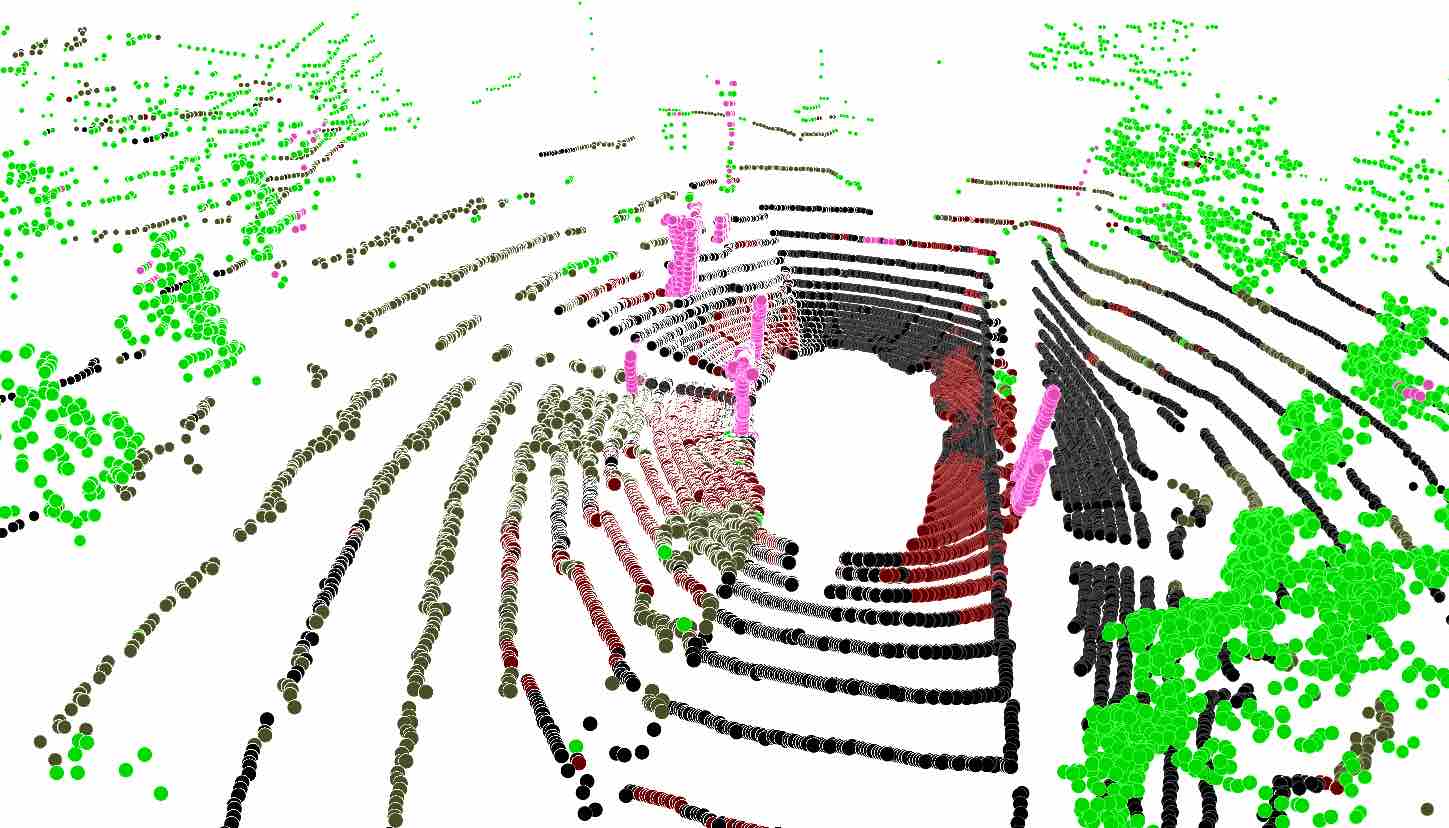}
        \end{overpic}\\
        \begin{overpic}[width=0.21\textwidth]{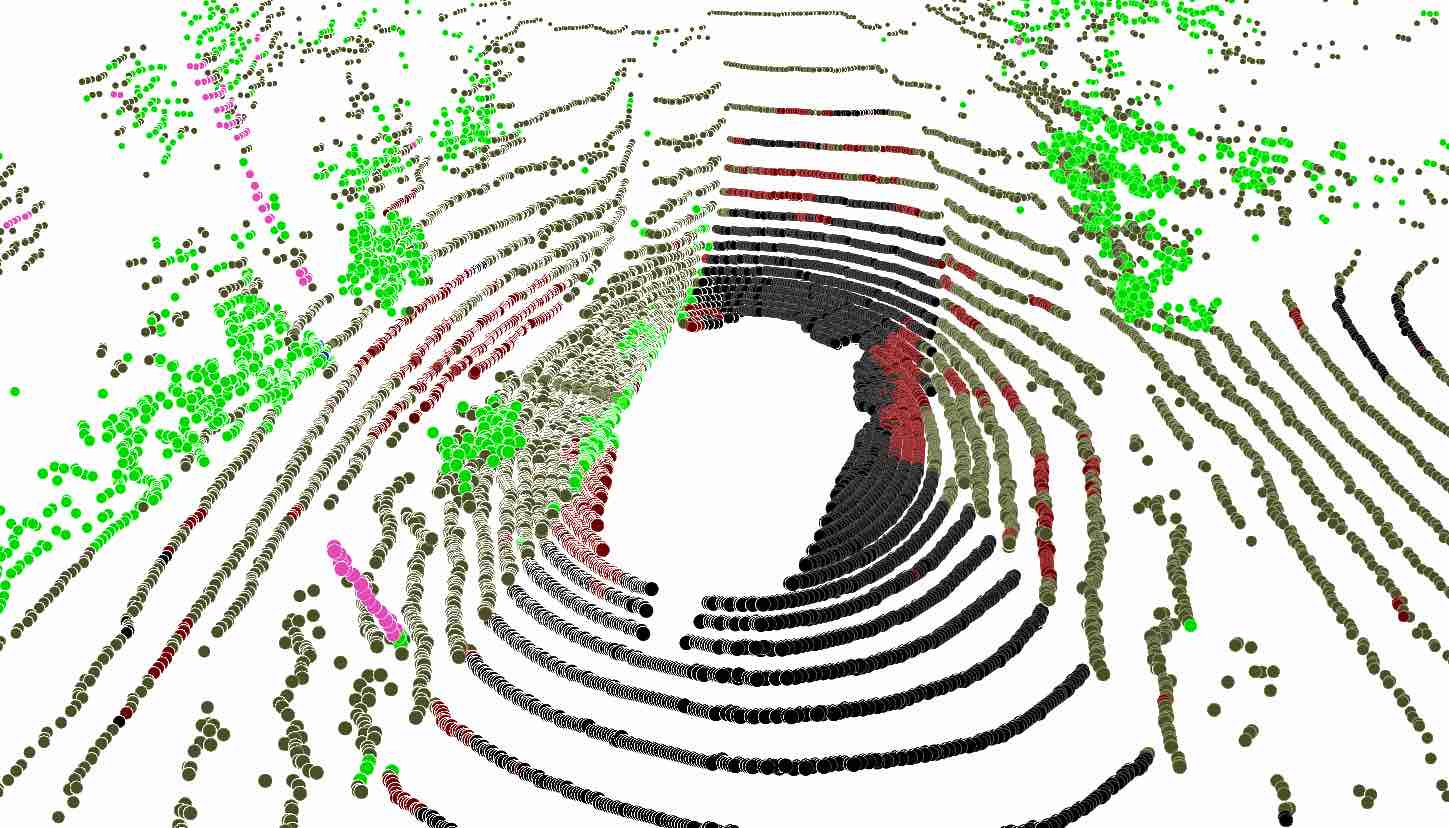}
        \end{overpic} &  
        \begin{overpic}[width=0.21\textwidth]{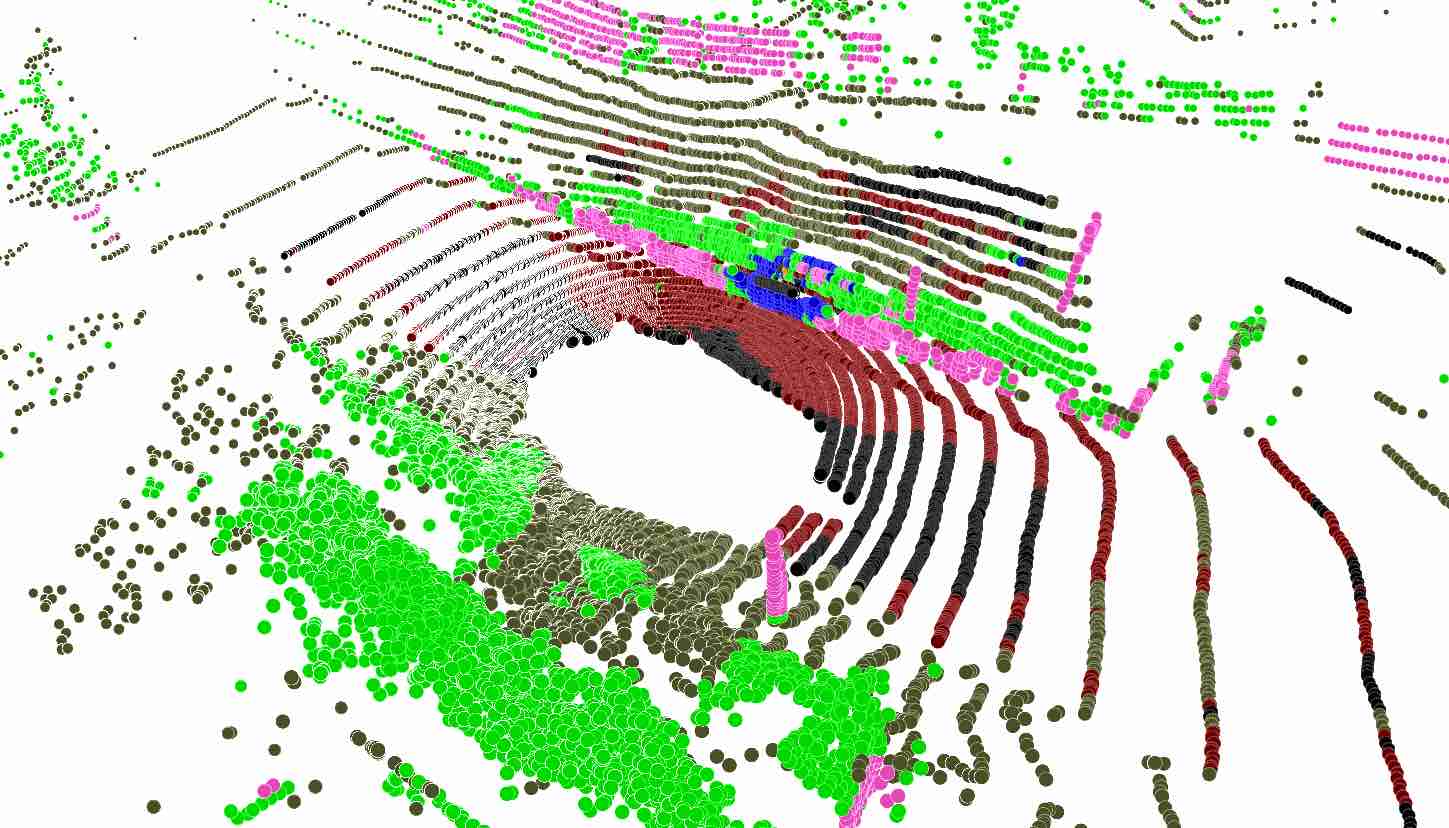}
        \end{overpic} &
        \begin{overpic}[width=0.21\textwidth]{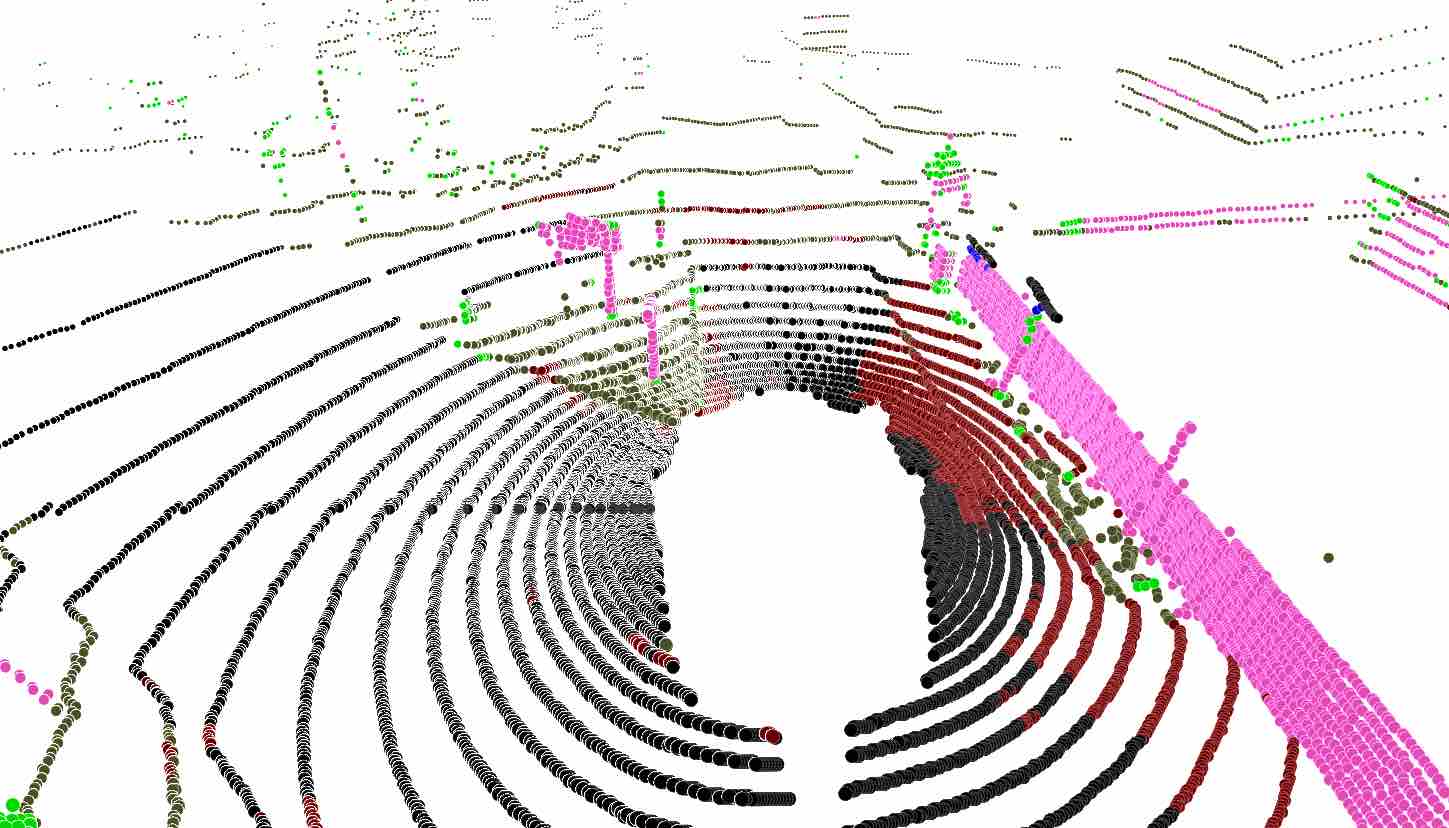}
        \end{overpic}& 
        \begin{overpic}[width=0.21\textwidth]{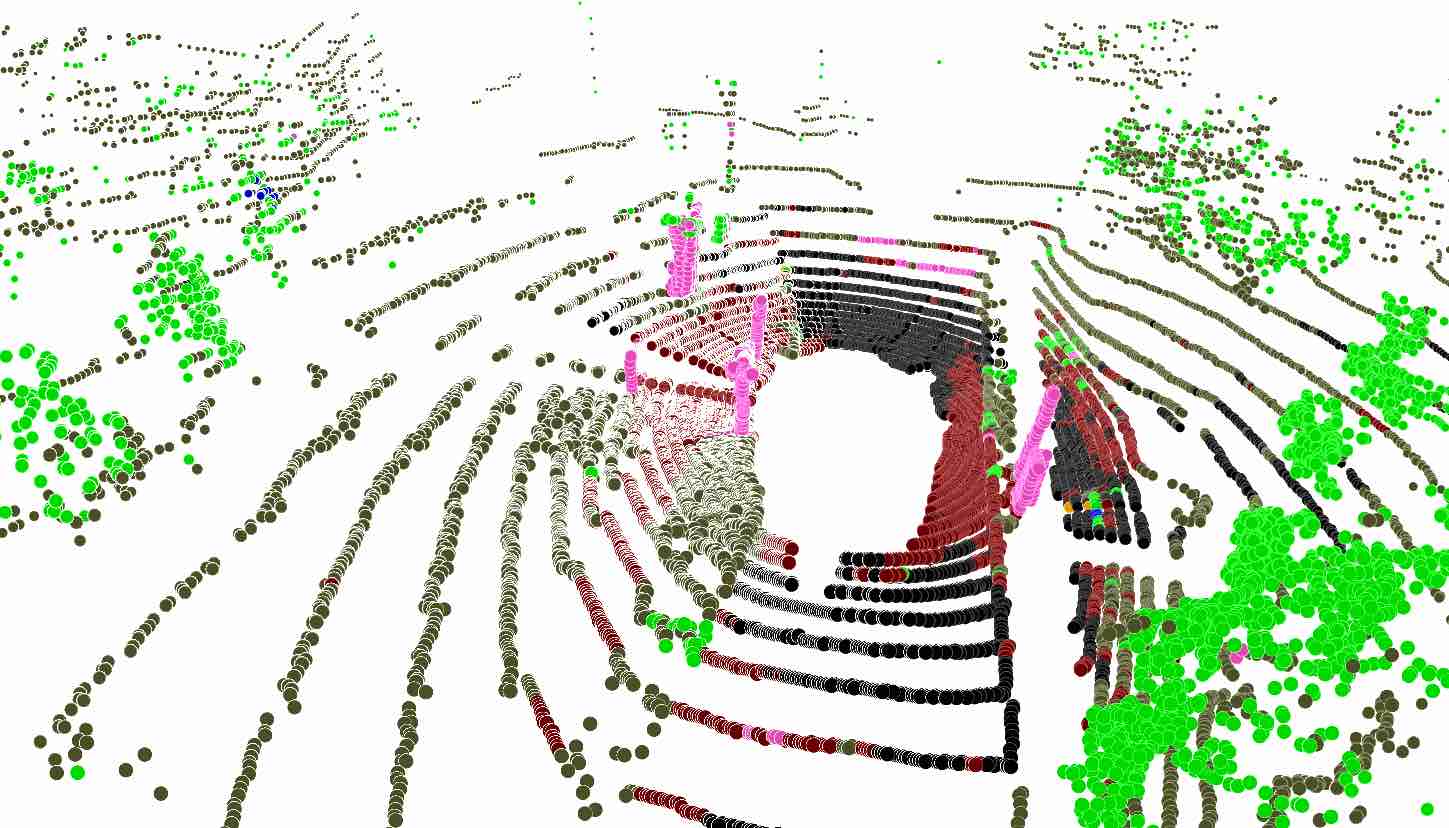}
        \end{overpic}\\
        \begin{overpic}[width=0.21\textwidth]{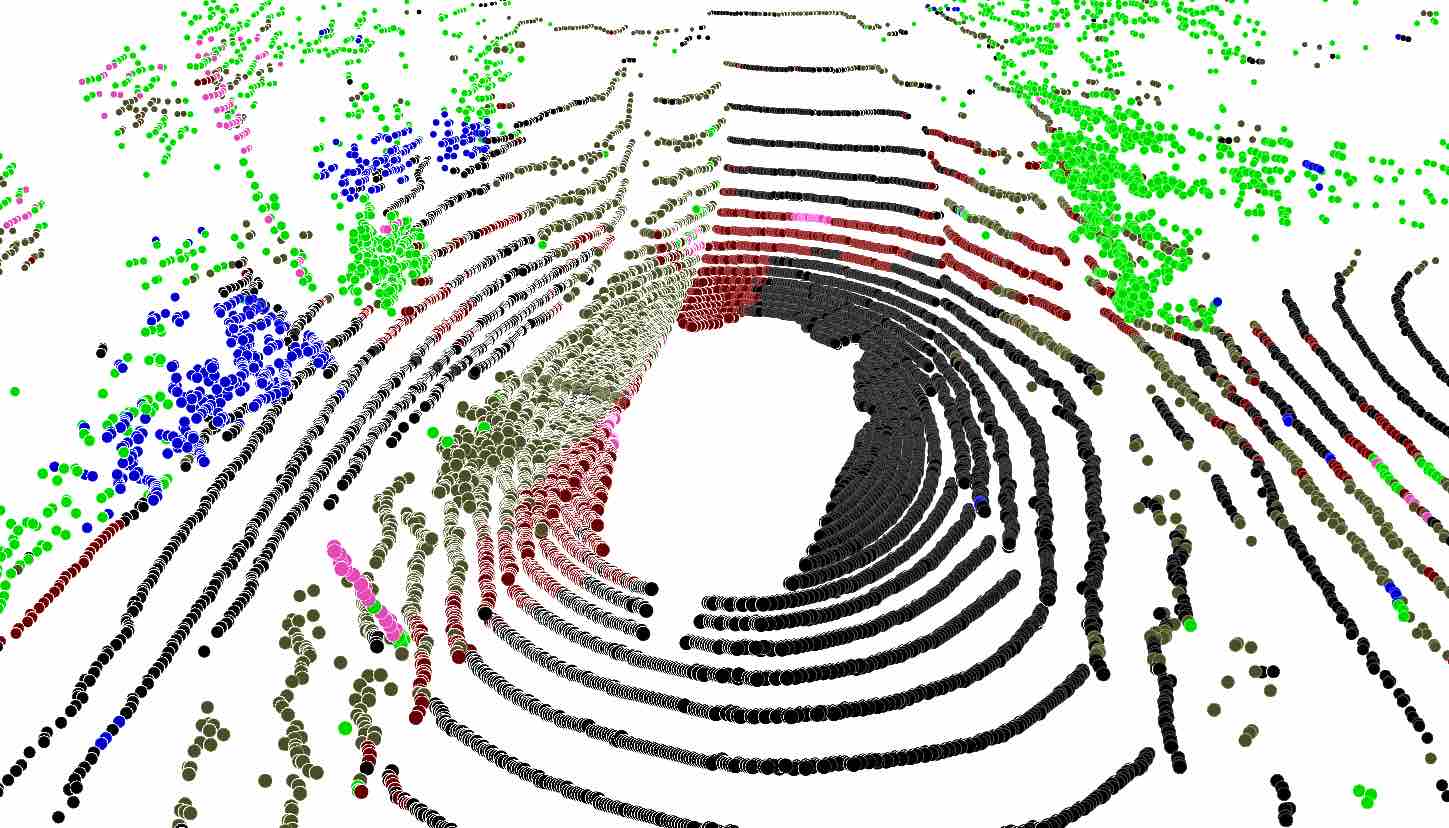}
        \end{overpic} &  
        \begin{overpic}[width=0.21\textwidth]{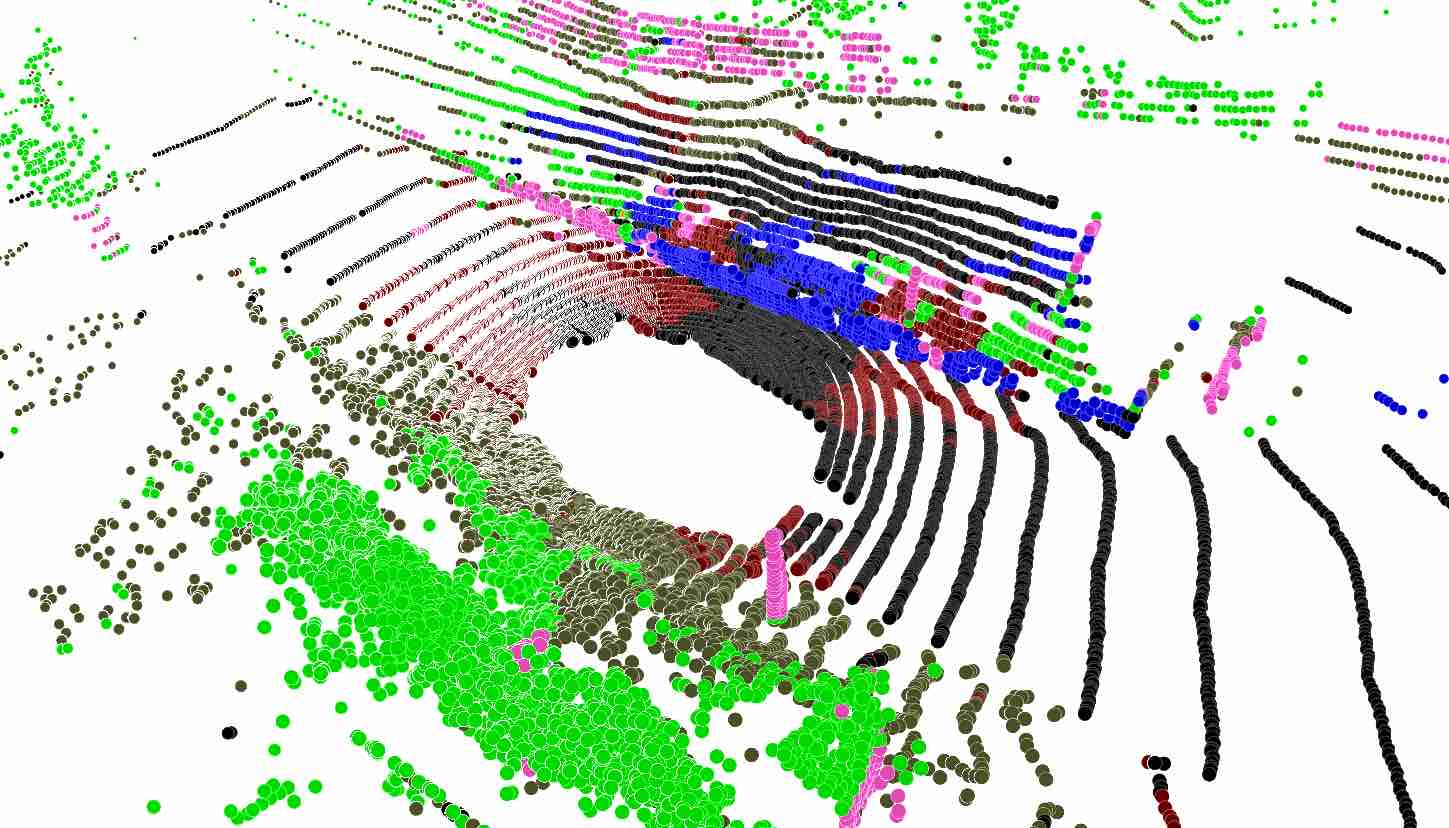}
        \end{overpic} &
        \begin{overpic}[width=0.21\textwidth]{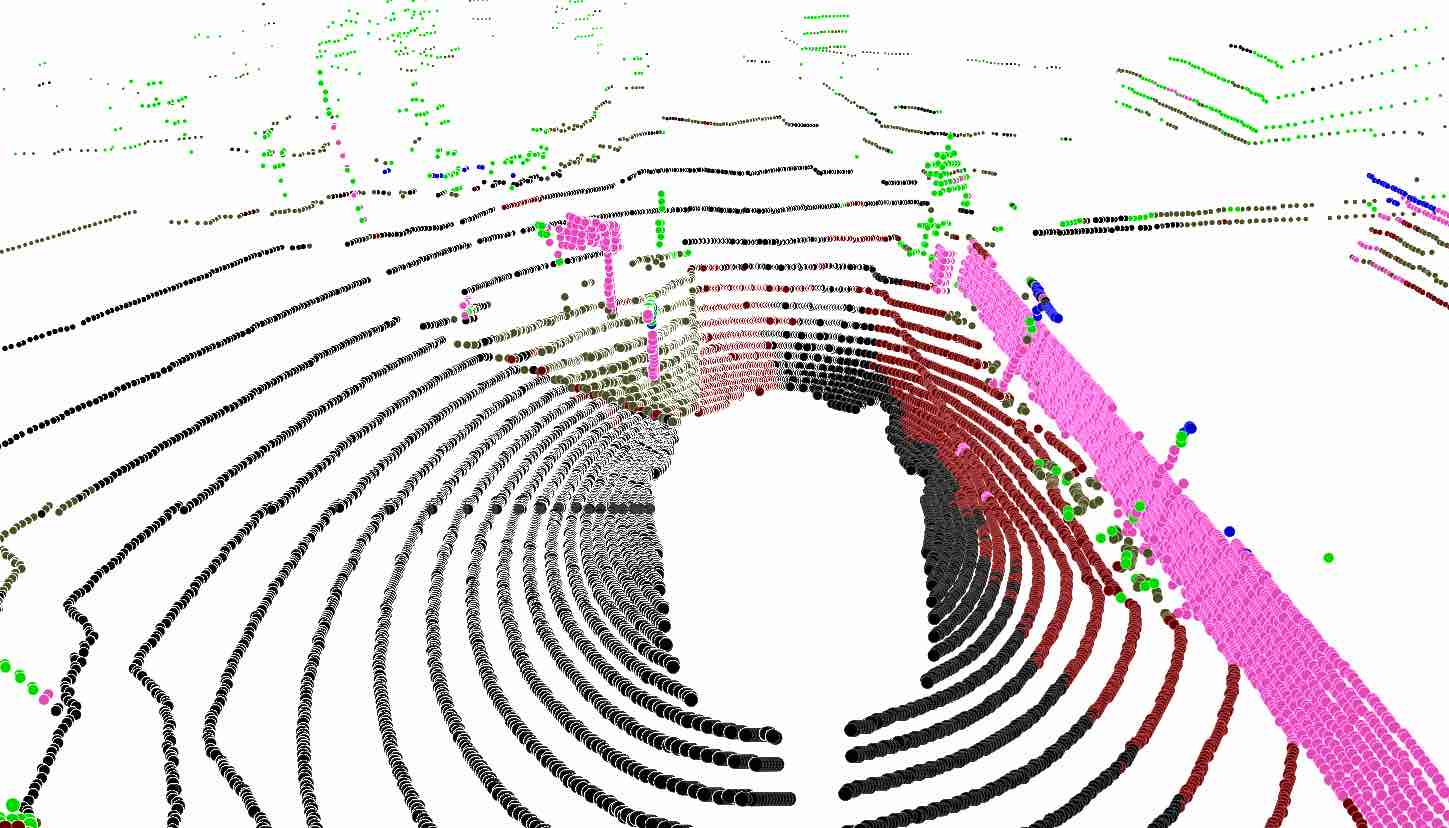}
        \end{overpic}& 
        \begin{overpic}[width=0.21\textwidth]{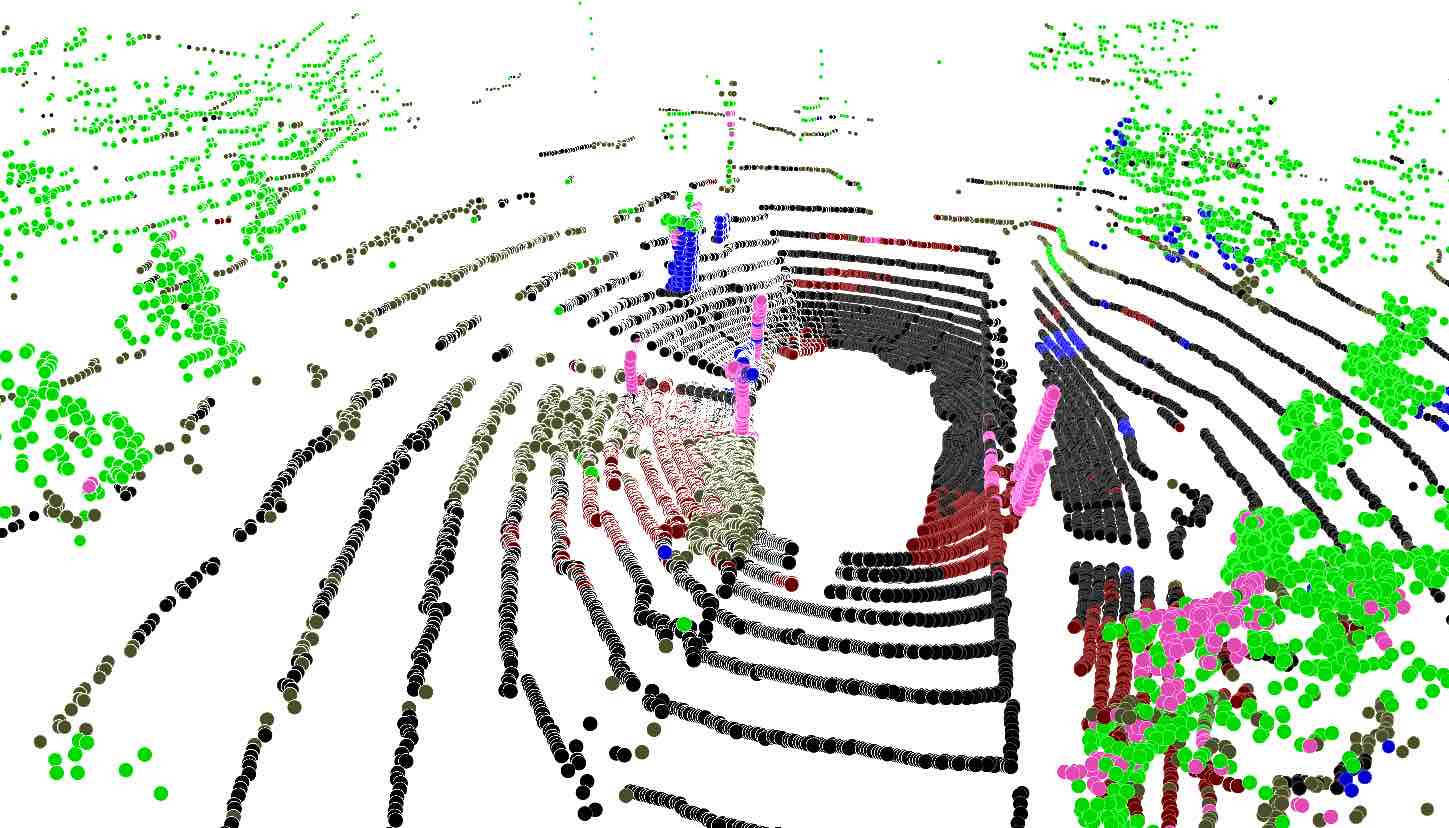}
        \end{overpic}\\
        \begin{overpic}[width=0.21\textwidth]{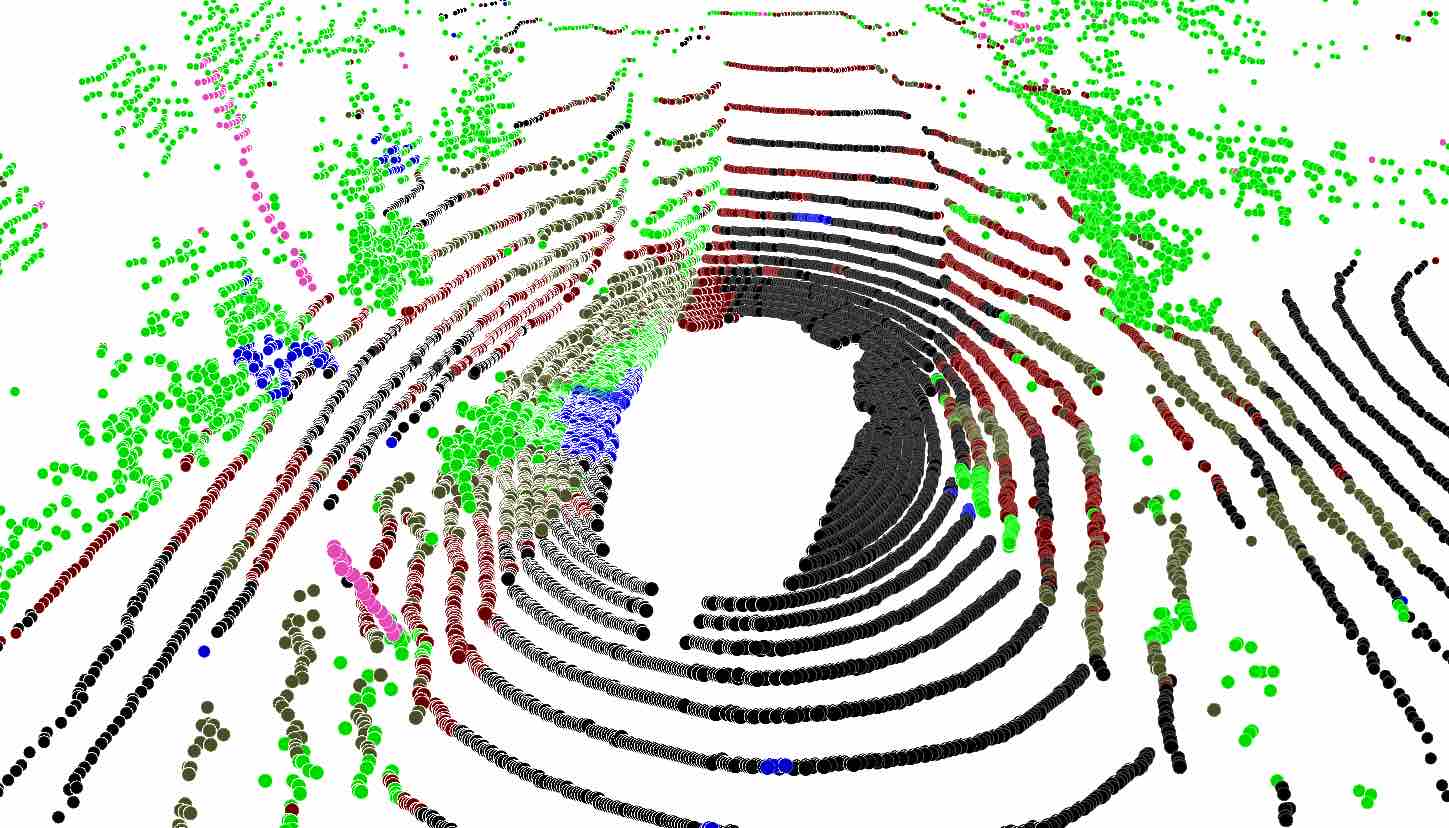}
        \end{overpic} &  
        \begin{overpic}[width=0.21\textwidth]{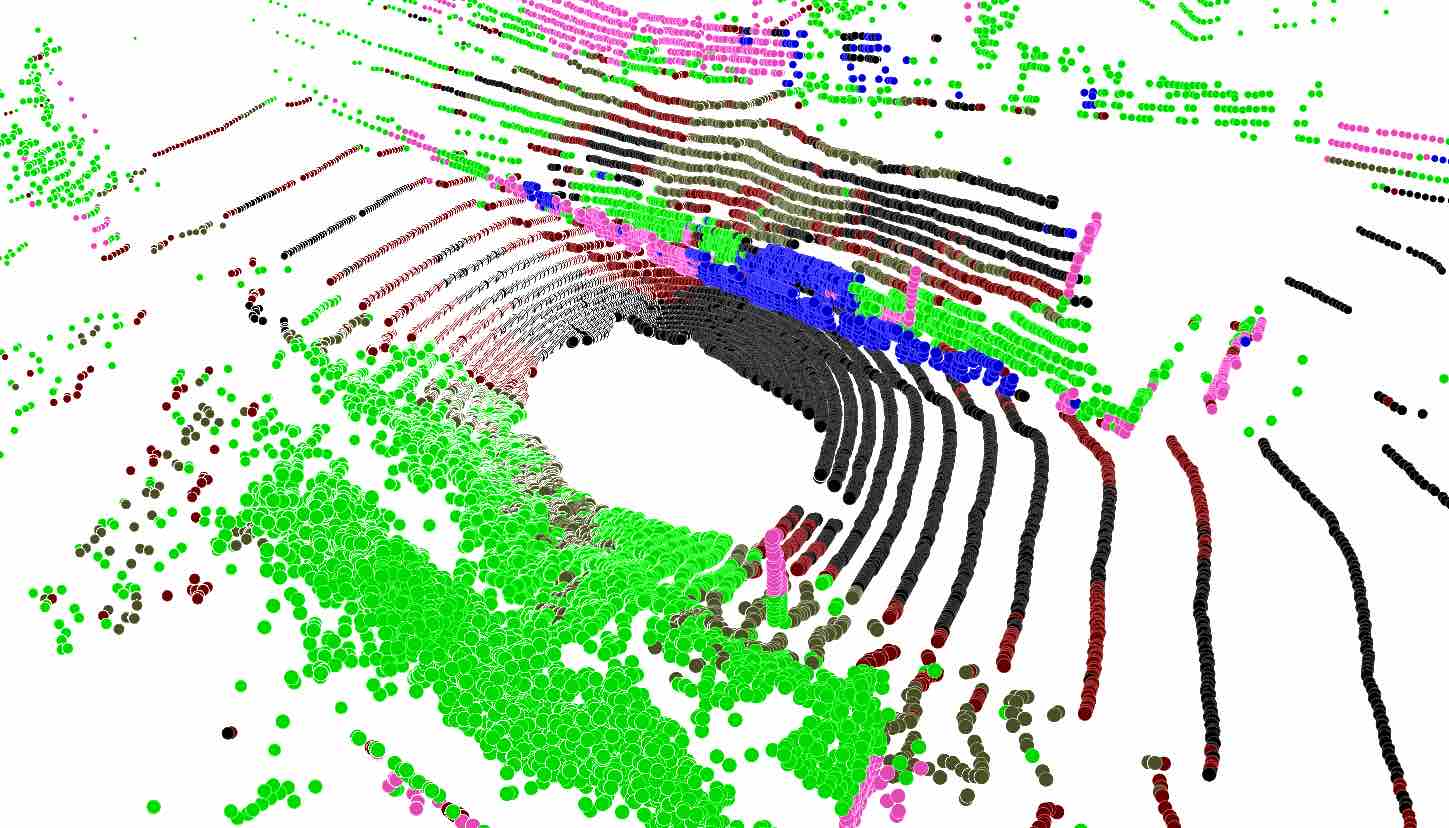}
        \end{overpic} &
        \begin{overpic}[width=0.21\textwidth]{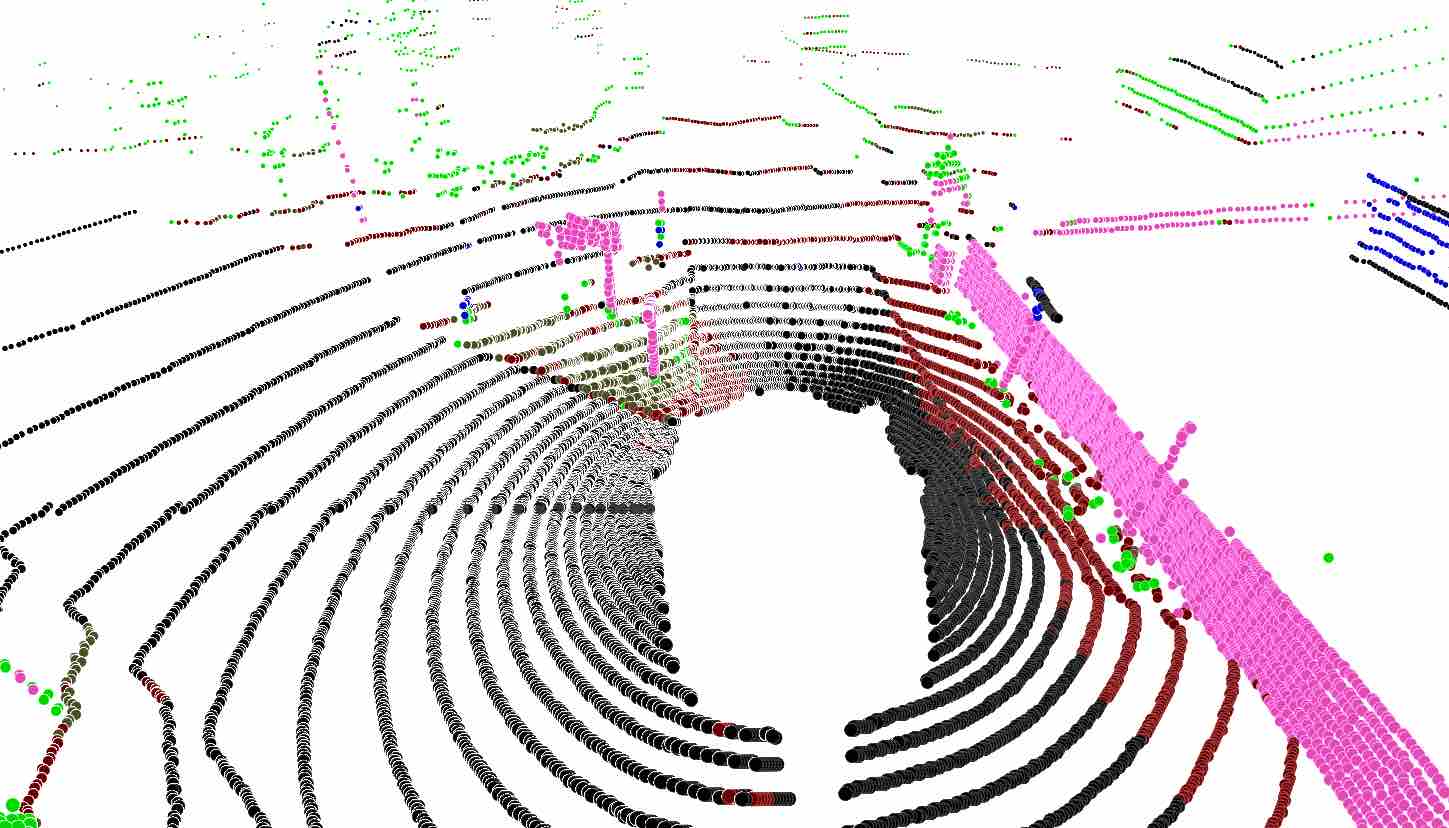}
        \end{overpic}& 
        \begin{overpic}[width=0.21\textwidth]{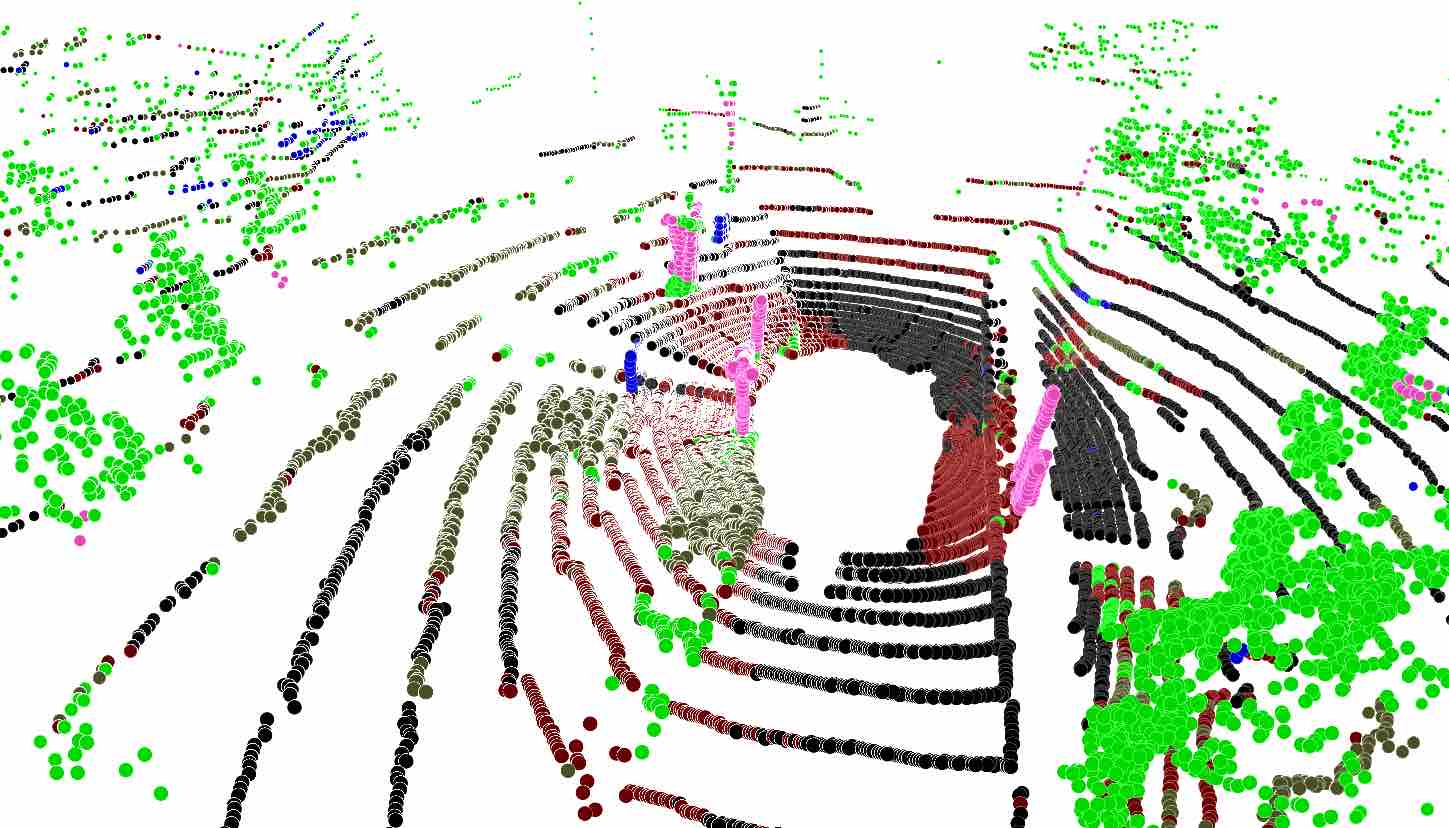}
        \end{overpic}\\
        \begin{overpic}[width=0.21\textwidth]{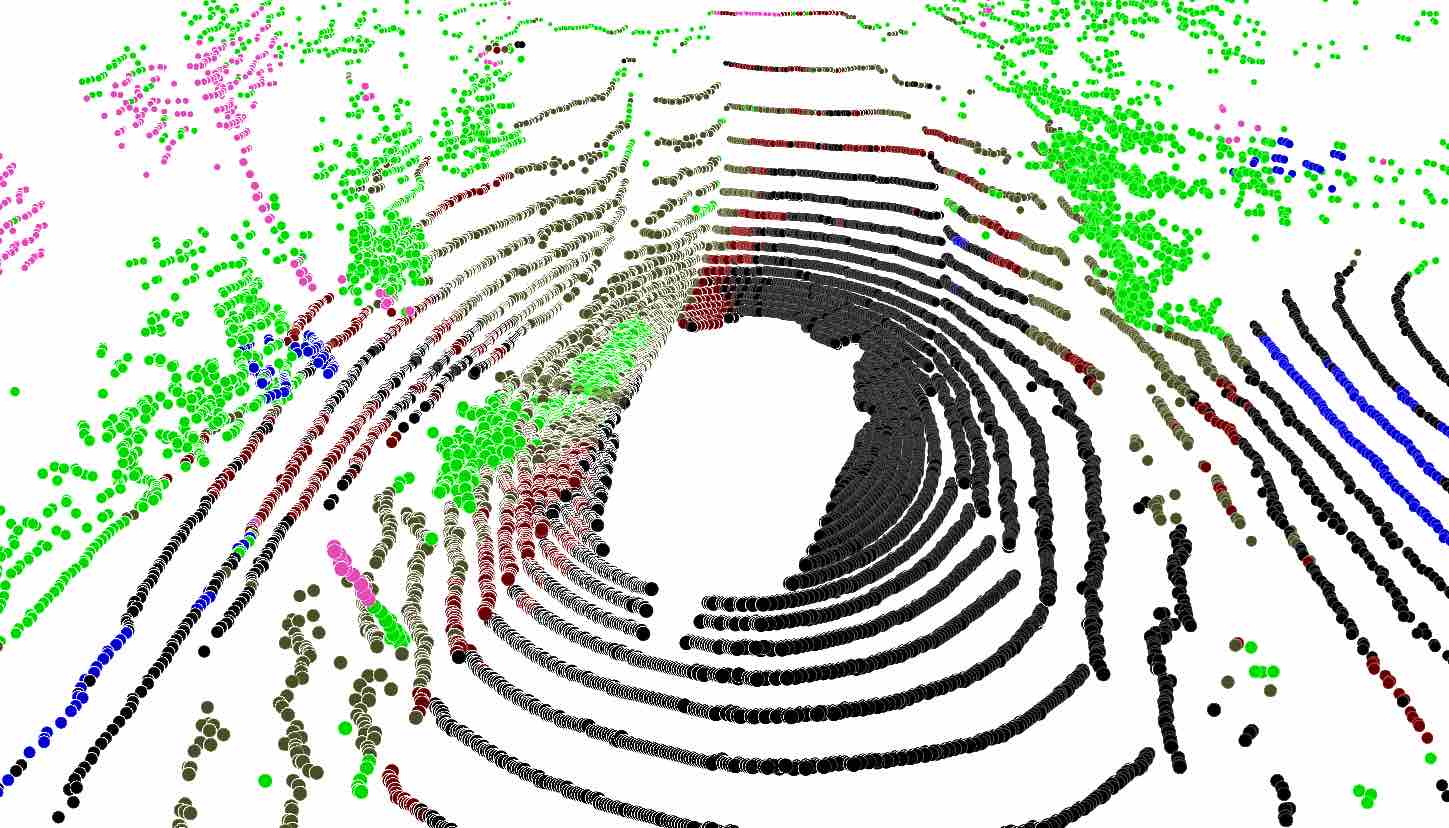}
        \end{overpic} &  
        \begin{overpic}[width=0.21\textwidth]{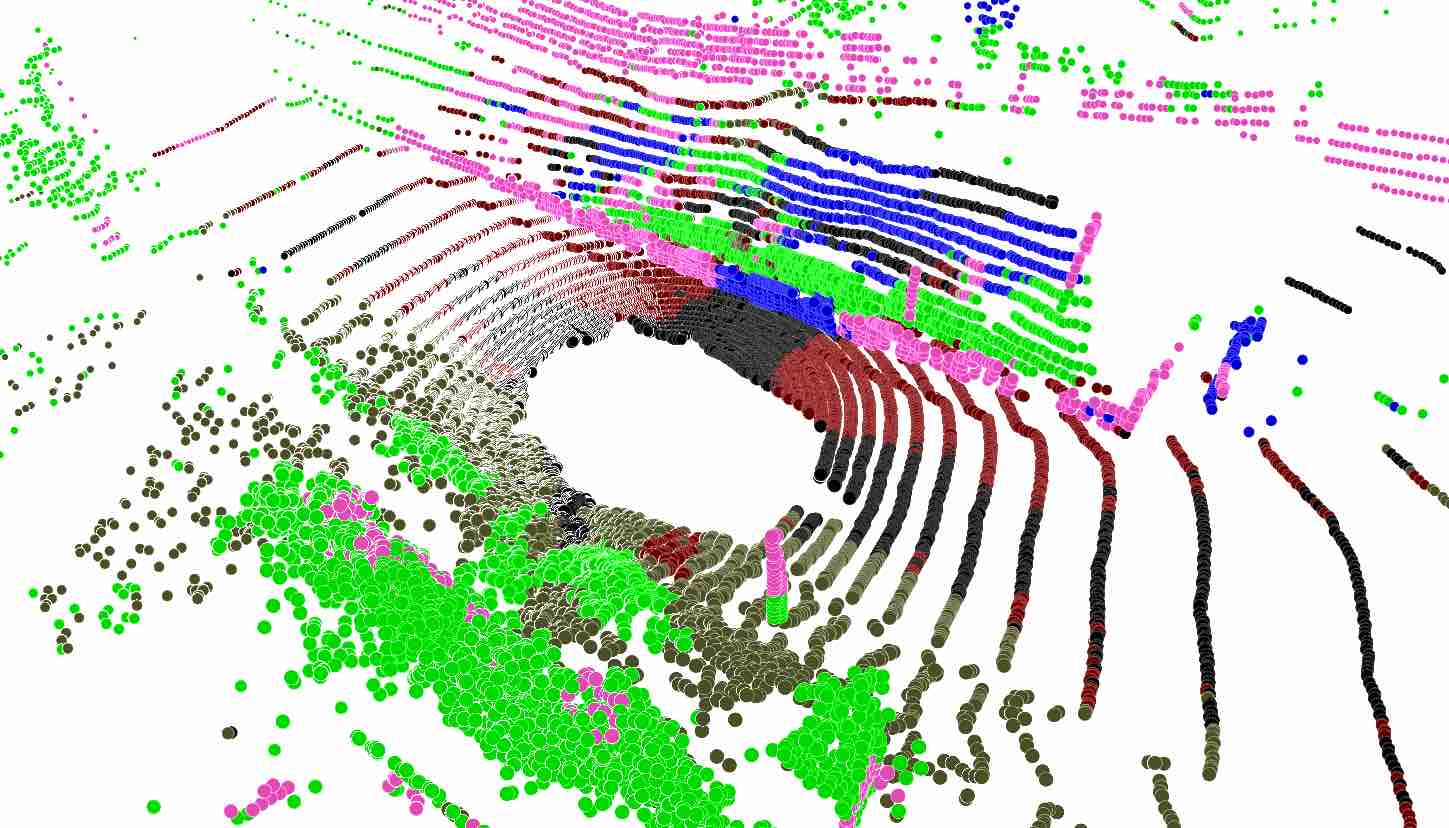}
        \end{overpic} &
        \begin{overpic}[width=0.21\textwidth]{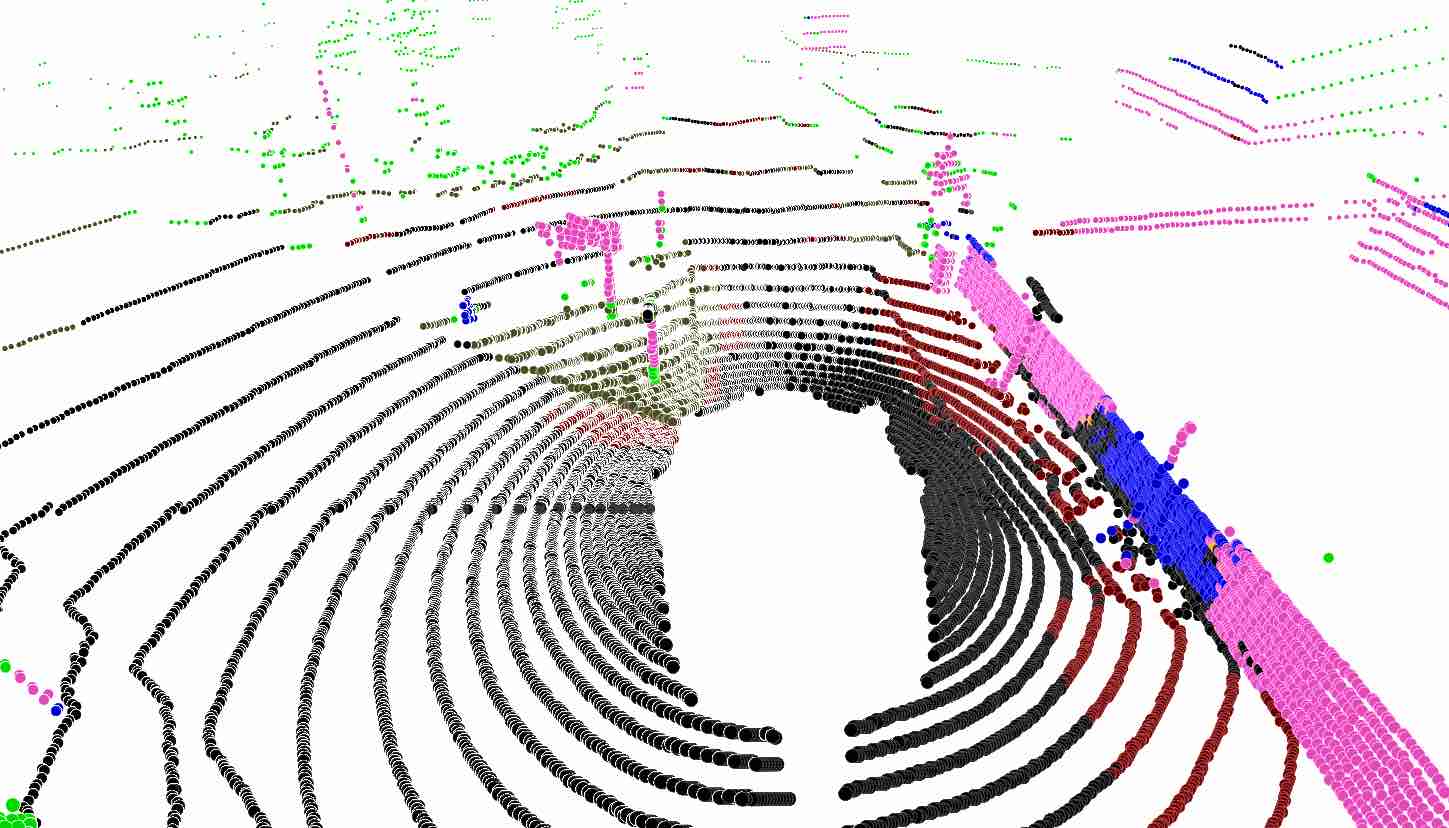}
        \end{overpic}& 
        \begin{overpic}[width=0.21\textwidth]{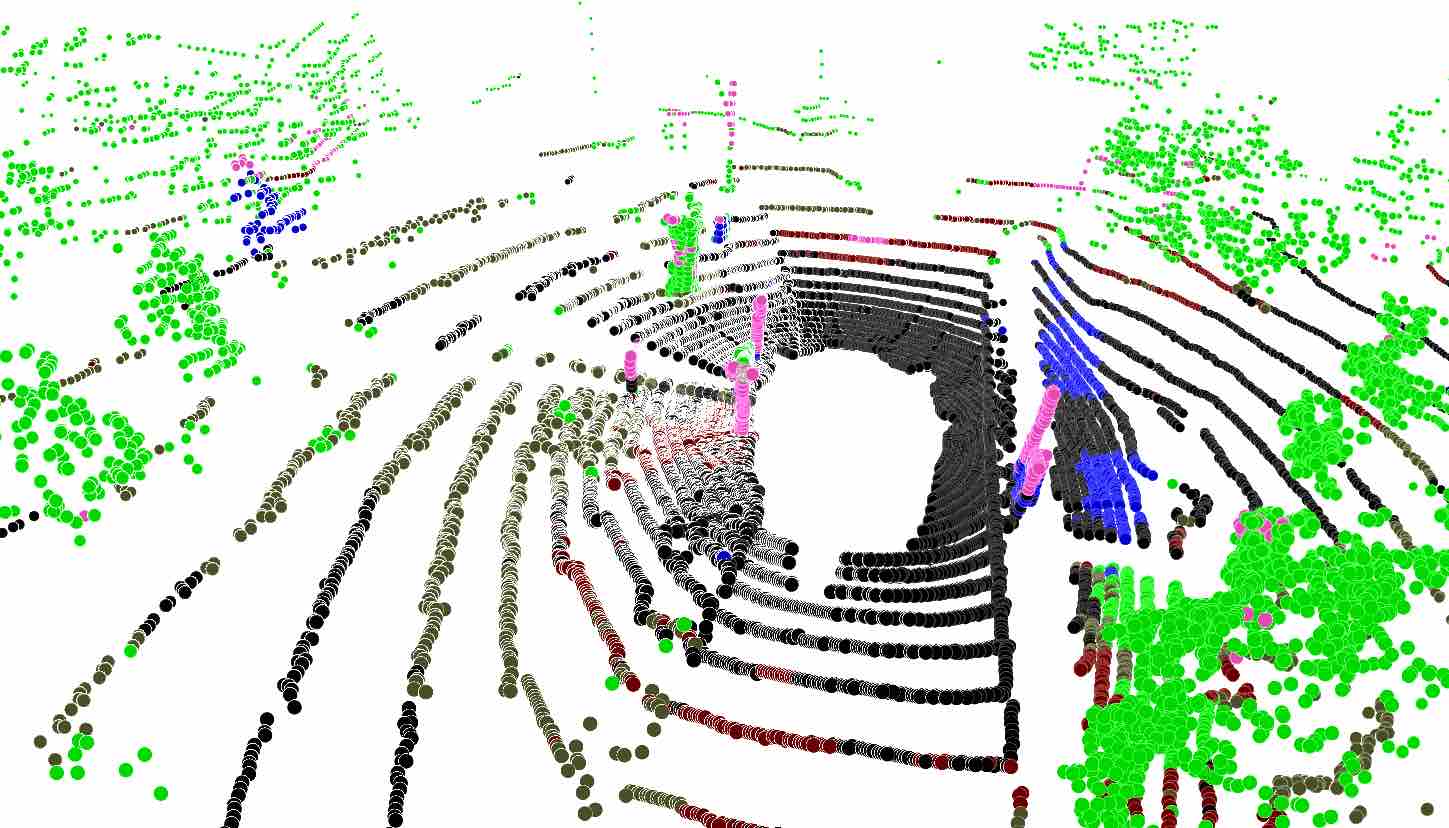}
        \end{overpic}\\
        \begin{overpic}[width=0.21\textwidth]{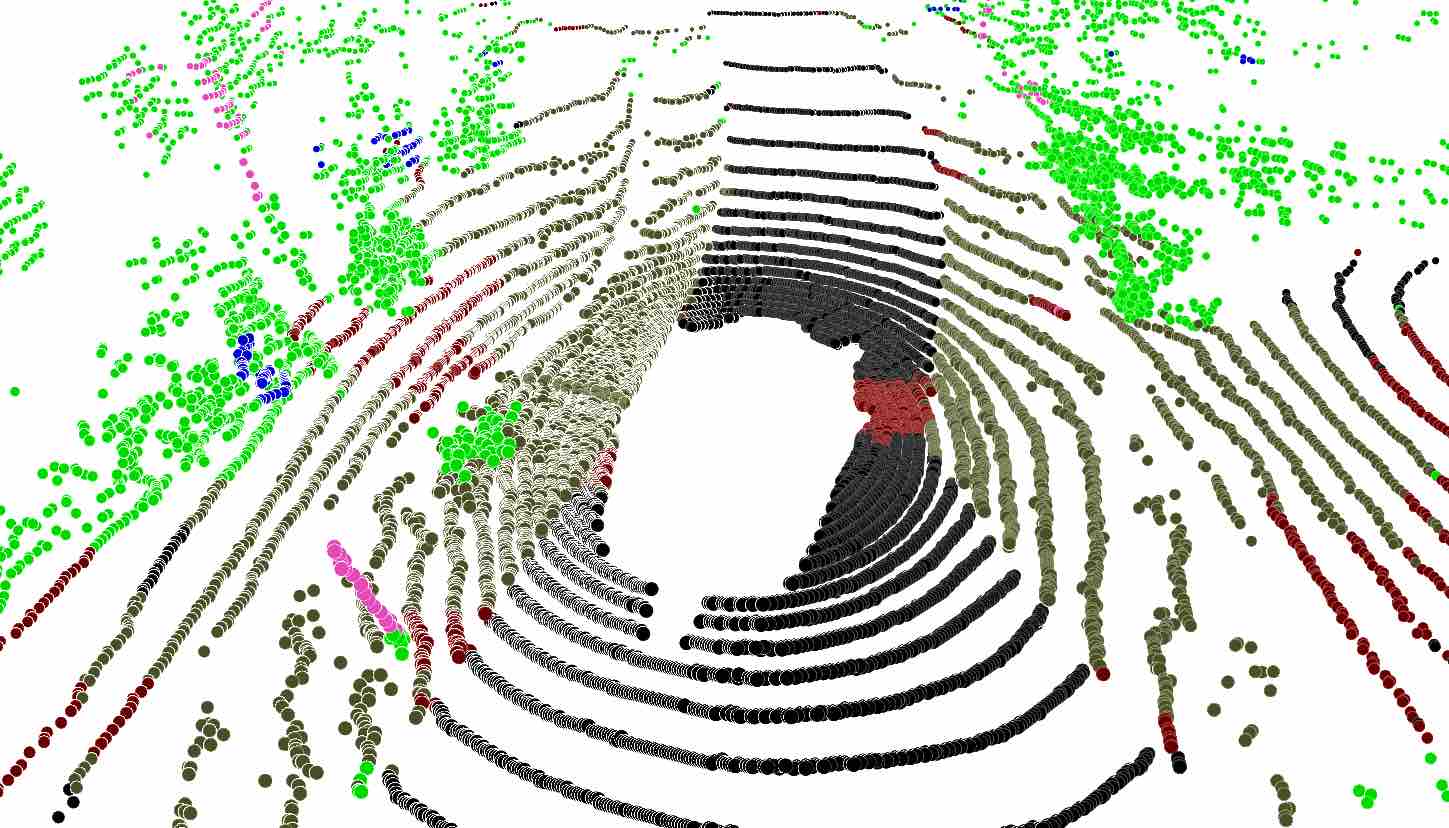}
        \put(-4.5,500){\rotatebox{90}{\color{black}\footnotesize \textbf{source}}}
        \put(-6,440){\rotatebox{90}{\color{black}\footnotesize \textbf{mix3D}}}
        \put(-6,375){\rotatebox{90}{\color{black}\footnotesize \textbf{p.cutmix}}}
        \put(-6,317){\rotatebox{90}{\color{black}\footnotesize \textbf{cosmix}}}
        \put(-6,260){\rotatebox{90}{\color{black}\footnotesize \textbf{ibn}}}
        \put(-6,200){\rotatebox{90}{\color{black}\footnotesize \textbf{robust.}}}
        \put(-5,140){\rotatebox{90}{\color{black}\footnotesize \textbf{sn}}}
        \put(-6,75){\rotatebox{90}{\color{black}\footnotesize \textbf{raycast}}}
        \put(-5,17){\rotatebox{90}{\color{black}\footnotesize \textbf{ours}}}
        \put(-6,-35){\rotatebox{90}{\color{black}\footnotesize \textbf{gt}}}
        \put(185, 542){\color{black}\footnotesize \textbf{nuScenes}}
        \end{overpic} &  
        \begin{overpic}[width=0.21\textwidth]{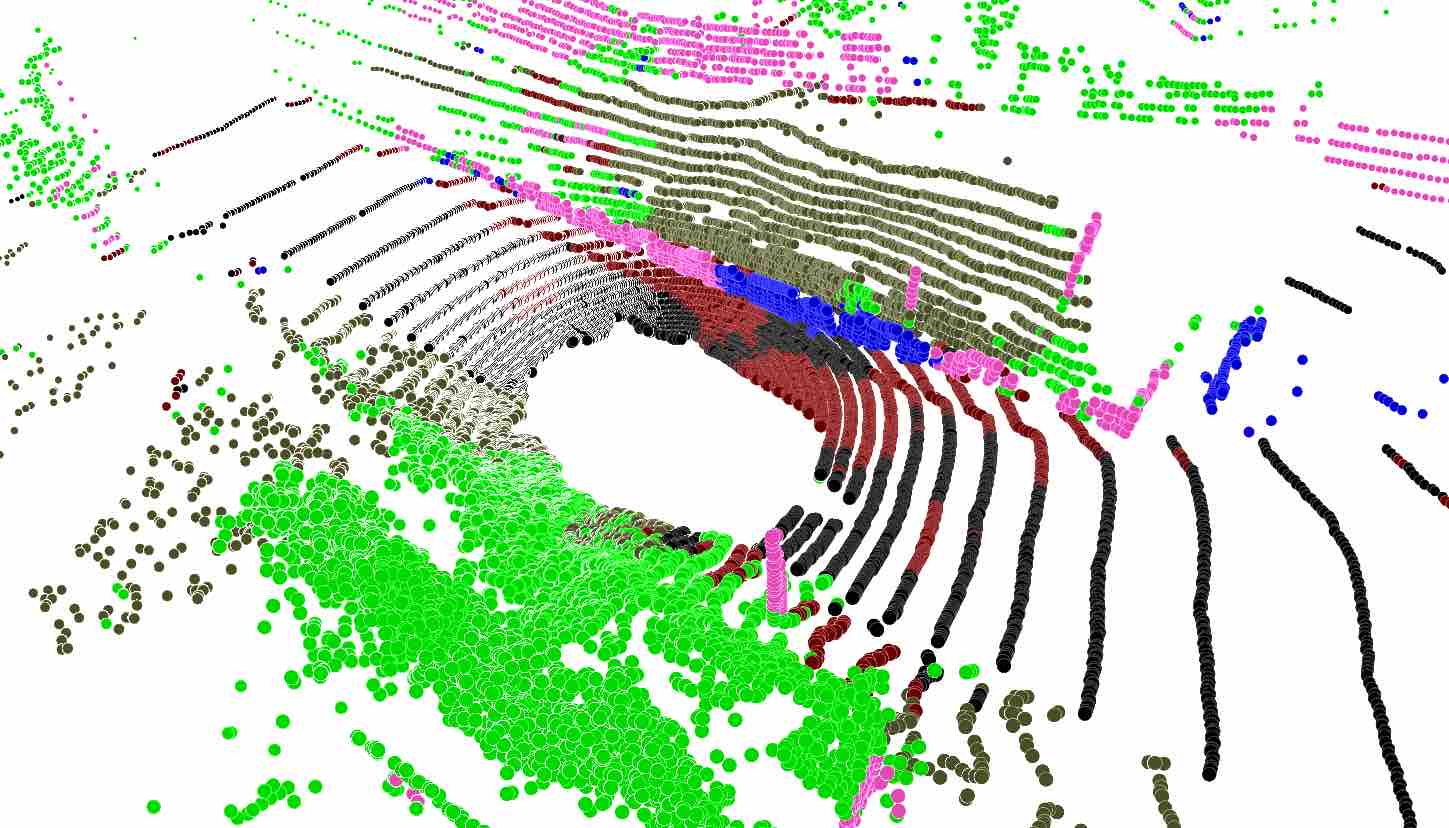}
        \end{overpic} &
        \begin{overpic}[width=0.21\textwidth]{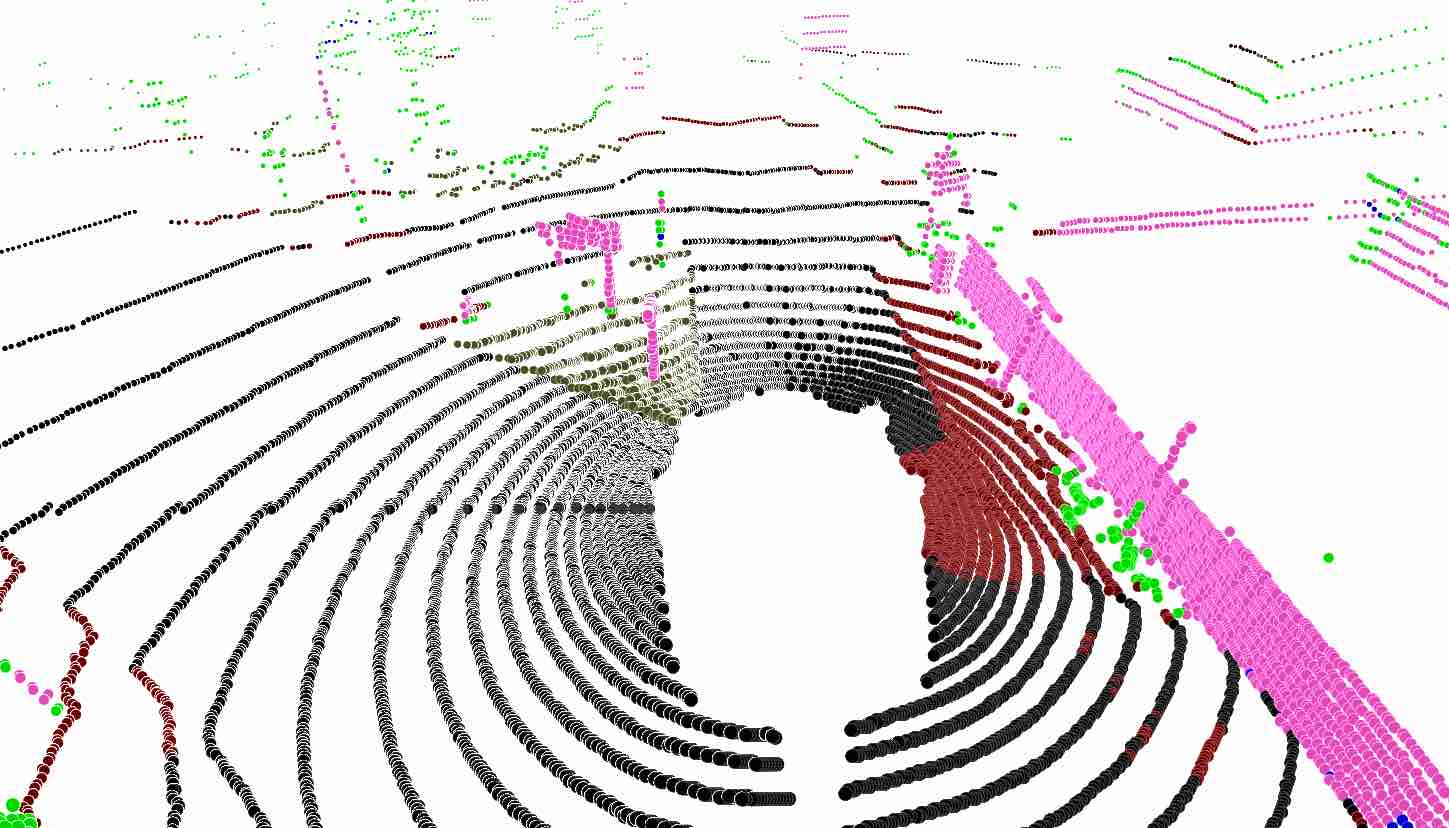}
        \end{overpic}& 
        \begin{overpic}[width=0.21\textwidth]{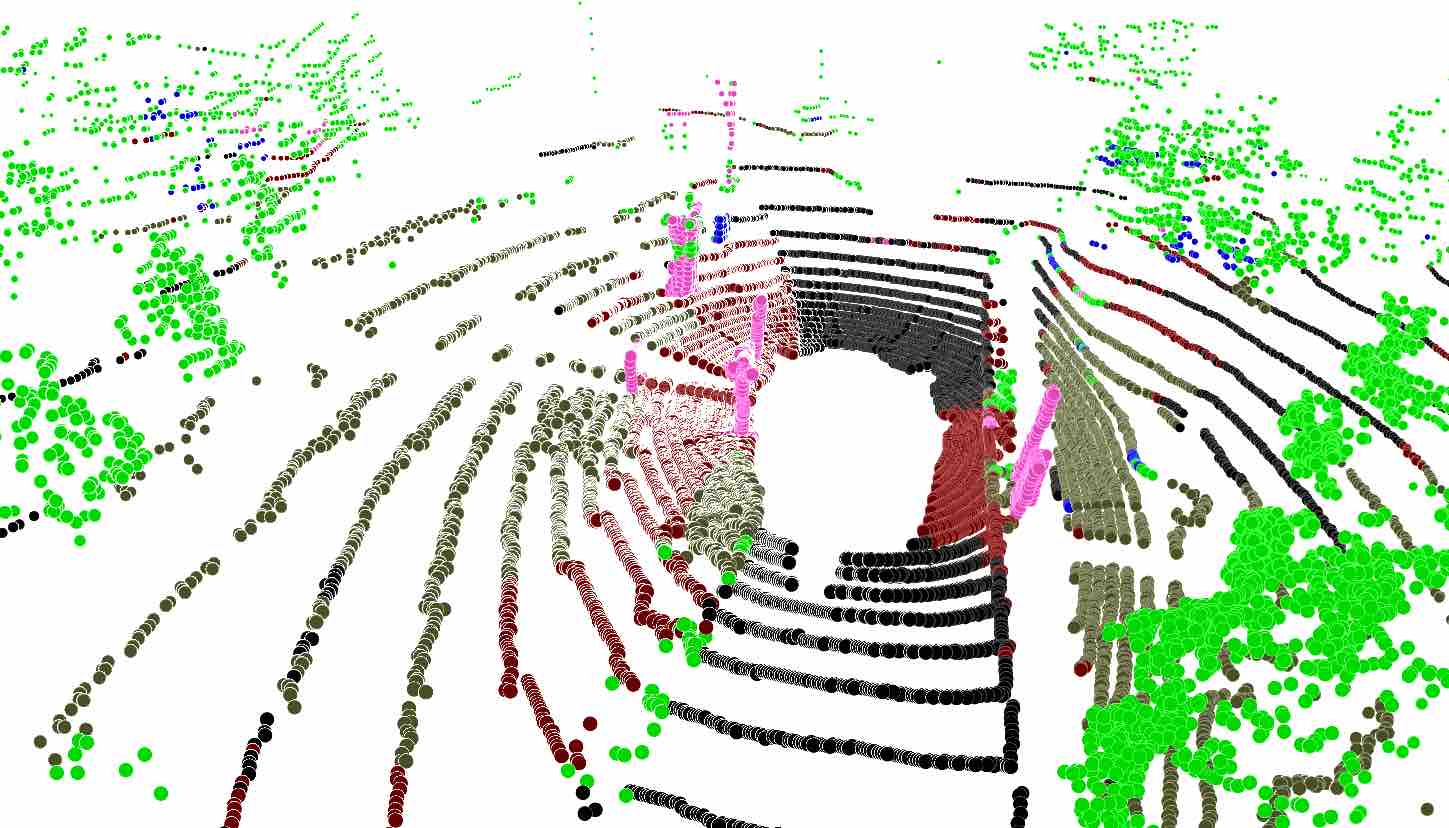}
        \end{overpic}\\
        \begin{overpic}[width=0.21\textwidth]{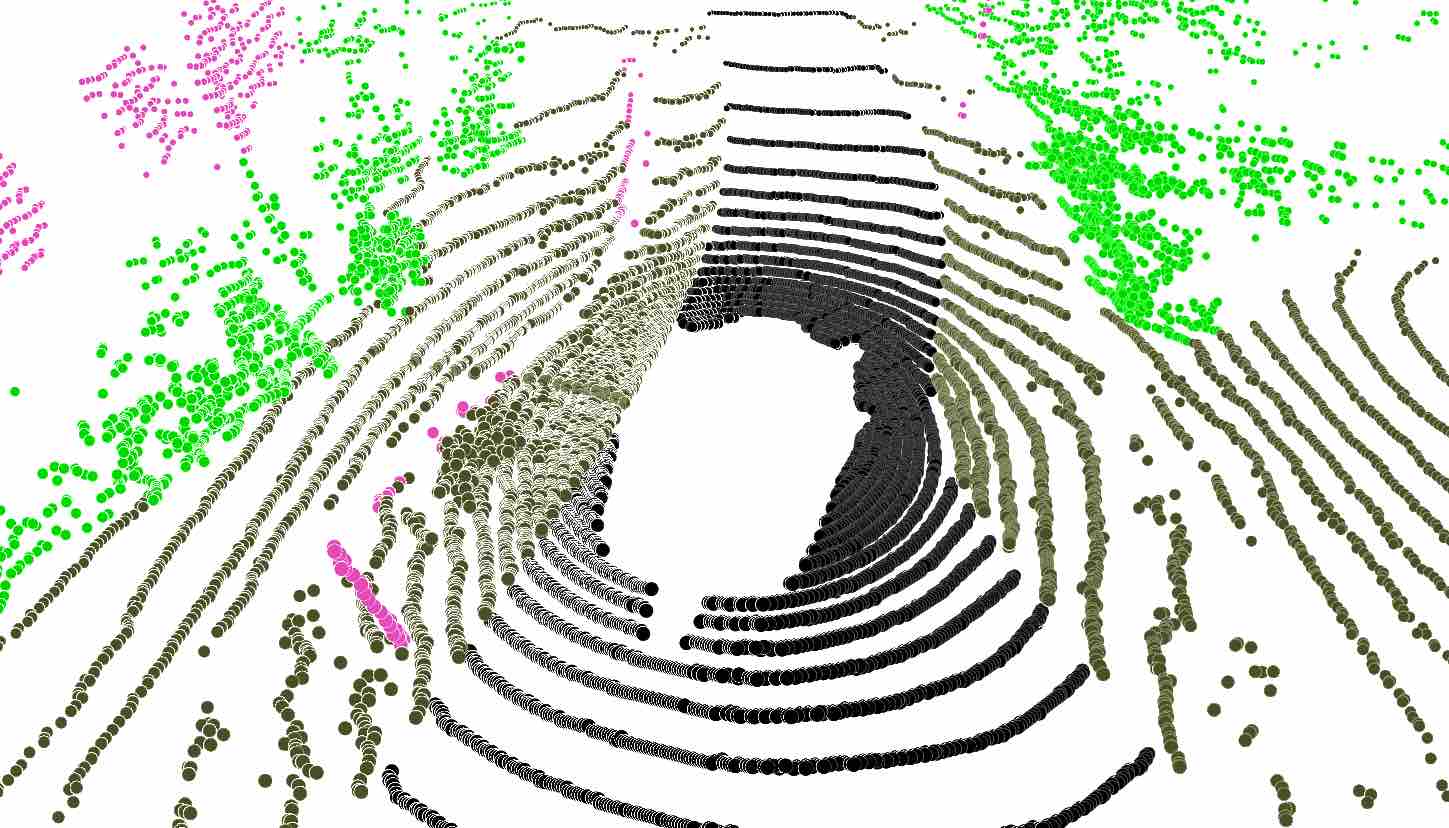}
        \end{overpic} &  
        \begin{overpic}[width=0.21\textwidth]{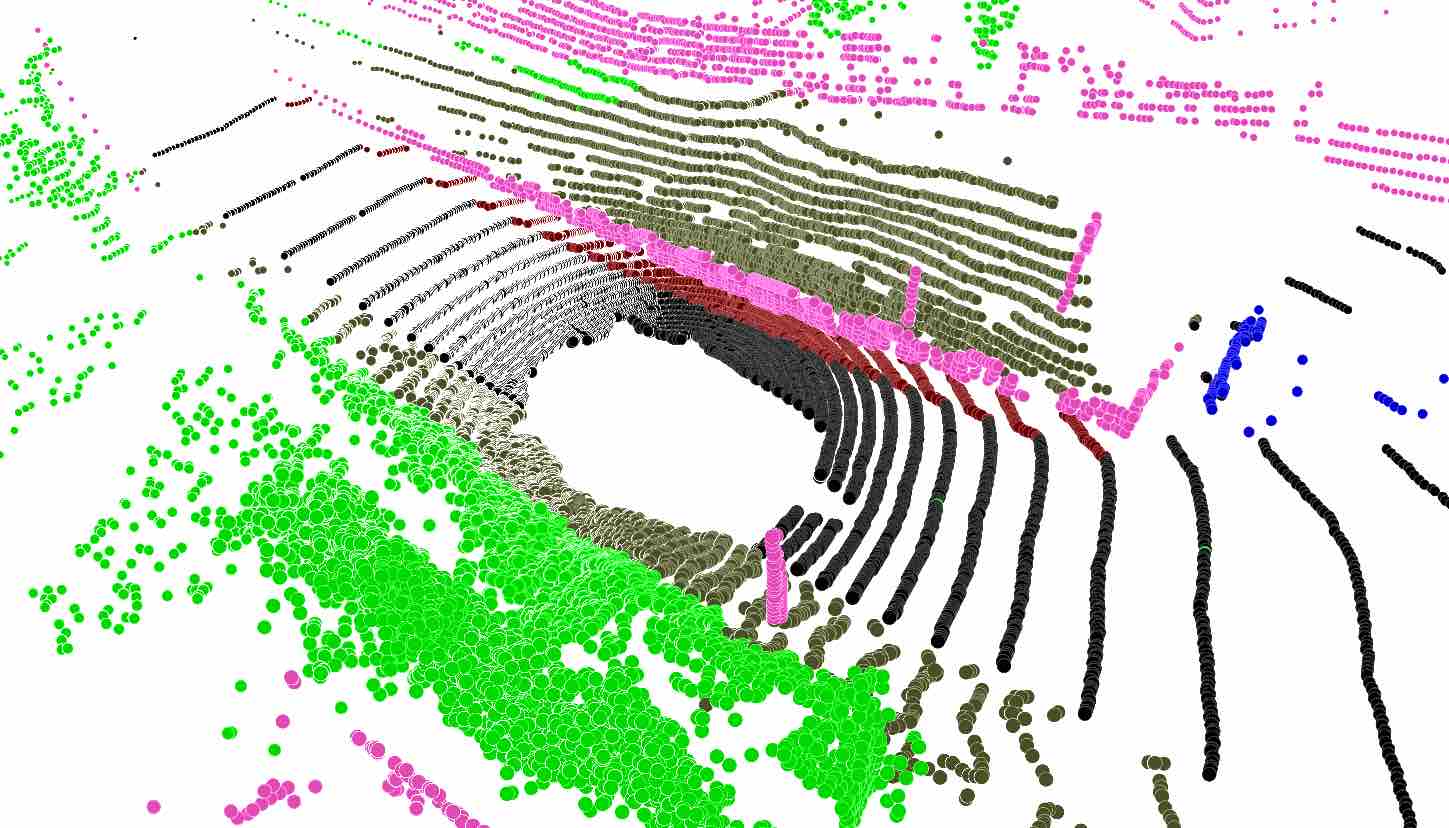}
        \end{overpic} &
        \begin{overpic}[width=0.21\textwidth]{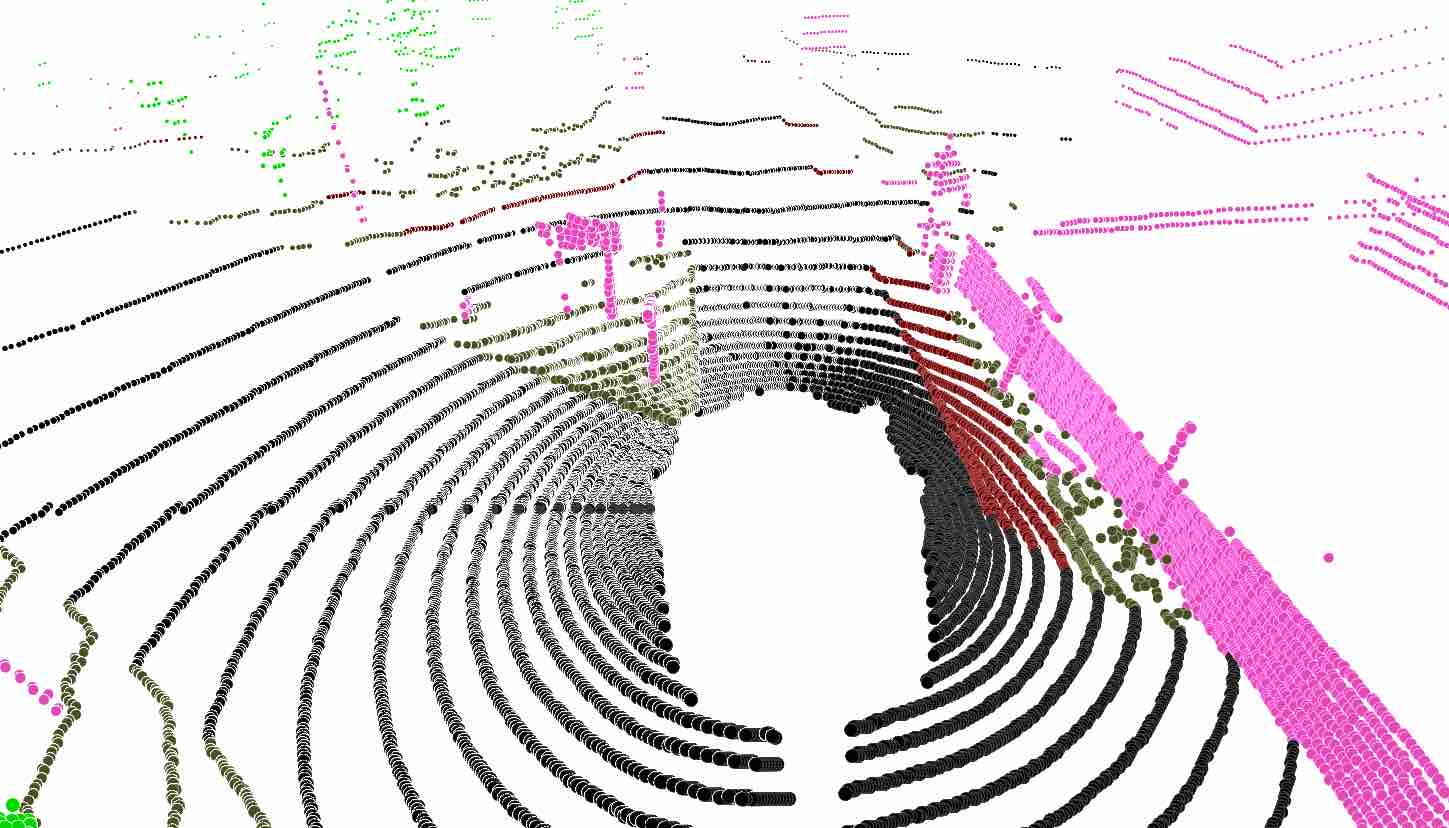}
        \end{overpic}& 
        \begin{overpic}[width=0.21\textwidth]{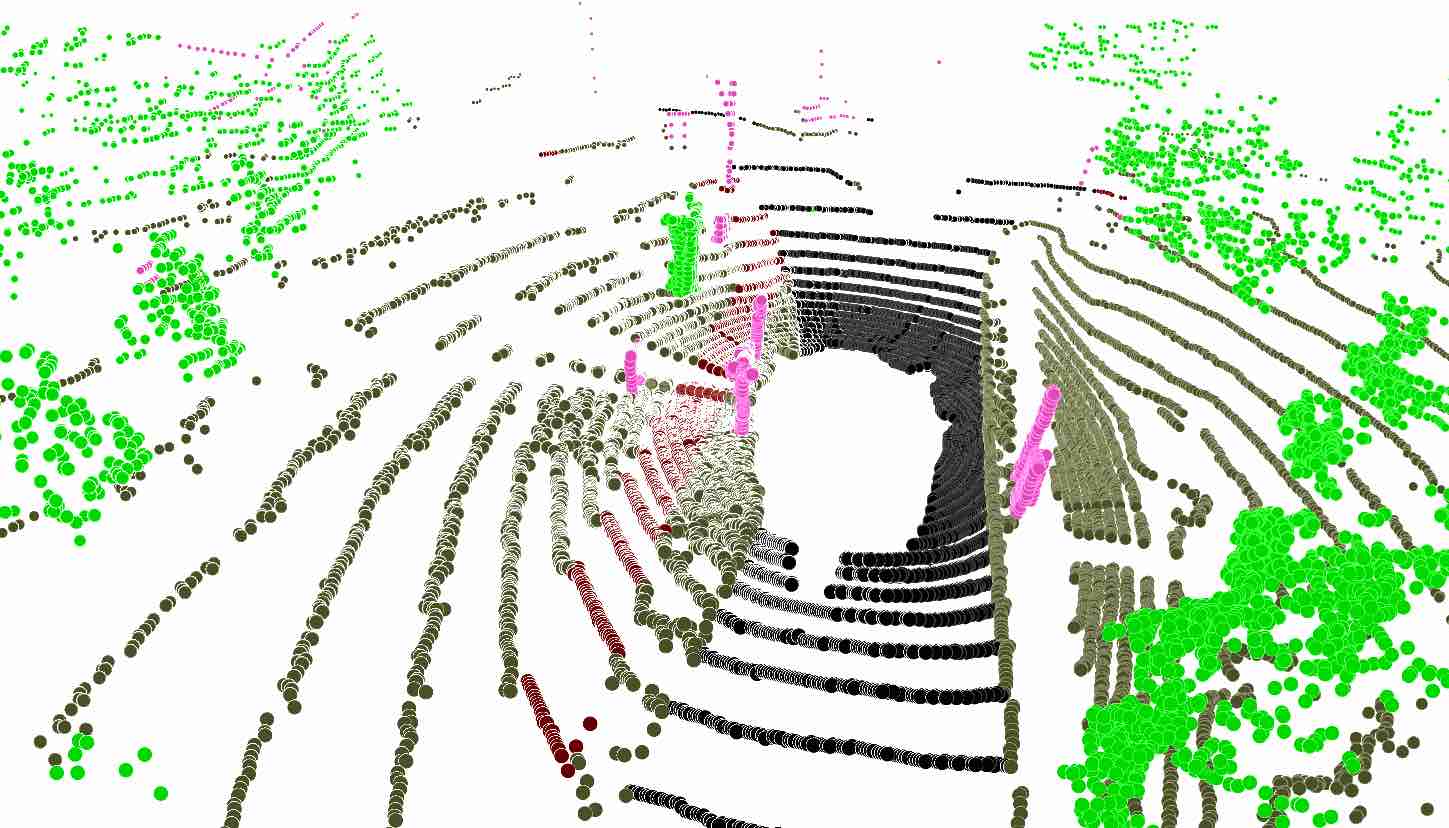}
        \end{overpic}\\
    \end{tabular}
    \vspace{-4mm}
    \caption{\textbf{Qualitative results.} \textit{Top:} SemanticKITTI$\to$nuScenes. \lidog improves over source and baselines, \eg, we observe the improvements in \textit{terrain}, \textit{road}, and \textit{manmade}.}
    \label{fig:supp_qualitative_kitti}
\end{figure*}

\begin{figure*}[t]
\centering
    \setlength\tabcolsep{1.pt}
    \begin{tabular}{cccc}
    \raggedright
        \begin{overpic}[width=0.21\textwidth]{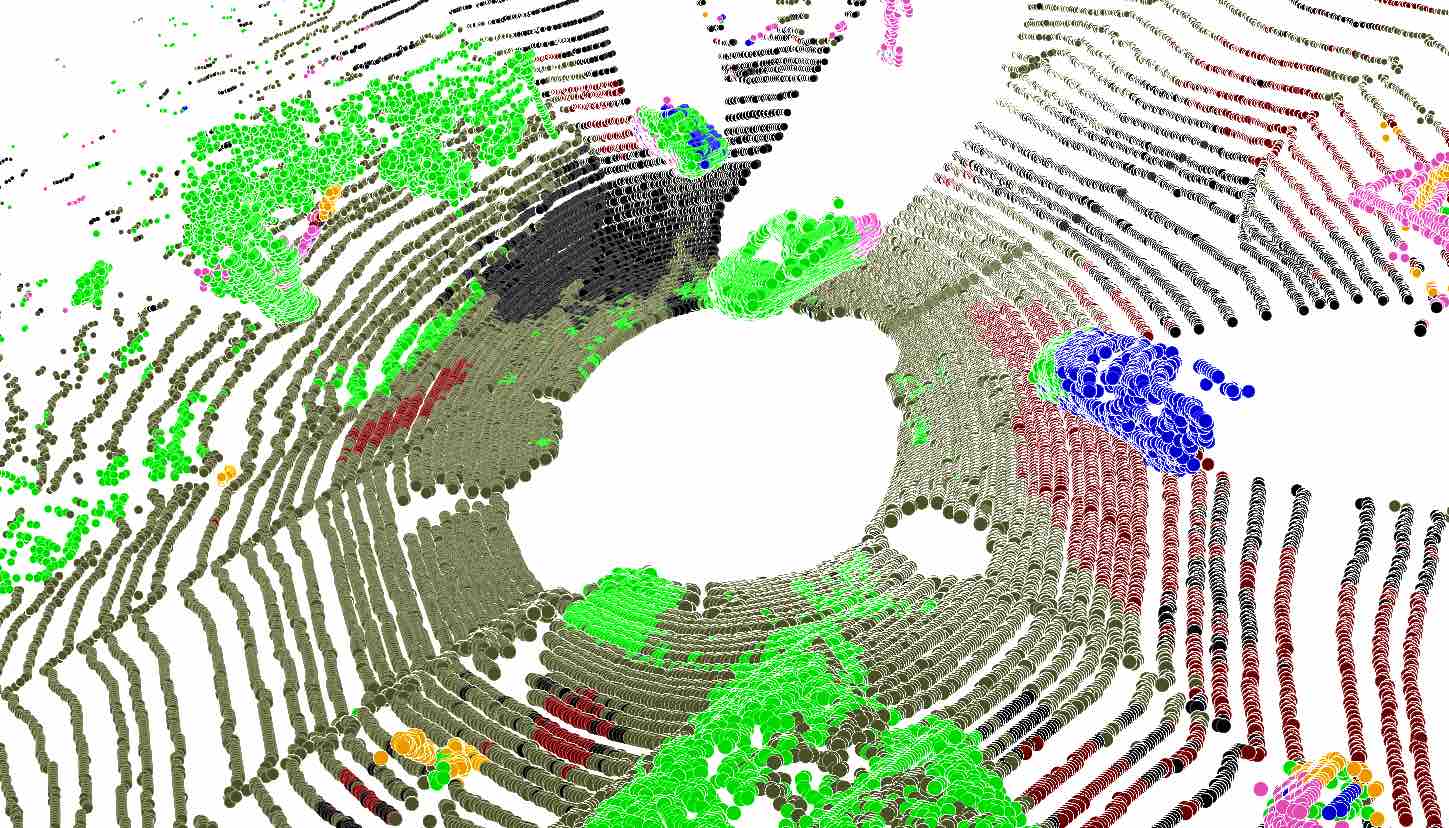}
        \end{overpic} &  
        \begin{overpic}[width=0.21\textwidth]{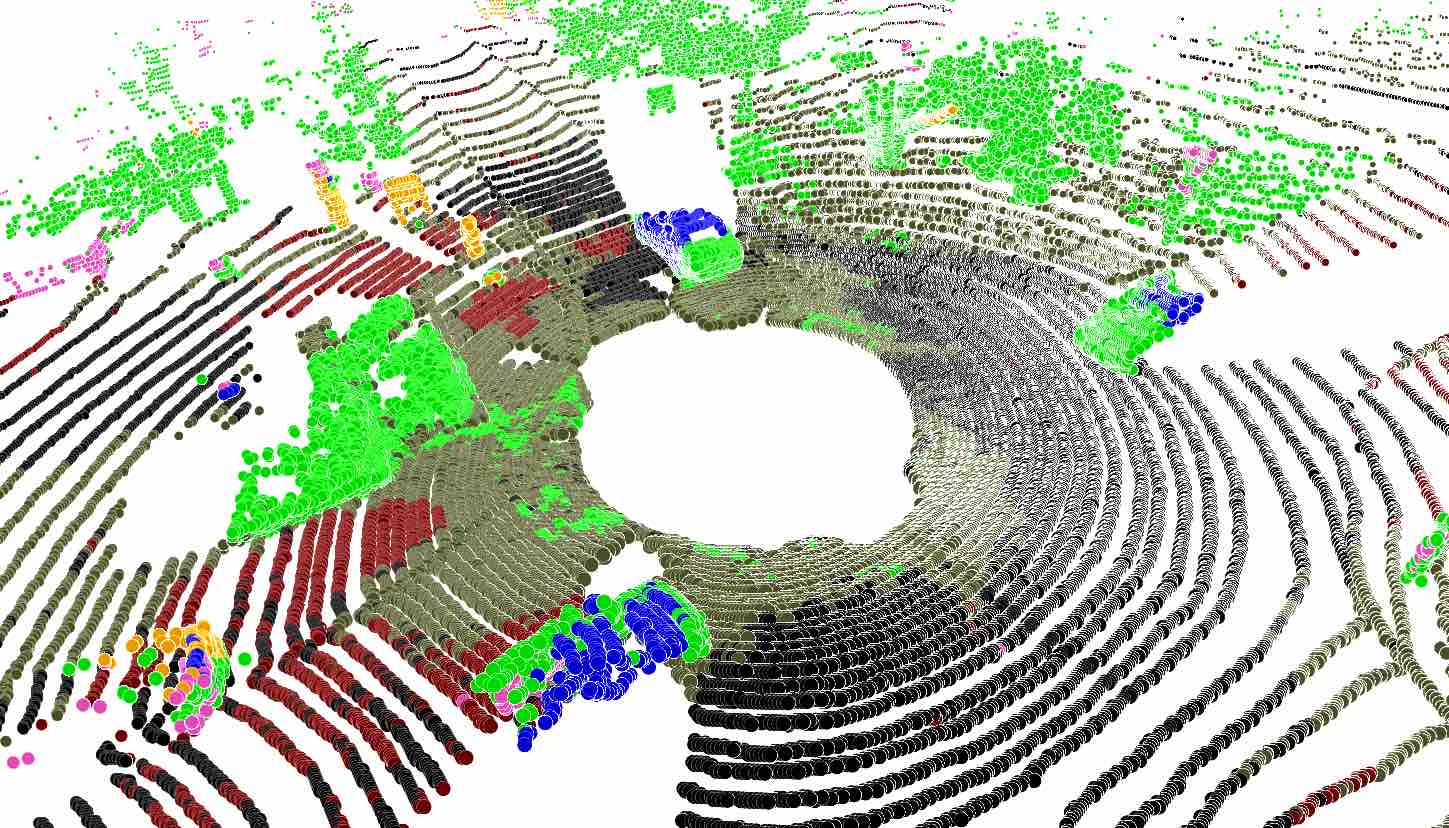}
        \end{overpic} &
        \begin{overpic}[width=0.21\textwidth]{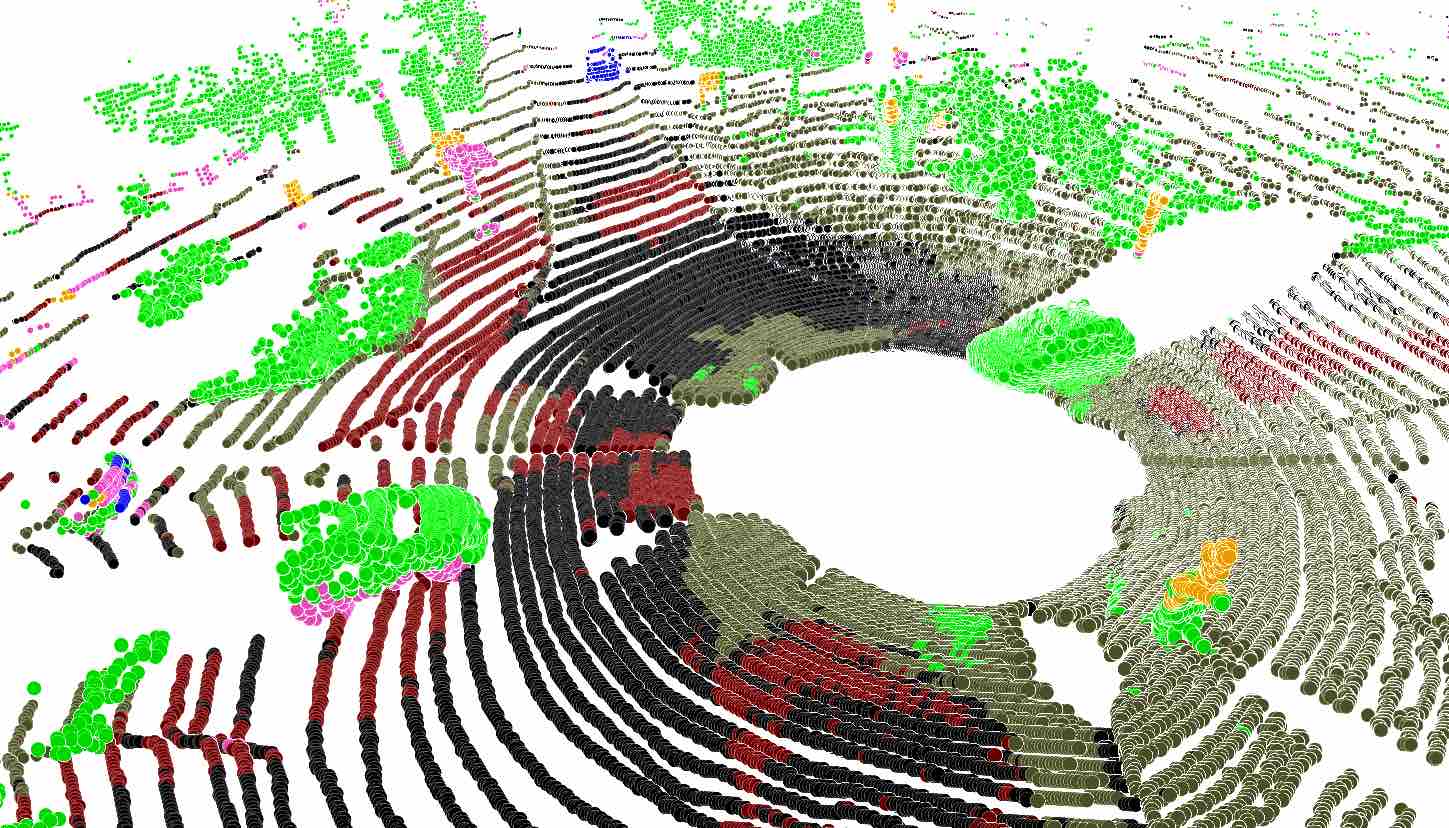}
        \end{overpic}& 
        \begin{overpic}[width=0.21\textwidth]{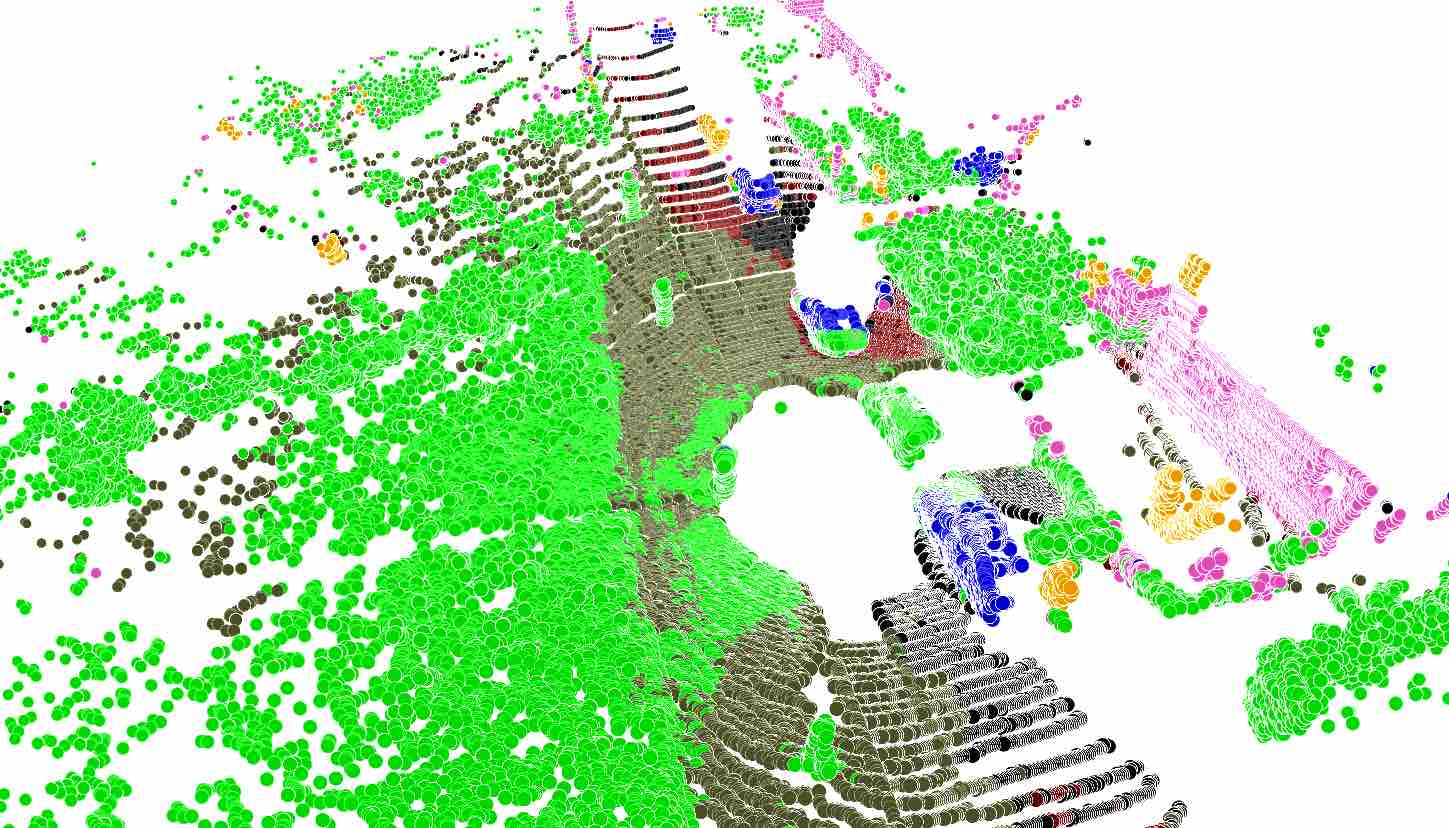}
        \end{overpic}\\
        \begin{overpic}[width=0.21\textwidth]{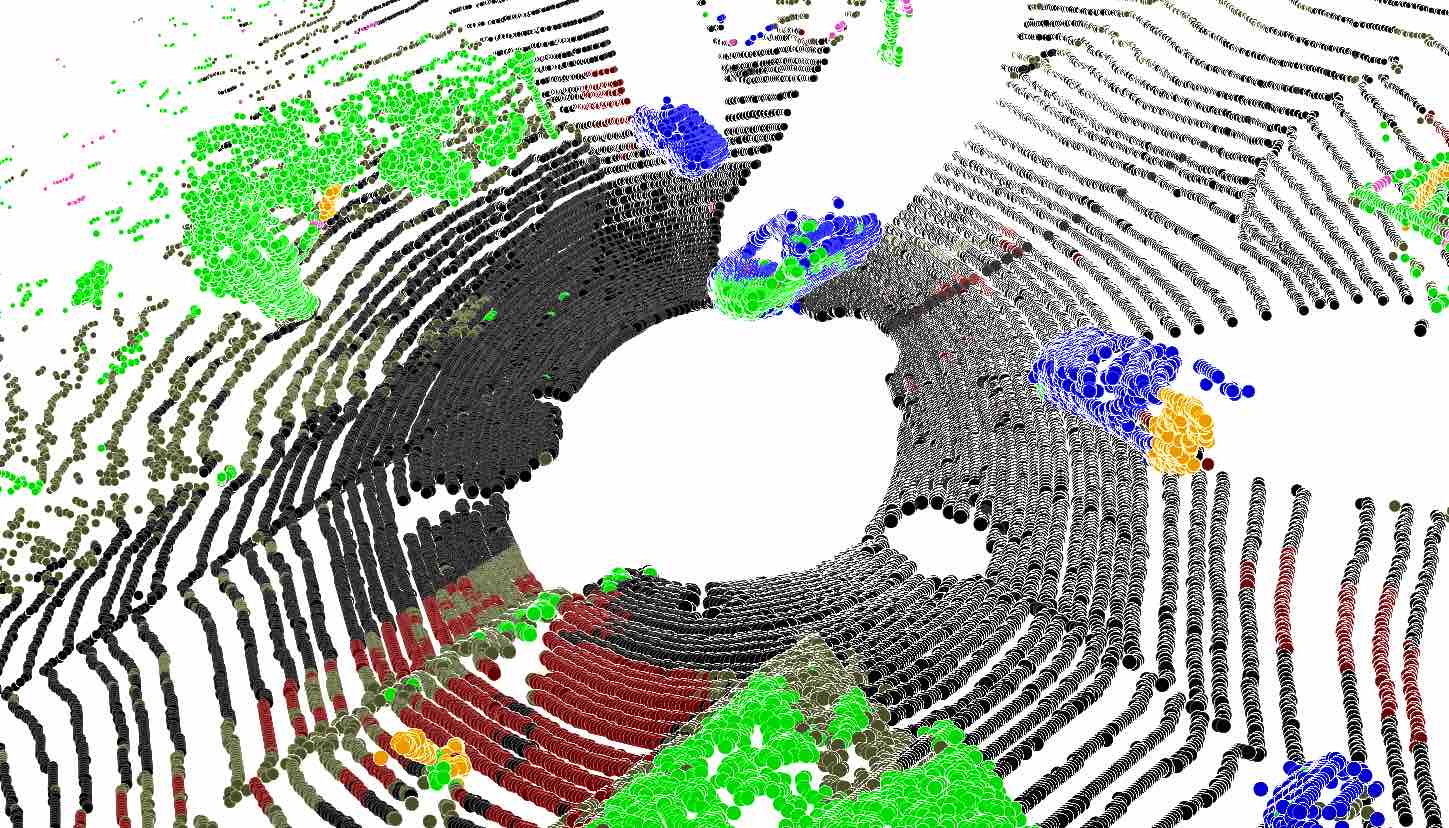}
        \end{overpic} &  
        \begin{overpic}[width=0.21\textwidth]{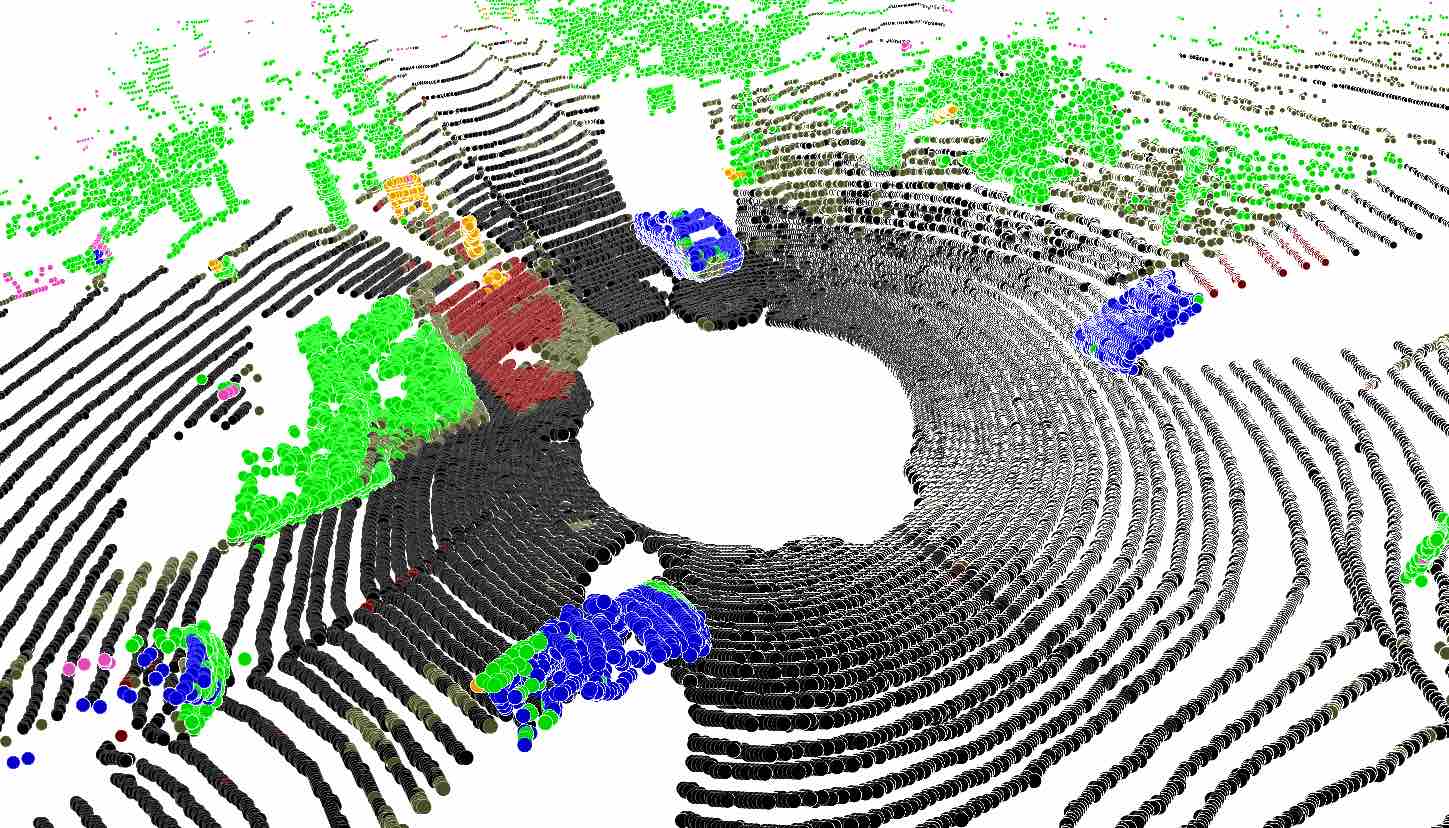}
        \end{overpic} &
        \begin{overpic}[width=0.21\textwidth]{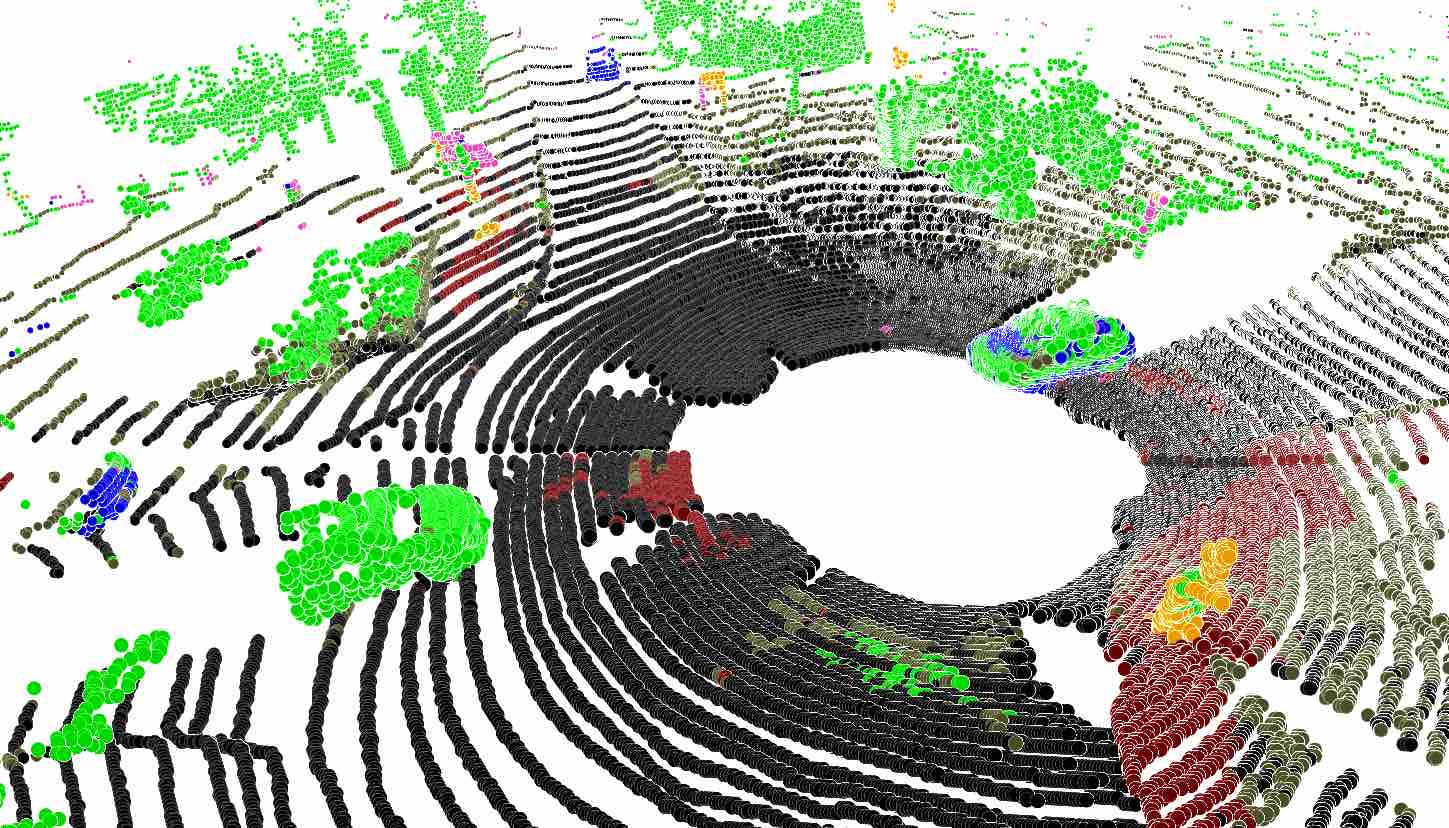}
        \end{overpic}& 
        \begin{overpic}[width=0.21\textwidth]{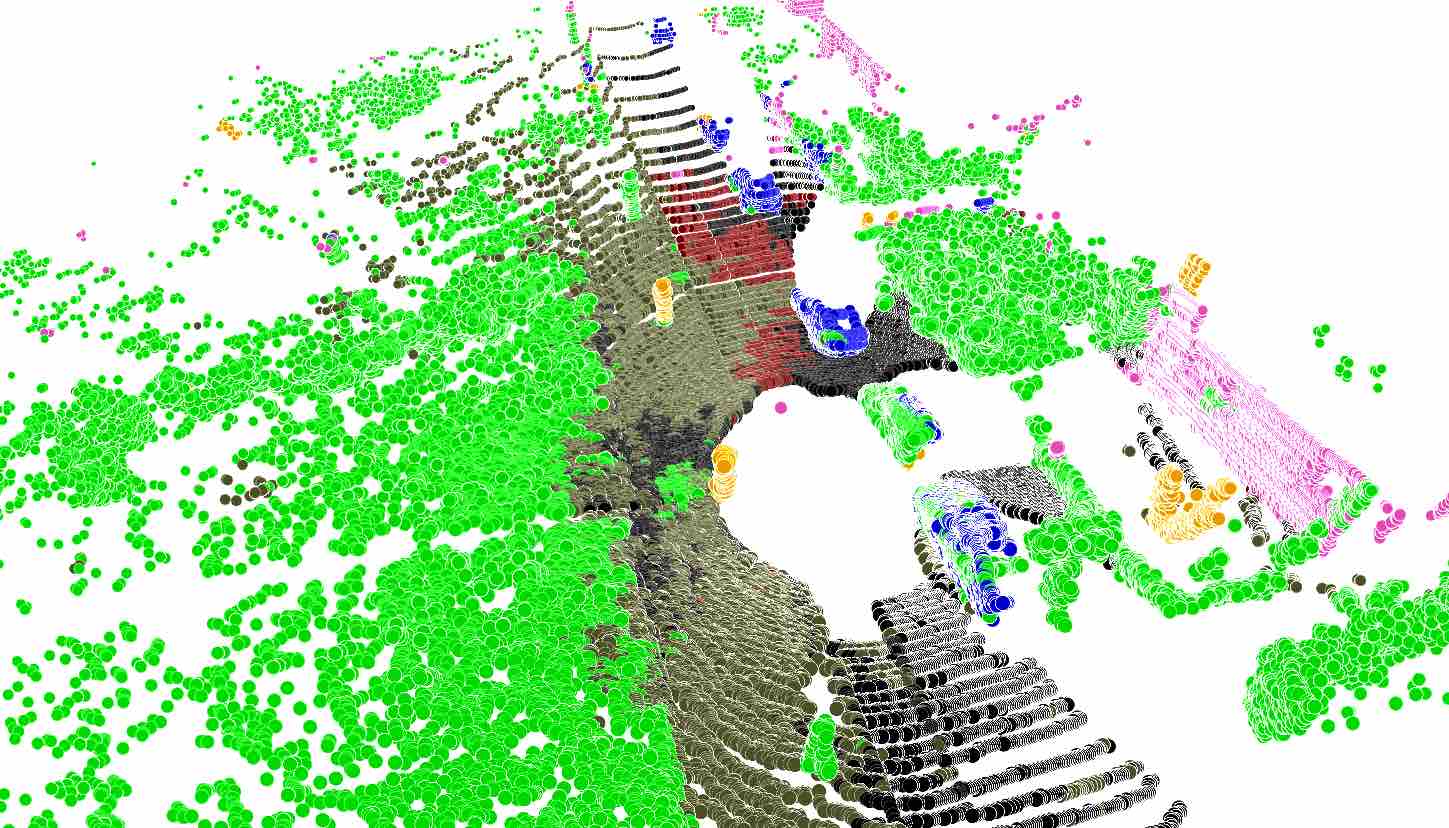}
        \end{overpic}\\
        \begin{overpic}[width=0.21\textwidth]{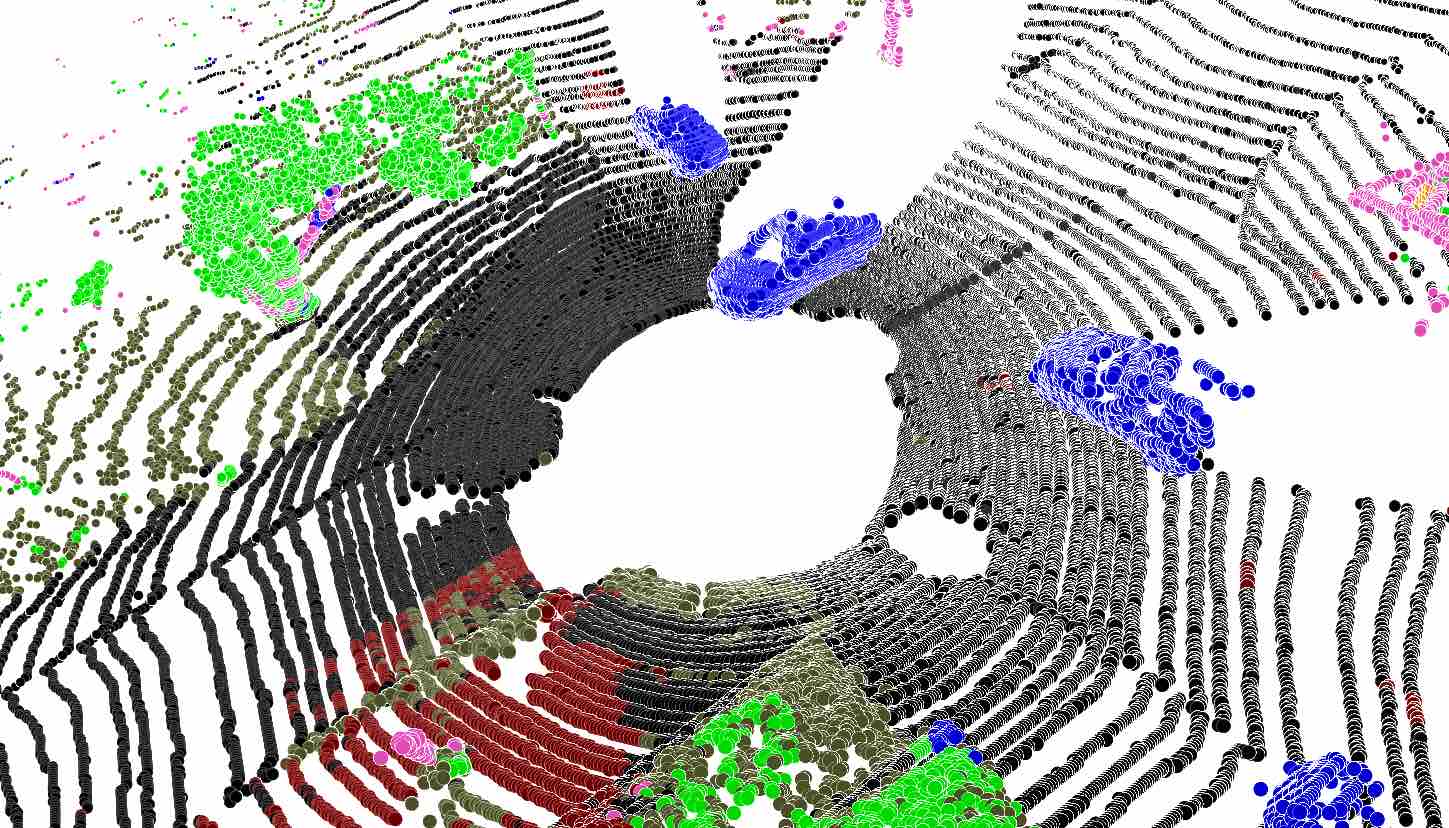}
        \end{overpic} &  
        \begin{overpic}[width=0.21\textwidth]{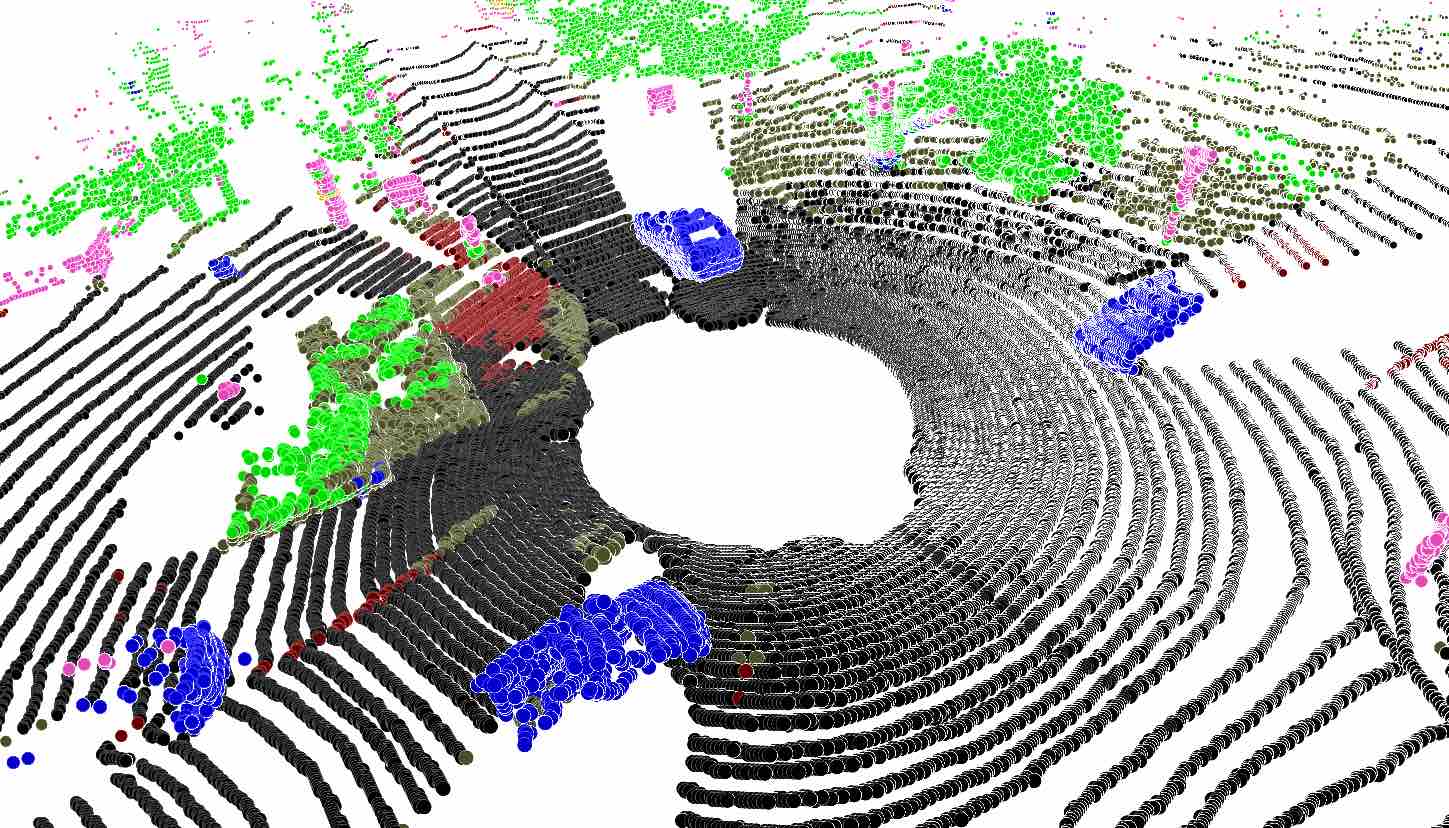}
        \end{overpic} &
        \begin{overpic}[width=0.21\textwidth]{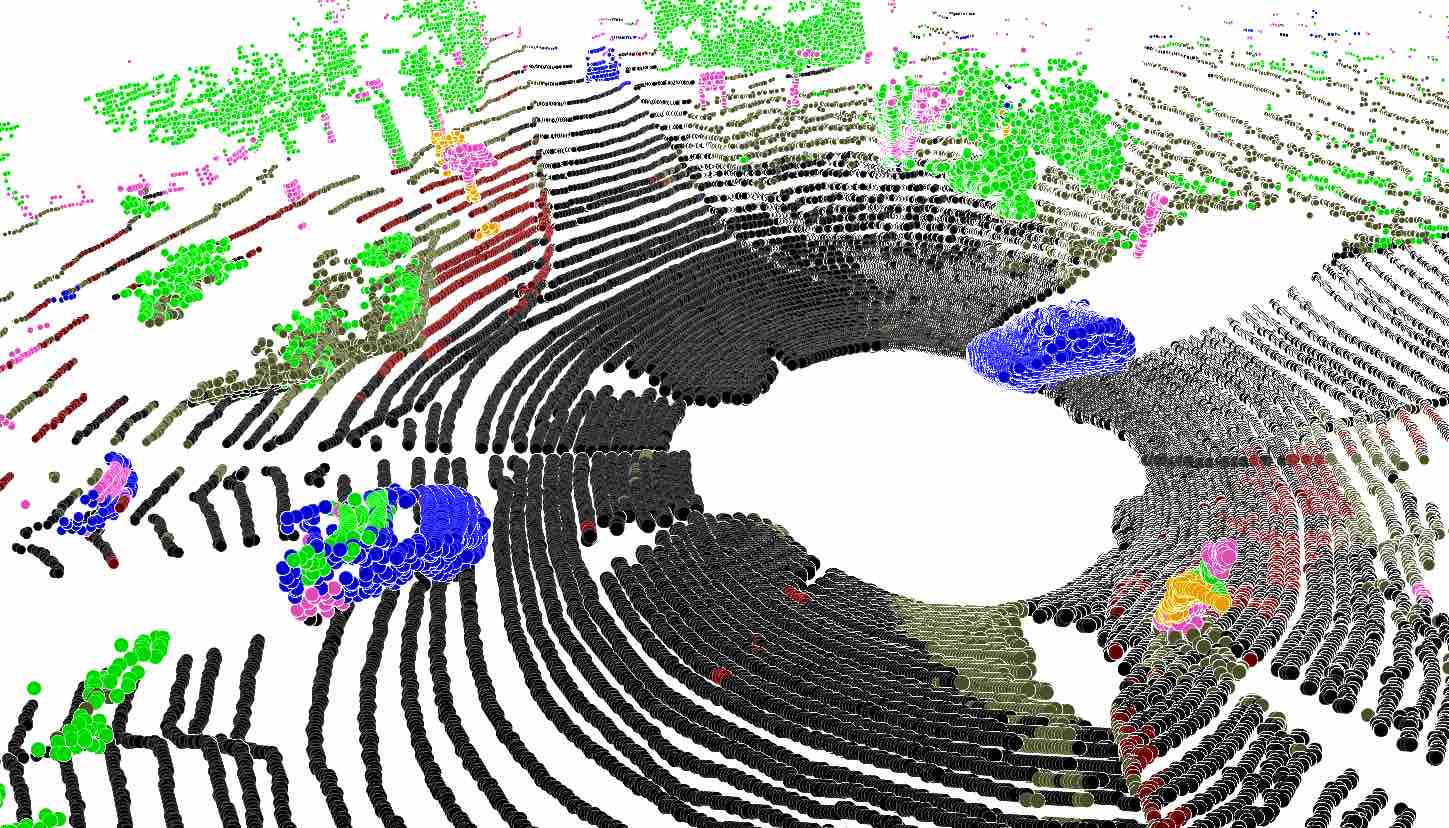}
        \end{overpic}& 
        \begin{overpic}[width=0.21\textwidth]{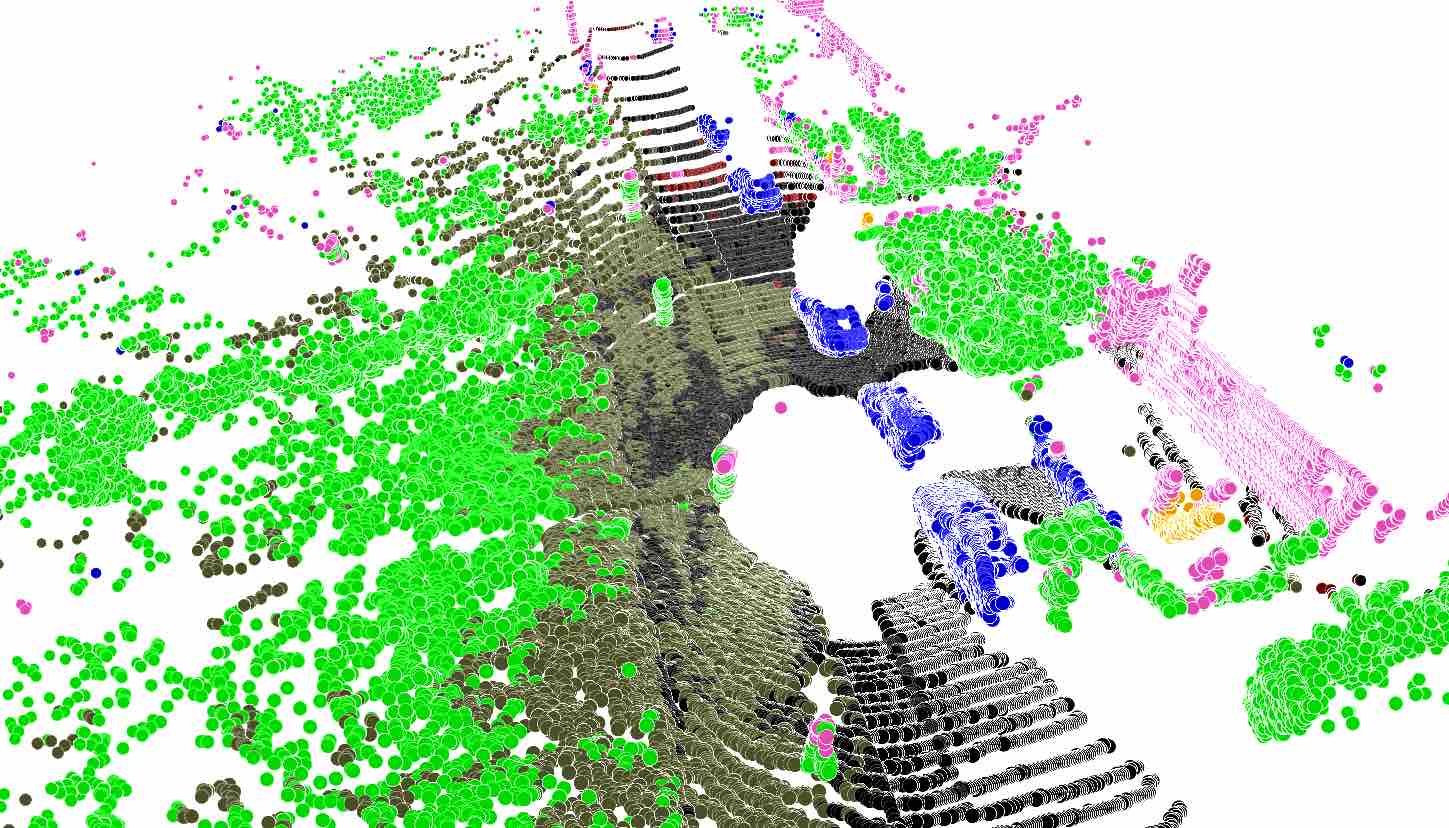}
        \end{overpic}\\
        \begin{overpic}[width=0.21\textwidth]{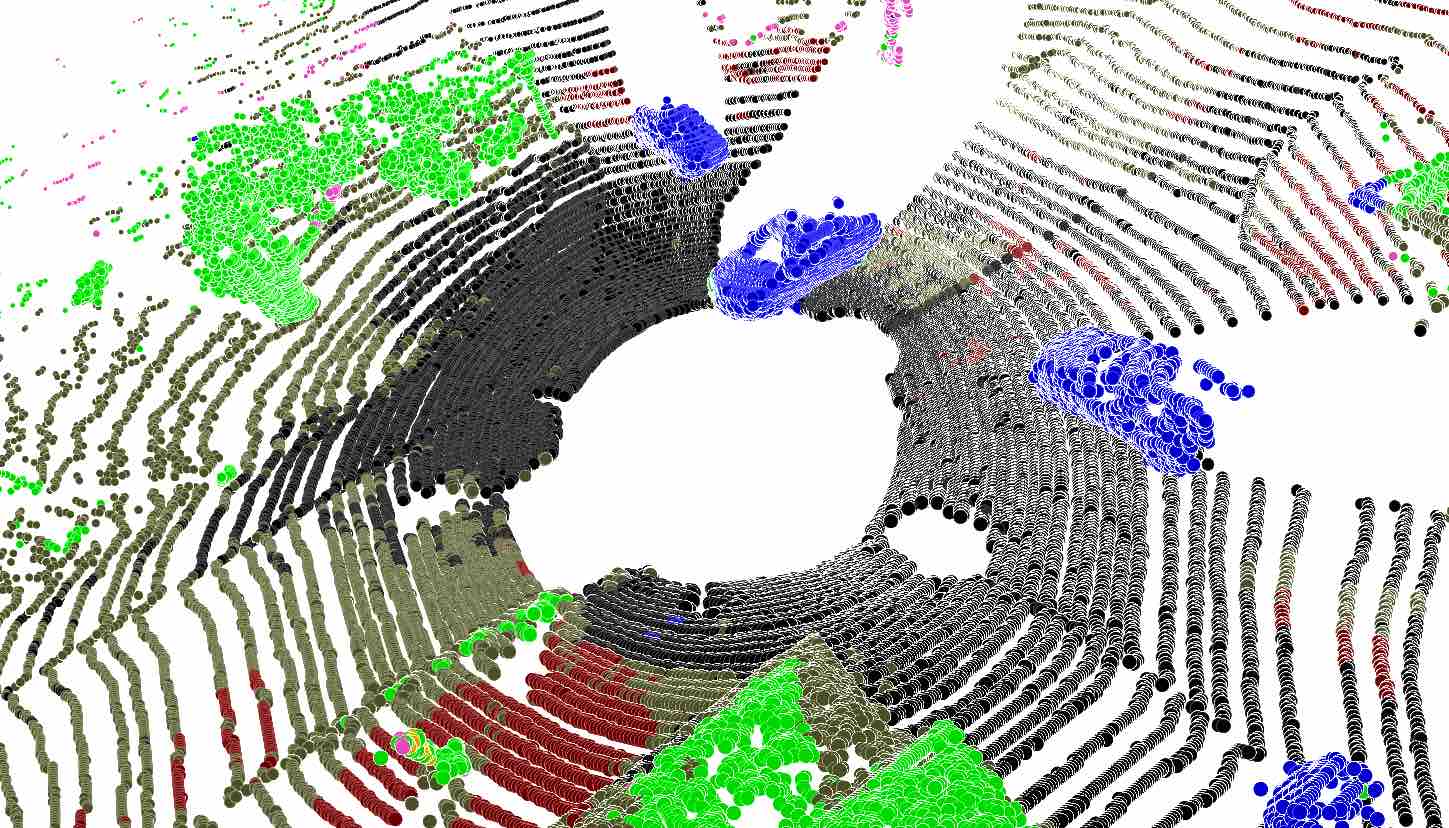}
        \end{overpic} &  
        \begin{overpic}[width=0.21\textwidth]{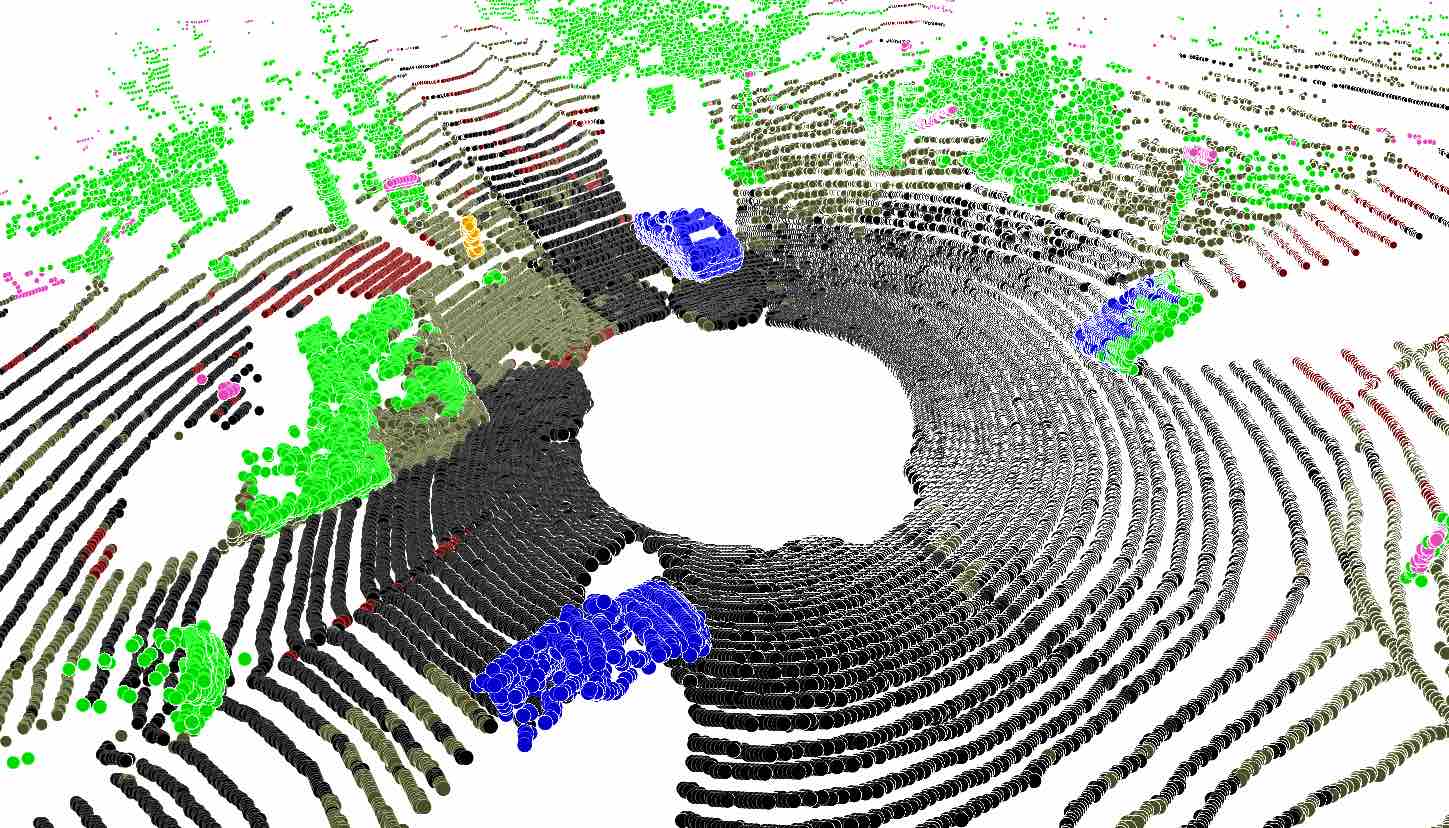}
        \end{overpic} &
        \begin{overpic}[width=0.21\textwidth]{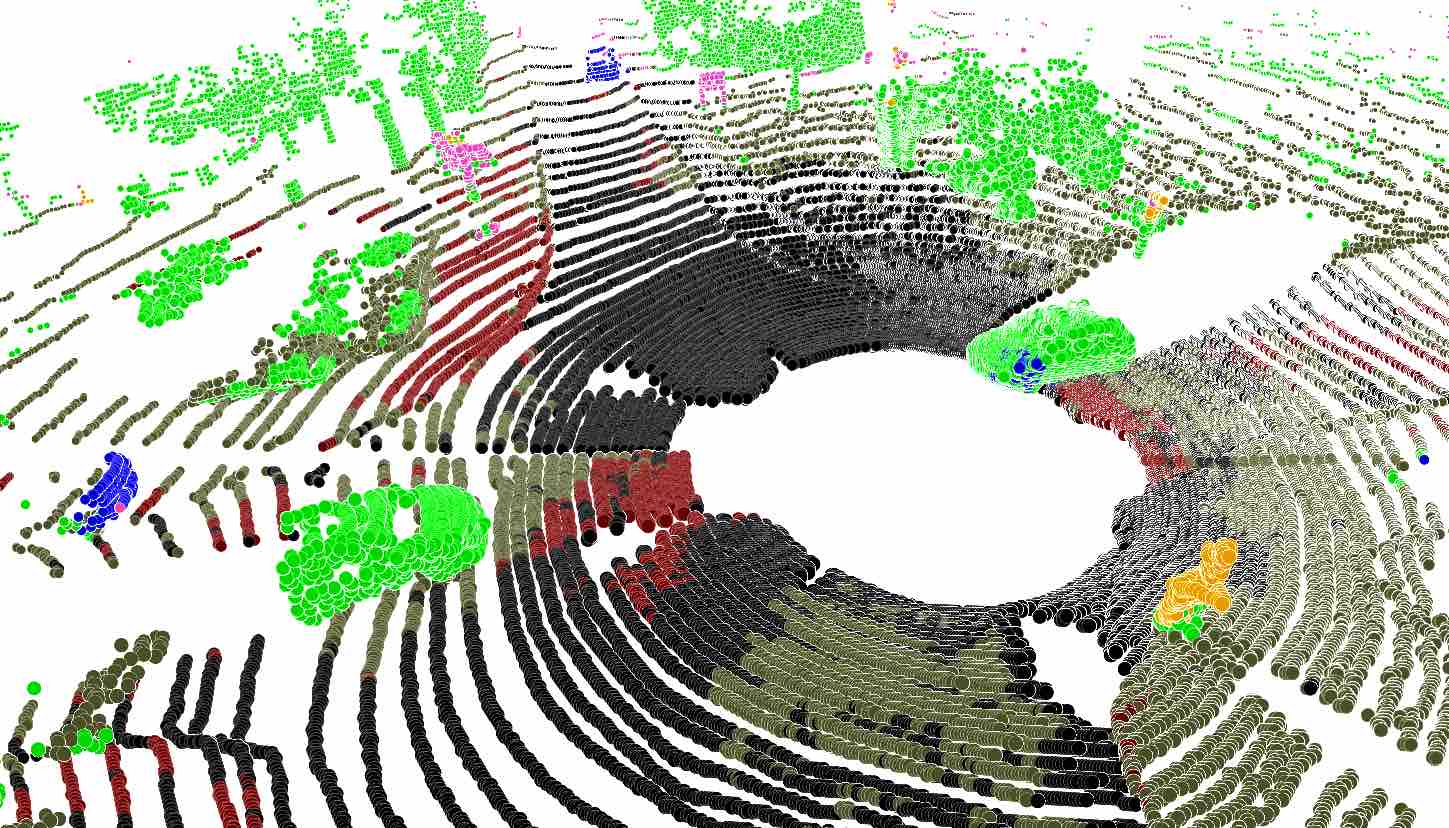}
        \end{overpic}& 
        \begin{overpic}[width=0.21\textwidth]{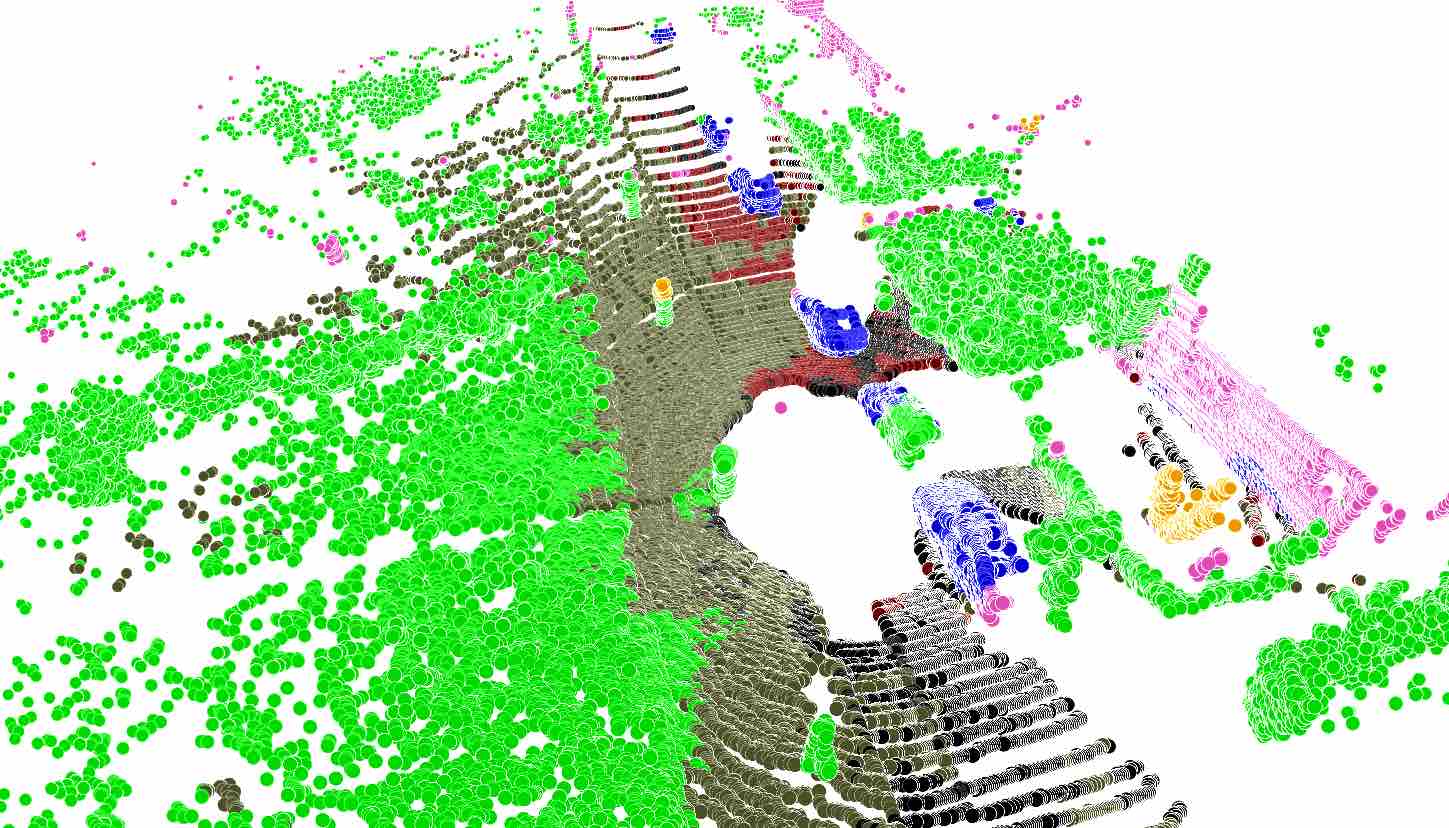}
        \end{overpic}\\
        \begin{overpic}[width=0.21\textwidth]{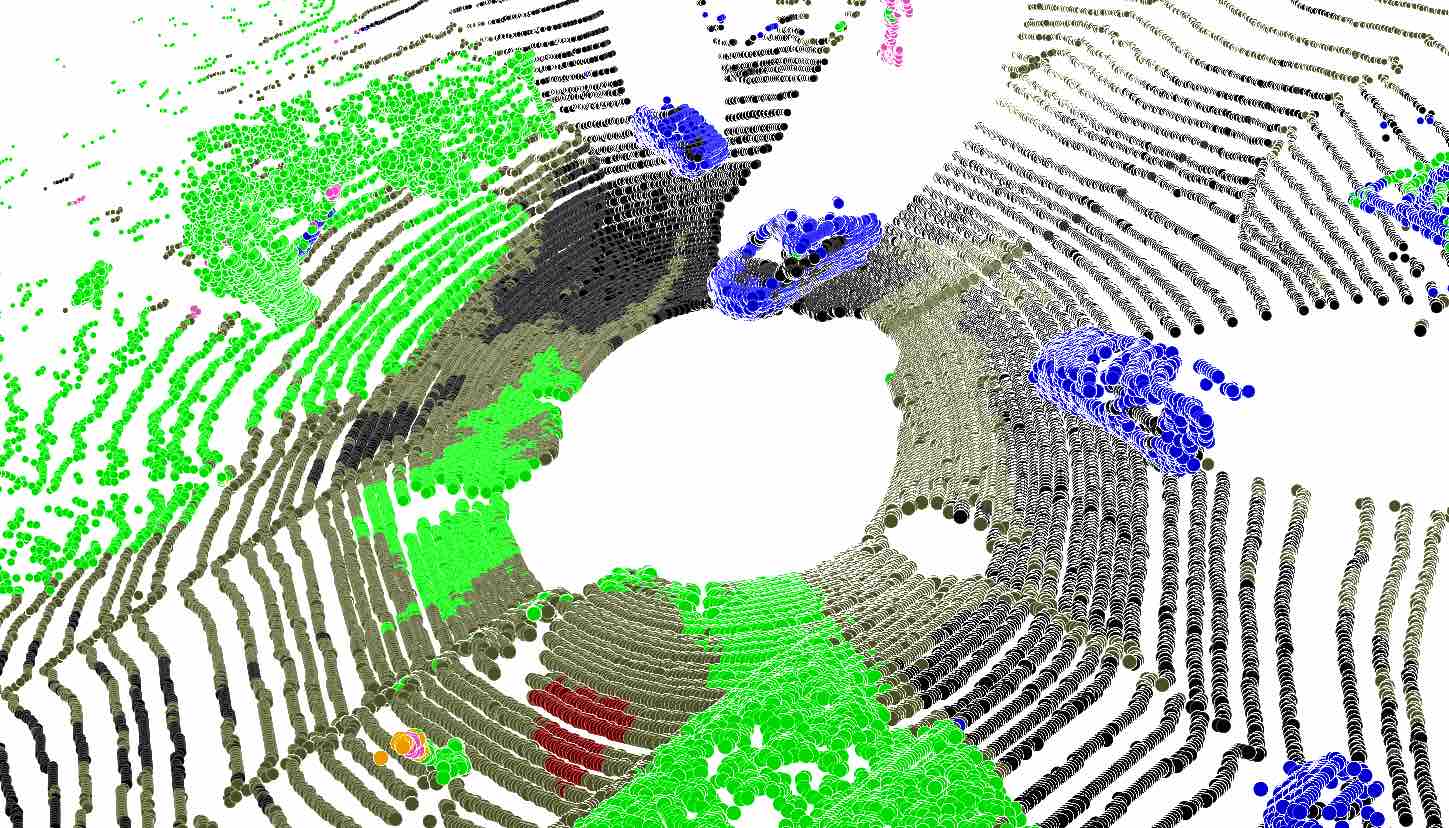}
        \end{overpic} &  
        \begin{overpic}[width=0.21\textwidth]{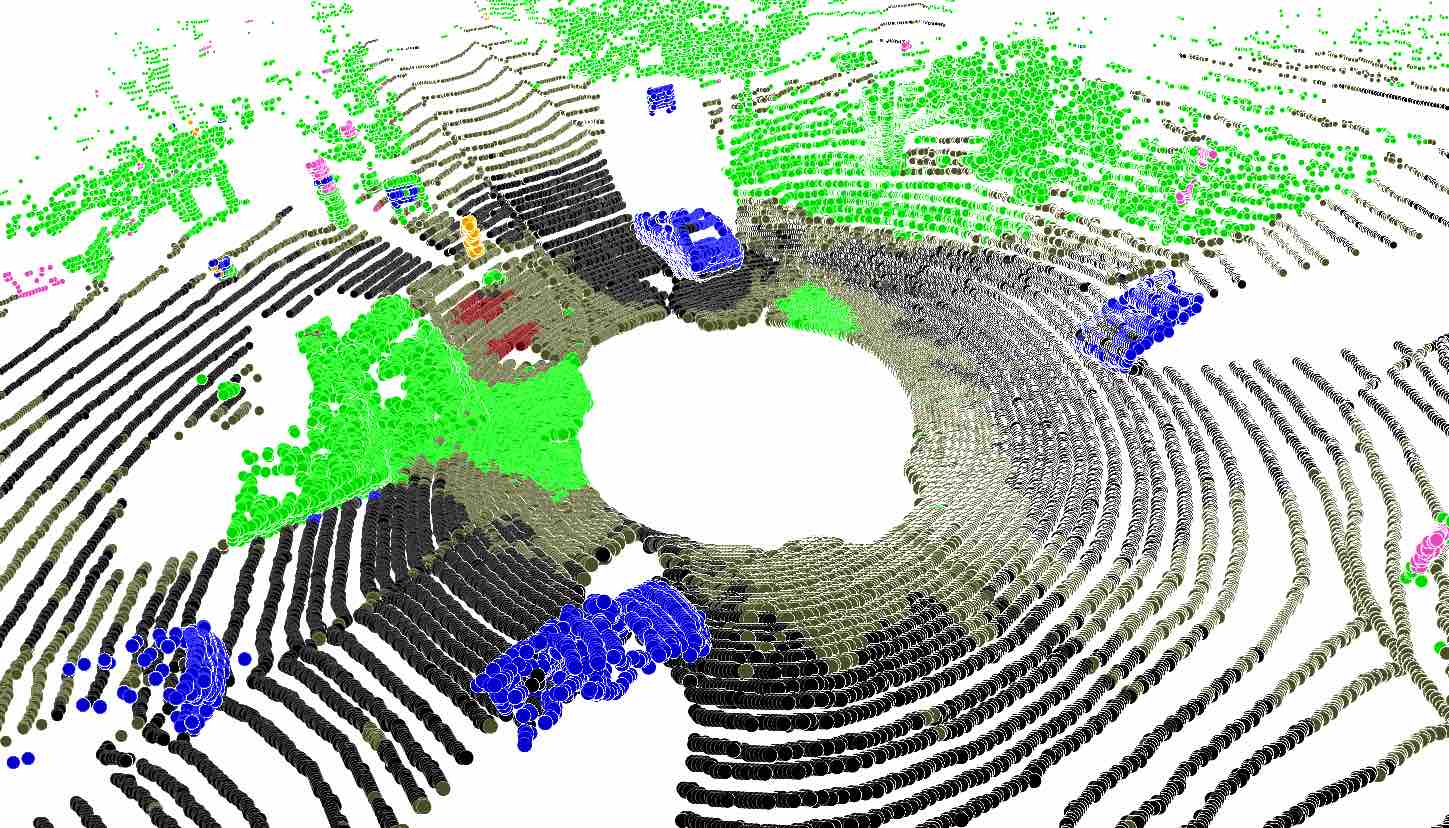}
        \end{overpic} &
        \begin{overpic}[width=0.21\textwidth]{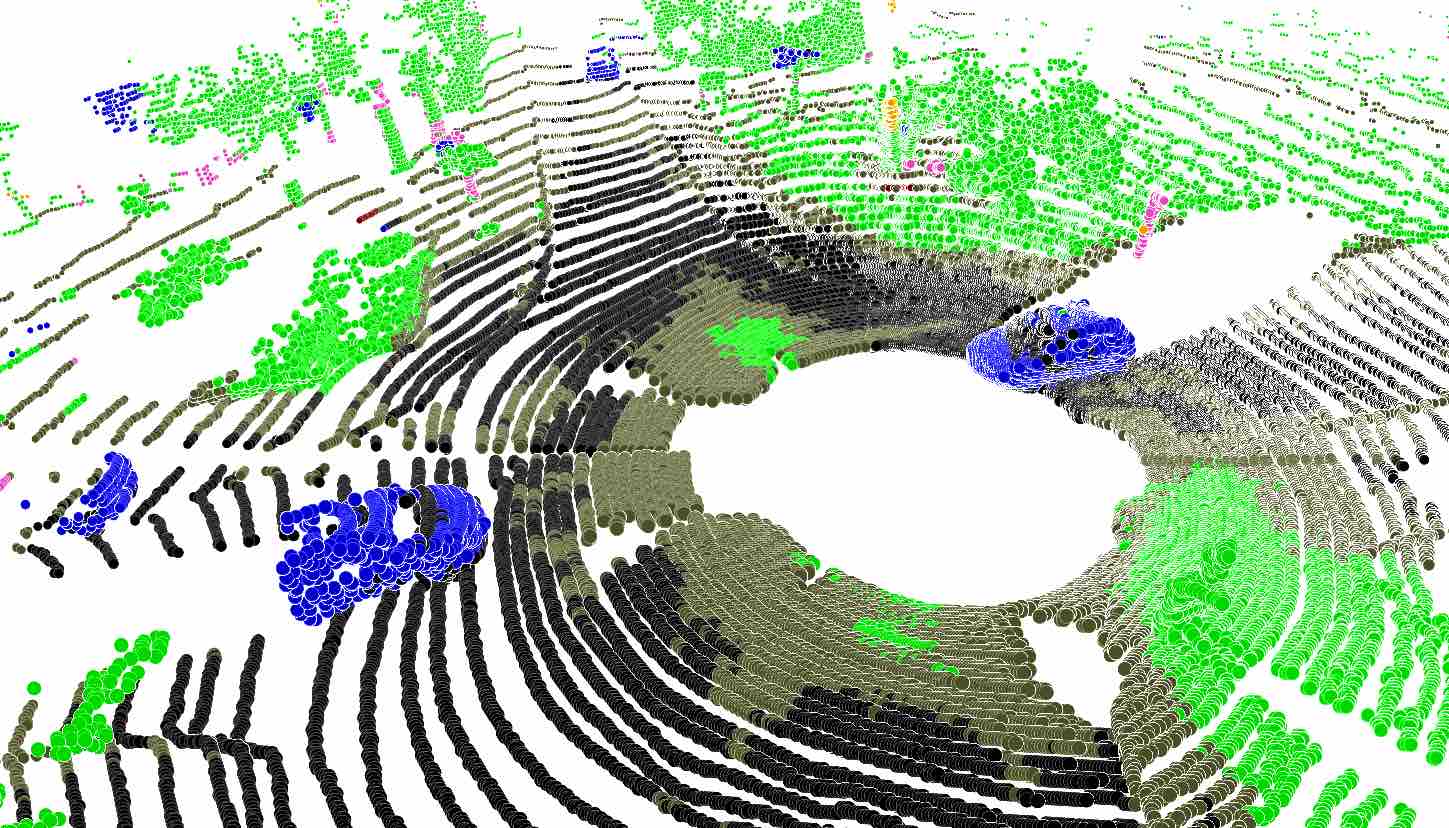}
        \end{overpic}& 
        \begin{overpic}[width=0.21\textwidth]{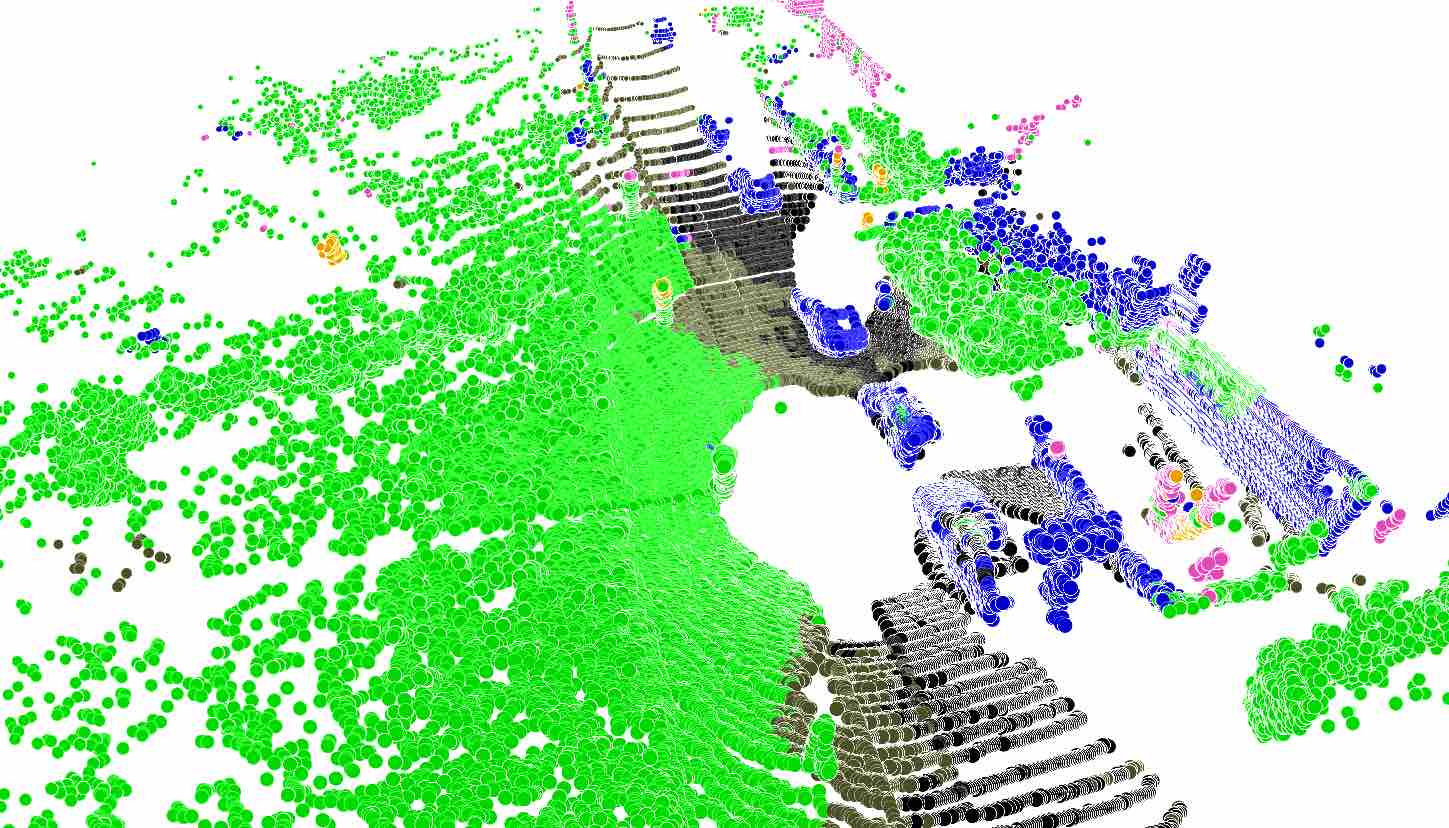}
        \end{overpic}\\
        \begin{overpic}[width=0.21\textwidth]{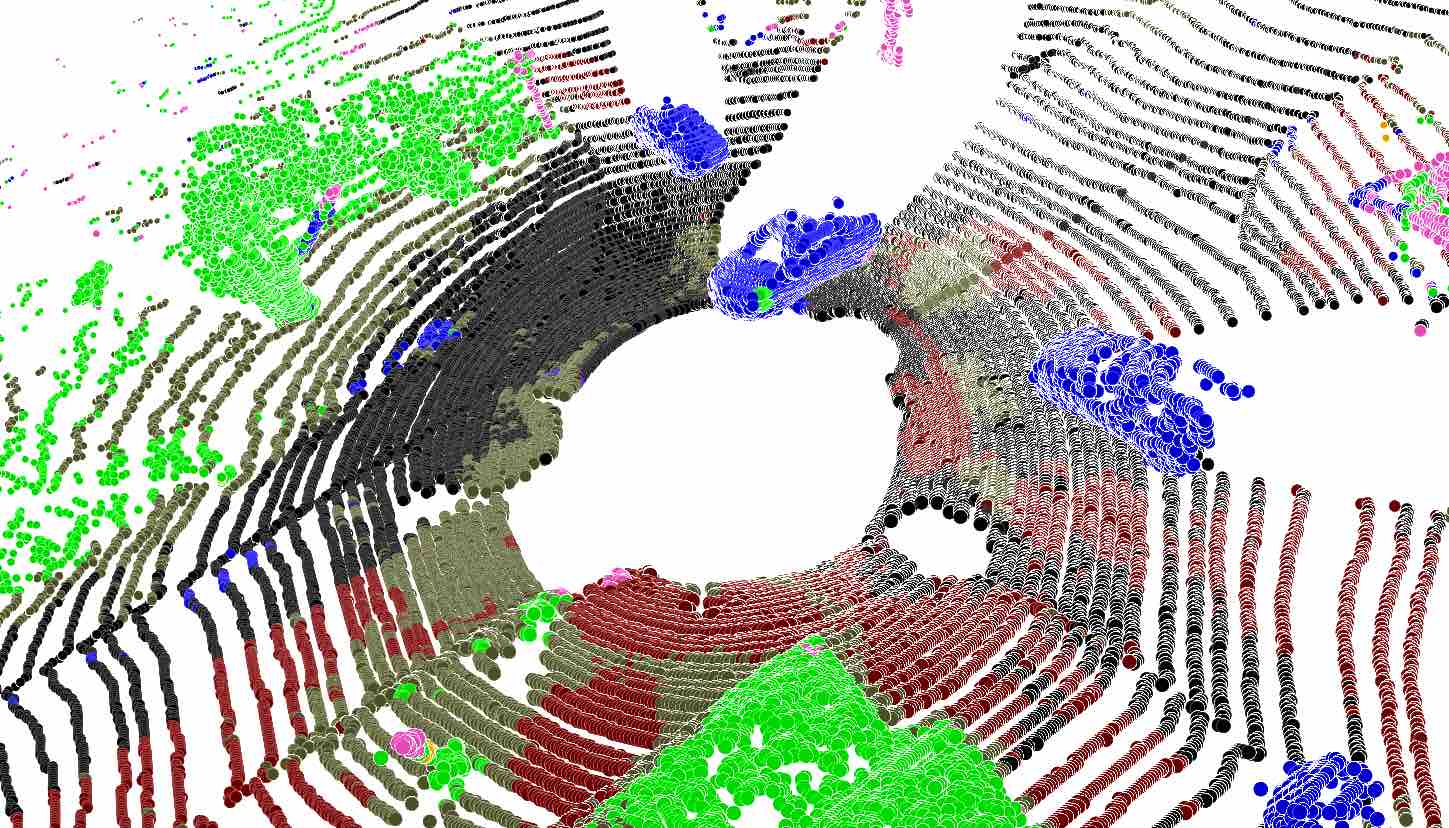}
        \end{overpic} &  
        \begin{overpic}[width=0.21\textwidth]{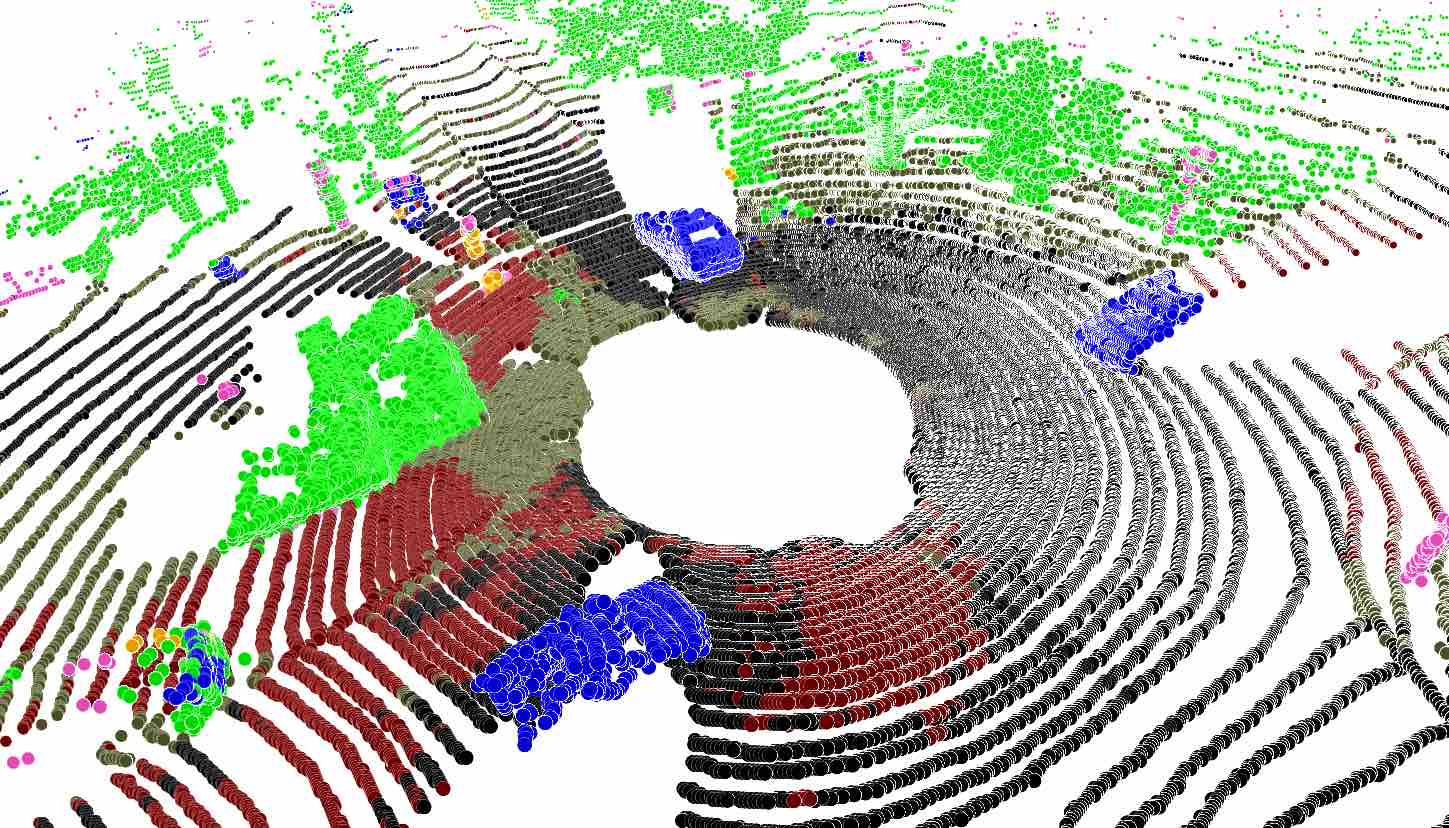}
        \end{overpic} &
        \begin{overpic}[width=0.21\textwidth]{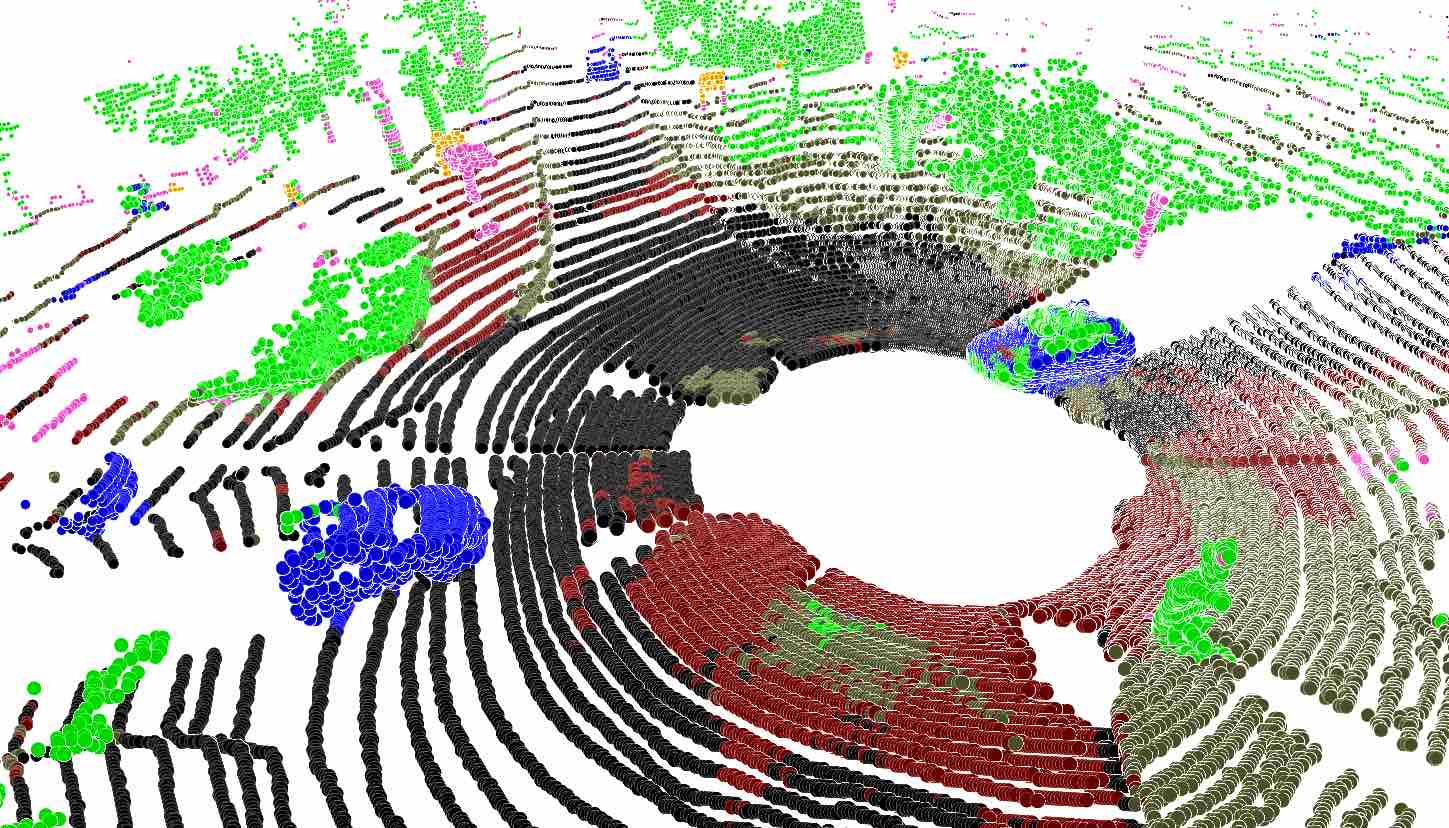}
        \end{overpic}& 
        \begin{overpic}[width=0.21\textwidth]{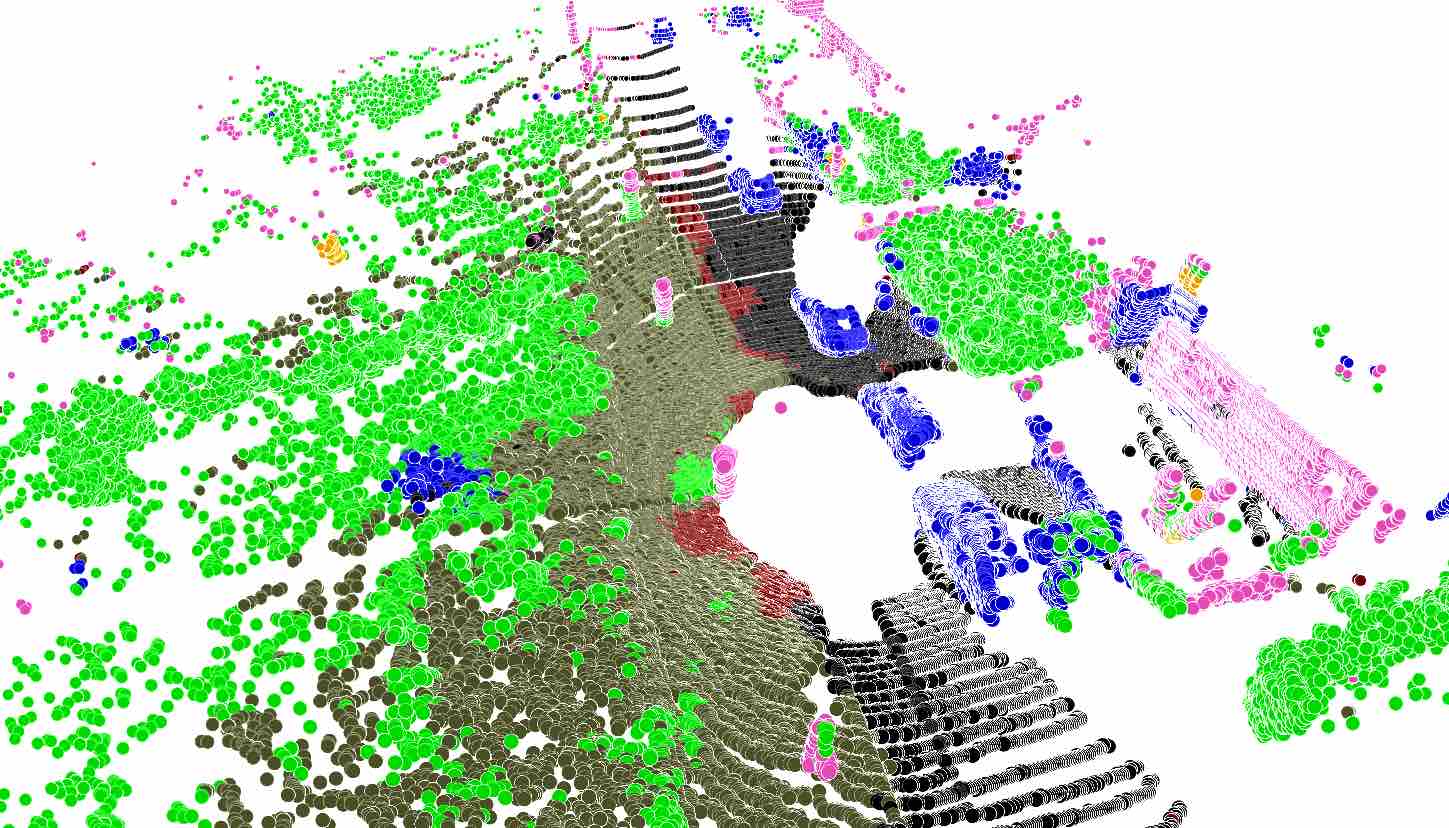}
        \end{overpic}\\
        \begin{overpic}[width=0.21\textwidth]{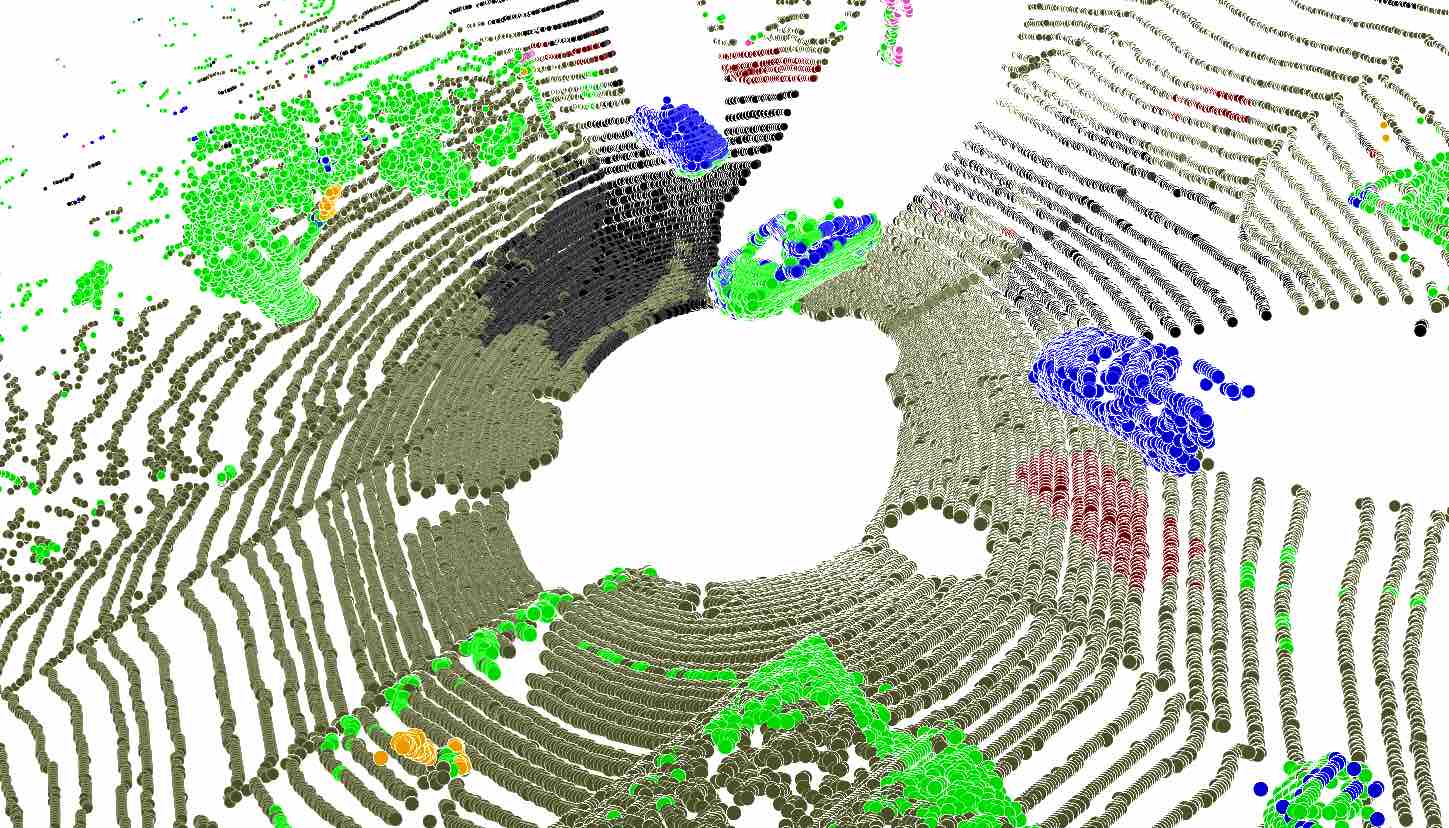}
        \end{overpic} &  
        \begin{overpic}[width=0.21\textwidth]{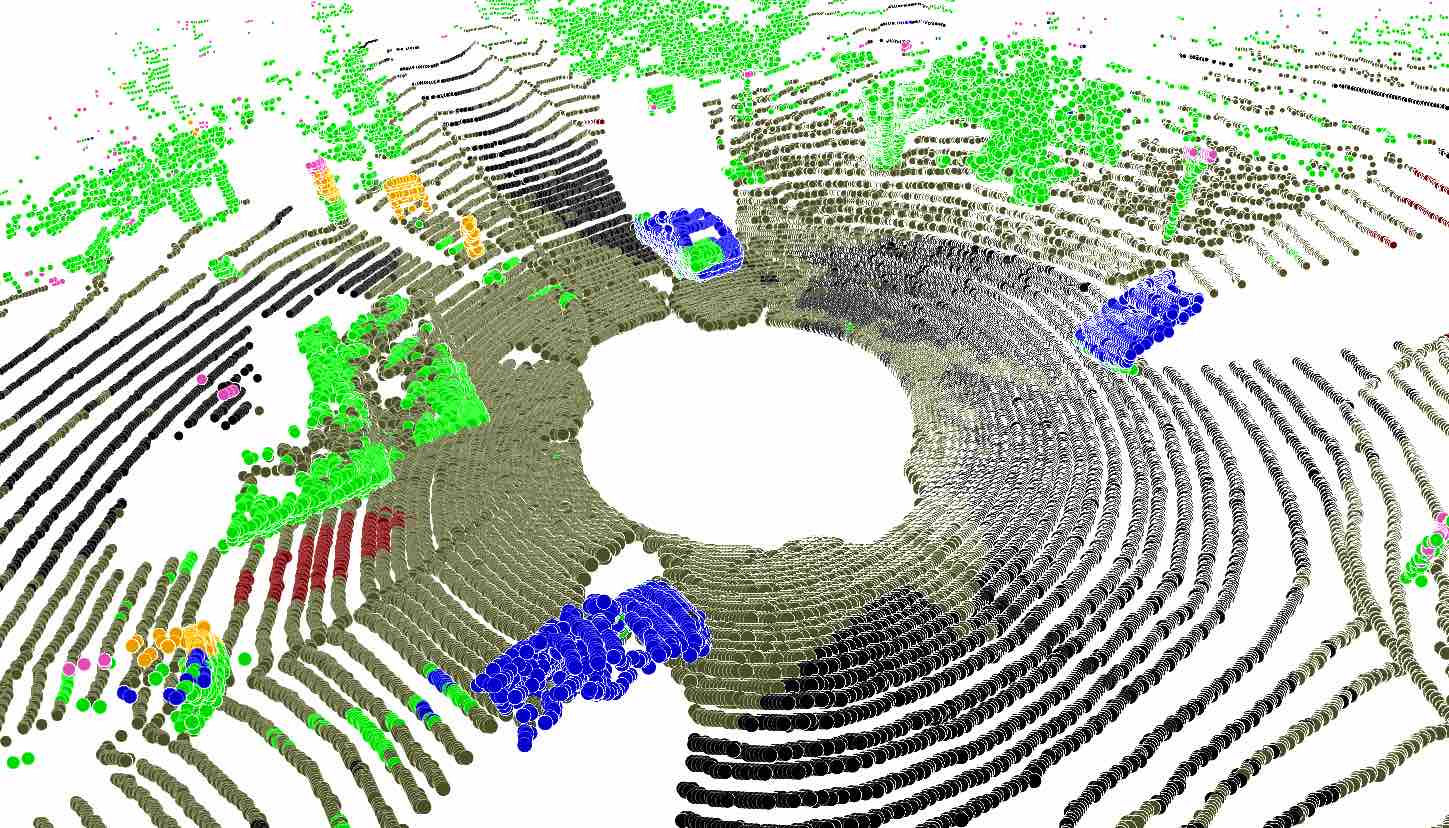}
        \end{overpic} &
        \begin{overpic}[width=0.21\textwidth]{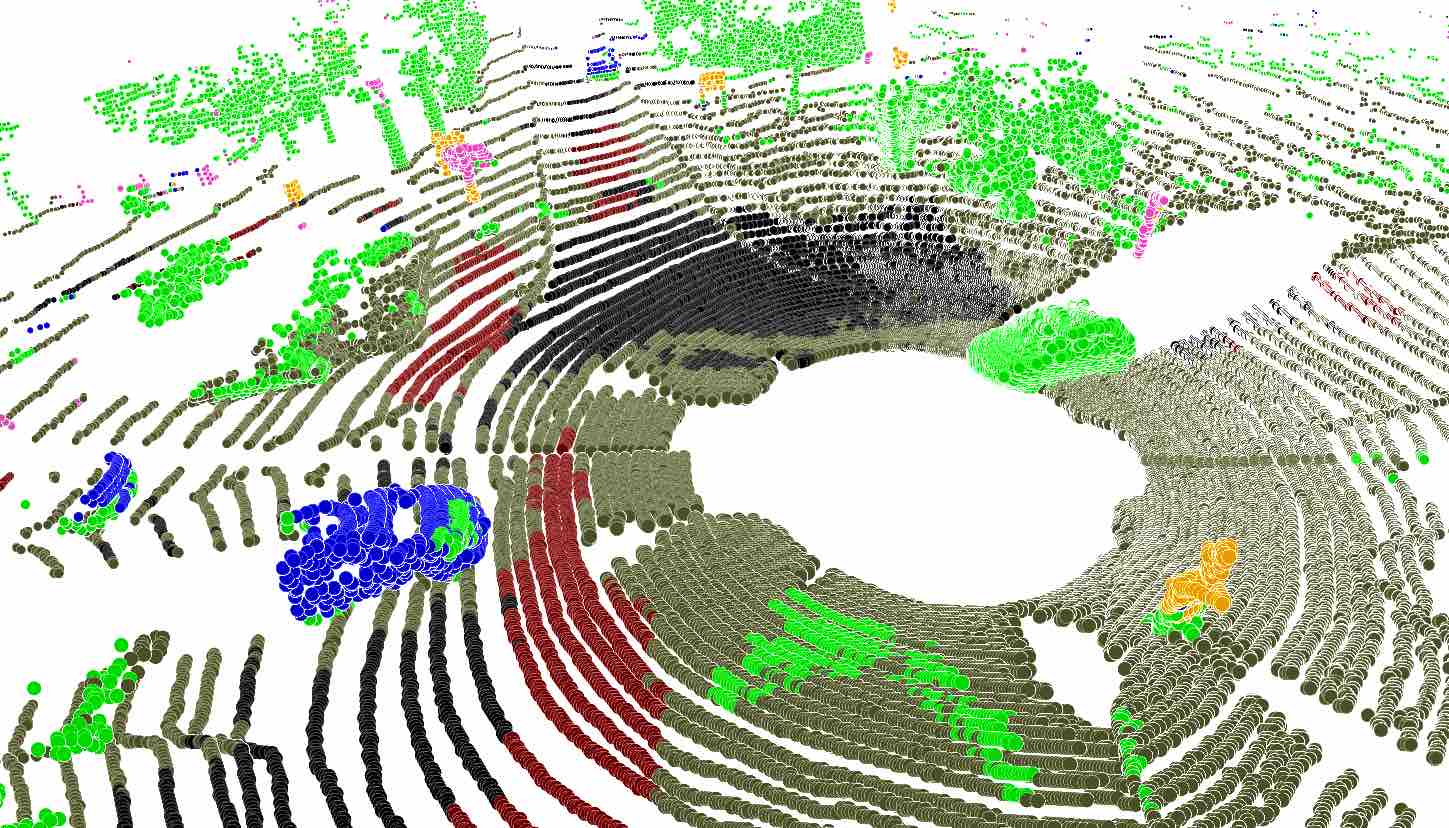}
        \end{overpic}& 
        \begin{overpic}[width=0.21\textwidth]{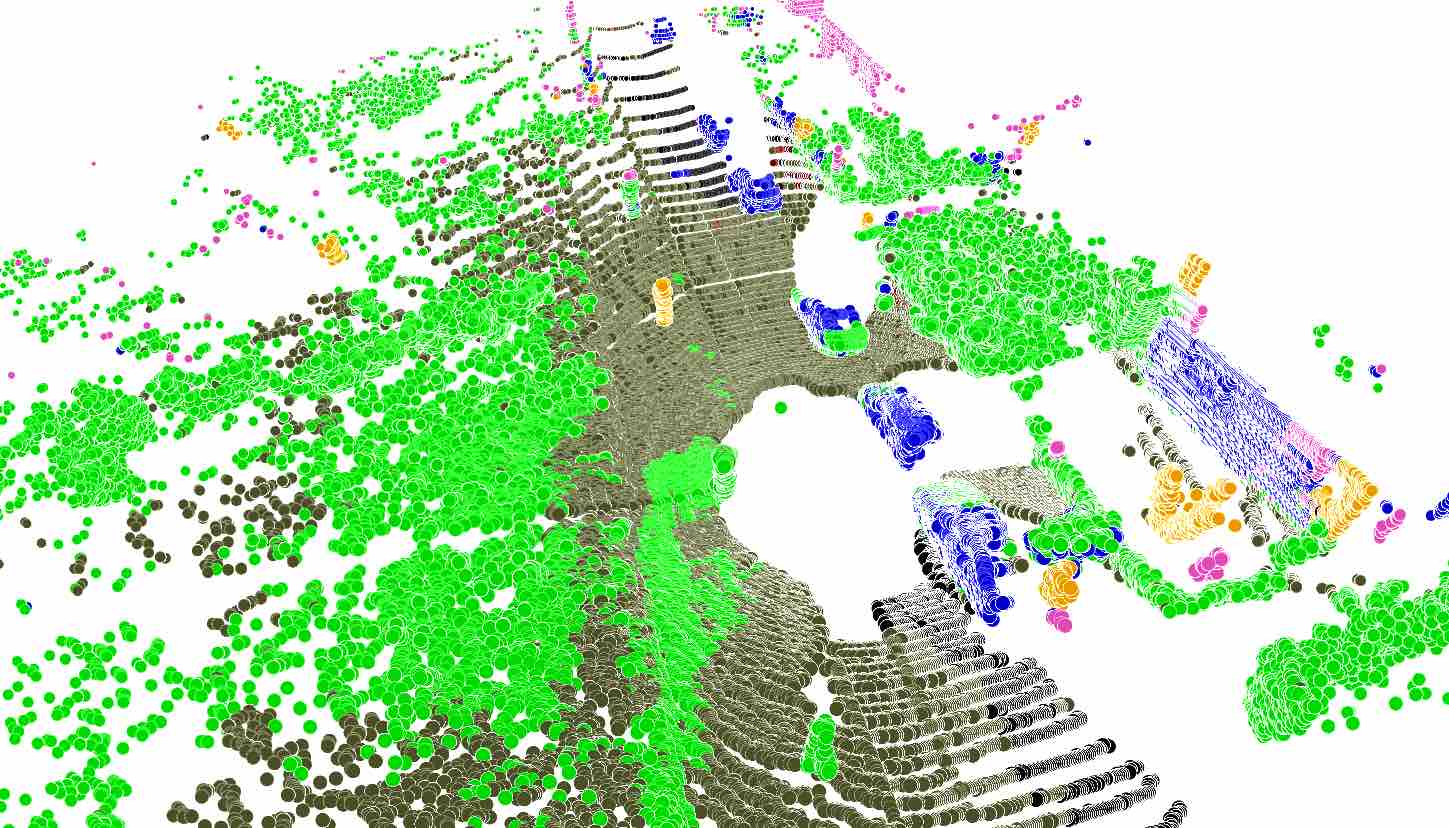}
        \end{overpic}\\
        \begin{overpic}[width=0.21\textwidth]{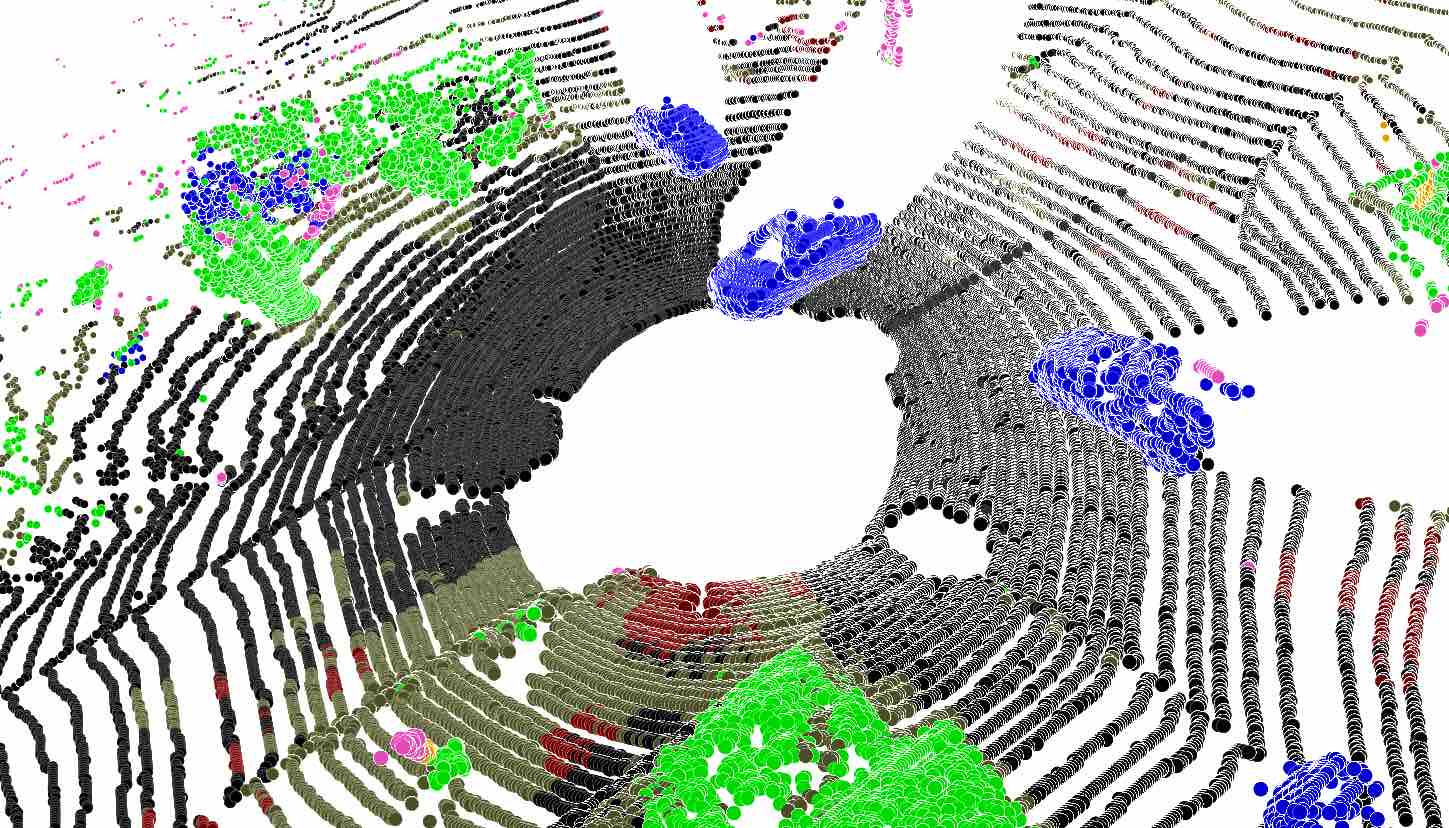}
        \end{overpic} &  
        \begin{overpic}[width=0.21\textwidth]{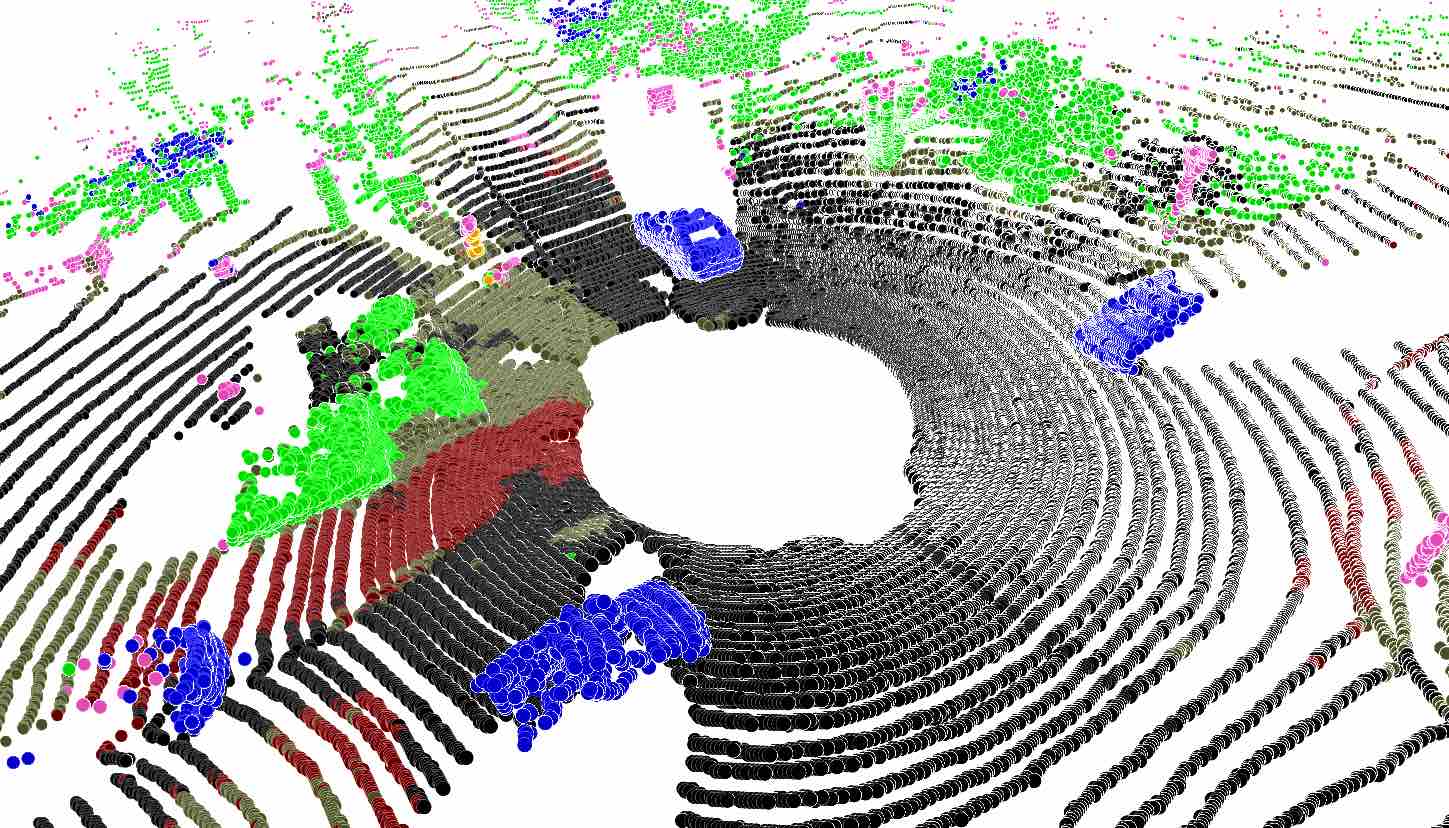}
        \end{overpic} &
        \begin{overpic}[width=0.21\textwidth]{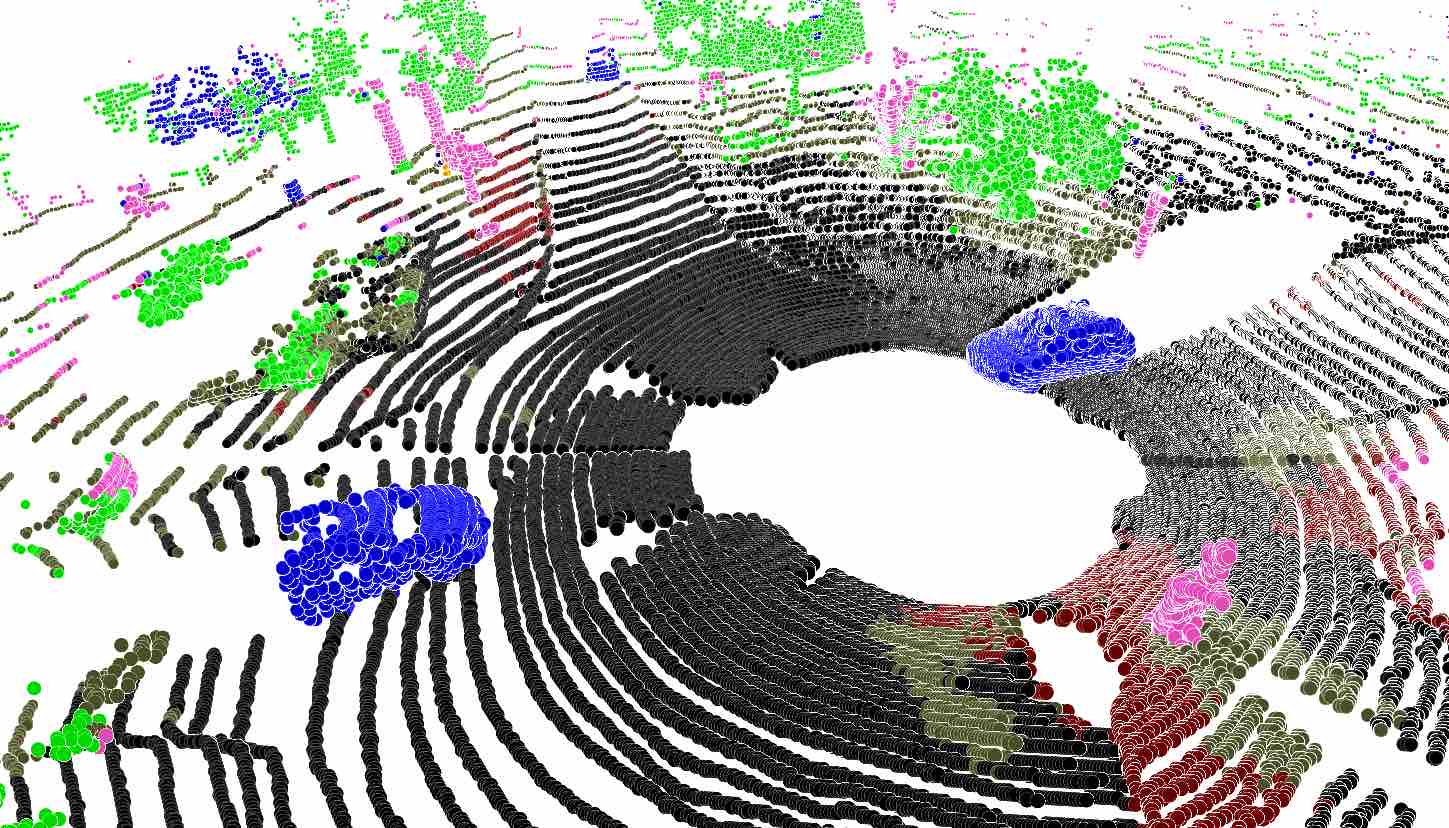}
        \end{overpic}& 
        \begin{overpic}[width=0.21\textwidth]{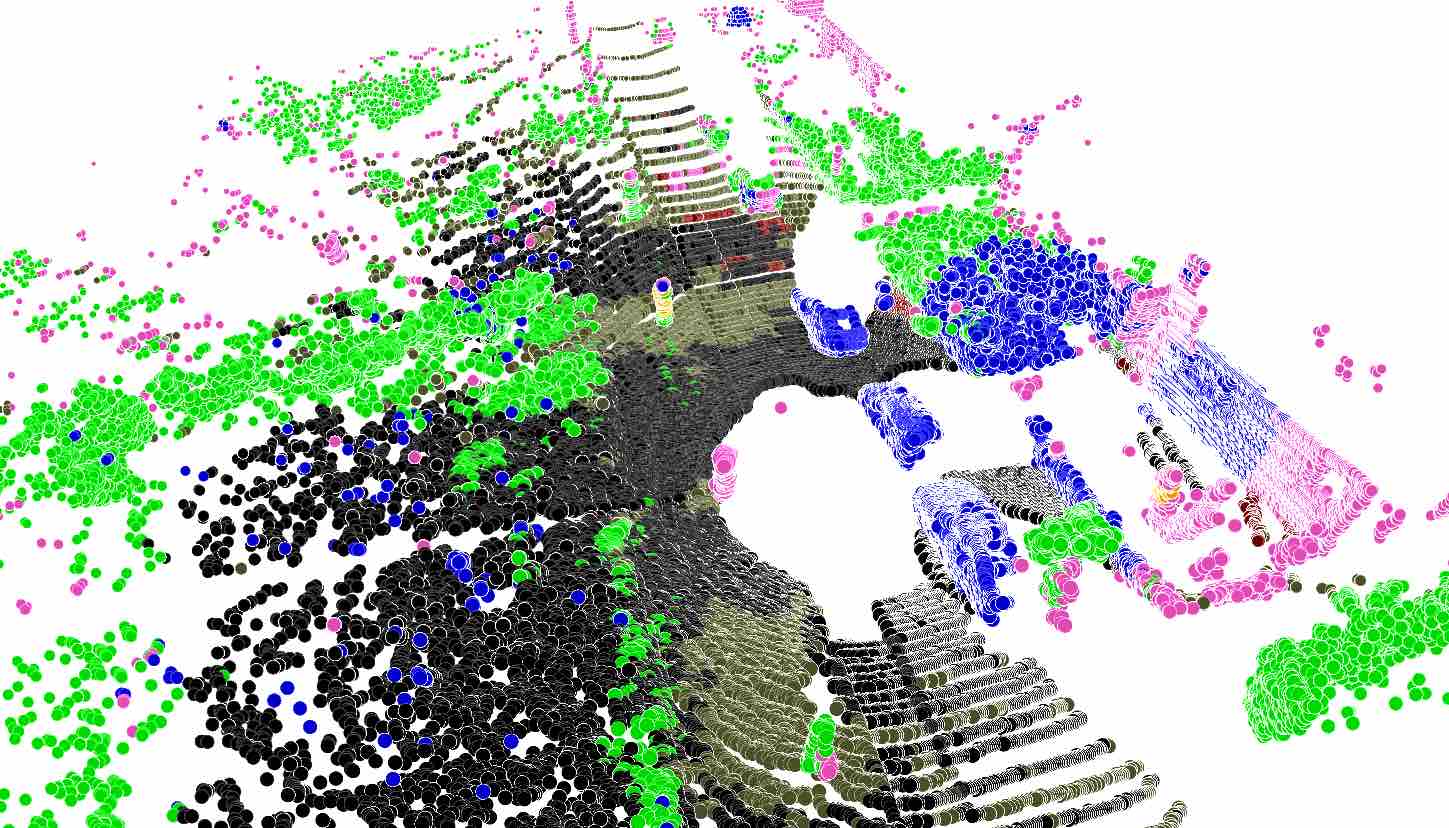}
        \end{overpic}\\
        \begin{overpic}[width=0.21\textwidth]{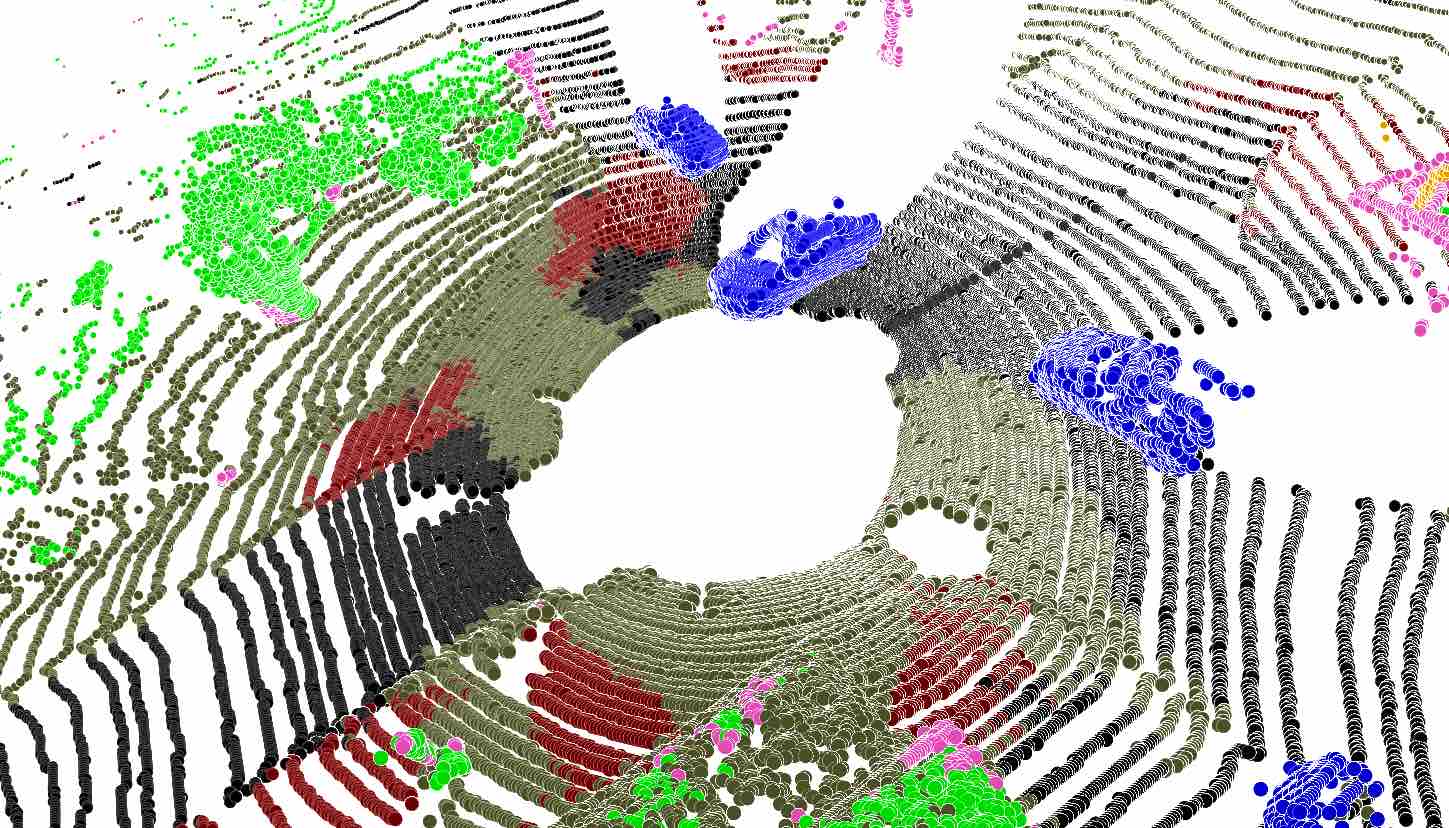}
        \put(-4.5,500){\rotatebox{90}{\color{black}\footnotesize \textbf{source}}}
        \put(-6,440){\rotatebox{90}{\color{black}\footnotesize \textbf{mix3D}}}
        \put(-6,375){\rotatebox{90}{\color{black}\footnotesize \textbf{p.cutmix}}}
        \put(-6,317){\rotatebox{90}{\color{black}\footnotesize \textbf{cosmix}}}
        \put(-6,260){\rotatebox{90}{\color{black}\footnotesize \textbf{ibn}}}
        \put(-6,200){\rotatebox{90}{\color{black}\footnotesize \textbf{robust.}}}
        \put(-5,140){\rotatebox{90}{\color{black}\footnotesize \textbf{sn}}}
        \put(-6,75){\rotatebox{90}{\color{black}\footnotesize \textbf{raycast}}}
        \put(-5,17){\rotatebox{90}{\color{black}\footnotesize \textbf{ours}}}
        \put(-6,-35){\rotatebox{90}{\color{black}\footnotesize \textbf{gt}}}
        \put(185, 542){\color{black}\footnotesize \textbf{SemanticKITTI}}
        \end{overpic} &  
        \begin{overpic}[width=0.21\textwidth]{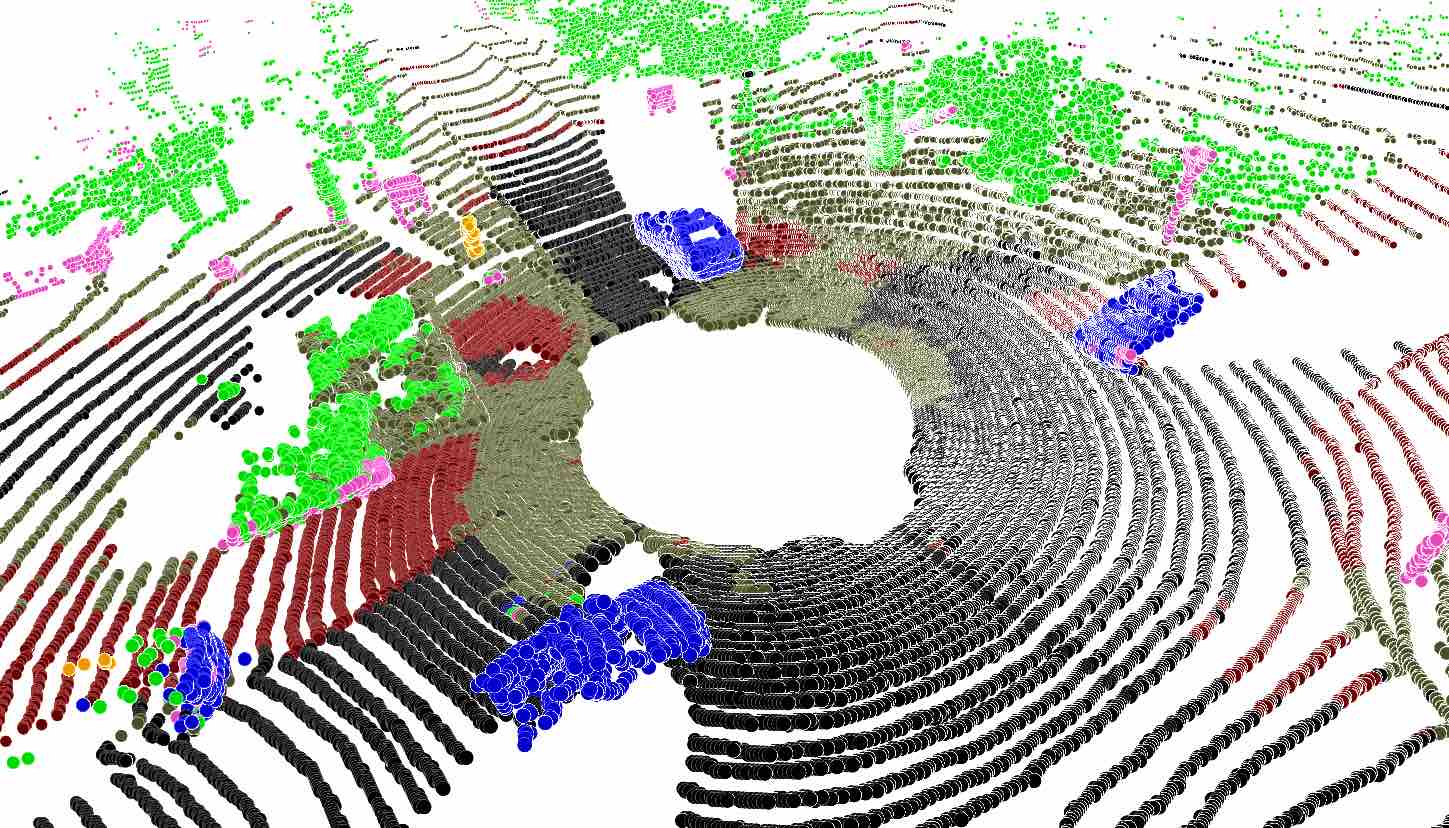}
        \end{overpic} &
        \begin{overpic}[width=0.21\textwidth]{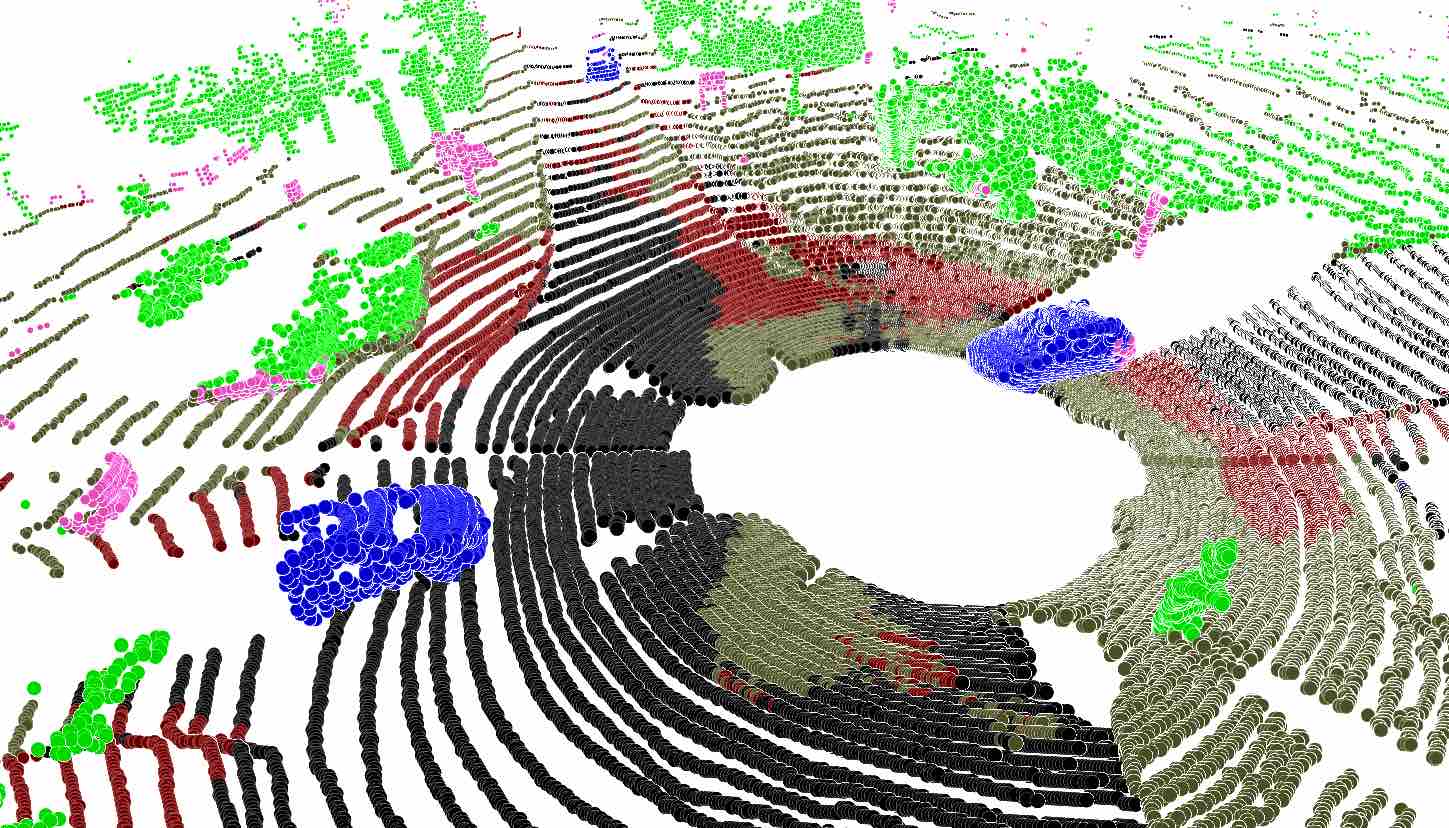}
        \end{overpic}& 
        \begin{overpic}[width=0.21\textwidth]{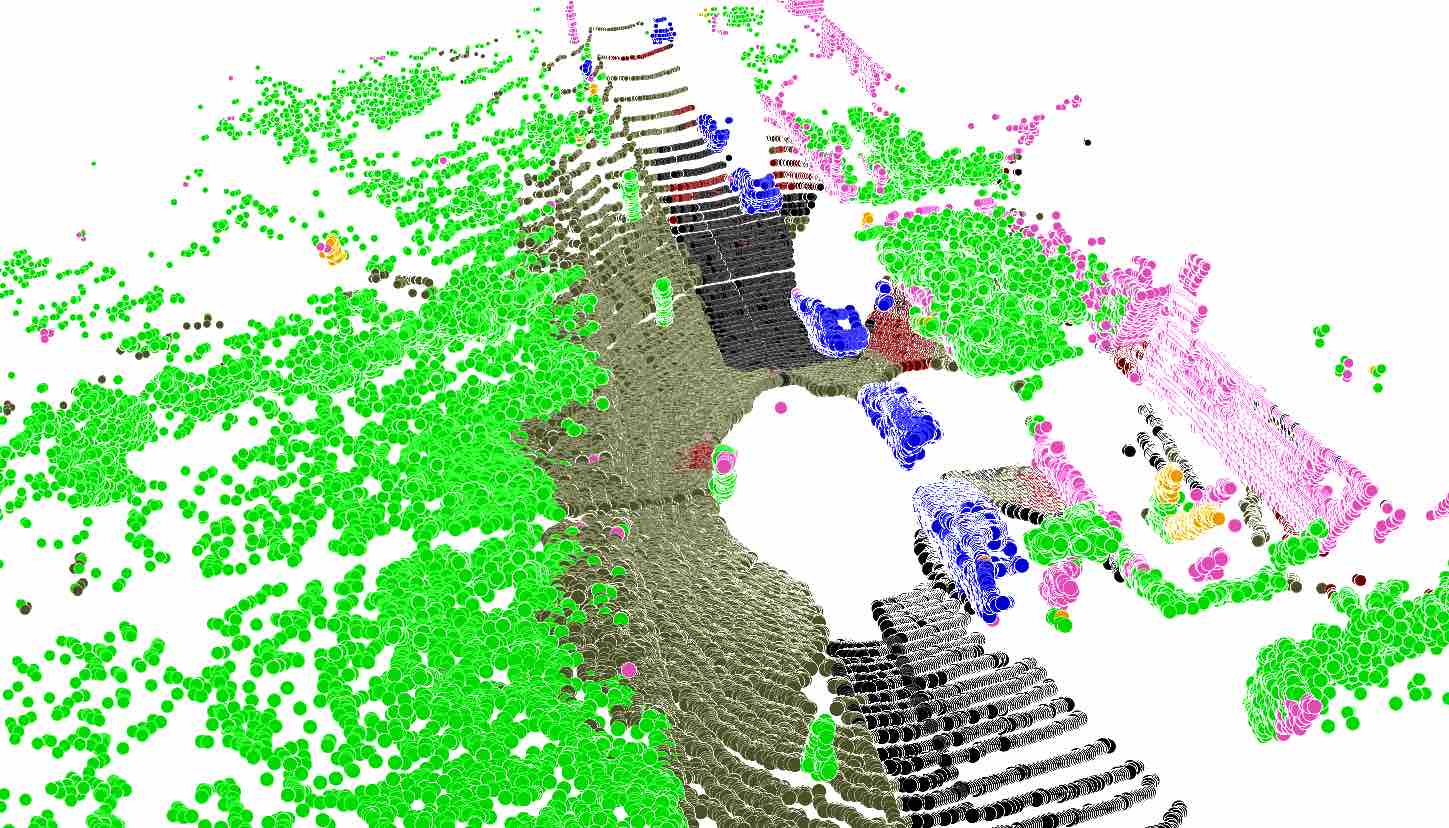}
        \end{overpic}\\
        \begin{overpic}[width=0.21\textwidth]{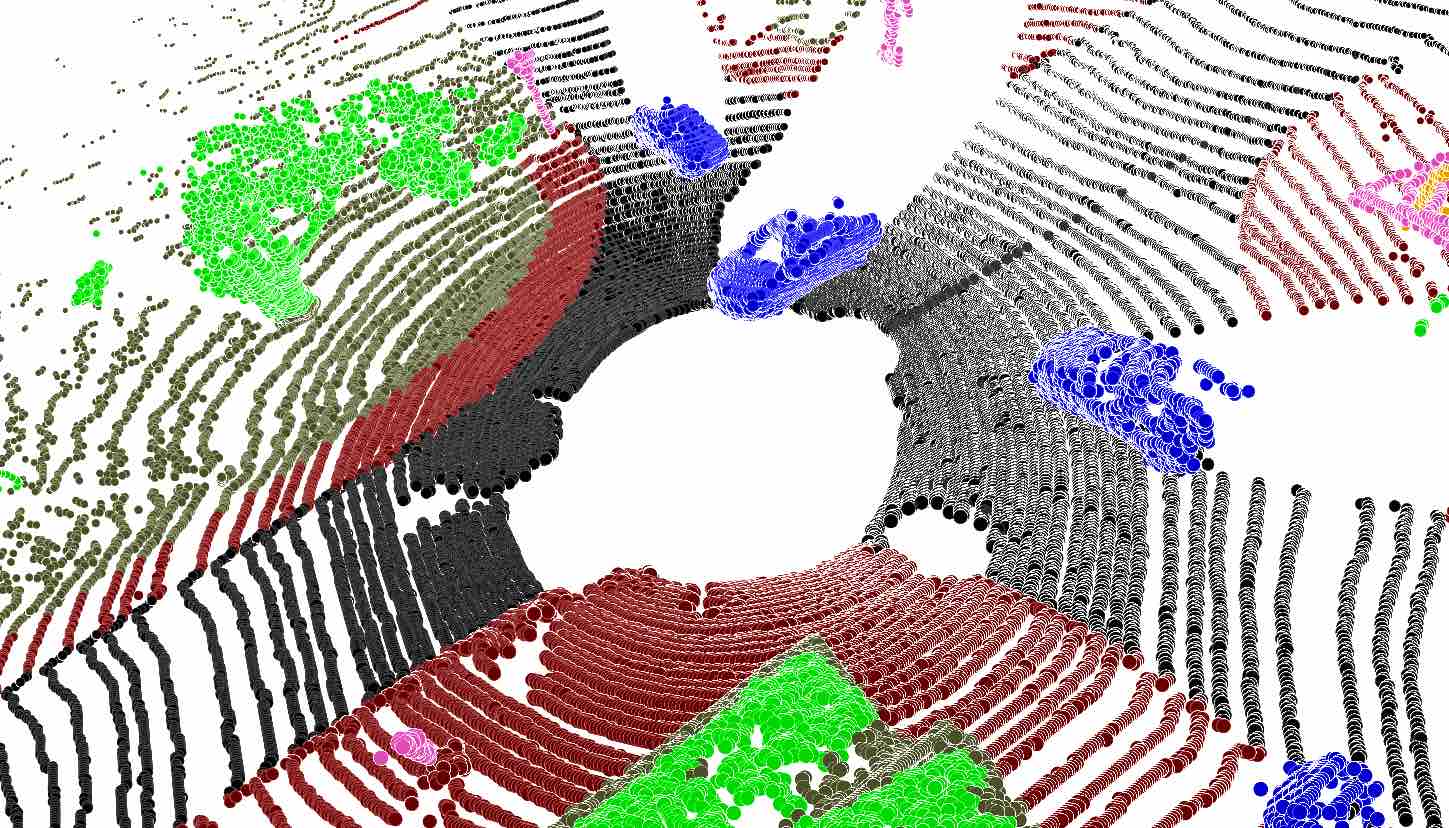}
        \end{overpic} &  
        \begin{overpic}[width=0.21\textwidth]{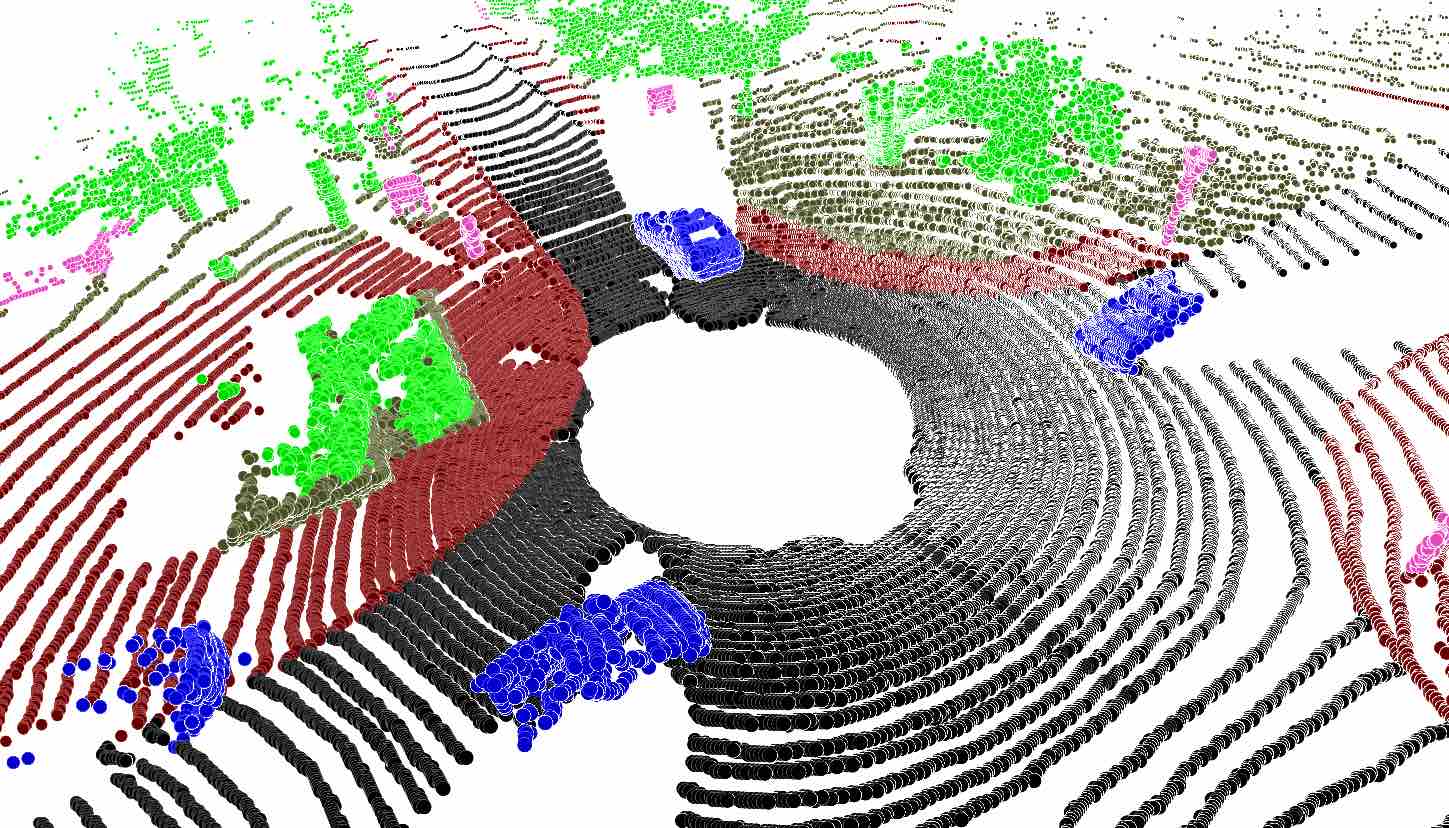}
        \end{overpic} &
        \begin{overpic}[width=0.21\textwidth]{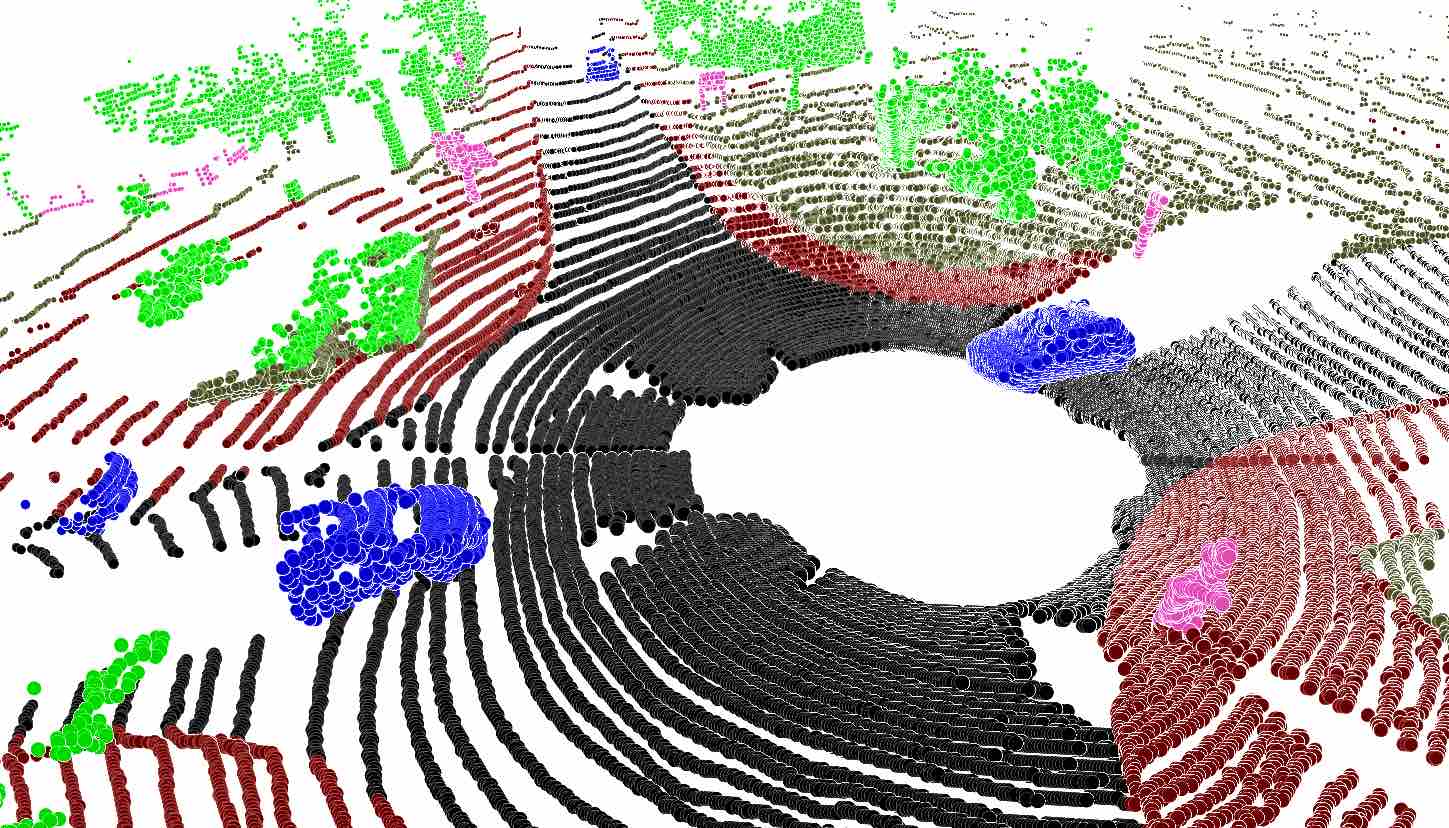}
        \end{overpic}& 
        \begin{overpic}[width=0.21\textwidth]{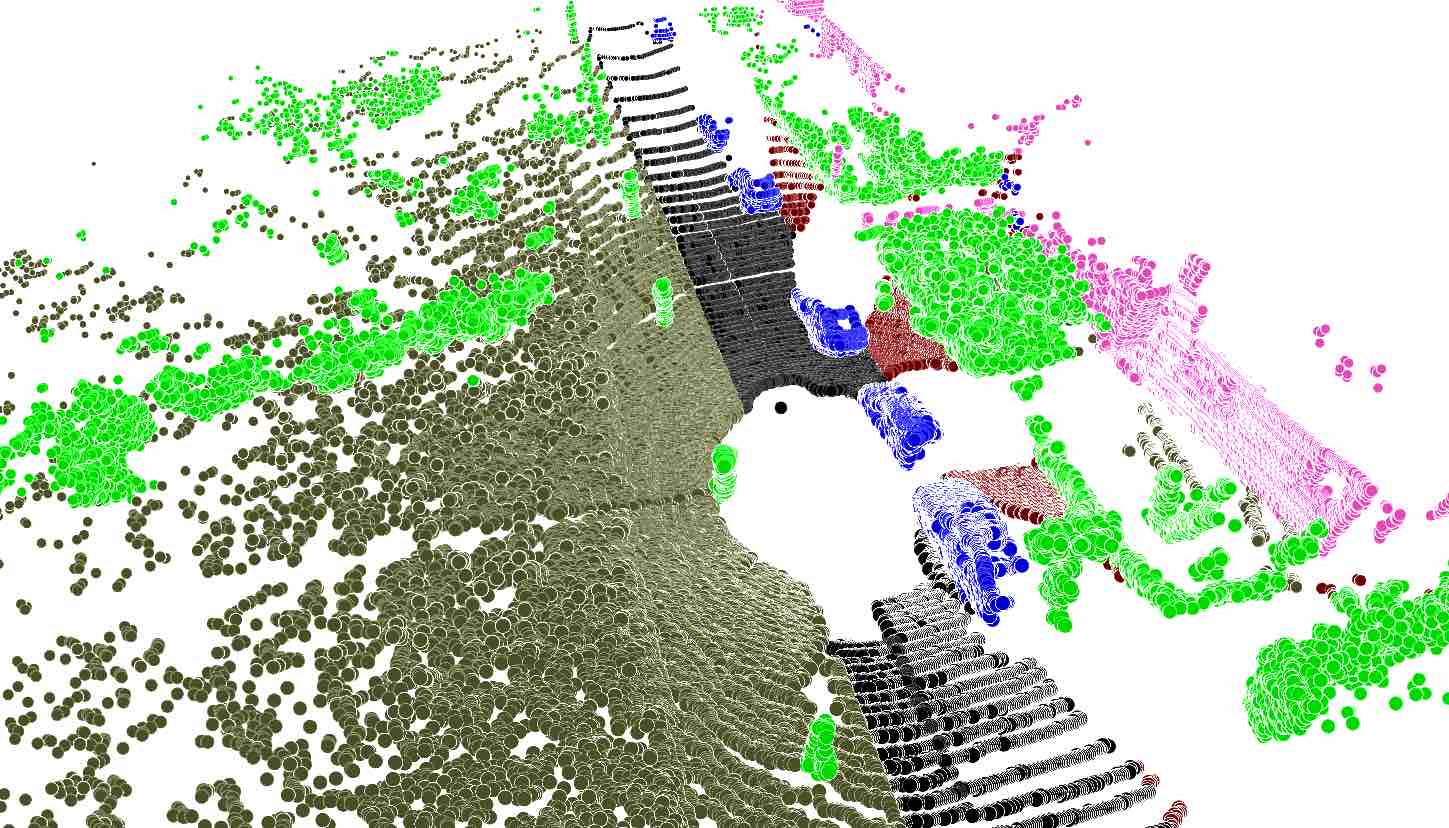}
        \end{overpic}\\
    \end{tabular}
    \vspace{-4mm}
    \caption{\textbf{Qualitative results.} \textit{Top:} nuScenes$\to$SemanticKITTI. \lidog improves over source and baselines, \eg, we observe the improvements in \textit{sidewalk} and \textit{road}.}
    \label{fig:supp_qualitative_nusc}
\end{figure*}

We report additional qualitative results for each baseline and in all the studied generalization directions. In Fig.~\ref{fig:supp_qualitative_synth4dkitti}-\ref{fig:supp_qualitative_multi} we show qualitative results in the Synth$\to$Real setting: Synth4D-kitti$\to$Real (Fig.~\ref{fig:supp_qualitative_synth4dkitti}), Synth4D-nuScenes$\to$Real (Fig.~\ref{fig:supp_qualitative_synth4dnusc}) and Synth4D-kitti$+$Synth4D-nuScenes$\to$Real (Fig.~\ref{fig:supp_qualitative_multi}).  In Fig.~\ref{fig:supp_qualitative_kitti}-\ref{fig:supp_qualitative_nusc} we show qualitative results in the SemanticKITTI$\to$nuScenes and nuScenes$\to$SemanticKITTI directions, respectively. 
Source predictions are often incorrect and spatially inconsistent. Baselines consistently improve over the source model performance. \lidog achieves the overall best performance with improved and more precise predictions. This can be seen in all the reported results, both \textit{synth$\to$real} and \textit{real$\to$real}.

\end{document}